\documentclass[acmtog,nonacm]{acmart}
\usepackage{appendix}

\AtBeginDocument{%
  }

\setcopyright{none}
\acmDOI{XXXXXXX.XXXXXXX}

\copyrightyear{2026}
\acmYear{2026}

\usepackage{amsmath}
\usepackage{comment}
\usepackage{multirow,bigdelim}
\usepackage{lipsum}
\usepackage{array}
\usepackage{wrapfig}

\usepackage{adjustbox}
\usepackage[percent]{overpic}
\usepackage{makecell}
\usepackage[normalem]{ulem}
\usepackage{blindtext}
\usepackage{xcolor}
\usepackage{soul}
\usepackage{cleveref}
\usepackage{amsfonts}

\makeatletter
\newcommand{\settitle}{\@maketitle}
\makeatother

\newcolumntype{C}[1]{>{\centering\let\newline\\\arraybackslash\hspace{0pt}}m{#1}}

\makeatletter
\DeclareRobustCommand\onedot{\futurelet\@let@token\@onedot}
\def\@onedot{\ifx\@let@token.\else.\null\fi\xspace}

\makeatother

\usepackage[bottom]{footmisc}
\usepackage{arydshln}

\makeatletter
\def\blfootnote{\xdef\@thefnmark{}\@footnotetext}
\makeatother

\usepackage{subcaption}
\usepackage{tikz}
\usetikzlibrary{arrows.meta}
\usepackage{multirow}

\makeatletter
\let\@authorsaddresses\@empty
\makeatother

\renewcommand\footnotetextcopyrightpermission[1]{}

\begin{document}

\title{On-the-fly Repulsion in the Contextual Space for Rich Diversity in Diffusion Transformers}

\author{Omer Dahary$^*$}
\email{omer11a@gmail.com}
\orcid{0000-0003-0448-9301}
\affiliation{%
    \institution{Tel Aviv University}
    \country{Israel}
}
\affiliation{%
    \institution{Snap Research}
    \country{Israel}
}

\author{Benaya Koren$^*$}
\email{benayakoren@gmail.com}
\orcid{0009-0007-9643-7256}
\affiliation{%
    \institution{Tel Aviv University}
    \country{Israel}
}

\author{Daniel Garibi}
\email{danielgaribi@mail.tau.ac.il}
\orcid{0009-0003-4261-6153}
\affiliation{%
    \institution{Tel Aviv University}
    \country{Israel}
}
\affiliation{%
    \institution{Snap Research}
    \country{Israel}
}

\author{Daniel Cohen-Or}
\email{cohenor@gmail.com}
\orcid{0000-0001-6777-7445}
\affiliation{%
    \institution{Tel Aviv University}
    \country{Israel}
}
\affiliation{%
    \institution{Snap Research}
    \country{Israel}
}

\begin{teaserfigure}
    \centering
    \includegraphics[width=1\linewidth]{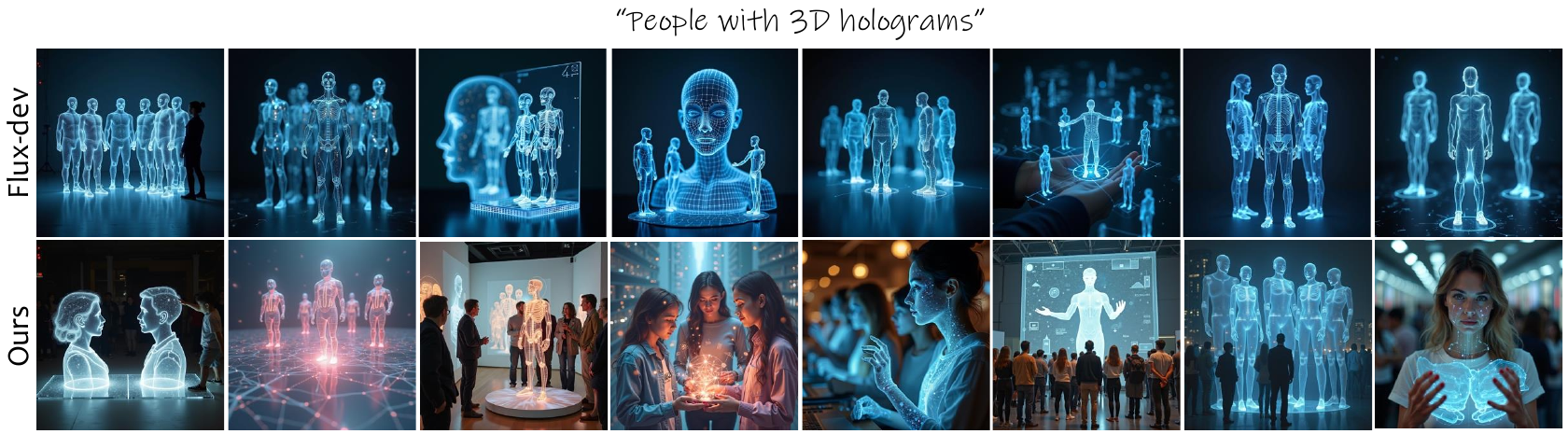}
    \caption{
    \textbf{Example results of our Contextual Space repulsion framework using Flux-dev.} The base model (top) typically converges on a narrow set of visual solutions. By applying semantic intervention within the internal multi-modal attention channels, our approach (bottom) produces a diverse set of images with minimal computational overhead.
    }
    \label{fig:teaser}
\end{teaserfigure}

\begin{abstract}

Modern Text-to-Image (T2I) diffusion models have achieved remarkable semantic alignment, yet they often suffer from a significant lack of variety, converging on a narrow set of visual solutions for any given prompt.
This typicality bias
presents a challenge for creative applications that require a wide range of generative outcomes. We identify a fundamental
trade-off
in current
approaches to diversity: modifying model inputs requires costly optimization to incorporate feedback from the generative path. In contrast, acting on spatially-committed intermediate latents tends to disrupt the forming visual structure, leading to artifacts.
In this work, we propose to apply repulsion in the Contextual Space as a novel framework for achieving rich diversity in Diffusion Transformers. By intervening in the multimodal attention channels, we apply
on-the-fly repulsion during the transformer's forward pass, injecting the intervention between blocks where text conditioning is enriched with emergent image structure. This allows for 
redirecting the guidance trajectory after it is structurally informed but before the composition is fixed.
Our results demonstrate that repulsion in the Contextual Space produces significantly richer diversity without sacrificing visual fidelity or semantic adherence.
Furthermore, our method
is uniquely efficient,
imposing a small
computational overhead
while remaining effective even in modern ``Turbo'' and distilled models where traditional trajectory-based interventions typically fail. 
Project page: \url{https://contextual-repulsion.github.io/}.
\end{abstract}

\maketitle
\makeatletter
\def\thefootnote{*}\footnotetext{Denotes equal contribution.}
\makeatother

\section{Introduction}
\label{sec:intro}
The rapid evolution of Text-to-Image (T2I) generative models has ushered in a new era of high-fidelity visual synthesis, where models now exhibit unprecedented alignment with complex textual prompts~\cite{rombach2022high, podell2023sdxl, esser2024scaling}. However, this progress has come at a significant cost: the reduction of generative diversity. As advanced generative models are increasingly optimized for precision and human preference, they tend to converge on a narrow set of ``typical'' visual solutions, a phenomenon often described as typicality bias~\cite{teotia2025dimcim}. %
Diversity is no longer a secondary metric; it has become a central research problem addressed by a growing body of work~\cite{um2025minority, morshed2025diverseflow, jalali2025sparke}. This is because the utility of generative AI depends on its ability to act as a creative partner that explores the vast manifold of human imagination. It should function as a generative engine rather than merely a sophisticated retrieval mechanism. 

The diversity problem is fundamentally difficult due to the structural tension between quality and variety. High-quality generation currently relies on strong conditioning signals, most notably Classifier-Free Guidance (CFG)~\cite{ho2022classifier}, which effectively sharpens the probability distribution around a single mode by suppressing nearby semantically valid alternatives. Consequently, restoring diversity requires an efficient mechanism to overcome this bias without degrading the structural integrity of the image or losing semantic adherence.

\begin{figure}[t]

    \setlength{\tabcolsep}{0.002\textwidth}
    \scriptsize
    \centering
    \begin{subfigure}{0.33\linewidth}
    \centering
    \includegraphics[width=1\linewidth]{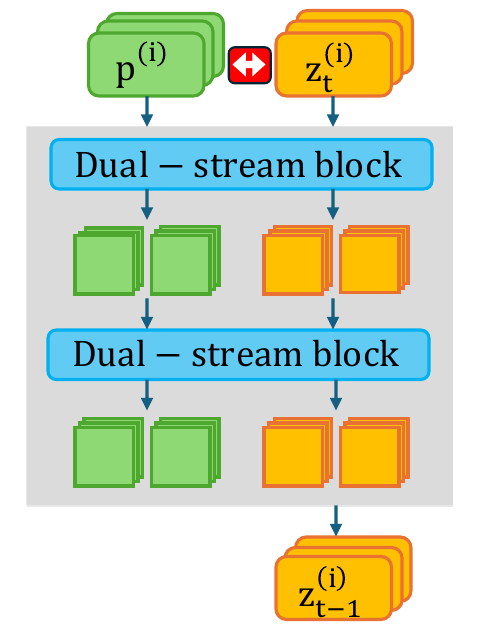}
    \caption{Upstream}
    \label{fig:upstream_scheme}
    \end{subfigure}%
    \begin{subfigure}{0.33\linewidth}
    \centering
    \includegraphics[width=1\linewidth]{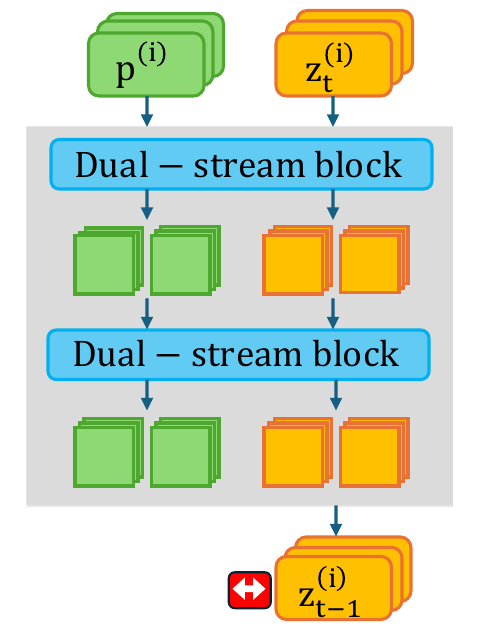}
    \caption{Downstream}
    \label{fig:downstream_scheme}
    \end{subfigure}%
    \begin{subfigure}{0.33\linewidth}
    \centering
    \includegraphics[width=1\linewidth]{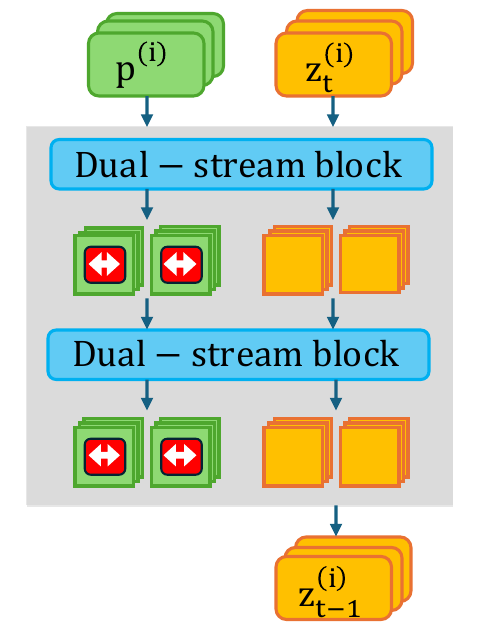}
    \caption{Ours}
    \label{fig:ours_scheme}
    \end{subfigure}
     \caption{
     \textbf{Conceptual comparison of diversity strategies in dual-stream DiT architectures.}
     Here $p^{(i)}$ denotes the prompt embedding for sample $i$, $z_t^{(i)}$ denotes the latent at timestep $t$ for sample $i$, and the \textbf{red double-arrow icon} indicates the point of diversity manipulation.
     \textbf{(a) Upstream}: Interventions on noise or prompt embeddings lack structural feedback from the emerging image. \textbf{(b) Downstream}: Repulsion in image latents acts on a fixed visual mode and can push samples off the data manifold, causing artifacts. \textbf{(c) Ours}: By applying on-the-fly repulsion within the \textbf{Contextual Space} (text-attention channels), we steer the model’s generative intent.
     This allows for a semantically driven intervention synchronized with the emergent visual structure.
    }
    
    \label{fig:schemes}
\end{figure}

Previous attempts to bridge the diversity-alignment gap can be categorized by their point of intervention within the denoising trajectory, as illustrated in Figure~\ref{fig:schemes}.
Upstream methods (Figure~\ref{fig:upstream_scheme}) attempt to solve the problem by altering initial conditions, such as noise seeds or prompt embeddings. However, these approaches are often decoupled from the actual generation process~\cite{sadat2023cads}; to achieve semantic grounding, they must either rely on noisy intermediate estimates~\cite{kim2025diverse} or employ optimization that incurs significant computational overhead~\cite{um2025minority,parmar2025scaling}. Conversely, downstream methods (Figure~\ref{fig:downstream_scheme}) enforce repulsion in the image latent space during denoising~\cite{corso2023particle, jalali2025sparke}.
While these can force variance, they often push samples outside the learned data manifold, resulting in catastrophic drops in visual fidelity and unnatural visual artifacts.

The core difficulty lies in an interventional trade-off: early interventions lack structural feedback, while late interventions face a committed visual mode. This is particularly acute in few-step "Turbo" models, where the generative path is decided almost instantly. Upstream methods require slow optimization to search for diversity-inducing initial conditions, while downstream repulsion arrives too late to steer the composition.

In this work, we present a novel approach that bypasses this trade-off by identifying and leveraging the \textit{Contextual Space} (Figure~\ref{fig:ours_scheme}),
which emerges inside the multimodal attention blocks of Diffusion Transformer (DiT) architectures~\cite{flux2024, esser2024scaling}. Unlike previous U-Net models where text conditioning remains a static external signal, these blocks facilitate a dynamic bidirectional exchange between text and image tokens, continuously updating the text representations in response to the evolving image. This interaction creates an ``enriched'' semantic representation that is both aware of the prompt and synchronized with emergent visual details~\cite{helbling2025conceptattention}.

By leveraging these enriched textual representations, our approach steers the model’s \textit{generative intent} to overcome the CFG mode collapse. By targeting these representations rather than raw pixels, we
preserve samples within the learned data manifold, avoiding the artifacts common in downstream interventions. To achieve this, we apply repulsion to the tokens as they pass between multimodal attention blocks. This intervention is performed on-the-fly during the transformer’s forward pass, at a stage where the emergent representation is already structurally informed but the final composition is not yet fixed. Intervening while the representation is still flexible
allows for steering that
remains
semantically driven yet image-aware. This enables the model to explore diverse paths while maintaining natural, high-quality results.

To demonstrate the efficacy of our approach, we conduct extensive experiments
across multiple DiT-based architectures.
We evaluate our results on the COCO benchmark using metrics for both visual quality and distributional variety. Our results show that repulsion in the Contextual Space consistently produces richer diversity without the mode collapse or semantic misalignment characteristic of prior work. Furthermore, we demonstrate that our method is uniquely efficient, requiring only a small computational overhead and no additional memory, making it compatible with the rapid inference requirements of modern distilled models.

\section{Related Work}
\label{sec:related}
\paragraph{Diffusion transformers.}
While foundational diffusion models predominantly utilized UNet-based architectures~\cite{rombach2022high, podell2023sdxl, ramesh2022hierarchical, saharia2022photorealistic, razzhigaev2023kandinskyimprovedtexttoimagesynthesis}, contemporary state-of-the-art text-to-image systems have largely shifted toward Diffusion Transformers (DiTs) as their backbone~\cite{esser2024scaling, flux2024, kong2025hunyuanvideosystematicframeworklarge, labs2025flux}. A key distinction lies in the conditioning mechanism: whereas UNets typically incorporate text via cross-attention layers, DiTs process text and image tokens concurrently within the transformer. This architecture employs multimodal attention blocks to facilitate bidirectional interaction, ensuring a unified integration of visual and textual information throughout the generation process.
A growing body of research has successfully employed this architecture across diverse downstream tasks~\cite{Avrahami_2025, tan2025ominicontrolminimaluniversalcontrol, garibi2025tokenverseversatilemulticonceptpersonalization, labs2025flux, dalva2024fluxspacedisentangledsemanticediting, kamenetsky2025saedittokenlevelcontrolcontinuous, zarei2025slidereditcontinuousimageediting}

Research addressing the diversity-alignment gap in Text-to-Image (T2I) models generally falls into two categories based on the stage and level of intervention: \textit{upstream} methods, which modify conditions prior to or in the earliest stages of the generative process, and \textit{downstream} methods, which manipulate the image latents throughout the denoising trajectory.

\paragraph{Upstream Interventions}
Upstream methods attempt to induce diversity by optimizing input conditions, namely the initial noise or text conditioning, before a stable image structure emerges. Purely decoupled interventions like CADS~\cite{sadat2023cads} inject prompt-agnostic noise into text embeddings, which often leads to semantic drifting due to a lack of structural feedback. To bridge this, methods like CNO~\cite{kim2025diverse} utilize the very first timestep's $\hat{x}_0$ prediction to force divergence, yet these estimates are frequently structurally unformed at high noise levels, providing an unstable signal for conceptual variety. Similarly, optimization-based regimes such as MinorityPrompt~\cite{um2025minority} and {Scalable Group Inference (SGI)}~\cite{parmar2025scaling} seek diversity-inducing initial conditions through iterative search; however, their heavy computational overhead makes them increasingly impractical for real-time applications or integration with fast-inference distilled models.

\paragraph{Downstream Interventions}
Downstream methods manipulate the latent trajectory throughout the denoising process, either through interacting particle systems or modified guidance schedules. 
The former, pioneered by {Particle Guidance (PG)}~\cite{corso2023particle}, uses kernel-based repulsion in the image latent space to force variance between samples, with subsequent works focusing on improving repulsion loss objectives~\cite{askari2024improving, morshed2025diverseflow, jalali2025sparke} or applying it to image restoration~\cite{cohen2023posterior}.
Despite these refinements, these methods operate on non-semantic representations, repelling low-level pixel-space features rather than semantic content. 
Importantly, semantic concepts in the image latent space are spatially entangled and not aligned across samples, so the same high-level attribute may correspond to different spatial locations and configurations in different generations.
As a result, repulsion in this space often pushes samples outside the learned manifold, leading to unnatural artifacts. In addition, such approaches lack sufficient trajectory depth to remain effective in modern distilled ``Turbo'' models; since the generative path is decided almost instantly, the remaining denoising trajectory is insufficient for late-stage repulsion to steer the model toward diverse modes.

Specifically for distilled models, Diversity Distillation~\cite{gandikota2026distilling} attempts to restore variety by matching the base model's distribution; however, it is limited by the teacher's own lack of diversity and requires a base model that may be inaccessible or costly at inference.
Alternatively, scheduling-based approaches like {Interval Guidance}~\cite{kynkaanniemi2024applying} preserve variety by modulating the CFG scale during denoising. However, because these rescaling schedules are fixed and independent of the model's internal state, they often reduce the prompt's influence before the model has sufficiently established semantic alignment to the prompt.

\medskip
A recurring limitation of these approaches is that their steering signals, whether derived from raw latents or external encoders, lack the semantic coherence necessary for meaningful control during the critical early stages of denoising. This forces an unfavorable trade-off: upstream intervention must incur significant computational overhead to find valid diversity-inducing paths, while downstream interventions occur on a committed visual mode where the composition is already fixed, often producing noise-level variance that pushes samples outside the learned manifold and results in unnatural artifacts. Our work departs from these by identifying a Contextual Space within Diffusion Transformers that is both semantically flexible and structurally informed. This allows us to redirect the guidance trajectory once the bidirectional exchange between text and image tokens has established a stable semantic signal, but before the model has fully converged on a specific generative outcome.%

\section{Method: Repulsion in the Contextual Space}
\label{sec:method}
In this section, we formalize our approach to generative diversity by shifting the intervention focus to the \textit{Contextual Space}. As identified in Section \ref{sec:related}, the core difficulty of existing methods lies in the timing and location of the repulsion: upstream methods act on unformed noise, while downstream methods act on a rigid latent manifold. 
Our central insight is that the Contextual Space, %
inherent to multimodal transformer architectures such as DiTs,
provides an effective environment for diversity interventions because it is structurally informed yet conceptually flexible.

\subsection{Defining the Contextual Space}
The Contextual Space is the high-dimensional manifold formed within the Multimodal Attention (MM-Attention) blocks of a DiT. Unlike the static text embeddings used in U-Net architectures, the DiT processing flow facilitates a bidirectional exchange between text features $f_{T}$ and image features $f_{I}$.

In each transformer block $l$, the resulting tokens undergo a structural transformation:
\begin{equation}
    \hat{f}_{T}^{(l)}, \hat{f}_{I}^{(l)} = \text{MM-Attn}(f_{T}^{(l-1)}, f_{I}^{(l-1)}).
\end{equation}
In this interaction, the text features $f_{T}$ guide the image tokens toward the prompt's semantic requirements. Simultaneously, the image features $f_{I}$ provide feedback regarding the spatial composition and emerging visual details, which the text features absorb to become uniquely tied to the specific image being formed. We therefore identify the resulting enriched text tokens $\hat{f}_{T}^{(l)}$ as the primary %
elements
of the Contextual Space.

A key advantage of this space is its inherent token ordering. Unlike the image latent space, where specific semantic content can shift spatially across different samples, the Contextual Space maintains a fixed semantic alignment across the sequence index. This
facilitates a consistent representation where each token index generally represents the same conceptual component across the entire batch, largely independent of its realized placement in the emergent image structure.

\subsection{The Mechanism of Contextual Repulsion}
We illustrate the positioning of our intervention in Figure~\ref{fig:ours_scheme}. Our key insight is that applying repulsion within the Contextual Space allows for the manipulation of \textit{generative intent}. By enforcing distance between batch samples in this space, we steer the model's high-level planning before it commits to a specific visual mode. To achieve this, we adopt the particle guidance framework~\cite{corso2023particle}, which treats a batch of $B$ samples as interacting particles. However, unlike prior work that applies guidance to the image latents $z_t$ (Figure~\ref{fig:downstream_scheme}), we apply the repulsive forces directly to the Contextual Space tokens $\hat{f}_T$ (Figure~\ref{fig:ours_scheme}).

Since the conditioning for each sample is initialized from the same unmodified prompt encoding at every timestep, the intervention mitigates the risk of permanent semantic drift. This common starting point
promotes a state where
contextual features remain closely aligned to the original prompt and directly comparable across the batch throughout the trajectory, allowing the repulsion to act as a force that differentiates how the same prompt is visually realized.

A critical advantage of our approach is that these forces are computed on-the-fly. Because we intervene directly on the internal activations, the method does not require backpropagating through the model layers, making it significantly more computationally efficient than optimization-based methods. Within each transformer block, we apply $M$ inner-block iterations to iteratively refine the token positions. Following the gradient-based guidance formulation~\cite{corso2023particle}, the updated state of the contextual tokens for a sample $i \in \{1, \dots, B\}$ after each iteration is given by:
\begin{equation}
    \hat{f}_{T, i}^{(l) \prime} = \hat{f}_{T, i}^{(l)} + \frac{\eta}{M} \nabla_{\hat{f}_{T, i}^{(l)}} \mathcal{L}_{div}(\{\hat{f}_{T, j}^{(l)}\}_{j=1}^B),
\end{equation}
where $\eta$ is the overall repulsion scale and $\mathcal{L}_{div}$ is a diversity loss defined over the batch of $B$ samples. To maintain diversity throughout the trajectory, we apply this repulsion across all transformer MM-blocks. 
However, since the initial stages of the denoising trajectory are the most crucial for the eventual semantic meaning and global composition~\cite{dahary2024yourself,dahary2025decisive,patashnik2023localizing, balaji2023ediffitexttoimagediffusionmodels, cao2025temporalscoreanalysisunderstanding, huberman2025image, yehezkel2025navigating}, and are also where strong guidance signals such as CFG most strongly bias the generative path, we restrict the intervention to a chosen interval of the first few timesteps.

\subsection{Diversity Objective}

The Contextual Space encodes global semantic intent shared across the batch, making diversity objectives based on batch-level similarity more appropriate than token-wise or local measures.
While our framework is flexible and can adopt various diversity losses defined in prior work~\cite{morshed2025diverseflow, jalali2025sparke}, we specifically utilize the Vendi Score~\cite{friedman2022vendi,askari2024improving} as our primary objective. The Vendi Score provides a principled way to measure the effective number of distinct samples in a batch by considering the eigenvalues of a similarity matrix. Formally, it is defined as the exponent of the von Neumann entropy of that matrix.

For simplicity, we represent each sample $i$ at block $l$ as a single vector $\mathbf{c}_i^{(l)} \in \mathbb{R}^{ND}$ by flattening the sequence of $N$ contextual tokens, each of dimension $D$.
For a batch of size $B$ represented by these flattened contextual vectors $\{\mathbf{c}_i^{(l)}\}_{i=1}^B$, we first define a kernel matrix $\mathbf{K} \in \mathbb{R}^{B \times B}$, where each entry $K_{ij}$ represents the similarity between samples $i$ and $j$. In our work, we use the cosine similarity as our kernel:
\begin{equation}
    K_{ij} = \frac{\langle \mathbf{c}_i^{(l)}, \mathbf{c}_j^{(l)} \rangle}{\|\mathbf{c}_i^{(l)}\| \|\mathbf{c}_j^{(l)}\|}
\end{equation}
To maximize diversity, we compute the eigenvalues $\{\lambda_k\}$ of the normalized kernel $\tilde{\mathbf{K}} = \frac{1}{B}\mathbf{K}$ and define our loss $\mathcal{L}_{div}$ as the negative von Neumann entropy:
\begin{equation}
    \mathcal{L}_{div} = -\sum_{k=1}^B \lambda_k \log \lambda_k
\end{equation}
This objective effectively pushes the tokens in the Contextual Space to span a higher-dimensional manifold, preventing the semantic collapse typically induced by CFG.

\section{The Contextual Space}
\label{sec:context_space_analysis}

In this section, we empirically examine the properties of the Contextual Space by analyzing how internal representations behave under controlled interpolation and extrapolation.
We focus on how semantic structure is preserved or degraded when steering representations in two internal spaces of the DiT: the VAE latent space and the contextual (enriched text) token space.
The goal is to characterize how each of these spaces reflects semantic variation when multiple samples are generated from the same prompt, and to assess their suitability for diversity control without introducing visual artifacts.

To examine this, we conduct an interpolation and extrapolation experiment across these two internal representation spaces.
We consider two prompts, ``a person with their pet'' and ``a mythical creature''. For each prompt, we generate two samples using different initial noise seeds, which we designate as a \textit{source image} and a \textit{target image}.
Maintaining the initial noise of the source image, we intervene during the denoising process by replacing its internal representation with a linear combination of the source and target representations%
\begin{equation}
    \mathbf{h}_{interp} = \mathbf{h}_{source} + \alpha (\mathbf{h}_{target} - \mathbf{h}_{source}),
\end{equation}
where $\mathbf{h}$ represents the representation in a given space, and $\alpha$ is the steering coefficient.
We compare this behavior across two distinct spaces: the VAE Latent Space ($z_t$) and our proposed Contextual Space (enriched text tokens $\hat{f}_T$).

\begin{figure}[t]
    \centering 
    \small

    \setlength{\tabcolsep}{0pt}
    \newcommand{\vertlabel}[1]{\rotatebox{90}{\textsf{#1}}}
    \newcommand{\imagewidth}{0.155\linewidth}

    \newcommand{\smallspace}{\hspace{4pt}}
    \newcommand{\largespace}{}

    \begin{tabular}{c@{\smallspace}c@{\largespace}c@{\smallspace}c@{\smallspace}c@{\largespace}c@{\largespace}}

    \multicolumn{6}{c}{``A mythical creature''} \\
    
    Target & \multicolumn{2}{c}{Interpolation 
    } & Source & \multicolumn{2}{c}{Extrapolation 
    } \\
    \includegraphics[width=\imagewidth]{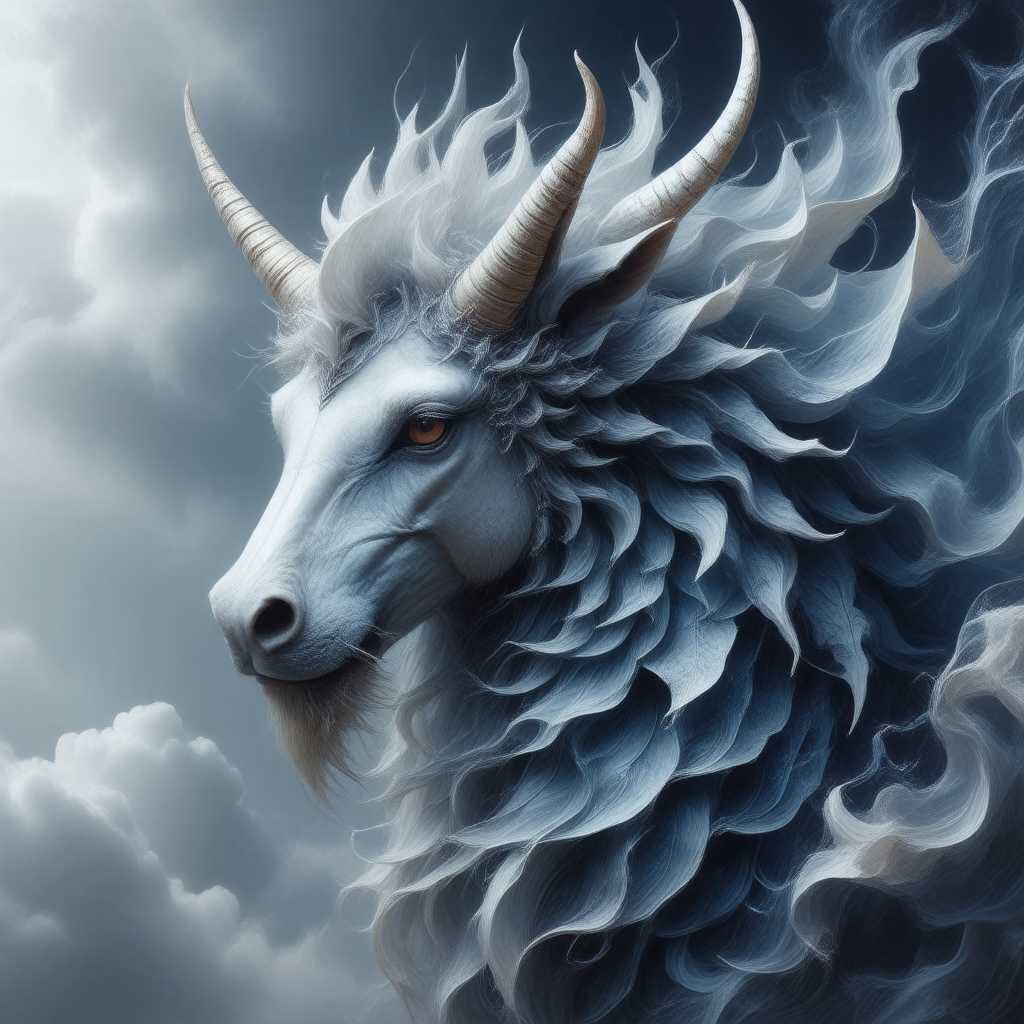} &
    \includegraphics[width=\imagewidth]{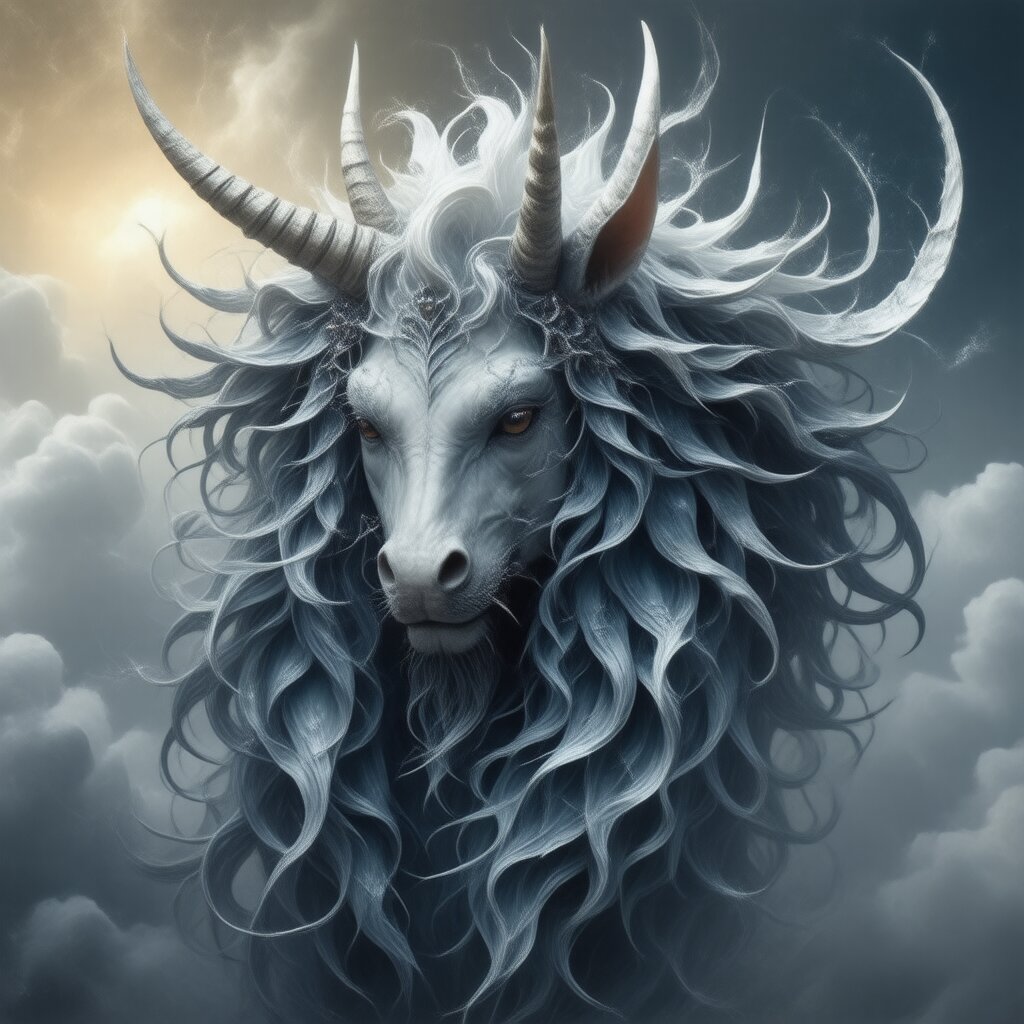} &
    \includegraphics[width=\imagewidth]{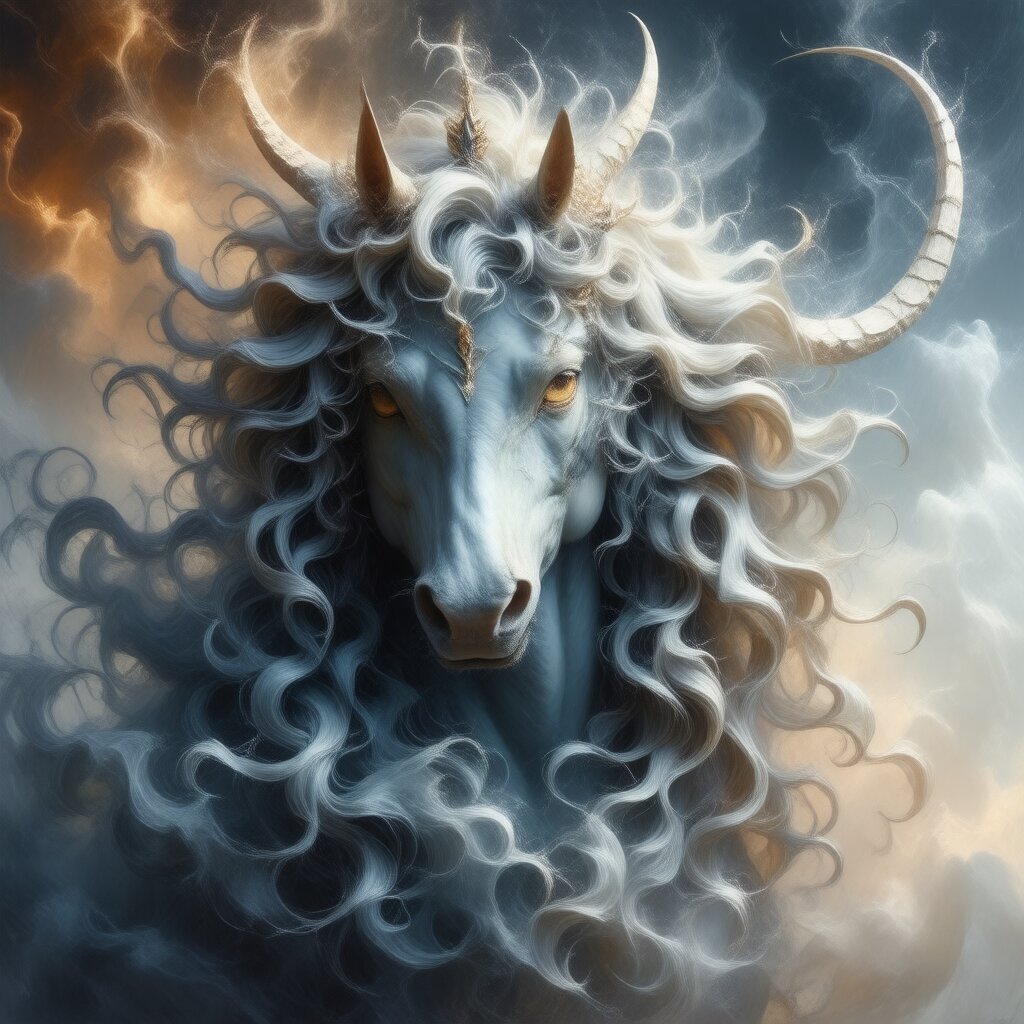} &
    \includegraphics[width=\imagewidth]{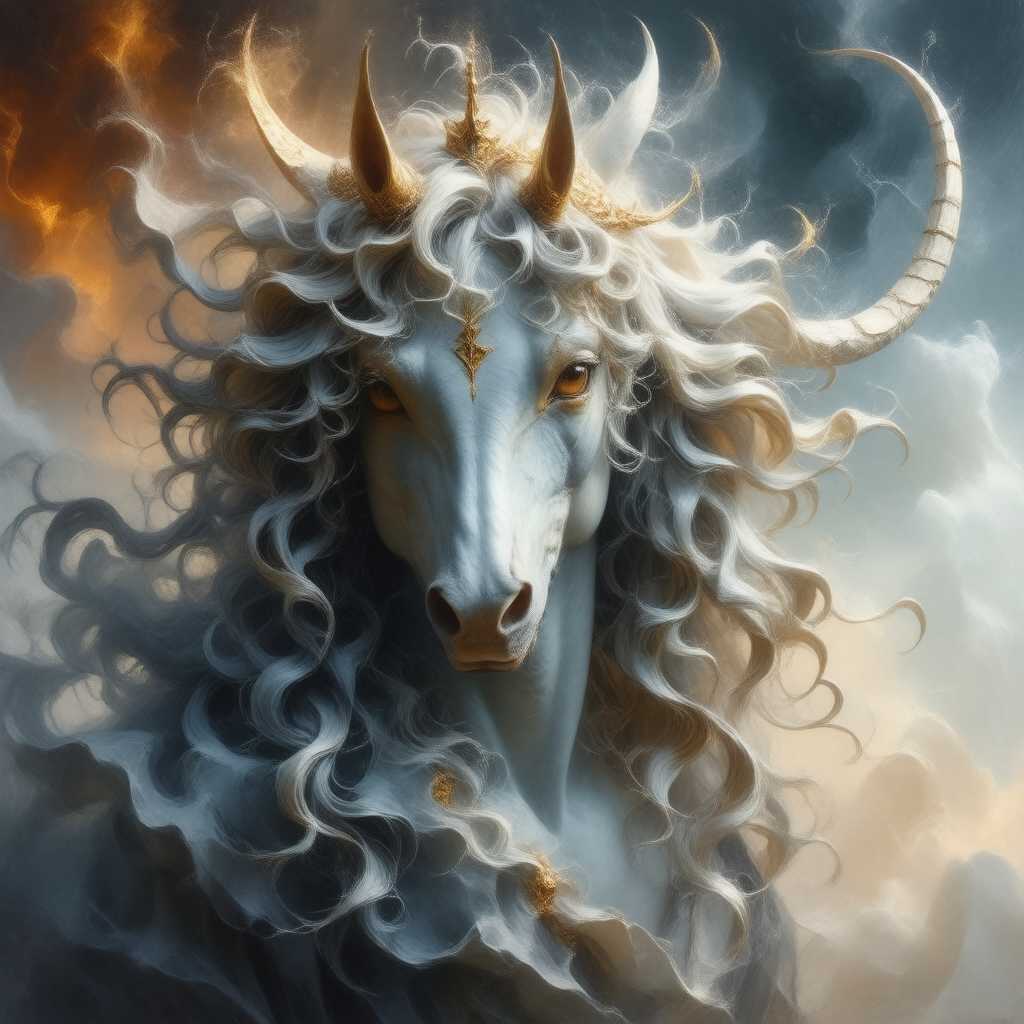} &
    \includegraphics[width=\imagewidth]{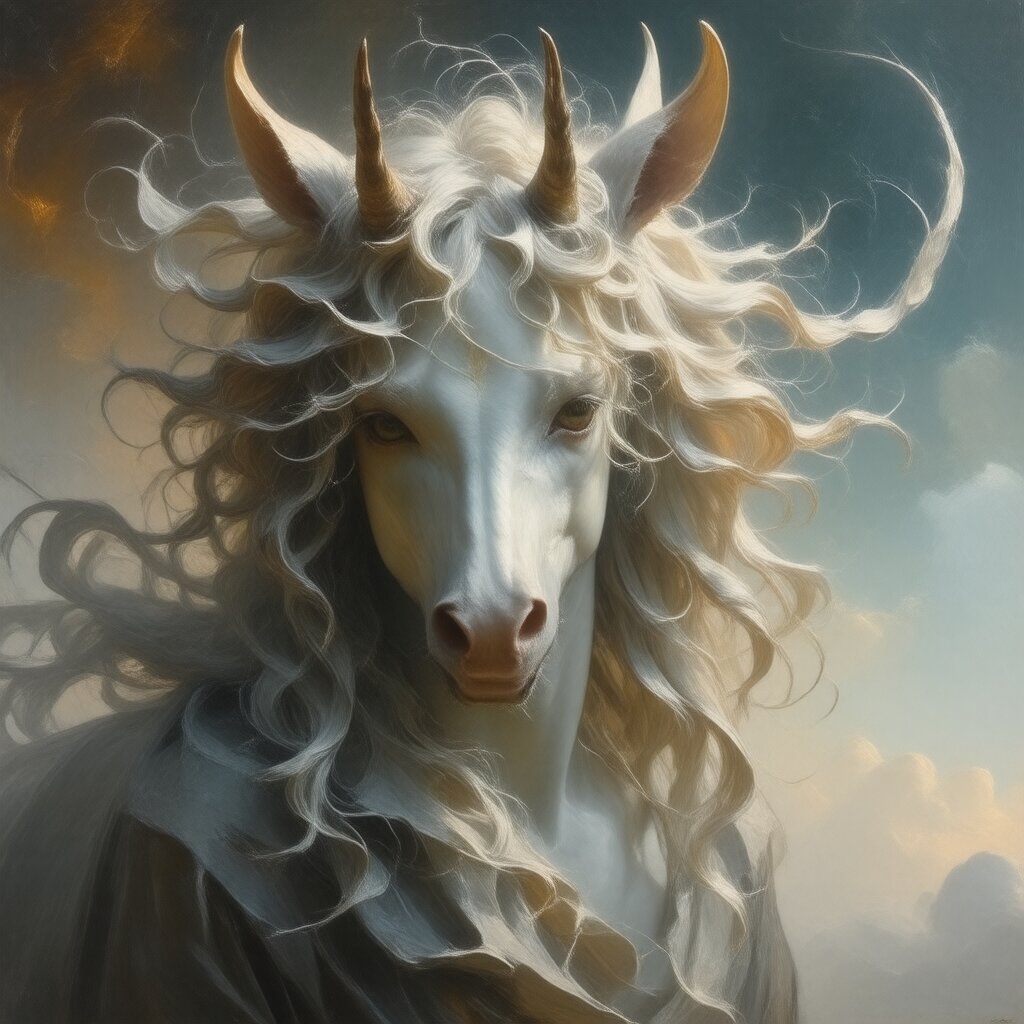} &
    \includegraphics[width=\imagewidth]{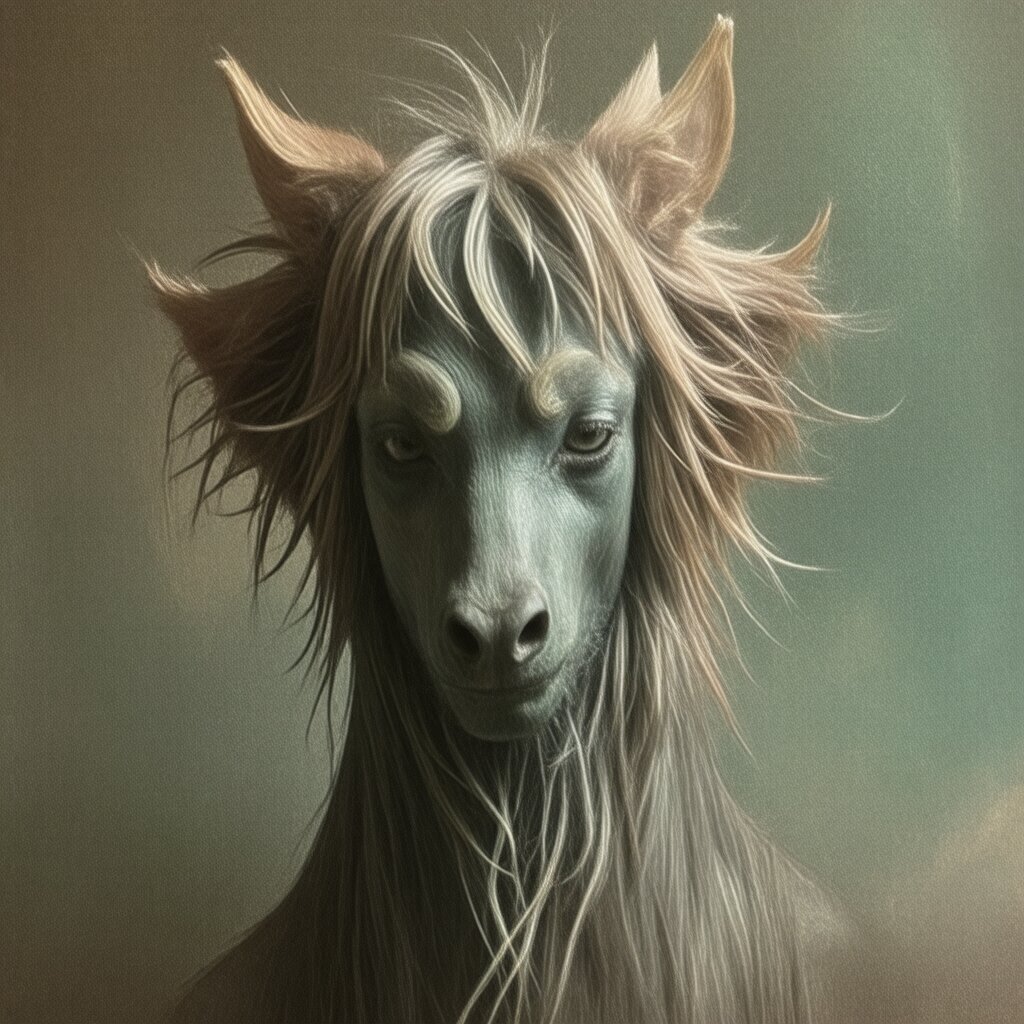} \\
    \multicolumn{6}{l}{\textbf{Contextual Space}} \\ \noalign{\smallskip}

    \includegraphics[width=\imagewidth]{images/analysis/interpolations/mythical/target.jpg} &
    \includegraphics[width=\imagewidth]{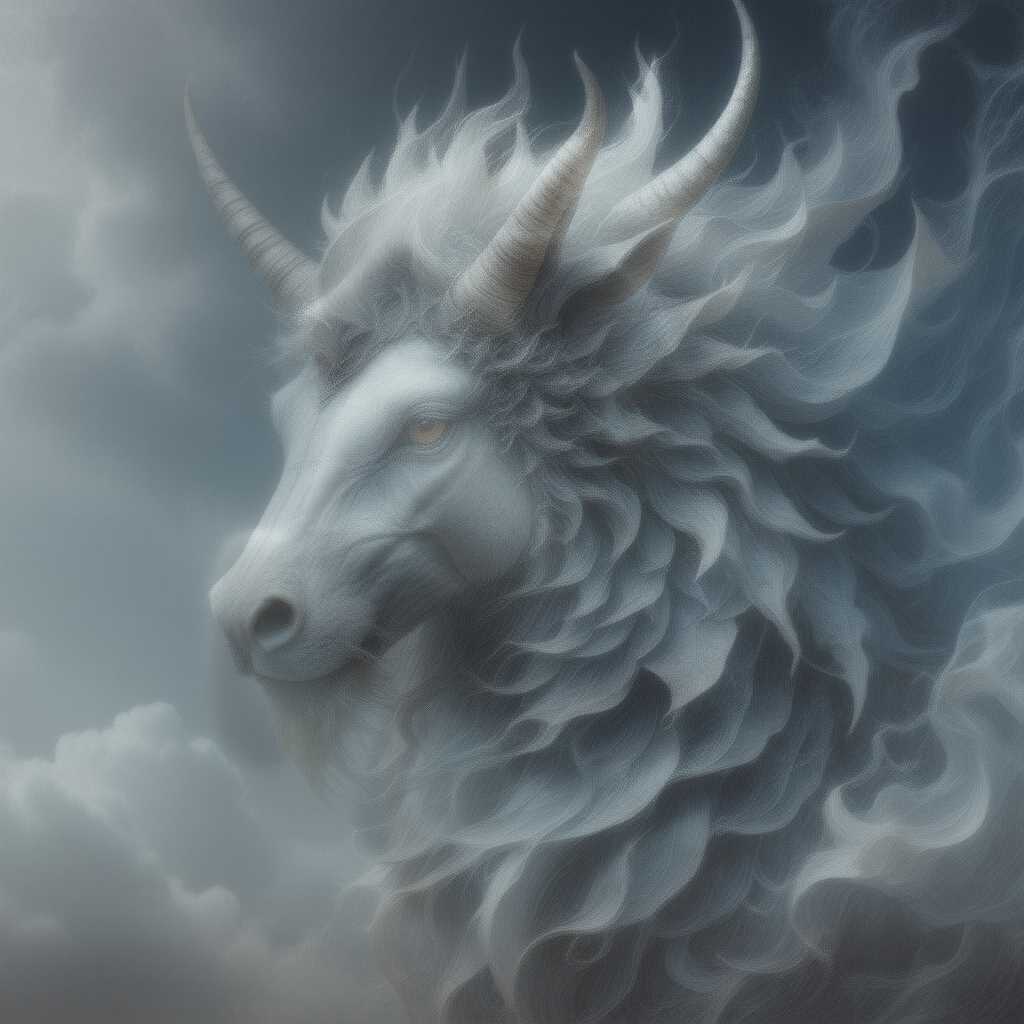} &
    \includegraphics[width=\imagewidth]{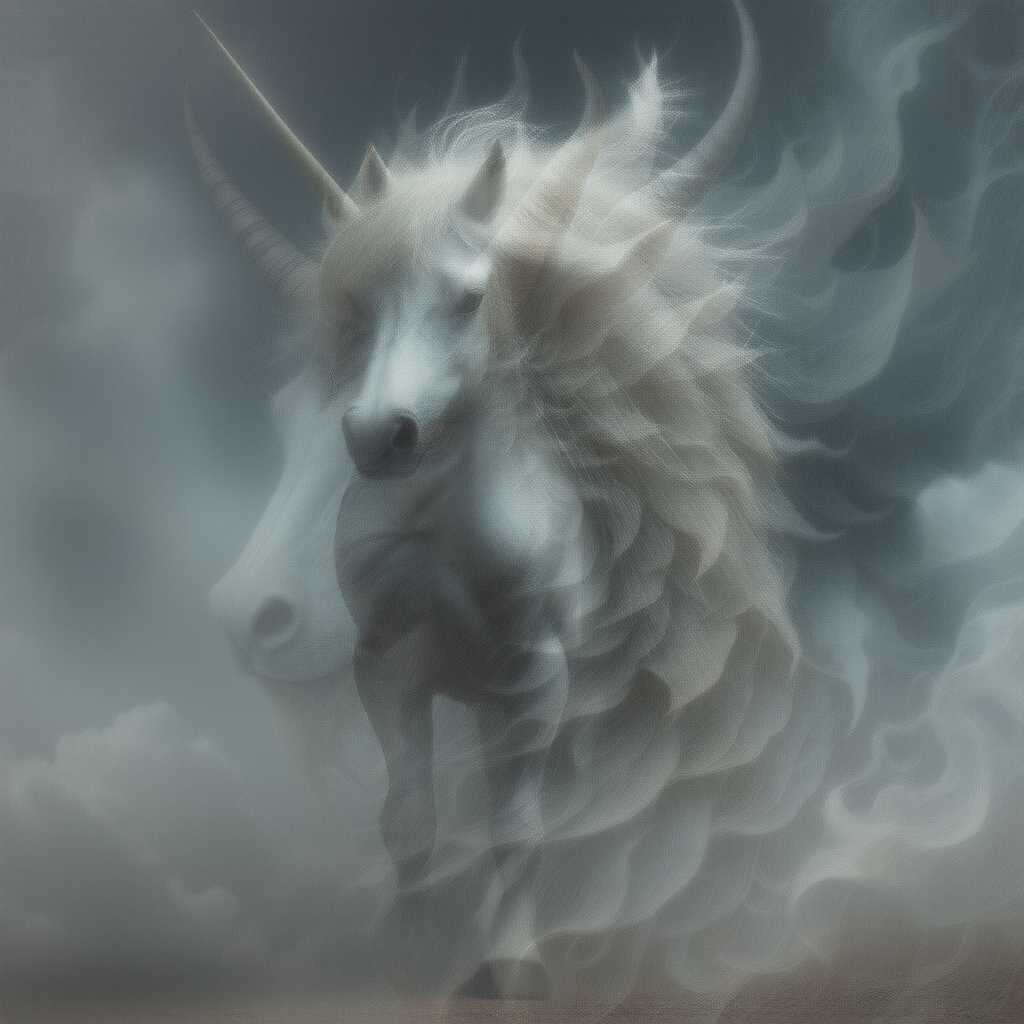} &
    \includegraphics[width=\imagewidth]{images/analysis/interpolations/mythical/source.jpg} &
    \includegraphics[width=\imagewidth]{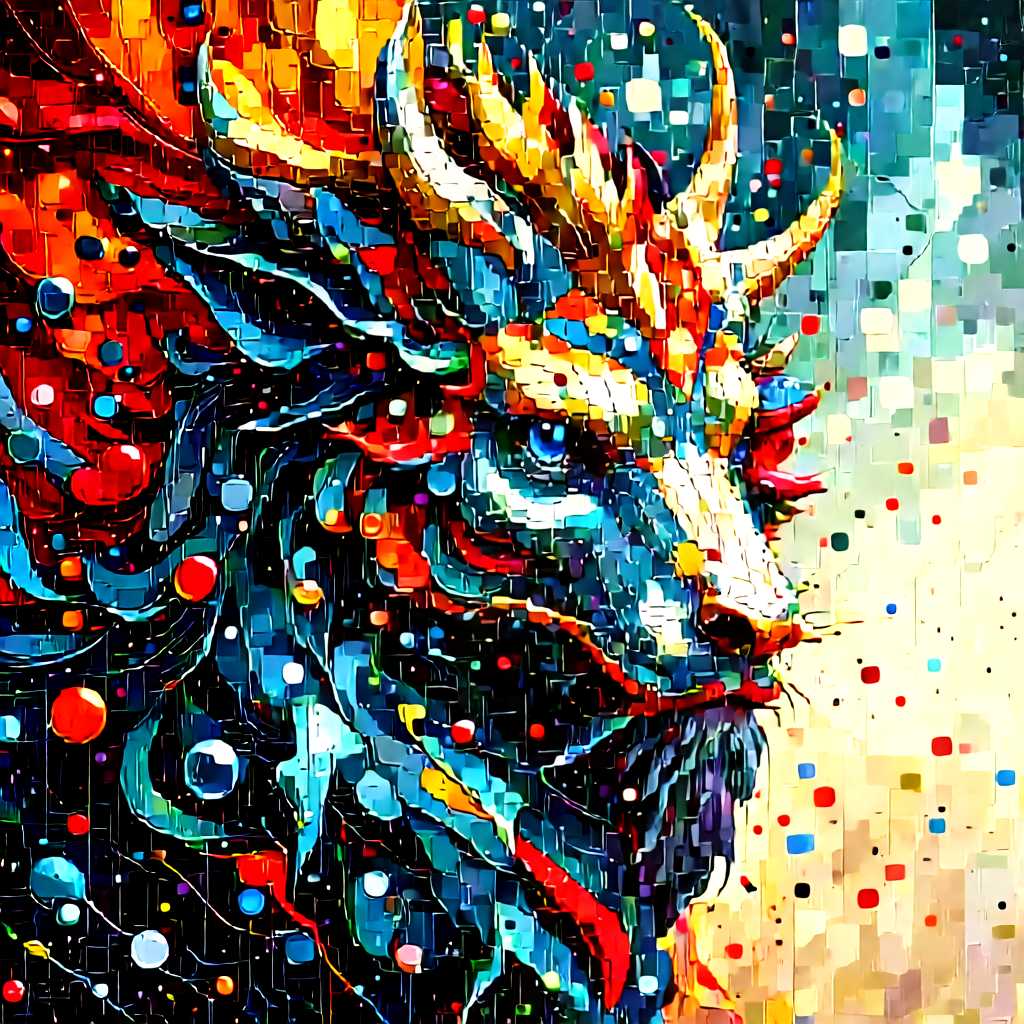} &
    \includegraphics[width=\imagewidth]{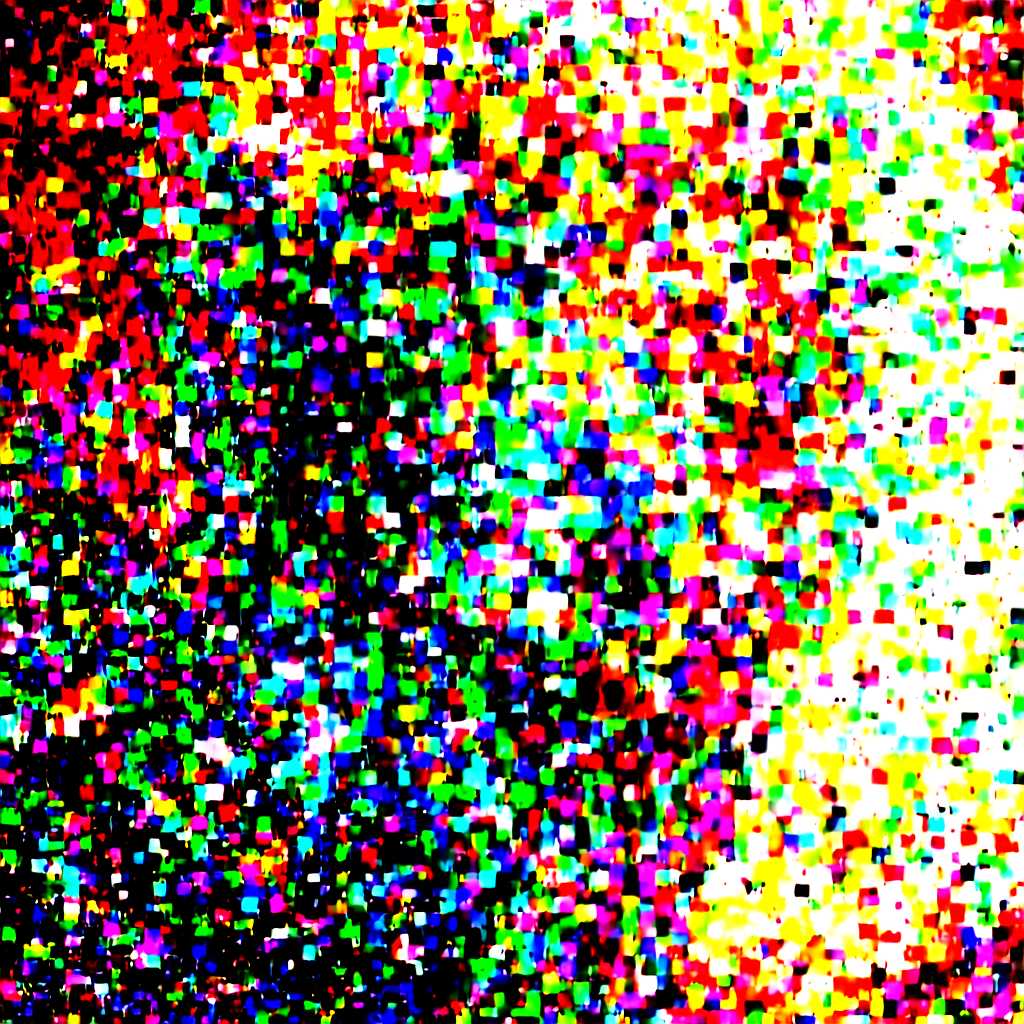} \\
    \multicolumn{6}{l}{\textbf{Latent Space}} \\
    \noalign{\smallskip}

    \multicolumn{6}{c}{``A person with their pet''} \\
    Target & \multicolumn{2}{c}{Interpolation 
    } & Source & \multicolumn{2}{c}{Extrapolation 
    } \\
    \includegraphics[width=\imagewidth]{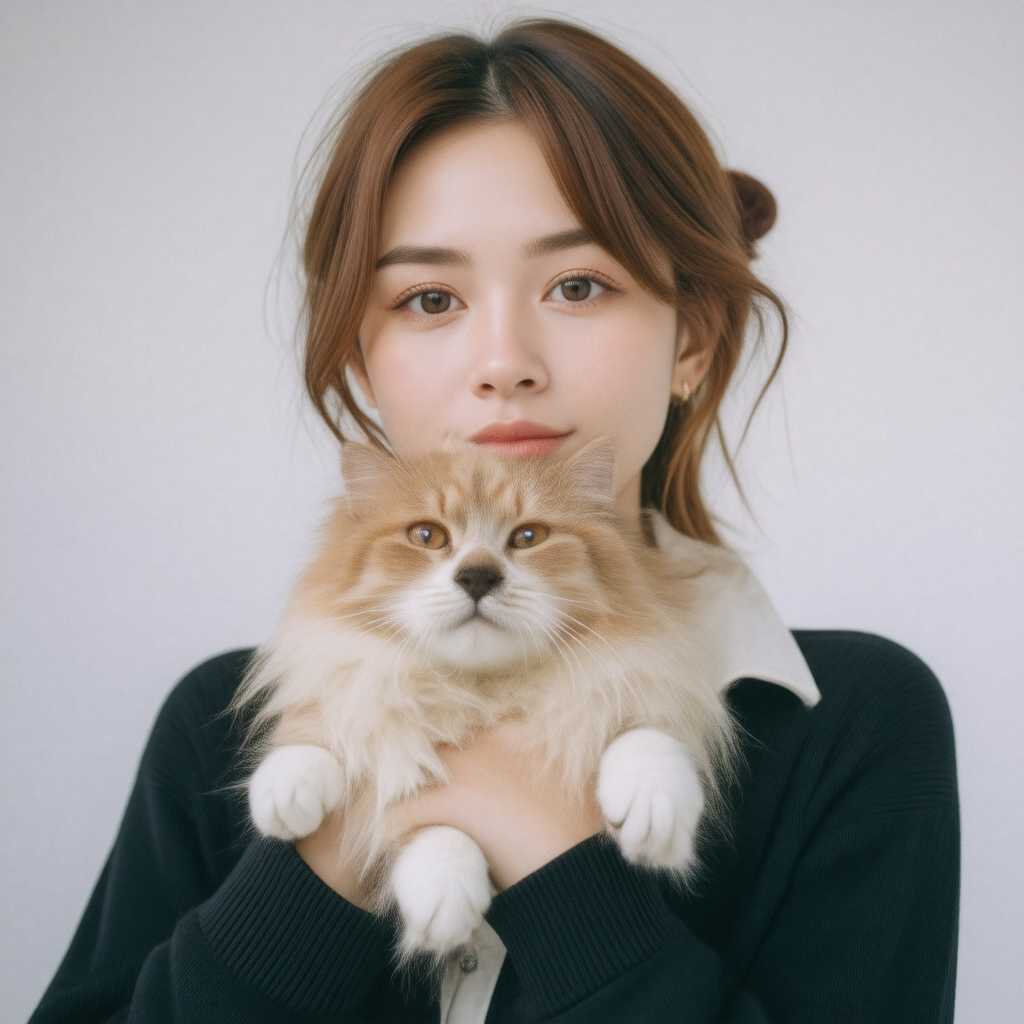} &
    \includegraphics[width=\imagewidth]{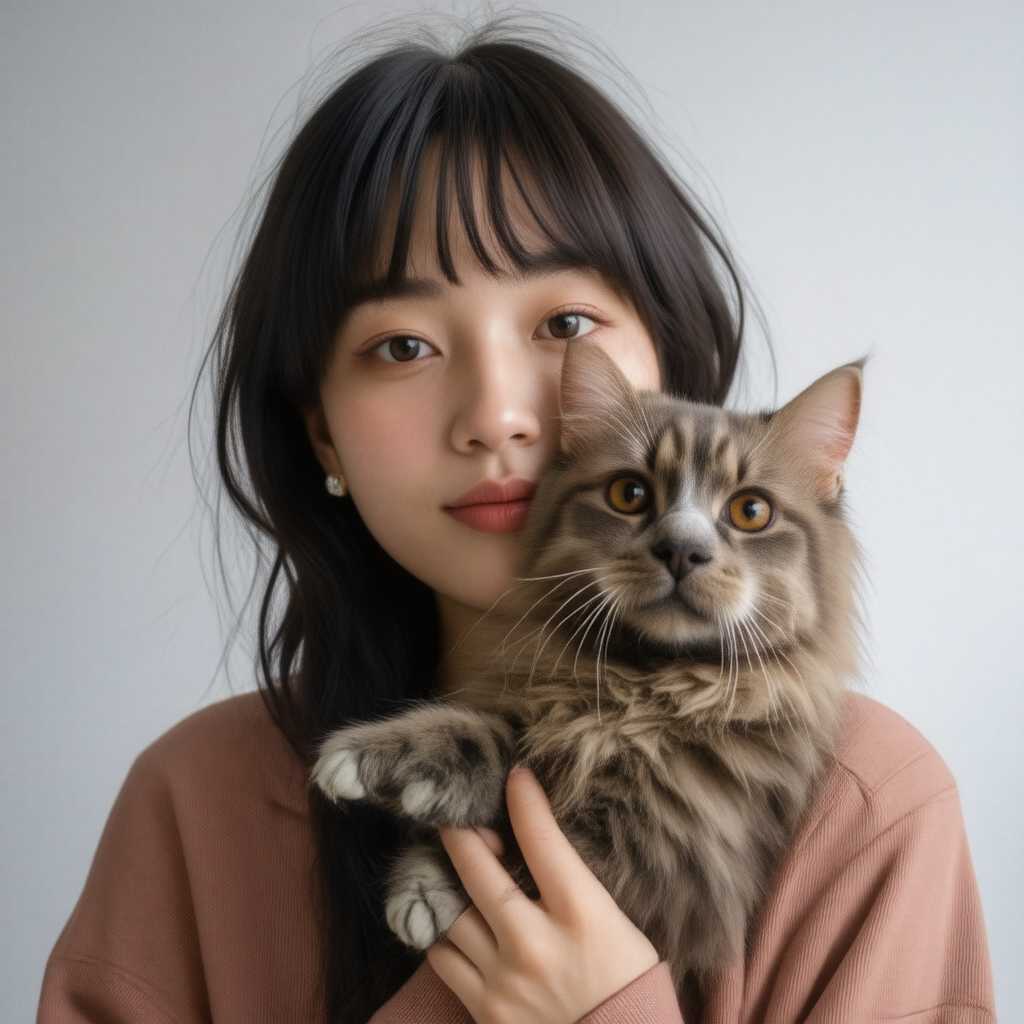} &
    \includegraphics[width=\imagewidth]{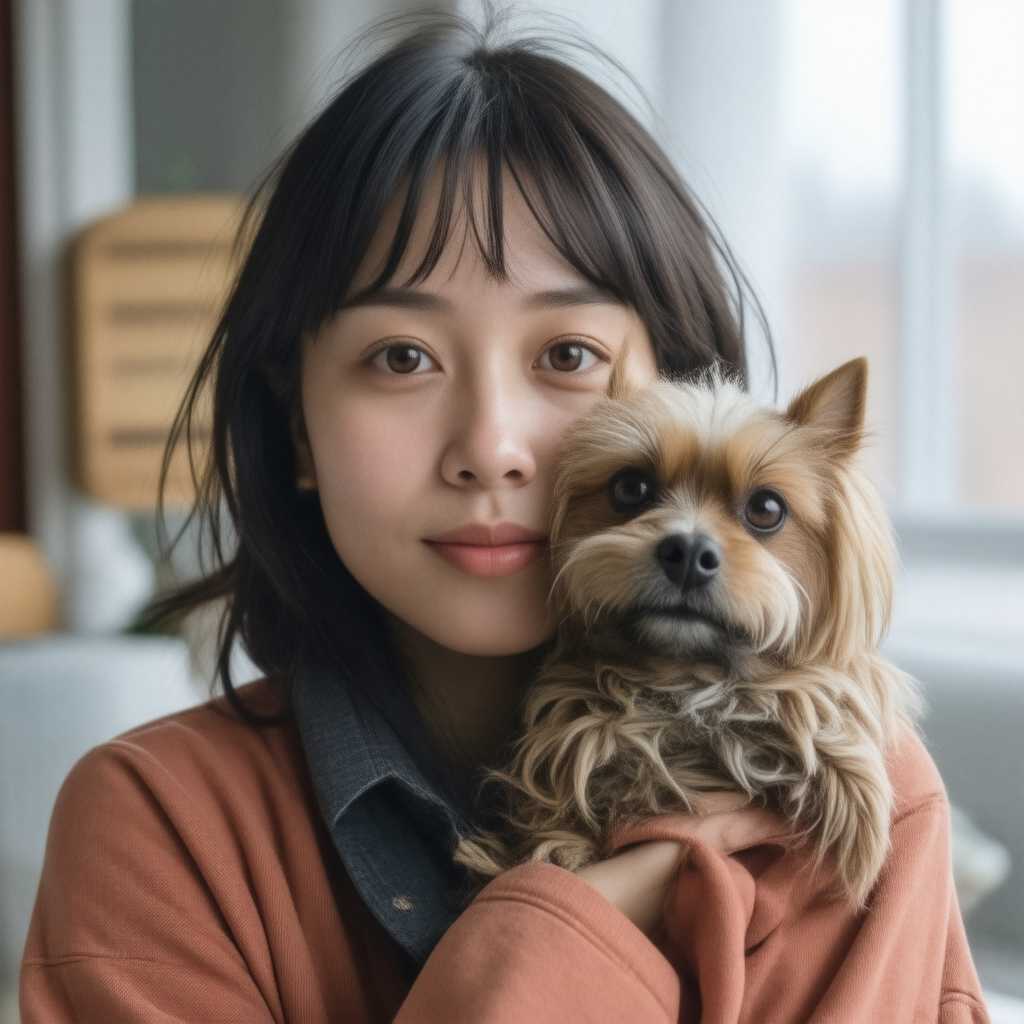} &
    \includegraphics[width=\imagewidth]{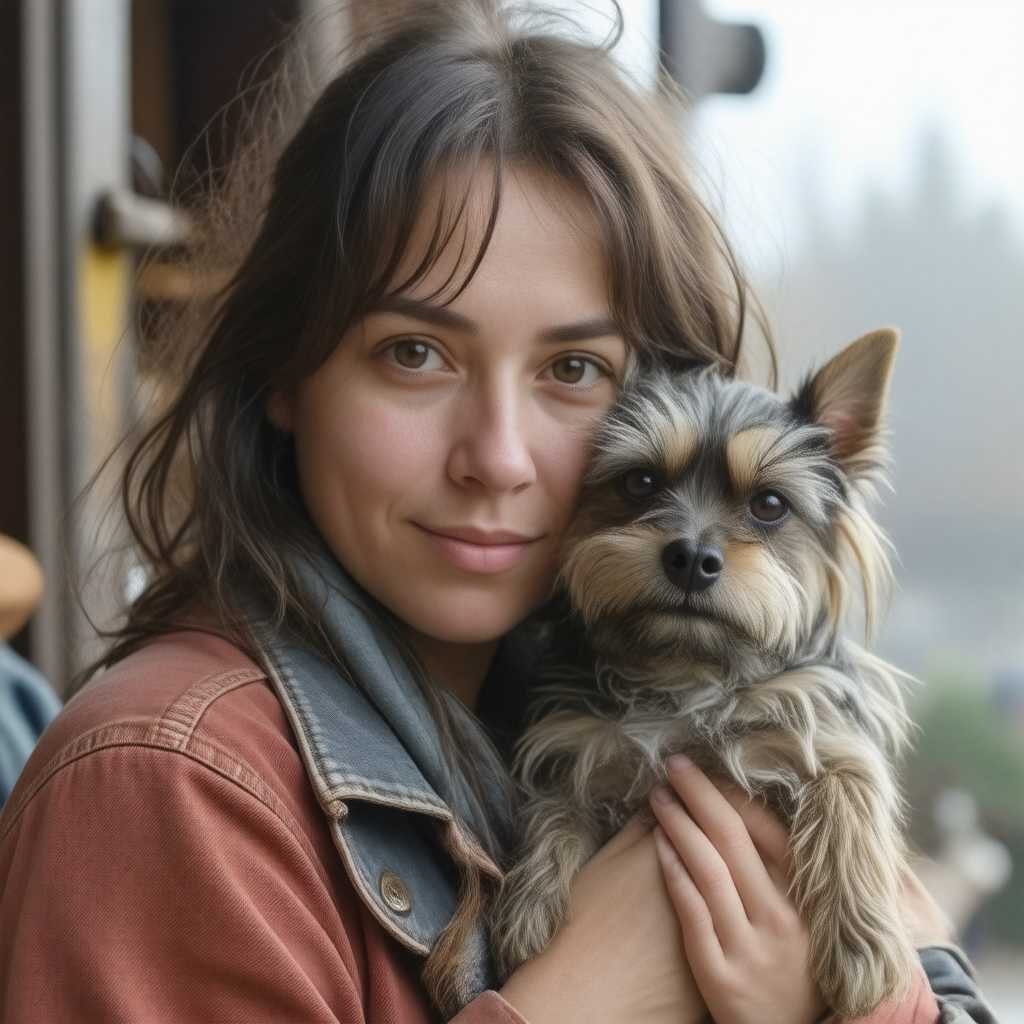} &
    \includegraphics[width=\imagewidth]{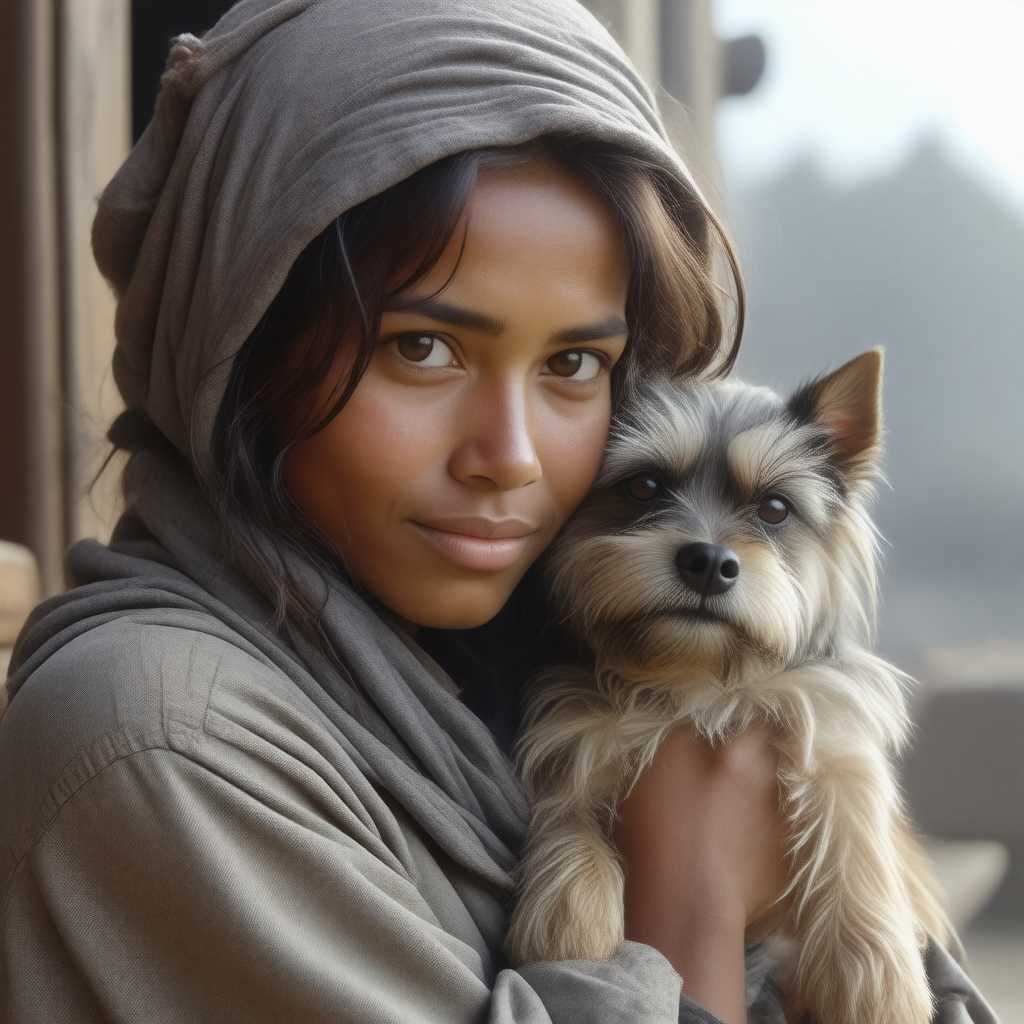} &
    \includegraphics[width=\imagewidth]{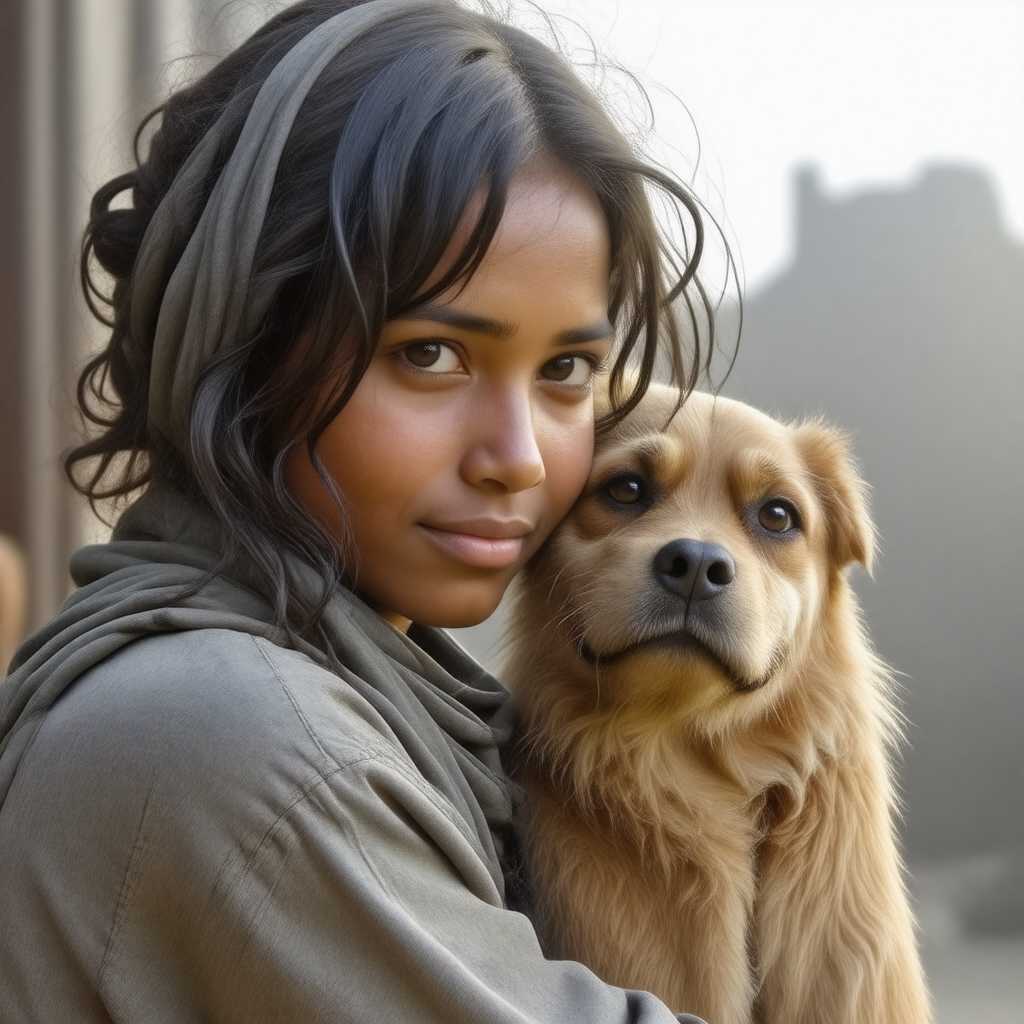} \\
    \multicolumn{6}{l}{\textbf{Contextual Space}} \\ \noalign{\smallskip}

    \includegraphics[width=\imagewidth]{images/analysis/interpolations/pet/target.jpg} &
    \includegraphics[width=\imagewidth]{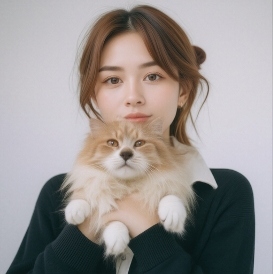} &
    \includegraphics[width=\imagewidth]{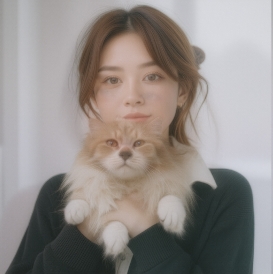} &
    \includegraphics[width=\imagewidth]{images/analysis/interpolations/pet/source.jpg} &
    \includegraphics[width=\imagewidth]{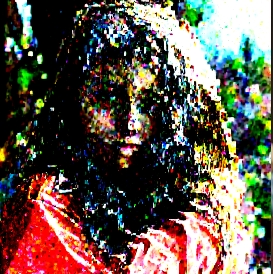} &
    \includegraphics[width=\imagewidth]{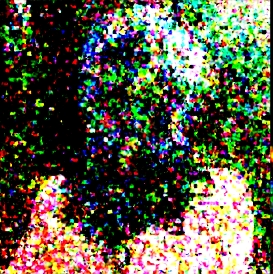} \\
    \multicolumn{6}{l}{\textbf{Latent Space}} \\ \noalign{\smallskip}
    
    \end{tabular}
    \caption{\textbf{Comparison of interpolation and extrapolation between the internal representations of two images.} Intermediate frames are generated by denoising the source image while linearly blending its internal features with those of the target; extrapolation extends this vector beyond the endpoints. While Latent Space interpolation leads to structural blurring and artifacts due to spatial misalignment, the Contextual Space maintains high visual fidelity. This demonstrates that the Contextual Space enables smooth semantic transitions by decoupling generative intent from fixed spatial structures.}
    \label{fig:interpolation_comparison}
\end{figure}

As illustrated in Figure~\ref{fig:interpolation_comparison}, the results highlight a fundamental difference in how these spaces handle semantic information. In the VAE Latent Space, representations are tied to the specific spatial grid and pixel-level layout of the sample. Since the source and target images are spatially unaligned (exhibiting different poses and compositions) interpolating between them results in a structural blur. The model attempts to resolve two conflicting geometries simultaneously, leading to incoherent overlays and ghostly artifacts. More critically, extrapolating in the VAE Latent Space quickly pushes the latents outside the learned data manifold, resulting in severe artifacts.

In contrast, performing the same operation within the Contextual Space yields a smooth semantic transition. Rather than blending pixels or geometries, the model reallocates visual elements in a coherent manner, gradually modifying appearance and composition while maintaining a sharp, high-fidelity structure. For instance, as we move from the source image toward the target, we observe a meaningful evolution in high-level appearance attributes of the subject, such as facial features and overall visual style, which shift naturally from the source toward the target.
In the bottom example, this transition applies coherently to each subject independently, with both the woman and the accompanying pet undergoing meaningful semantic changes (e.g., the pet gradually shifting from a dog-like to a cat-like appearance).
Throughout this interpolation, the pre-trained weights retain the generated images on-manifold, preserving structural integrity and visual plausibility.

Furthermore, the Contextual Space maintains its integrity during extrapolation, where the shifts remain semantically consistent with the direction of the steering vector ($\mathbf{h}_{target} - \mathbf{h}_{source}$). As shown in the right-most columns of Figure~\ref{fig:interpolation_comparison}, applying extrapolation ($\alpha < 0$) relative to the target does not lead to manifold collapse. Instead, it generates a semantically meaningful extrapolation: In the top
example, extrapolation progressively removes the creature’s horns and beast-like features, producing a plausible semantic evolution rather than noise or collapse. In the bottom
example, the woman's features evolve toward a darker tone, effectively moving away from the characteristics of the reference. Simultaneously, the pet’s appearance is modified in a logically consistent manner, such as deepening the coat color and shifting the ears to a more drooping shape. These observations suggest that the Contextual Space encodes global semantic features independently of a fixed spatial grid. Intervening in this space enables the modification of high-level attributes while the transformer's attention mechanisms maintain the structural coherence of the output.

\section{Experiments}
\label{sec:experiments}
To evaluate the generality of our approach, we conduct experiments across three state-of-the-art Diffusion Transformer (DiT) architectures that span distinct design choices and sampling regimes: Flux-dev~\cite{flux2024}, a guidance-distilled model; SD3.5-Turbo, distilled for high-speed, few-step inference; and SD3.5-Large~\cite{esser2024scaling}, a standard non-distilled model. Together, these models cover a broad spectrum of modern DiT variants, allowing us to demonstrate that Contextual Space repulsion is broadly applicable and not tied to a specific architecture, training regime, or sampling budget.

We compare our Contextual Space repulsion against representative diversity-enhancing baselines, including upstream methods that modify initial conditions such as CADS~\cite{sadat2023cads} and SGI~\cite{parmar2025scaling}, as well as downstream methods that intervene in the latent space, including Particle Guidance~\cite{corso2023particle} and SPARKE~\cite{jalali2025sparke}. Full implementation details and hyperparameter settings are provided in Appendix~\ref{sec:implementation_details}.

\subsection{Qualitative Results}
\paragraph{Flux-dev results.}

\begin{figure}
    \centering
    
    \setlength{\tabcolsep}{0.5pt} \renewcommand{\arraystretch}{0.5} %
    \newcommand{\imgwidth}{%
        0.112\textwidth 
    }
    \newcommand{\vertlabel}[1]{\raisebox{2.5em}{\rotatebox{90}{\scriptsize\textbf{#1}}}}
    \begin{tabular}{c c c c c}
        \vertlabel{Flux} & \includegraphics[width=\imgwidth]{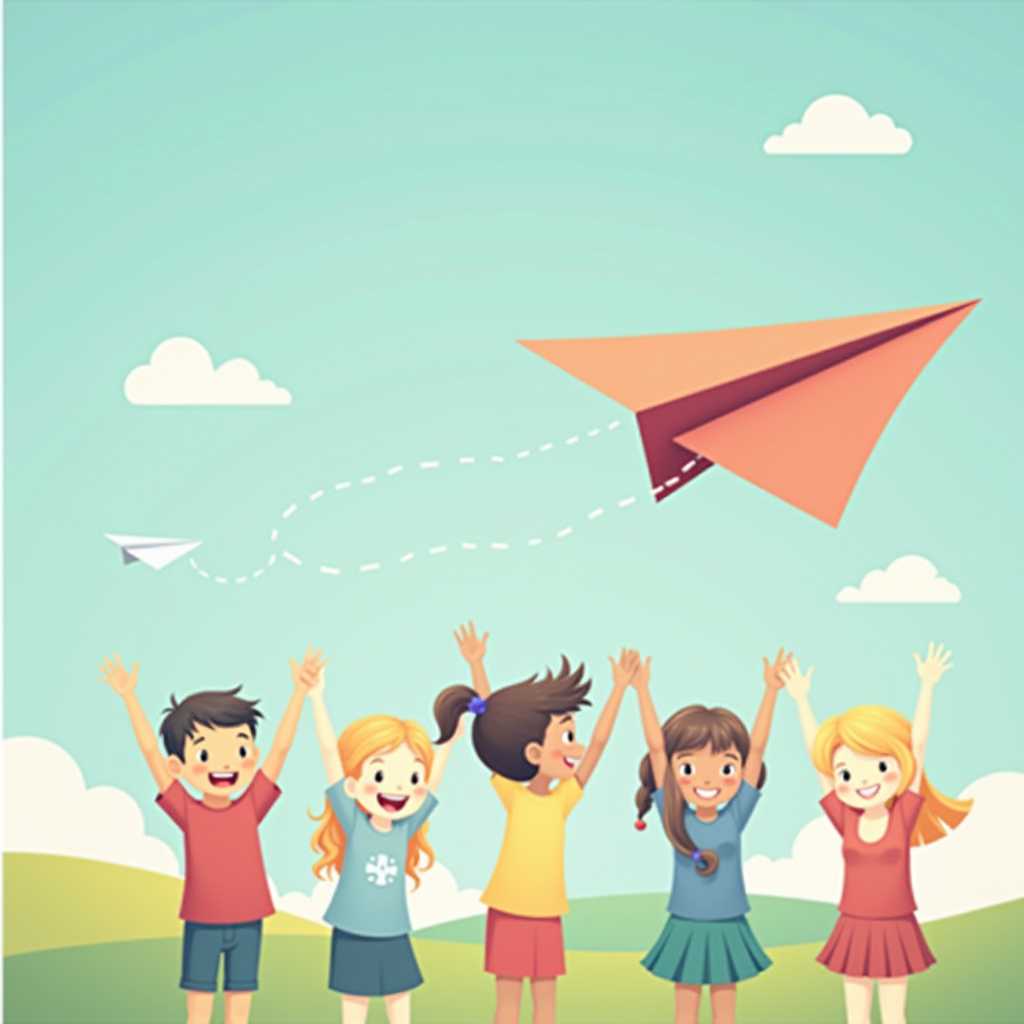} & \includegraphics[width=\imgwidth]{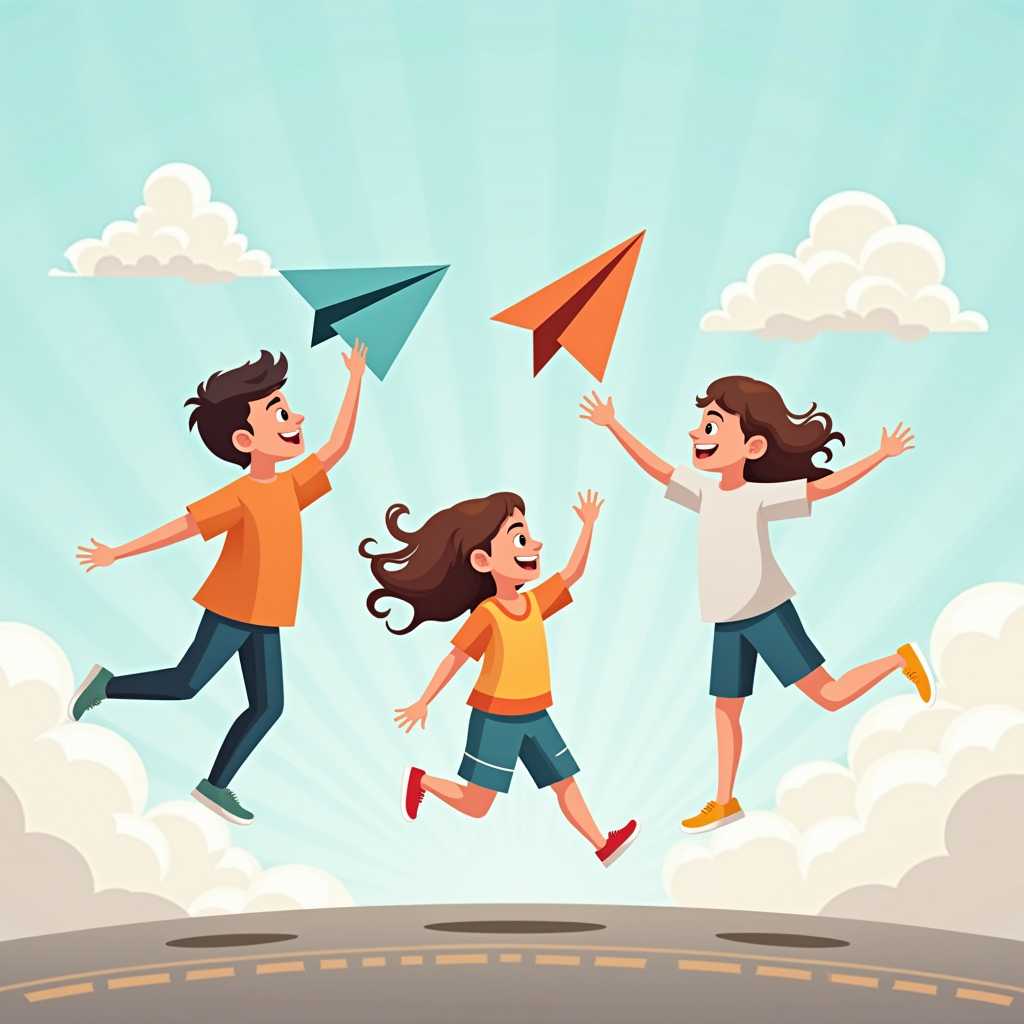} & \includegraphics[width=\imgwidth]{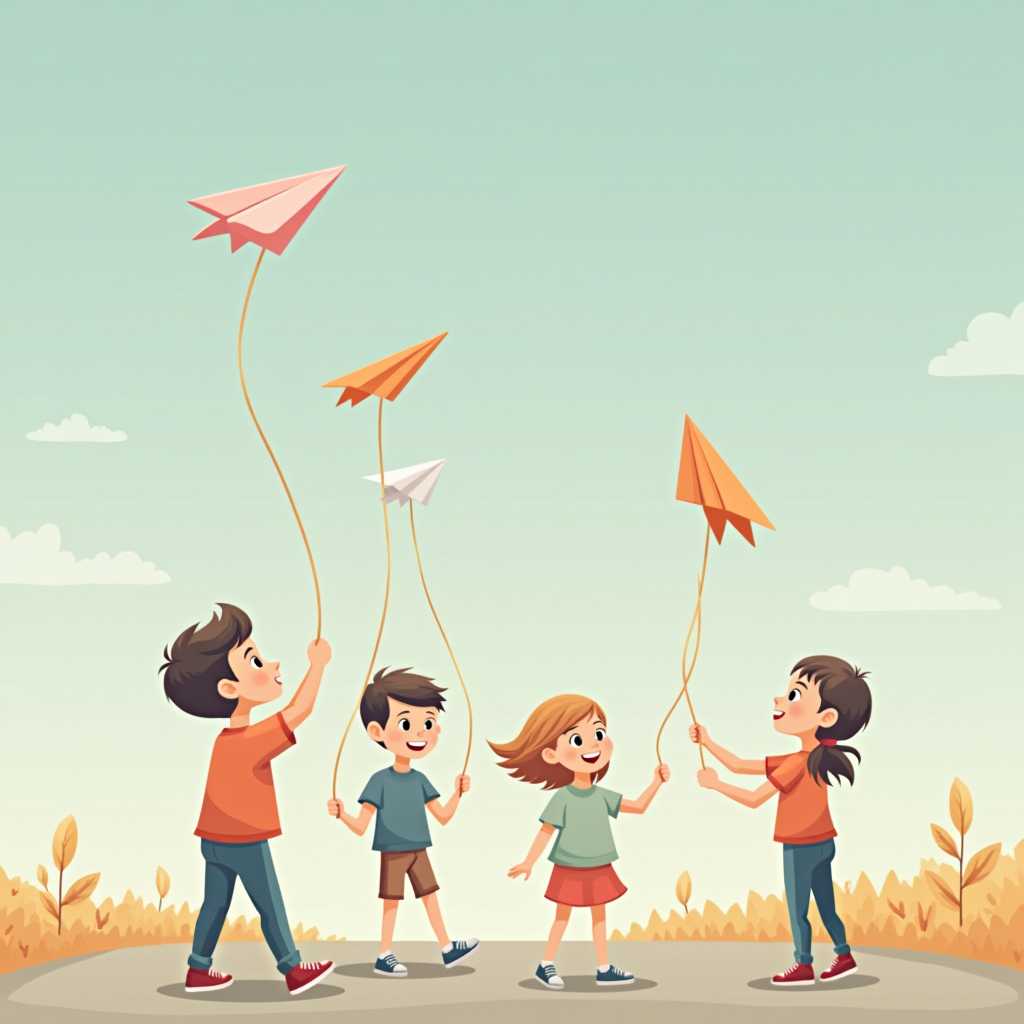} & \includegraphics[width=\imgwidth]{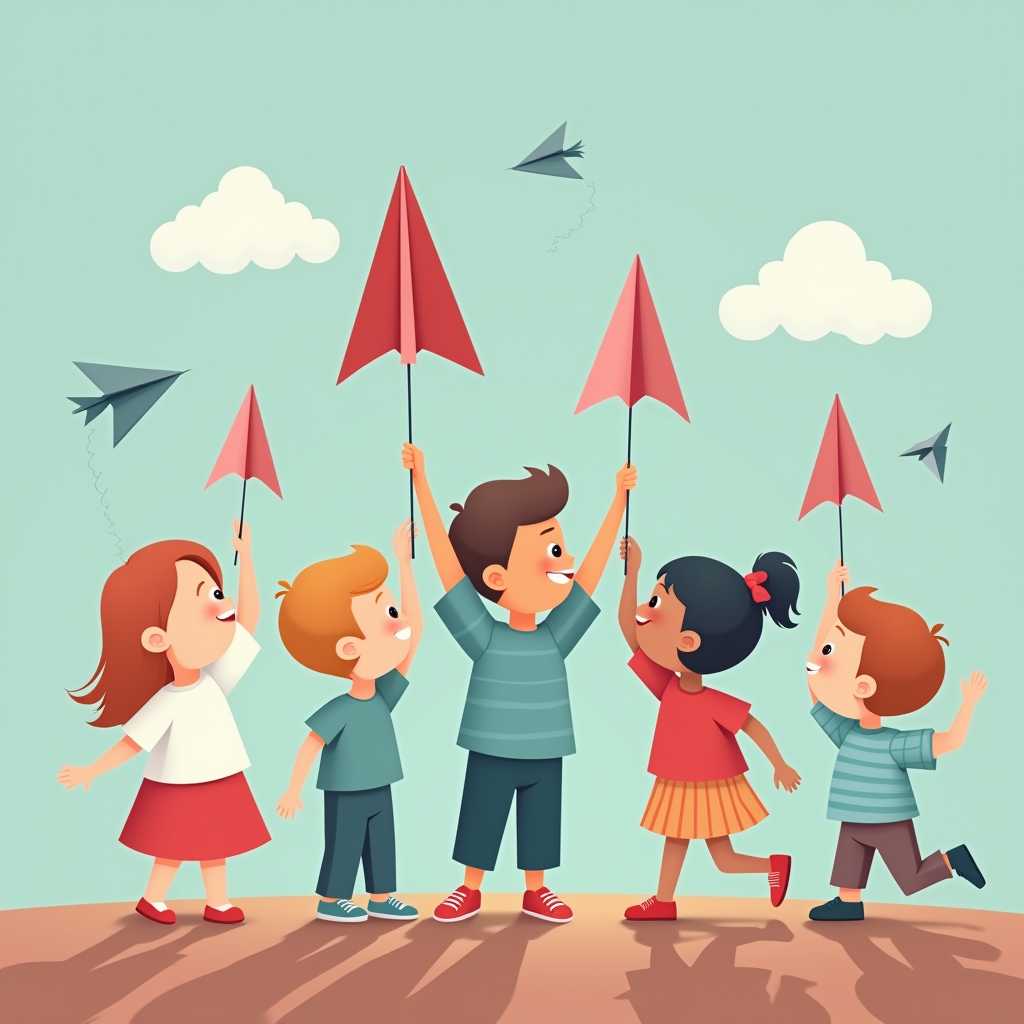} \\[-1pt]
        \vertlabel{Ours} & \includegraphics[width=\imgwidth]{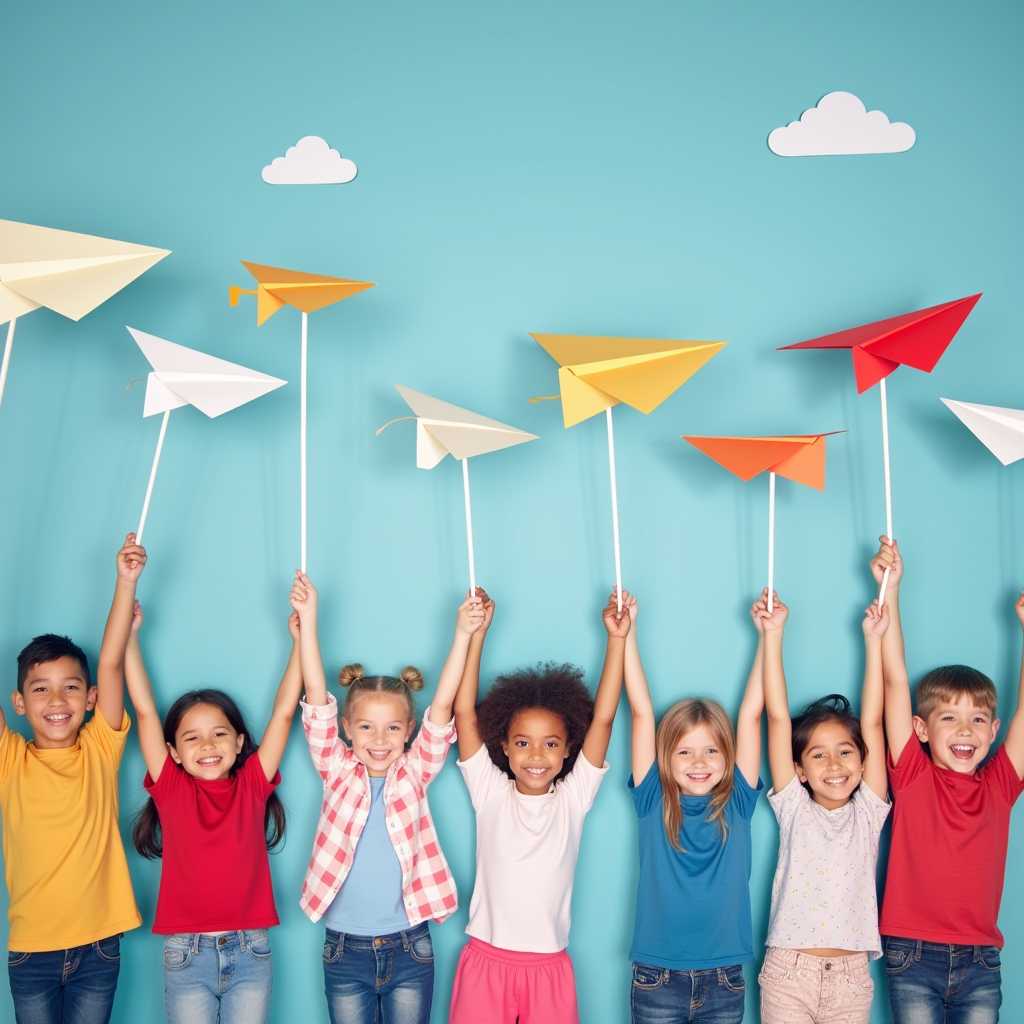} & \includegraphics[width=\imgwidth]{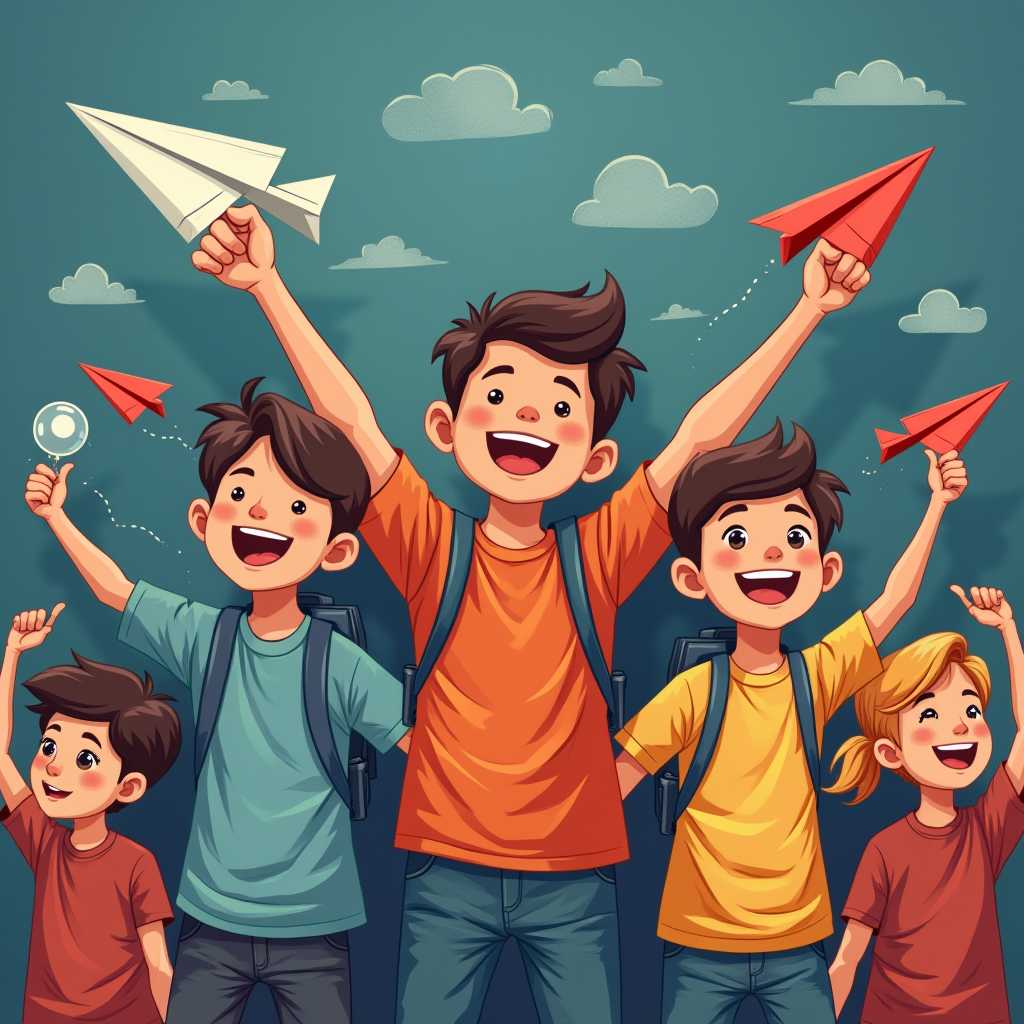} & \includegraphics[width=\imgwidth]{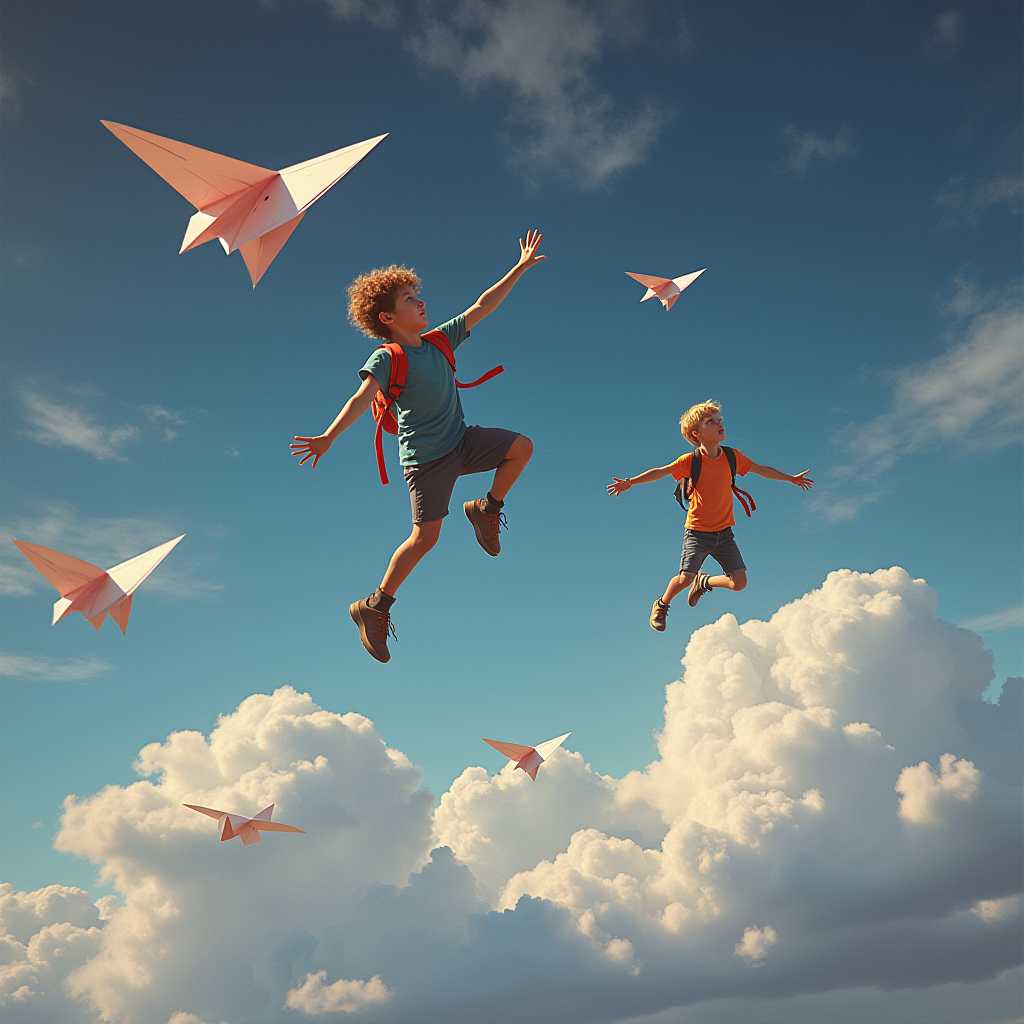} & \includegraphics[width=\imgwidth]{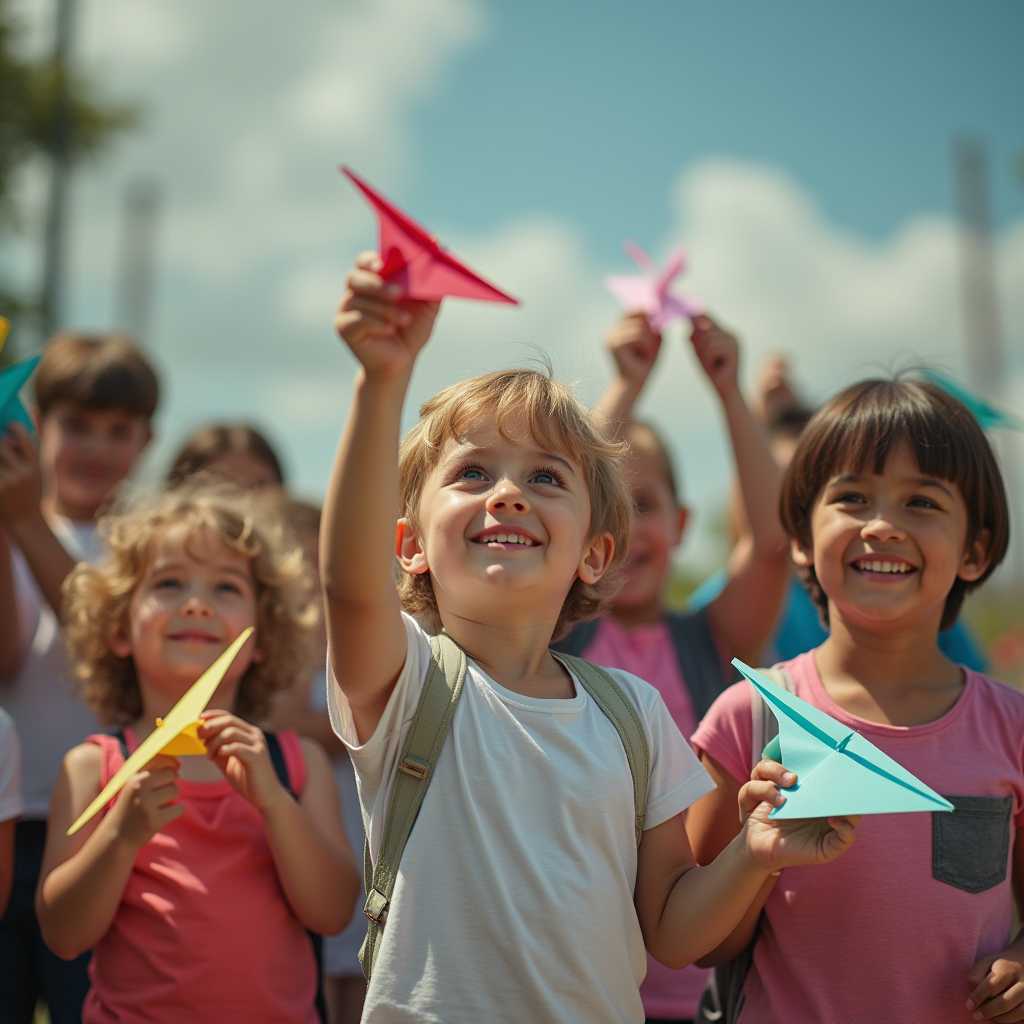} \\
        \multicolumn{5}{c}{\vspace{2pt}\small ``Kids with paper airplanes'' \vspace{8pt}} \\
        \vertlabel{Flux} & \includegraphics[width=\imgwidth]{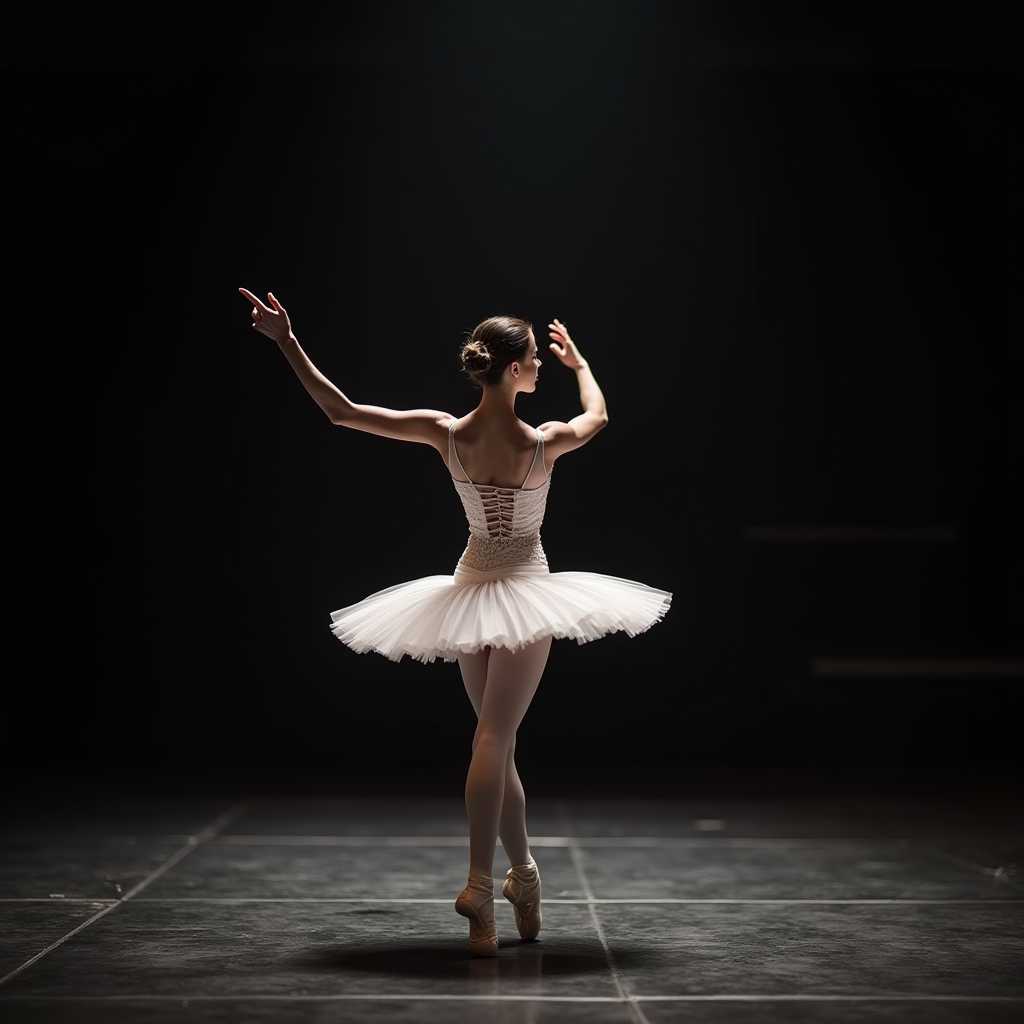} & \includegraphics[width=\imgwidth]{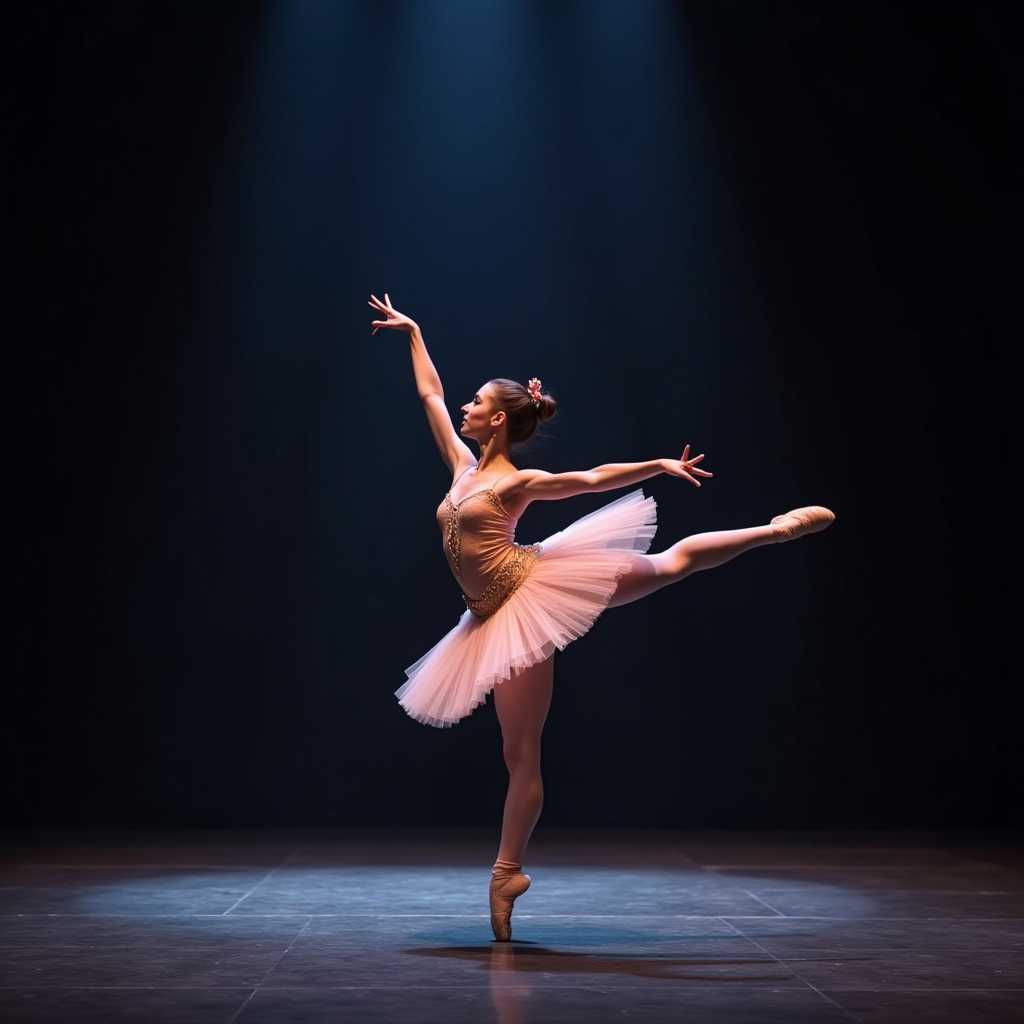} & \includegraphics[width=\imgwidth]{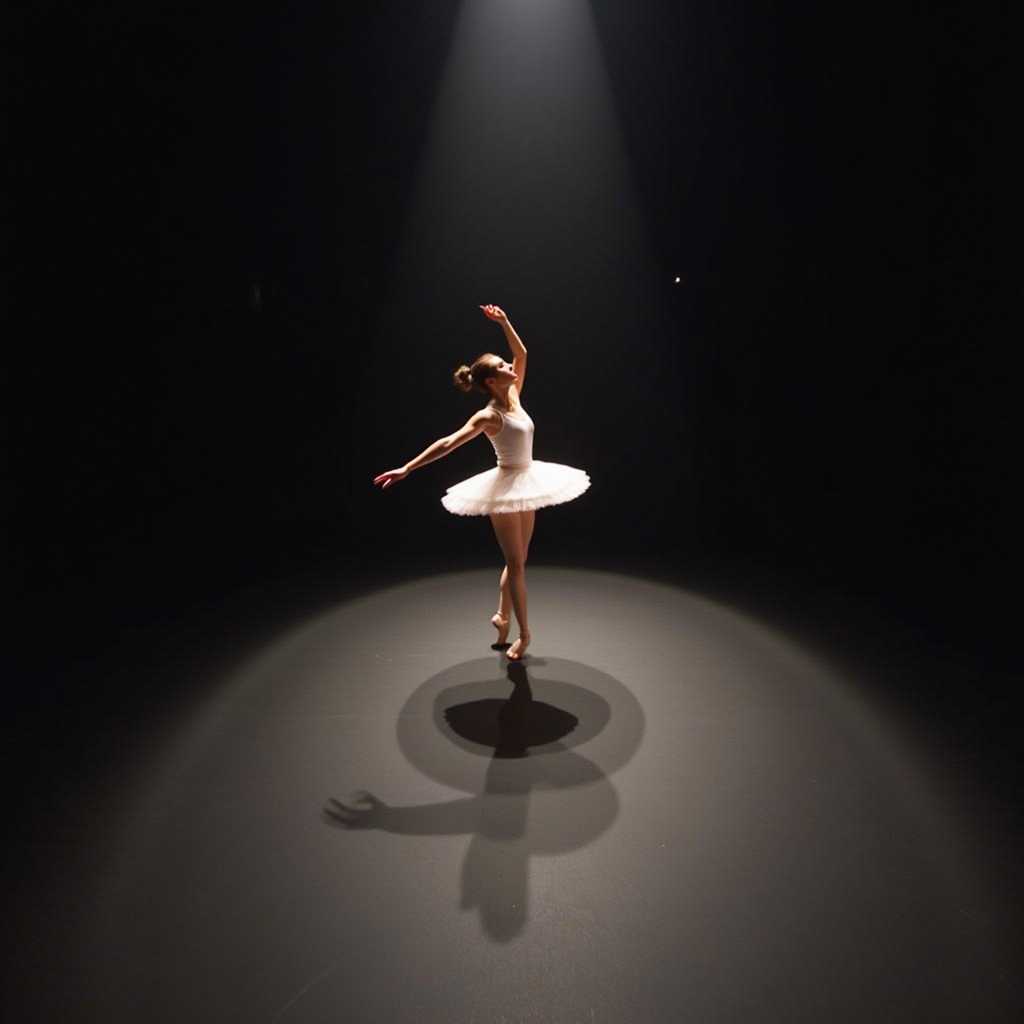} & \includegraphics[width=\imgwidth]{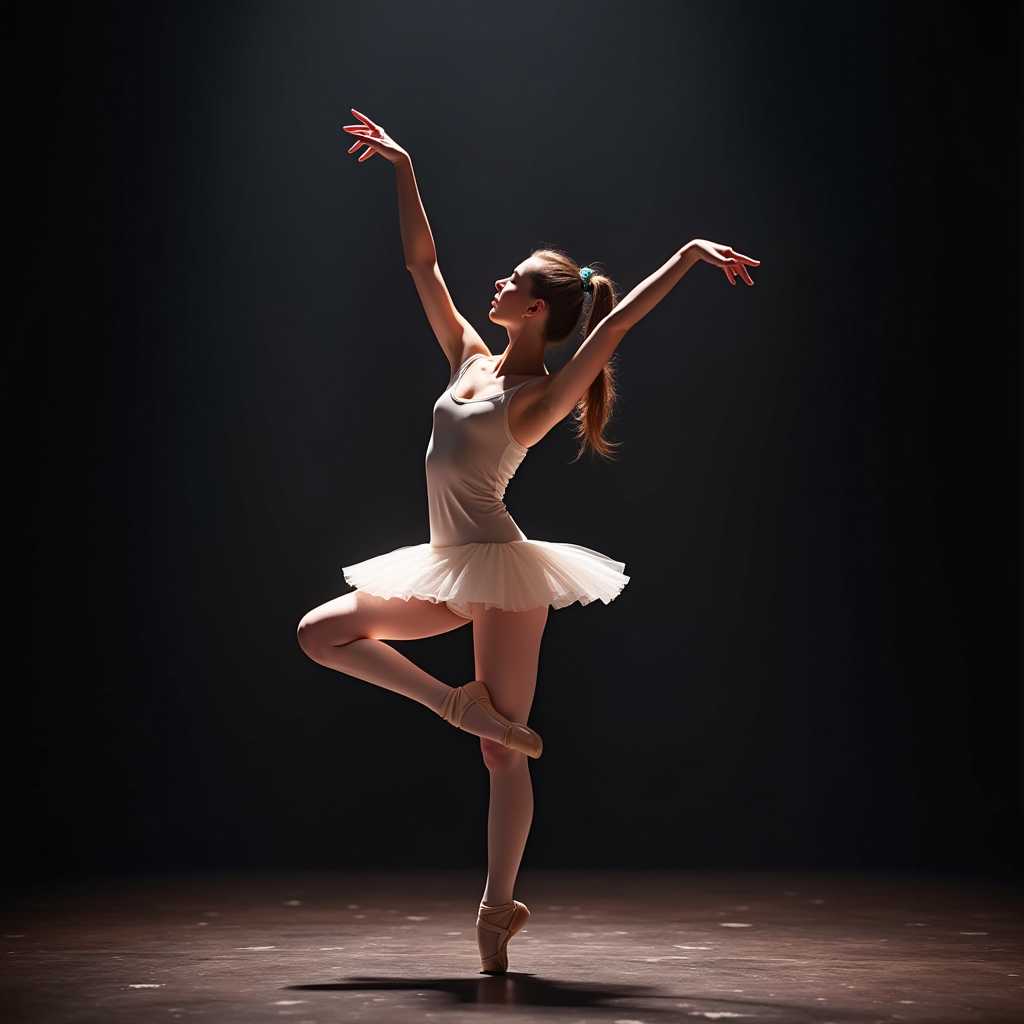} \\[-1pt]
        \vertlabel{Ours} & \includegraphics[width=\imgwidth]{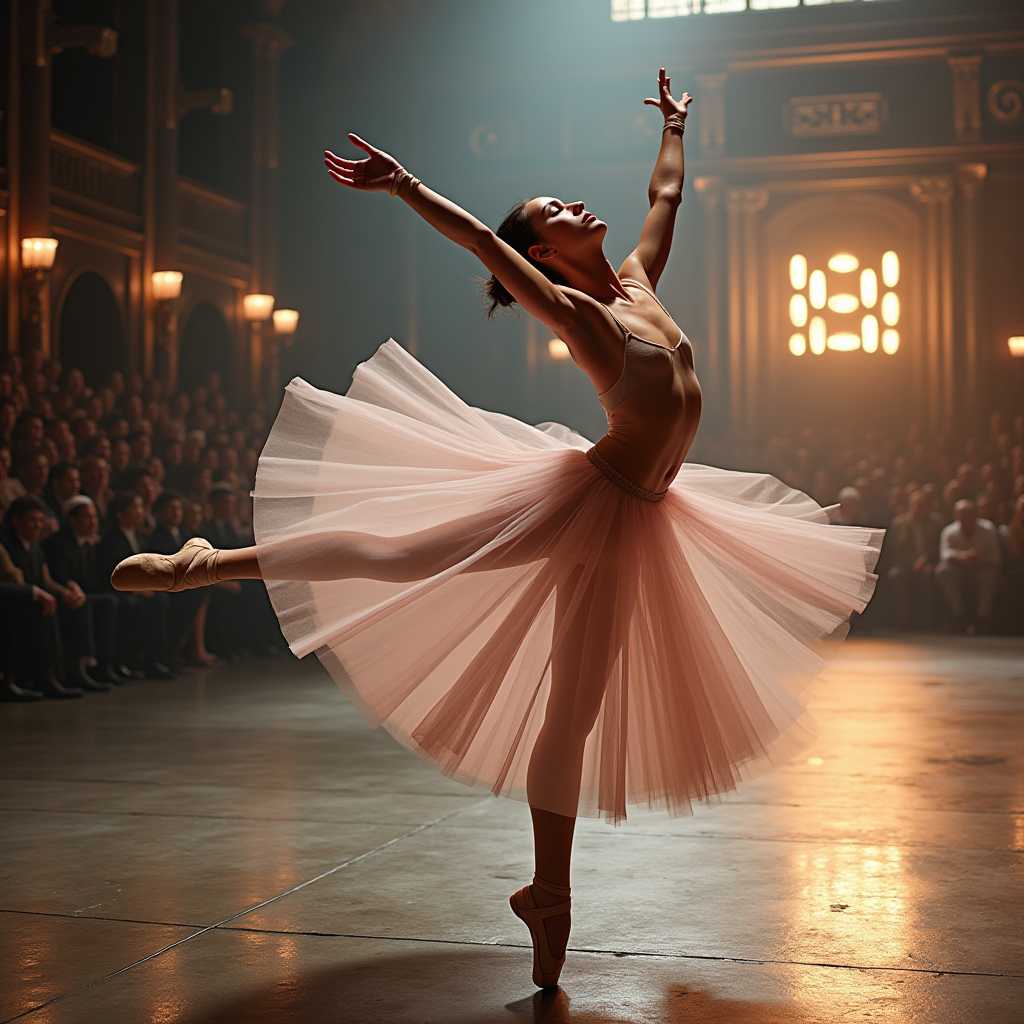} & \includegraphics[width=\imgwidth]{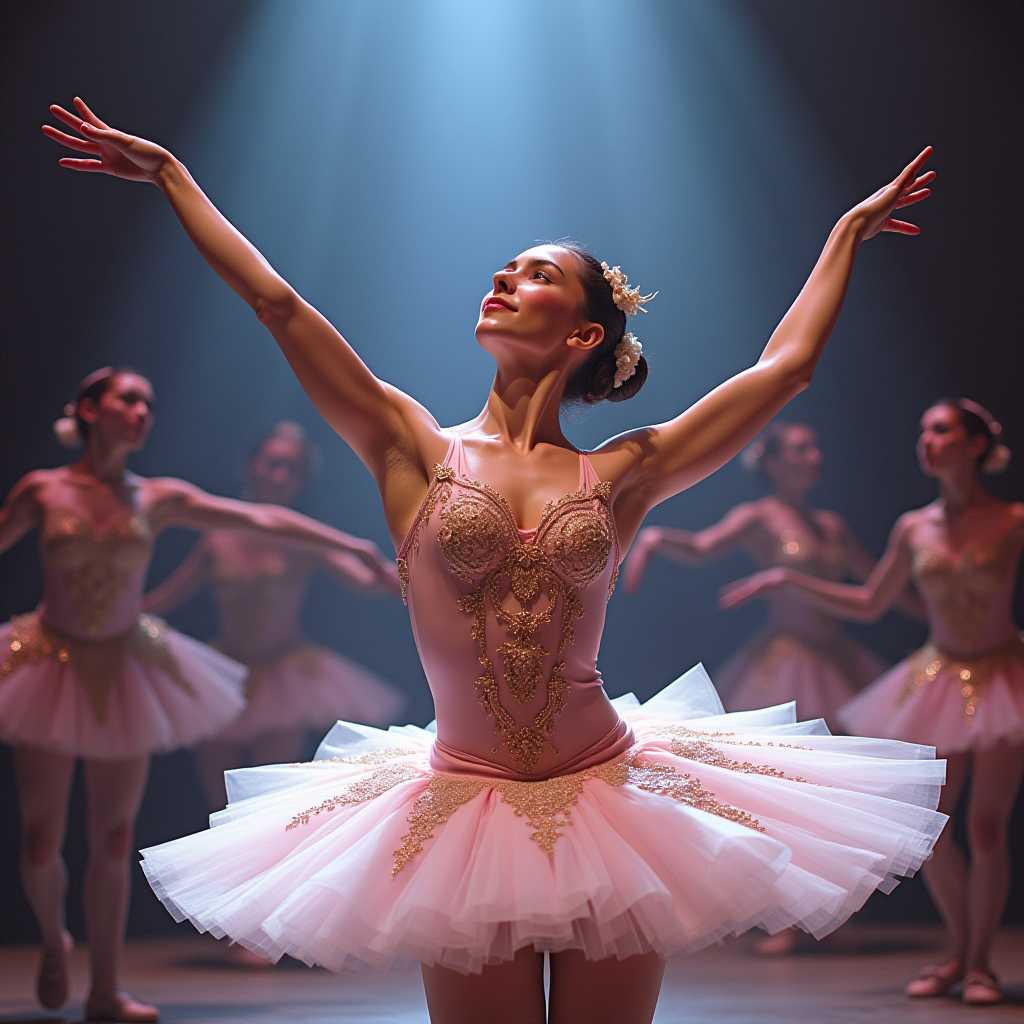} & \includegraphics[width=\imgwidth]{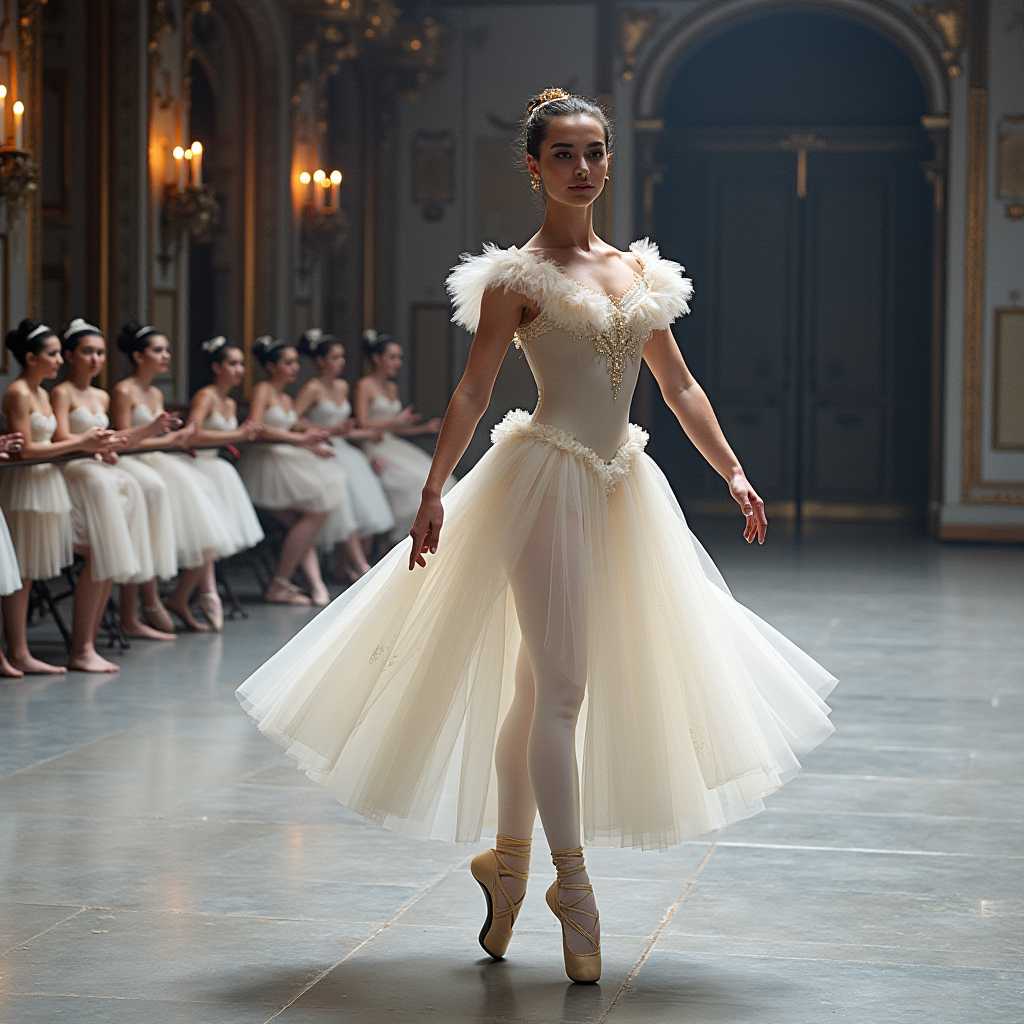} & \includegraphics[width=\imgwidth]{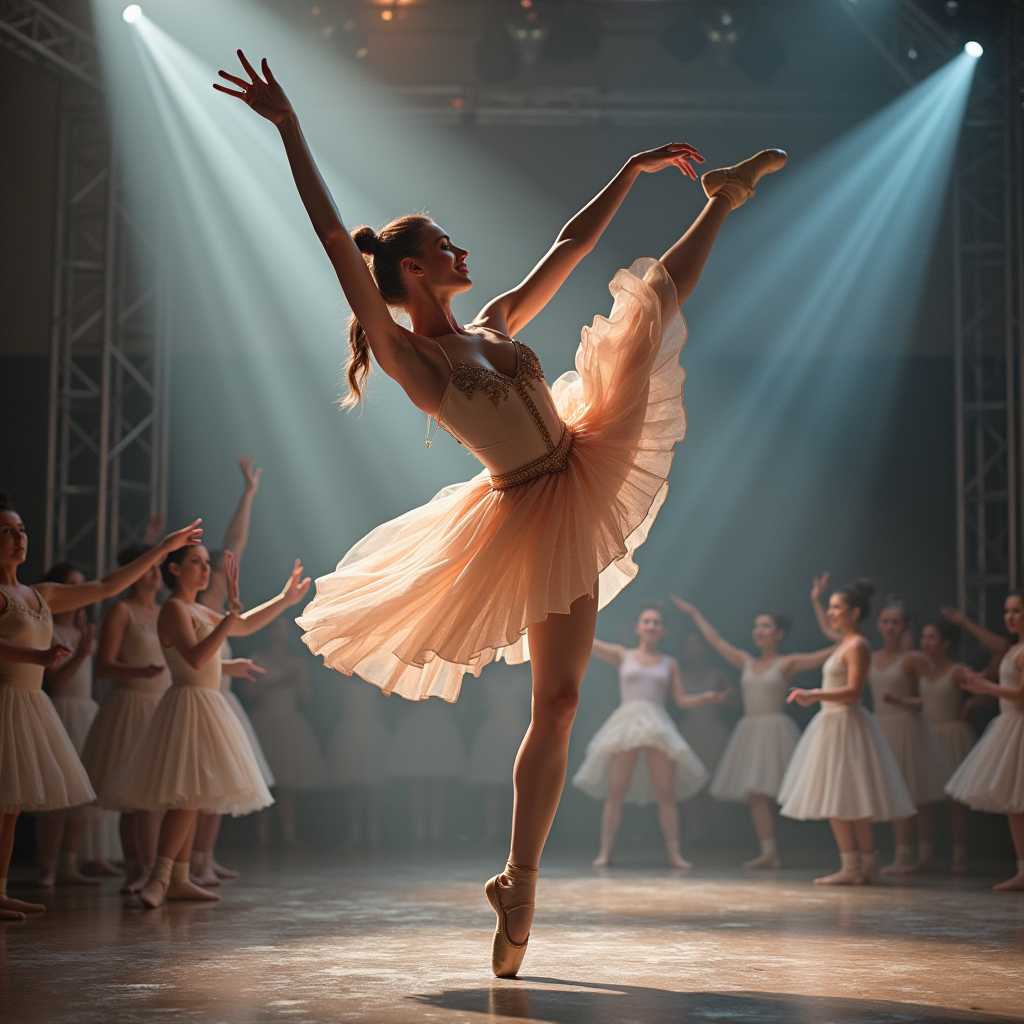} \\
        \multicolumn{5}{c}{\vspace{2pt}\small ``A ballet dancer on stage'' \vspace{8pt}} \\
    \end{tabular}
    \vspace{-20pt}
    \caption{\textbf{Qualitative results.} For each prompt, we compare the base model results (top) to our results (bottom).}
    \label{fig:flux_short}
\end{figure}

We compare our results with the base Flux-dev model in Figures~\ref{fig:flux_short} and \ref{fig:eights};
additional comparisons with Flux-dev, SD3.5-Large and SD3.5-Turbo are provided in Appendix~\ref{sec:additional_qualitative_results}.
For an objective comparison, all qualitative results use the same fixed seed to sample a batch of distinct initial noises.
Despite this, the base model typically produces a narrow and repetitive range of outputs for many prompts.
 As shown in Figure~\ref{fig:eights}, our method alleviates typicality biases, such as the barely visible or harsh lighting seen in the ``musician'' and ``scientist'' examples. Furthermore, it
generates a diverse array of compositions, arrangements, and camera angles for the ``painter'' and ``stadium'' prompts.

\paragraph{Baseline comparisons.}

\begin{figure*}
    \centering
    \setlength{\tabcolsep}{0.5pt} 
    \renewcommand{\arraystretch}{0.5} 
    \newcommand{\imgwidth}{0.24\linewidth}
    \newcommand{\vertlabel}[2]{\raisebox{#2}{\rotatebox{90}{\scriptsize\textbf{#1}}}}

    \begin{minipage}{0.48\textwidth}
        \centering
        \begin{tabular}{c c c c c}
            \vertlabel{Ours}{2.5em} & 
            \includegraphics[width=\imgwidth, height=\imgwidth]{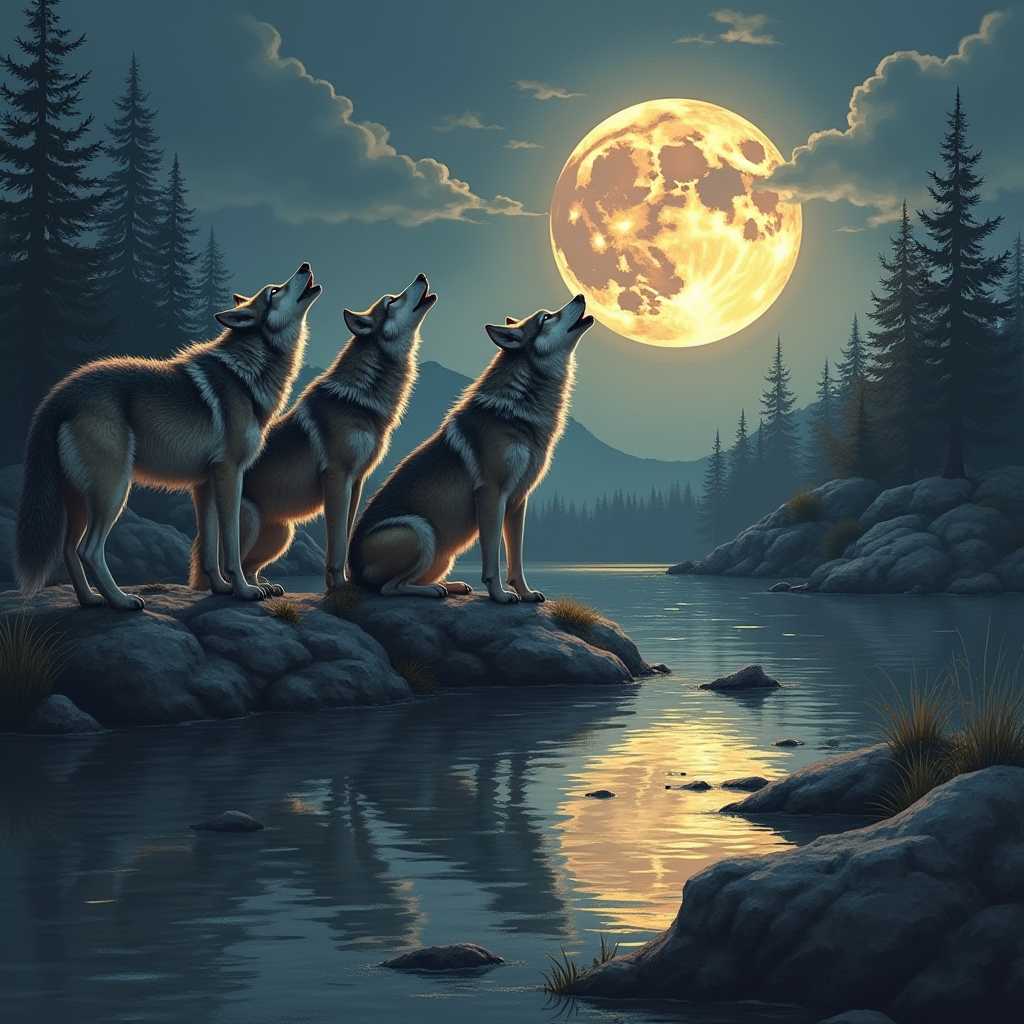} &
            \includegraphics[width=\imgwidth, height=\imgwidth]{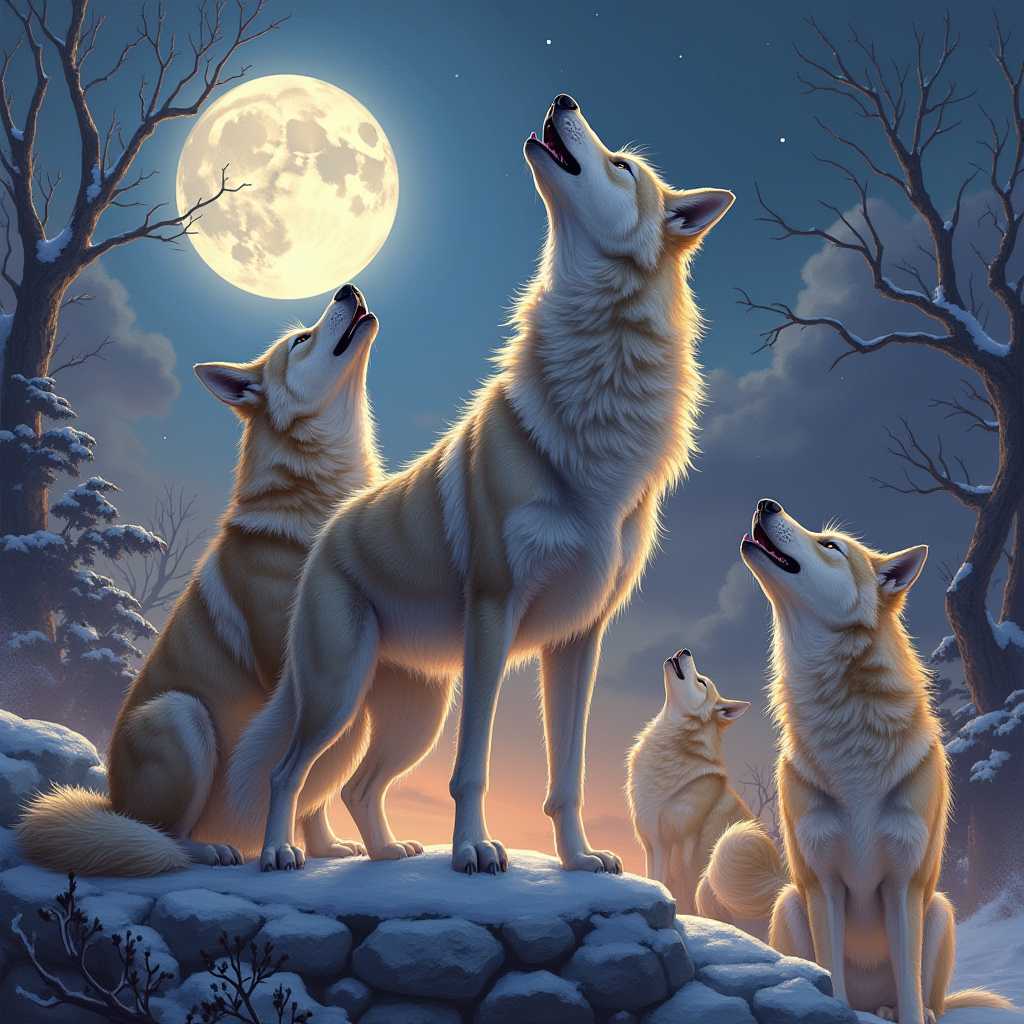} &
            \includegraphics[width=\imgwidth, height=\imgwidth]{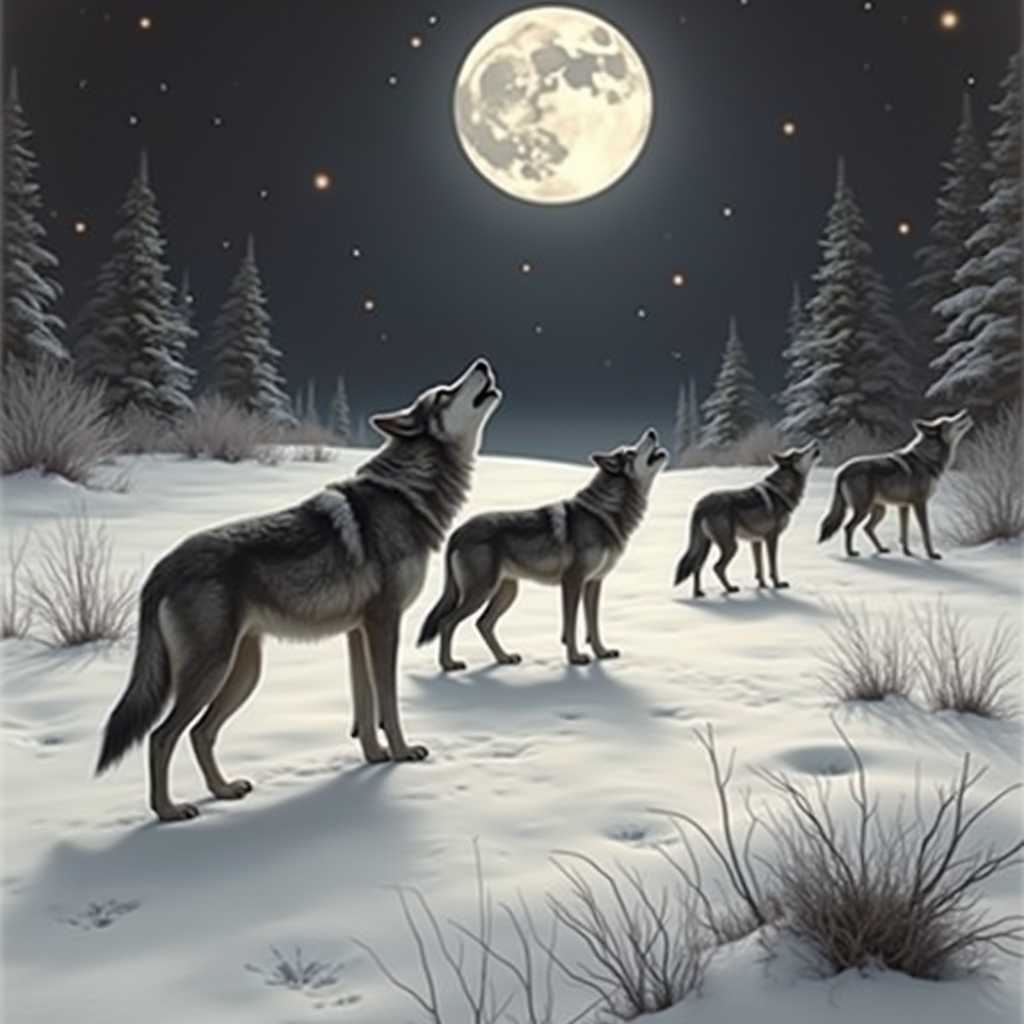} &
            \includegraphics[width=\imgwidth, height=\imgwidth]{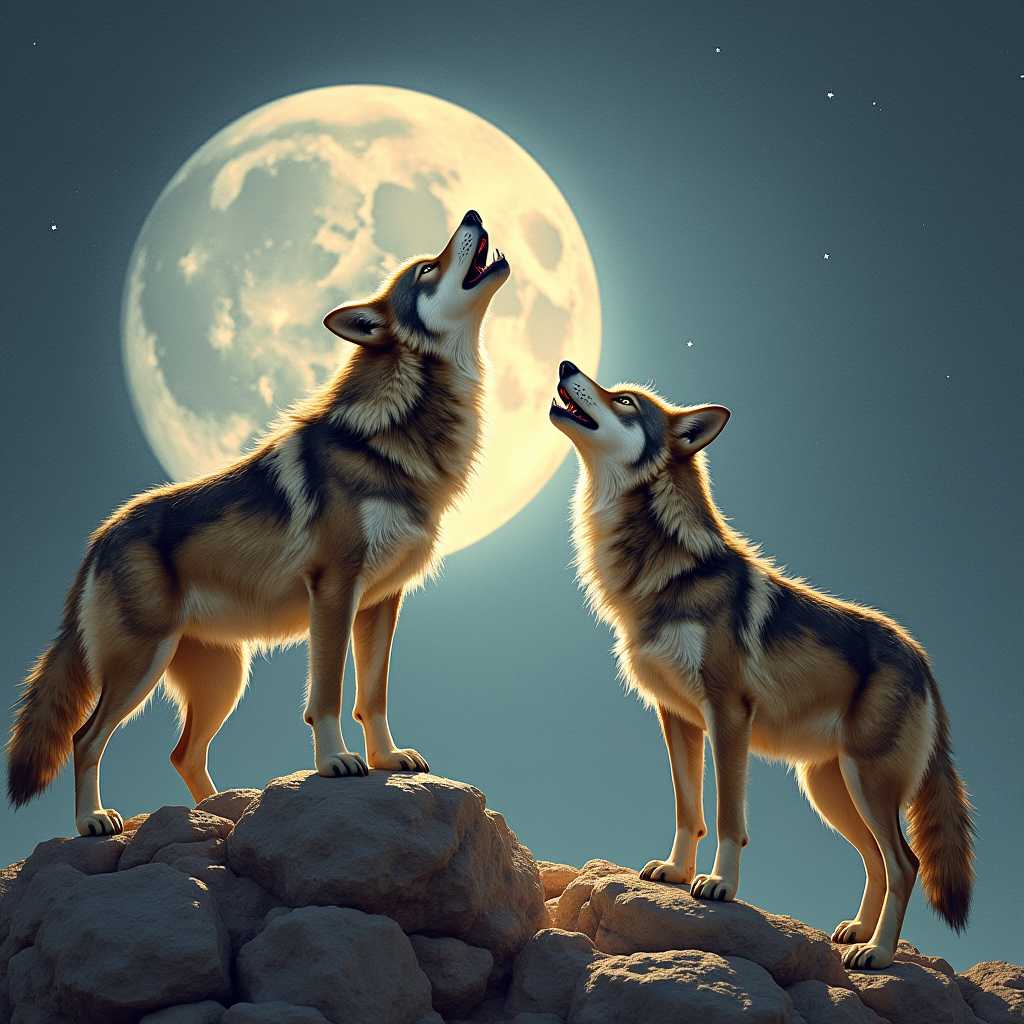} \\

            \vertlabel{SGI}{2.5em} & 
            \includegraphics[width=\imgwidth, height=\imgwidth]{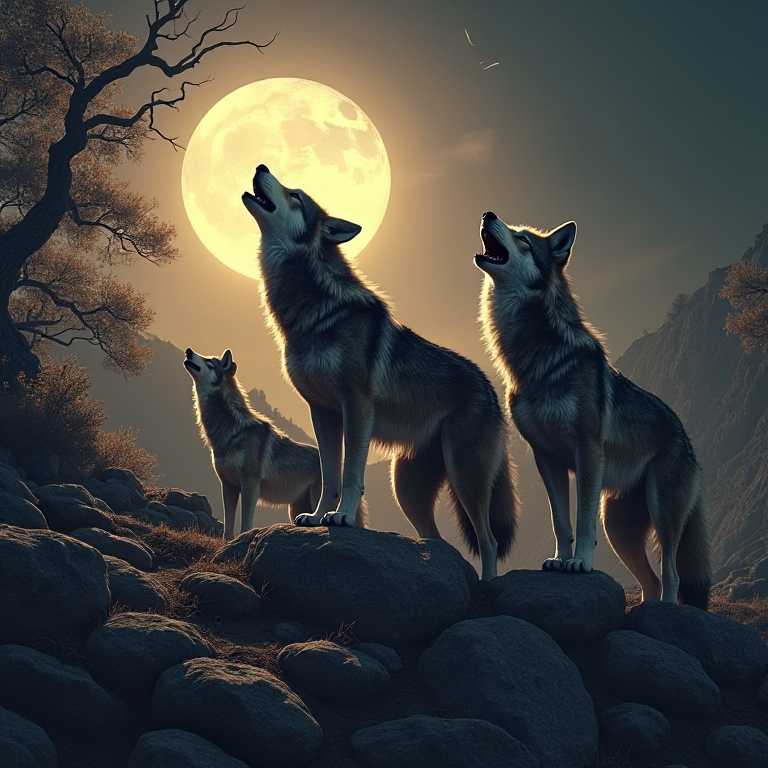} &
            \includegraphics[width=\imgwidth, height=\imgwidth]{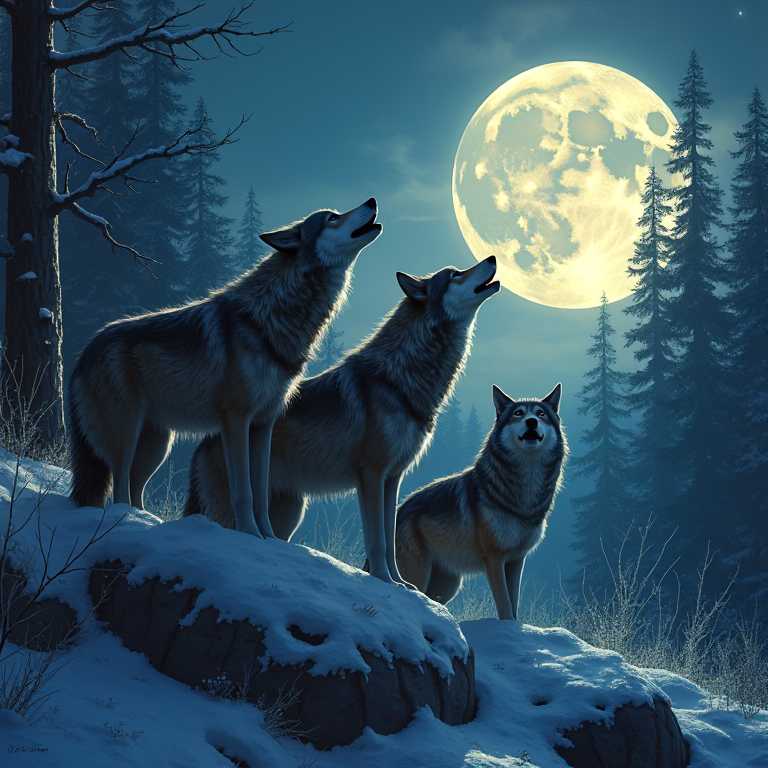} &
            \includegraphics[width=\imgwidth, height=\imgwidth]{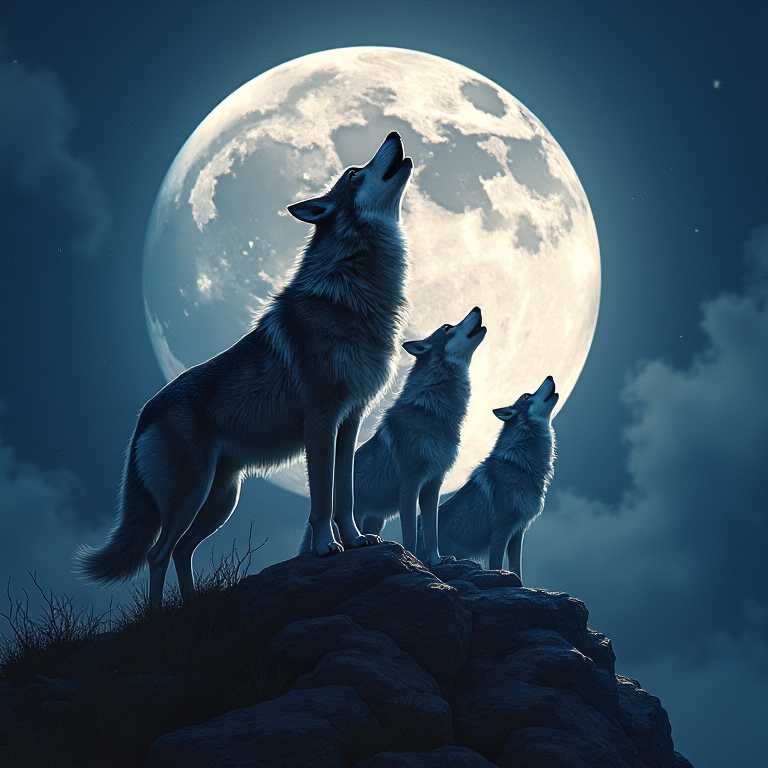} &
            \includegraphics[width=\imgwidth, height=\imgwidth]{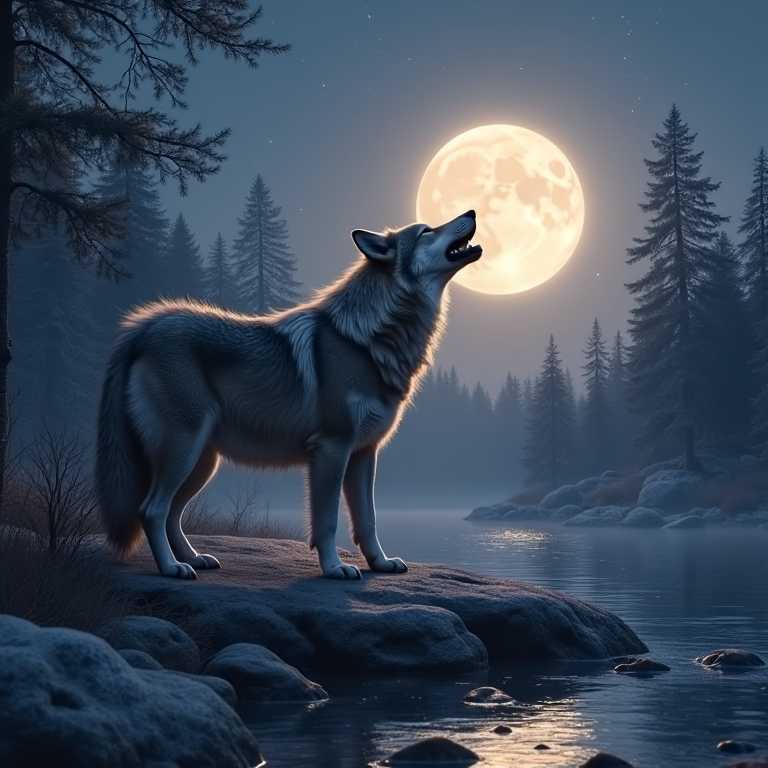} \\
            
            \vertlabel{CADS}{2.5em} & 
            \includegraphics[width=\imgwidth, height=\imgwidth]{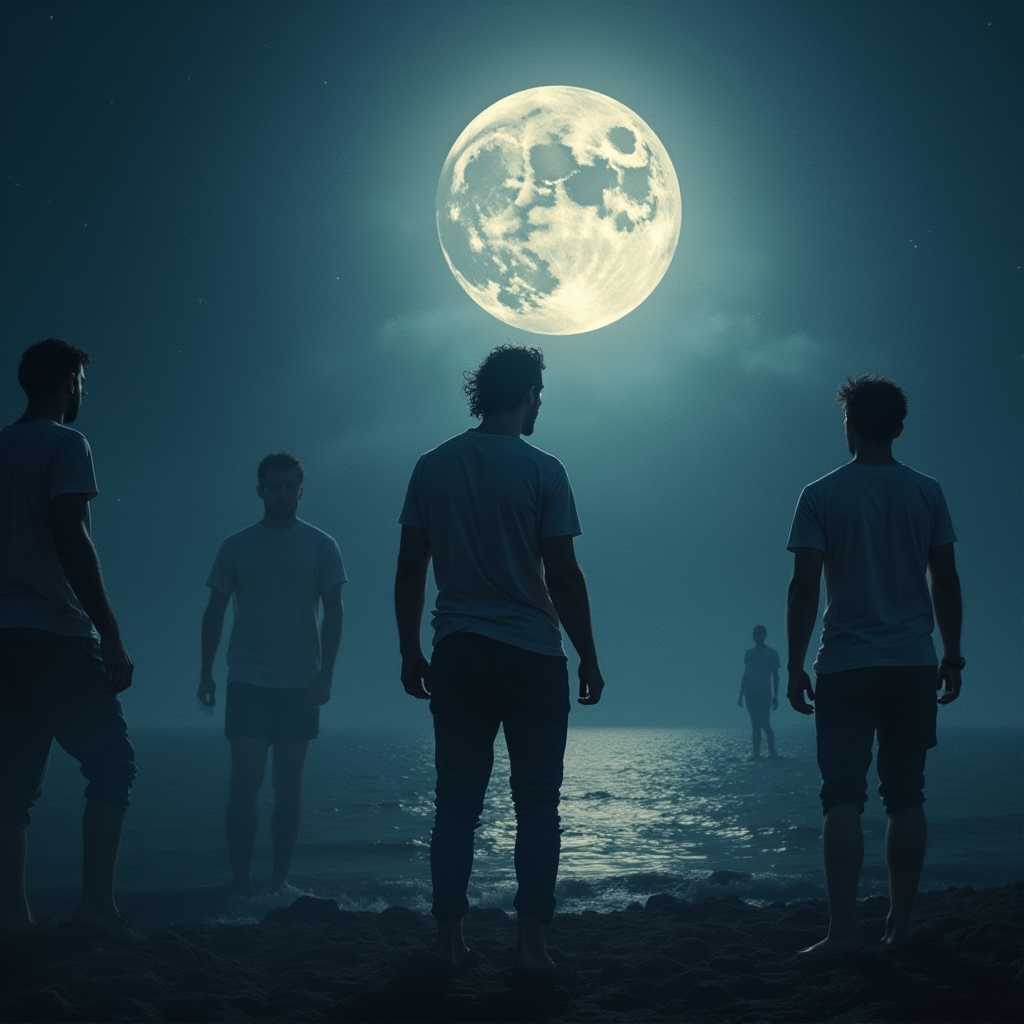} &
            \includegraphics[width=\imgwidth, height=\imgwidth]{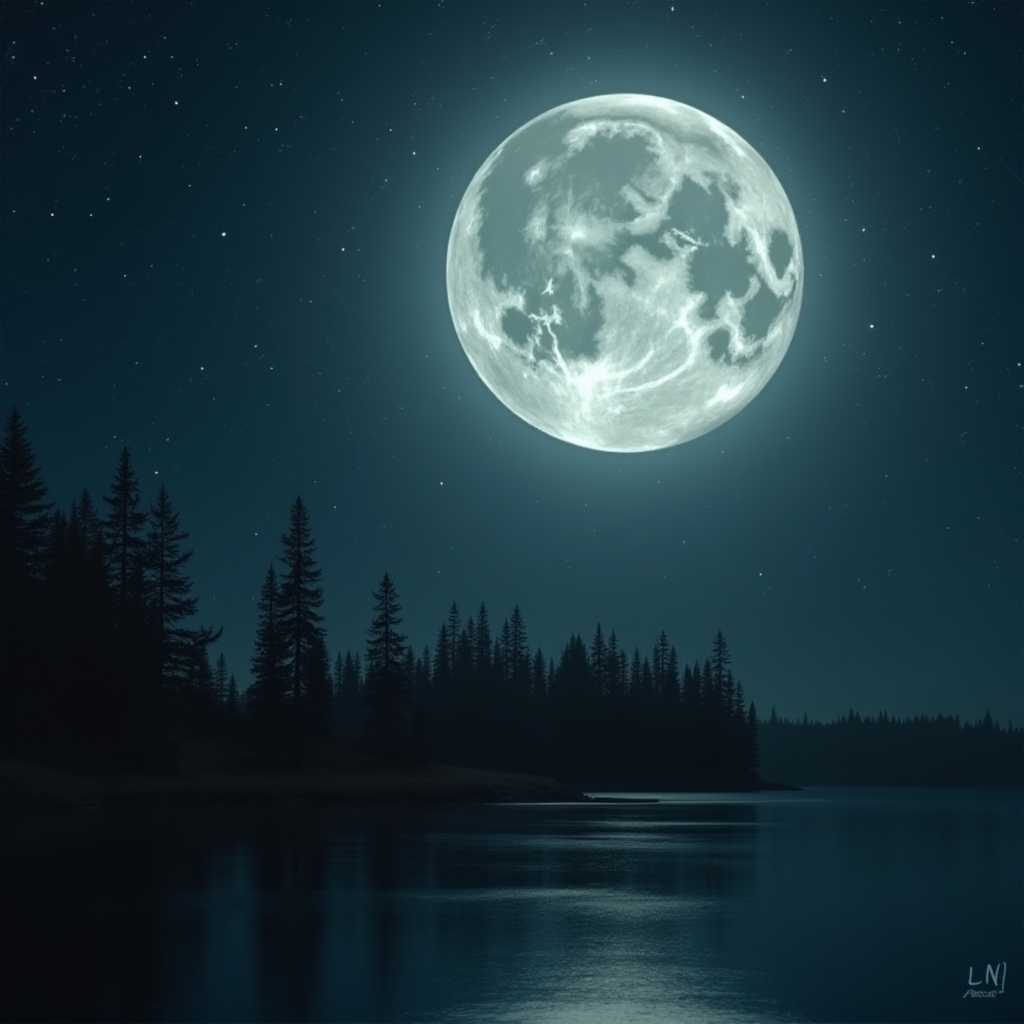} &
            \includegraphics[width=\imgwidth, height=\imgwidth]{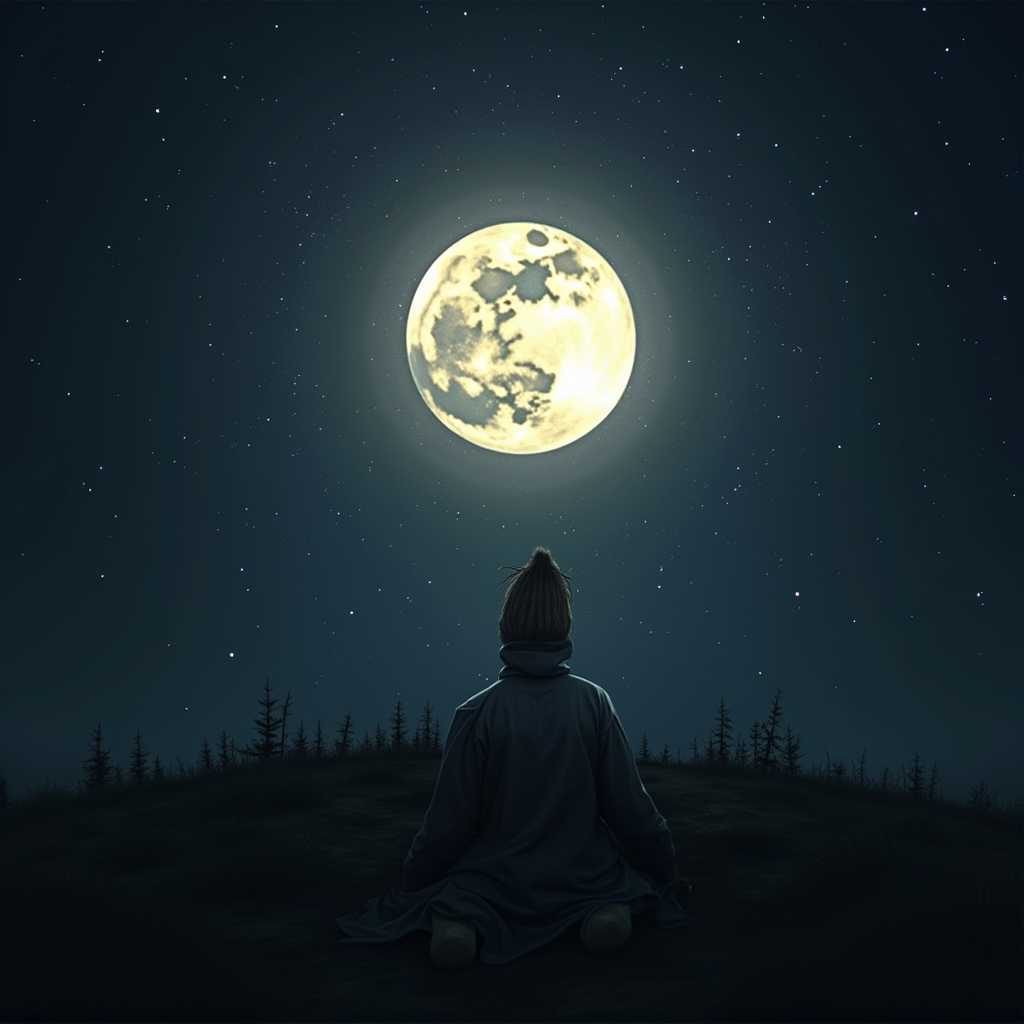} &
            \includegraphics[width=\imgwidth, height=\imgwidth]{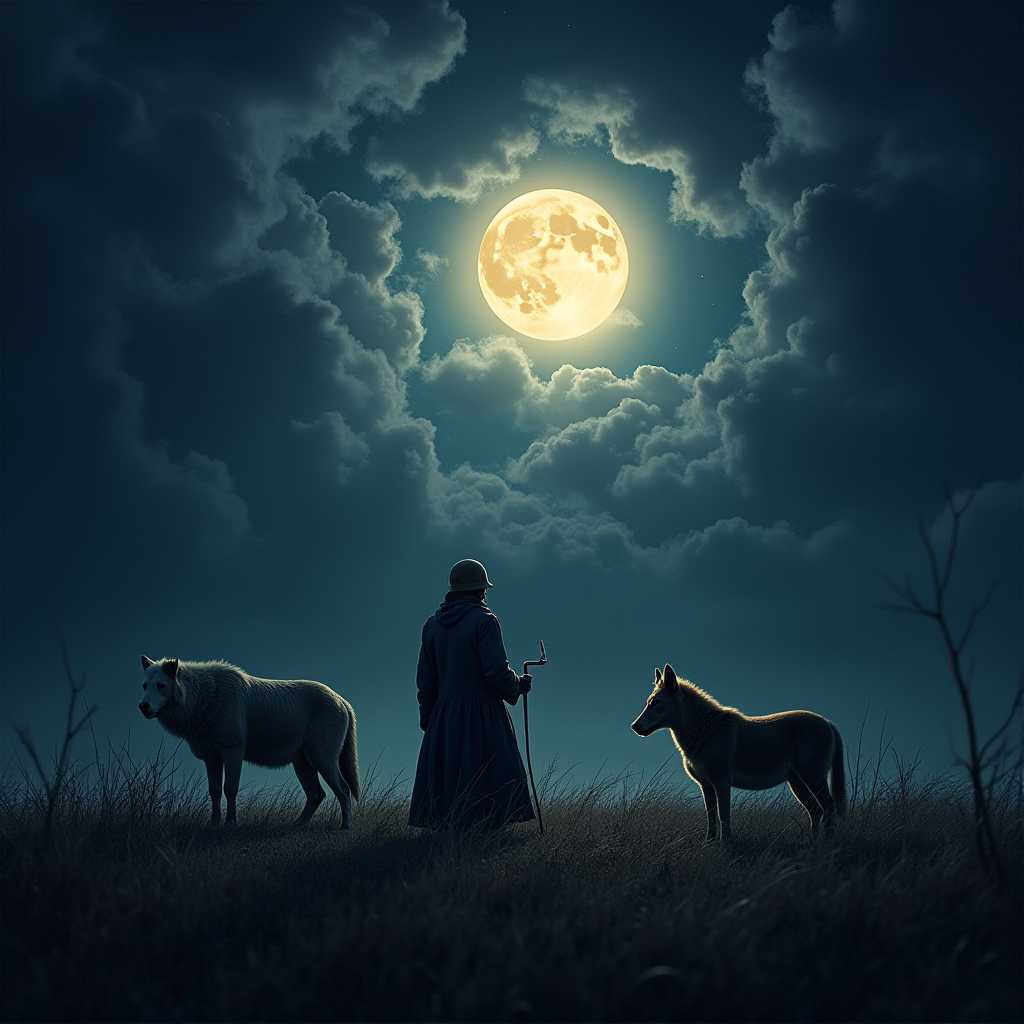} \\

                        \vertlabel{SPARKE}{2.5em} & %
            \includegraphics[width=\imgwidth, height=\imgwidth]{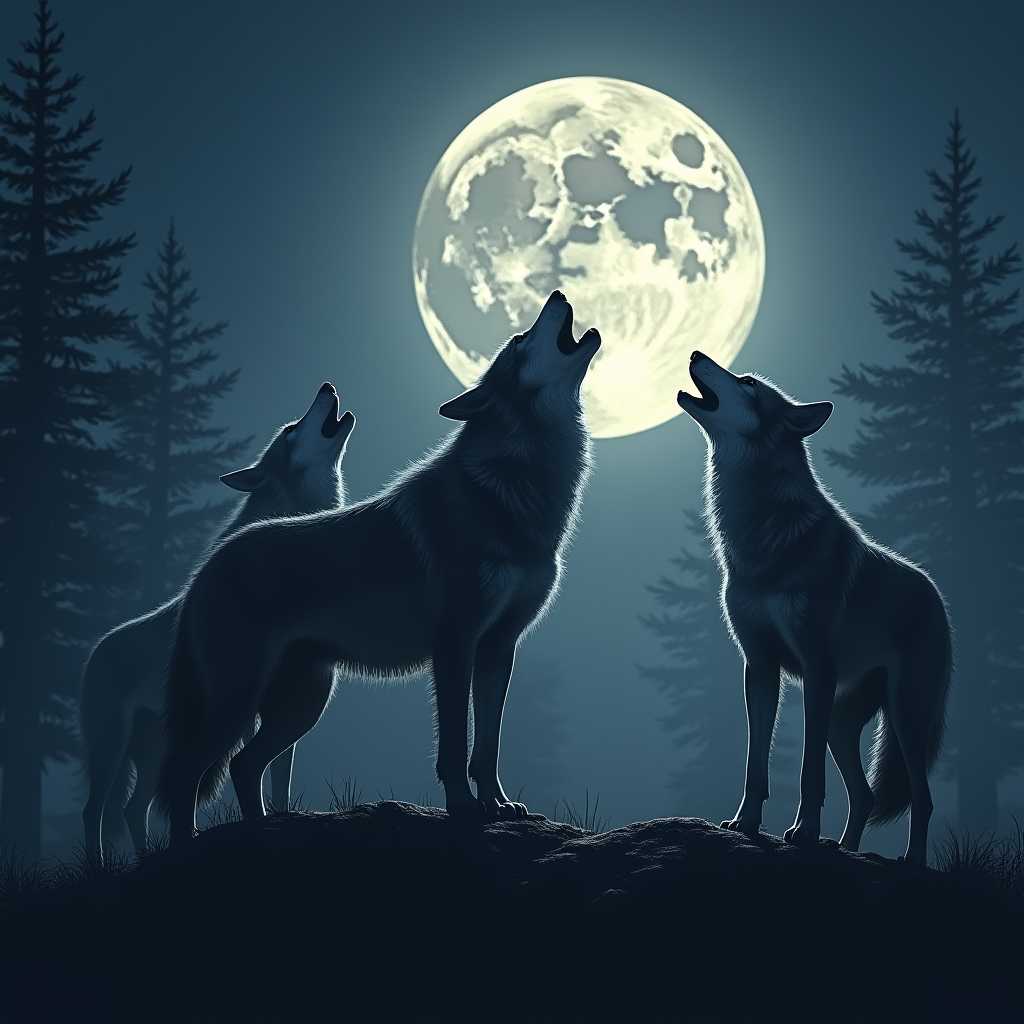} &
            \includegraphics[width=\imgwidth, height=\imgwidth]{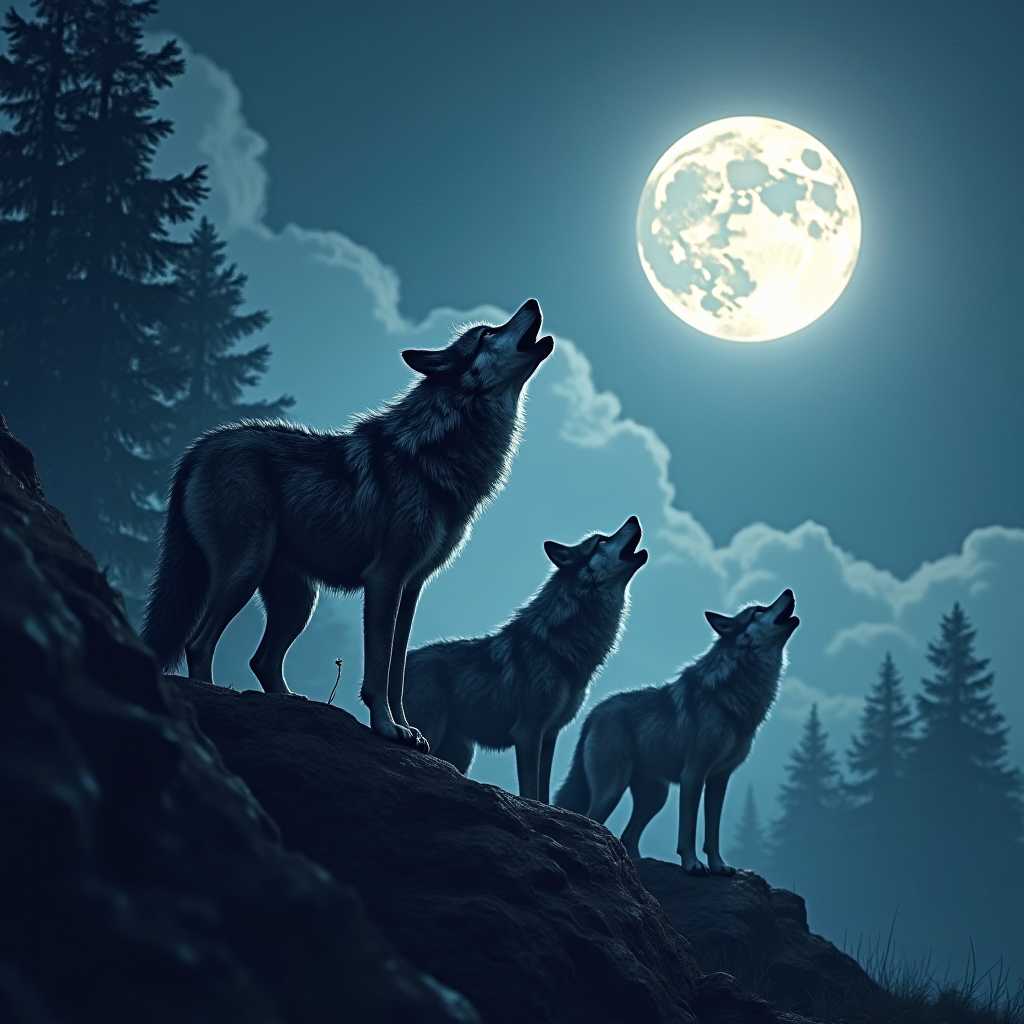} &
            \includegraphics[width=\imgwidth, height=\imgwidth]{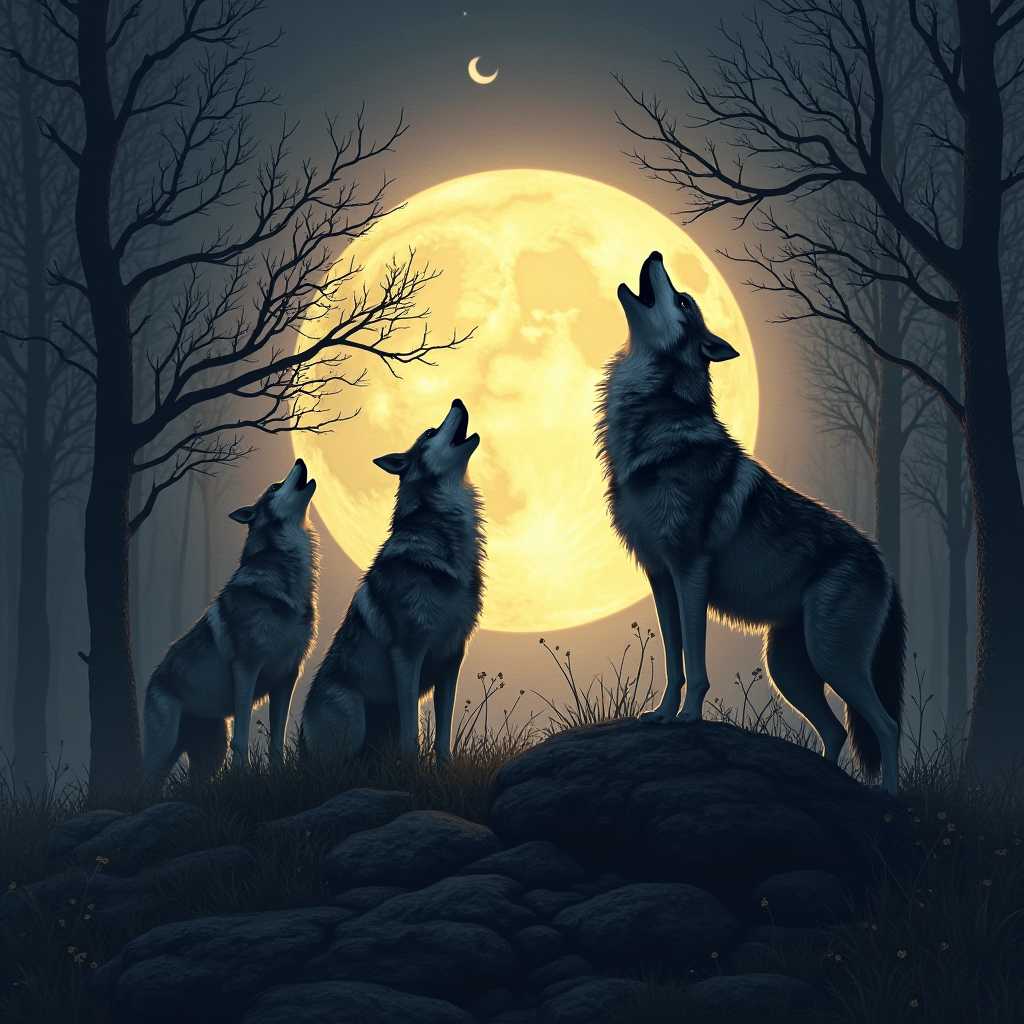} &
            \includegraphics[width=\imgwidth, height=\imgwidth]{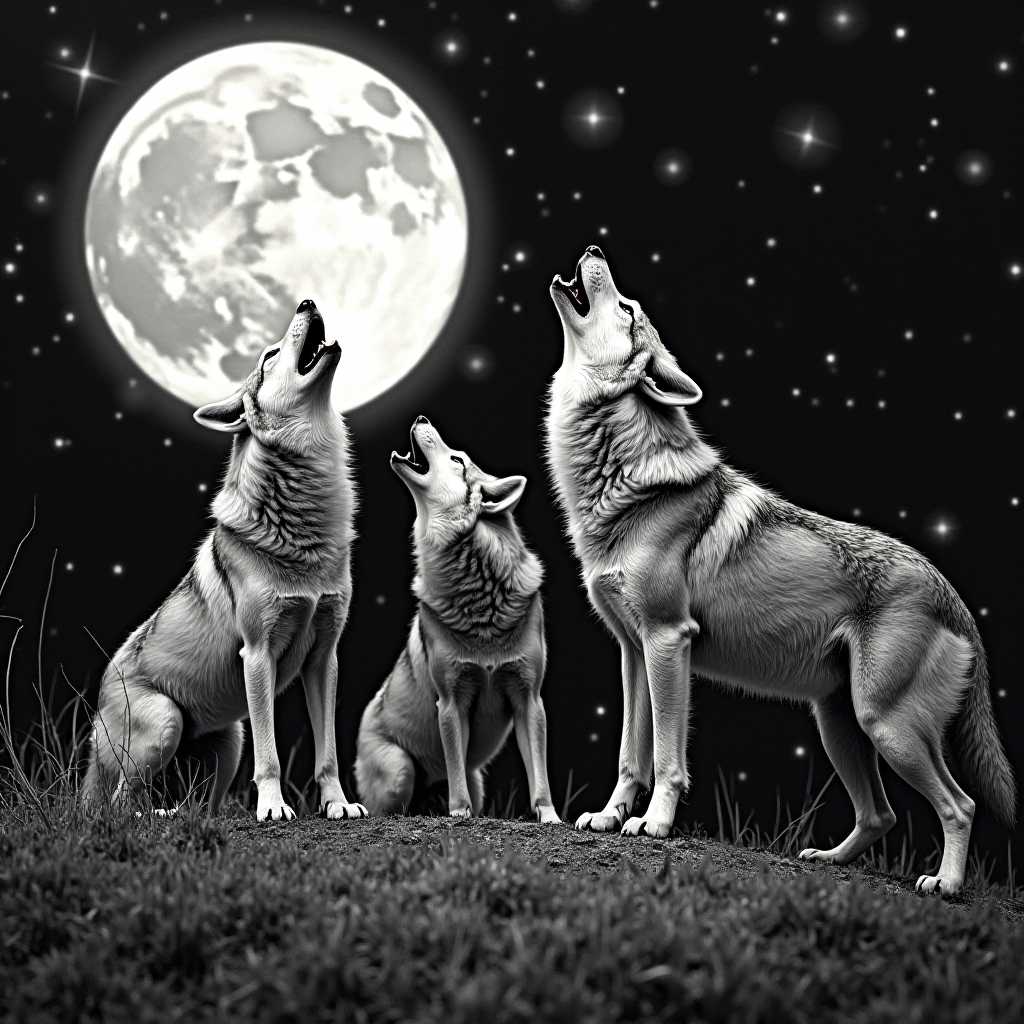} \\

            \vertlabel{PG}{2.5em} & %
            \includegraphics[width=\imgwidth, height=\imgwidth]{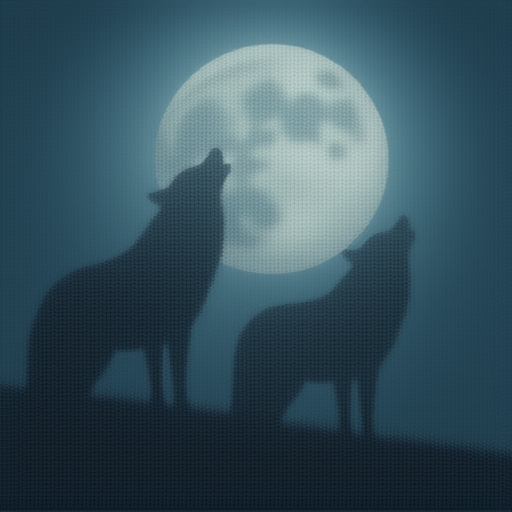} &
            \includegraphics[width=\imgwidth, height=\imgwidth]{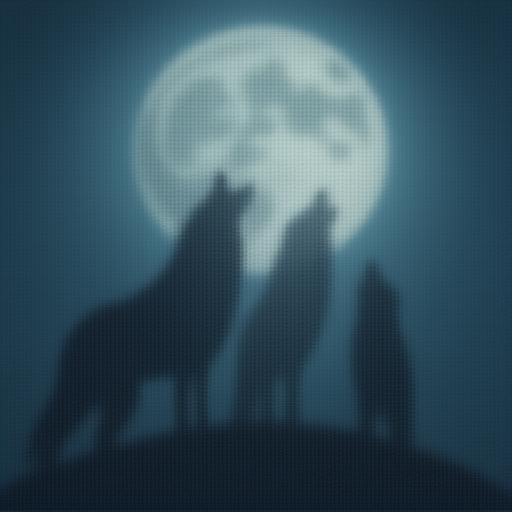} &
            \includegraphics[width=\imgwidth, height=\imgwidth]{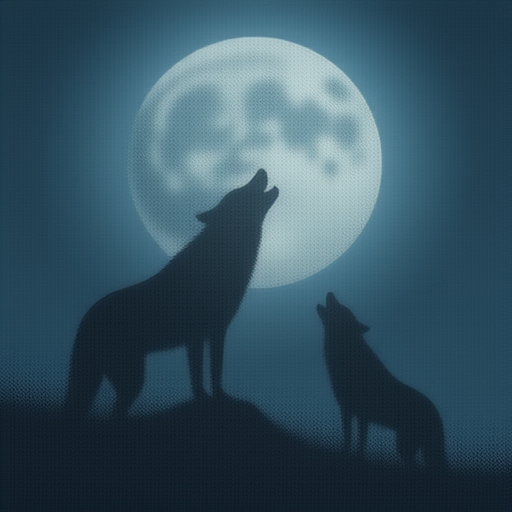} &
            \includegraphics[width=\imgwidth, height=\imgwidth]{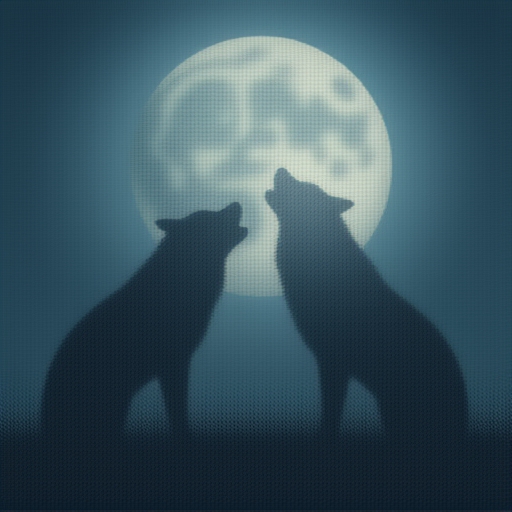} \\

            \multicolumn{5}{c}{\vspace{2pt}\small ``A wolf pack howling at the moon''} \\

            \vertlabel{Ours}{2.5em} & 
            \includegraphics[width=\imgwidth, height=\imgwidth]{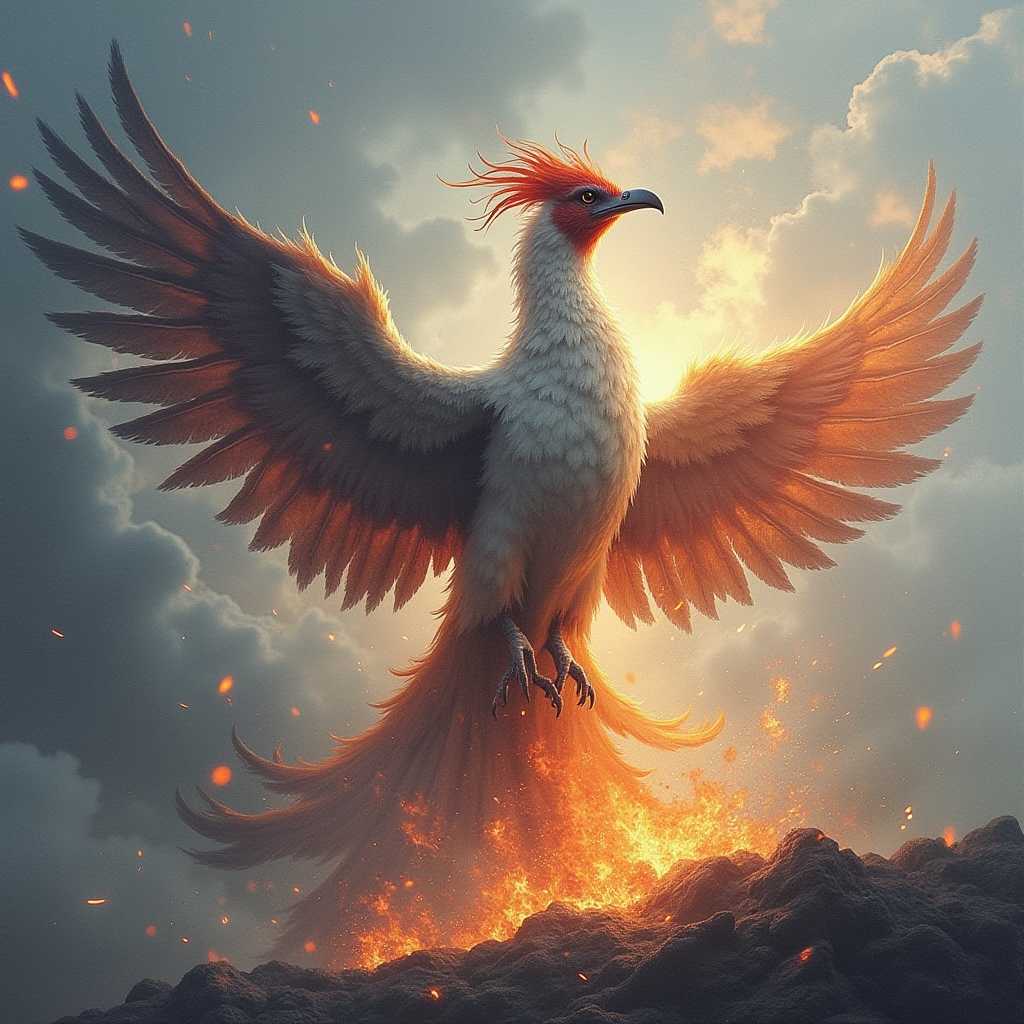} &
            \includegraphics[width=\imgwidth, height=\imgwidth]{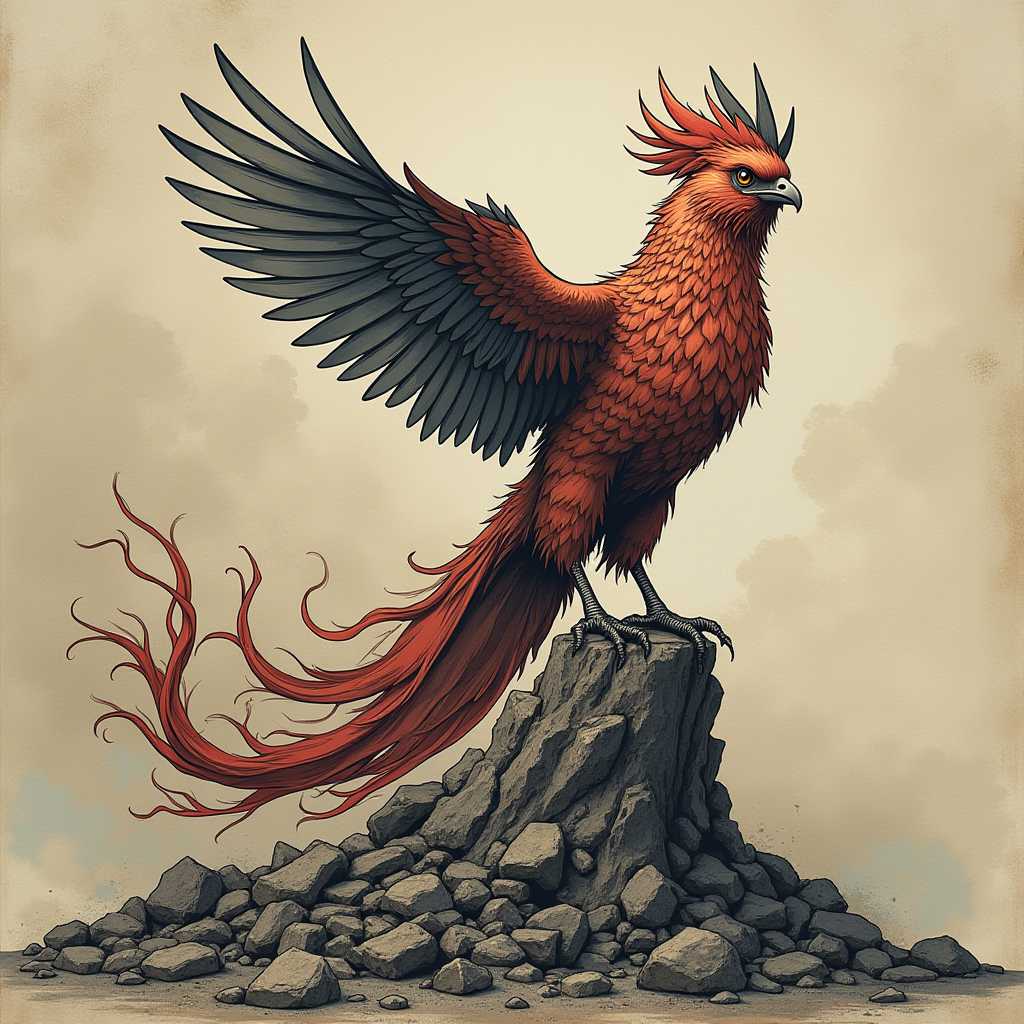} &
            \includegraphics[width=\imgwidth, height=\imgwidth]{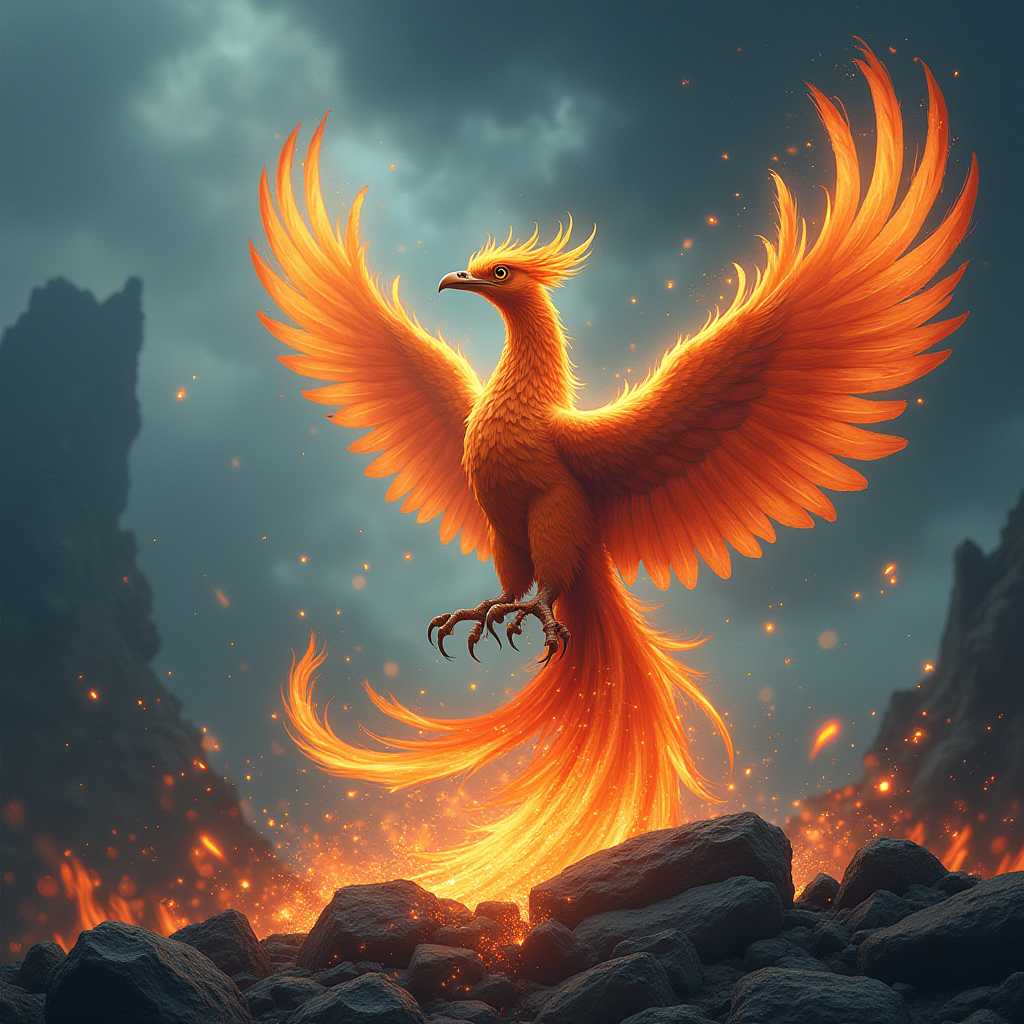} &
            \includegraphics[width=\imgwidth, height=\imgwidth]{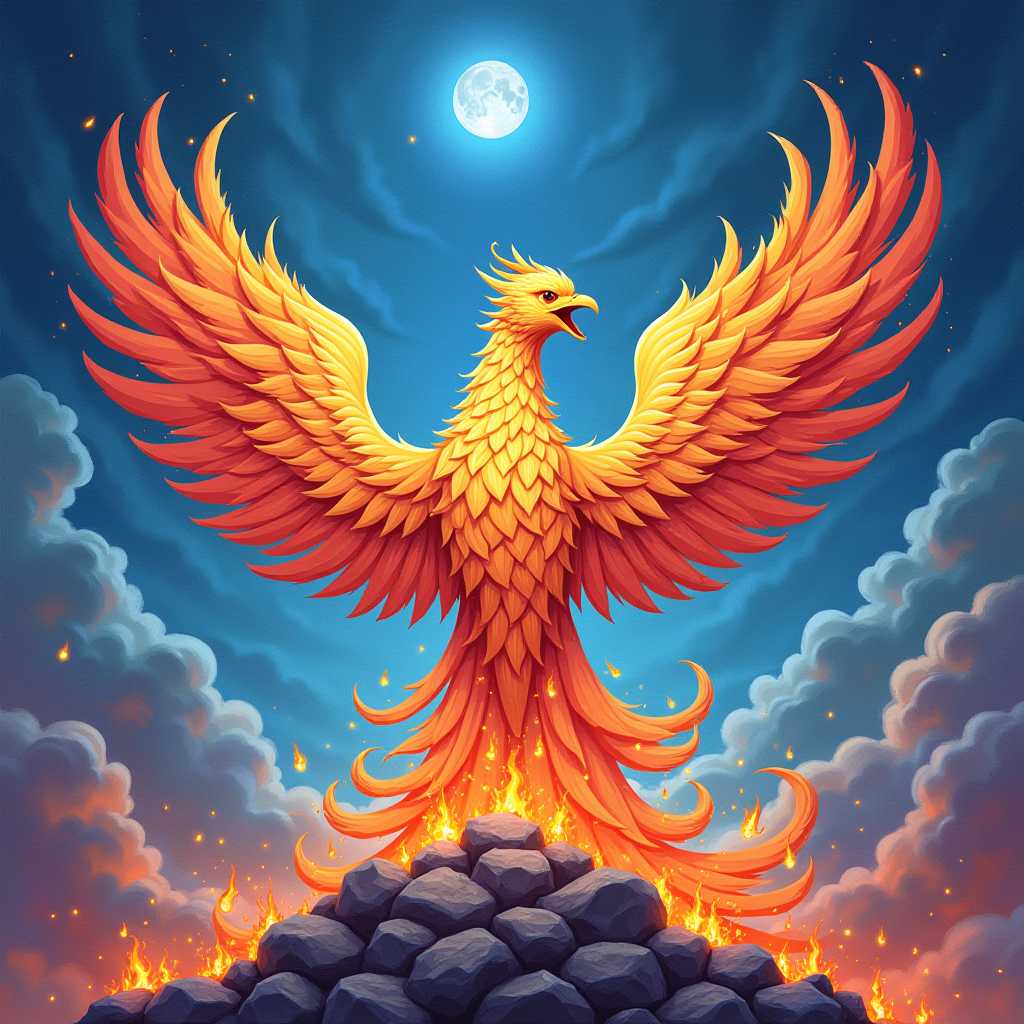} \\

            \vertlabel{SGI}{2.5em} & 
            \includegraphics[width=\imgwidth, height=\imgwidth]{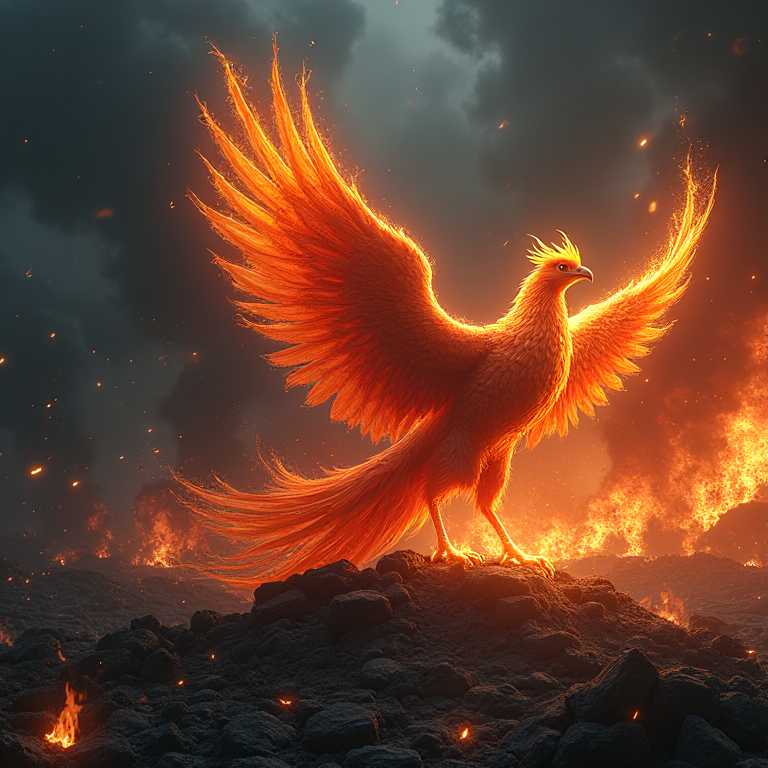} &
            \includegraphics[width=\imgwidth, height=\imgwidth]{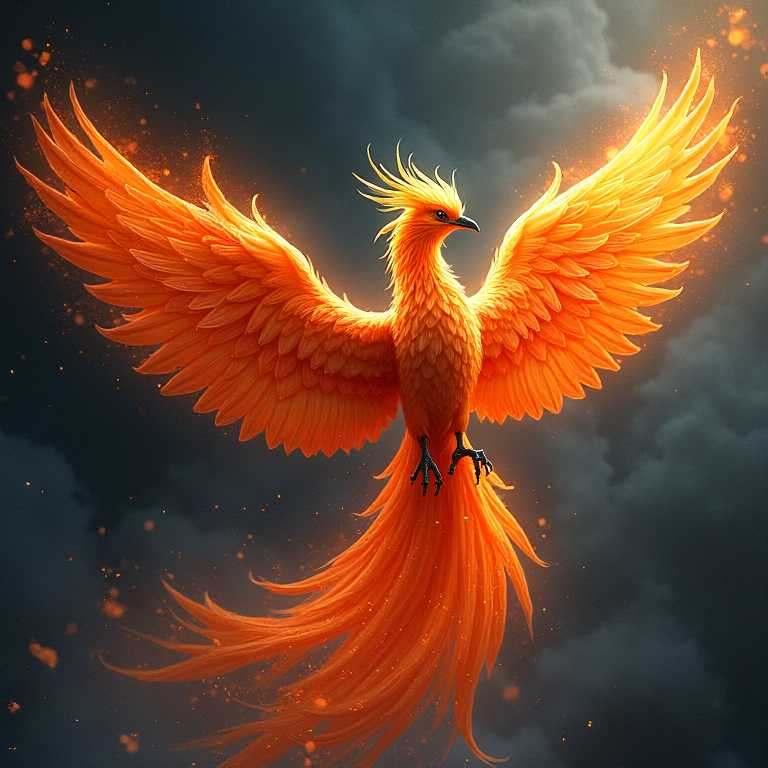} &
            \includegraphics[width=\imgwidth, height=\imgwidth]{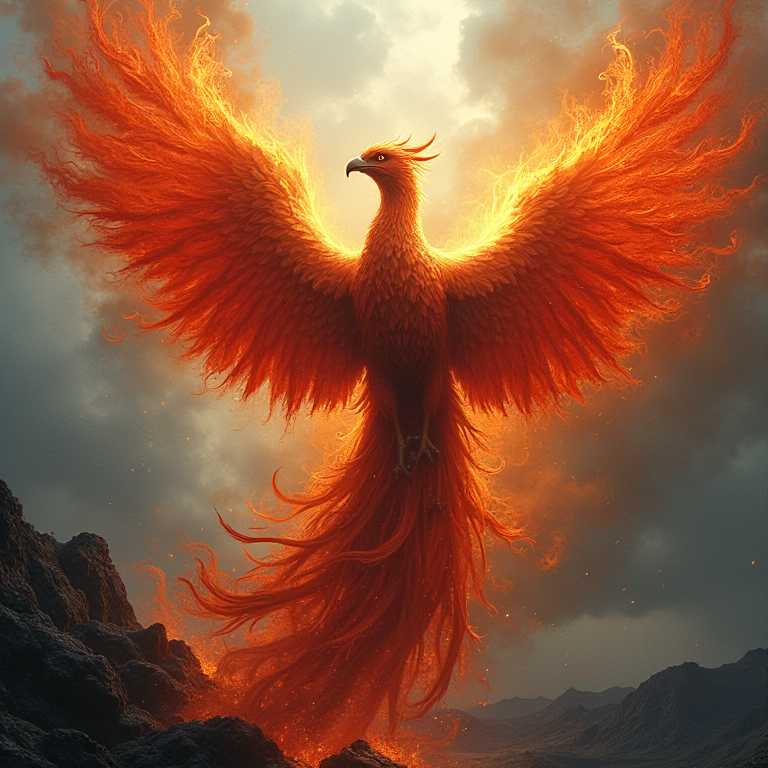} &
            \includegraphics[width=\imgwidth, height=\imgwidth]{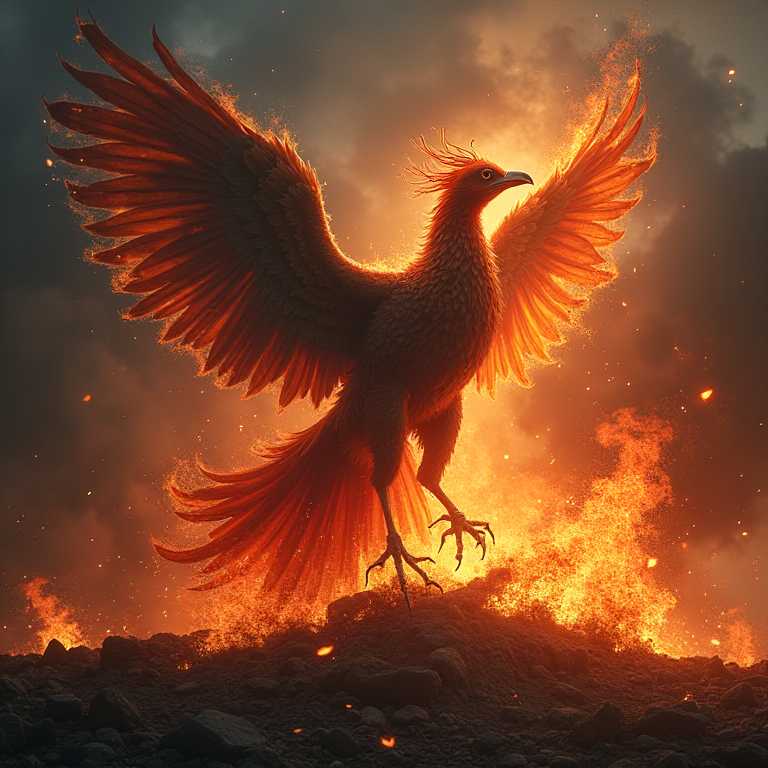} \\
            
            \vertlabel{CADS}{2.5em} & 
            \includegraphics[width=\imgwidth, height=\imgwidth]{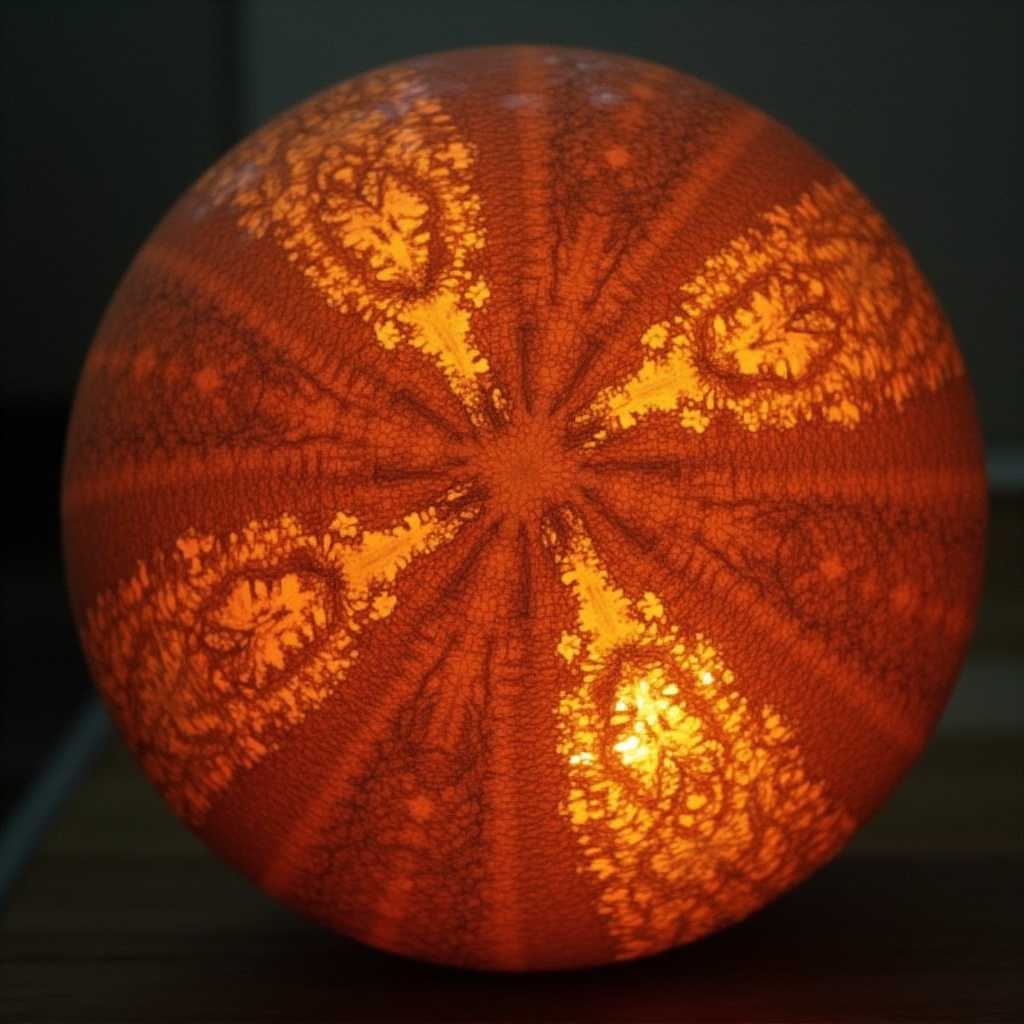} &
            \includegraphics[width=\imgwidth, height=\imgwidth]{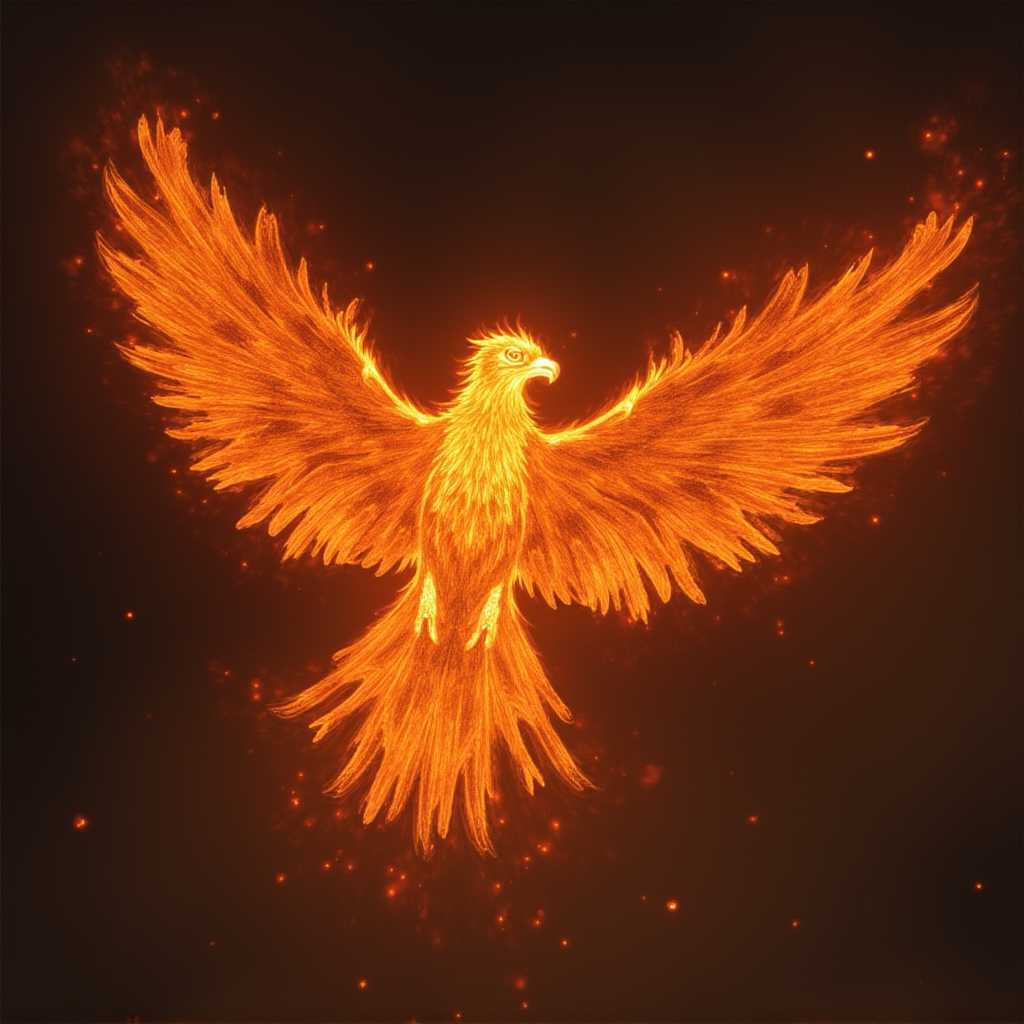} &
            \includegraphics[width=\imgwidth, height=\imgwidth]{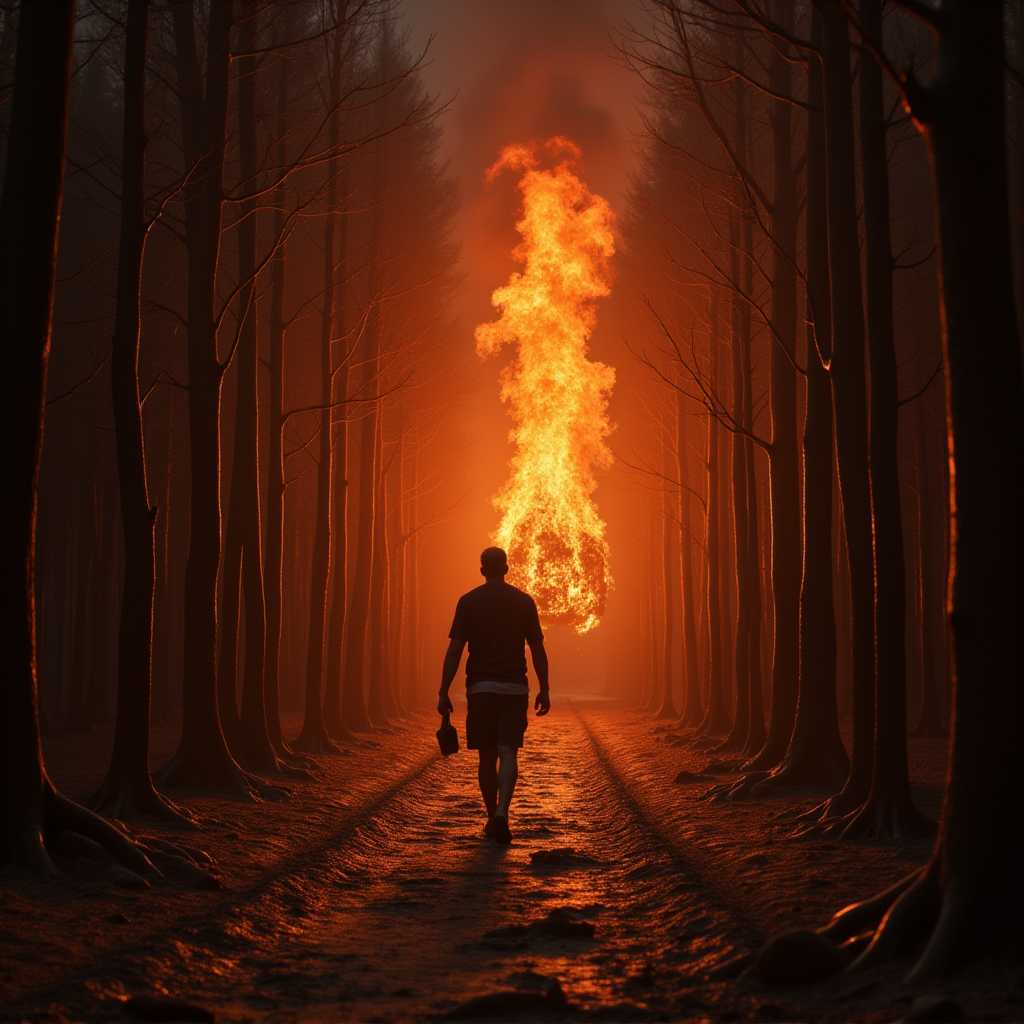} &
            \includegraphics[width=\imgwidth, height=\imgwidth]{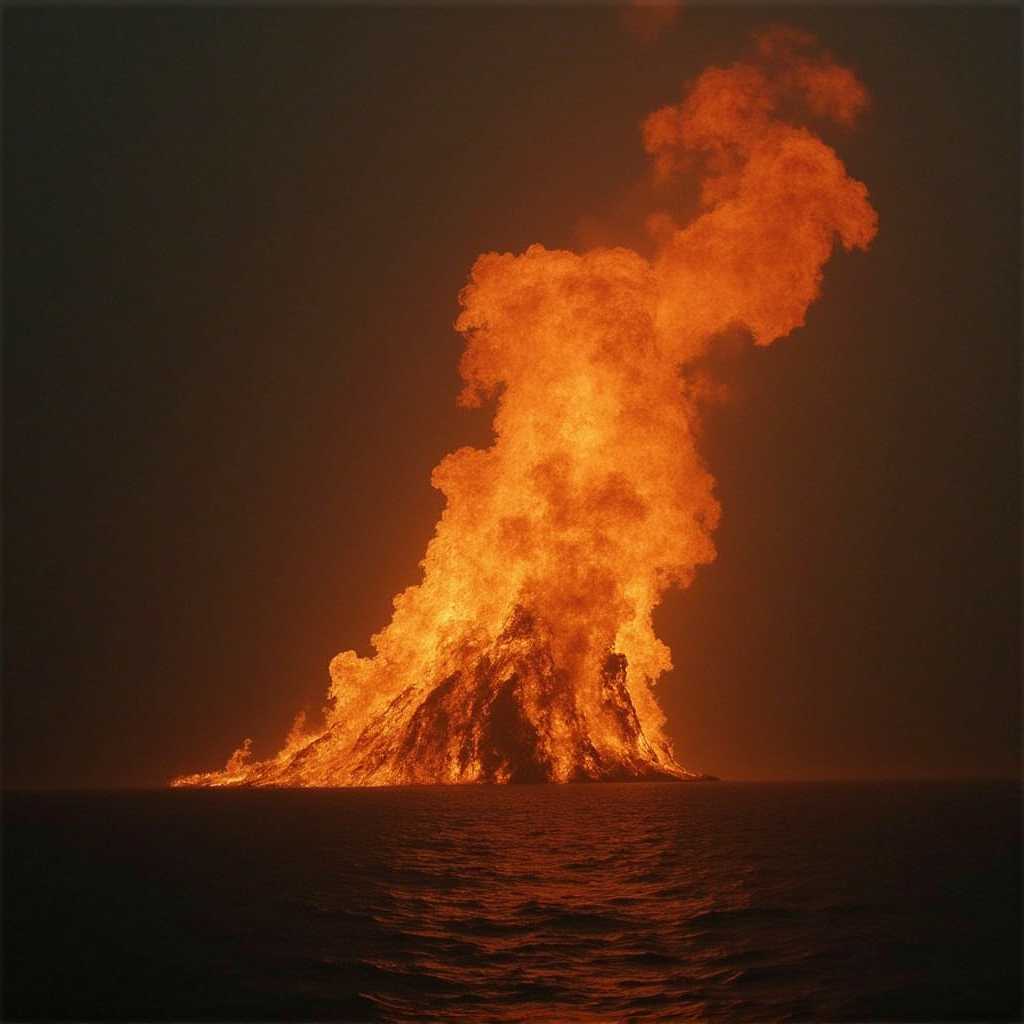} \\

            \vertlabel{SPARKE}{2.5em} & %
            
            \includegraphics[width=\imgwidth, height=\imgwidth]{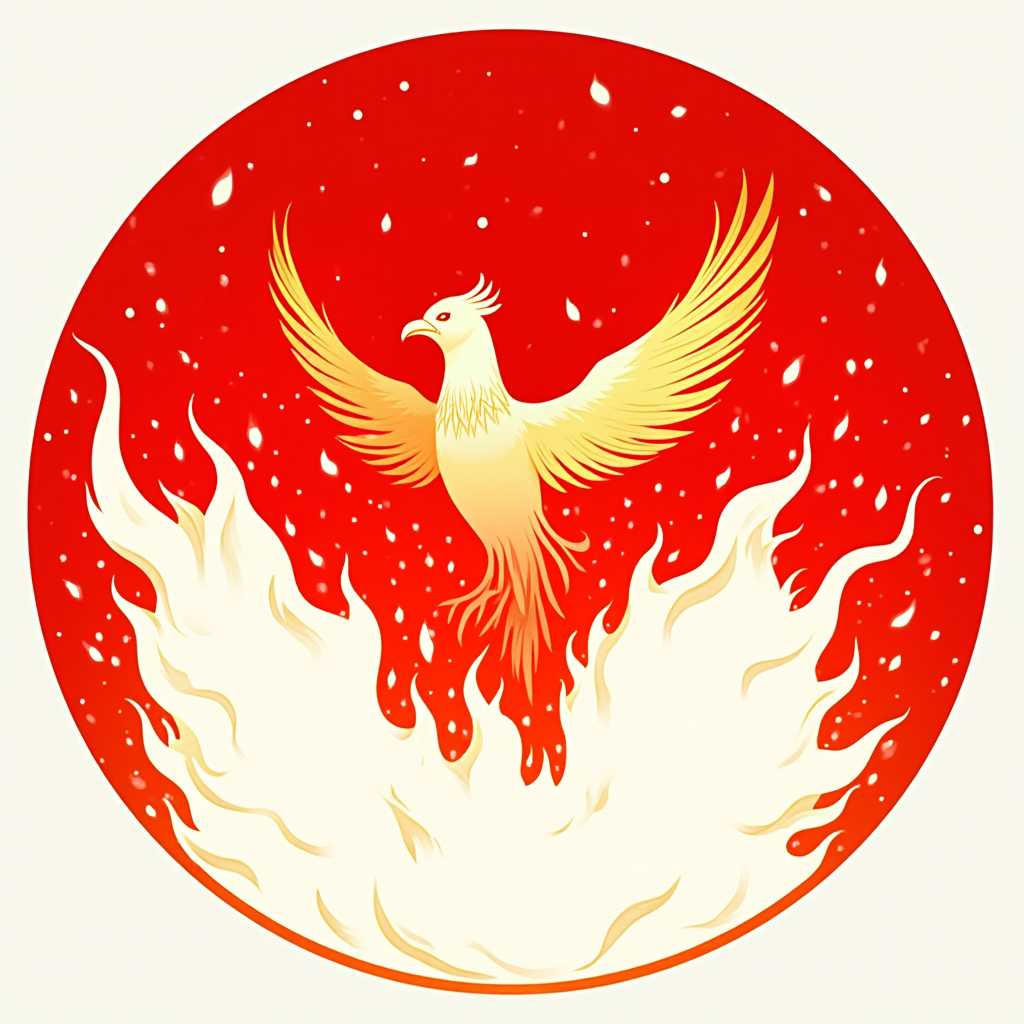} &
            \includegraphics[width=\imgwidth, height=\imgwidth]{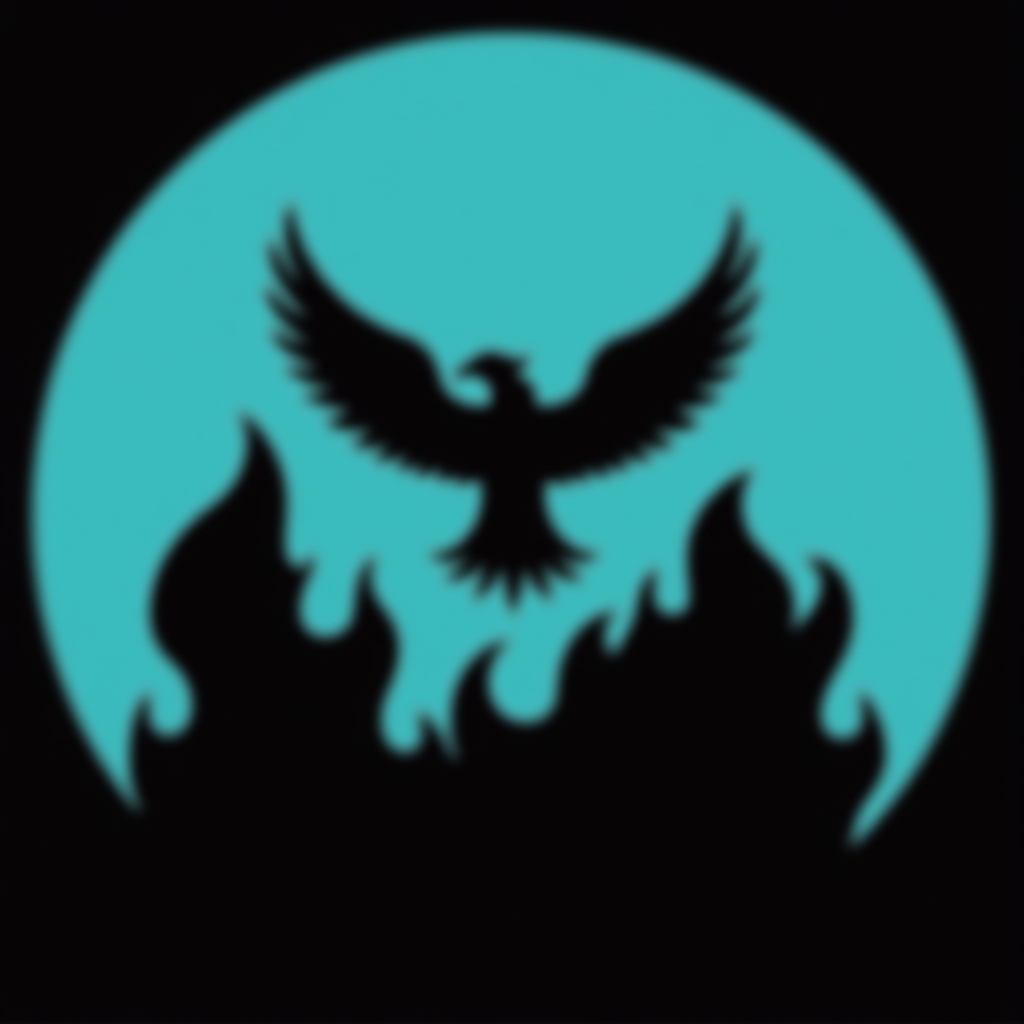} &
            \includegraphics[width=\imgwidth, height=\imgwidth]{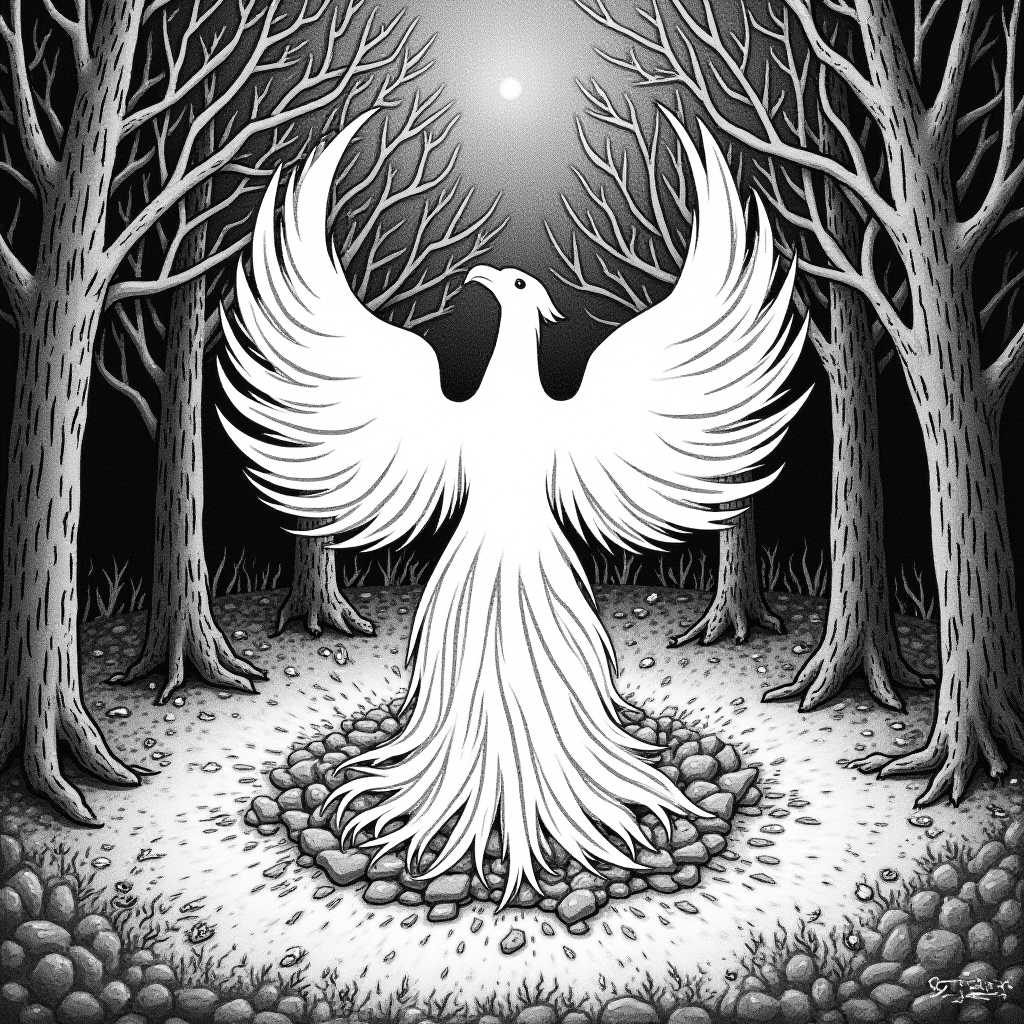} &
            \includegraphics[width=\imgwidth, height=\imgwidth]{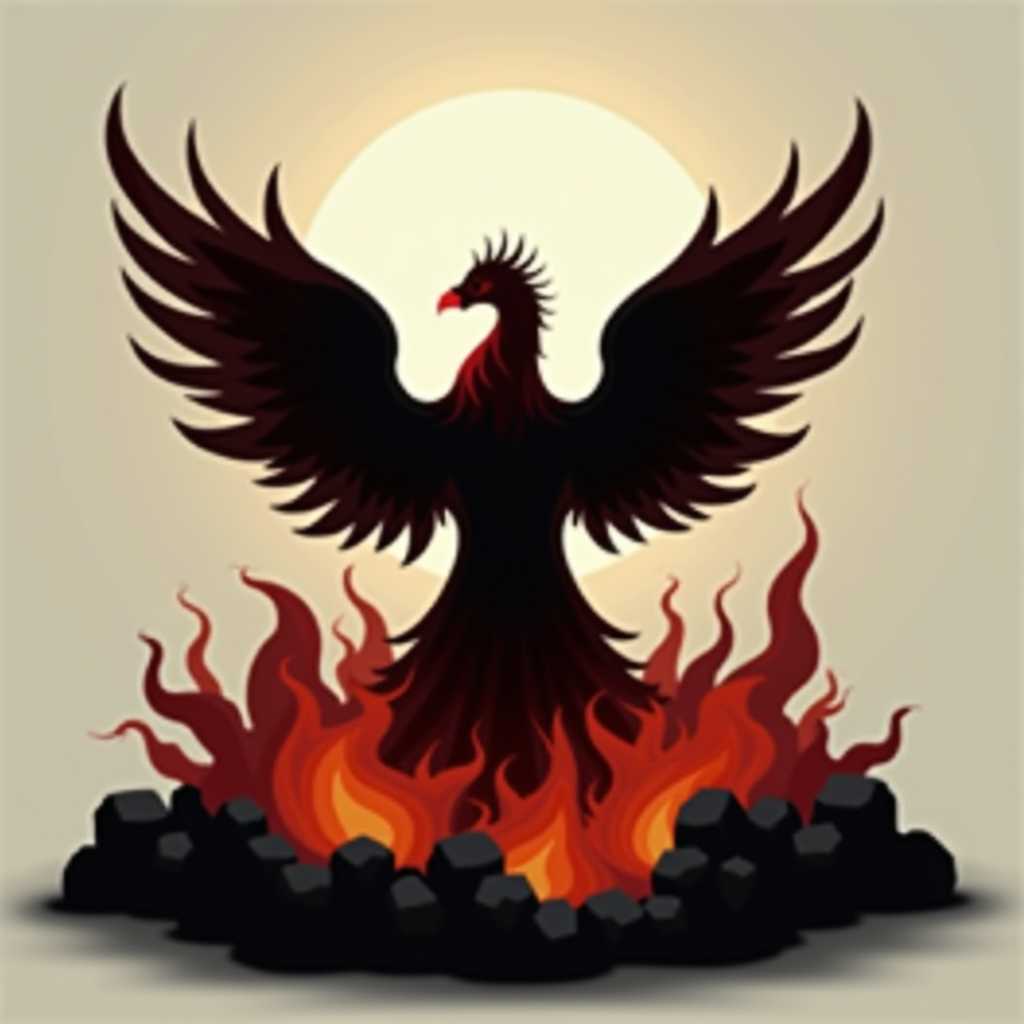} \\
            
            \vertlabel{PG}{2.5em} & %
            \includegraphics[width=\imgwidth, height=\imgwidth]{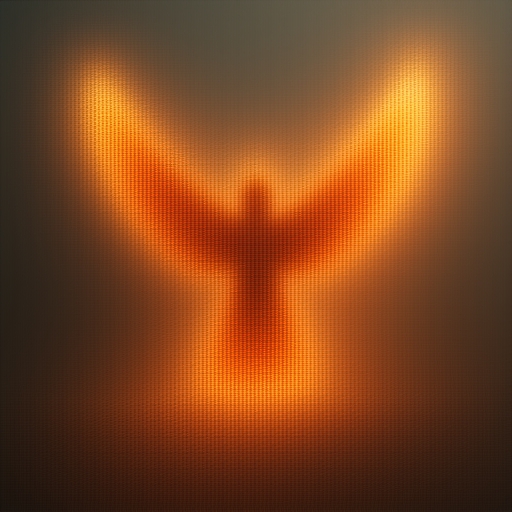} &
            \includegraphics[width=\imgwidth, height=\imgwidth]{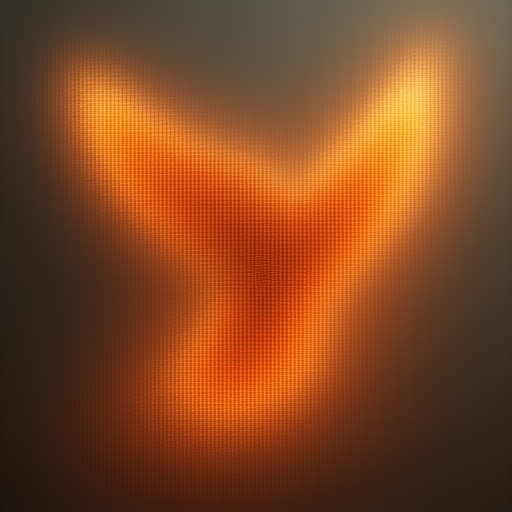} &
            \includegraphics[width=\imgwidth, height=\imgwidth]{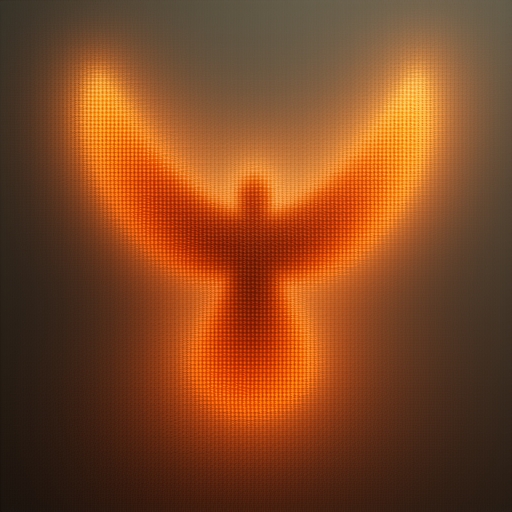} &
            \includegraphics[width=\imgwidth, height=\imgwidth]{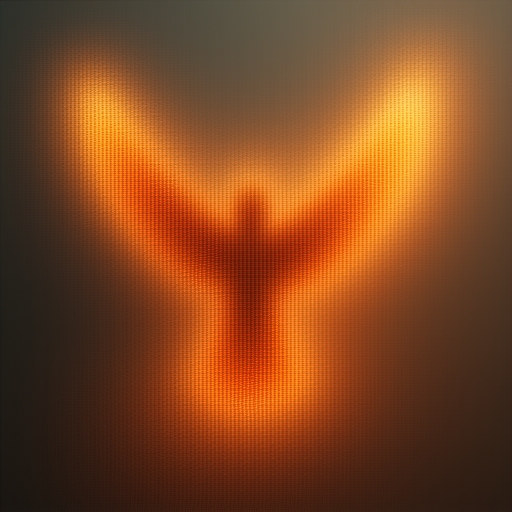} \\

            \multicolumn{5}{c}{\vspace{2pt}\small ``A phoenix rising from ashes''}
        \end{tabular}
    \end{minipage}
    \hfill
    \begin{minipage}{0.48\textwidth}
        \centering
        \begin{tabular}{c c c c c}
            \vertlabel{Ours}{2.5em} & 
            \includegraphics[width=\imgwidth, height=\imgwidth]{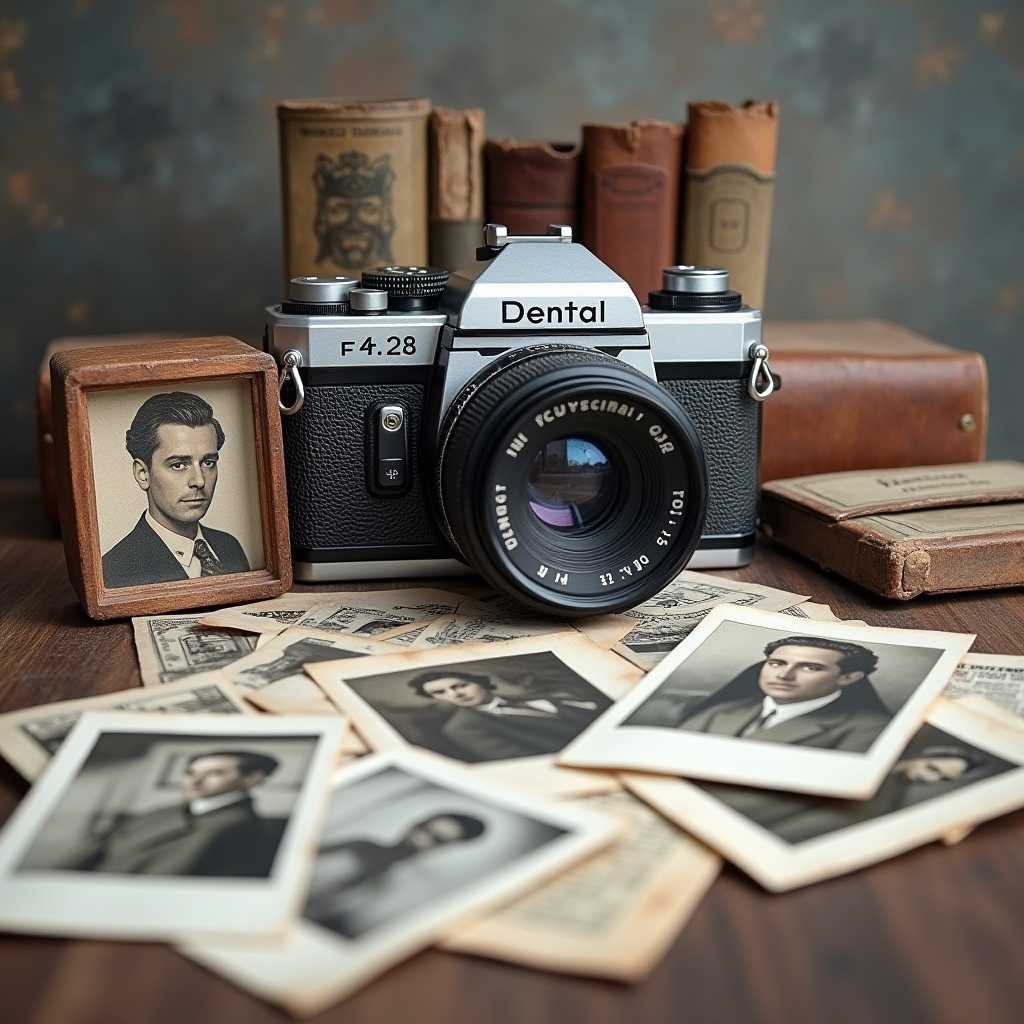} &
            \includegraphics[width=\imgwidth, height=\imgwidth]{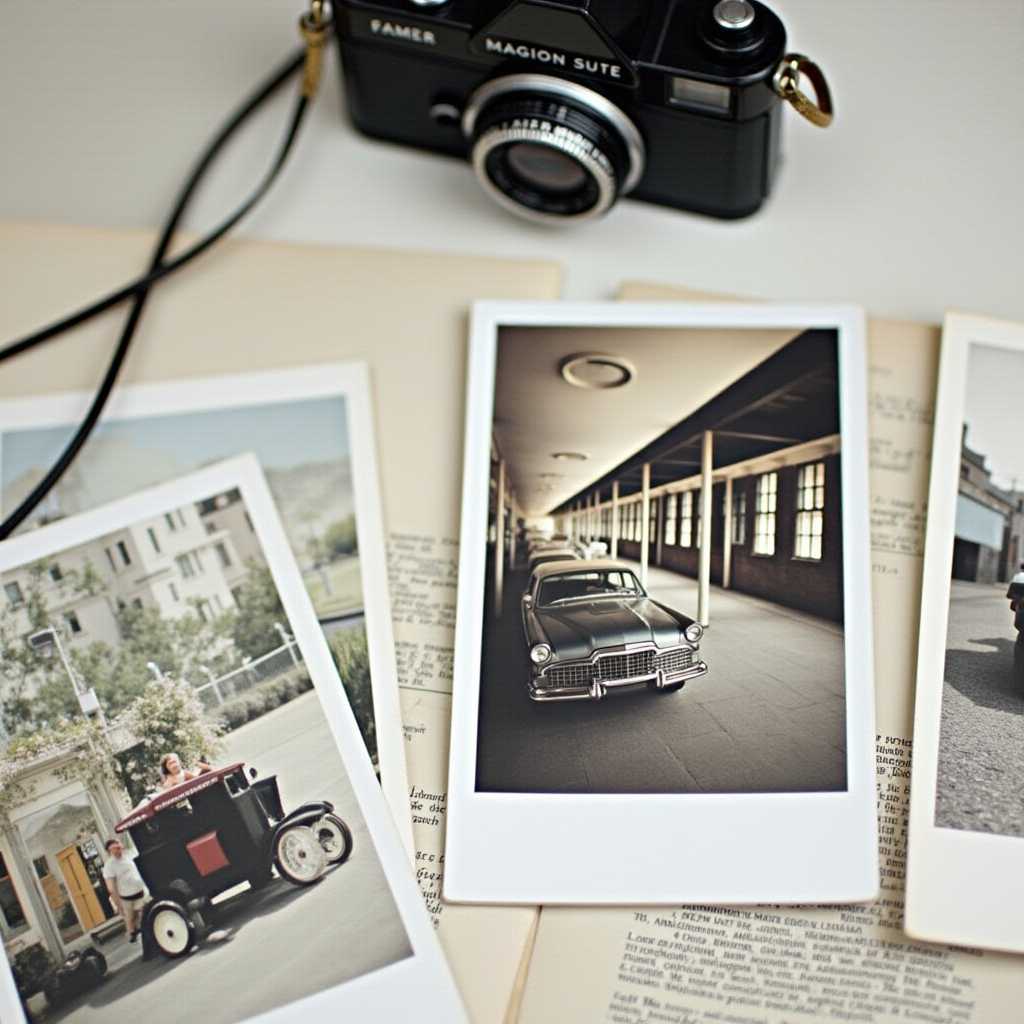} &
            \includegraphics[width=\imgwidth, height=\imgwidth]{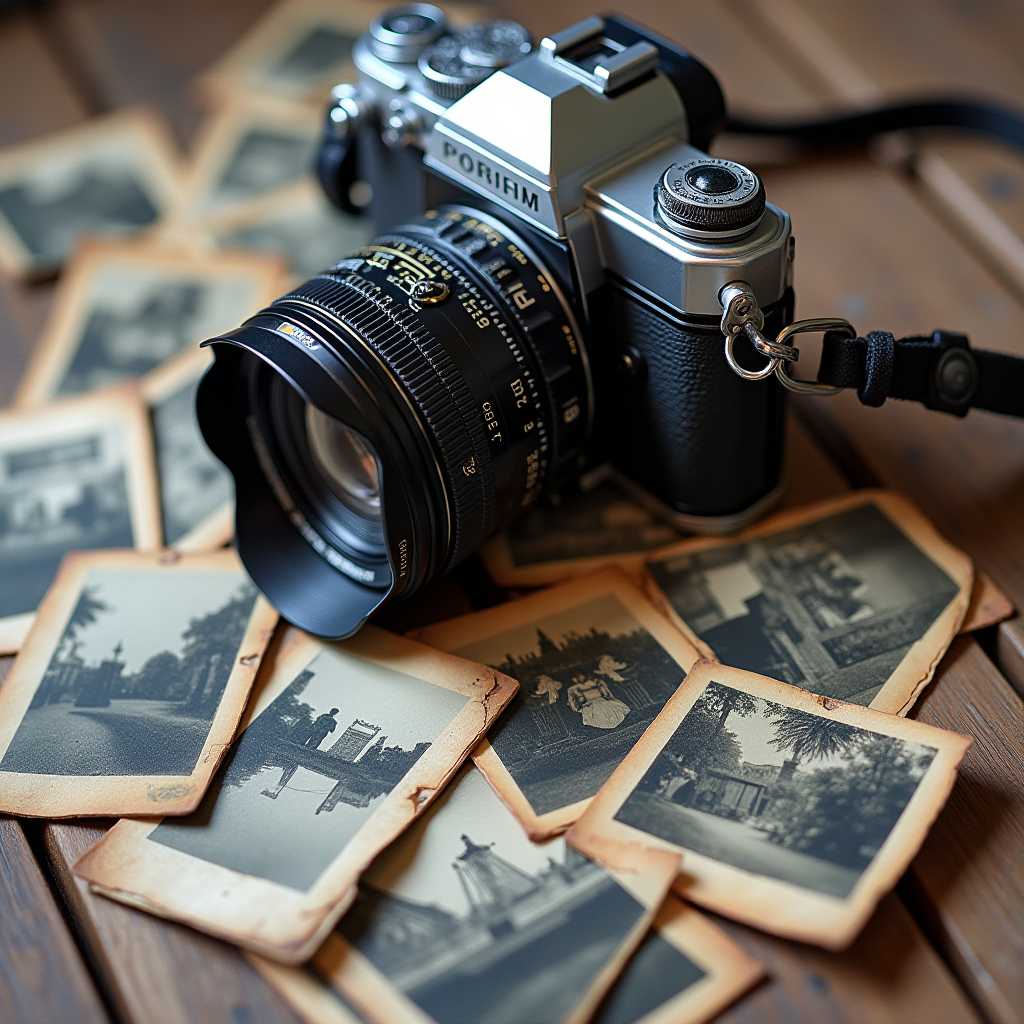} &
            \includegraphics[width=\imgwidth, height=\imgwidth]{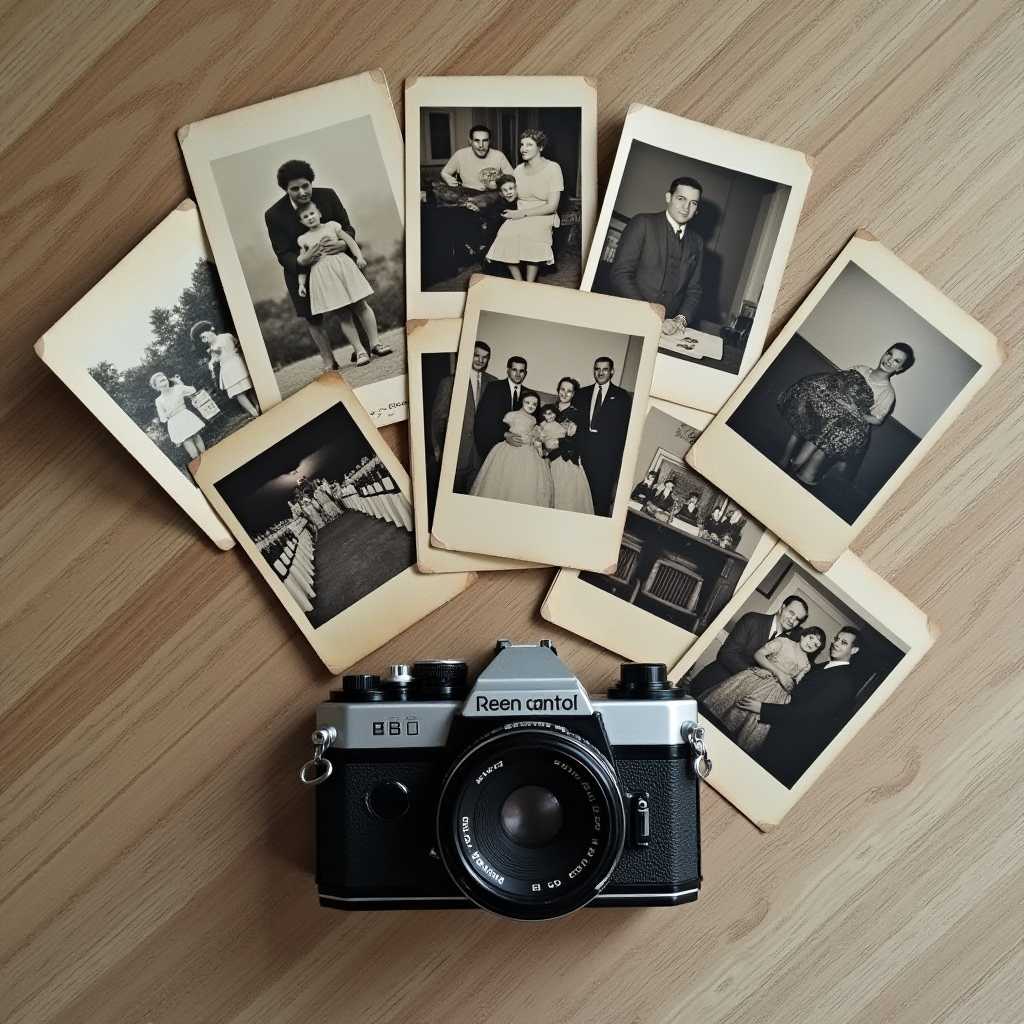} \\

            \vertlabel{SGI}{2.5em} & 
            \includegraphics[width=\imgwidth, height=\imgwidth]{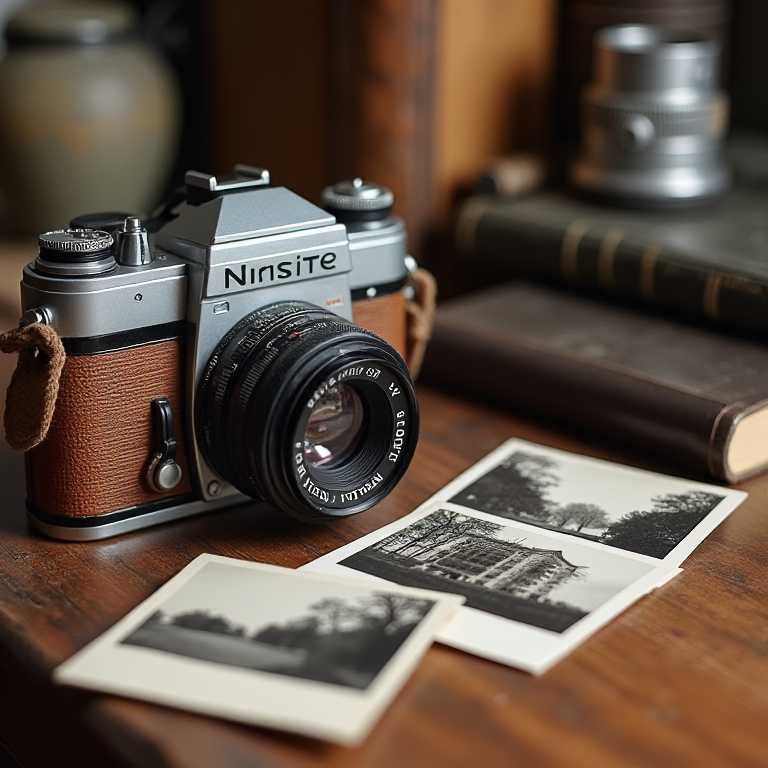} &
            \includegraphics[width=\imgwidth, height=\imgwidth]{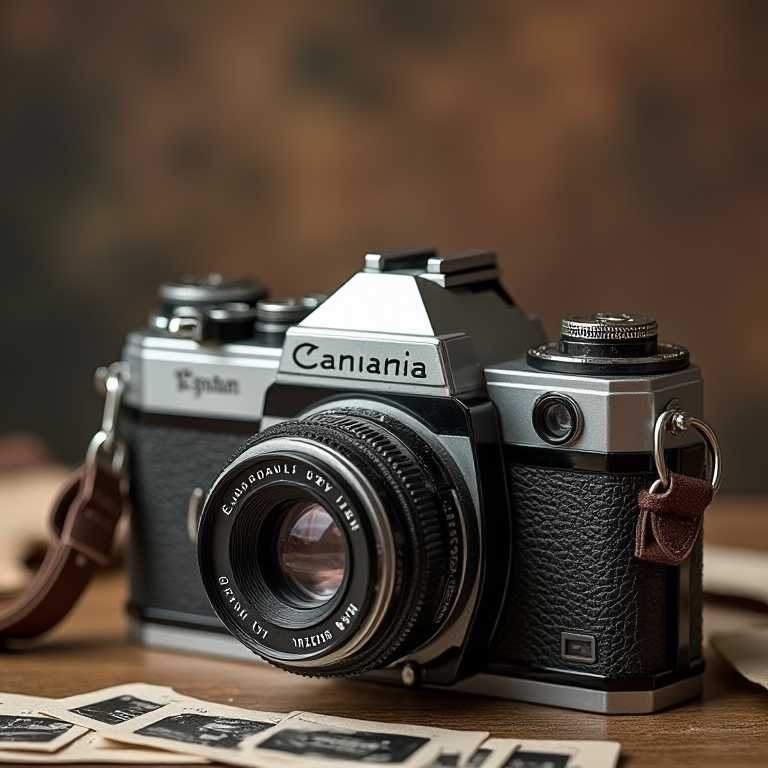} &
            \includegraphics[width=\imgwidth, height=\imgwidth]{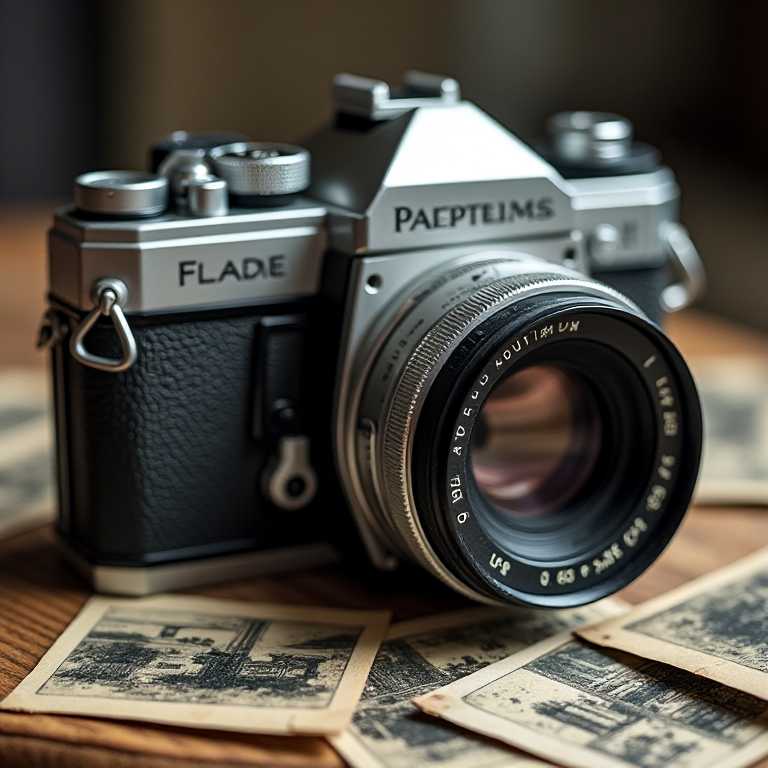} &
            \includegraphics[width=\imgwidth, height=\imgwidth]{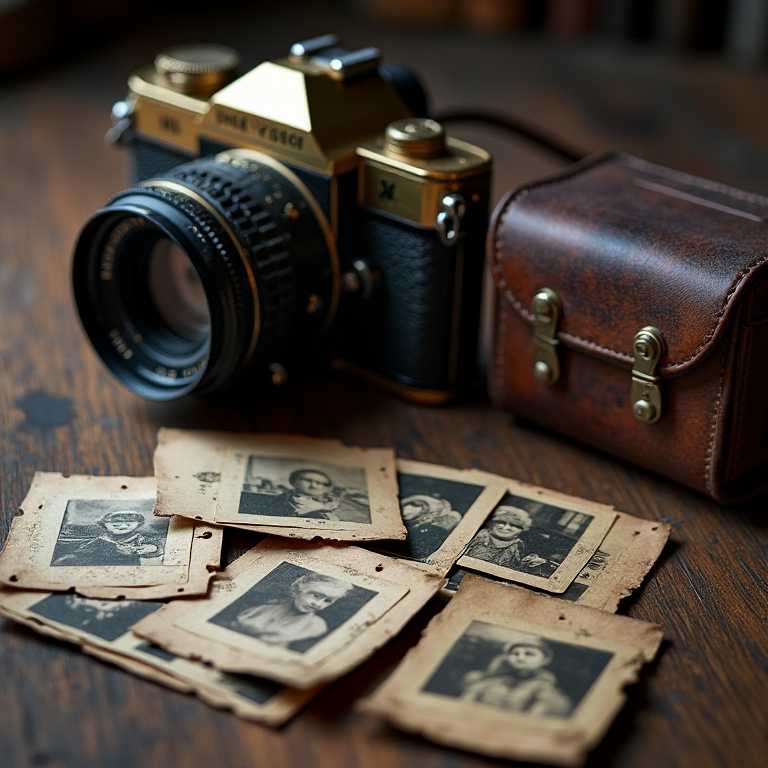} \\
            
            \vertlabel{CADS}{2.5em} & 
            \includegraphics[width=\imgwidth, height=\imgwidth]{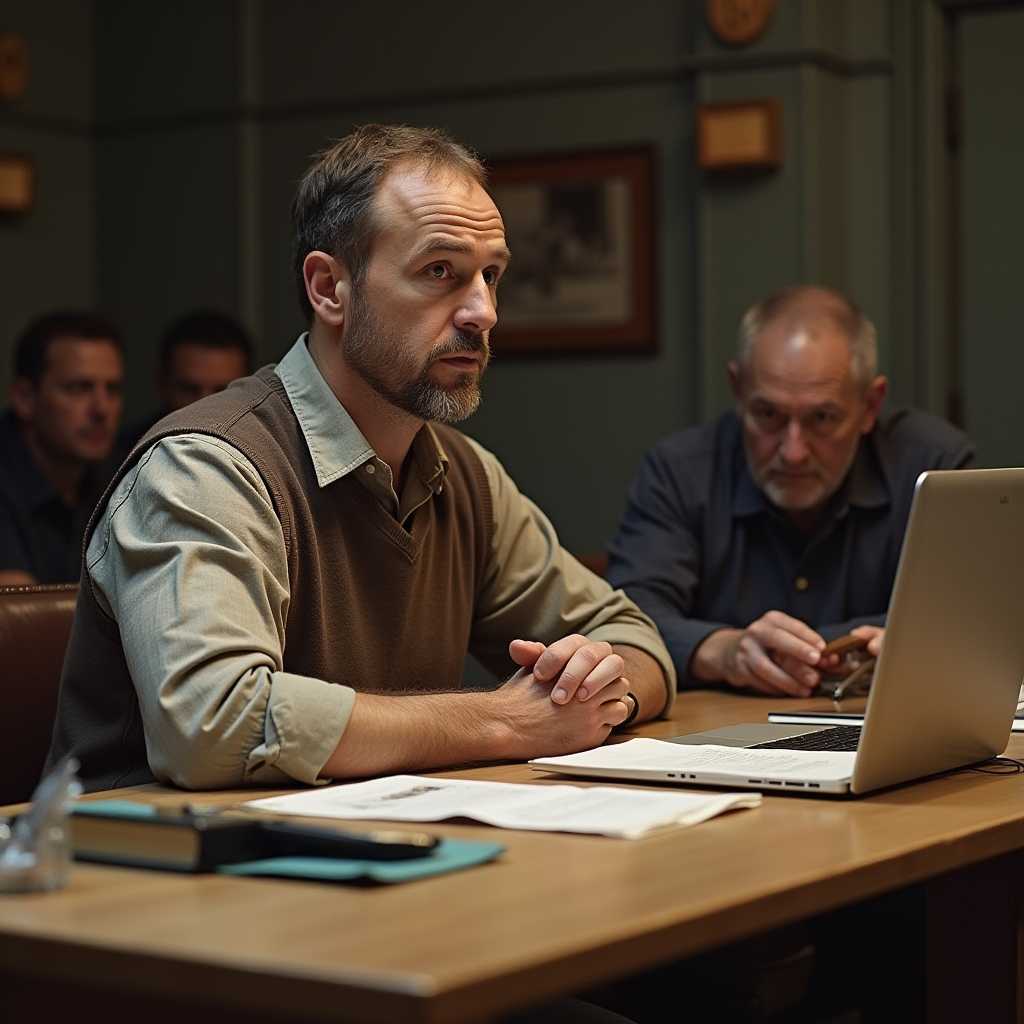} &
            \includegraphics[width=\imgwidth, height=\imgwidth]{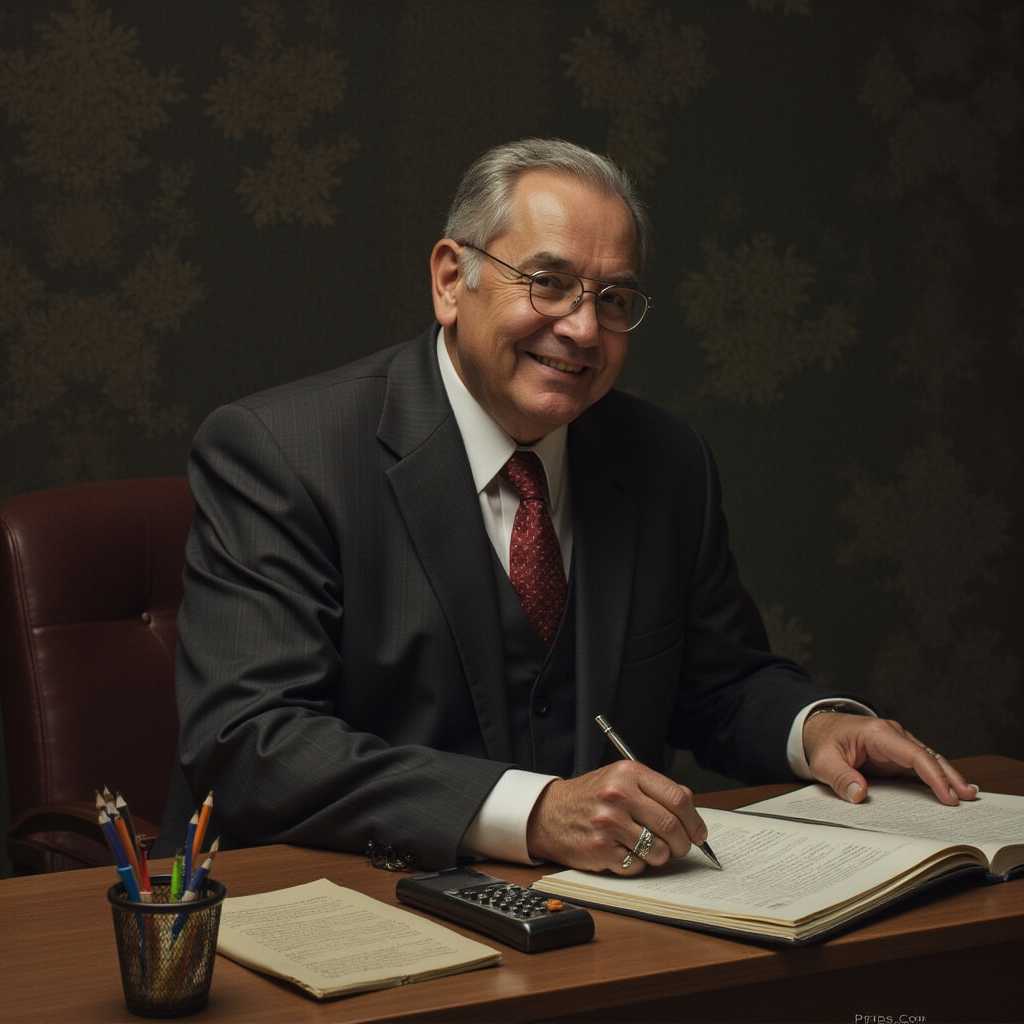} &
            \includegraphics[width=\imgwidth, height=\imgwidth]{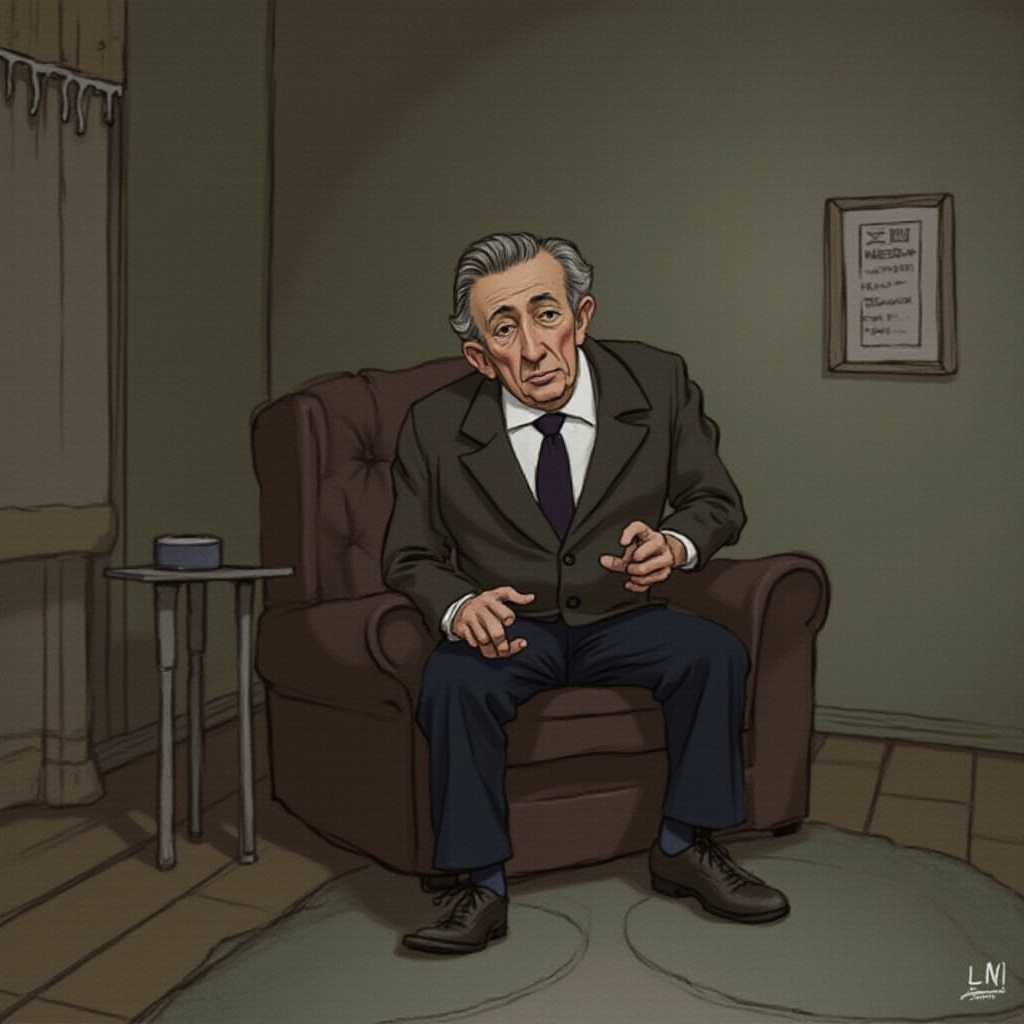} &
            \includegraphics[width=\imgwidth, height=\imgwidth]{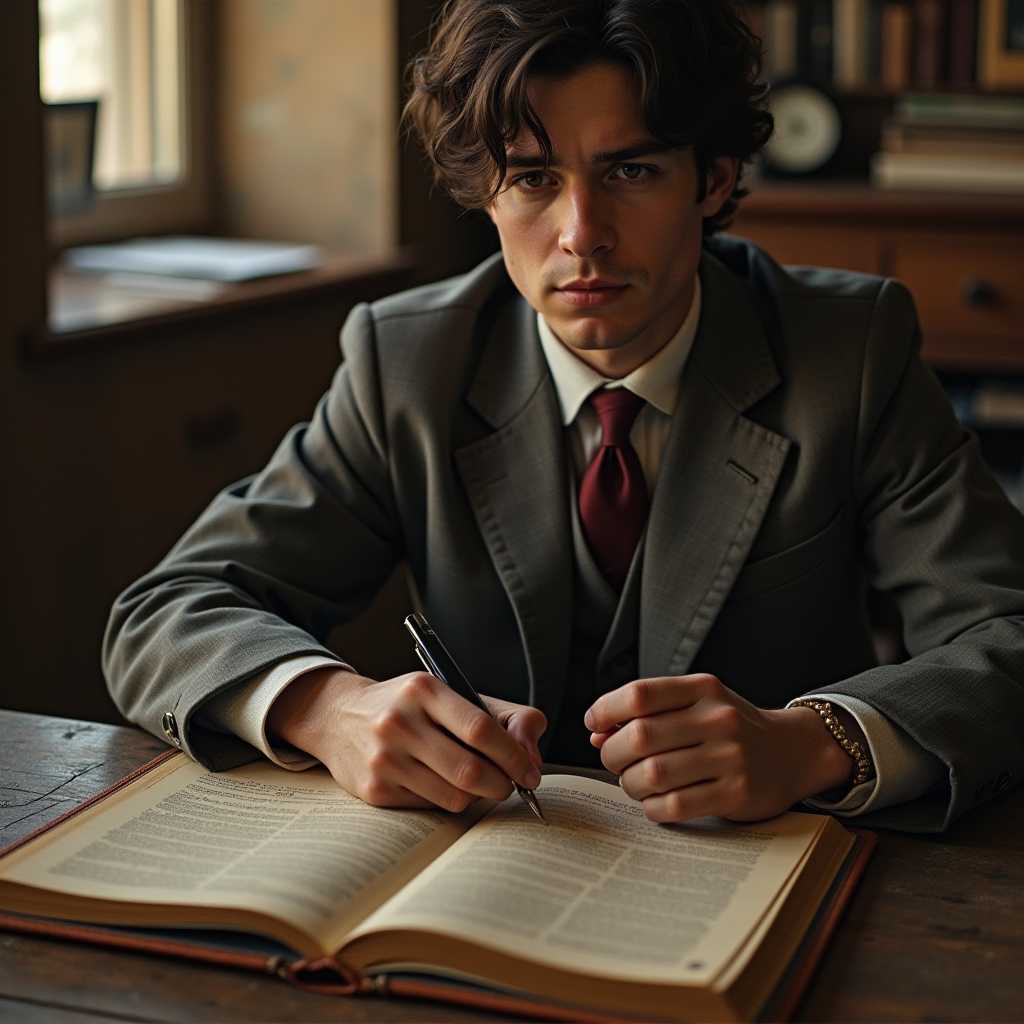} \\

                        \vertlabel{SPARKE}{2.5em} & %
            \includegraphics[width=\imgwidth, height=\imgwidth]{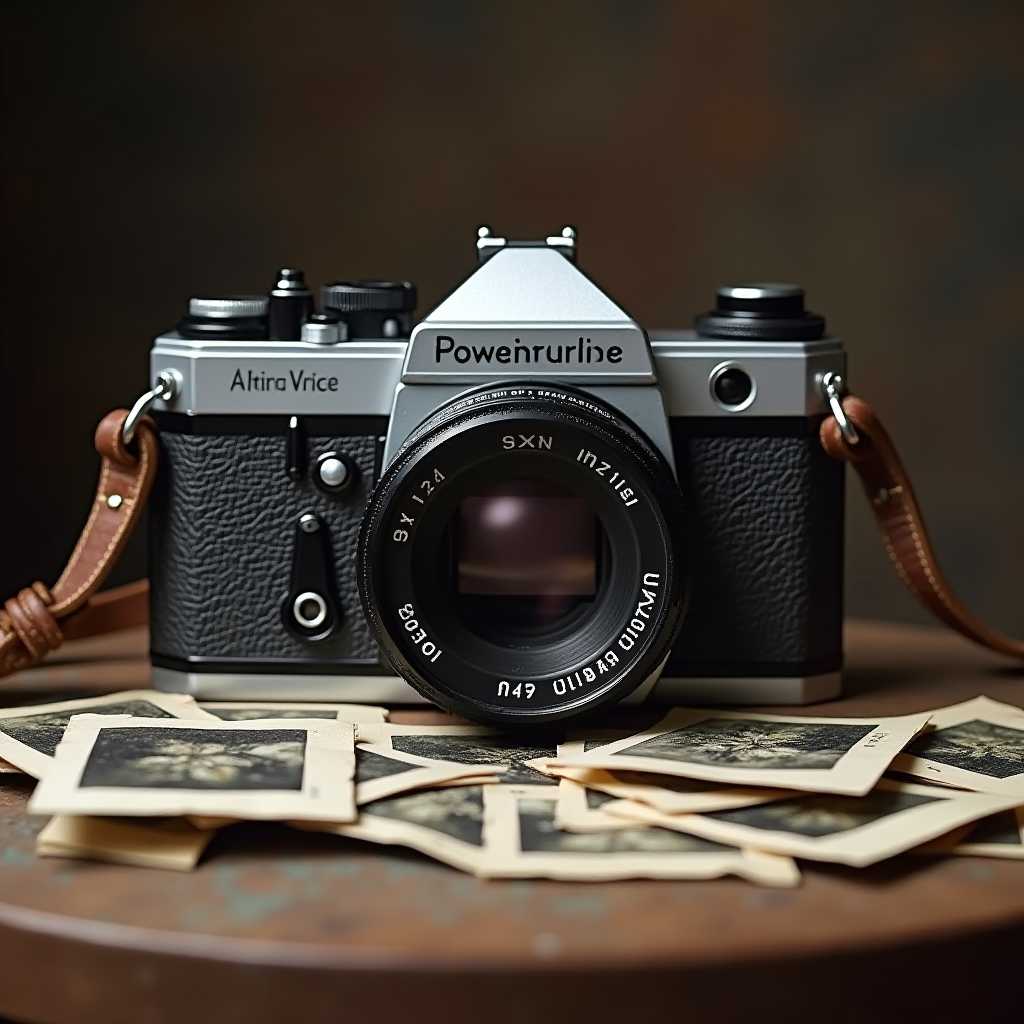} &
            \includegraphics[width=\imgwidth, height=\imgwidth]{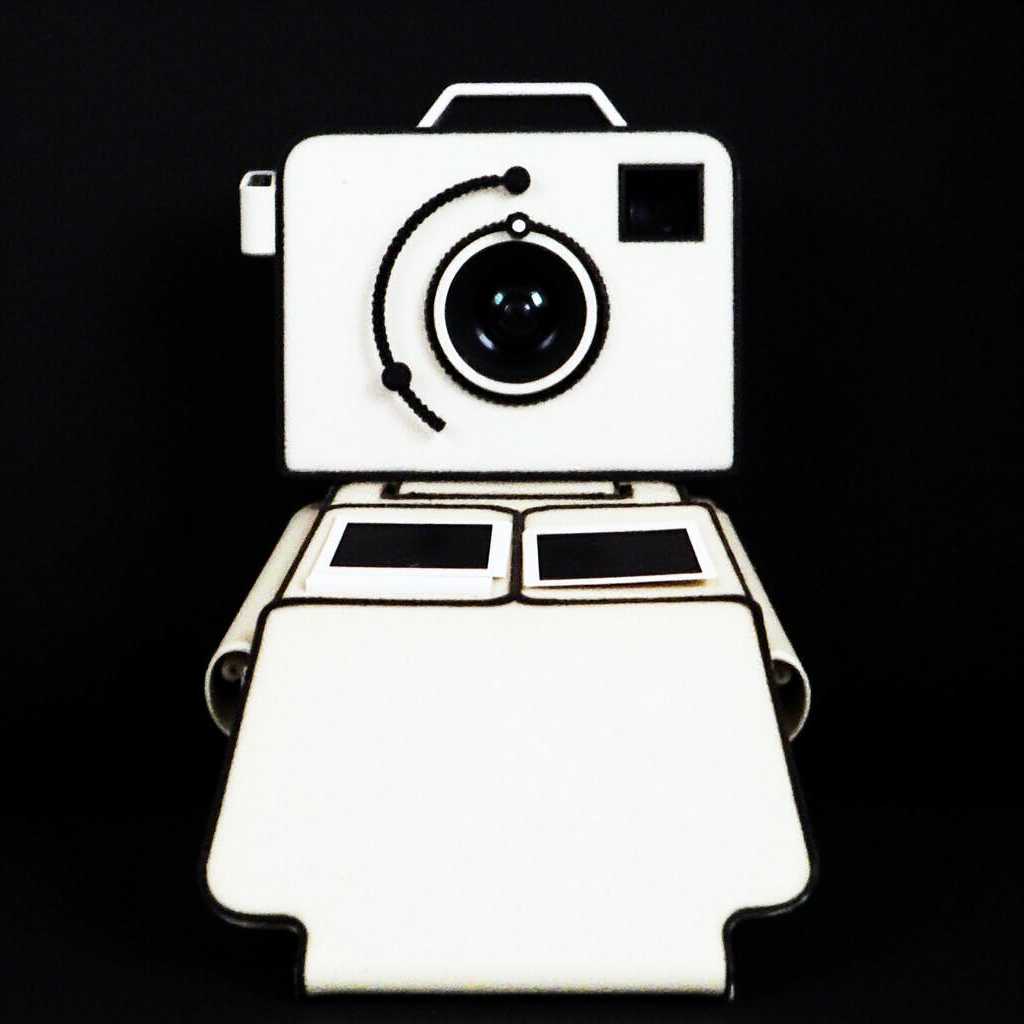} &
            \includegraphics[width=\imgwidth, height=\imgwidth]{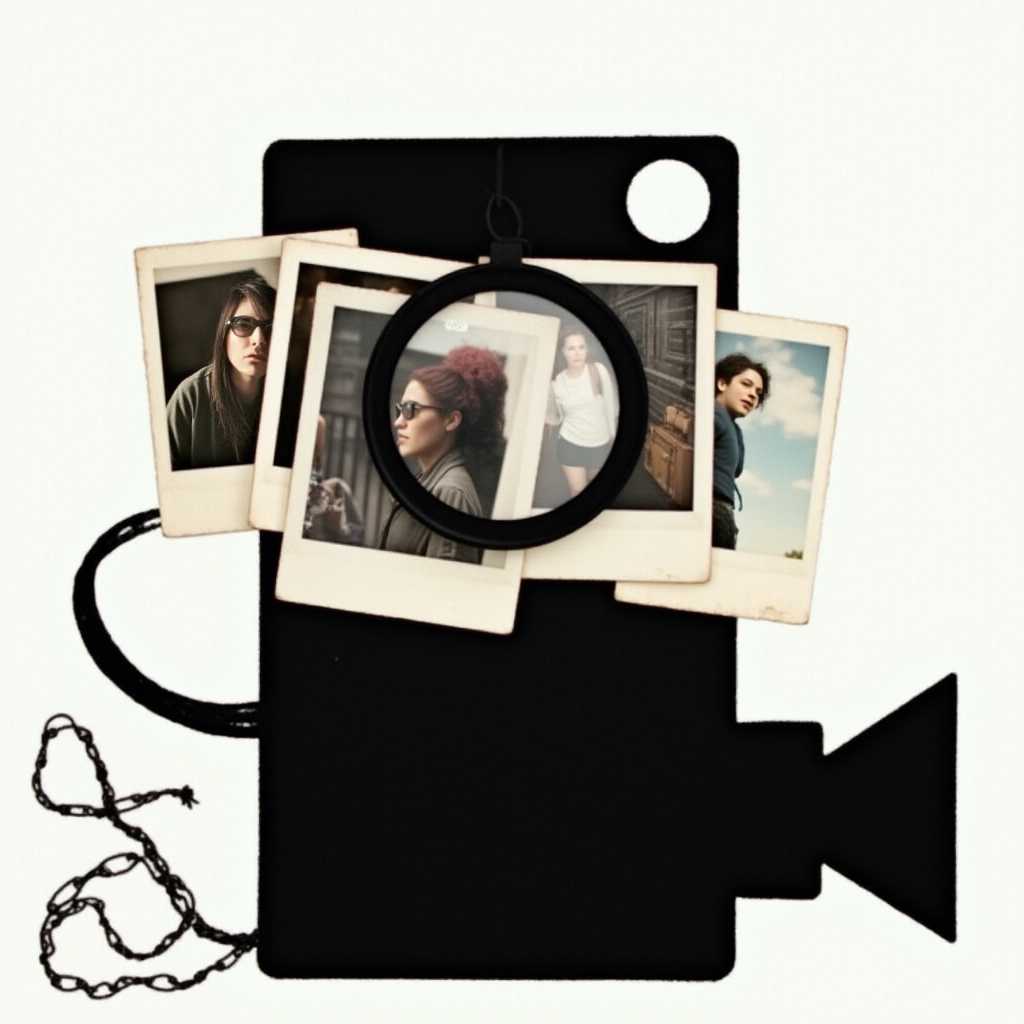} &
            \includegraphics[width=\imgwidth, height=\imgwidth]{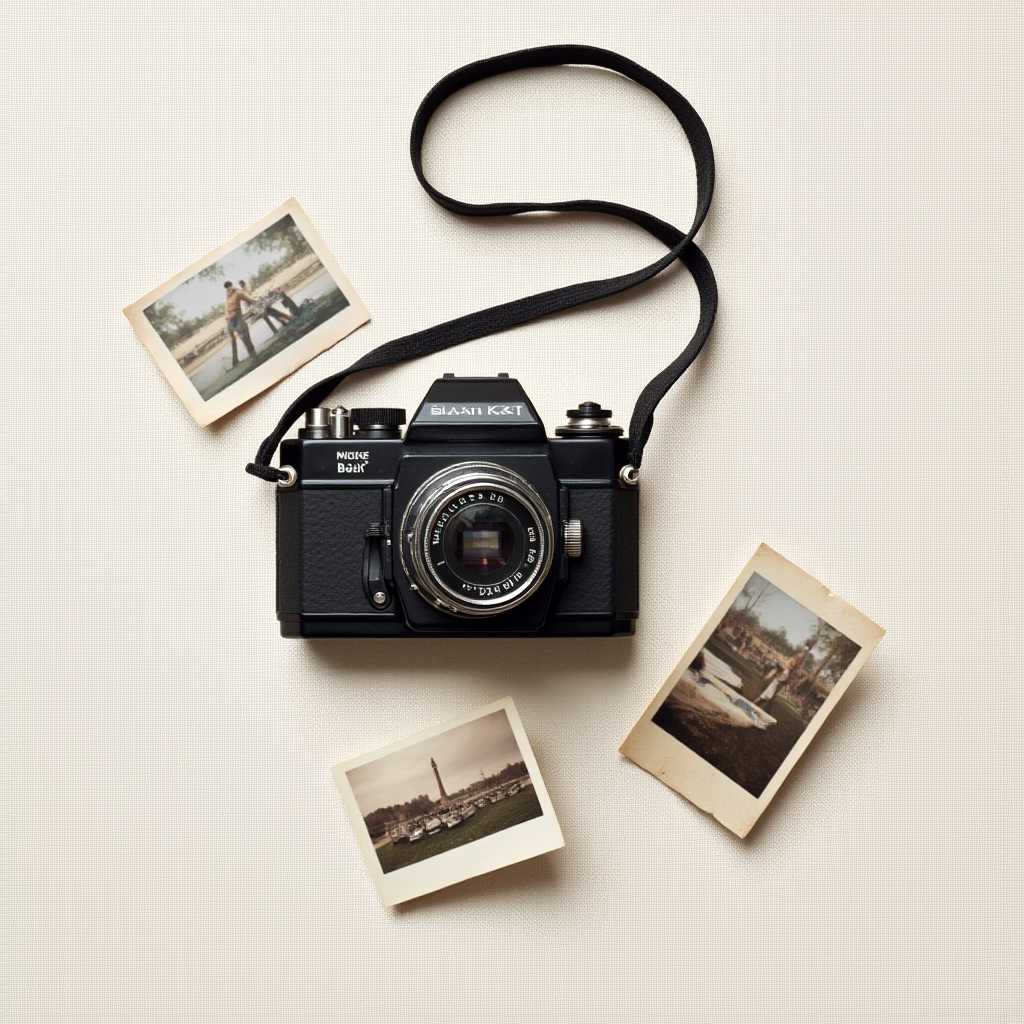} \\

            \vertlabel{PG}{2.5em} & %
            \includegraphics[width=\imgwidth, height=\imgwidth]{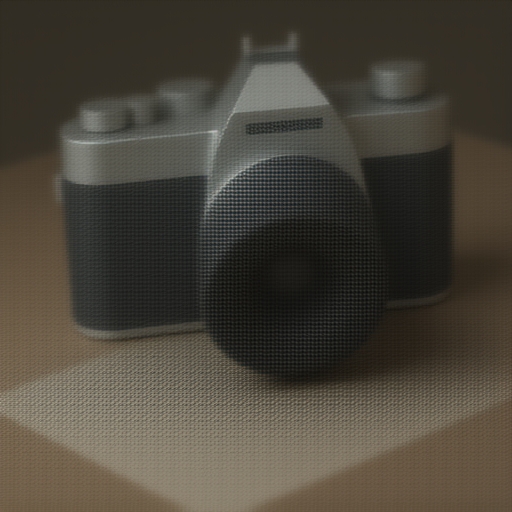} &
            \includegraphics[width=\imgwidth, height=\imgwidth]{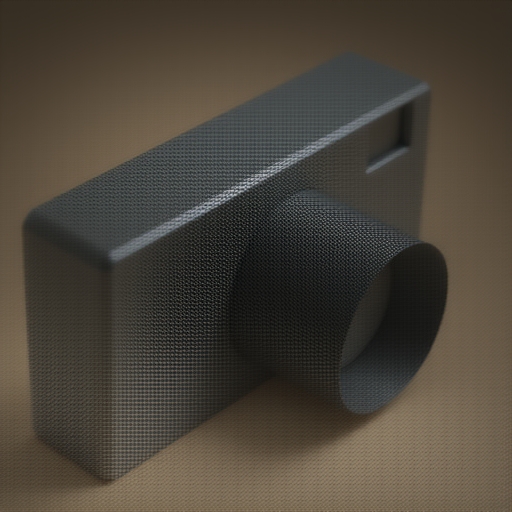} &
            \includegraphics[width=\imgwidth, height=\imgwidth]{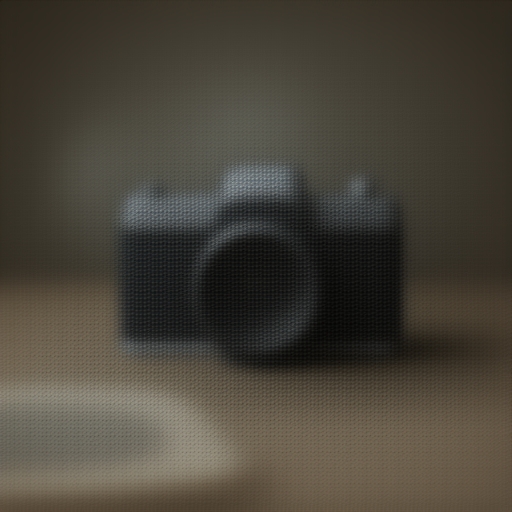} &
            \includegraphics[width=\imgwidth, height=\imgwidth]{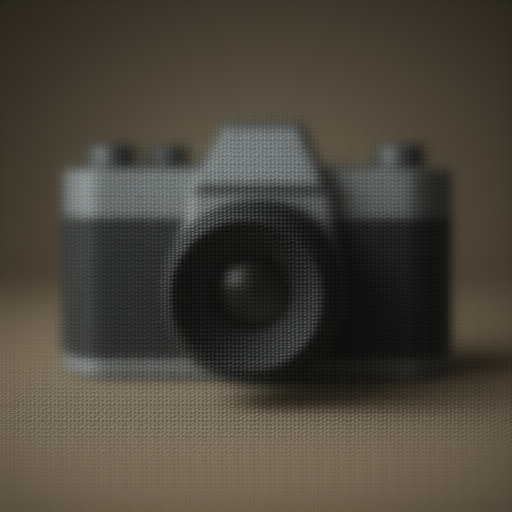} \\
            
            \multicolumn{5}{c}{\vspace{2pt}\small ``A camera with old photographs''} \\

            \vertlabel{Ours}{2.5em} & 
            \includegraphics[width=\imgwidth, height=\imgwidth]{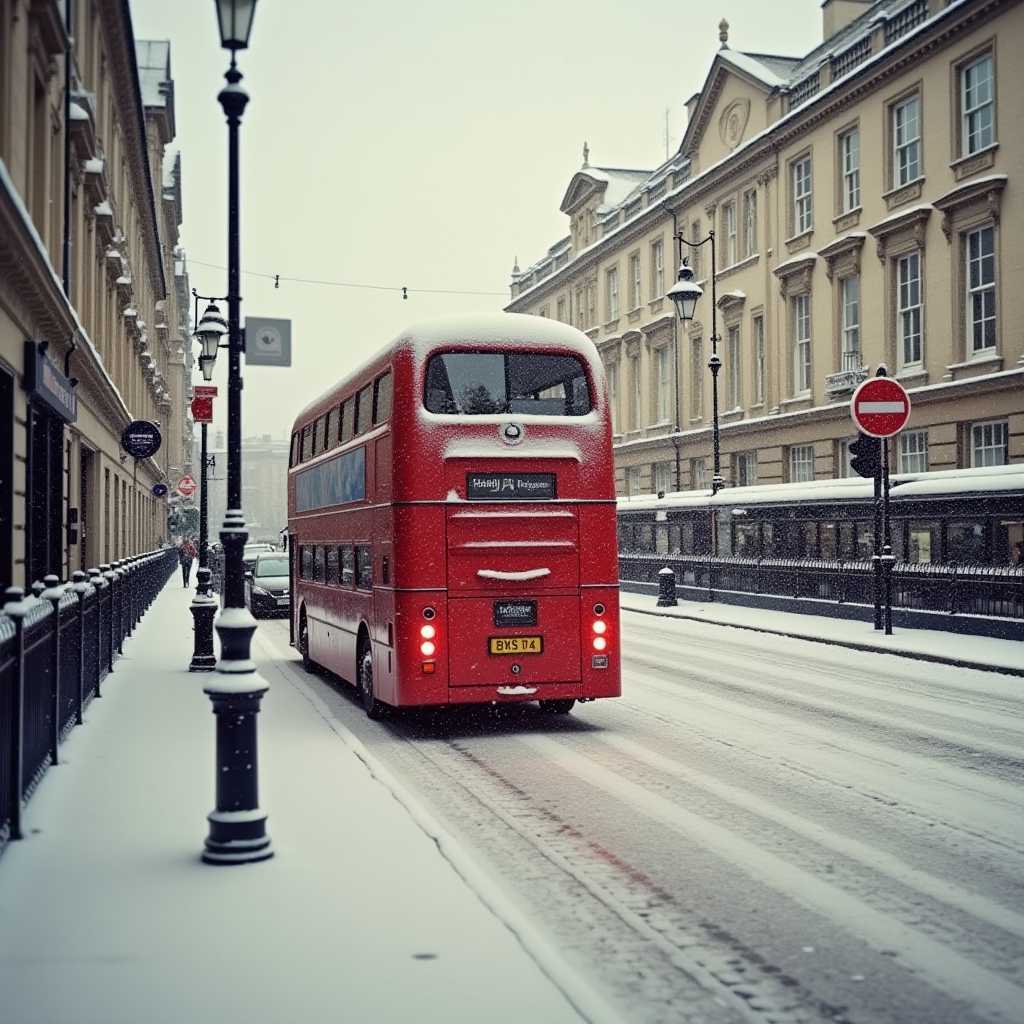} &
            \includegraphics[width=\imgwidth, height=\imgwidth]{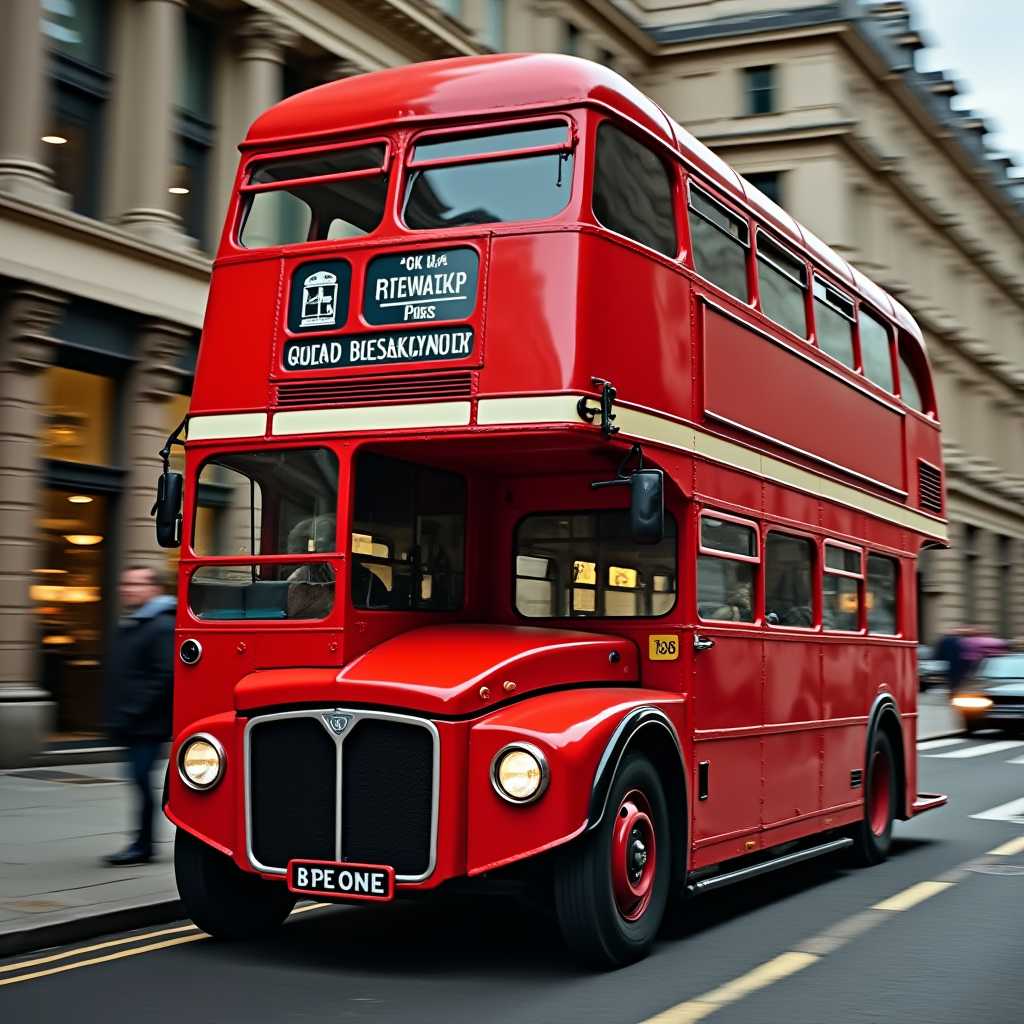} &
            \includegraphics[width=\imgwidth, height=\imgwidth]{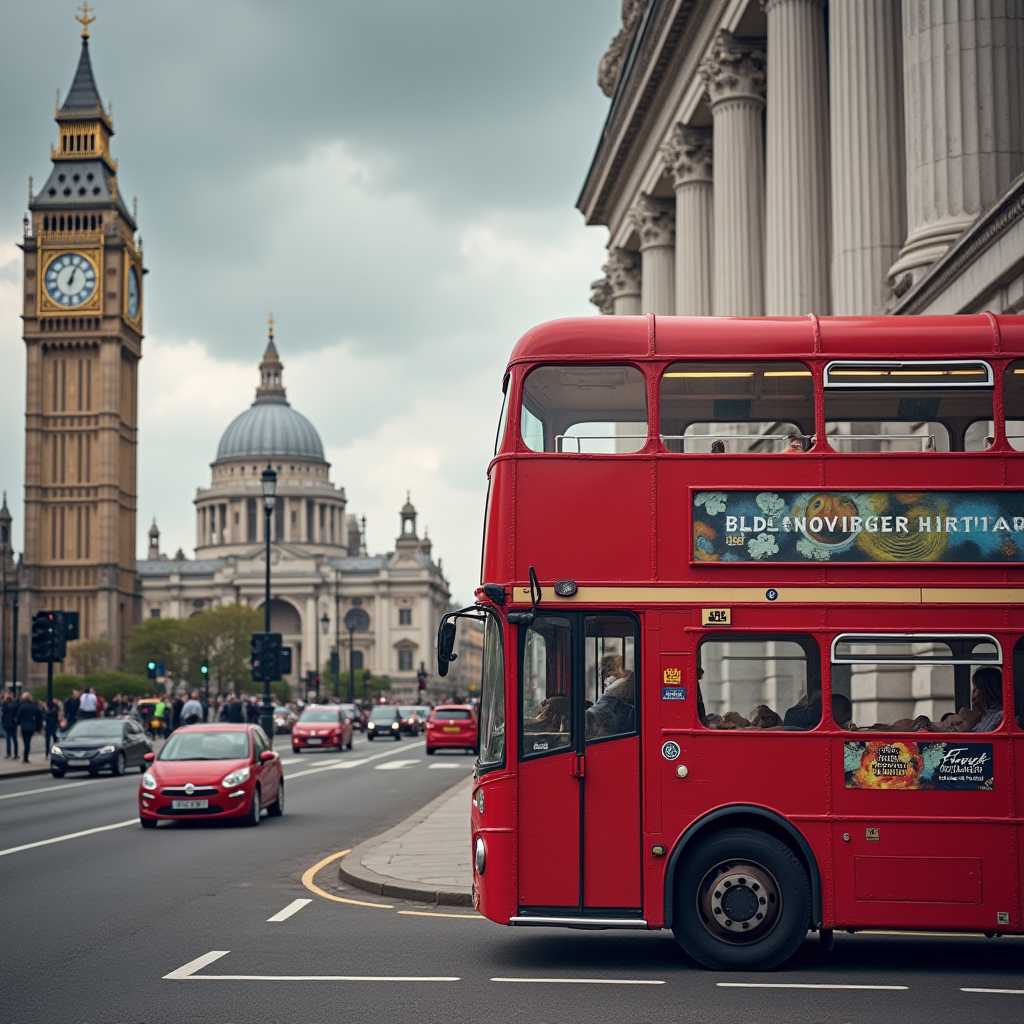} &
            \includegraphics[width=\imgwidth, height=\imgwidth]{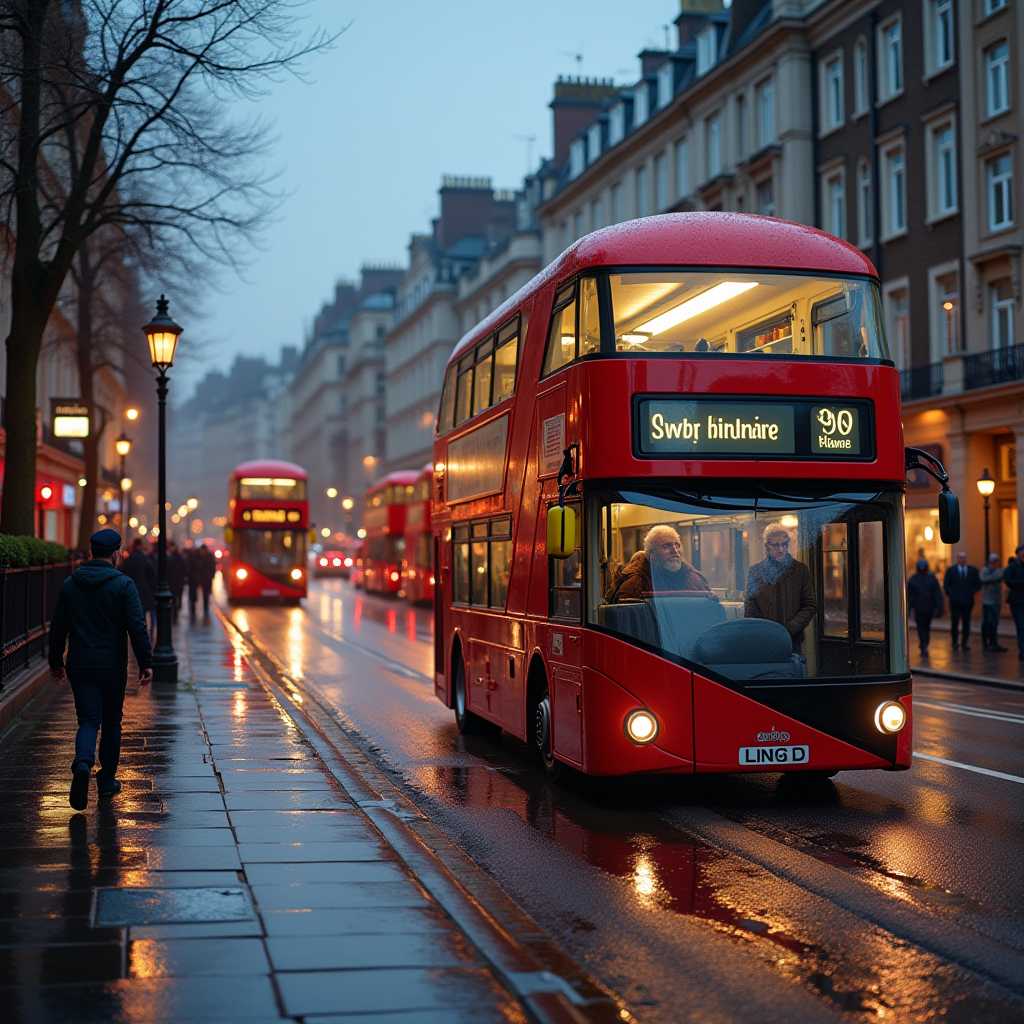} \\

            \vertlabel{SGI}{2.5em} & 
            \includegraphics[width=\imgwidth, height=\imgwidth]{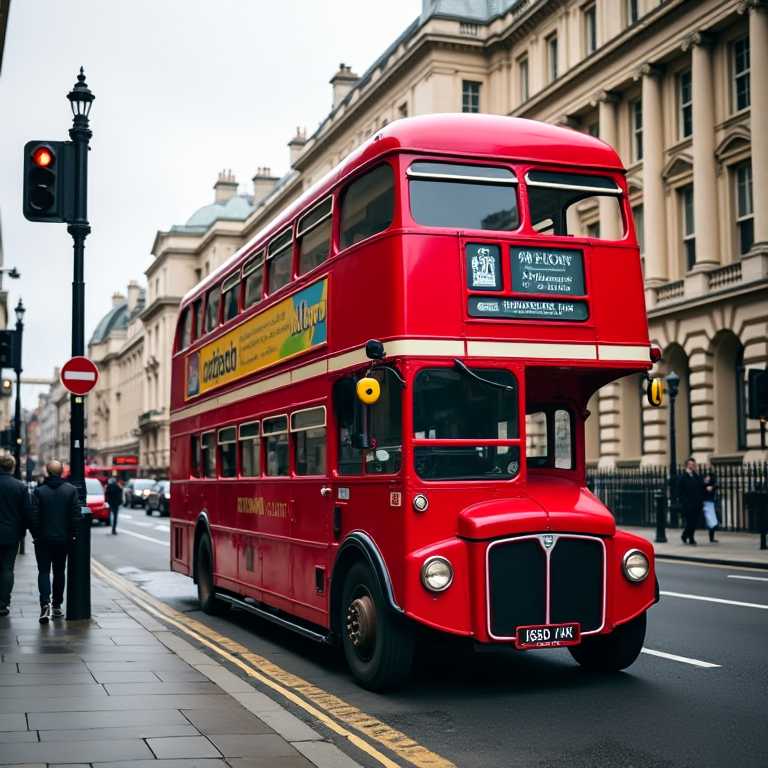} &
            \includegraphics[width=\imgwidth, height=\imgwidth]{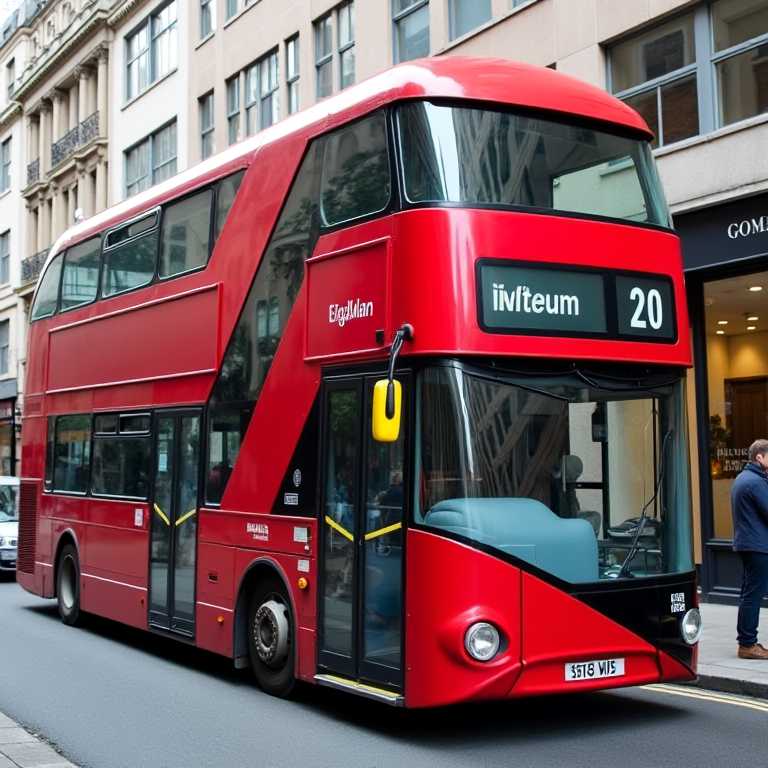} &
            \includegraphics[width=\imgwidth, height=\imgwidth]{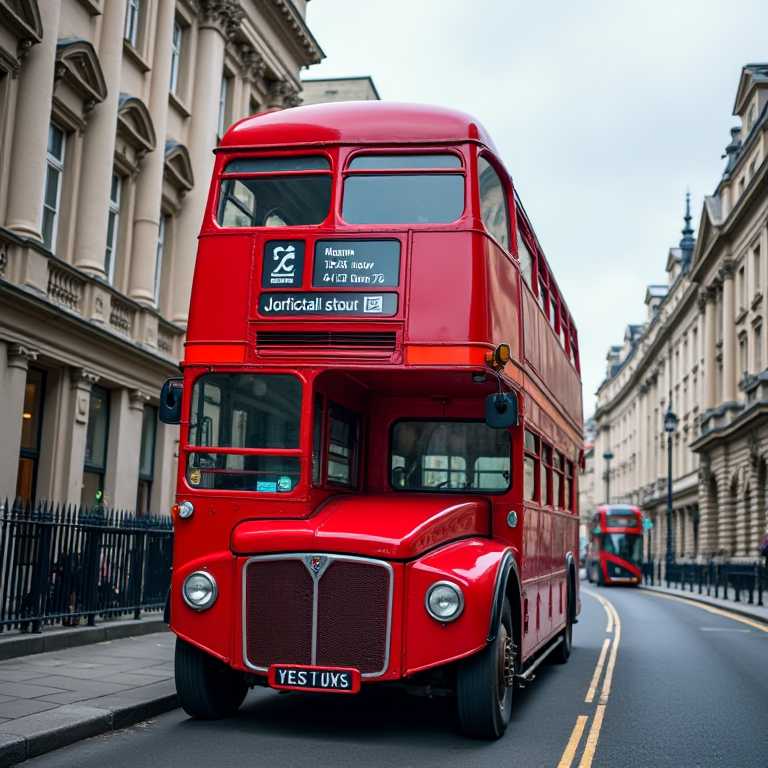} &
            \includegraphics[width=\imgwidth, height=\imgwidth]{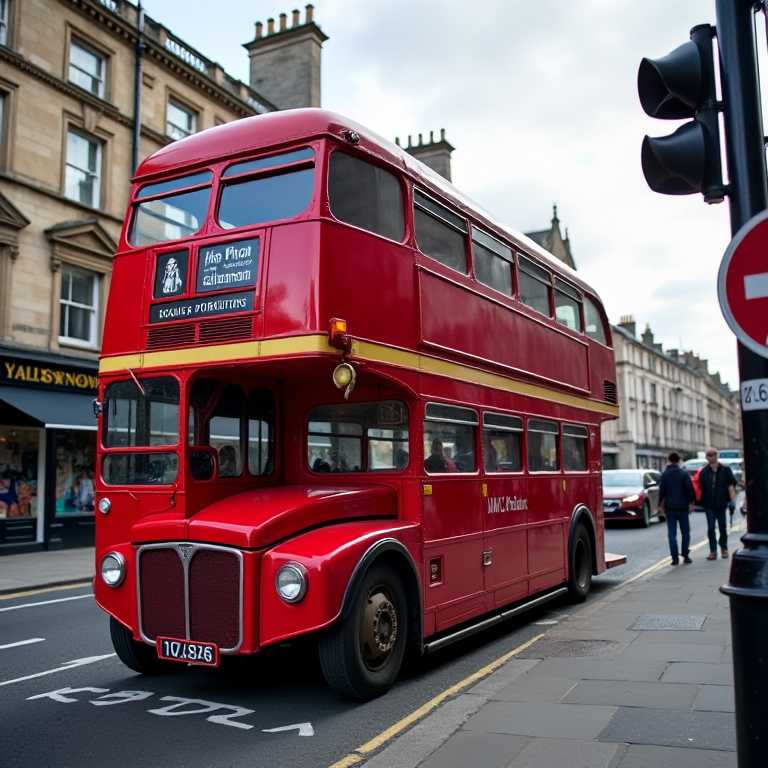} \\
            
            \vertlabel{CADS}{2.5em} & 
            \includegraphics[width=\imgwidth, height=\imgwidth]{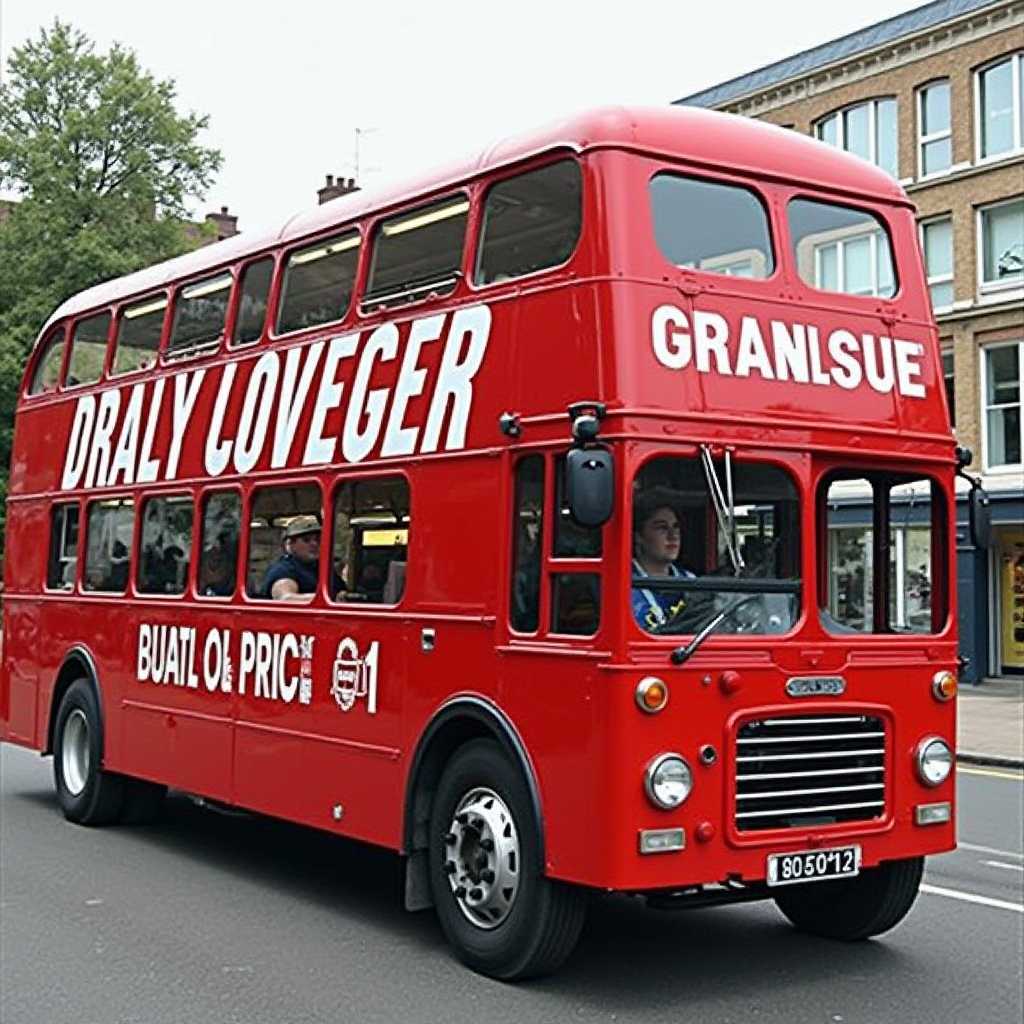} &
            \includegraphics[width=\imgwidth, height=\imgwidth]{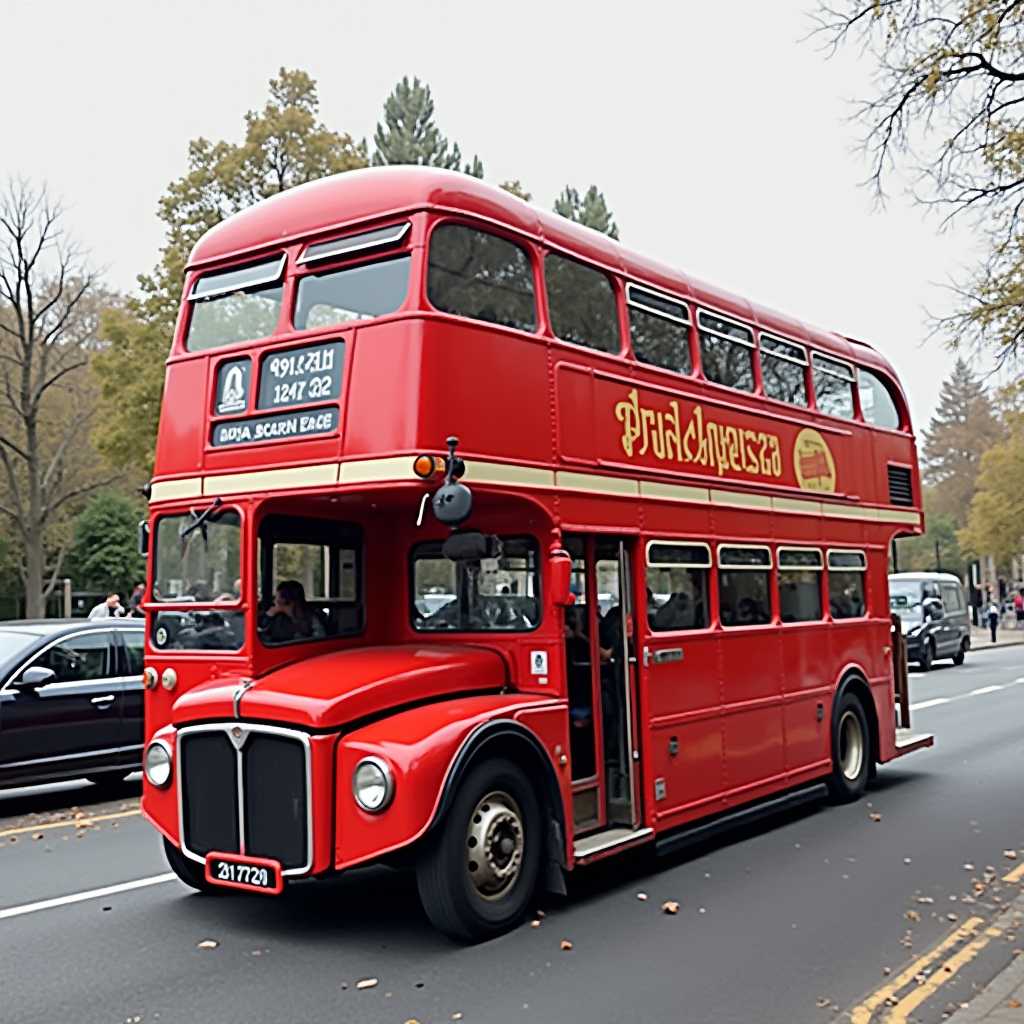} &
            \includegraphics[width=\imgwidth, height=\imgwidth]{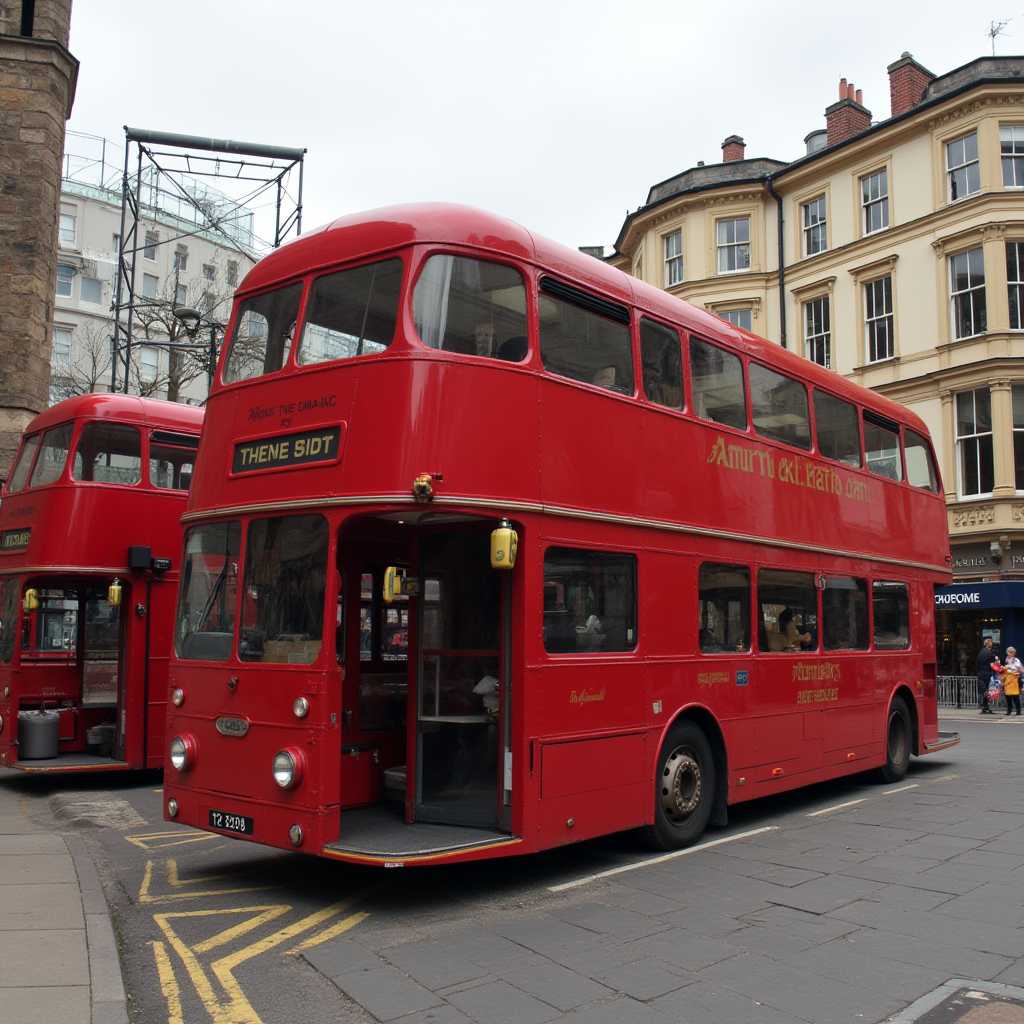} &
            \includegraphics[width=\imgwidth, height=\imgwidth]{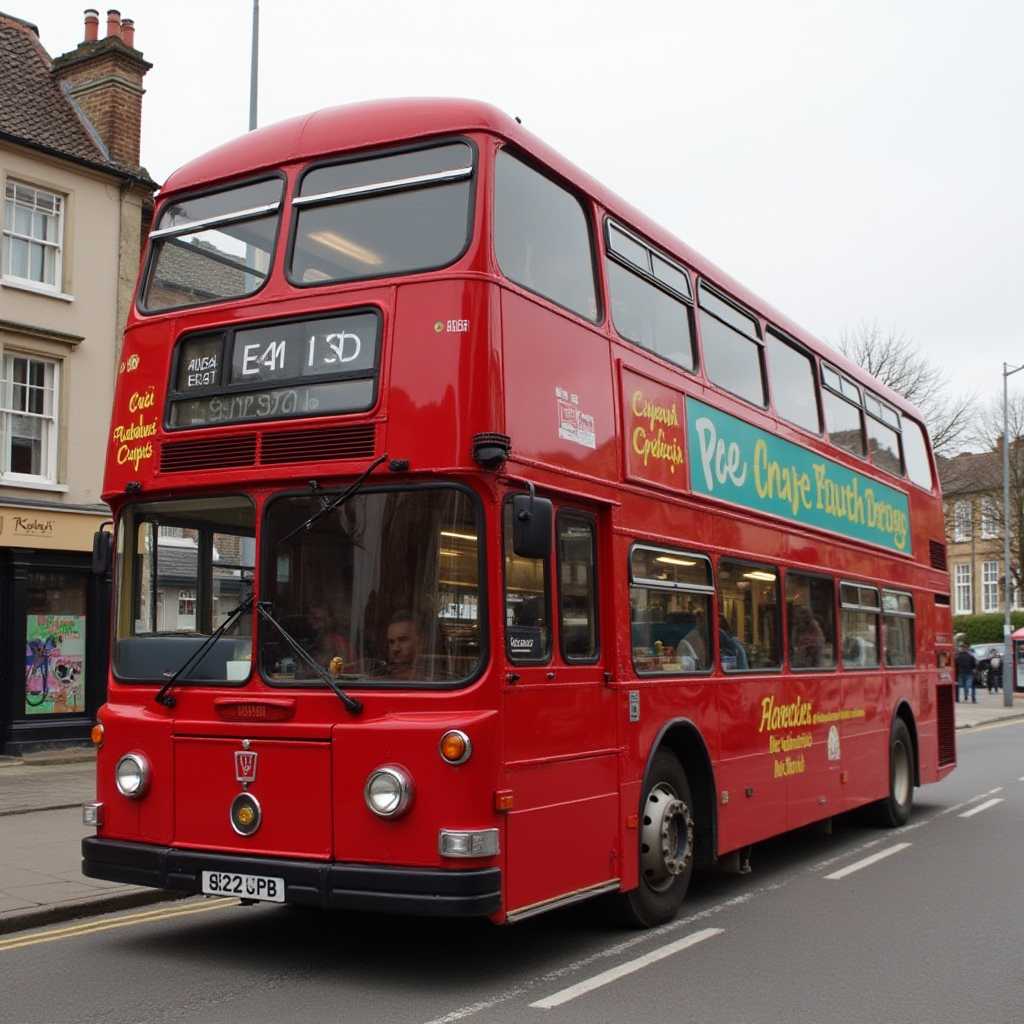} \\

                        \vertlabel{SPARKE}{2.5em} & %
            \includegraphics[width=\imgwidth, height=\imgwidth]{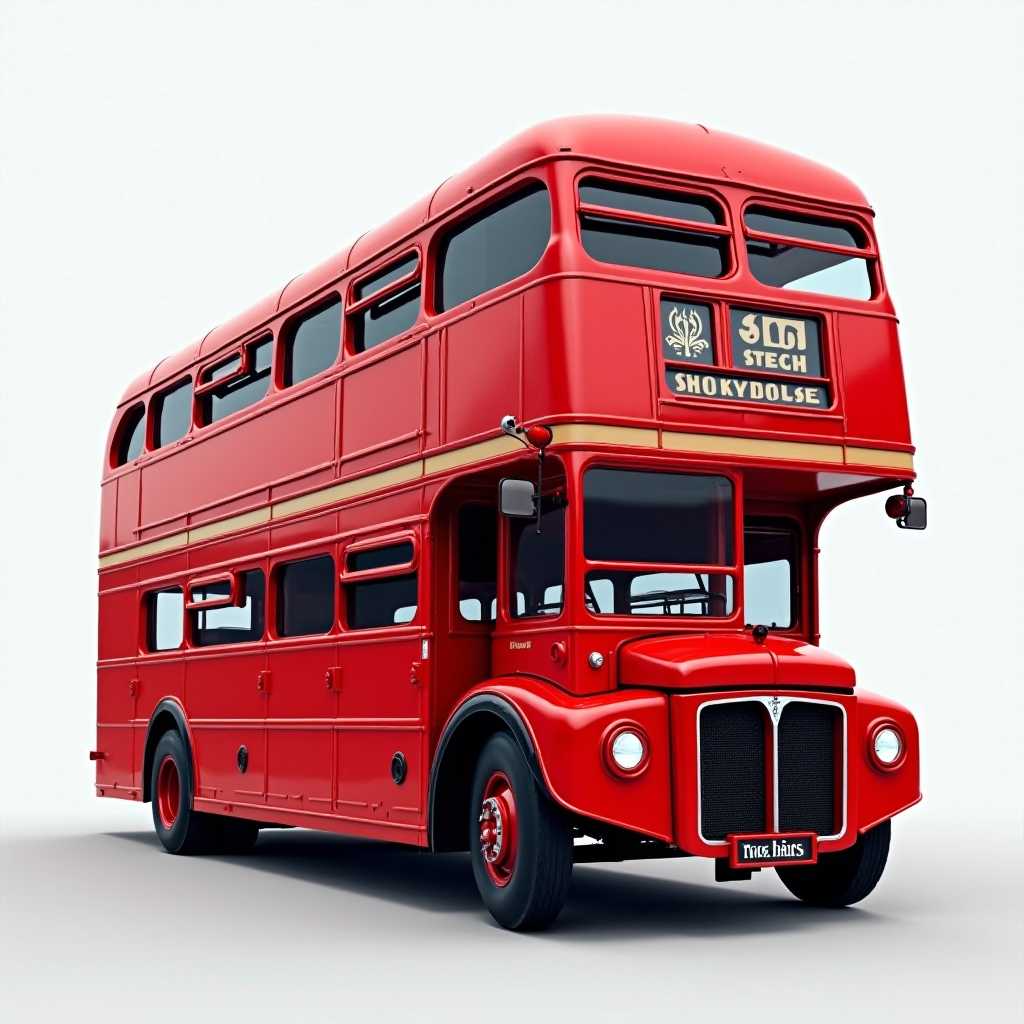} &
            \includegraphics[width=\imgwidth, height=\imgwidth]{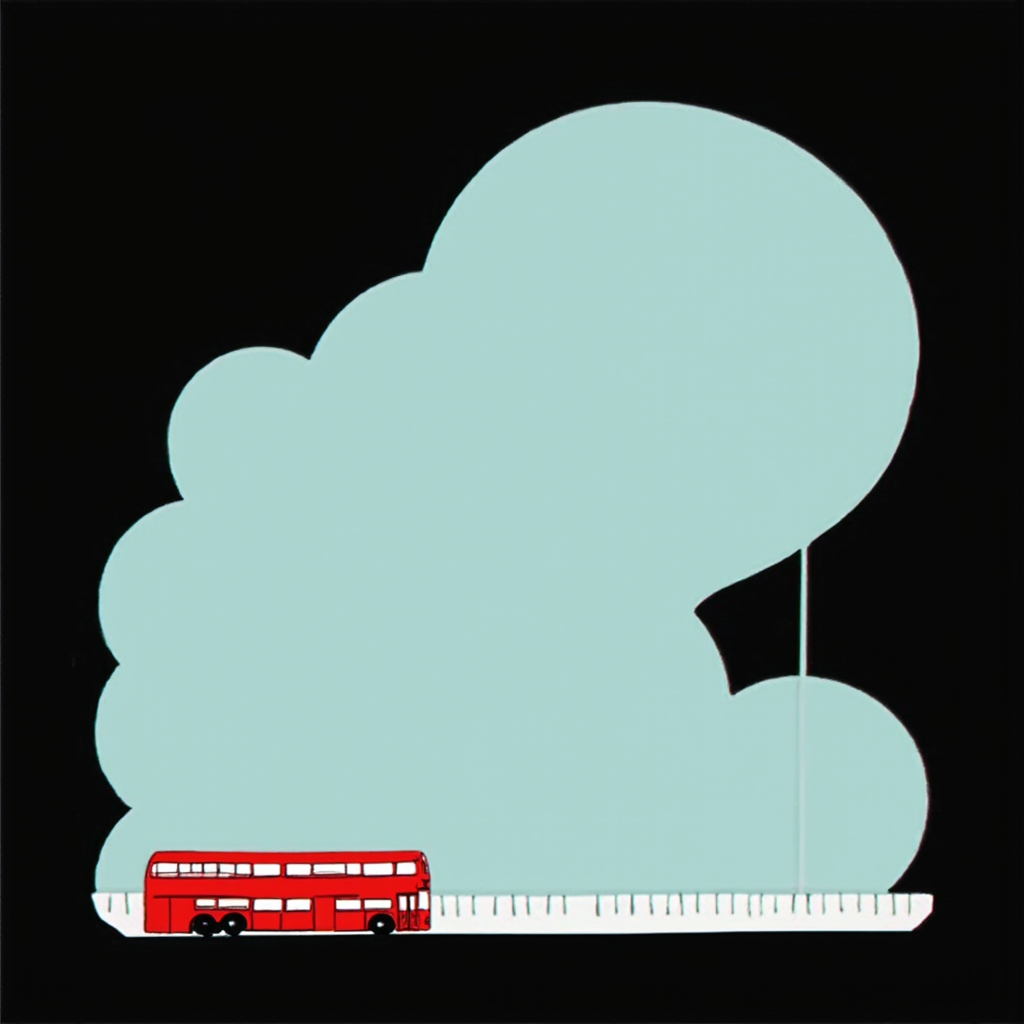} &
            \includegraphics[width=\imgwidth, height=\imgwidth]{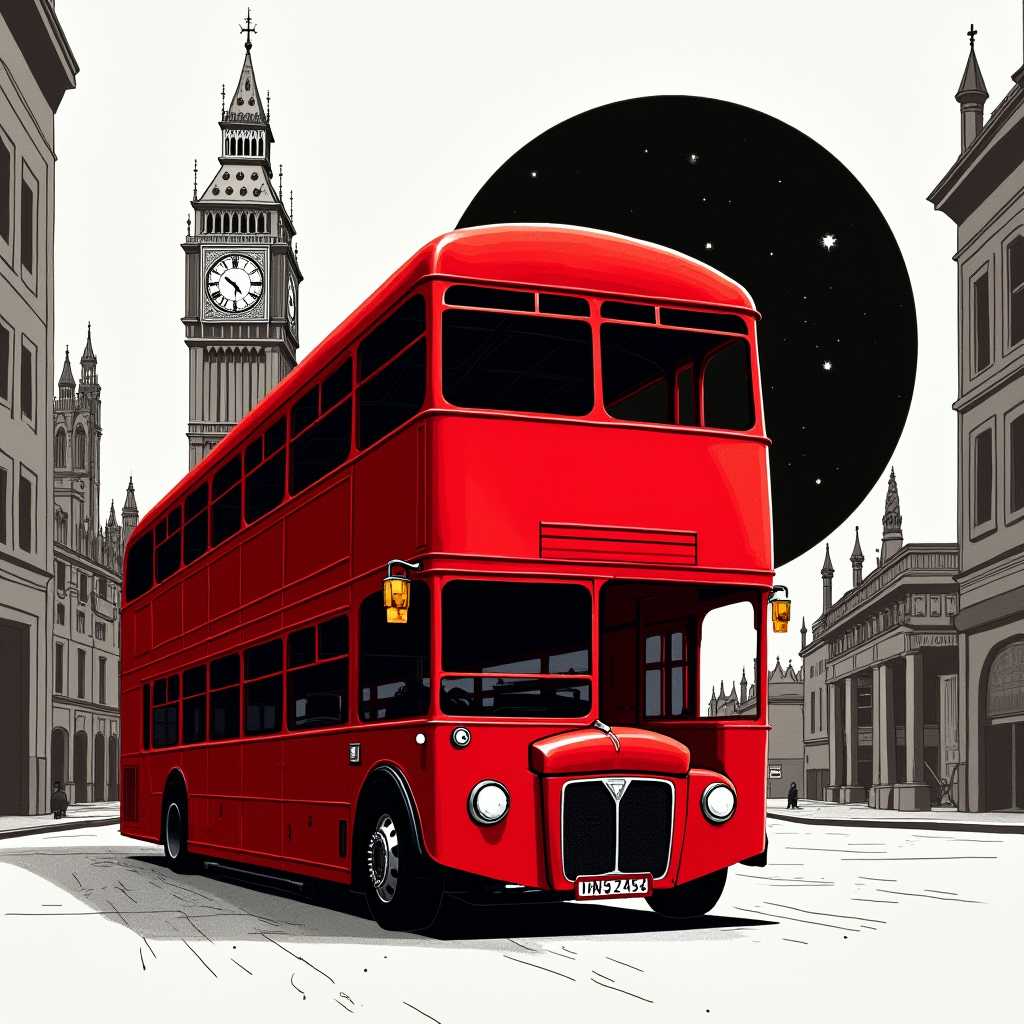} &
            \includegraphics[width=\imgwidth, height=\imgwidth]{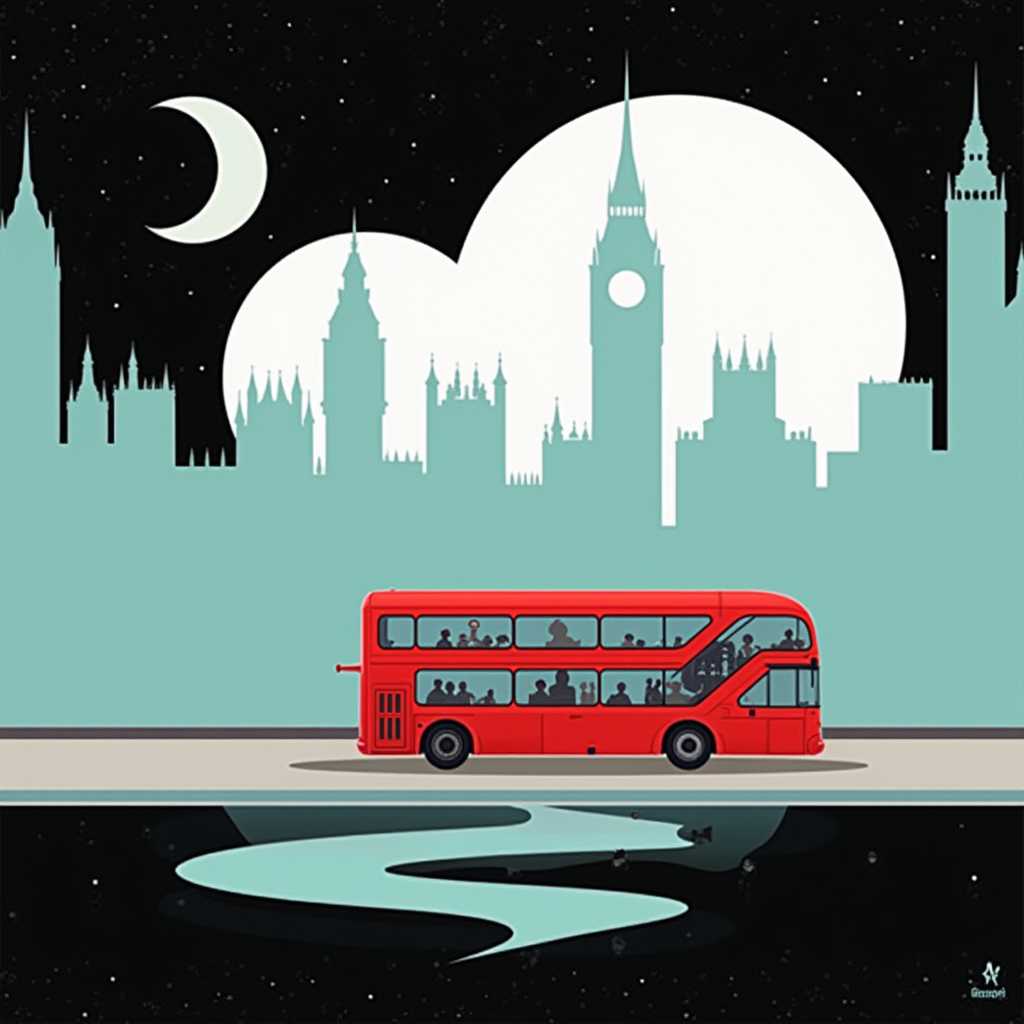} \\

            \vertlabel{PG}{2.5em} & %
            \includegraphics[width=\imgwidth, height=\imgwidth]{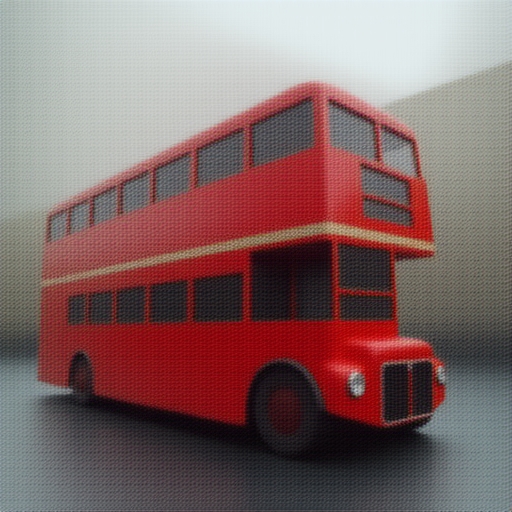} &
            \includegraphics[width=\imgwidth, height=\imgwidth]{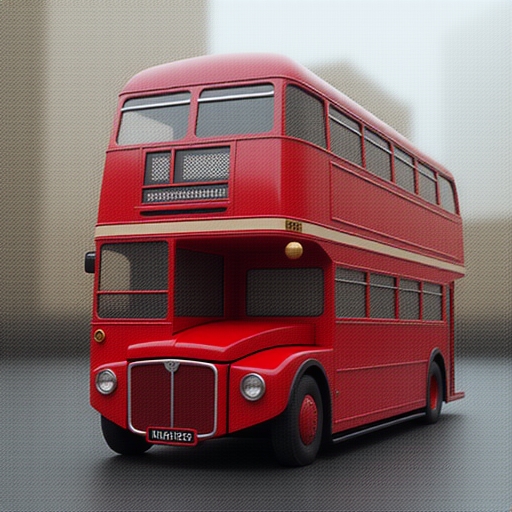} &
            \includegraphics[width=\imgwidth, height=\imgwidth]{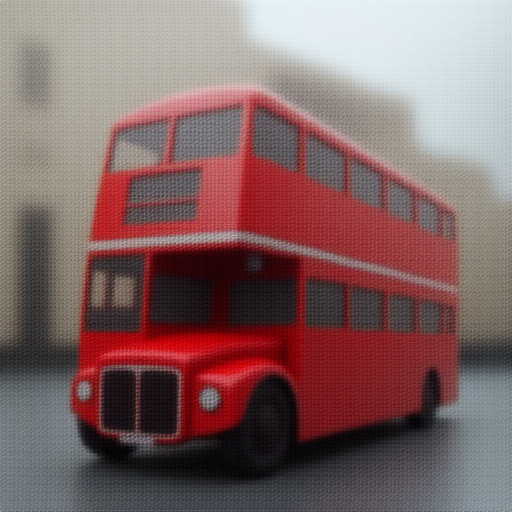} &
            \includegraphics[width=\imgwidth, height=\imgwidth]{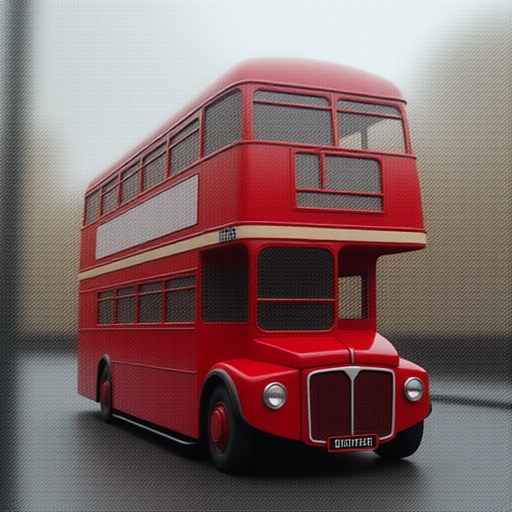} \\

            \multicolumn{5}{c}{\vspace{2pt}\small ``A red London double-decker bus''}
        \end{tabular}
    \end{minipage}

    \caption{\textbf{Qualitative comparison of our Contextual Repulsion approach against baseline methods.} %
    Each quadrant displays four generated samples per method for a given prompt.}
    \label{fig:comparisons}
\end{figure*}

We present qualitative comparisons against the baselines in Figure~\ref{fig:comparisons}.
As illustrated, downstream methods like PG and SPARKE often introduce visual artifacts because they intervene directly in the VAE latent space. For instance, in the ``red bus''
example, PG fails to modify the image structure, while SPARKE succeeds in moving objects but leaves patterned ``holes'' in their original locations.

In contrast, upstream methods maintain higher image quality, though they face different trade-offs. CADS frequently leads to semantic drift, where diversity is achieved through weak prompt alignment (e.g., replacing ``photographs'' with people, or a ``phoenix`` with a bonfire). SGI, which filters a large set of initial noise candidates through optimization, achieves both high quality and prompt adherence by minimizing intervention. However, SGI often struggles to produce high variation for prompts where the base model lacks inherent diversity, resulting in repetitive subject appearances and compositions (e.g., the ``red bus'').
 
Our method achieves richer diversity even with challenging prompts, without sacrificing alignment or quality. Interestingly, the axes of variation adapt to each prompt: for the ``phoenix,'' the model alternates between artistic styles; for the ``bus,'' it varies weather and pose;
and for the ``camera with old photographs'' and ``wolf pack,'' it generates unique compositions and object arrangements.

\begin{figure}[t]
    \centering
    \small
    
    \setlength{\tabcolsep}{0pt}
    \newcommand{\vertlabel}[2]{\raisebox{#2}{\rotatebox{90}{{#1}}}}
    \newcommand{\imagewidth}{0.19\linewidth}

    \begin{tabular}{c@{\hspace{2pt}}c@{\hspace{4pt}}c@{}c@{}c@{}c@{}}

        \vertlabel{Flux Kontext}{0.3em} &
        \includegraphics[width=\imagewidth]{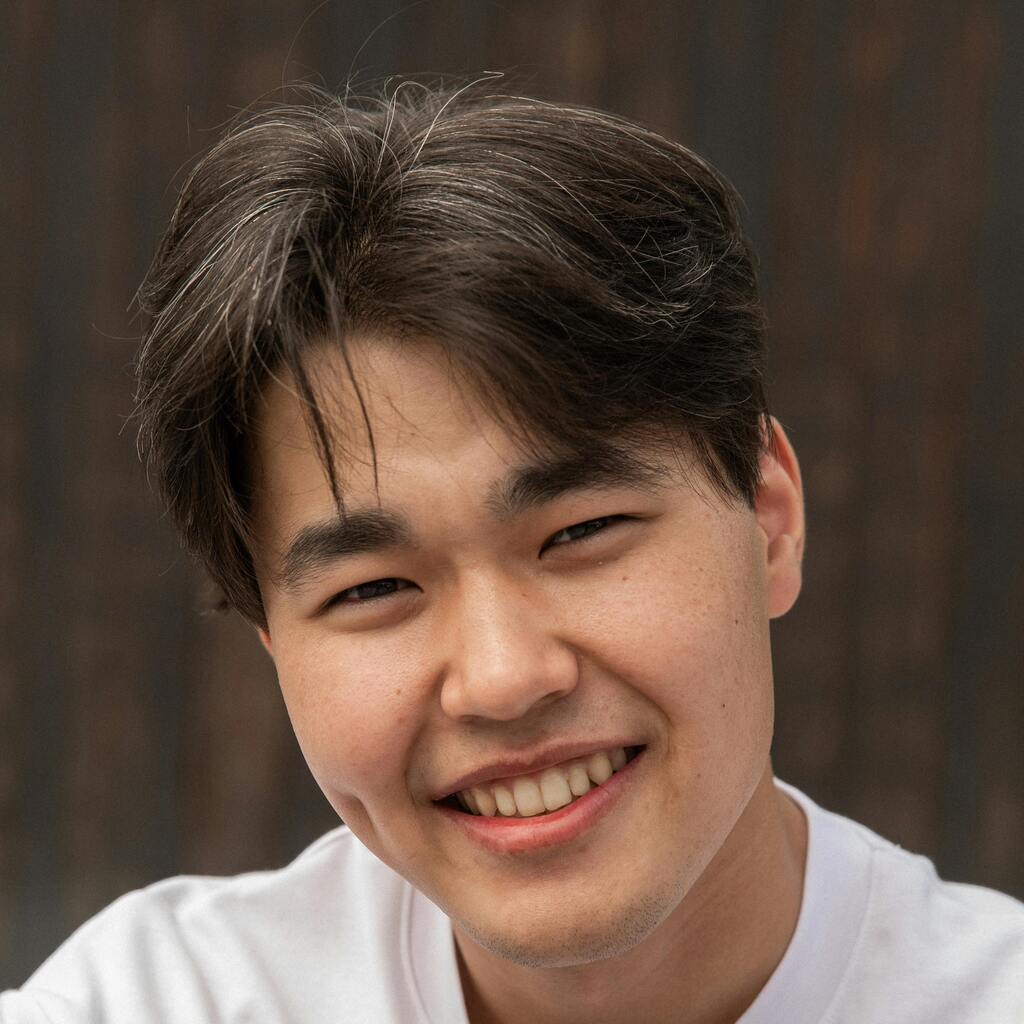} &
        \includegraphics[width=\imagewidth]{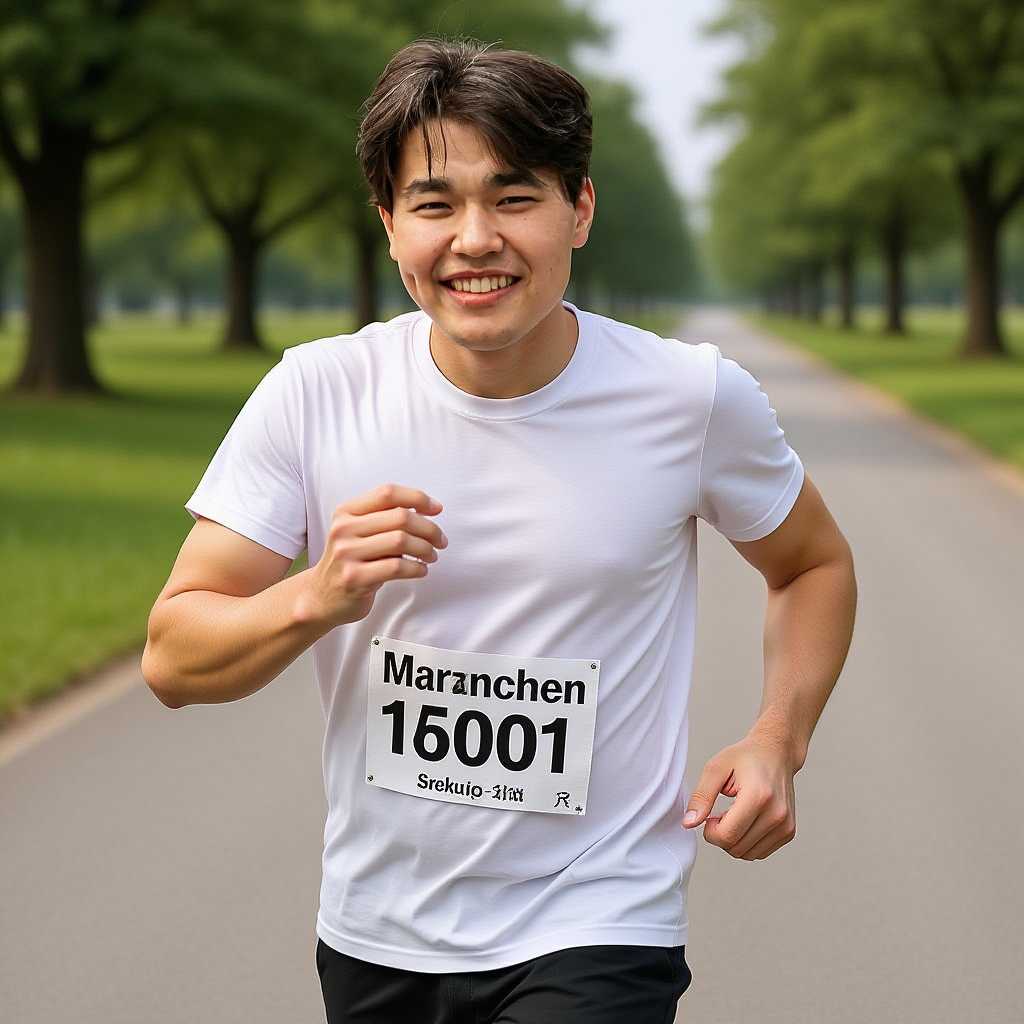} &
        \includegraphics[width=\imagewidth]{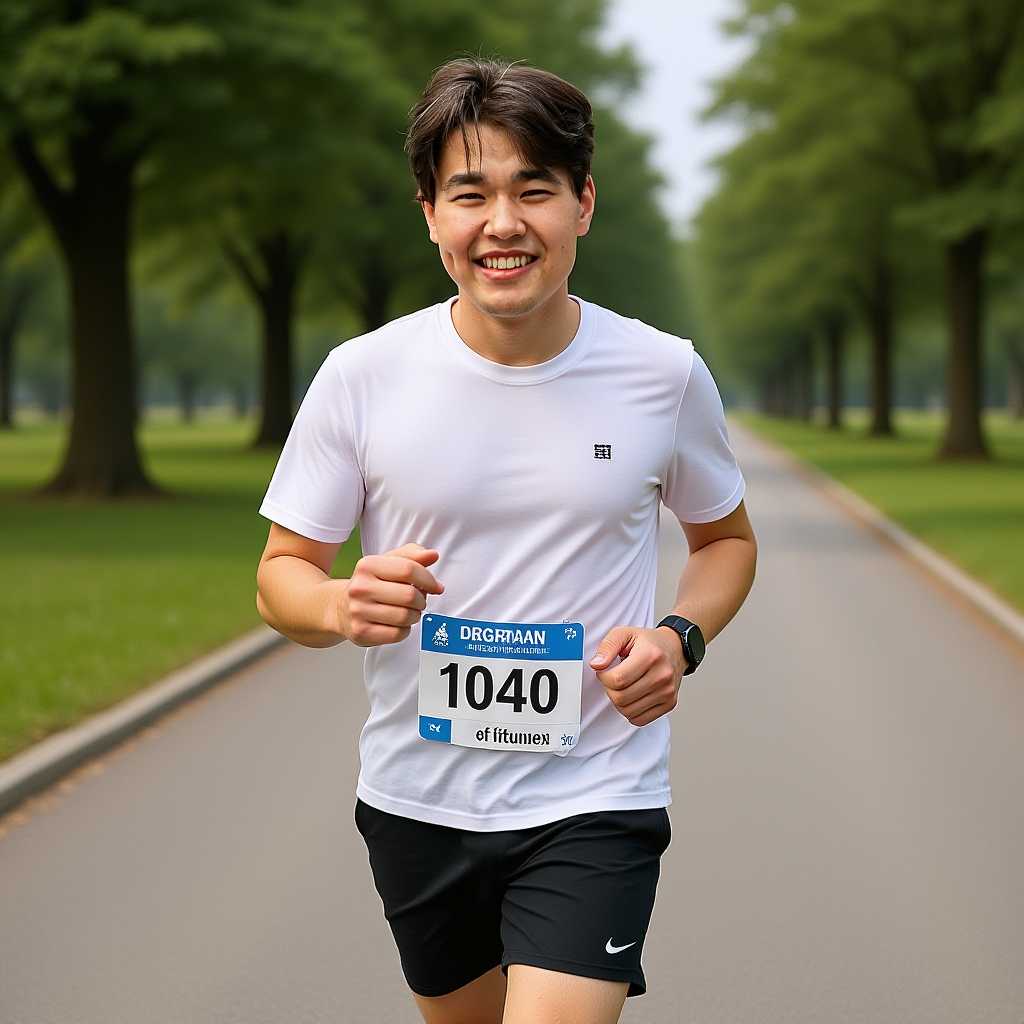} &
        \includegraphics[width=\imagewidth]{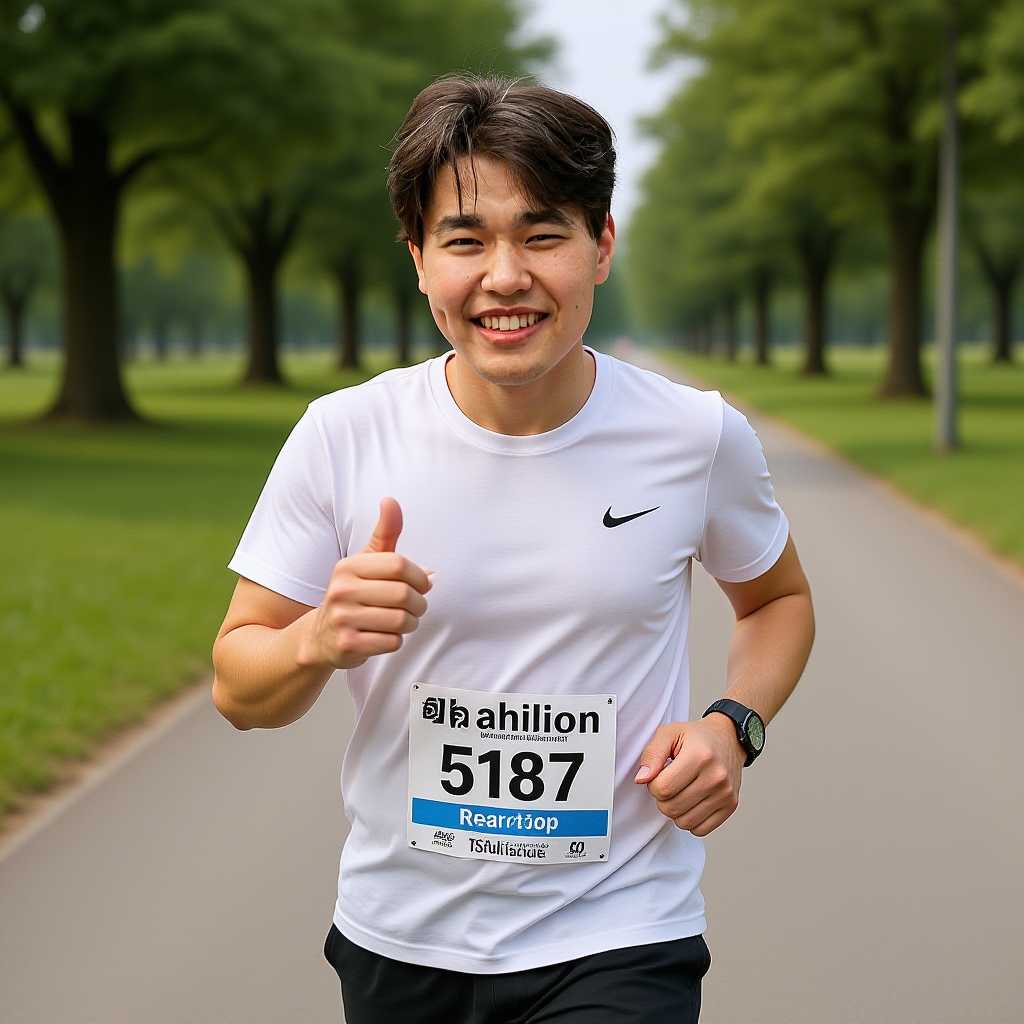} &
        \includegraphics[width=\imagewidth]{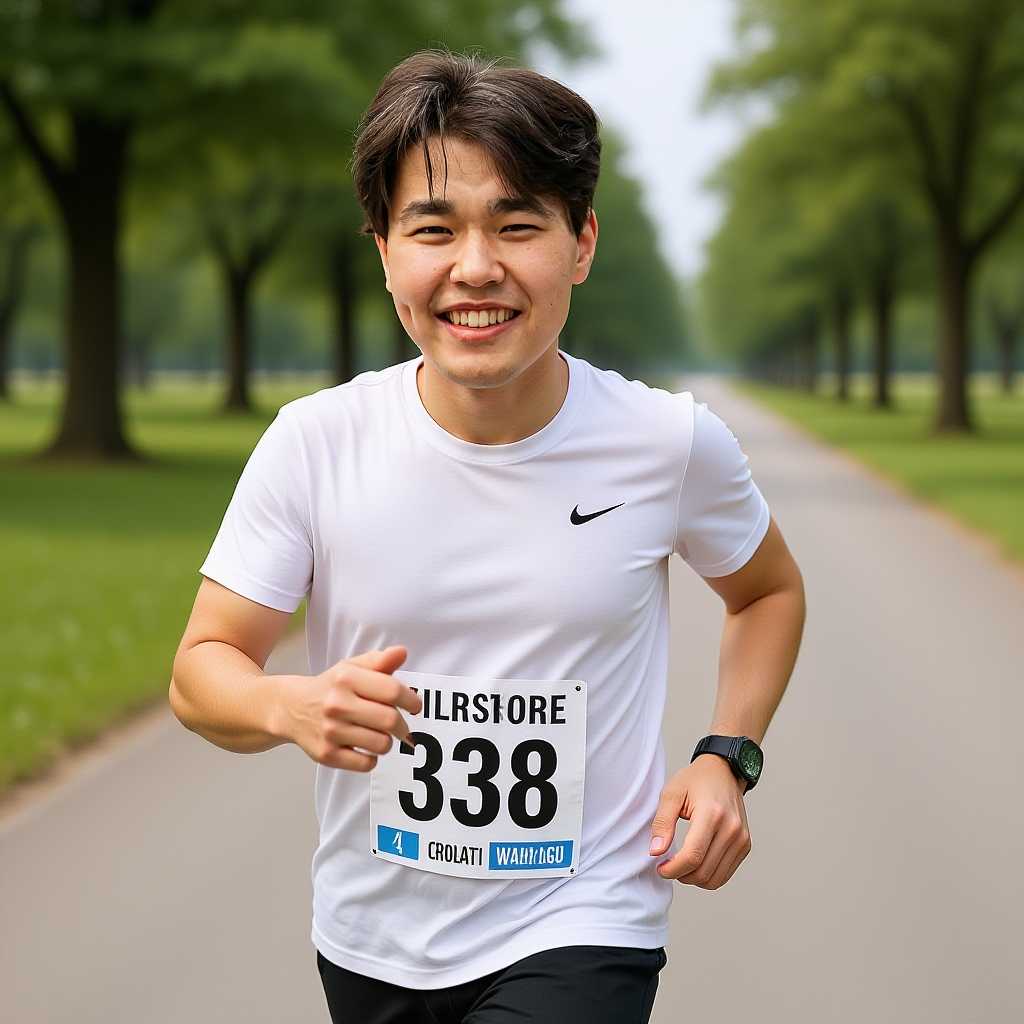} \\

        \vertlabel{Ours}{1.9em} &
        \includegraphics[width=\imagewidth]{images/kontext_figure/a-person-running-marathon/2.jpg} &
        \includegraphics[width=\imagewidth]{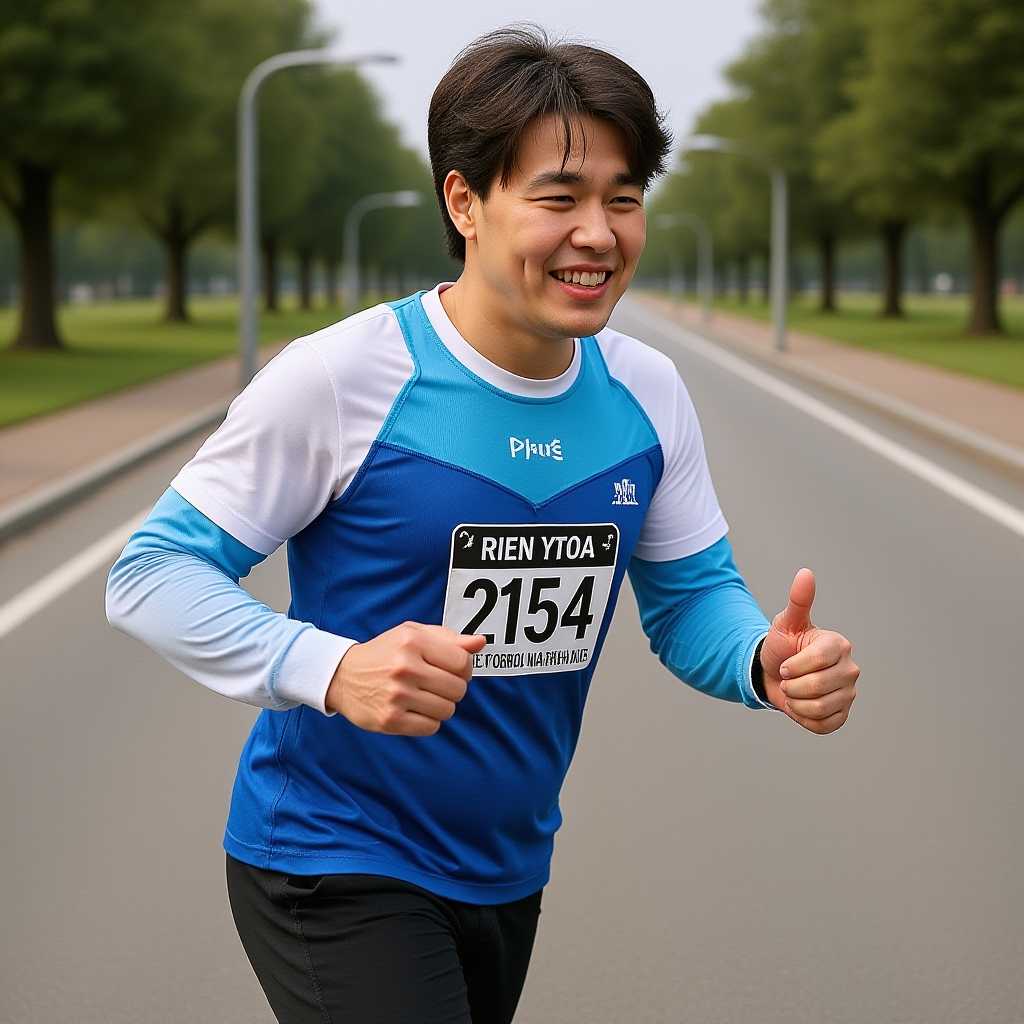} &
        \includegraphics[width=\imagewidth]{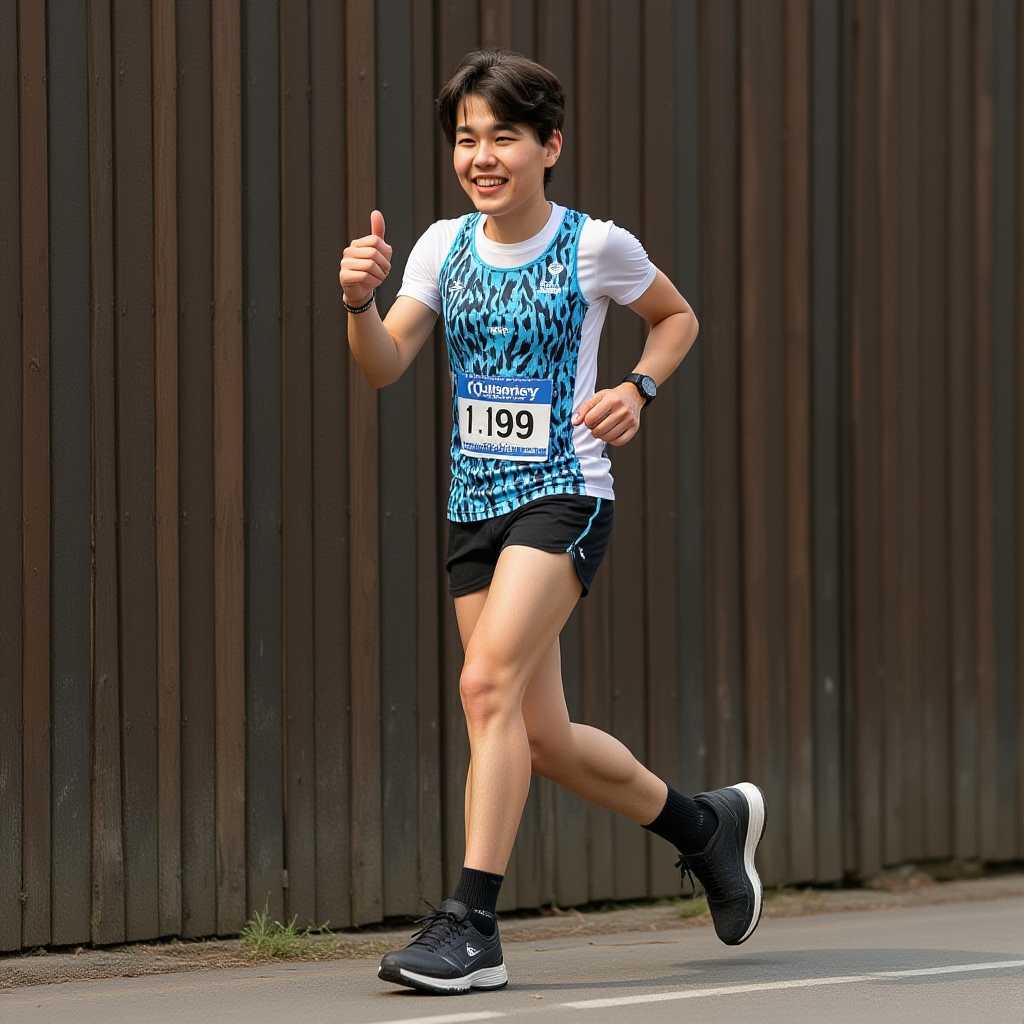} &
        \includegraphics[width=\imagewidth]{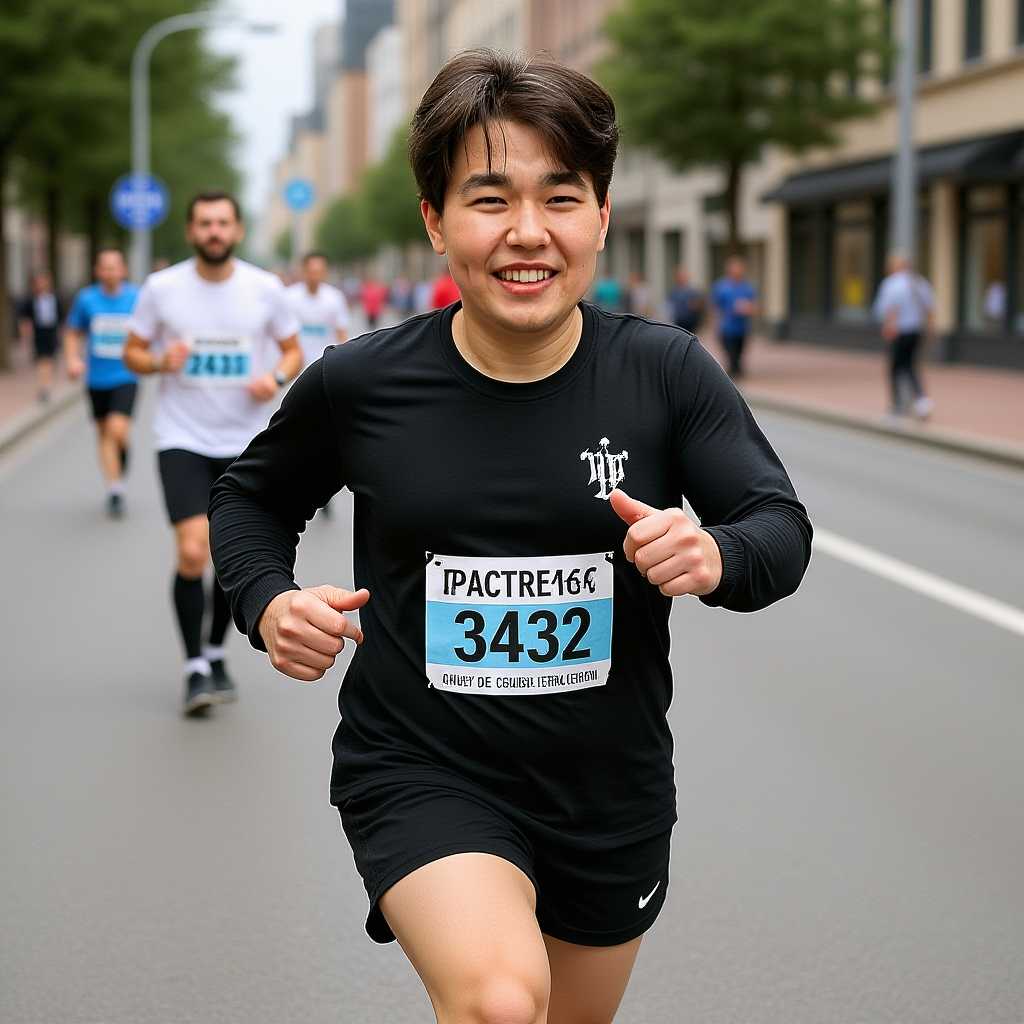} &
        \includegraphics[width=\imagewidth]{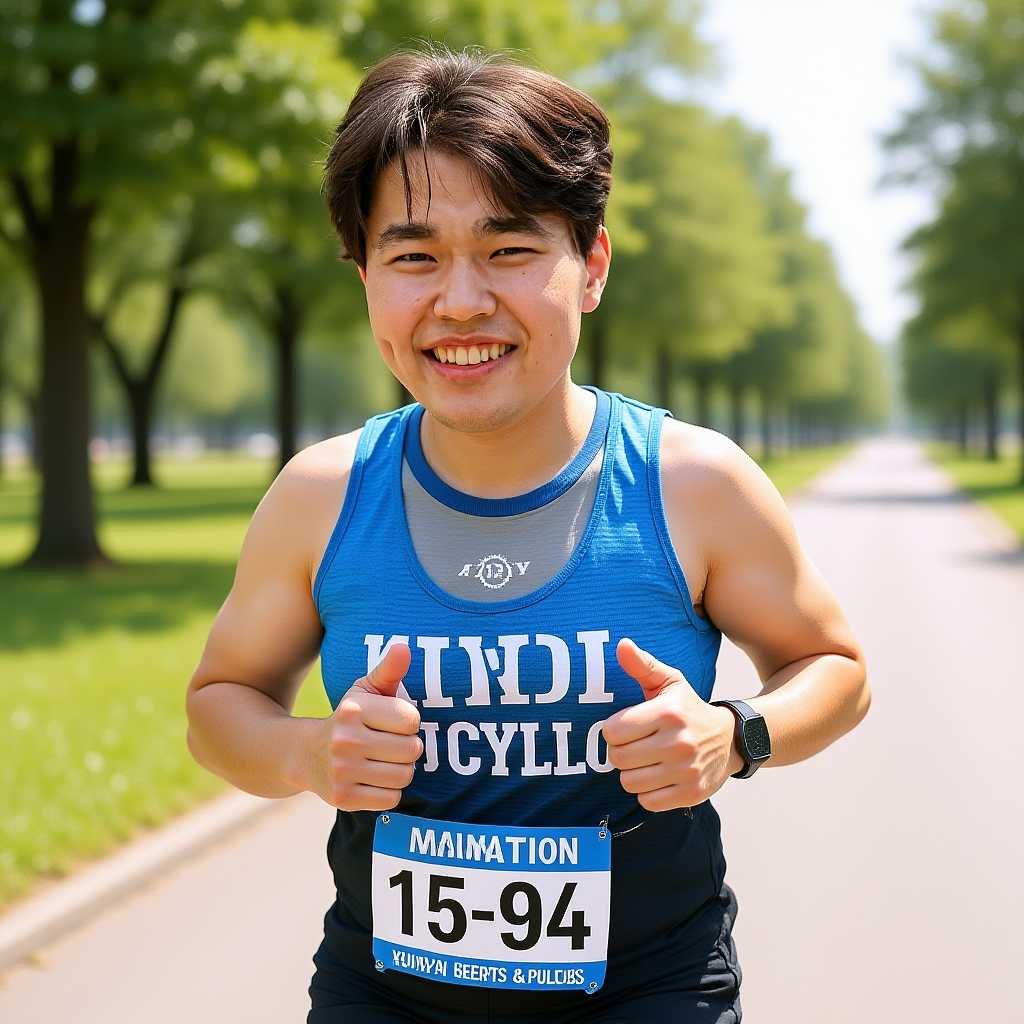} \\
        &
        Input Image & \multicolumn{4}{c}{``a person running a marathon''} \\

    \end{tabular}
    \caption{
        \textbf{Integration with image editing models.} We demonstrate that our method can be successfully integrated into Flux-Kontext to generate high-quality diverse results.}
    \label{fig:kontext_integration}
\end{figure}

\paragraph{Example result on Flux-Kontext.}

In Figure~\ref{fig:kontext_integration}, we demonstrate that our method generalizes beyond text-to-image generation and can be applied \emph{out of the box} to image editing models, specifically Flux~Kontext~\cite{labs2025flux}. Perhaps surprisingly, this requires no modification to the model or to our intervention strategy: we apply the exact same Contextual Space repulsion within the editing instruction stream. While the base editing model produces nearly identical edits across different random seeds, our approach yields diverse yet coherent edit realizations, all while preserving the intended edit semantics and maintaining the visual integrity of the original image. This result highlights that contextual repulsion operates at a level of abstraction that is compatible with both generation and editing paradigms, despite being developed specifically for text-to-image models.

\subsection{Quantitative Results}

\paragraph{Diversity-Quality trade-off.}

\begin{figure}[ t ]
    \centering
    \scriptsize
    \includegraphics[width=\linewidth]{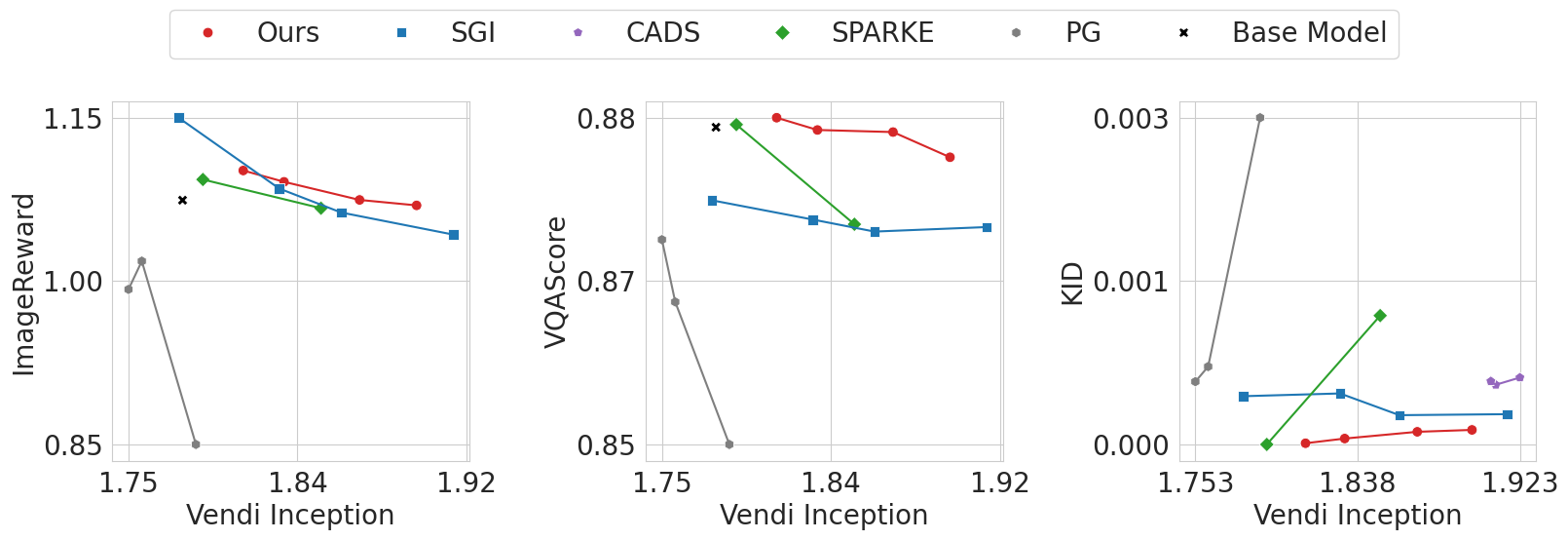}
    \caption{\textbf{Quantitative evaluation.} Pareto frontiers comparing our method against baseline methods using Flux-dev. We evaluate the trade-off between semantic diversity (Vendi Score) and three performance axes: (Left) Human Preference [ImageReward $\uparrow$], (Middle) Prompt Alignment [VQAScore $\uparrow$], and (Right) Distributional Fidelity [KID $\downarrow$]. Our method (red) achieves a superior frontier across all metrics.}
    \label{fig:flux_quant}
\end{figure}

We evaluated our method using 1,000 prompts sampled from the MS-COCO 2017 validation set, generating four images per prompt for a total of 4,000 images per configuration. To provide a holistic view of the diversity-quality trade-off, we utilize the Vendi Inception Score~\cite{friedman2022vendi, szegedy2017inception} to measure high-level semantic diversity %
alongside three primary quality and alignment axes: ImageReward~\cite{xu2023imagereward} for human preference, VQAScore~\cite{lin2024evaluating} for fine-grained prompt adherence, and Kernel Inception Distance (KID)~\cite{binkowski2018demystifying} for distributional fidelity.
By plotting the Pareto frontier
of the diversity score versus each of these
metrics, we can analyze how effectively each method navigates the tension between generative variety and visual fidelity.

To map the Pareto frontiers, we systematically vary the control hyperparameters for each baseline: the guidance scale for PG and SPARKE, the noise intensity for CADS, and the number of initial noise candidates for SGI. Specific hyperparameter configurations are provided in Appendix~\ref{sec:implementation_details}.

As shown in Figure~\ref{fig:flux_quant}, our method achieves a superior trade-off on Flux-dev. Notably, while our method exceeds the performance of SGI, the strongest baseline, it does so with drastically lower computational overhead (see Table~\ref{tab:runtimes}).
Results for additional models, including SD3.5-Turbo and SD3.5-Large, are provided in Appendix~\ref{sec:additional_quntitative_results}.

\paragraph{Runtime.} %
\label{para:runtime}

Many existing diversity methods rely on costly downstream signals, either through gradient-based optimization or by selecting from large pools of candidate latents. Both strategies impose substantial time overhead. By avoiding these mechanisms entirely, our approach provides a markedly more efficient solution, increasing runtime by only 20\%–30\% relative to the base model (Table~\ref{tab:runtimes}).

\begin{table}[t]
    \centering
    \small
    \caption{\textbf{Runtime comparison for generating a group of four images.} Our method provides a significant speedup over optimization-based diversity methods like SGI while maintaining a low overhead relative to the base model.}
    \label{tab:runtimes}
    \begin{tabular}{llccc}
        \toprule
        \multicolumn{2}{l}{Method} & SD3.5-Large & SD3.5-Turbo & Flux-dev \\
        \midrule
        \multicolumn{2}{l}{Base Model} & 13.83s & 4.18s & 10.34s \\
        \multicolumn{2}{l}{\textbf{Ours (Contextual)}} & \textbf{18.12s} & \textbf{5.52s} & \textbf{12.80s} \\
        \multirow{4}{*}{SGI} & 8 Candidates & 66.79s & 13.15s & 47.47s \\
                             & 16 Candidates & 76.79s & 23.73s & 56.32s \\
                             & 32 Candidates & 101.44s & 46.15s & 75.39s \\
                             & 64 Candidates & 145.14s & 91.30s & 113.99s \\
        \bottomrule
    \end{tabular}
\end{table}

\paragraph{User study.}

\begin{figure}[ t ]
    \includegraphics[width=1\linewidth]{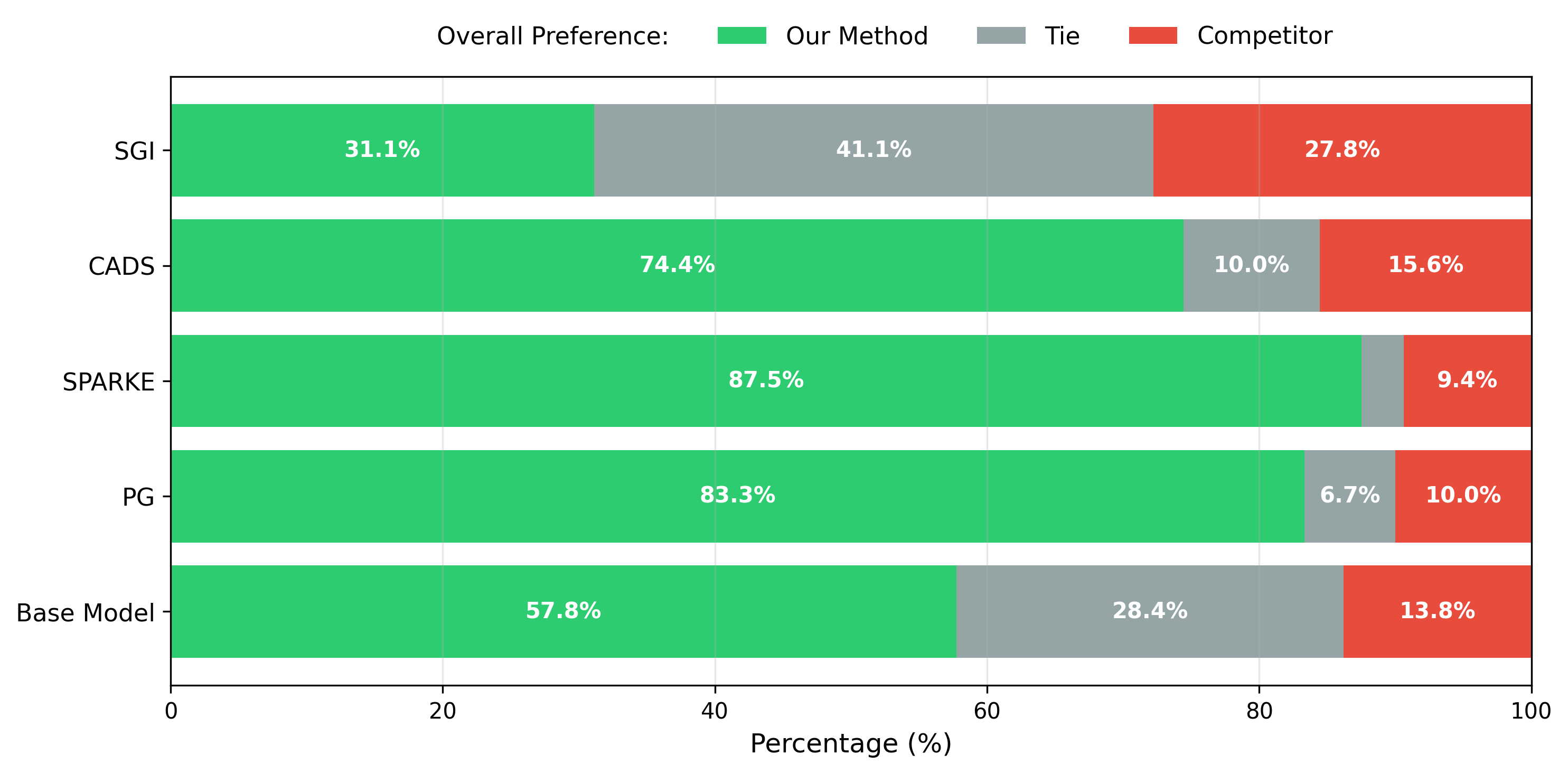}
    \caption{{\textbf{Overall user preference comparison.}
    Distribution of user choices
    comparing our method with five competing approaches.
    Bars indicate the percentage of cases in which users preferred our results (green), preferred competing methods (red), or rated both equally (gray).
    }}
    \label{fig:user_study}
\end{figure}

Standard quantitative metrics often fail to capture the nuances of generative diversity. These evaluators are typically trained on datasets dominated by common visual patterns, leading them to favor ``typical'' or average cases as more aesthetically pleasing or prompt-adherent. Consequently, methods that successfully push for greater diversity and creative interpretation may be unfairly penalized by these metrics, even when the resulting variations are highly desirable to human users. To address this limitation and provide a more meaningful assessment of our method, we conducted a user study.

We utilized ChatGPT to generate 40 diverse prompts across various categories. For each prompt, participants were presented with two batches of 8 images (16 images total): one batch generated by our method and the other by a competing method or the base model (Flux-dev). Participants were tasked with performing a side-by-side comparison to determine which batch: (i) Exhibited greater visual and semantic diversity; (ii) Maintained higher image quality; (iii) Demonstrated better prompt adherence; and (iv) Was preferred overall.

We collected 450 responses from 45 participants. Figure~\ref{fig:user_study} reports the overall user preference results of this study, with the full preference table provided in Appendix~\ref{par:user_study_table}. Overall, our method achieves higher user preference than all competing approaches. The only exception is SGI, where preferences are closely matched, with a slight advantage for our method. Importantly, these gains are achieved with minimal runtime overhead, as demonstrated in Table \ref{tab:runtimes}. 

\subsection{Ablation Studies}

We evaluate the impact of the repulsion scale and the specific representation space used for intervention below, with further hyperparameter analyses provided in Appendix~\ref{sec:more_ablations}.

\paragraph{Repulsion scale ablation.}

\begin{figure}[t]

    \centering
    \setlength{\tabcolsep}{0.5pt} \renewcommand{\arraystretch}{0.5} 
    \newcommand{\imgwidth}{
    0.09
    \textwidth}
    \newcommand{\vertlabel}[1]{%
      \raisebox{1.5em}{\rotatebox{90}{\scriptsize\textbf{#1}}}%
    }
    \begin{tabular}{c c c c c c}
        
        \vertlabel{$\eta=0$} & 
        \includegraphics[width=\imgwidth, height=\imgwidth]{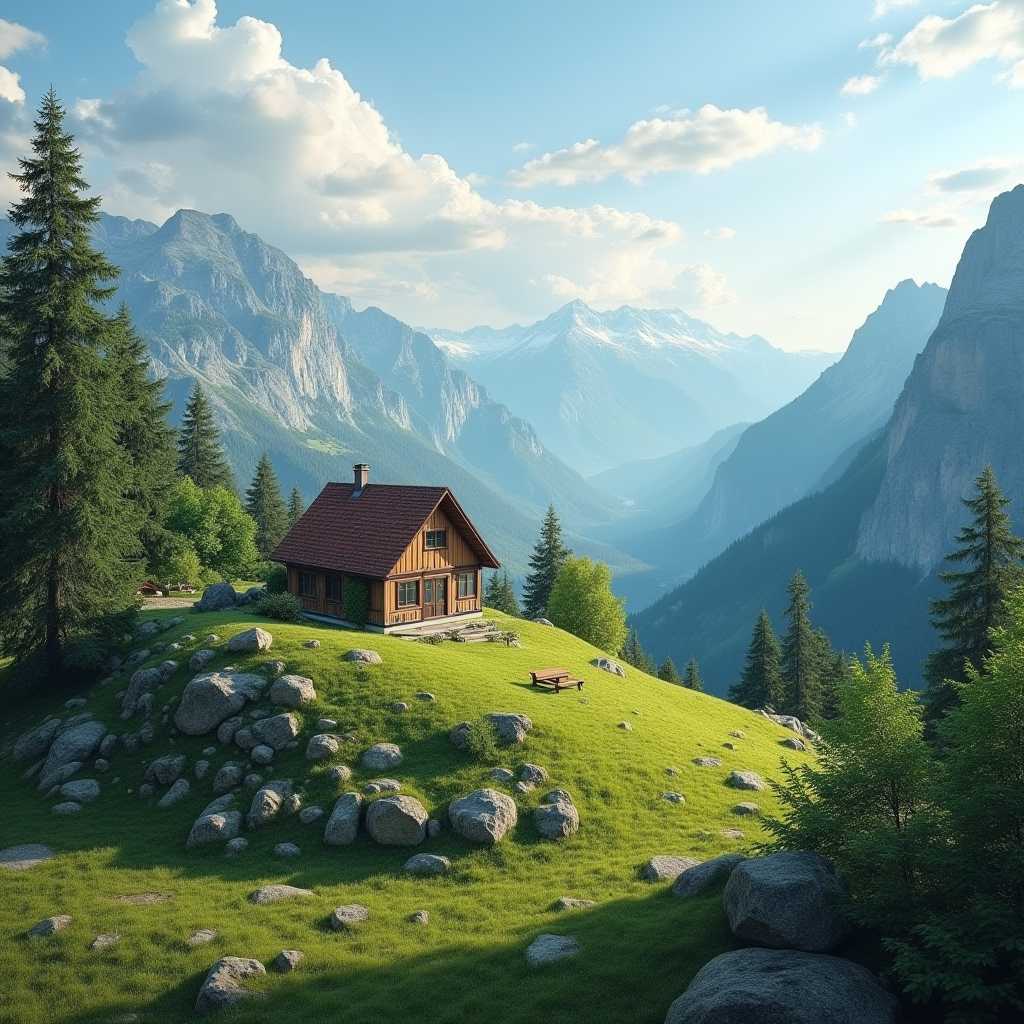} &
        \includegraphics[width=\imgwidth, height=\imgwidth]{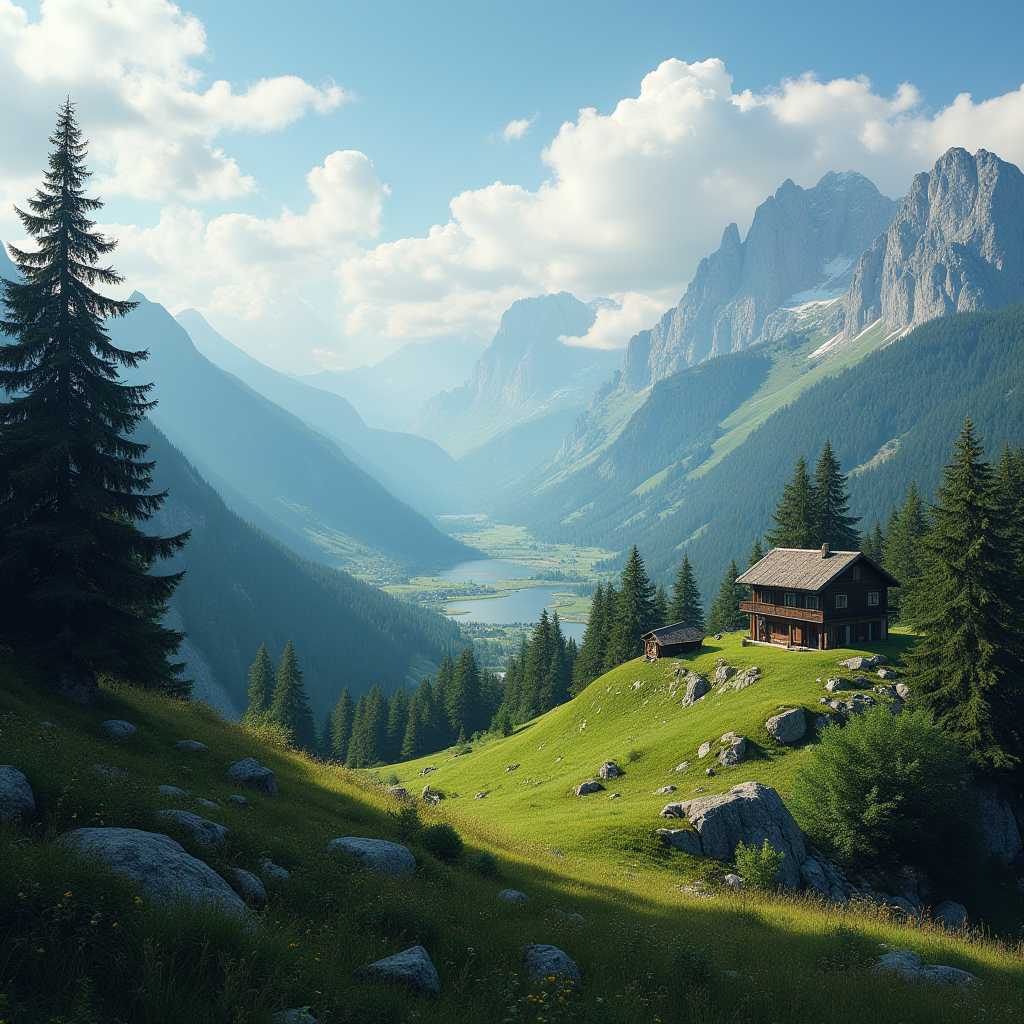} &
        \includegraphics[width=\imgwidth, height=\imgwidth]{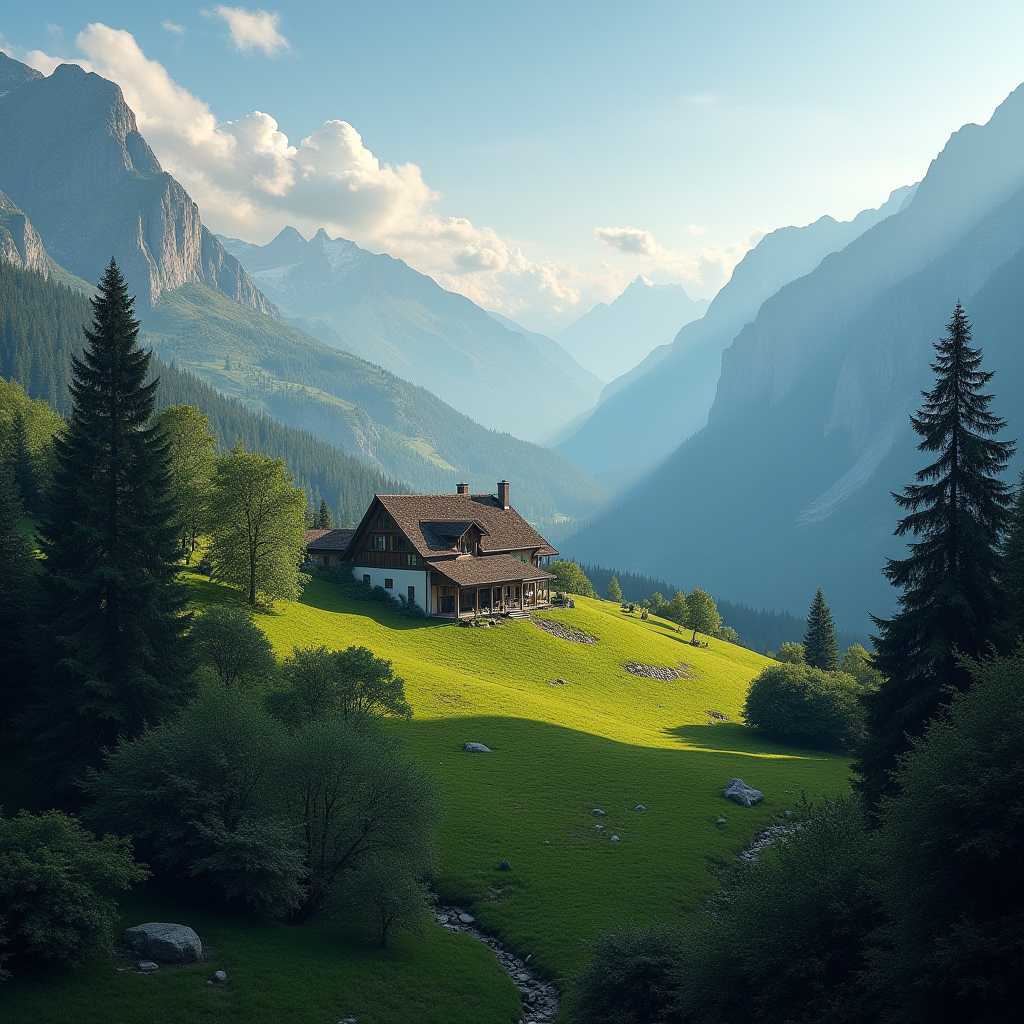} &
        \includegraphics[width=\imgwidth, height=\imgwidth]{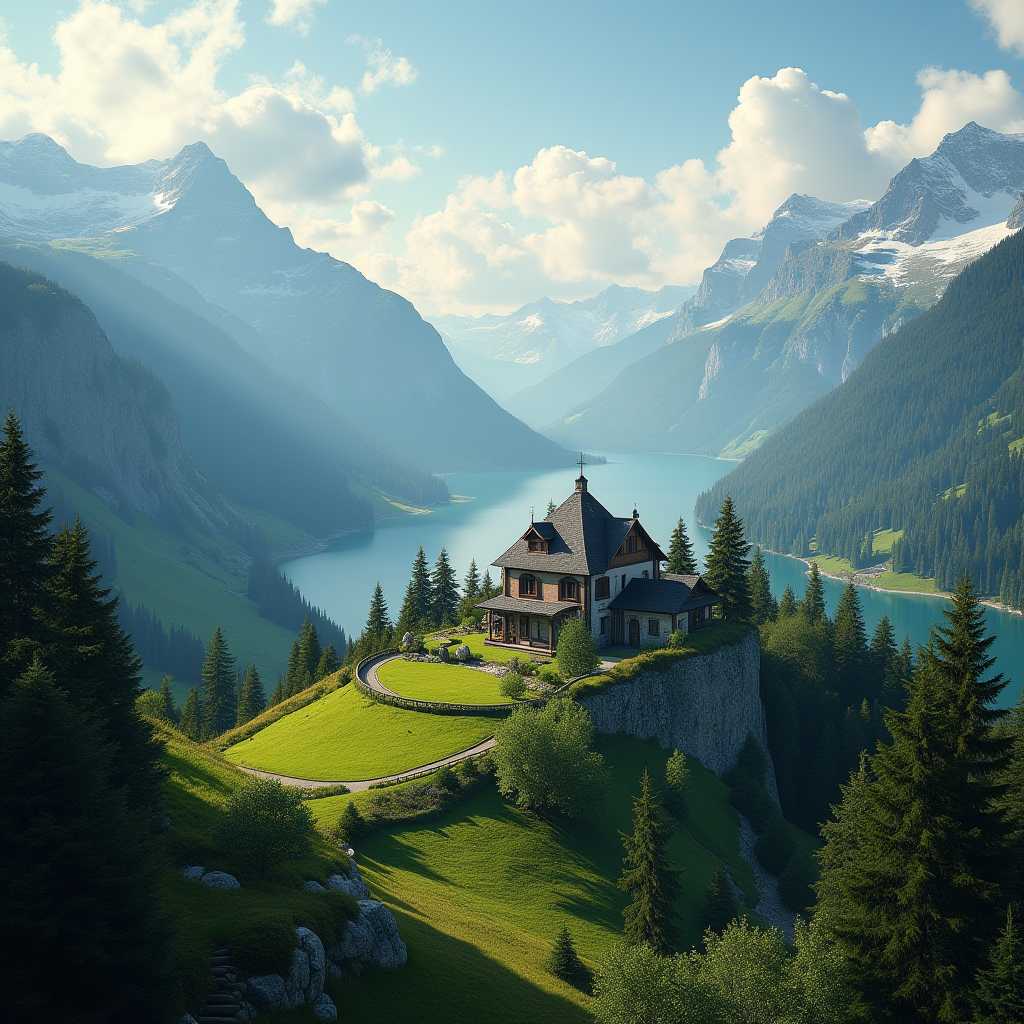} &
        \includegraphics[width=\imgwidth, height=\imgwidth]{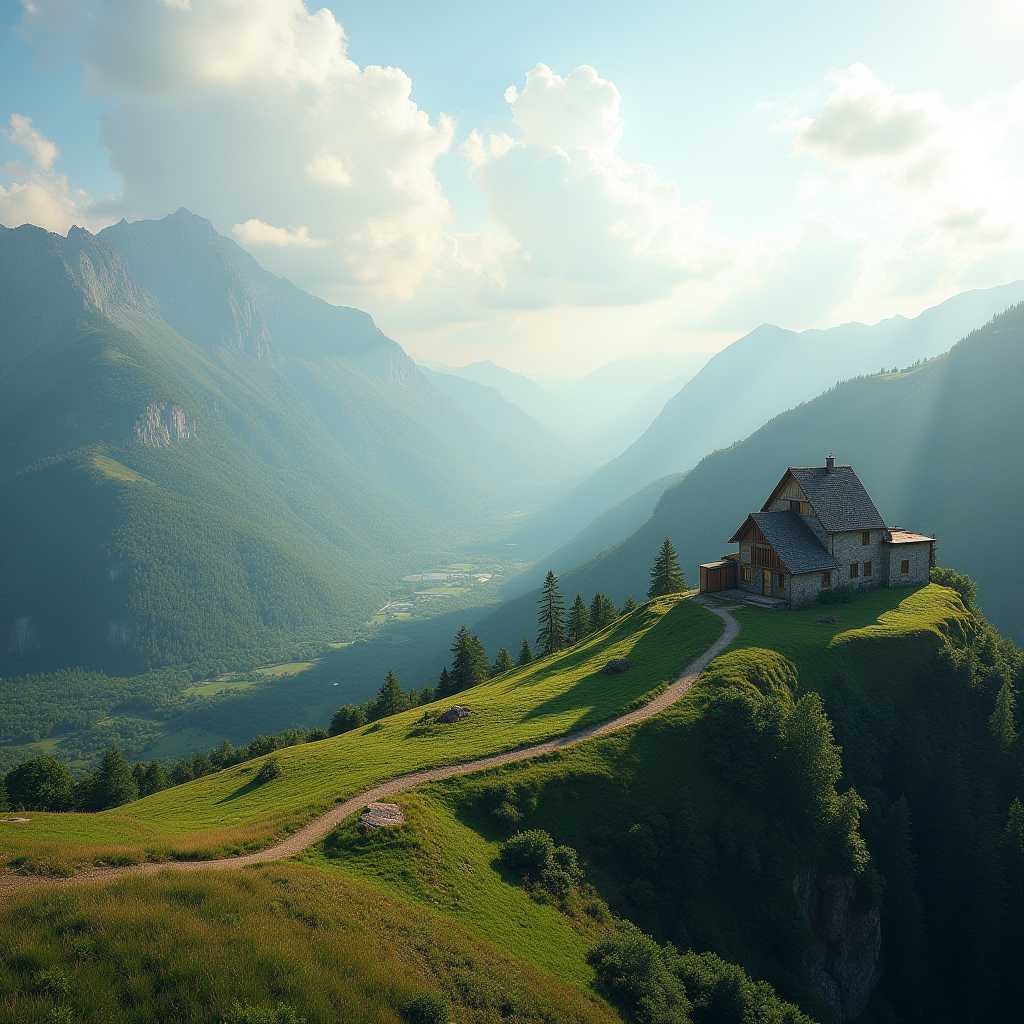} \\[-1pt]
        \vertlabel{$\eta=1e9$} & 
        \includegraphics[width=\imgwidth, height=\imgwidth]{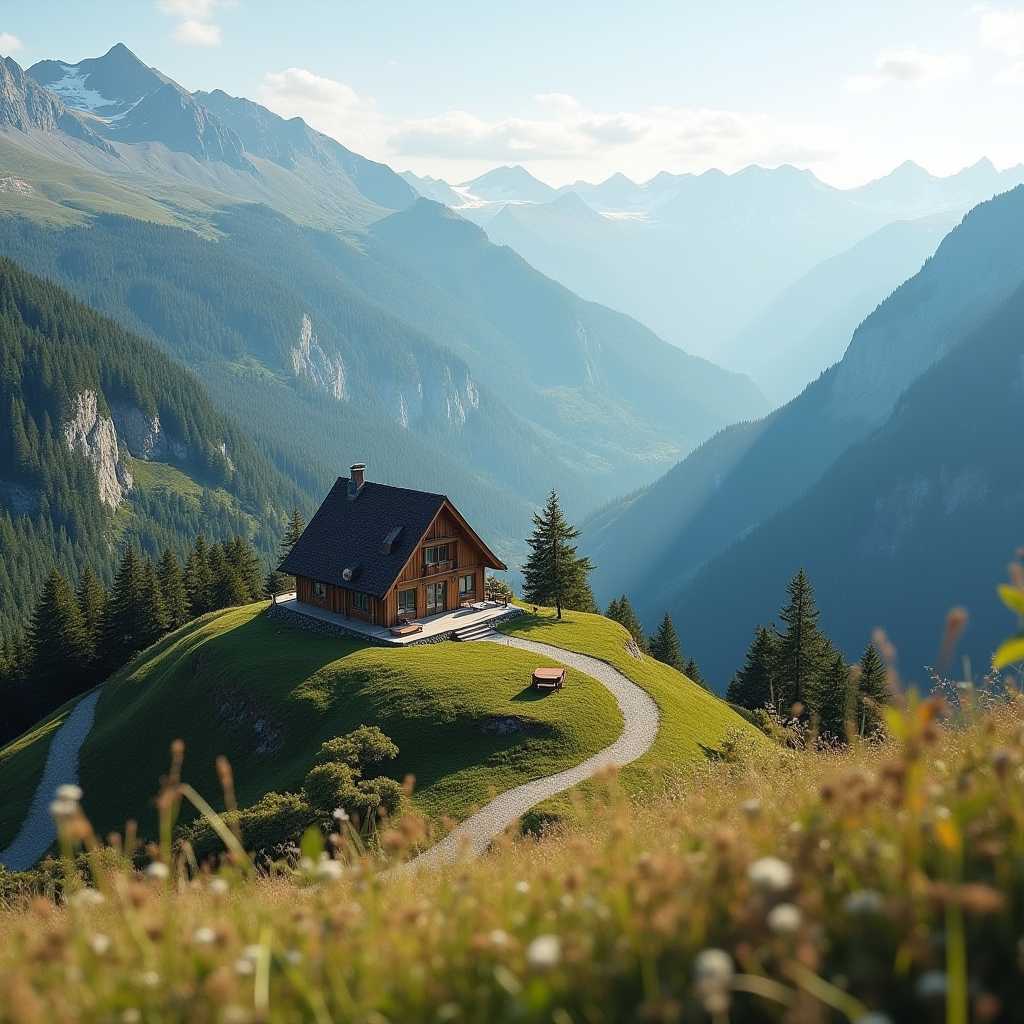} &
        \includegraphics[width=\imgwidth, height=\imgwidth]{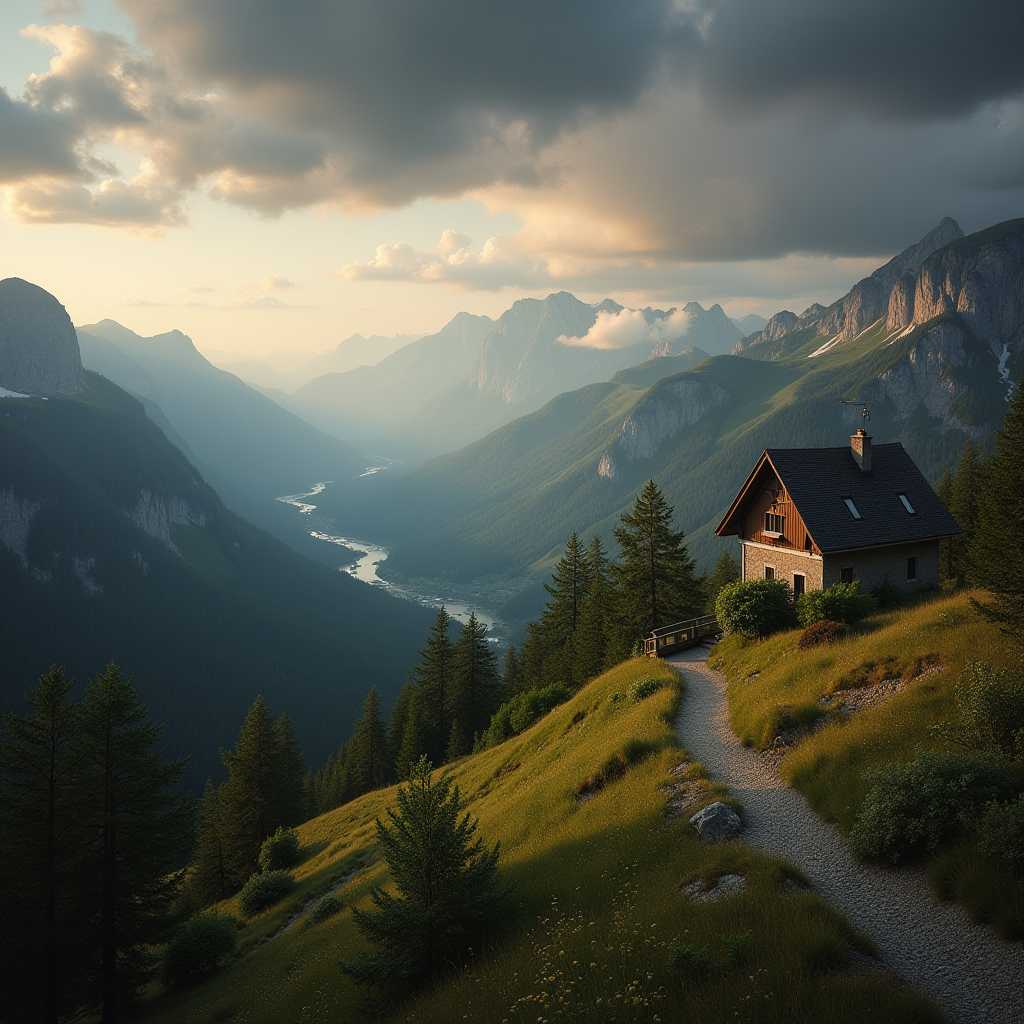} &
        \includegraphics[width=\imgwidth, height=\imgwidth]{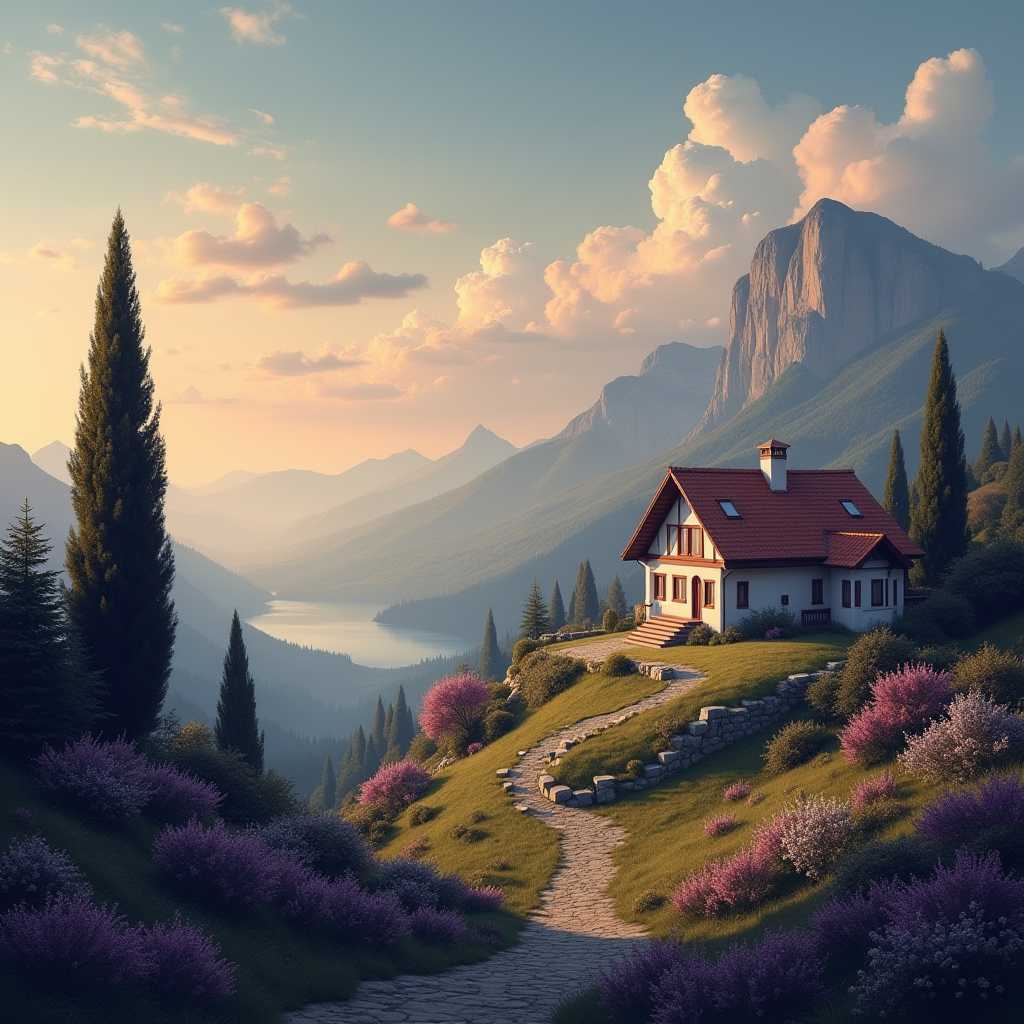} &
        \includegraphics[width=\imgwidth, height=\imgwidth]{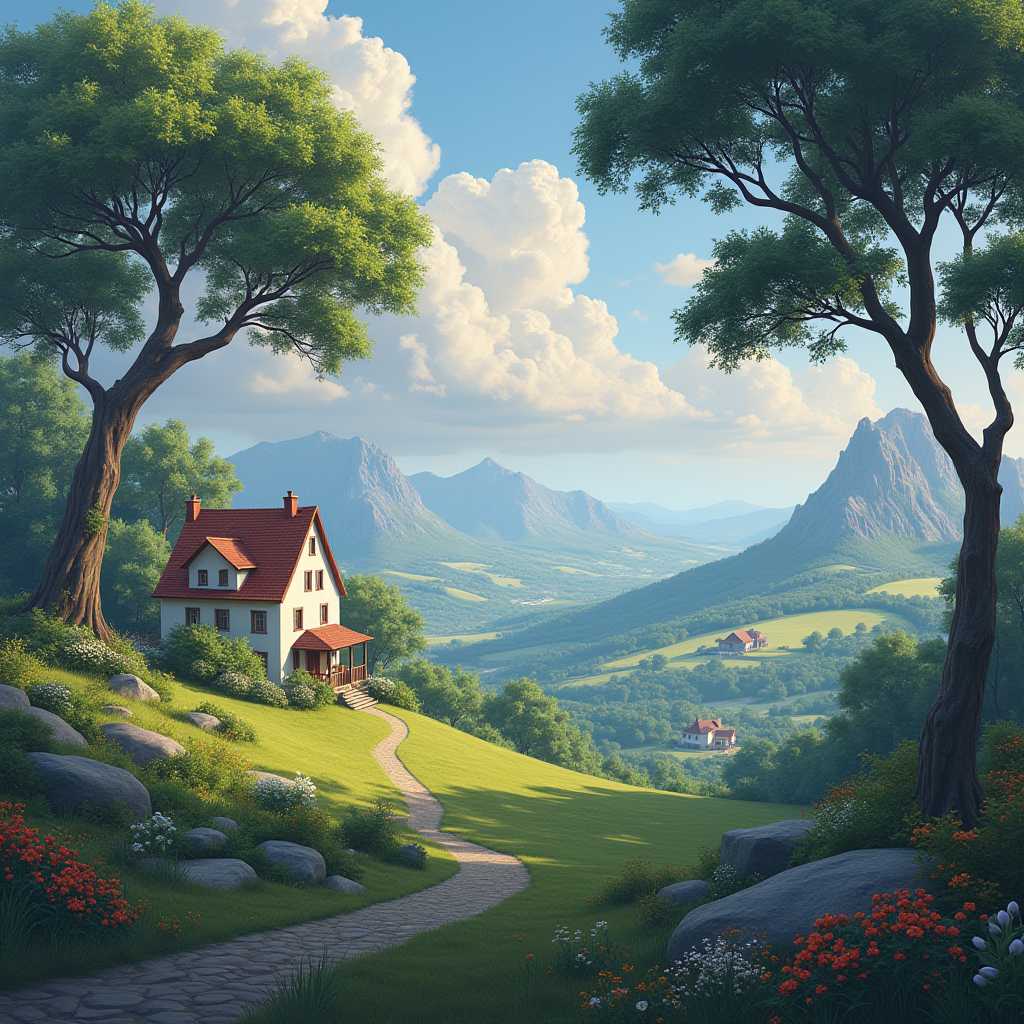} &
        \includegraphics[width=\imgwidth, height=\imgwidth]{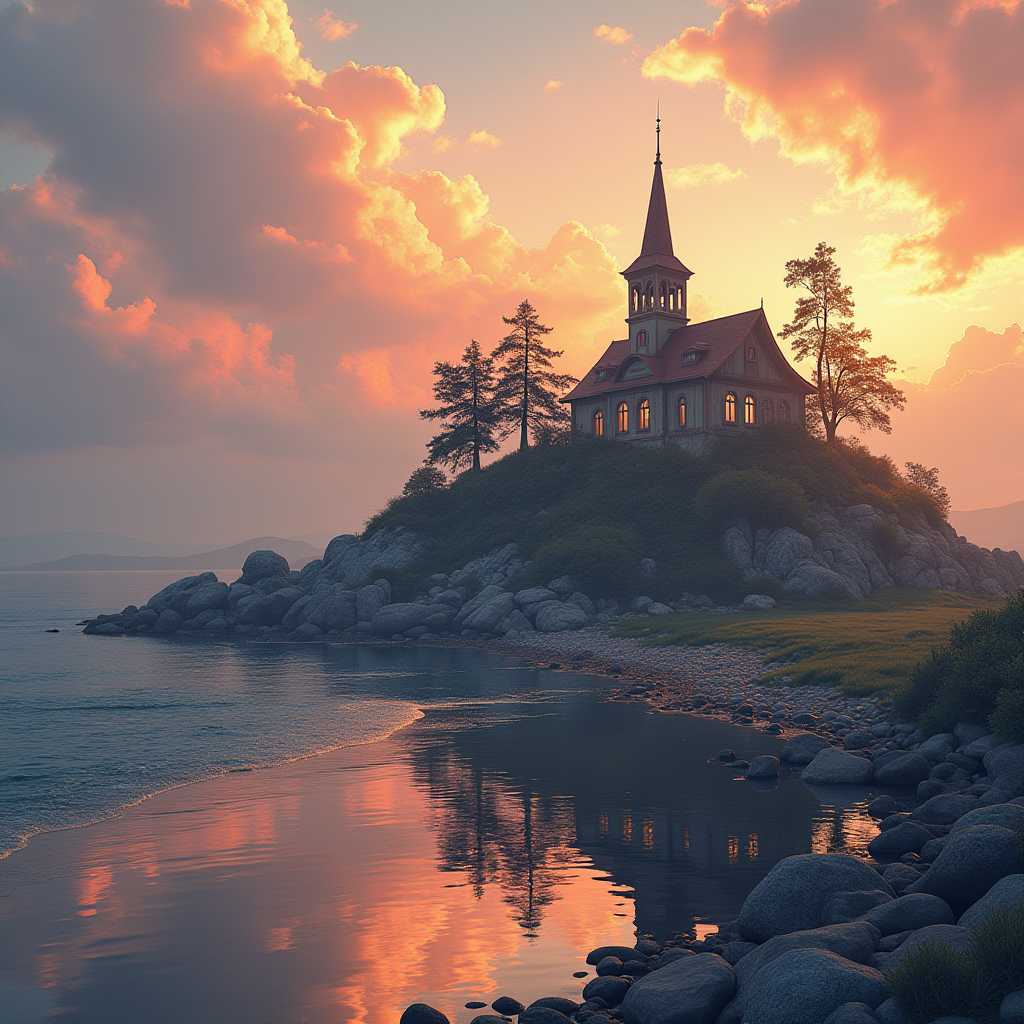} \\[-1pt]
        \vertlabel{$\eta=2e9$} & 
        \includegraphics[width=\imgwidth, height=\imgwidth]{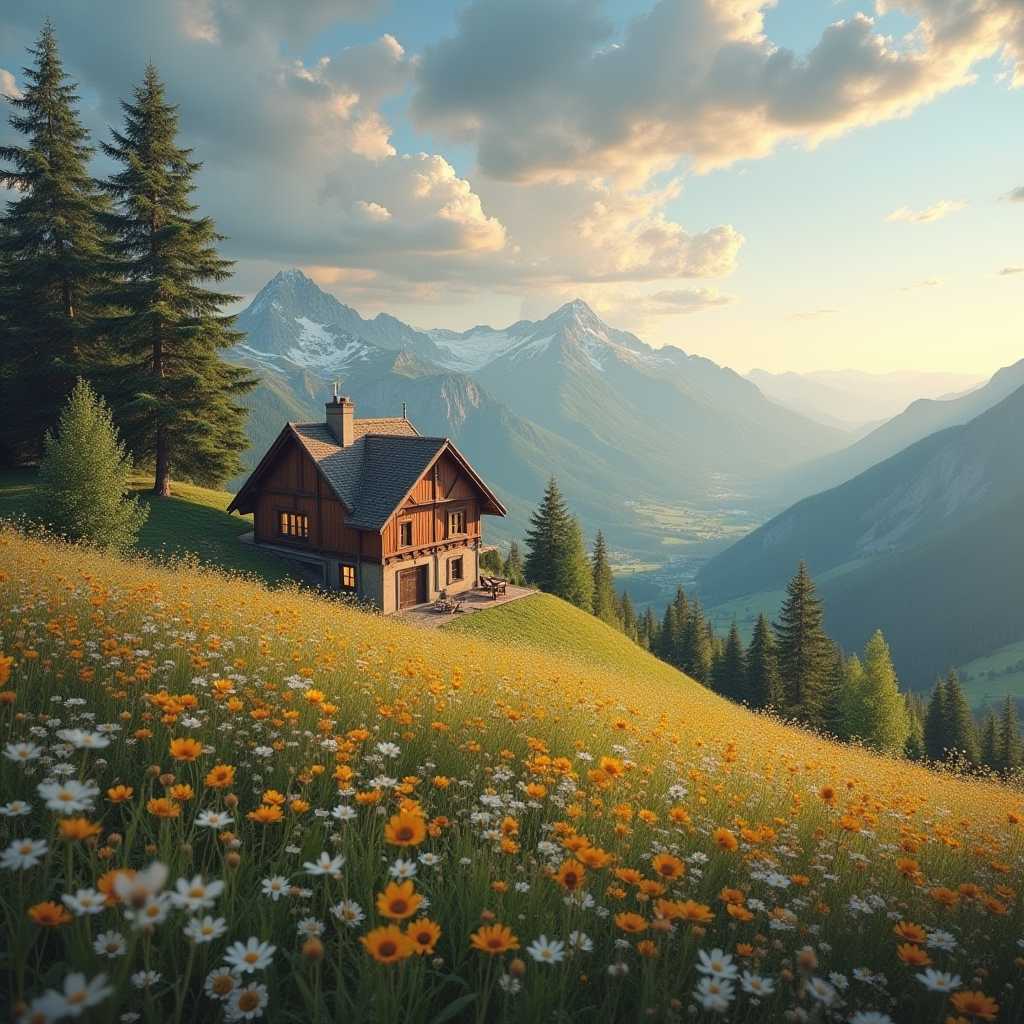} &
        \includegraphics[width=\imgwidth, height=\imgwidth]{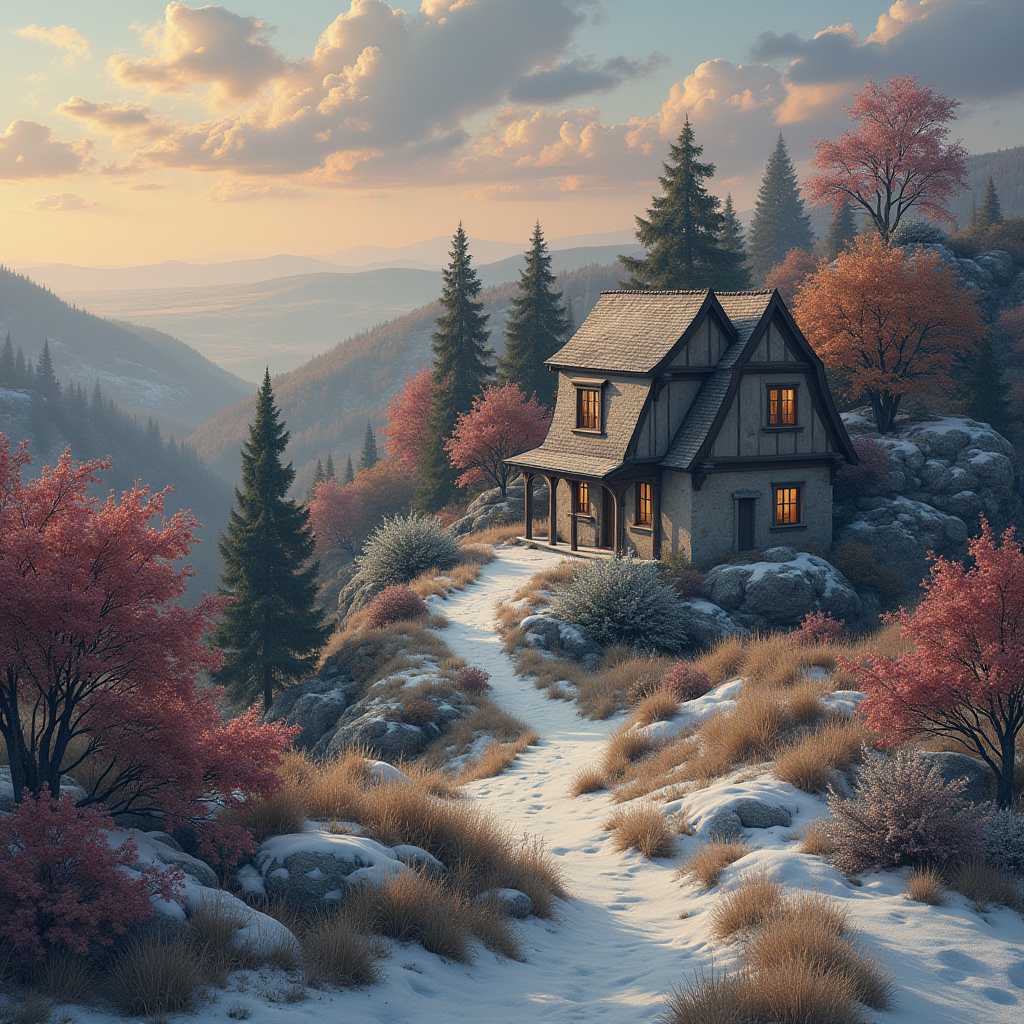} &
        \includegraphics[width=\imgwidth, height=\imgwidth]{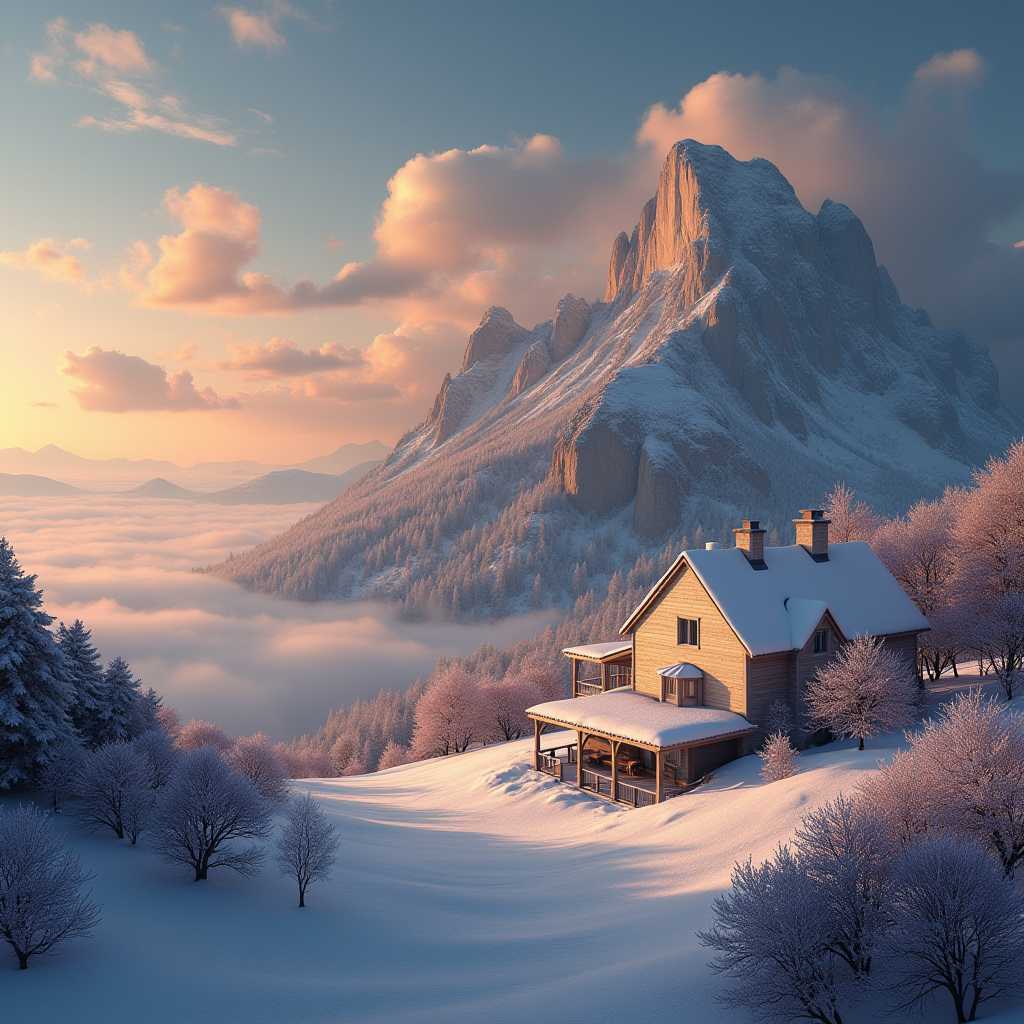} &
        \includegraphics[width=\imgwidth, height=\imgwidth]{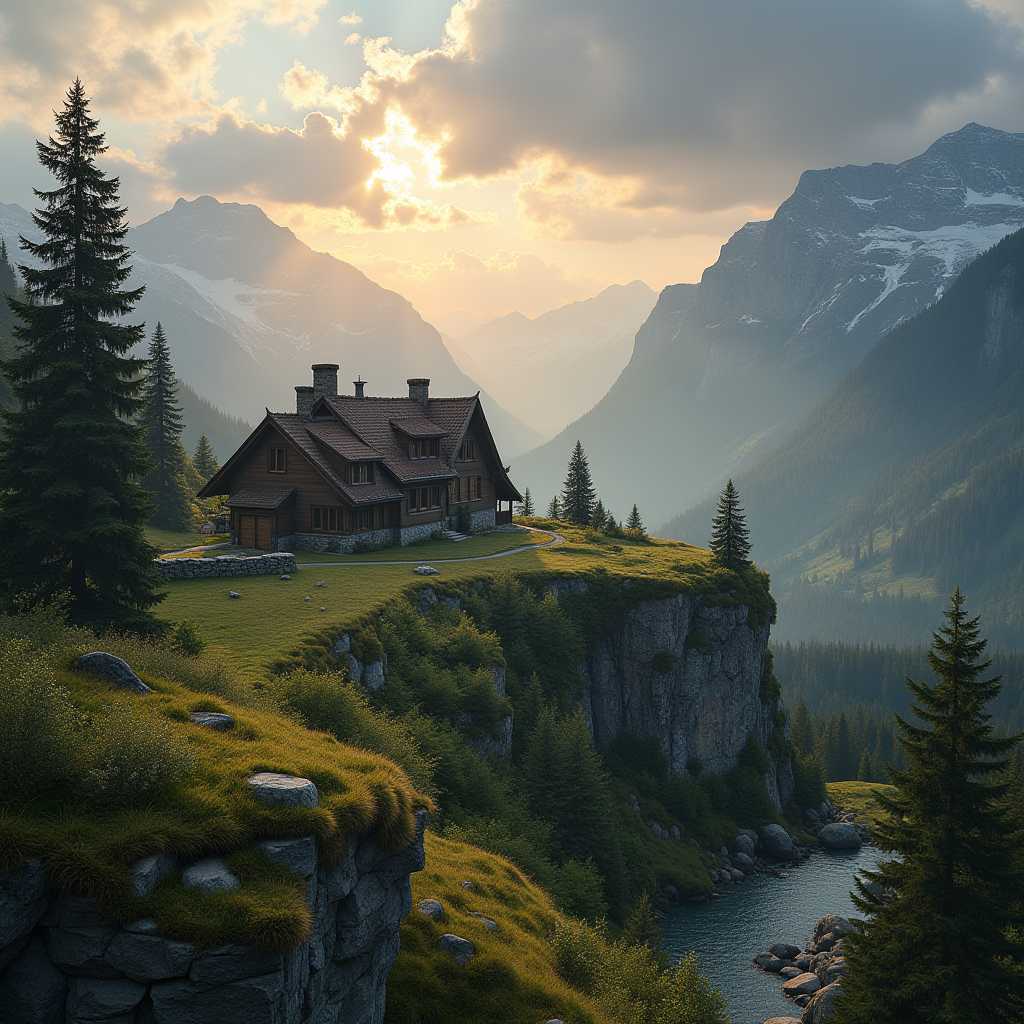} &
        \includegraphics[width=\imgwidth, height=\imgwidth]{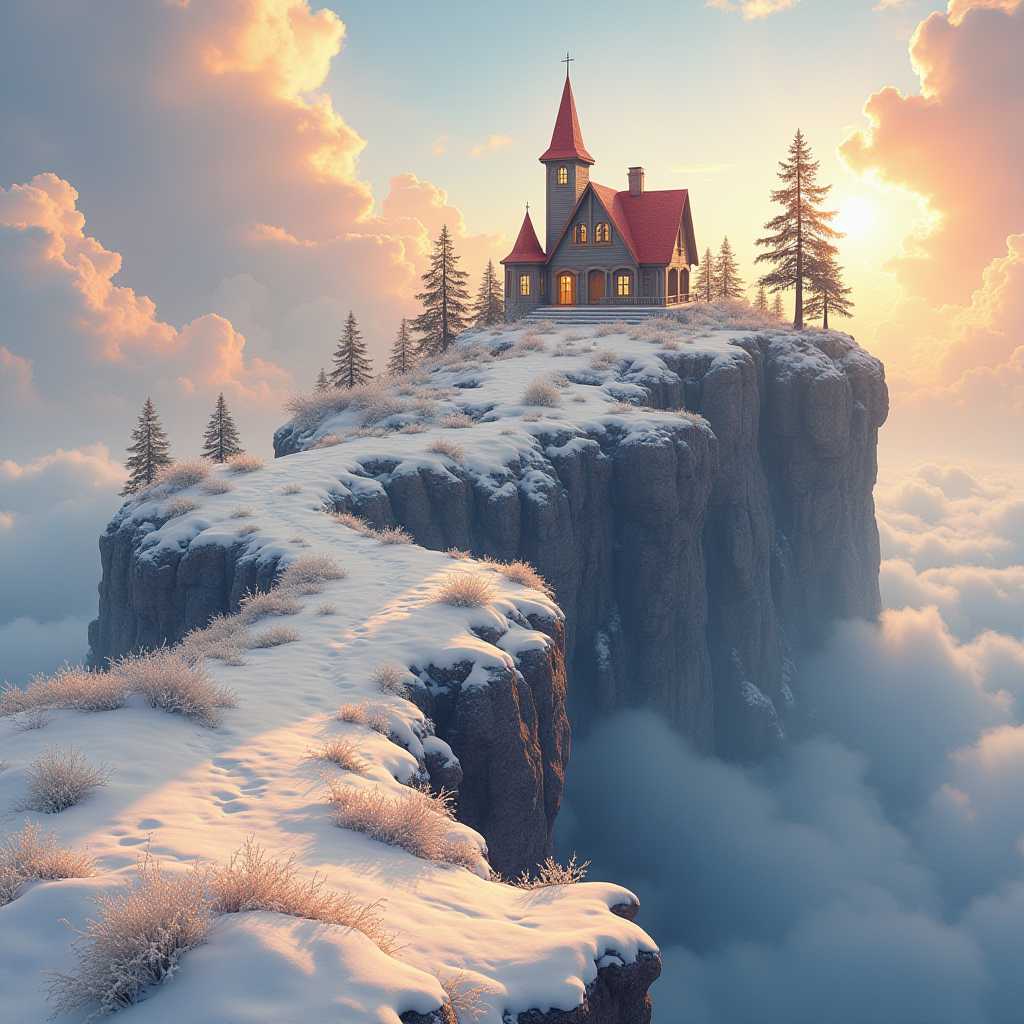} \\[-1pt]
        \vertlabel{$\eta=5e9$} & 
        \includegraphics[width=\imgwidth, height=\imgwidth]{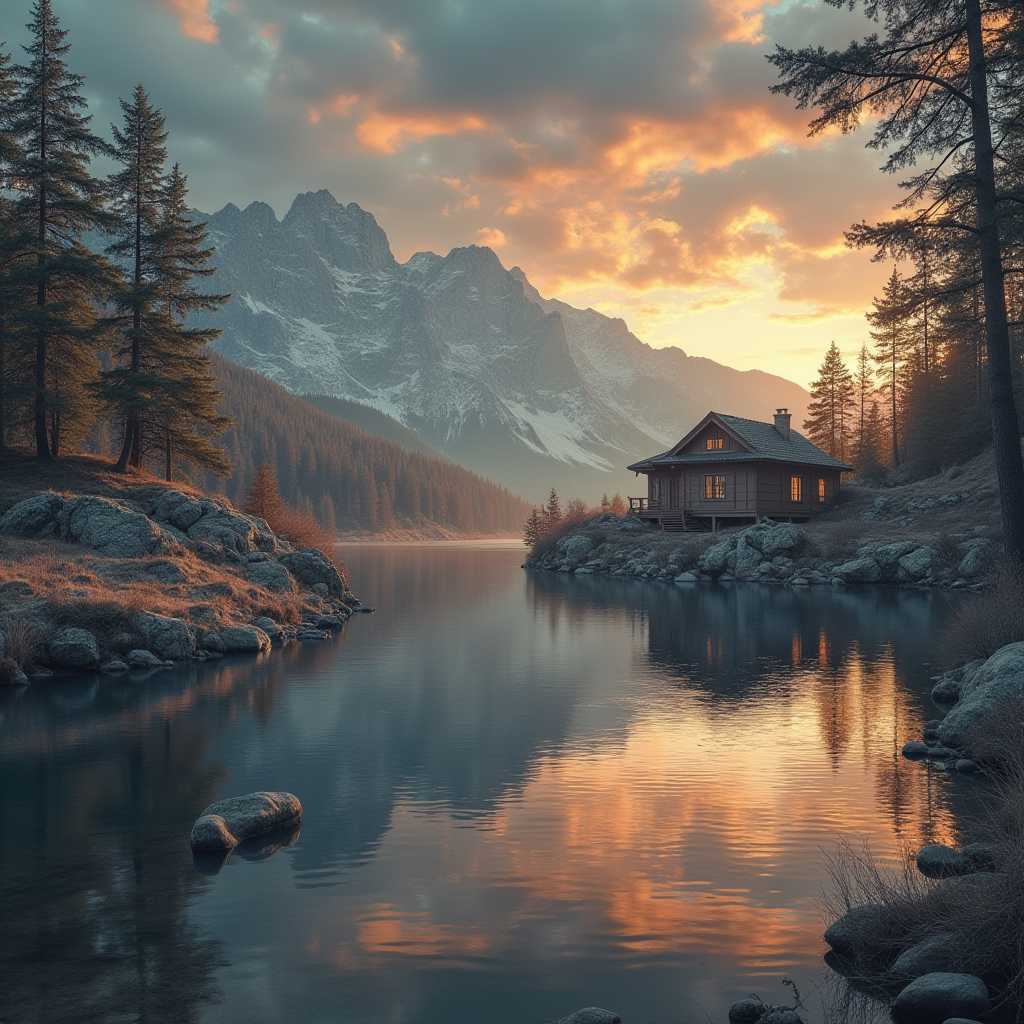} &
        \includegraphics[width=\imgwidth, height=\imgwidth]{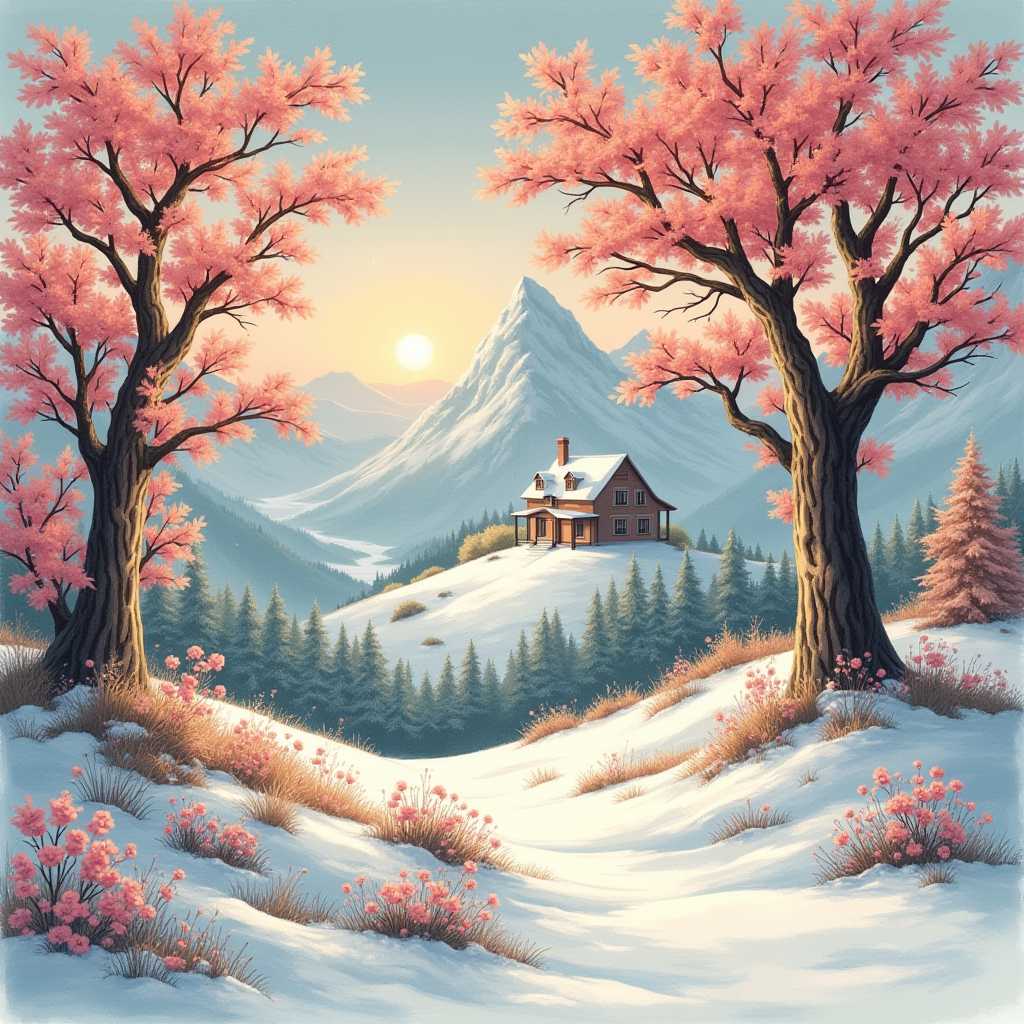} &
        \includegraphics[width=\imgwidth, height=\imgwidth]{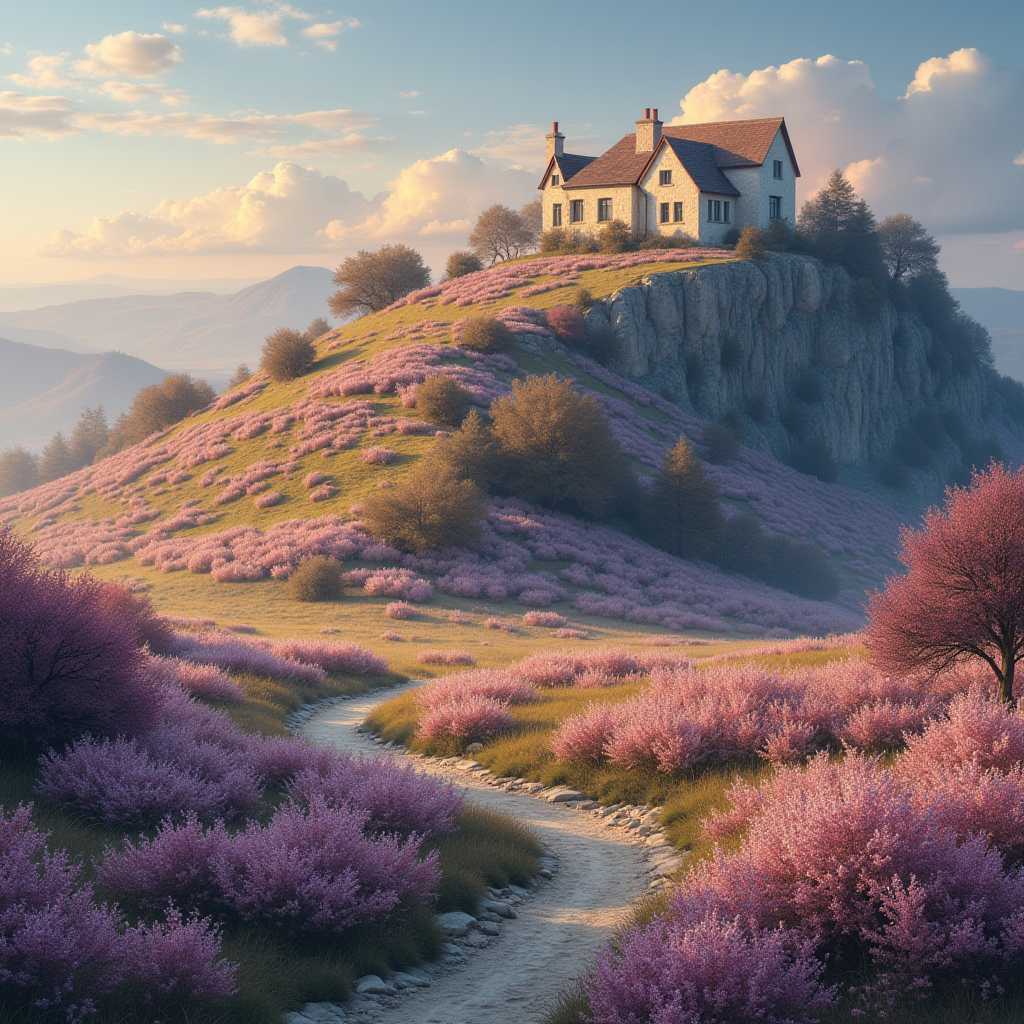} &
        \includegraphics[width=\imgwidth, height=\imgwidth]{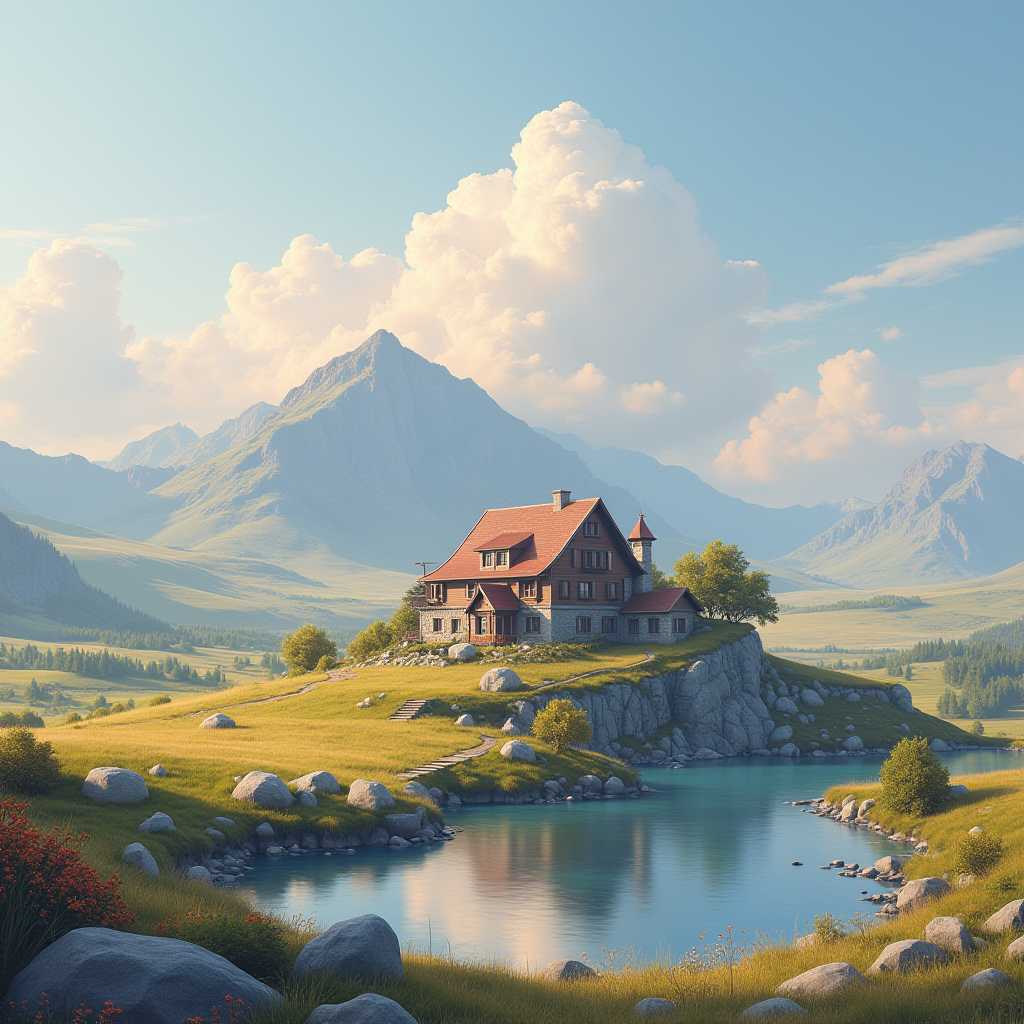} &
        \includegraphics[width=\imgwidth, height=\imgwidth]{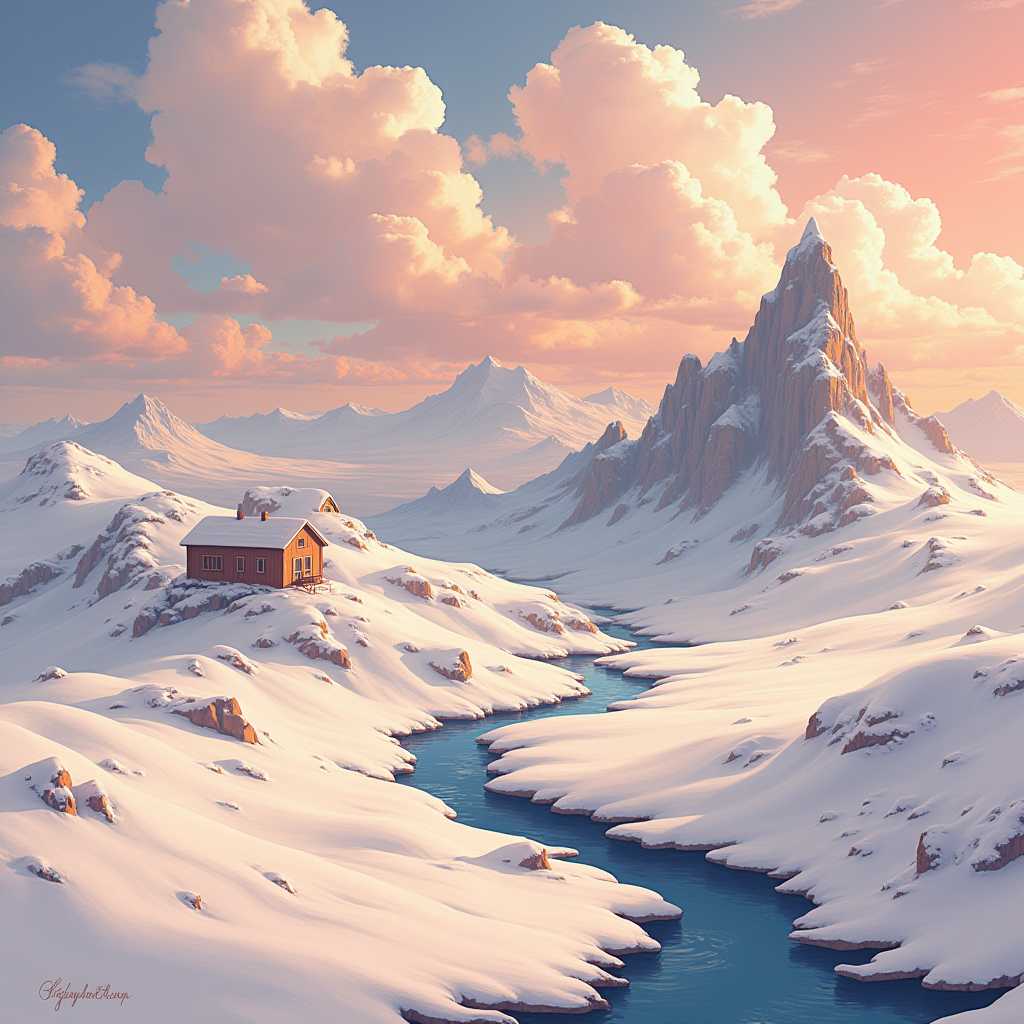} \\[-1pt]
        \vertlabel{$\eta=8e9$} & 
        \includegraphics[width=\imgwidth, height=\imgwidth]{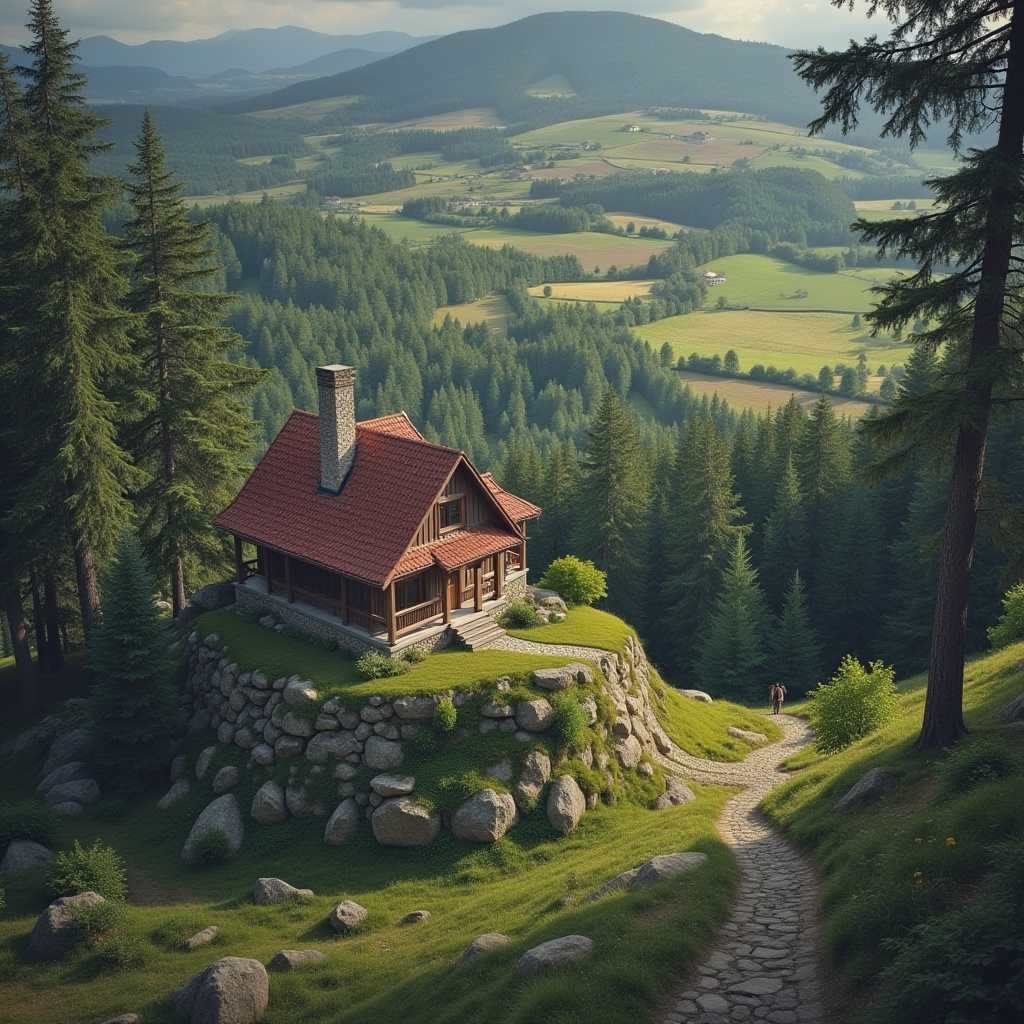} &
        \includegraphics[width=\imgwidth, height=\imgwidth]{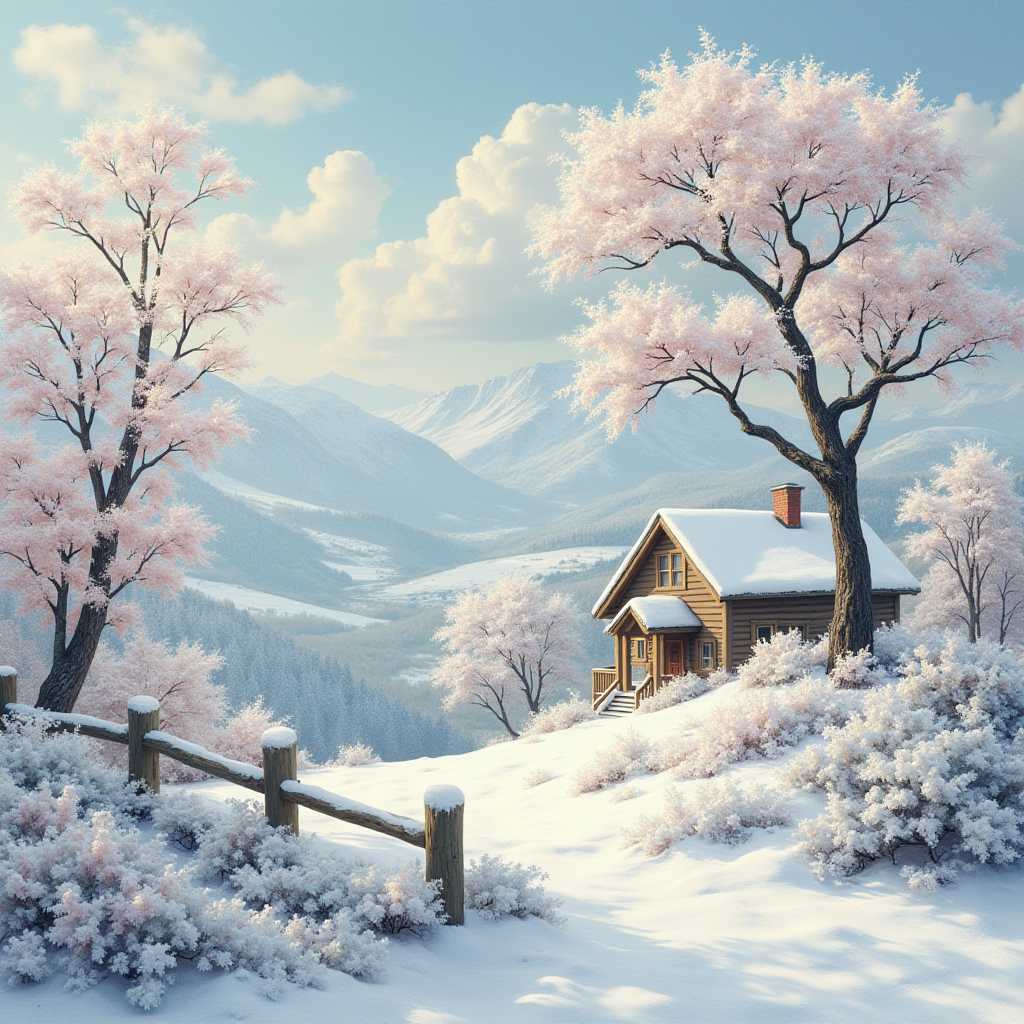} &
        \includegraphics[width=\imgwidth, height=\imgwidth]{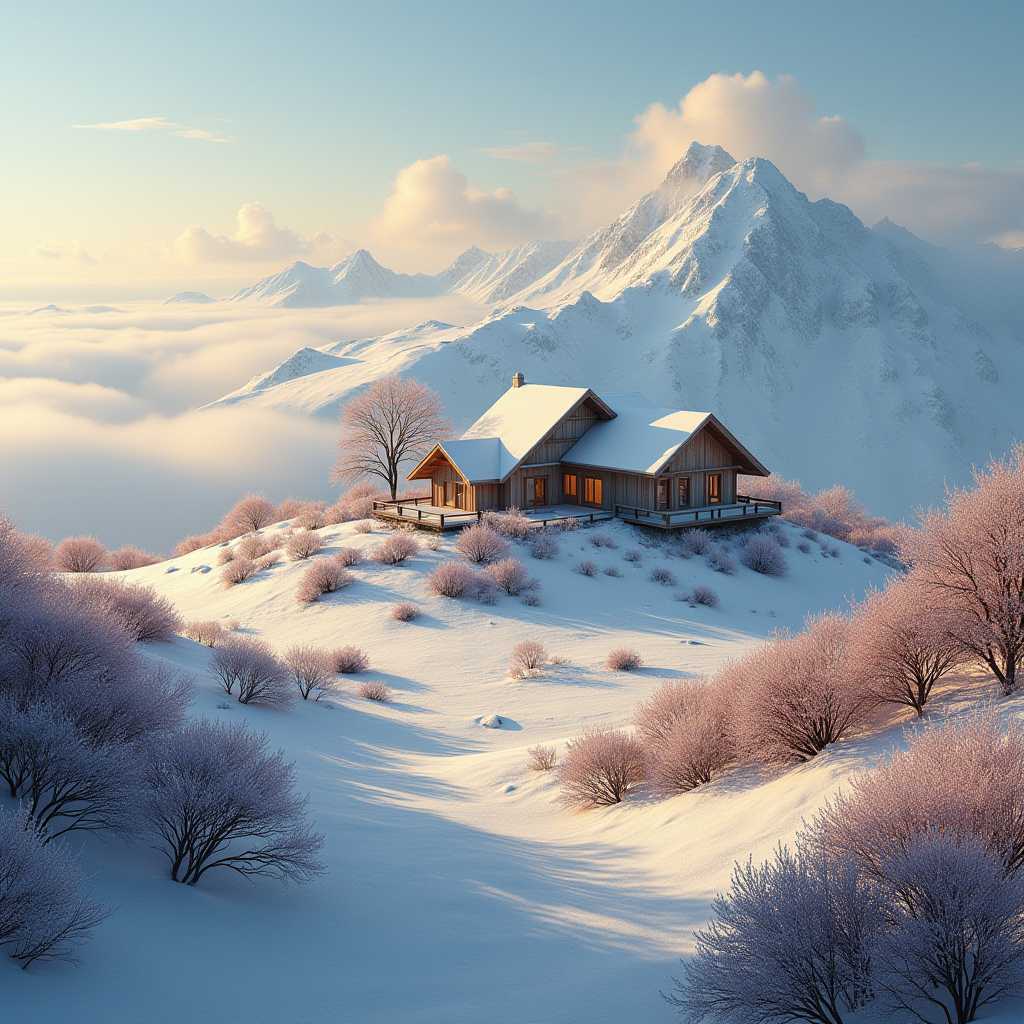} &
        \includegraphics[width=\imgwidth, height=\imgwidth]{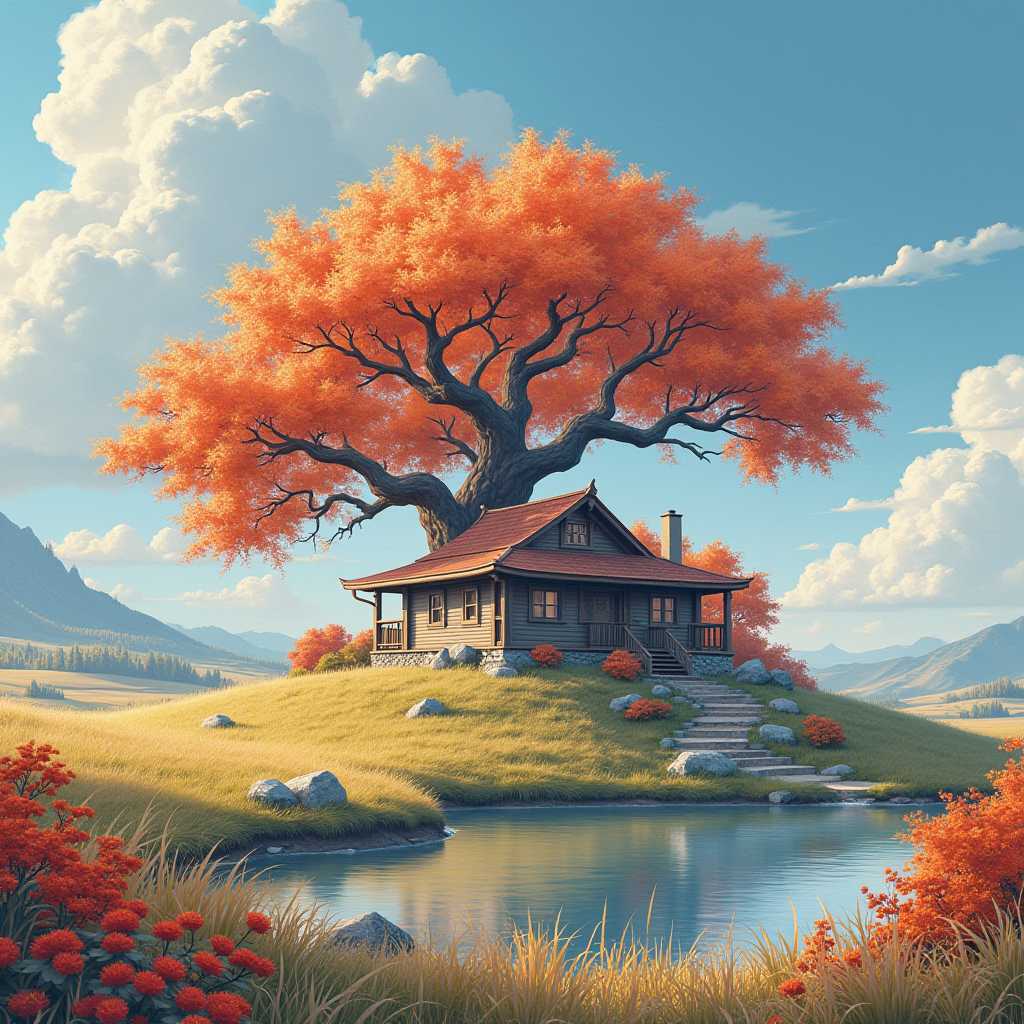} &
        \includegraphics[width=\imgwidth, height=\imgwidth]{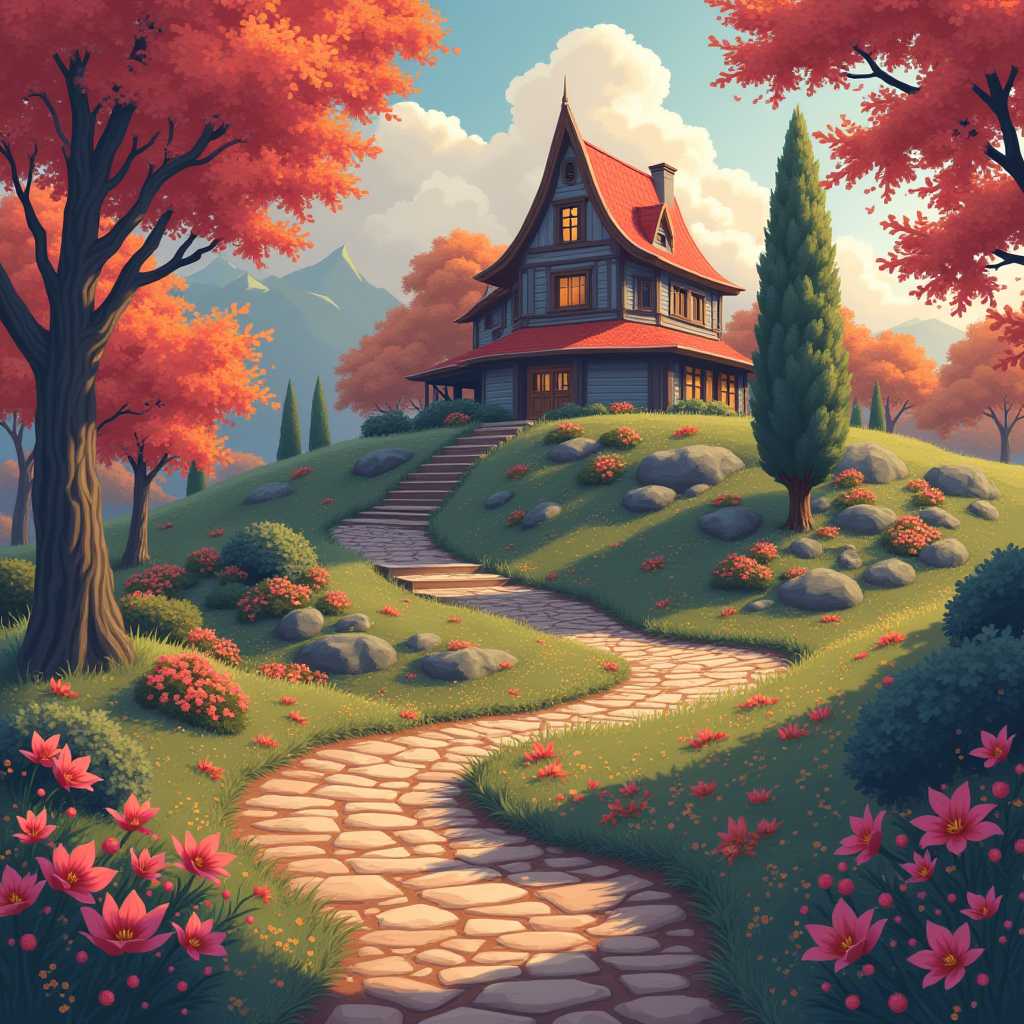} \\
        \multicolumn{6}{c}{\vspace{2pt}\small``A breathtaking view of a distant house in beautiful scenery''} \\

    \end{tabular}
    \caption{\textbf{Ablation of the repulsion scale $\eta$.} We visualize the impact of the repulsion scale on our results. At $\eta=0$ (top row), the base model exhibits low diversity, producing similar architectural styles and environments across multiple seeds. As $\eta$ increases, our Contextual Space repulsion introduces progressively larger variations, while maintaining high image quality and prompt alignment.}
    \label{fig:ablation_scale}

\end{figure}

In Figure~\ref{fig:ablation_scale}, we ablate the effect of the repulsion scale $\eta$. The top row ($\eta=0$) represents the base Flux-dev generations, which exhibit a narrow interpretation of the prompt; each image displays a similar-looking house in nearly identical environments. In each subsequent row, we show the results of our method with an increasing repulsion scale. As can be seen, higher values of $\eta$  generally yield greater diversity, introducing structural changes like adding a tower to the house, altering the landscape with a lake, or shifting the scene's season.

\paragraph{Repulsion space ablation.}

\begin{figure*}[t]
    \centering
    \scriptsize
    \includegraphics[width=\linewidth]{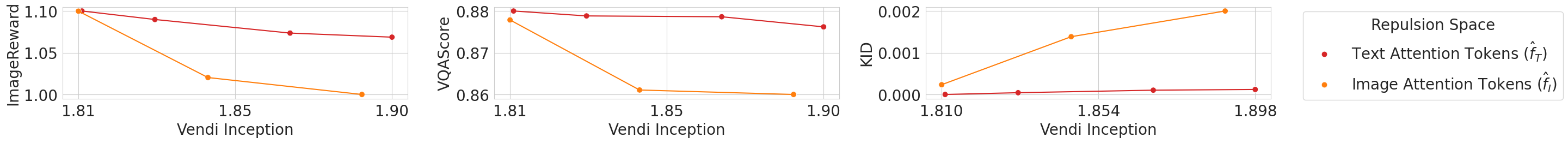}
    \caption{\textbf{Ablation of Repulsion Space.} Pareto frontiers comparing repulsion applied to text attention tokens (Contextual Space, $\hat{f}_T$) versus image attention tokens ($\hat{f}_I$) within the Flux-dev architecture. We evaluate the trade-off between semantic diversity (Vendi Score) and three performance axes: (Left) Human Preference [ImageReward $\uparrow$], (Middle) Prompt Alignment [VQAScore $\uparrow$], and (Right) Distributional Fidelity [KID $\downarrow$]. Our method (red) achieves a superior frontier across all metrics.}
    \label{fig:token_ablation}
\end{figure*}

To isolate the efficacy of intervening in the Contextual Space ($\hat{f}_T$), we compare our framework against an identical repulsion mechanism applied instead to the image attention tokens ($\hat{f}_I$) within the multimodal blocks (i.e., the dual-stream blocks in Flux). As illustrated in Figure~\ref{fig:token_ablation}, repulsion in the Contextual Space produces a significantly more robust Pareto frontier, yielding superior human preference (ImageReward), distributional fidelity (KID), and prompt alignment (VQAScore). Notably, while the image-token baseline exhibits sharp performance degradation as diversity increases, our method maintains a shallower decline across all metrics. This suggests that the Contextual Space is better suited for navigating semantic diversity while strictly preserving the integrity of samples within the learned conditional manifold.

\begin{figure}[h]
    \centering
    \setlength{\tabcolsep}{0.5pt} \renewcommand{\arraystretch}{0.5} %
    \newcommand{\imgwidth}{%
        0.11\textwidth
    }
    \newcommand{\labelImage}{%
        \raisebox{2.5em}{%
            \rotatebox{90}{\tiny\textbf{Image}}%
        }%
    }
    \newcommand{\labelContext}{%
        \raisebox{1.8em}{%
            \rotatebox{90}{\tiny\textbf{Contextual}}%
        }%
    }
    \begin{tabular}{c c c c c}
        \labelImage & \includegraphics[width=\imgwidth]{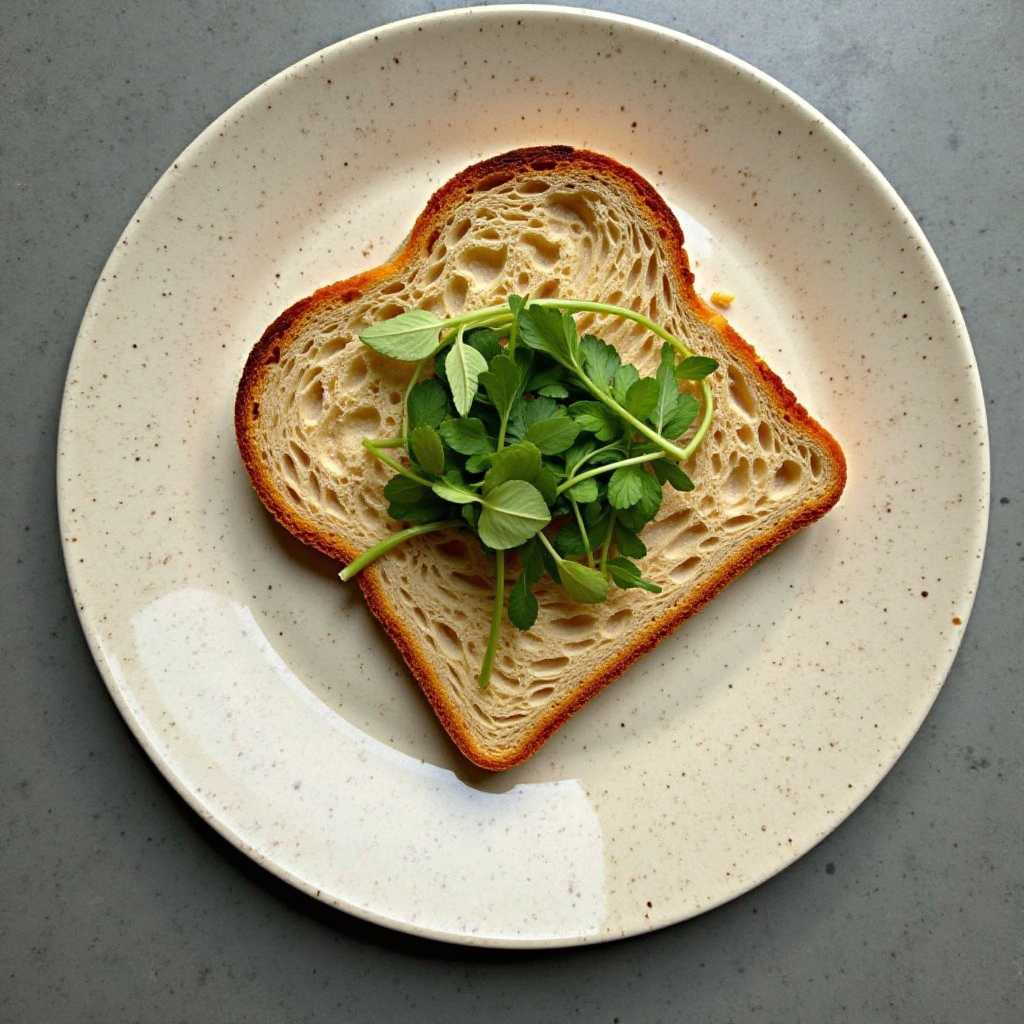} & \includegraphics[width=\imgwidth]{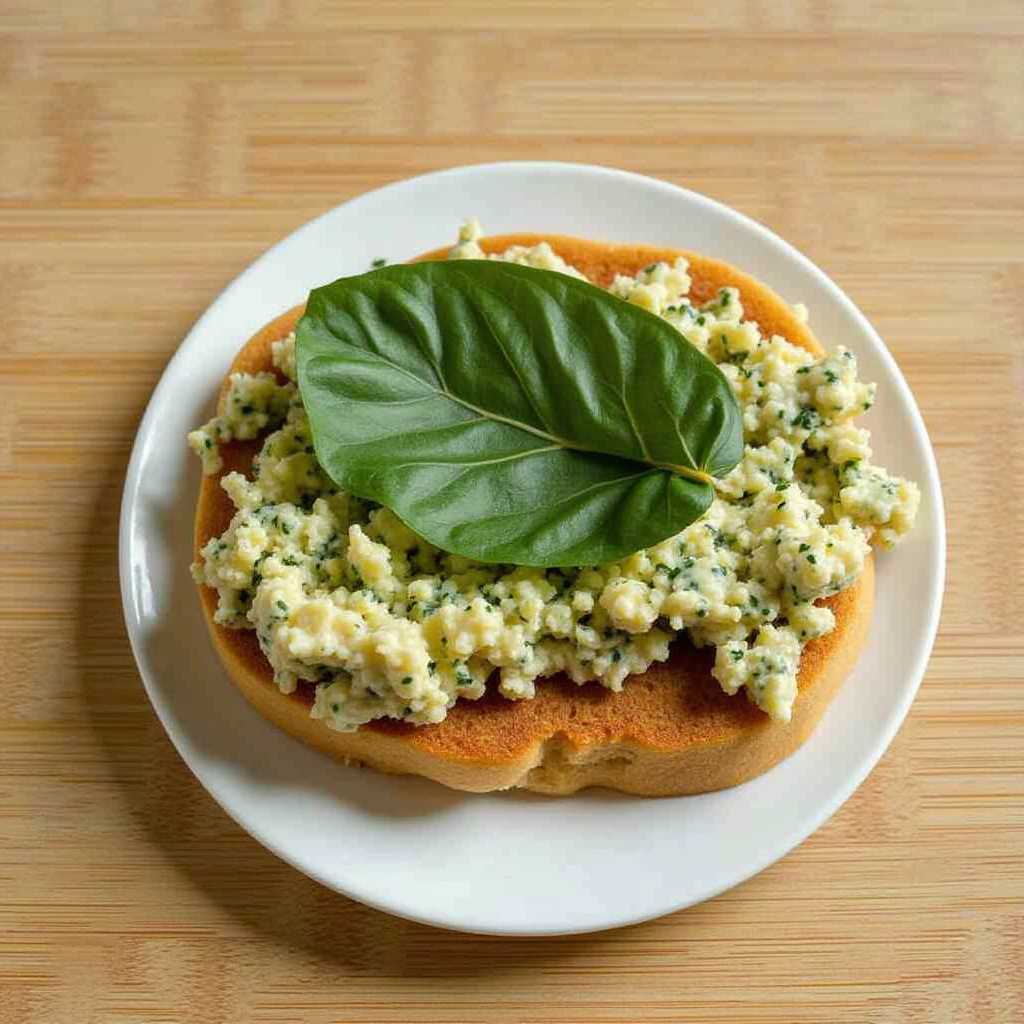} & \includegraphics[width=\imgwidth]{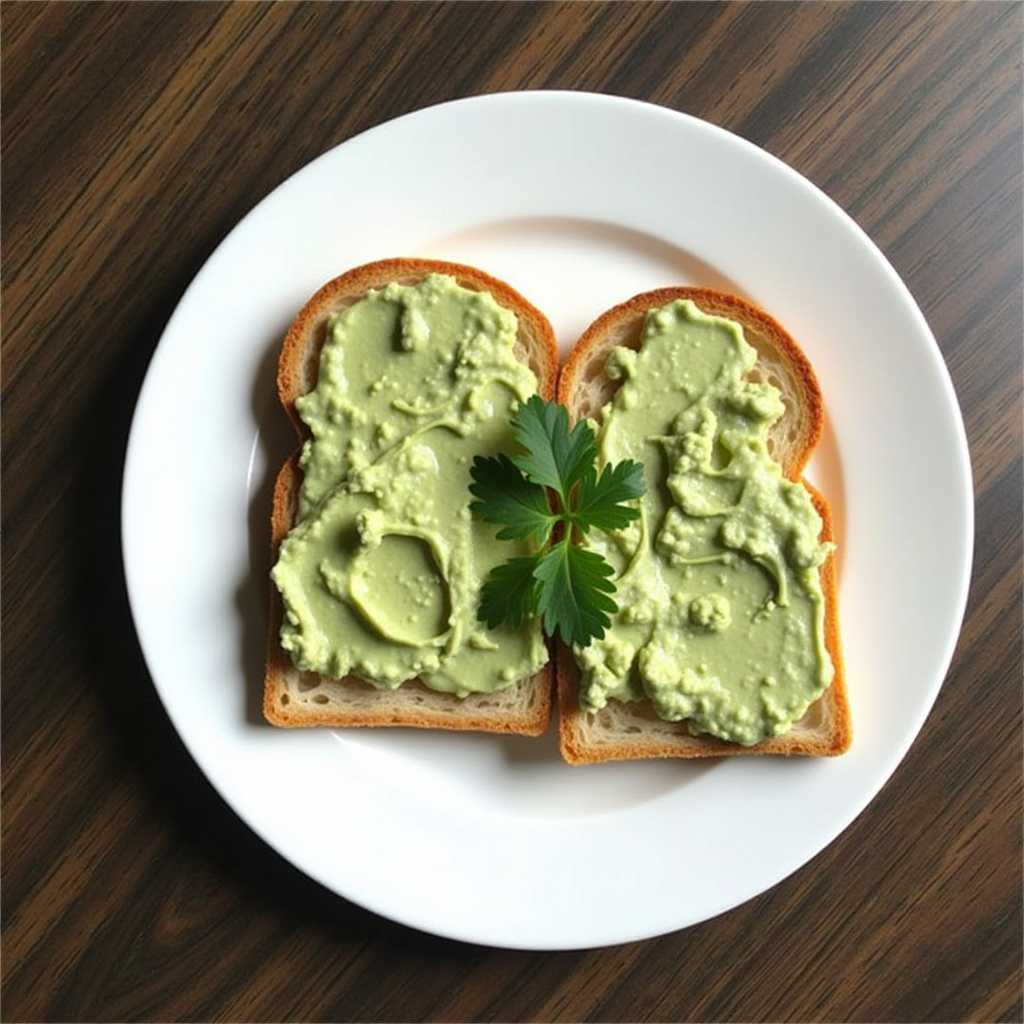} & \includegraphics[width=\imgwidth]{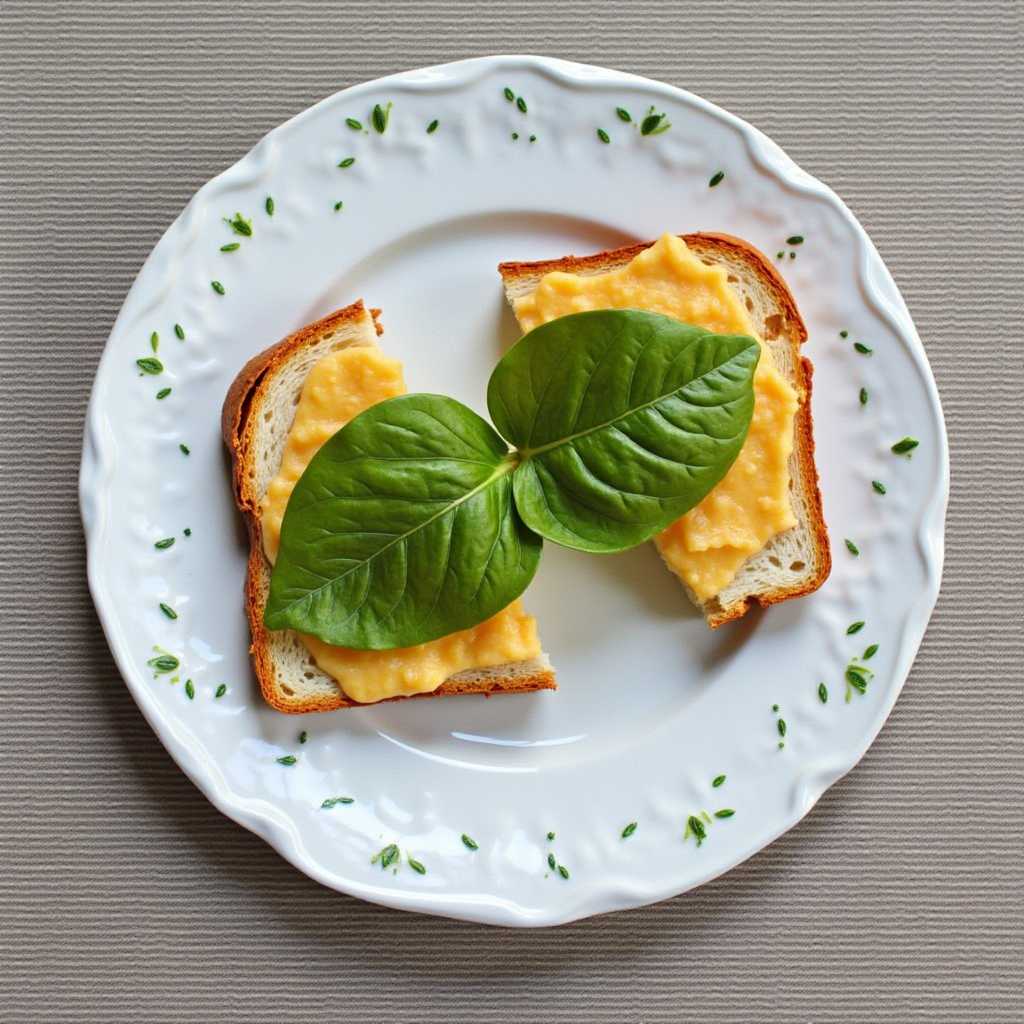} \\[-1pt]
        \labelContext & \includegraphics[width=\imgwidth]{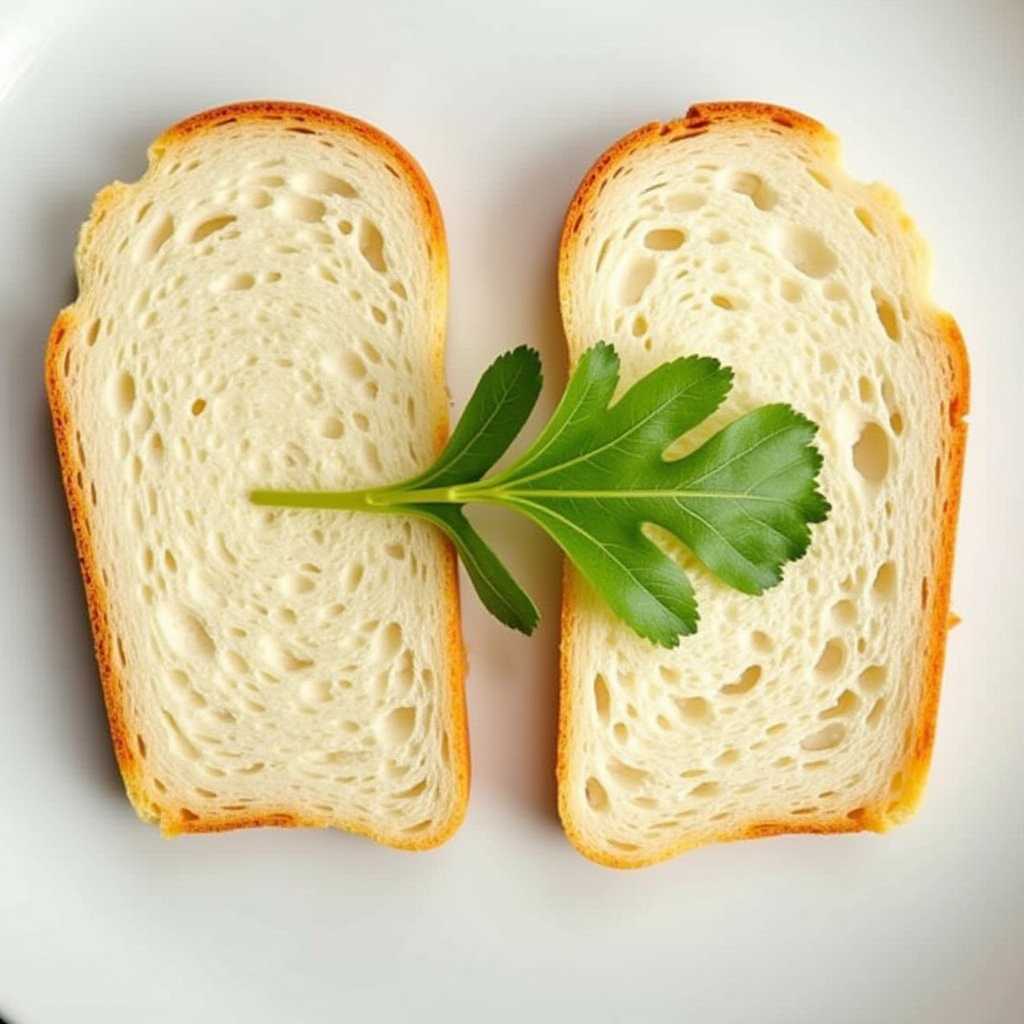} & \includegraphics[width=\imgwidth]{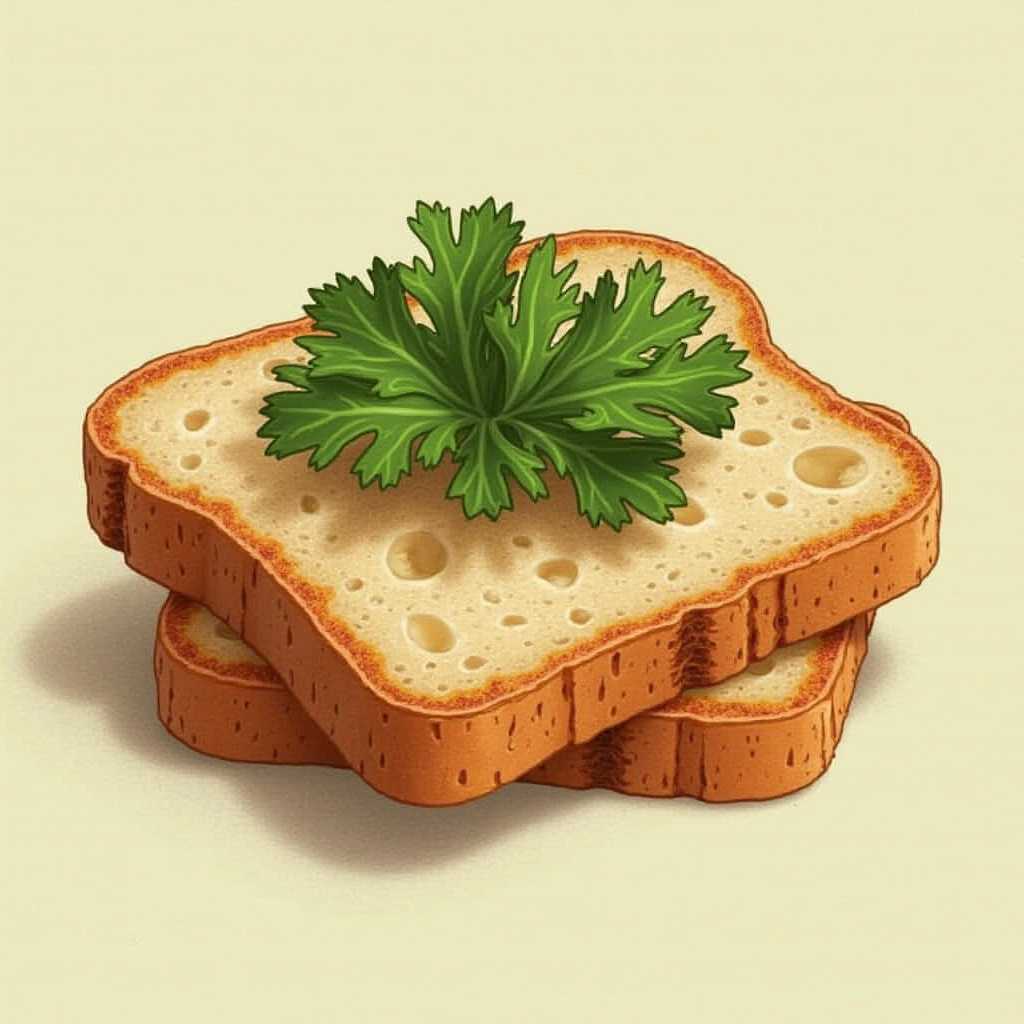} & \includegraphics[width=\imgwidth]{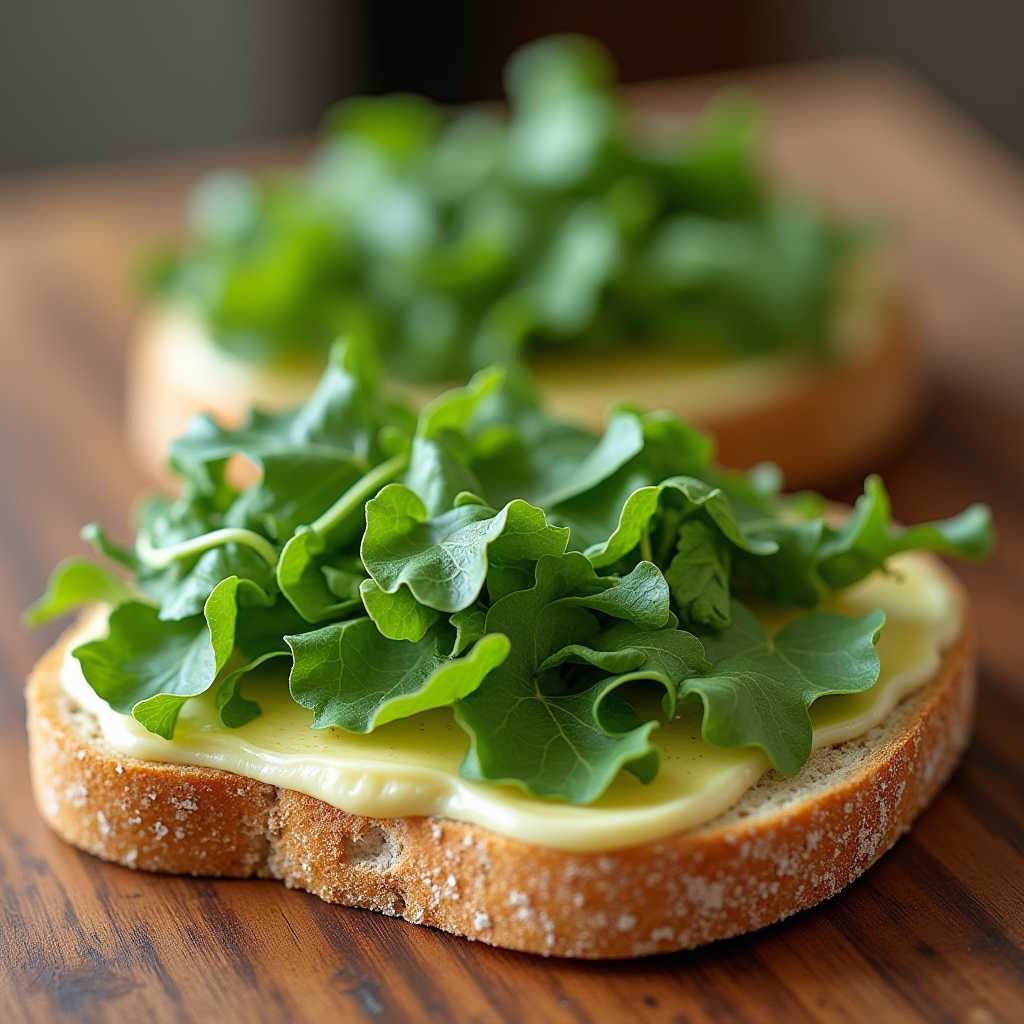} & \includegraphics[width=\imgwidth]{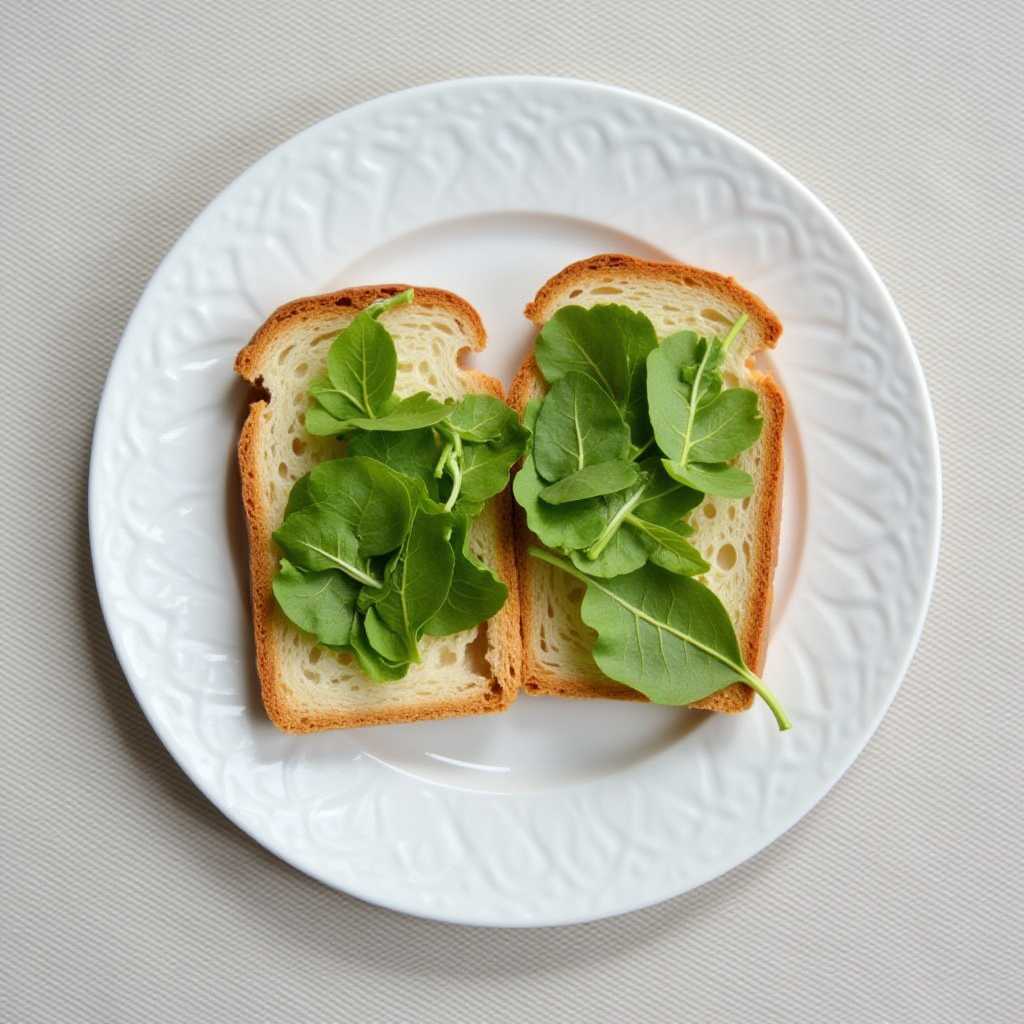} \\
        & \multicolumn{4}{c}{\vspace{2pt}\small ``Two pieces of bread with a leafy green on top of it''} \\
        \\
        \labelImage & \includegraphics[width=\imgwidth]{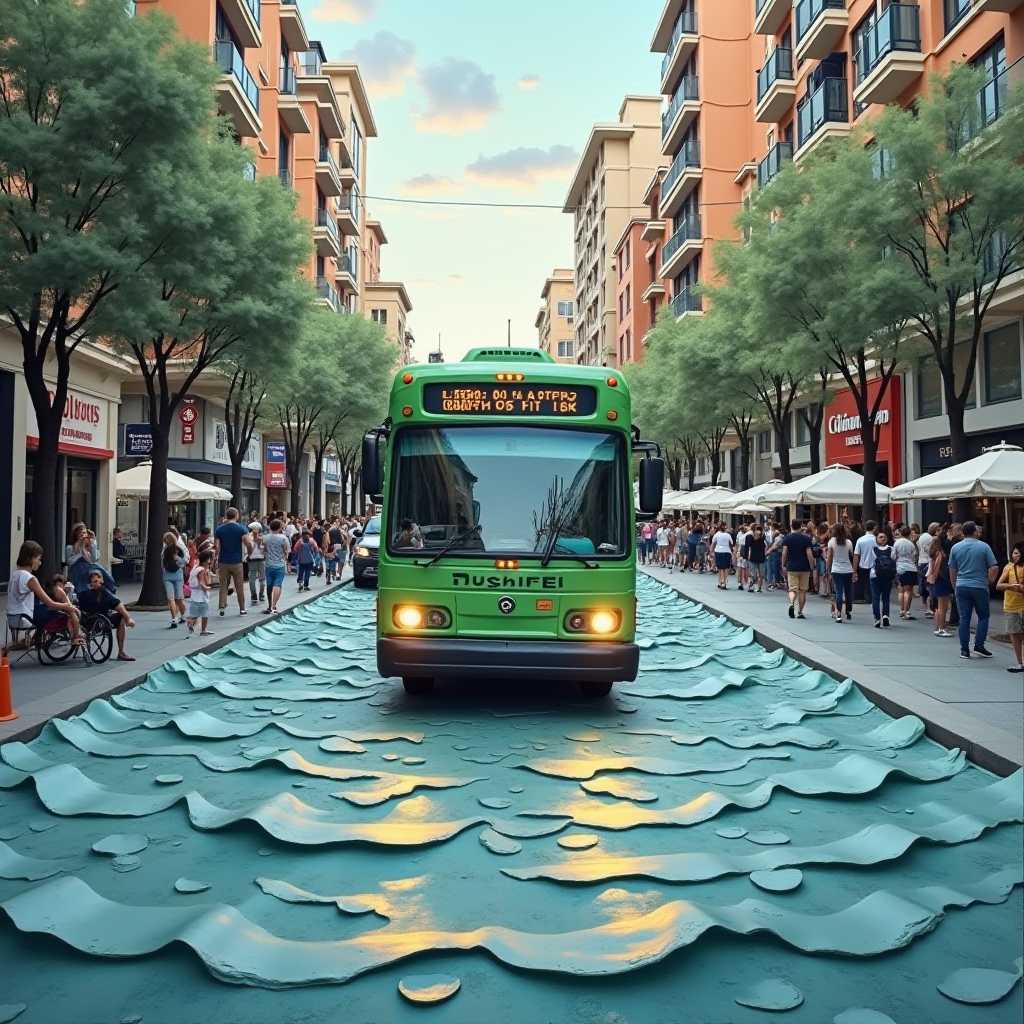} & \includegraphics[width=\imgwidth]{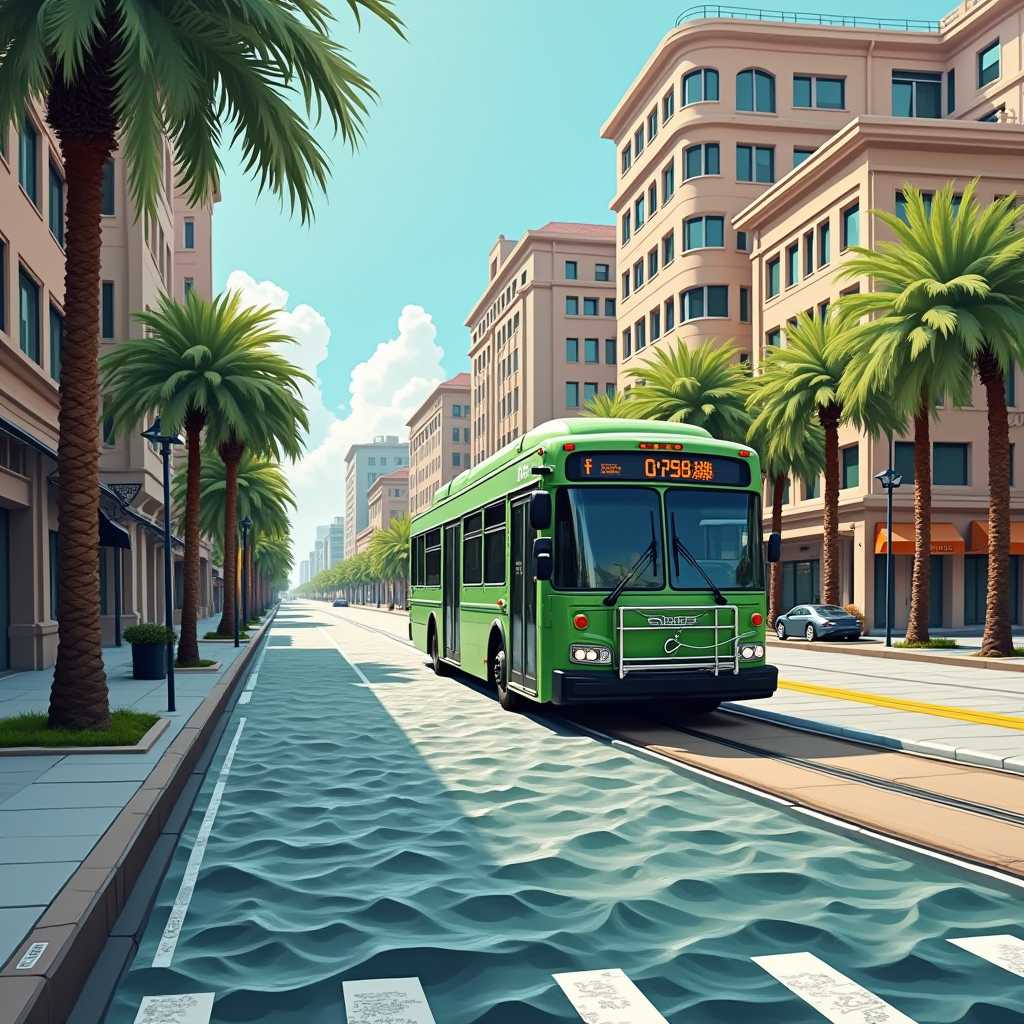} & \includegraphics[width=\imgwidth]{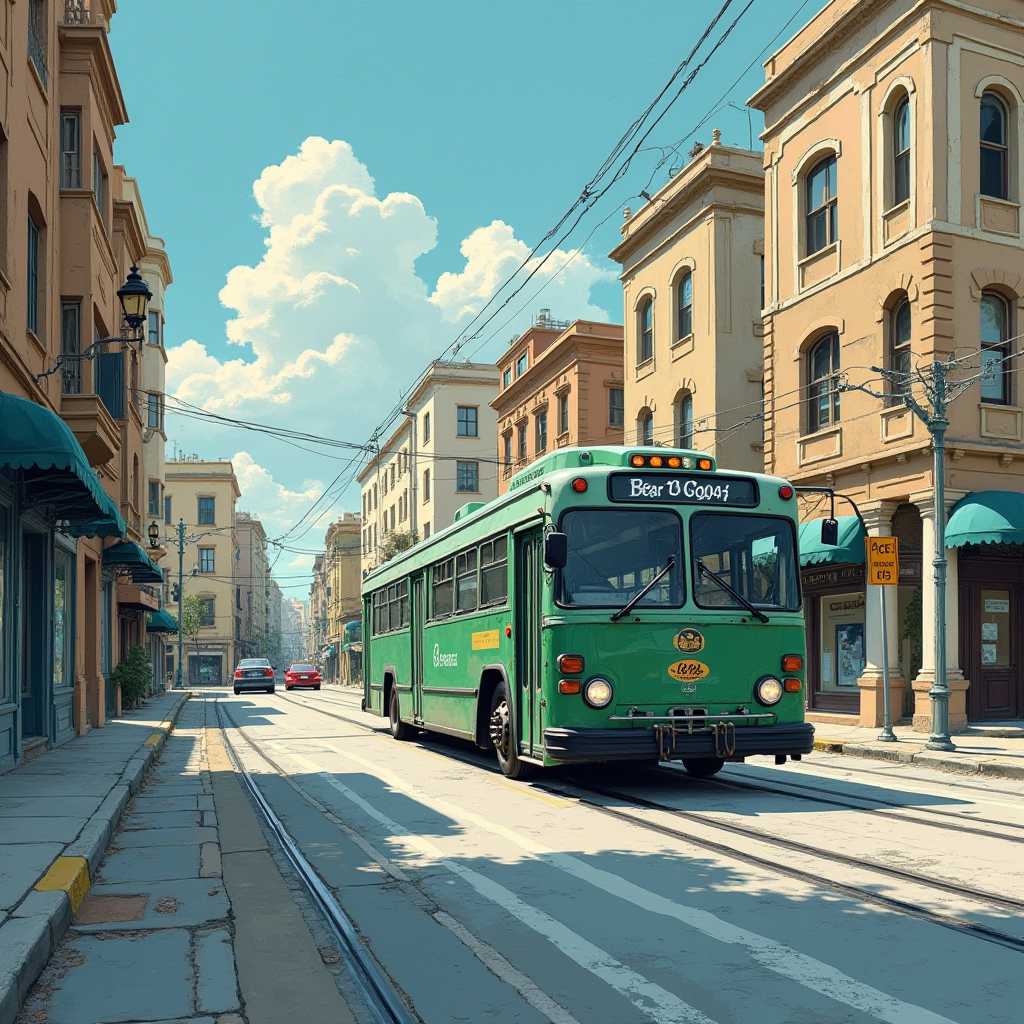} & \includegraphics[width=\imgwidth]{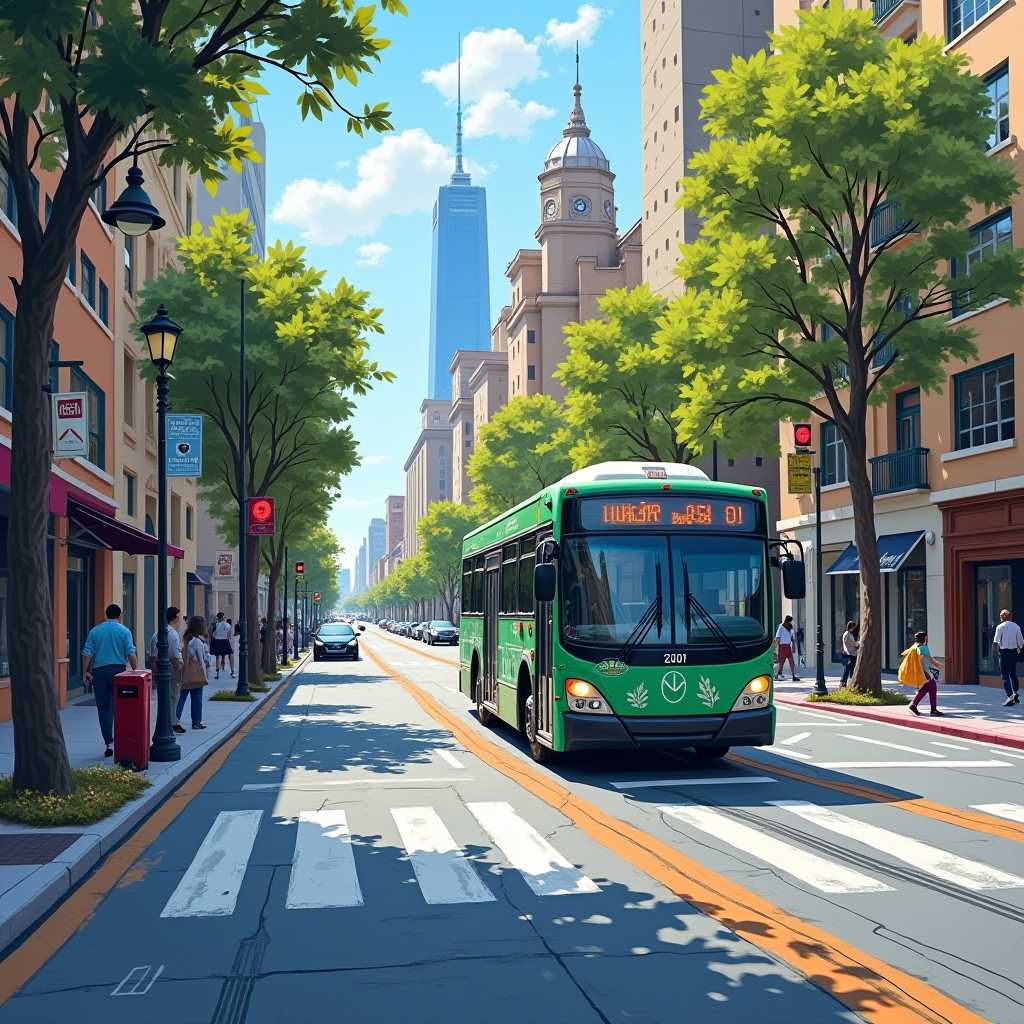} \\[-1pt]
        \labelContext & \includegraphics[width=\imgwidth]{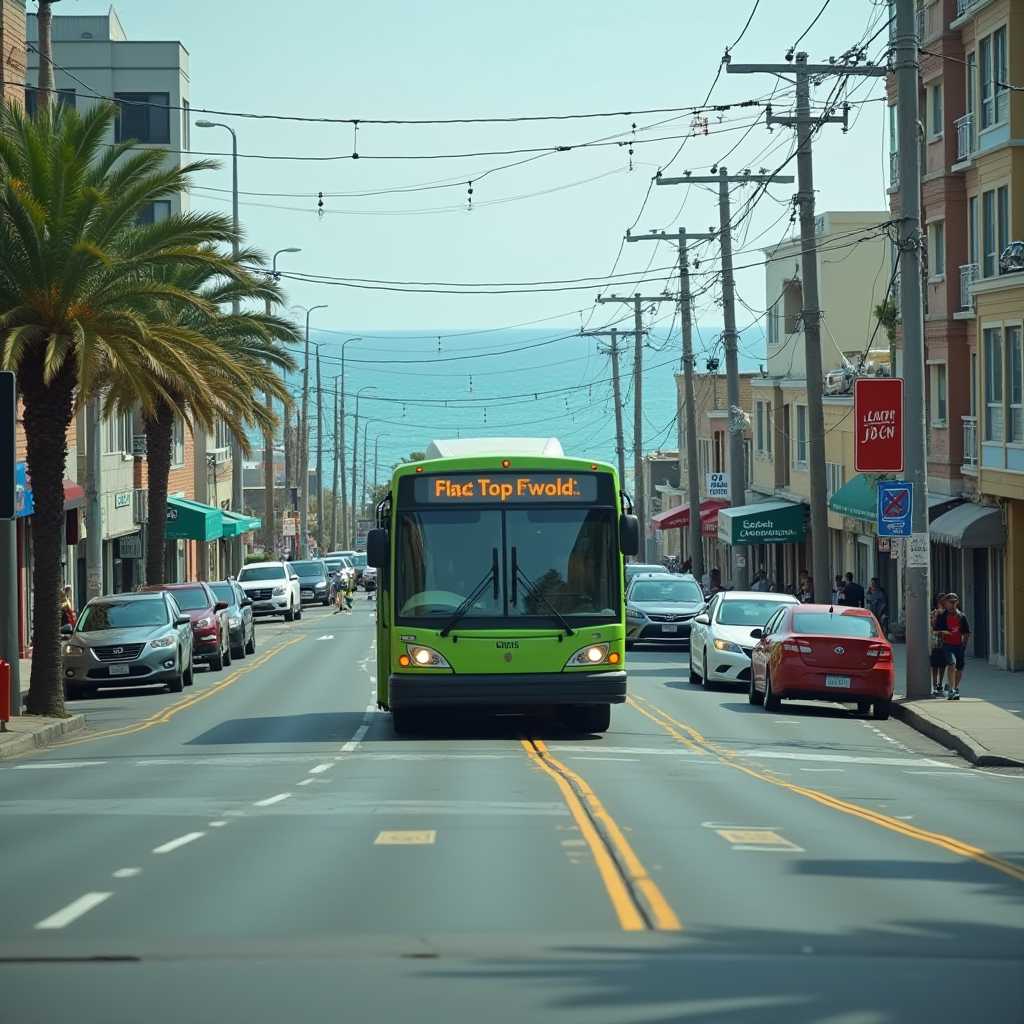} & \includegraphics[width=\imgwidth]{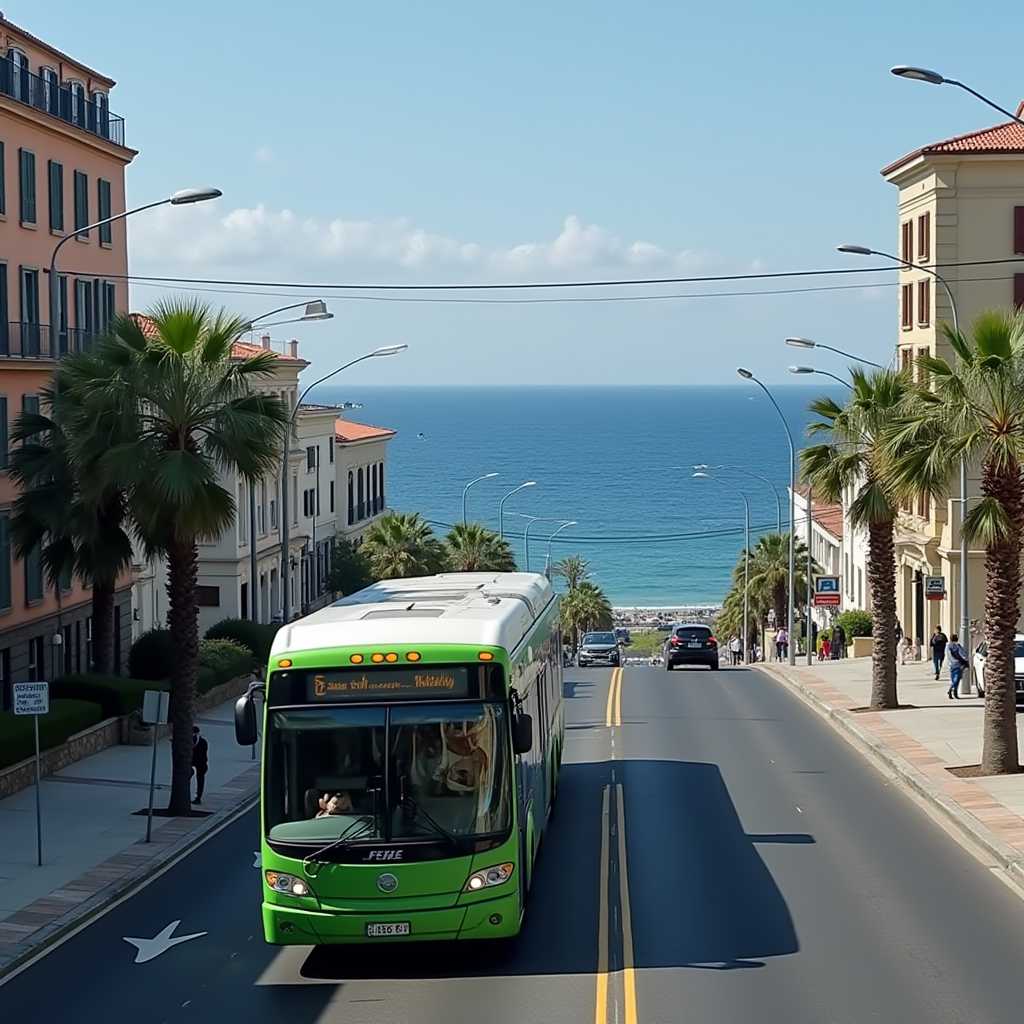} & \includegraphics[width=\imgwidth]{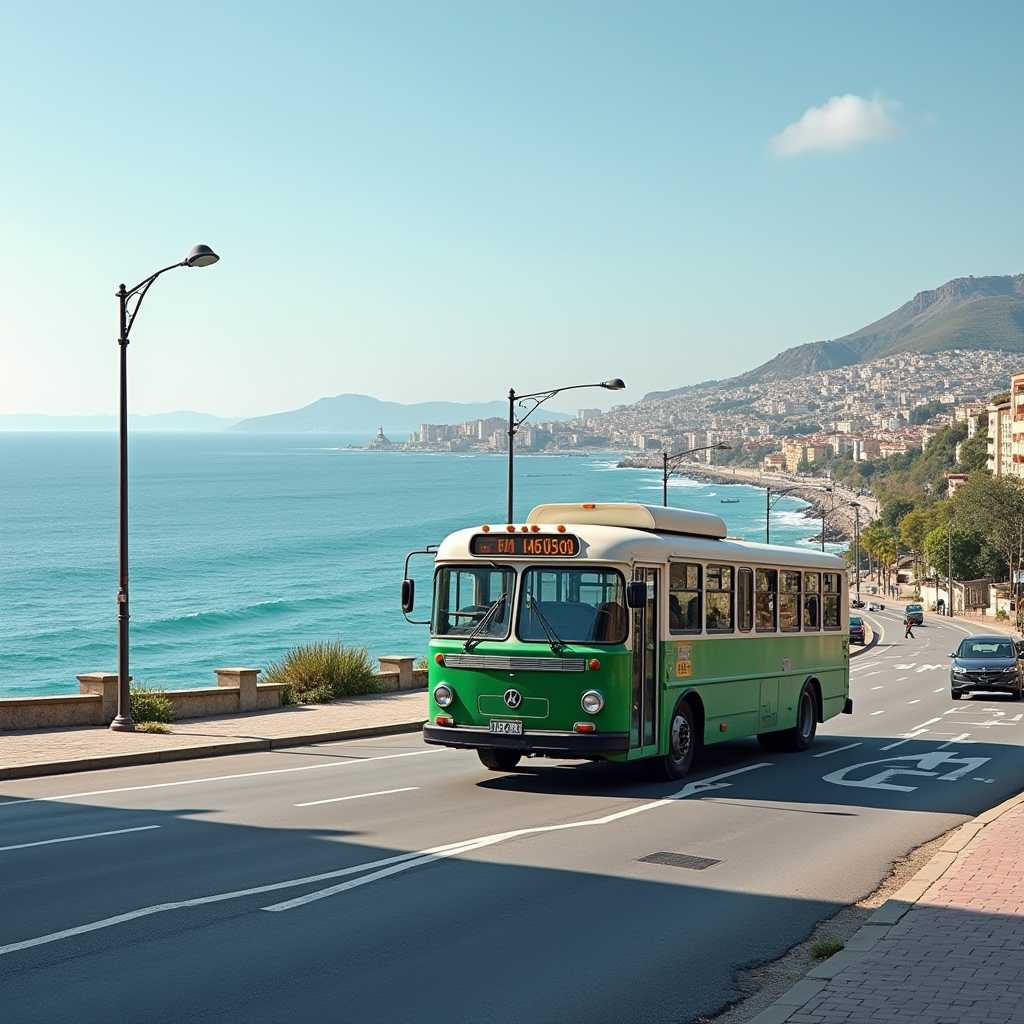} & \includegraphics[width=\imgwidth]{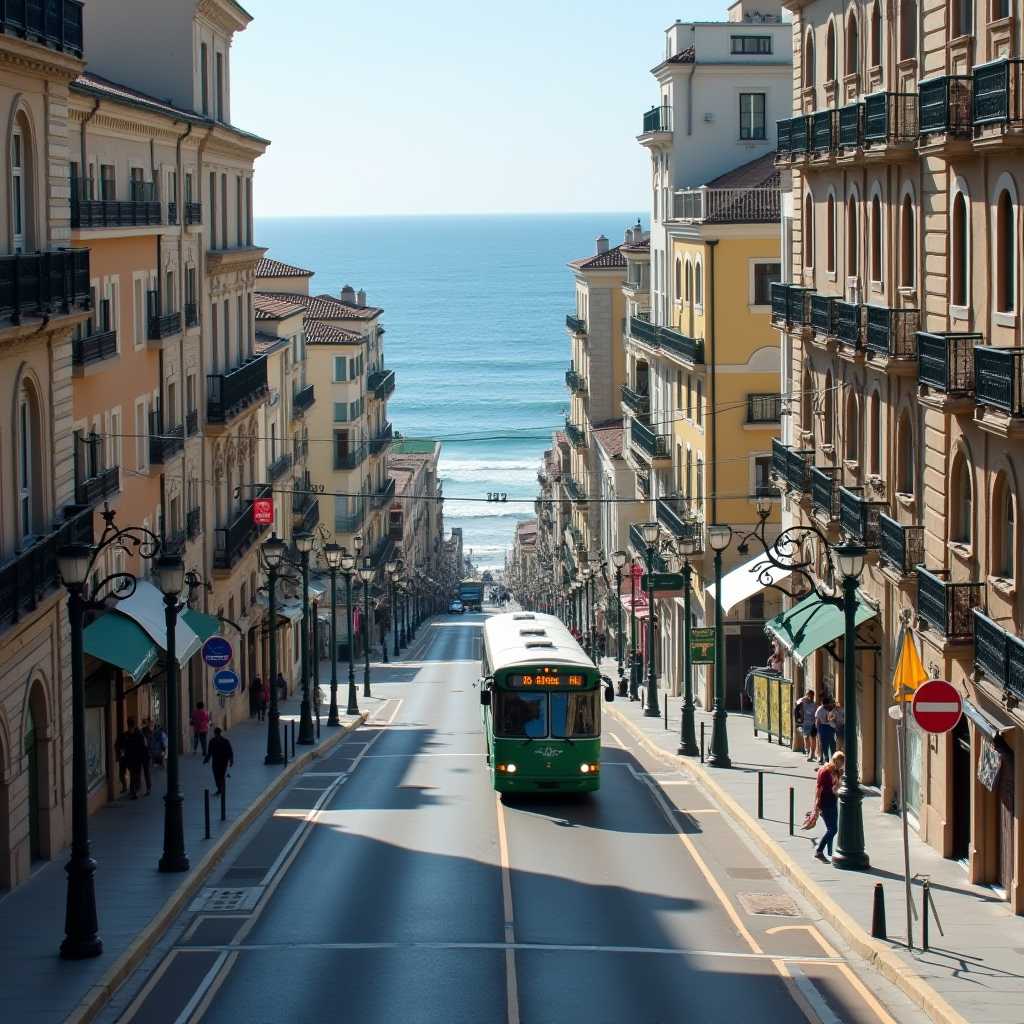} \\
        & \multicolumn{4}{c}{\vspace{2pt}\small ``A city street scene with a green bus coming up a street, with ocean''} \\
    \end{tabular}
    \caption{\textbf{Qualitative Ablation of Repulsion Space.} For each prompt, we compare repulsion applied in the image attention space (Image) versus our Contextual Space (Contextual). 
    While image-space repulsion is limited by spatial rigidity, our method achieves more varied compositions.
    }
    \label{fig:ablations_qual}
\end{figure}

Figure~\ref{fig:ablations_qual} provides qualitative examples. As can be seen, applying repulsion in the image token space ($\hat{f}_I$) often results in stagnant layouts due to its spatial rigidity; this forces the repulsion to artificially promote diversity by modifying local textures, leading to artifacts such as the sea blending unnaturally into the road in the ``street'' example.
In contrast, intervening in the contextual space ($\hat{f}_T$) tends to promote varied compositions while maintaining alignment and quality.

\section{Conclusions}
\label{sec:conclusions}
At a high level, this work highlights the Contextual Space in Diffusion Transformers as a particularly effective place to intervene when aiming for diversity. The Contextual Space sits between text and image: the representations already encode rich semantic intent shaped by the emerging image, yet they are not spatially locked in. Unlike image latents, this space is not tied to a spatial grid, so samples can be pushed apart semantically without tearing geometry or introducing visual artifacts. At the same time, unlike early text embeddings, it is structurally informed, meaning that interventions meaningfully influence what the model actually generates.

Applying on-the-fly repulsion in this space allows diversity to be increased in a controlled way, without sacrificing visual quality or relying on heavy optimization with significant computational cost. More broadly, this points to the importance of intervening at the right representational level, where decisions are still flexible, but already grounded in the image being formed.

\paragraph{Limitations} Contextual repulsion increases diversity but does not provide direct control over which attributes will vary, and may sometimes favor coarse semantic changes over fine, user-specified ones. In addition, the intervention is focused on early to mid stages of generation; how to best coordinate it with later stages, or combine it with other control mechanisms, remains an open question.

\paragraph{Future directions} An interesting direction for future work is to investigate whether a user-provided textual cue, such as “color” or “size”, can be used to guide the repulsion along a specific semantic direction in the Contextual Space. Instead of encouraging diversity in an unconstrained manner, the idea would be to bias the repulsive forces so that samples spread primarily along attributes associated with the given word. This could enable a more controlled and interpretable form of diversity, where variation is focused on selected semantic aspects while other parts of the generation remain stable.

\begin{acks}
We would like to thank Or Patashnik, Yuval Alaluf, Nir Goren, Maya Vishnevsky, Sara Dorfman, Shelly Golan, Saar Huberman, and Jackson Wang for their early feedback and insightful discussions.
We also thank the anonymous reviewers for their thorough and constructive comments, which helped improve this work.

This research was supported in part by the Israel Science Foundation (Grants No. 2492/20 and 1473/24) and by Len Blavatnik and the Blavatnik Family Foundation.
\end{acks}

\bibliographystyle{ACM-Reference-Format}
\bibliography{main}

\clearpage
\begin{figure*}
    \centering
    \setlength{\tabcolsep}{0.5pt} \renewcommand{\arraystretch}{0.5} \newcommand{\imgwidth}{0.12\textwidth}
    \newcommand{\vertlabel}[1]{\raisebox{2.5em}{\rotatebox{90}{\scriptsize\textbf{#1}}}}
    \begin{tabular}{c c c c c c c c c}

        \vertlabel{Flux} & \includegraphics[width=\imgwidth]{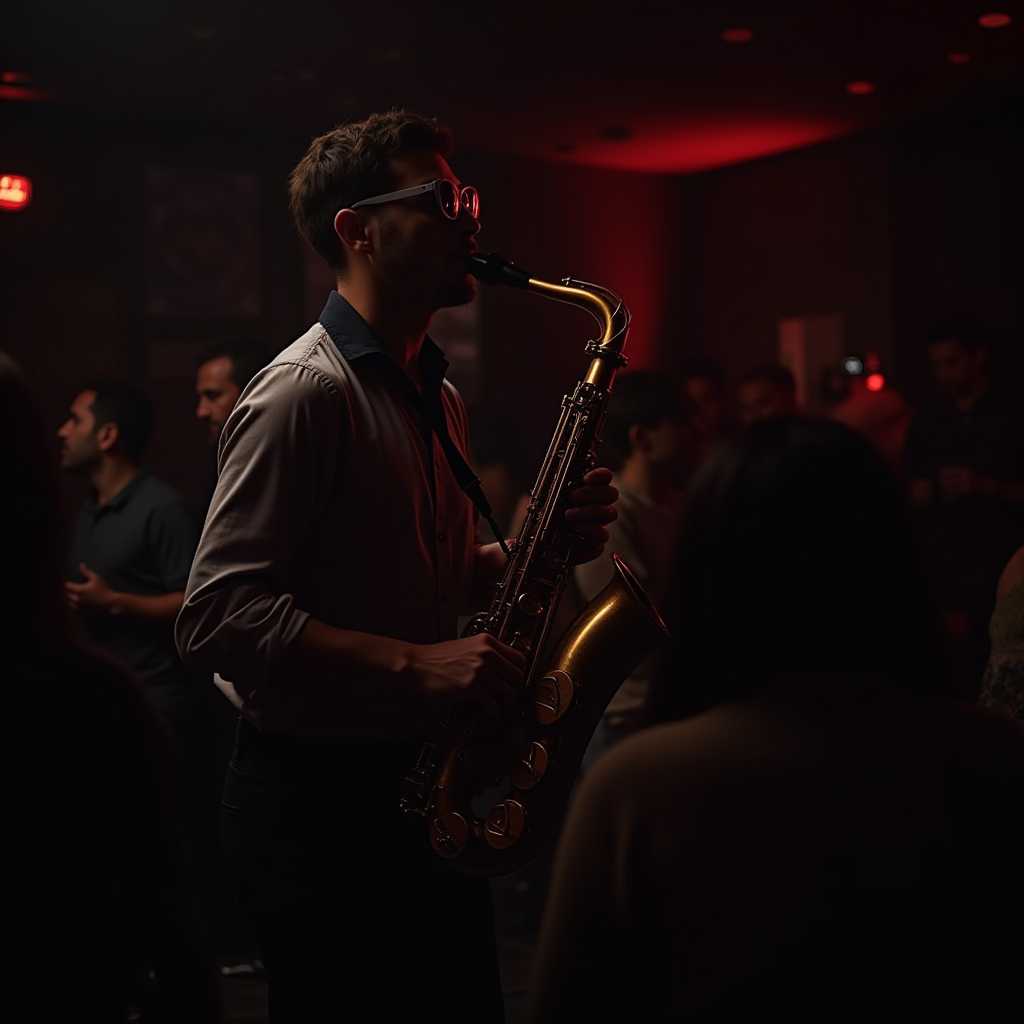} & \includegraphics[width=\imgwidth]{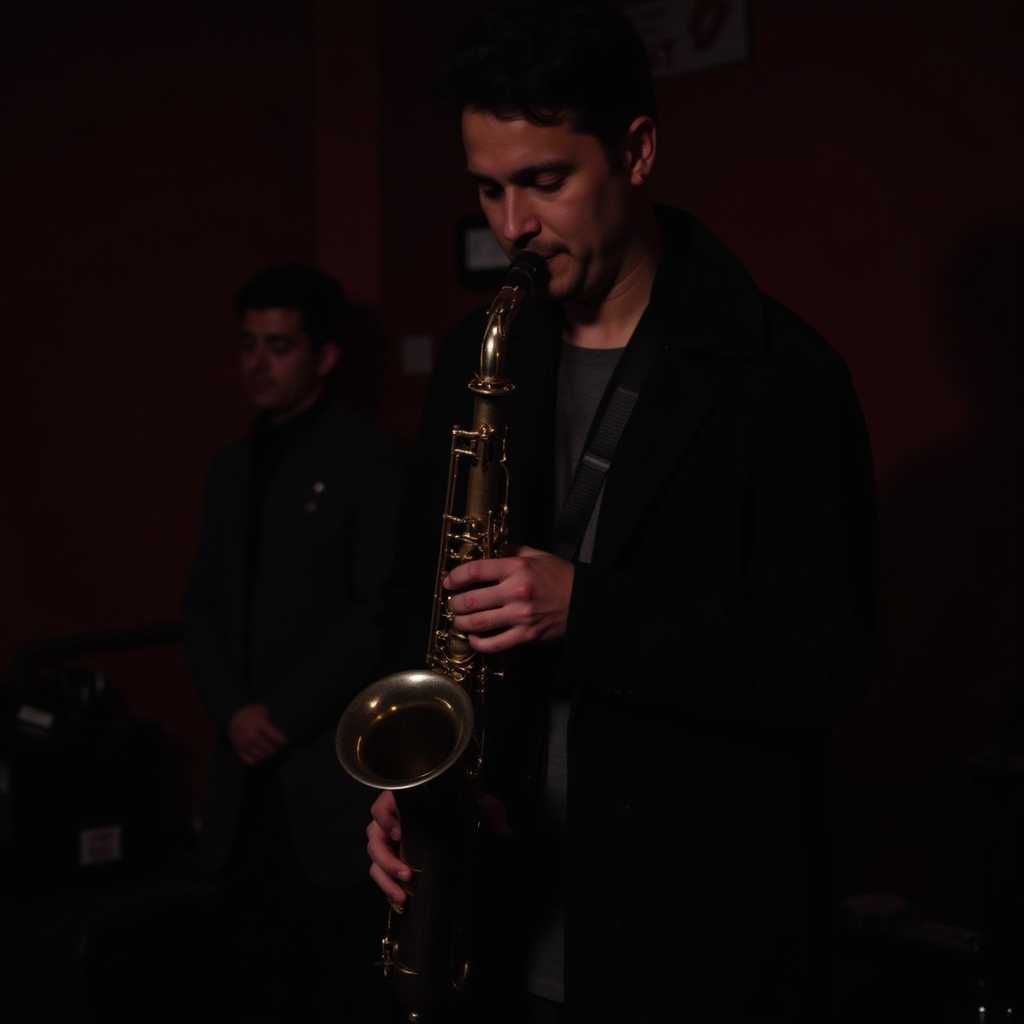} & \includegraphics[width=\imgwidth]{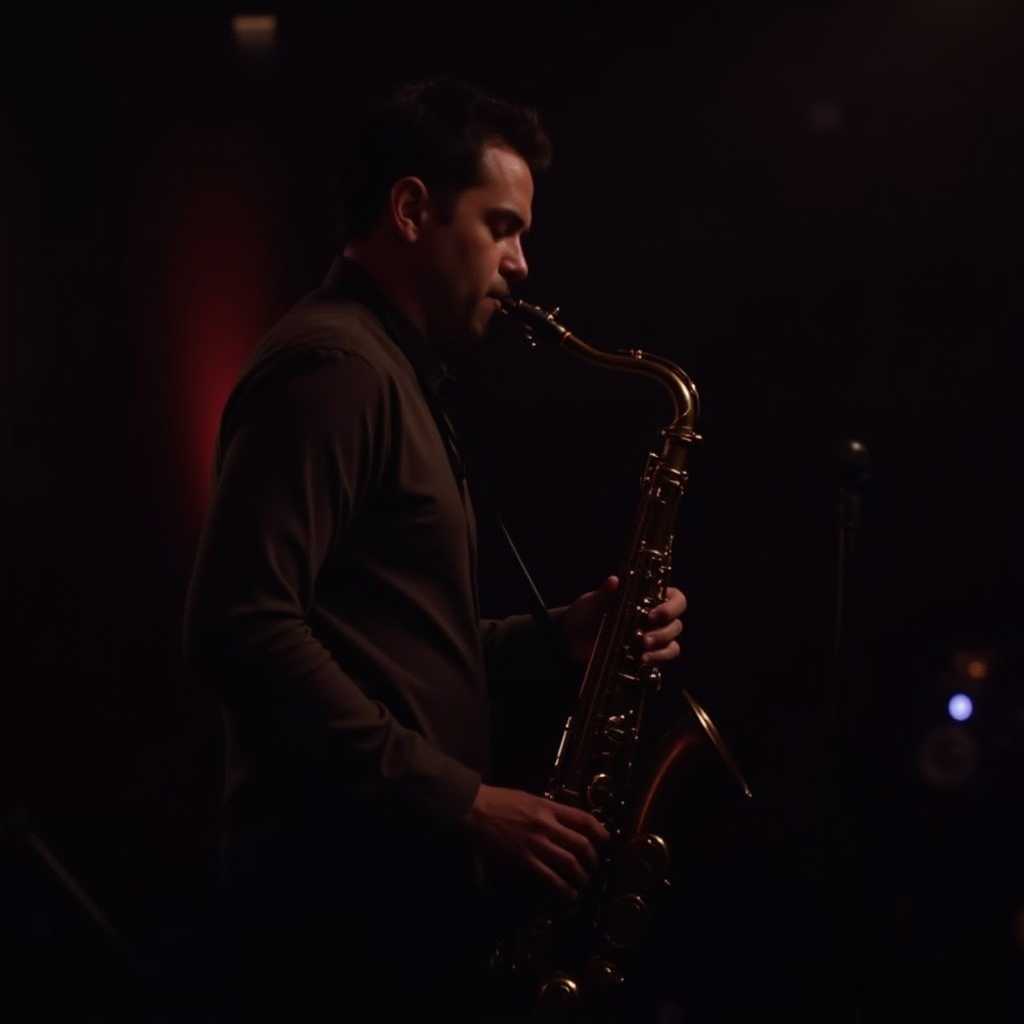} & \includegraphics[width=\imgwidth]{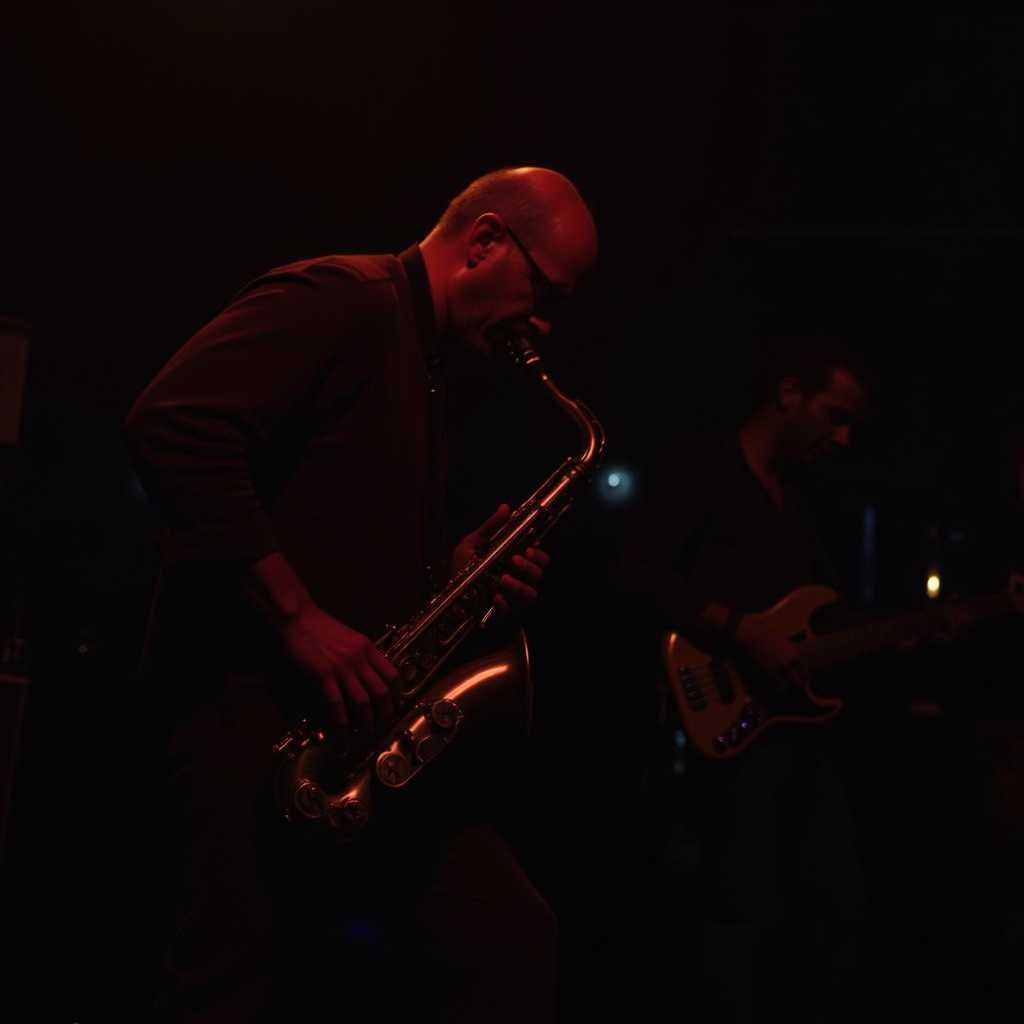} & \includegraphics[width=\imgwidth]{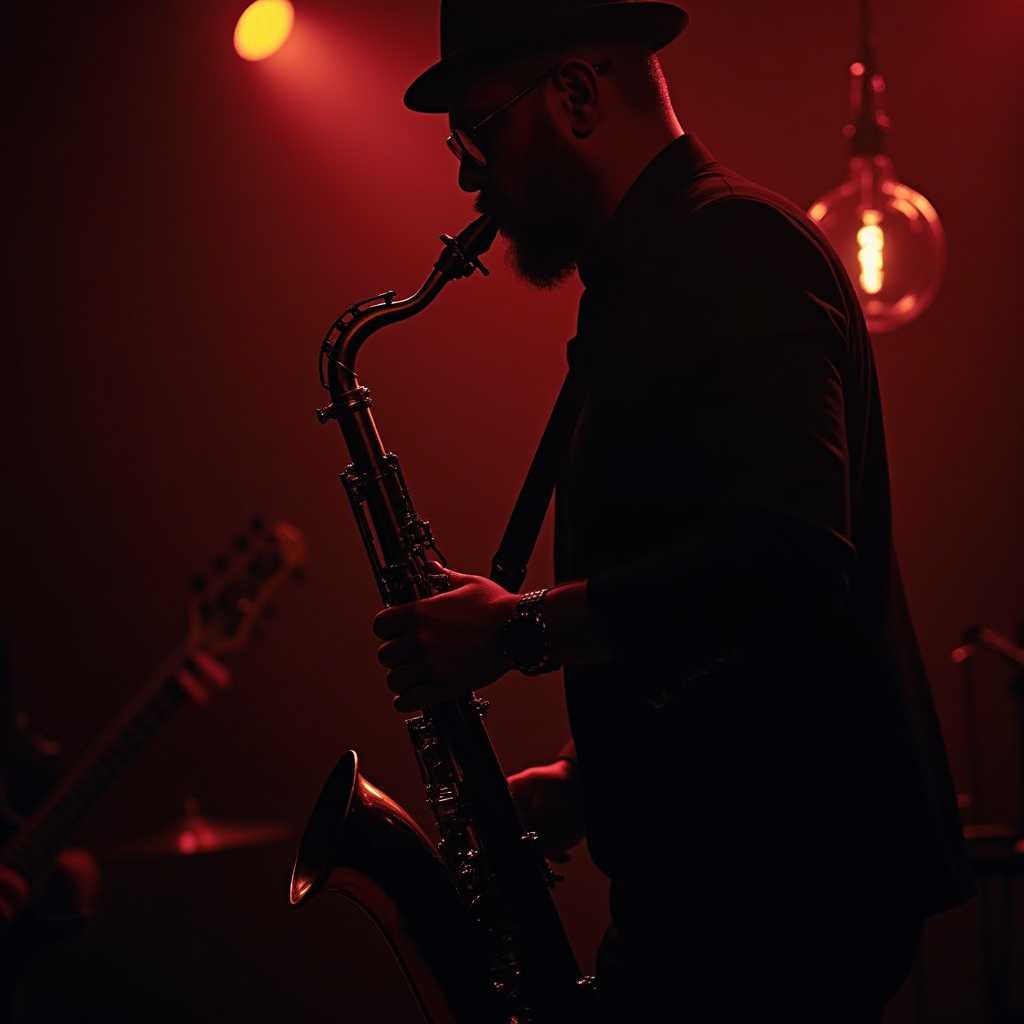} & \includegraphics[width=\imgwidth]{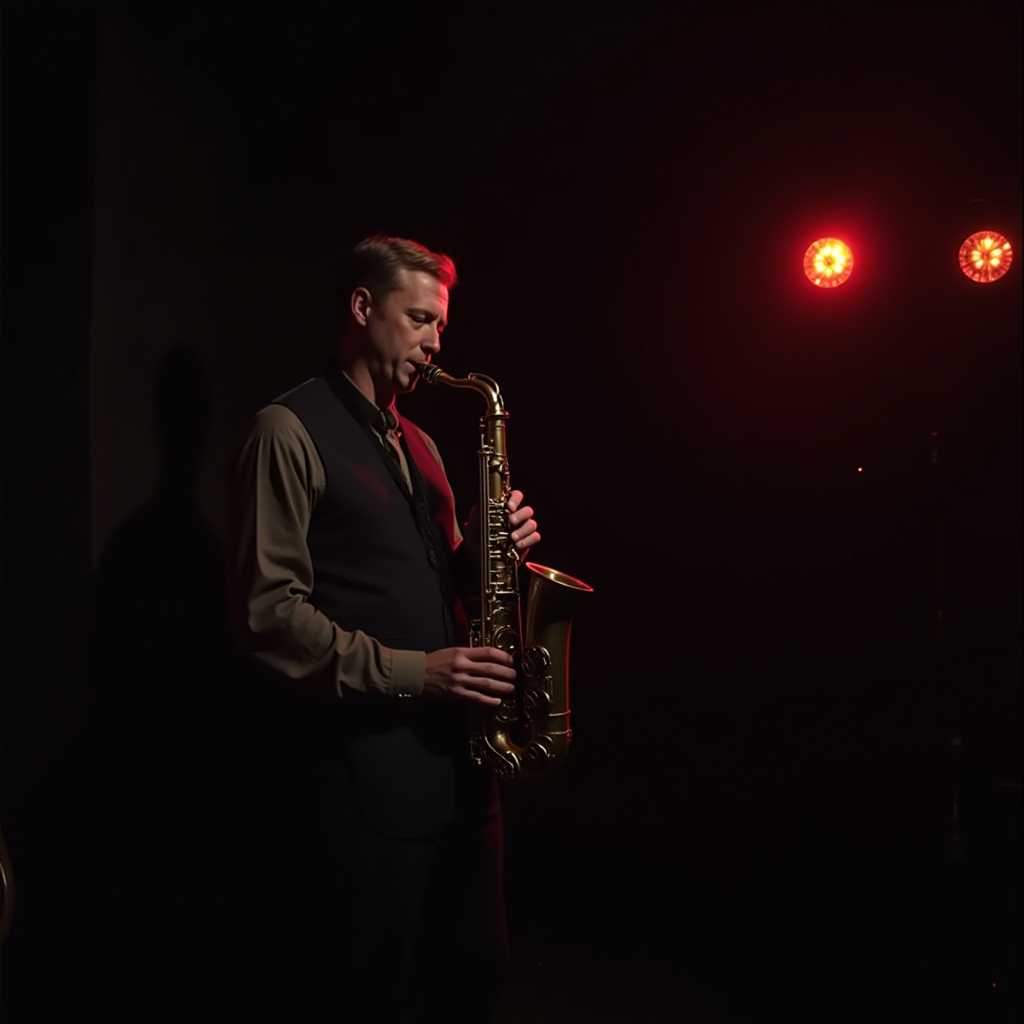} & \includegraphics[width=\imgwidth]{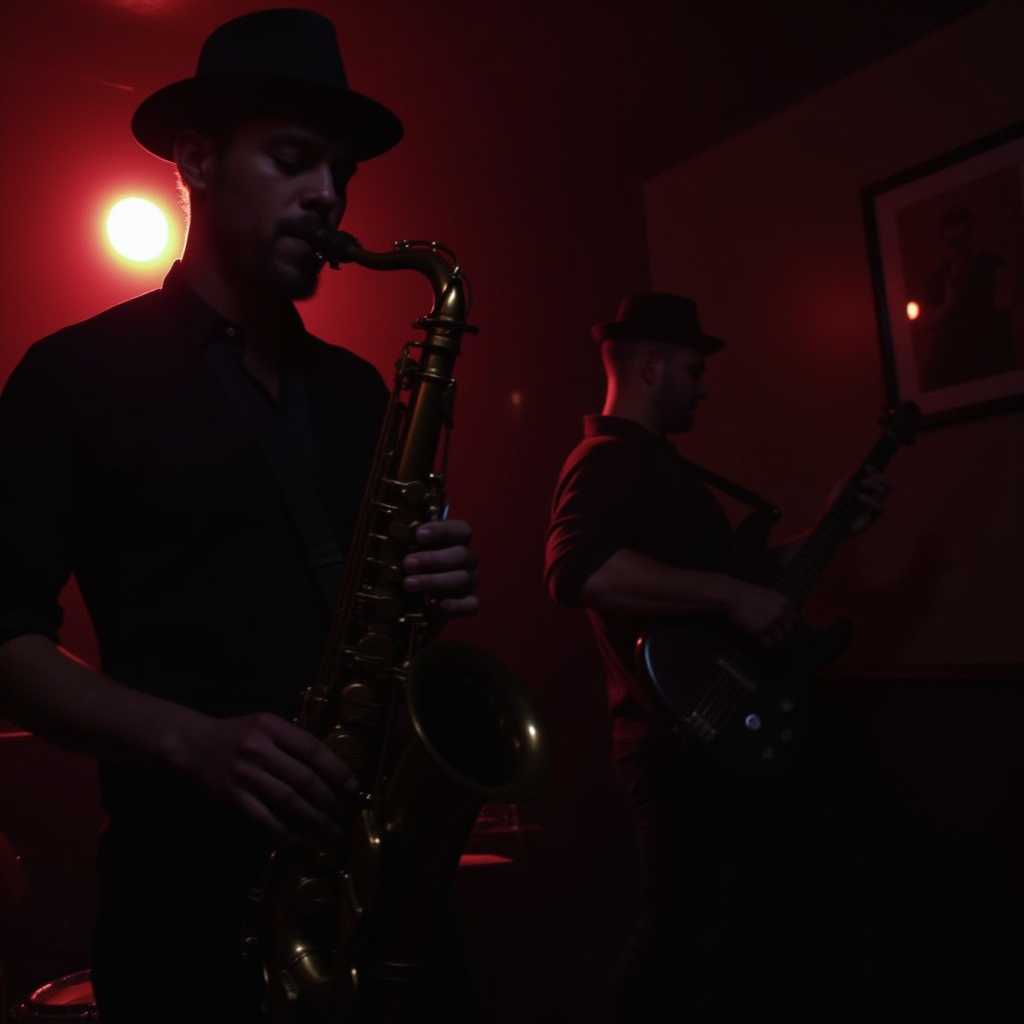} & \includegraphics[width=\imgwidth]{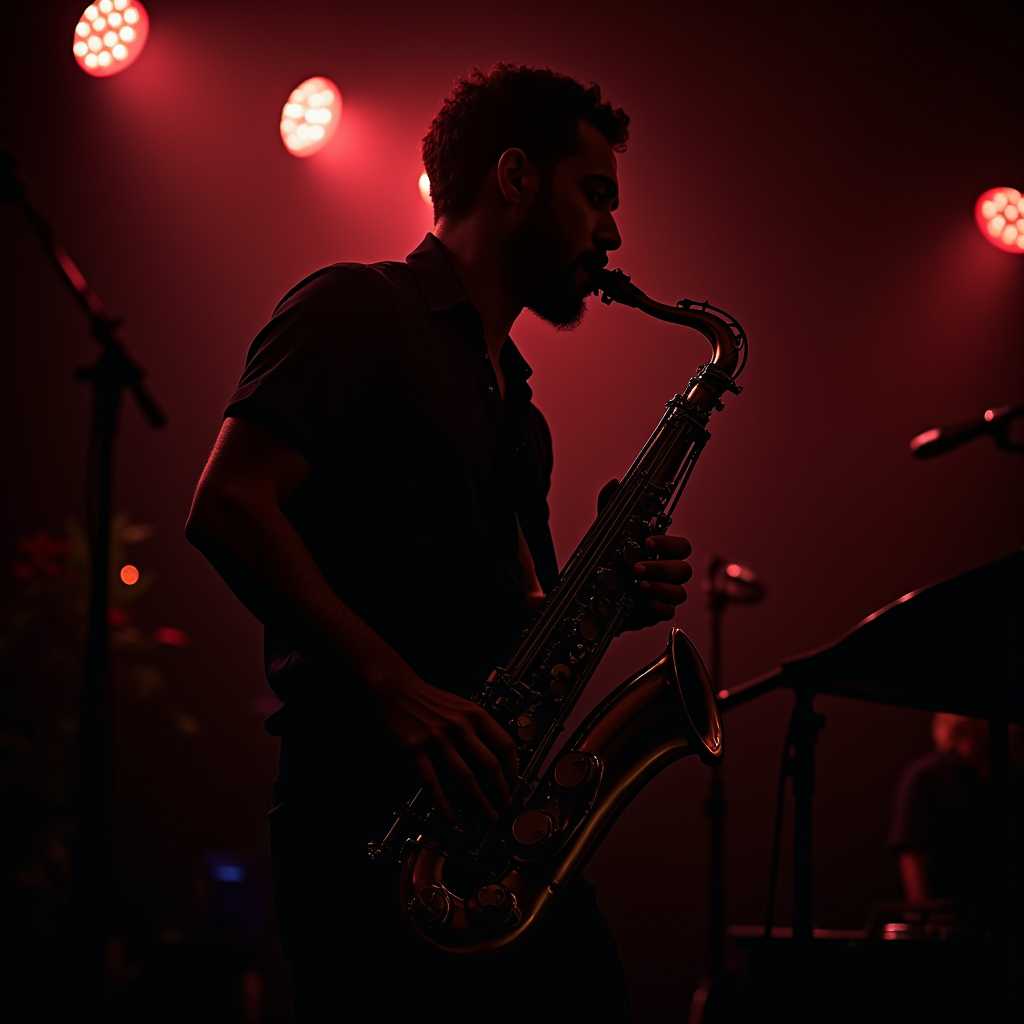} \\[-1pt]
        \vertlabel{Ours} & \includegraphics[width=\imgwidth]{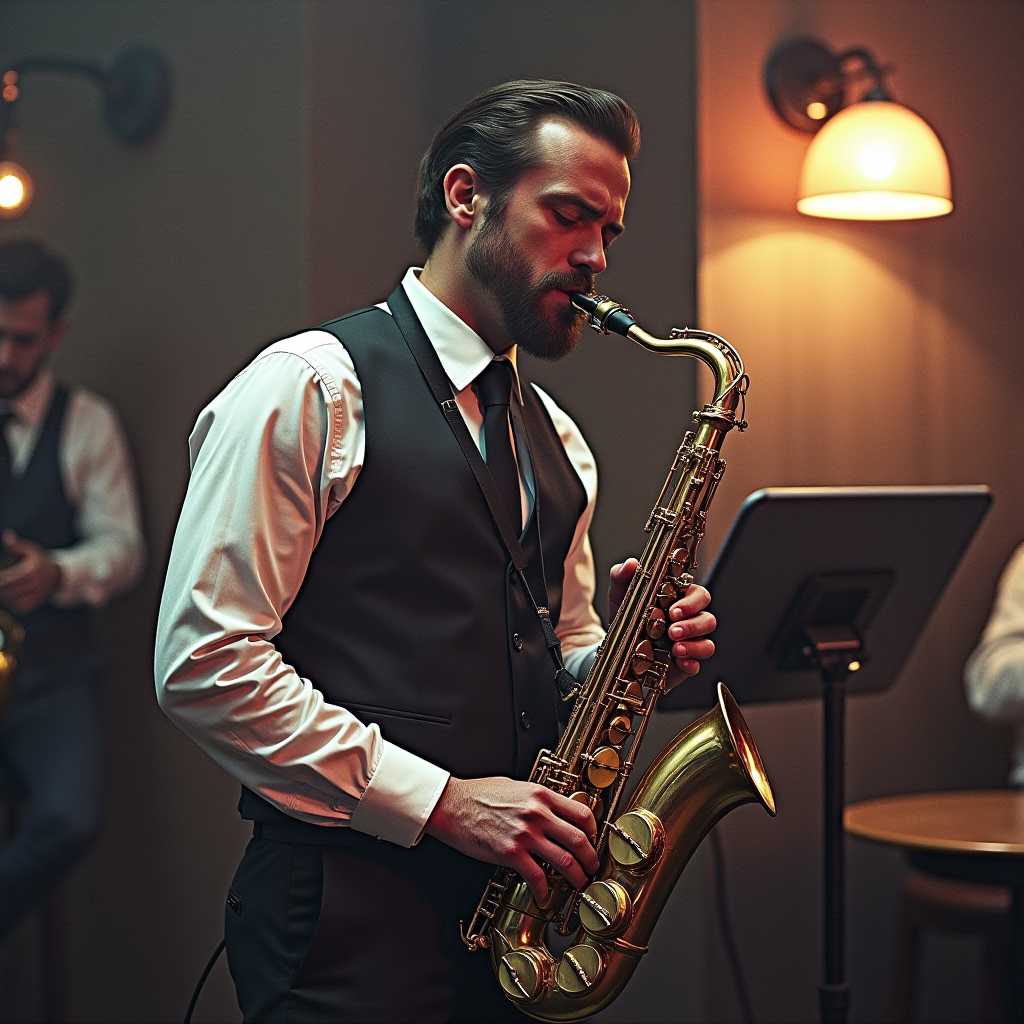} & \includegraphics[width=\imgwidth]{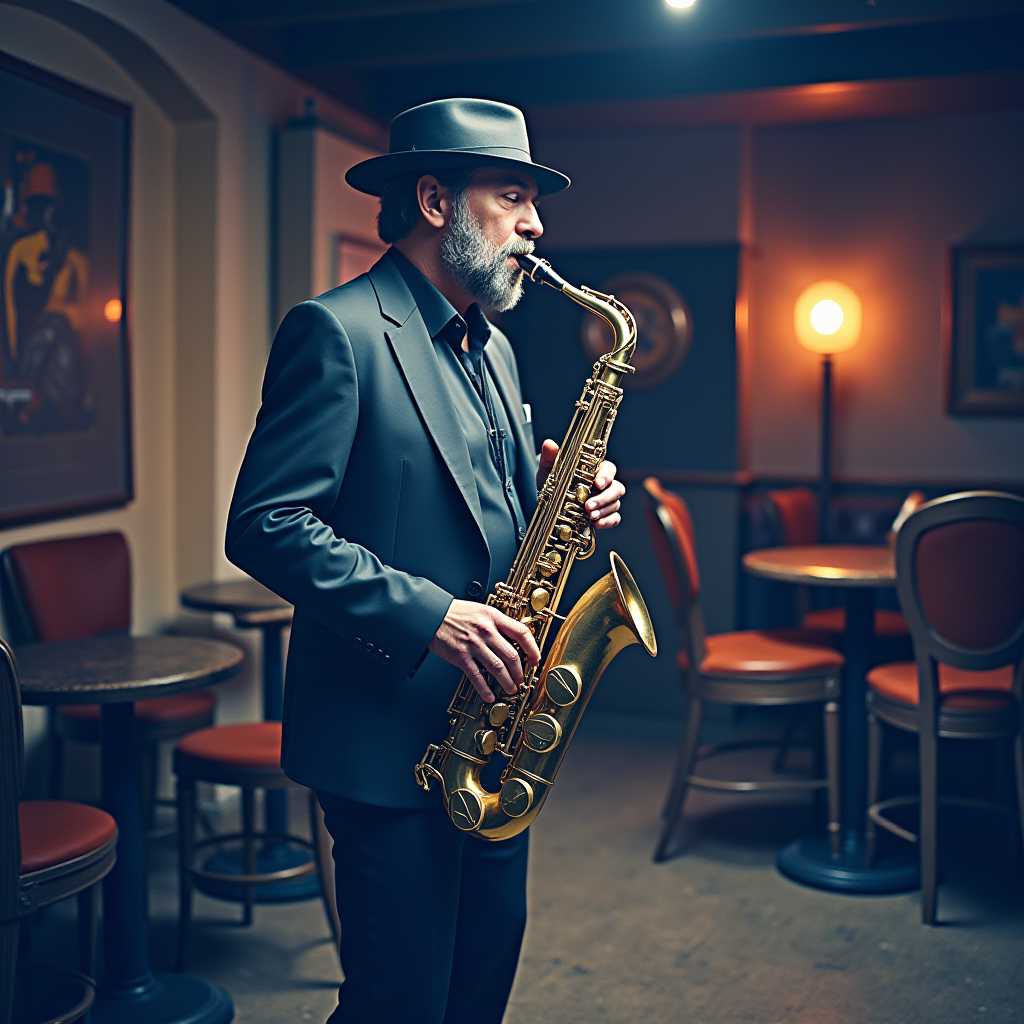} & \includegraphics[width=\imgwidth]{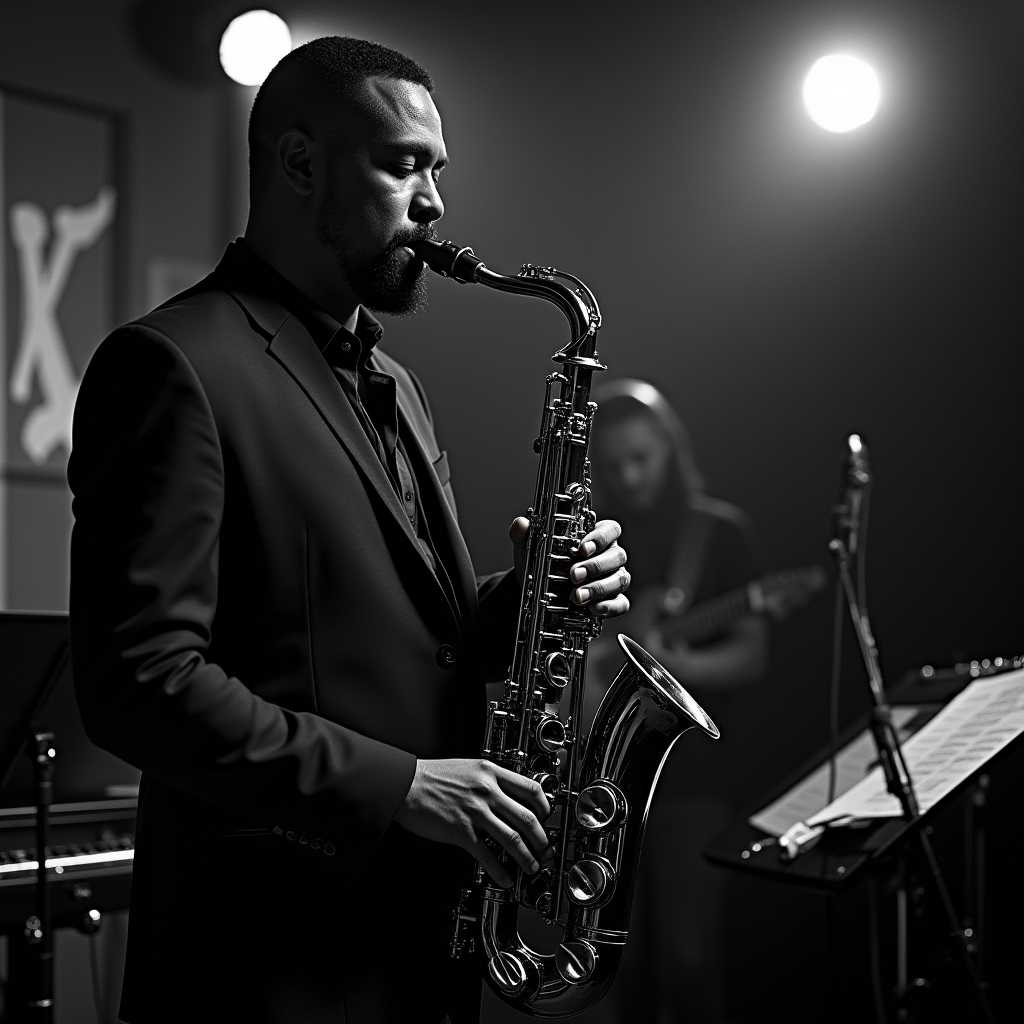} & \includegraphics[width=\imgwidth]{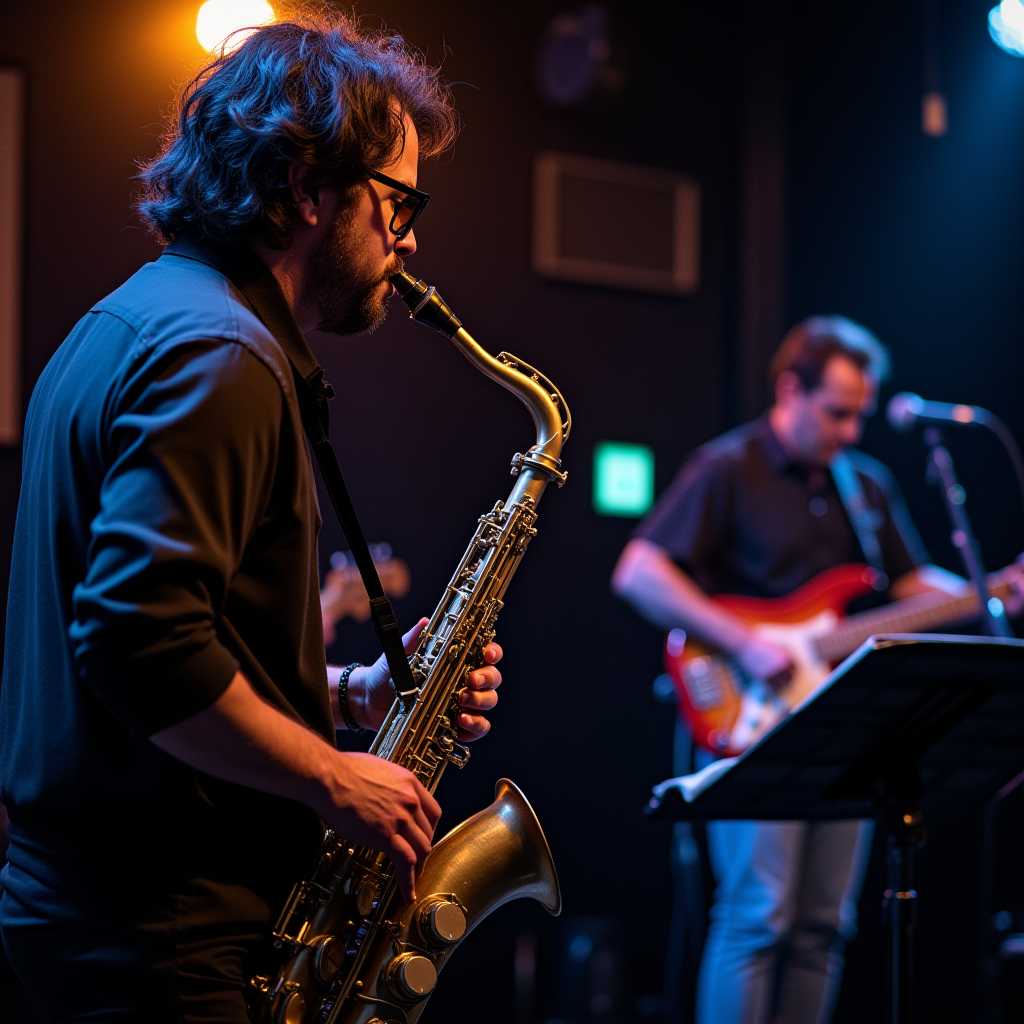} & \includegraphics[width=\imgwidth]{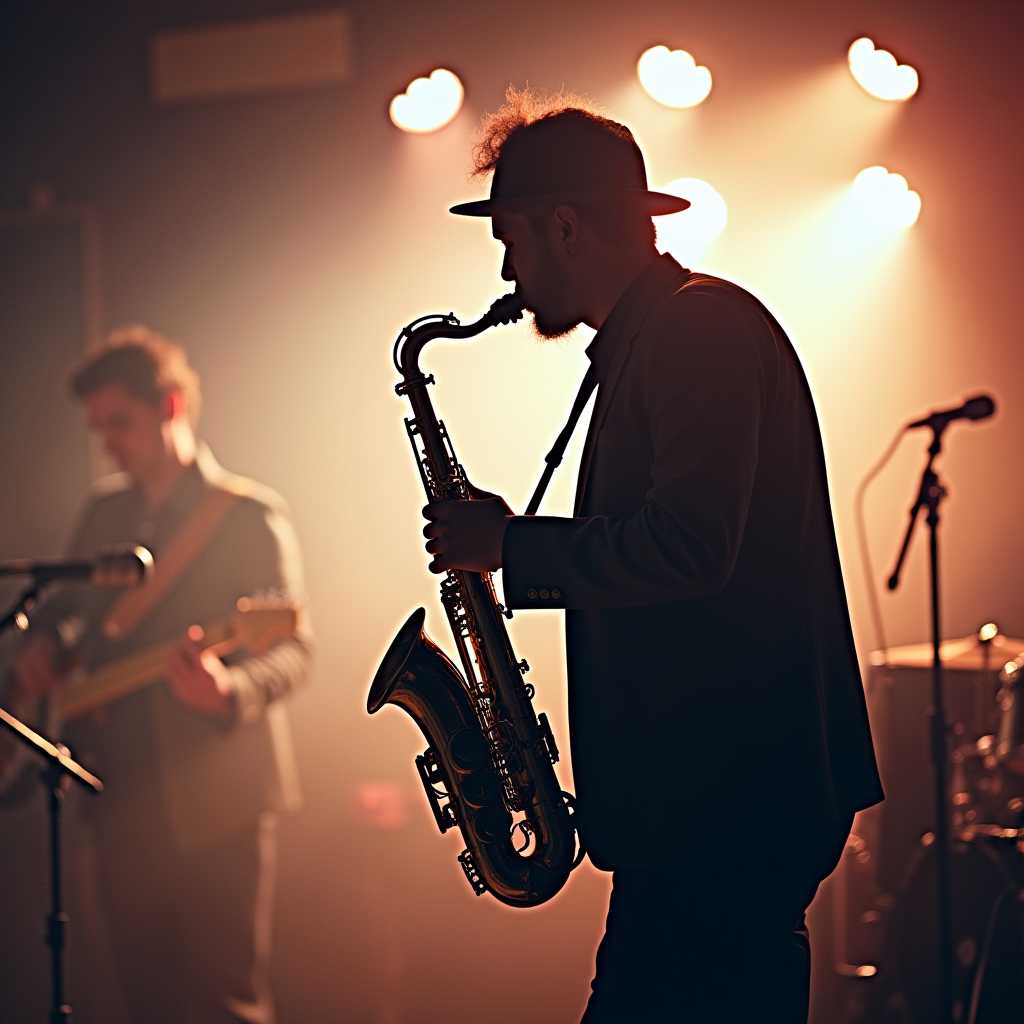} & \includegraphics[width=\imgwidth]{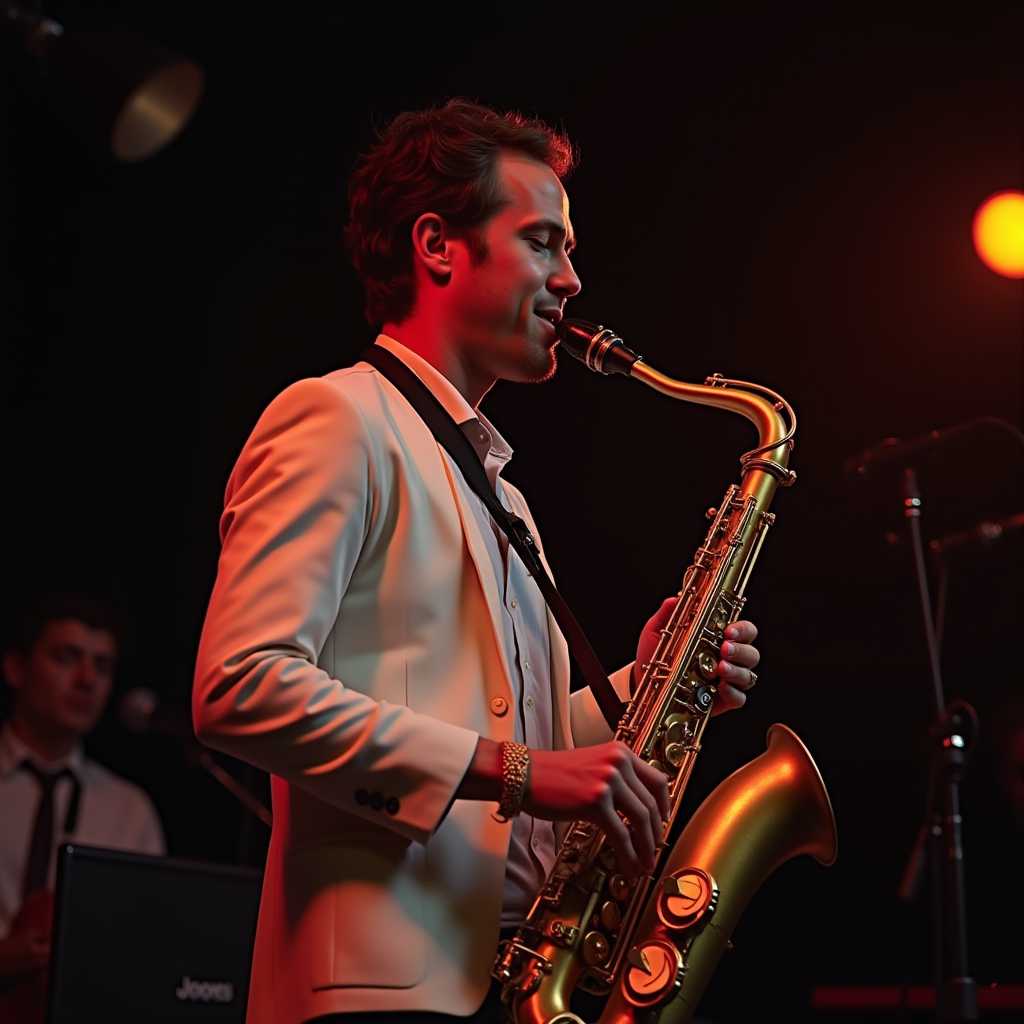} & \includegraphics[width=\imgwidth]{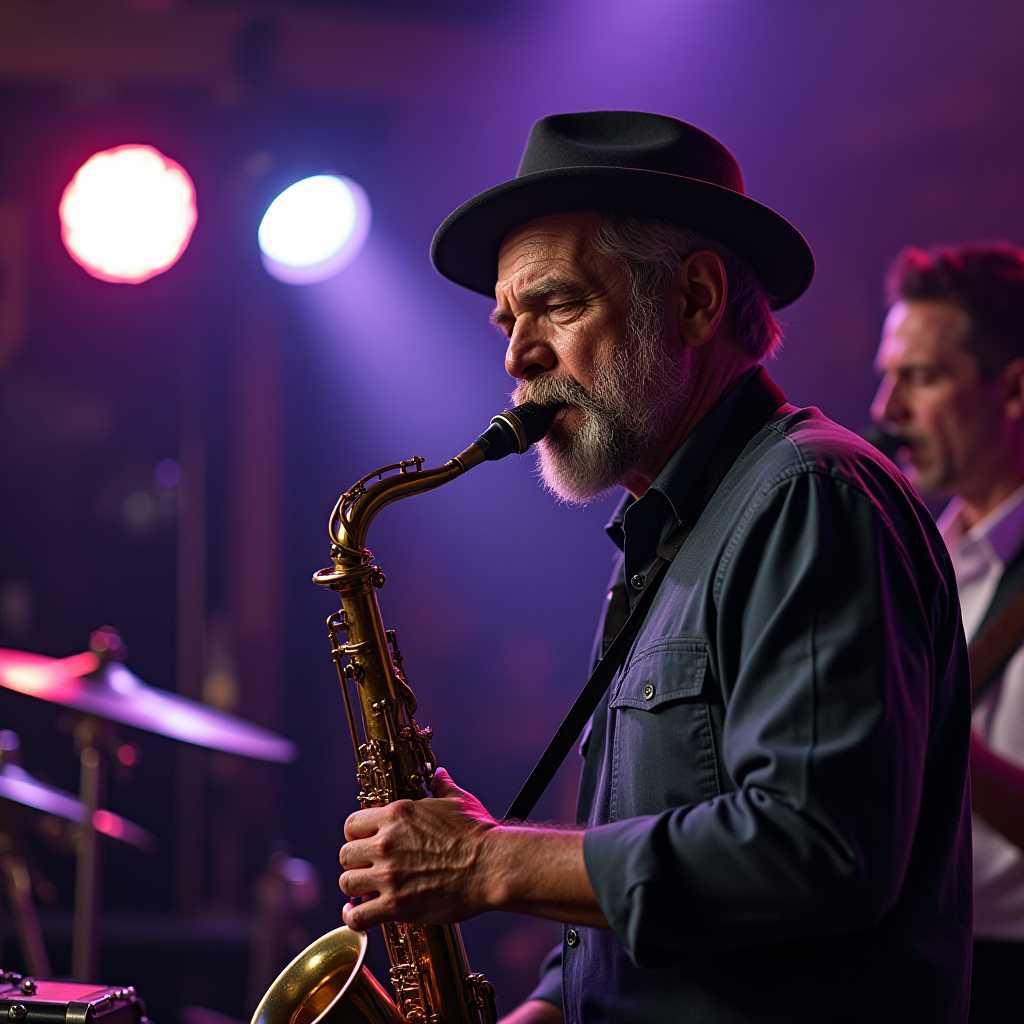} & \includegraphics[width=\imgwidth]{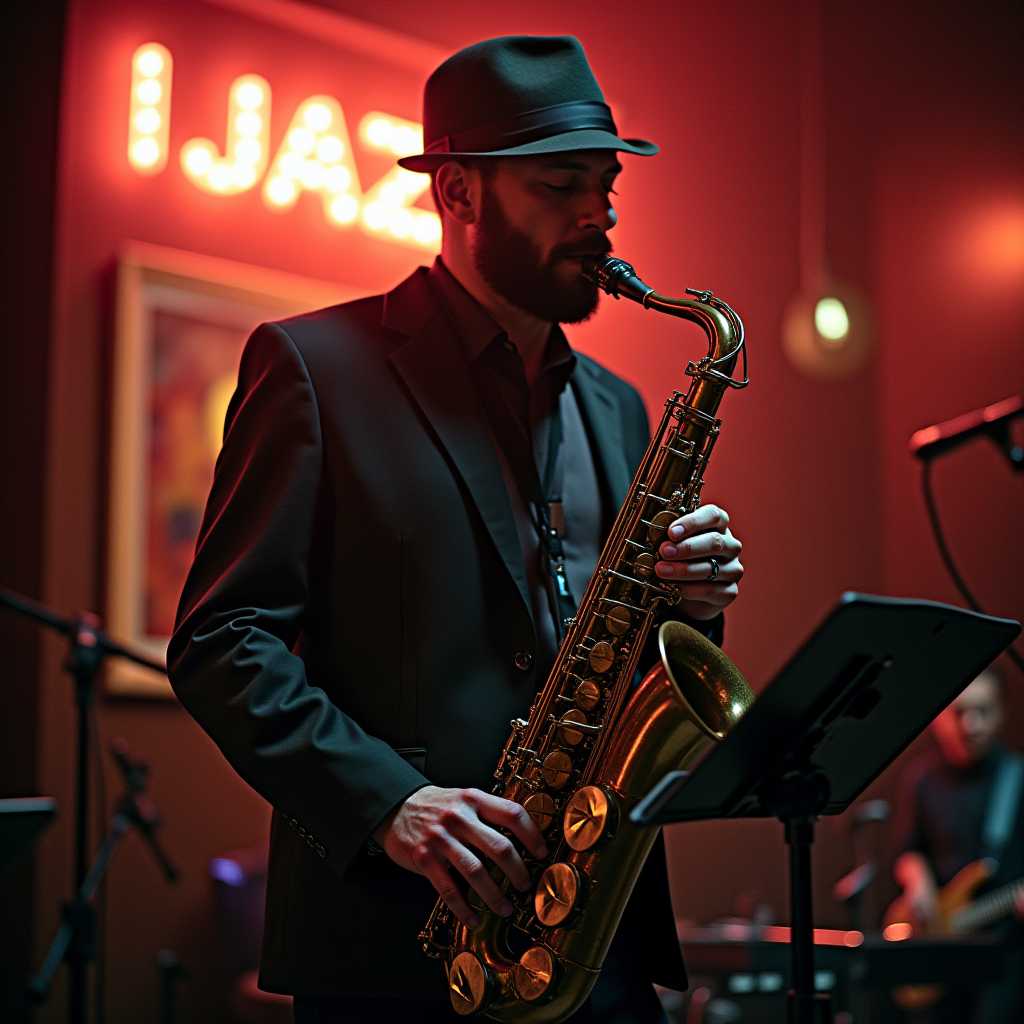} \\
        \multicolumn{9}{c}{\vspace{2pt}\small ``A jazz musician playing saxophone in a dimly lit club'' \vspace{8pt}} \\

        \vertlabel{Flux} & \includegraphics[width=\imgwidth]{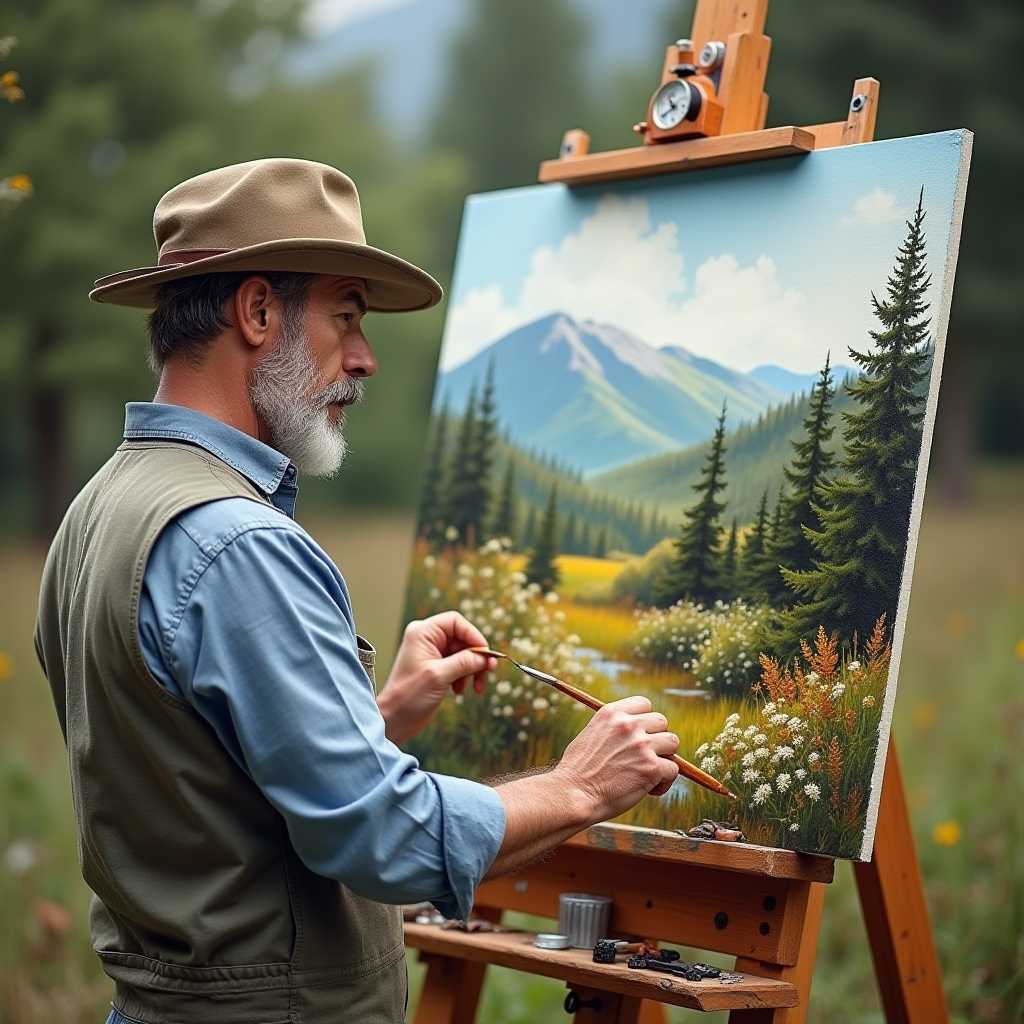} & \includegraphics[width=\imgwidth]{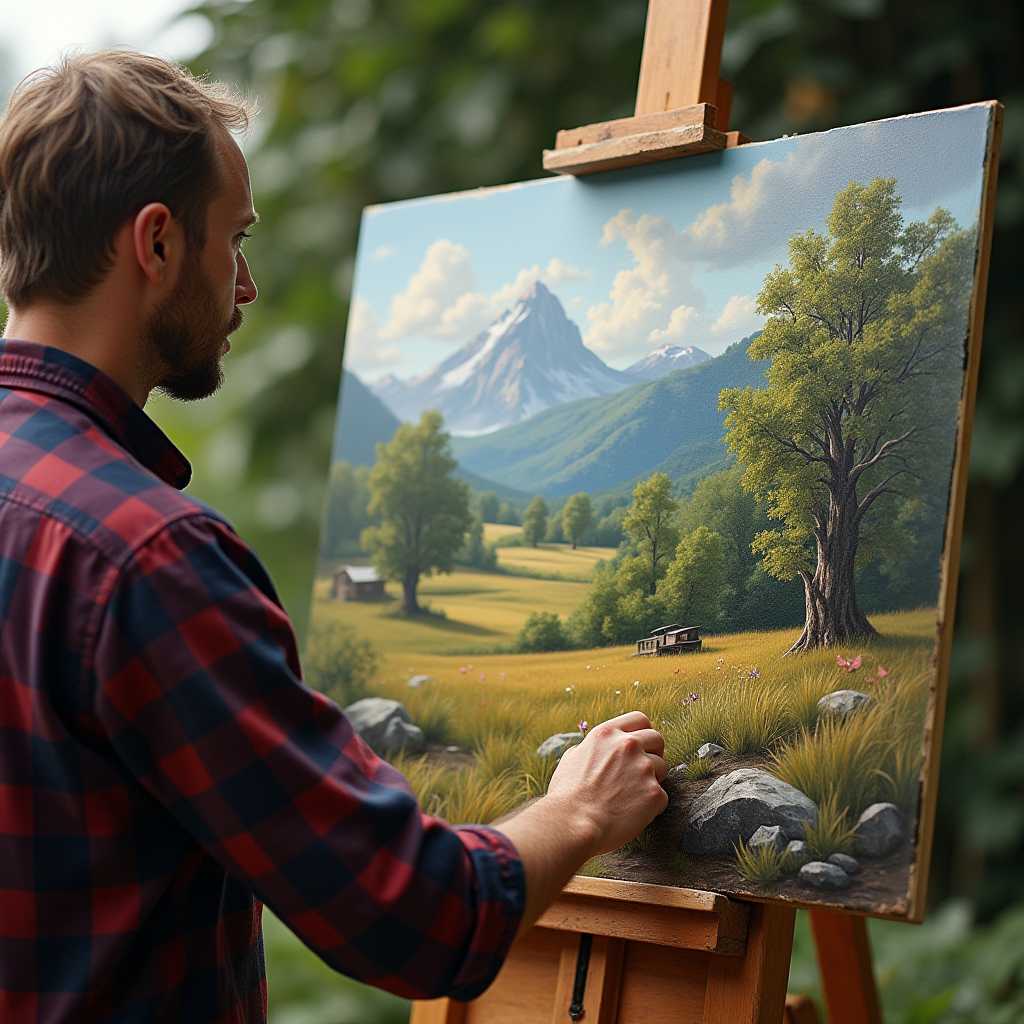} & \includegraphics[width=\imgwidth]{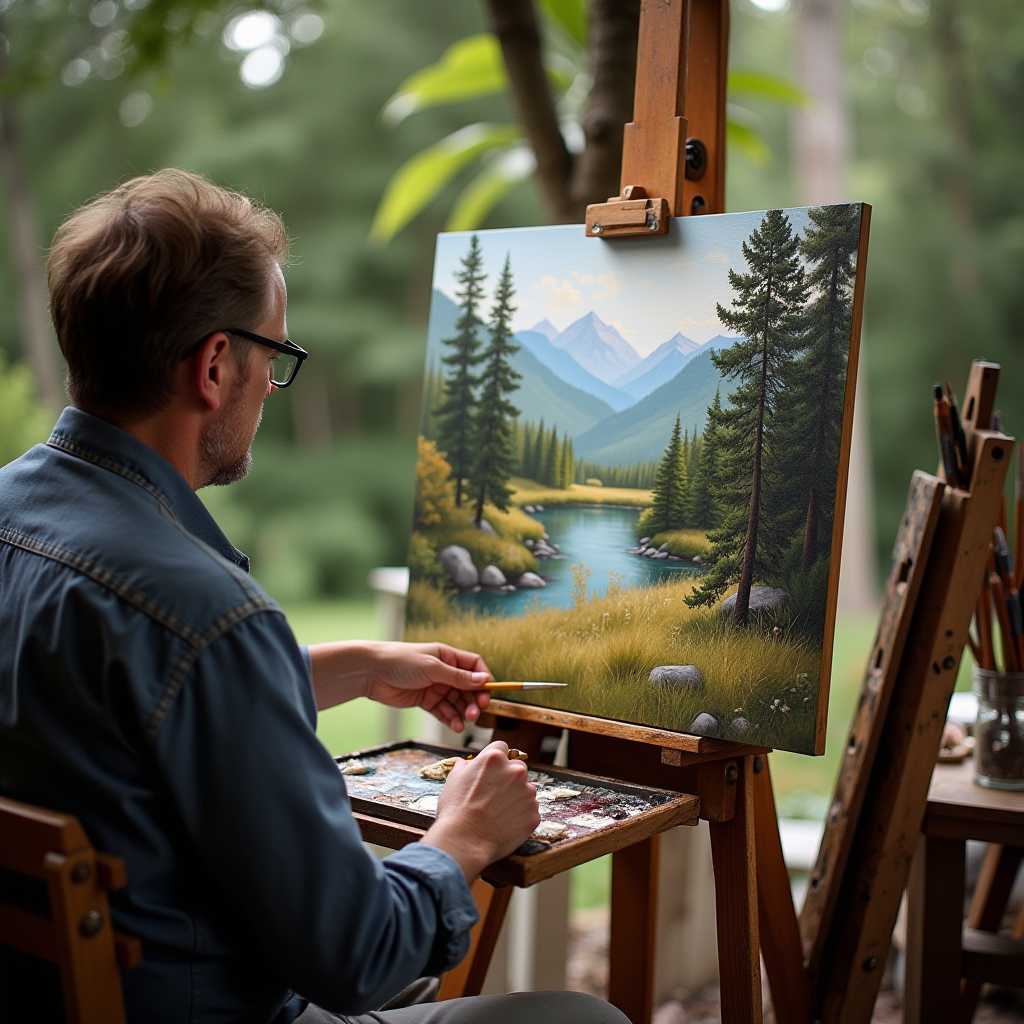} & \includegraphics[width=\imgwidth]{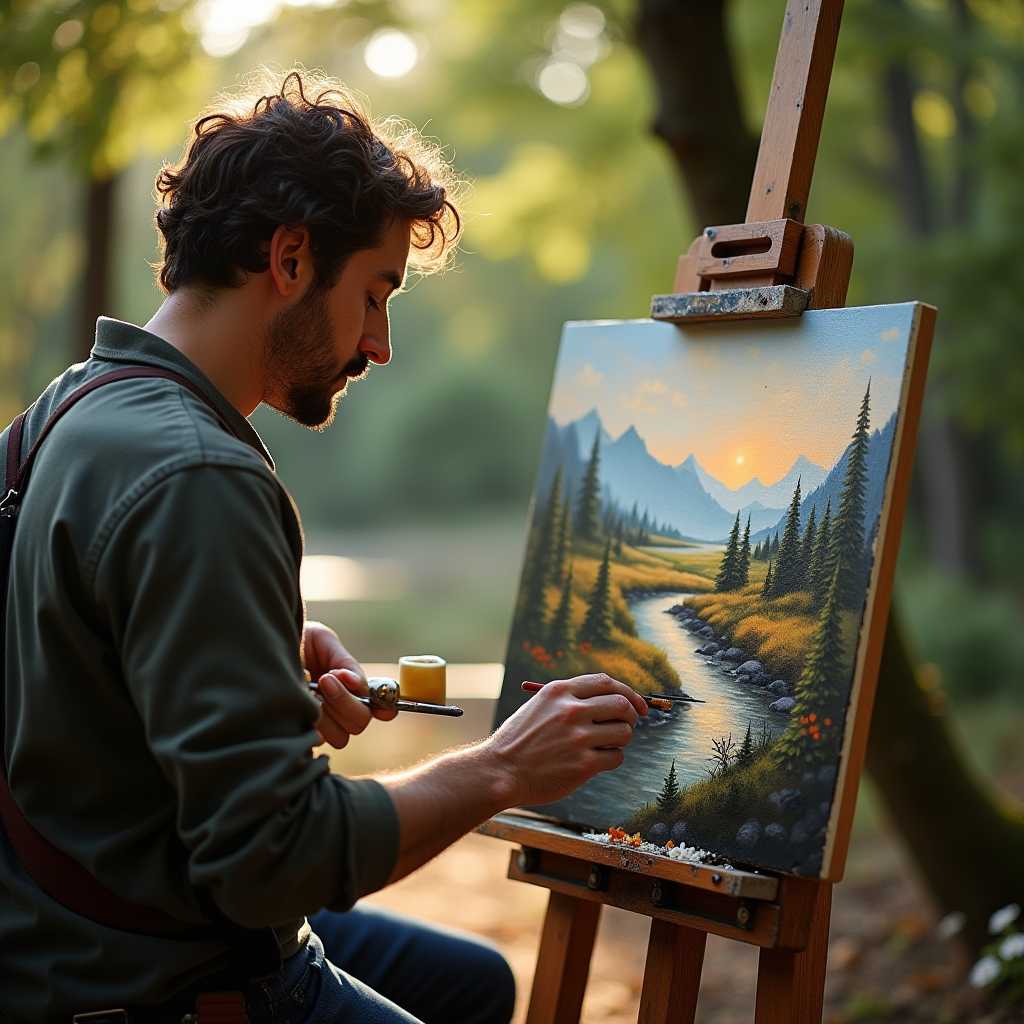} & \includegraphics[width=\imgwidth]{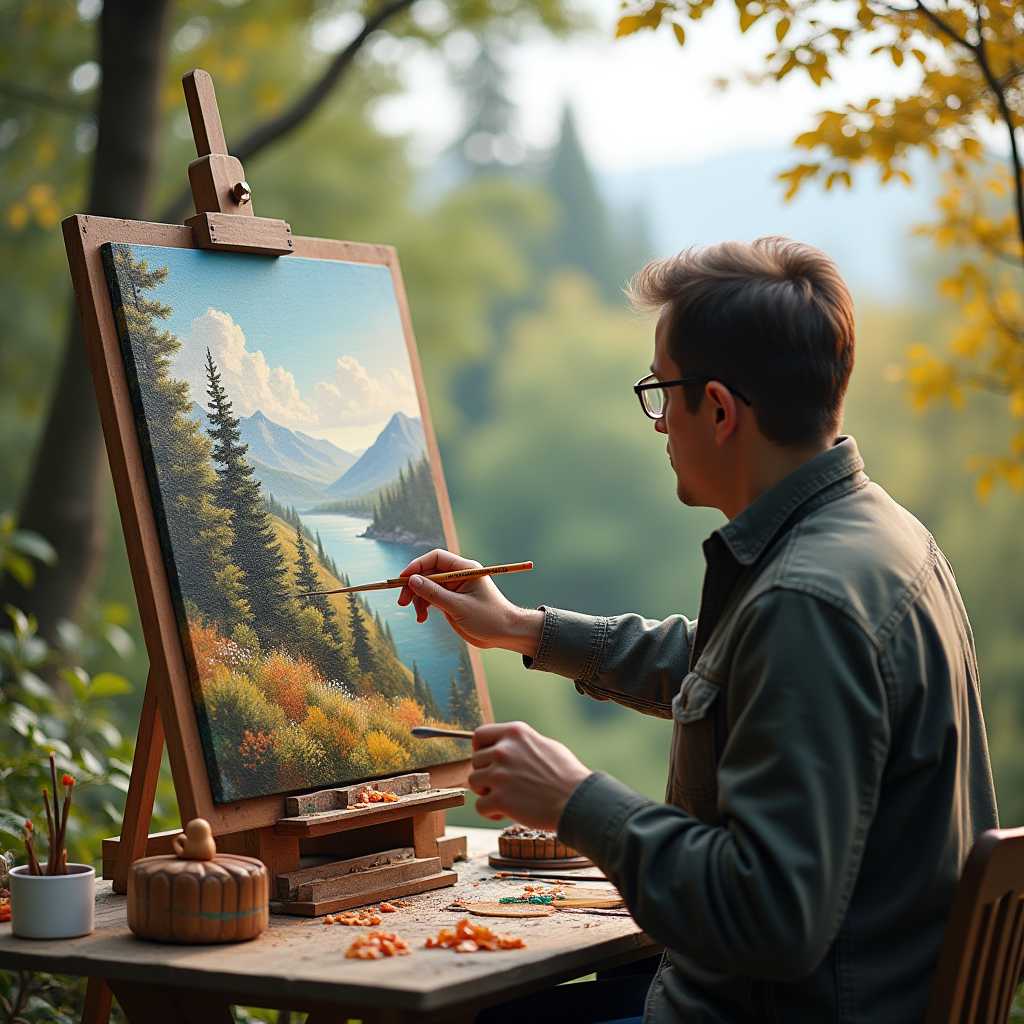} & \includegraphics[width=\imgwidth]{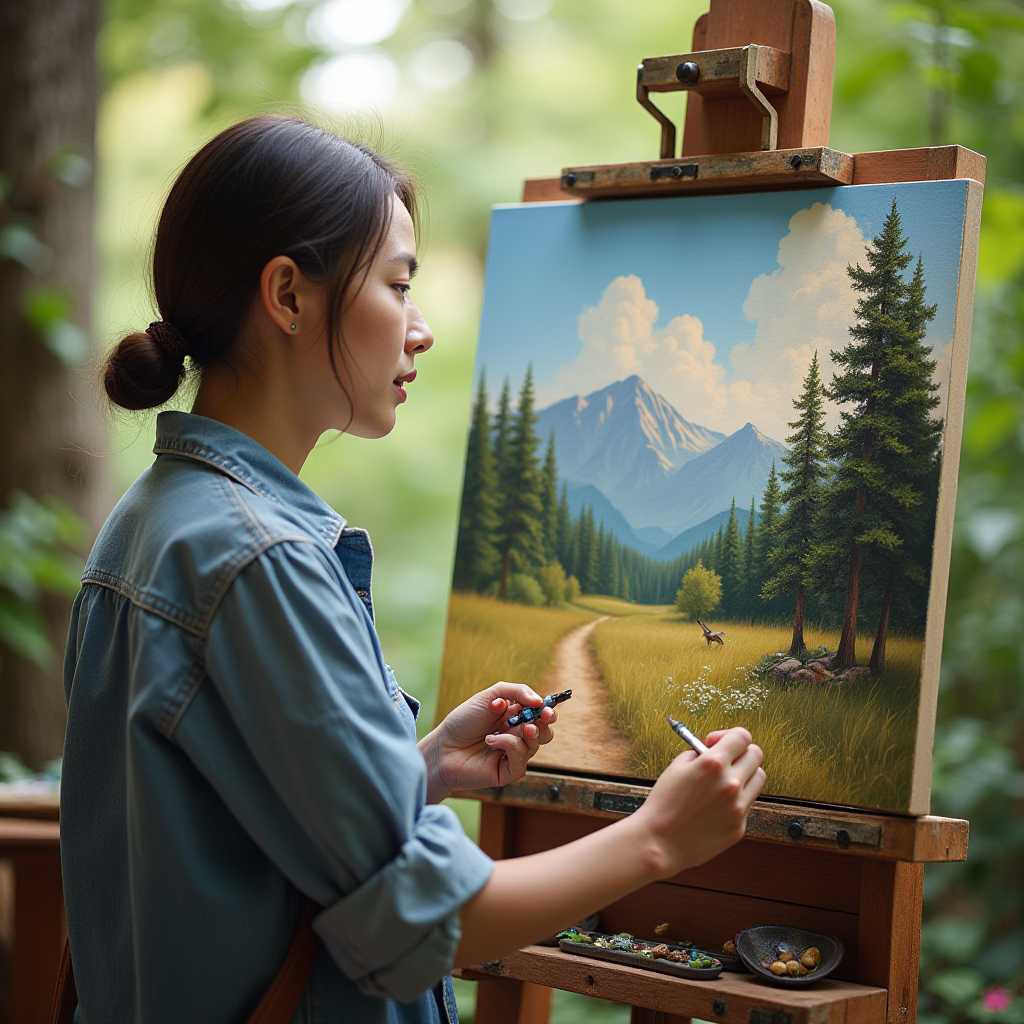} & \includegraphics[width=\imgwidth]{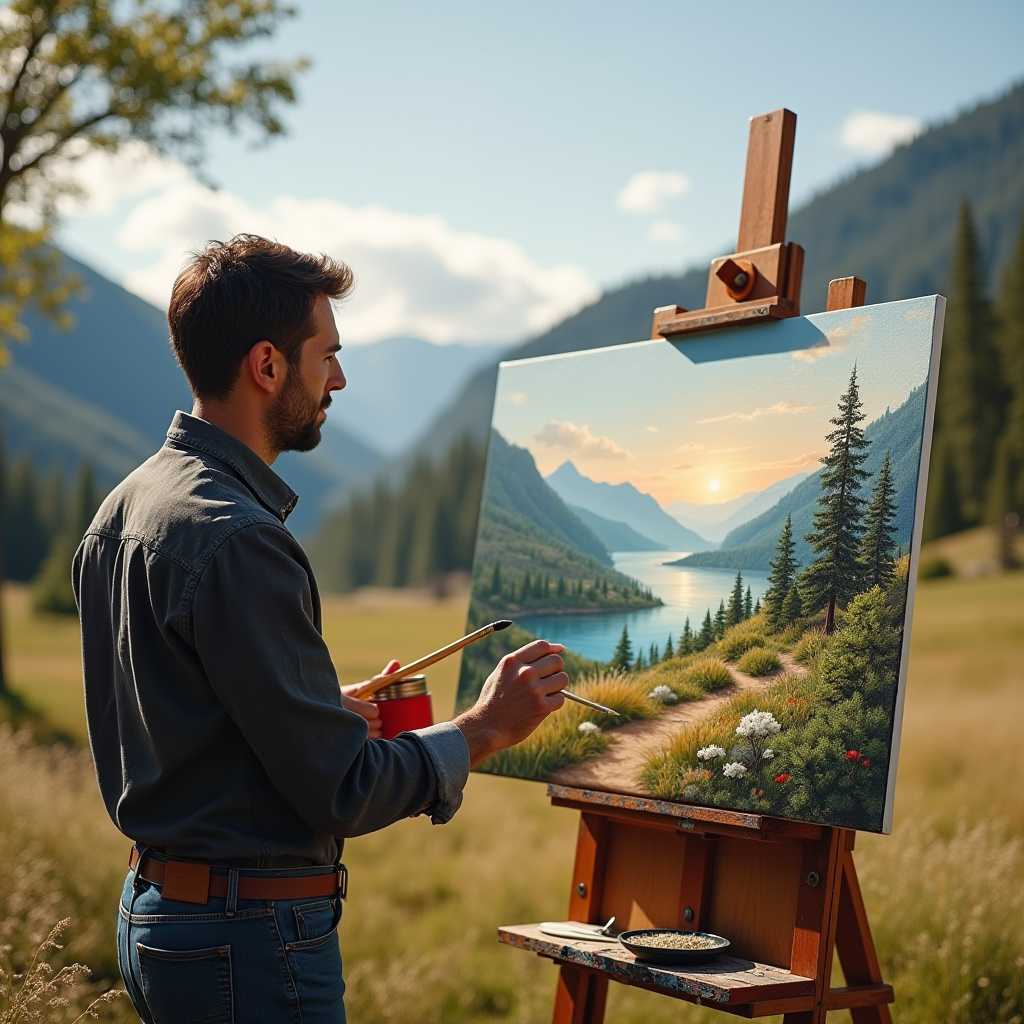} & \includegraphics[width=\imgwidth]{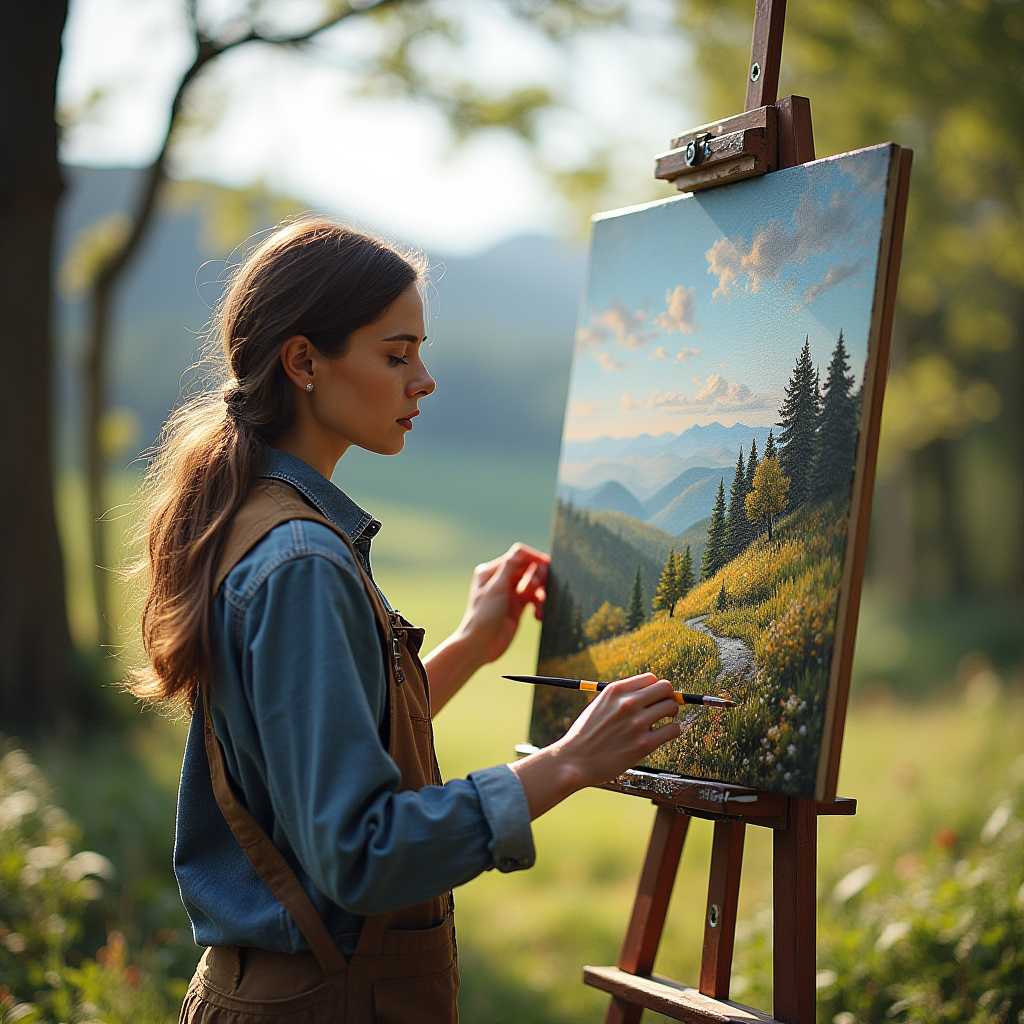} \\[-1pt]
        \vertlabel{Ours} & \includegraphics[width=\imgwidth]{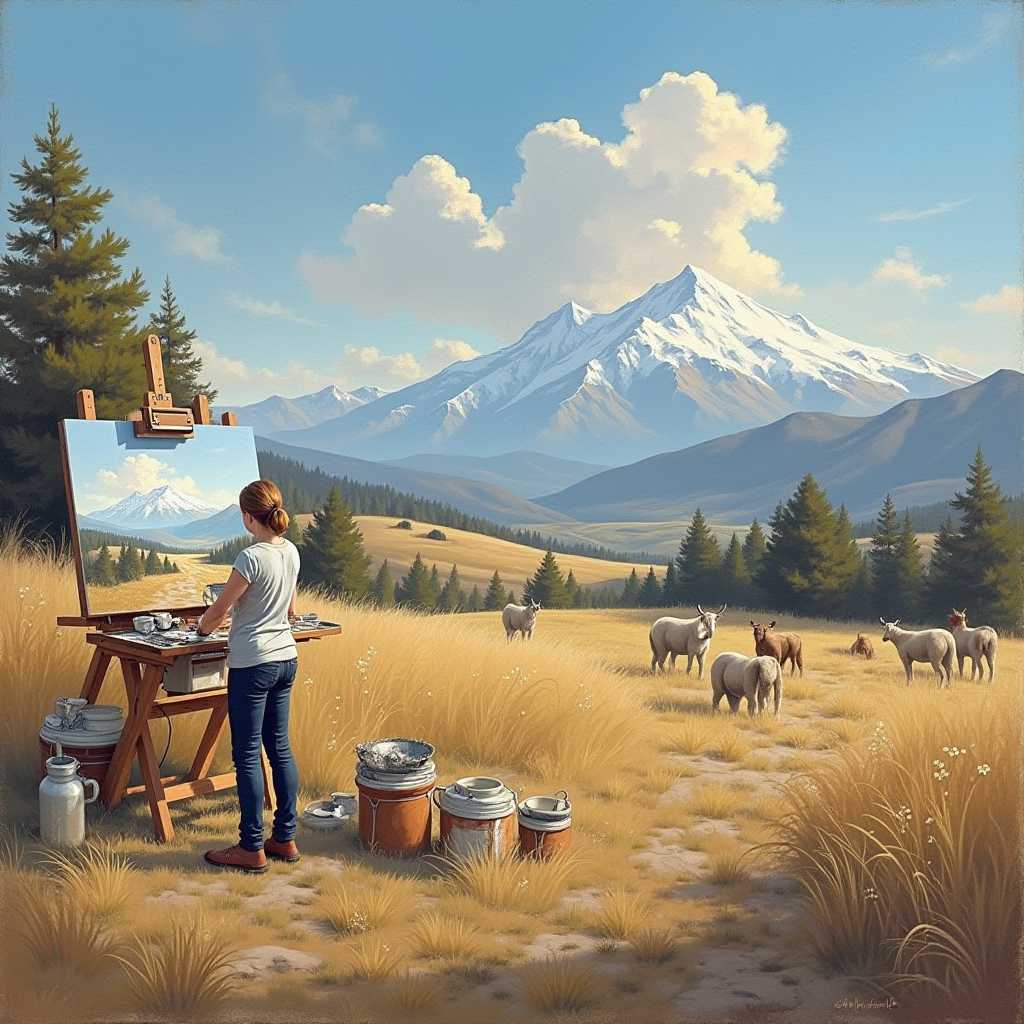} & \includegraphics[width=\imgwidth]{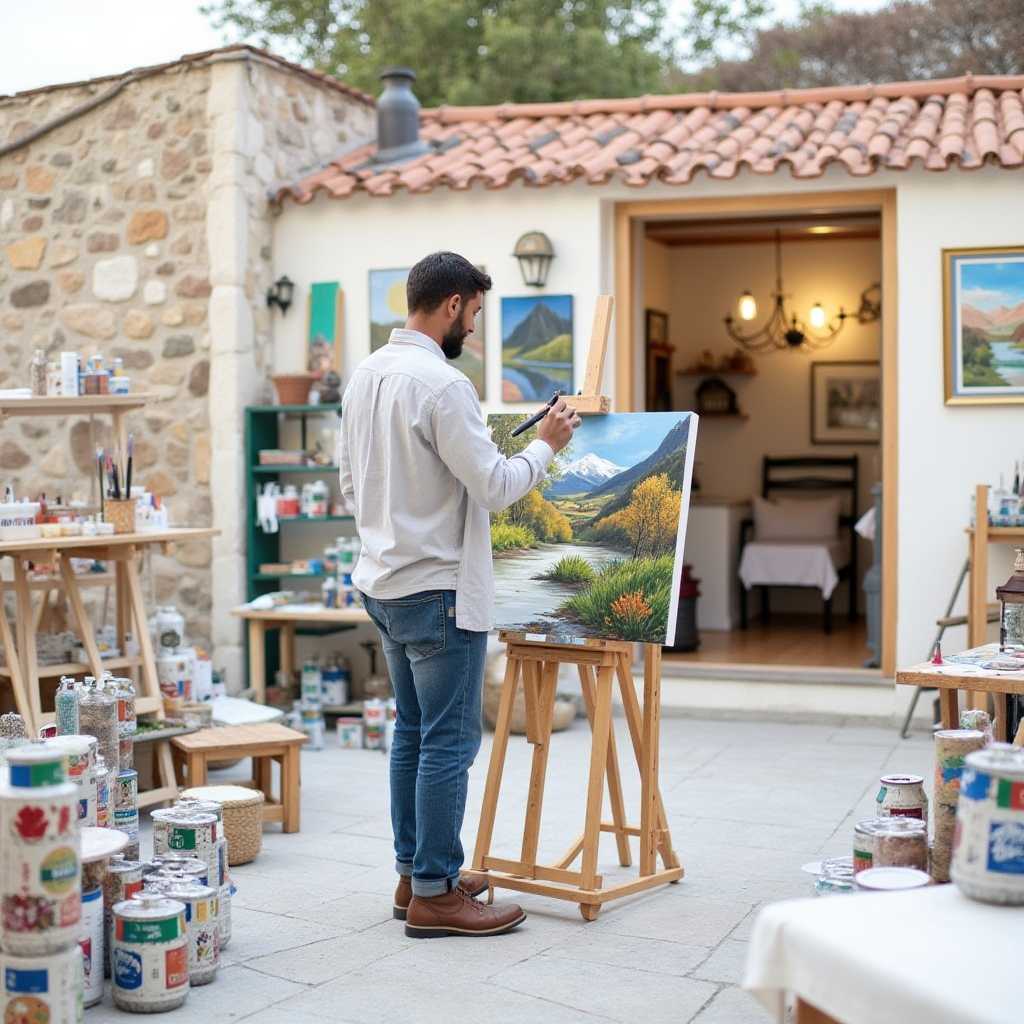} & \includegraphics[width=\imgwidth]{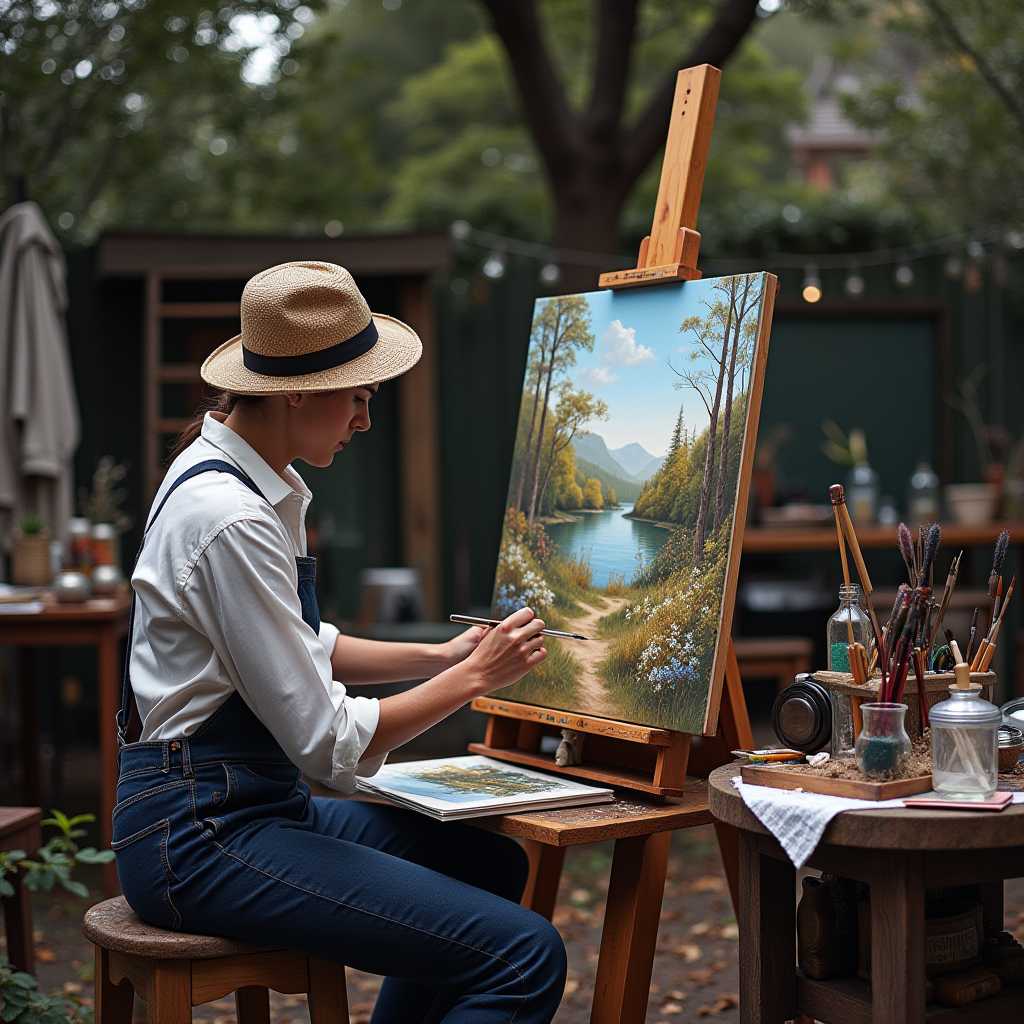} & \includegraphics[width=\imgwidth]{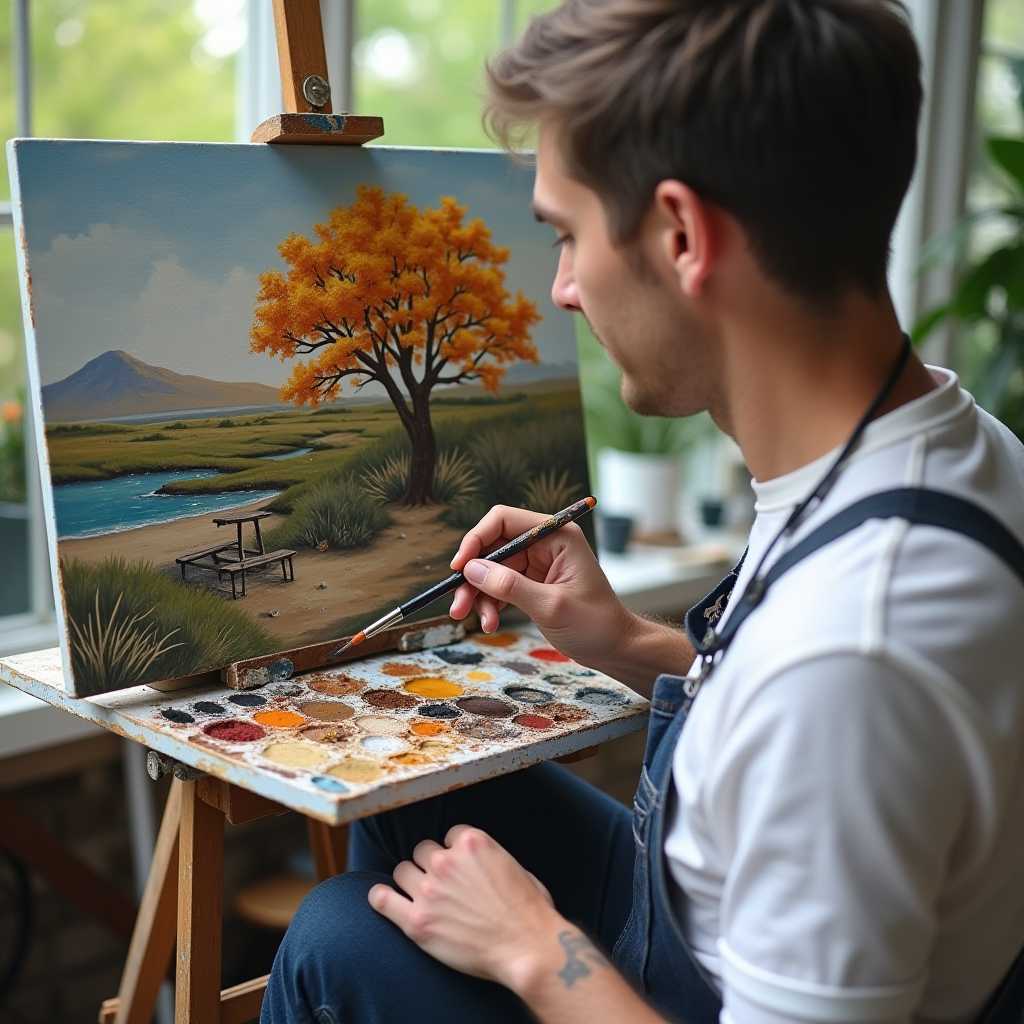} & \includegraphics[width=\imgwidth]{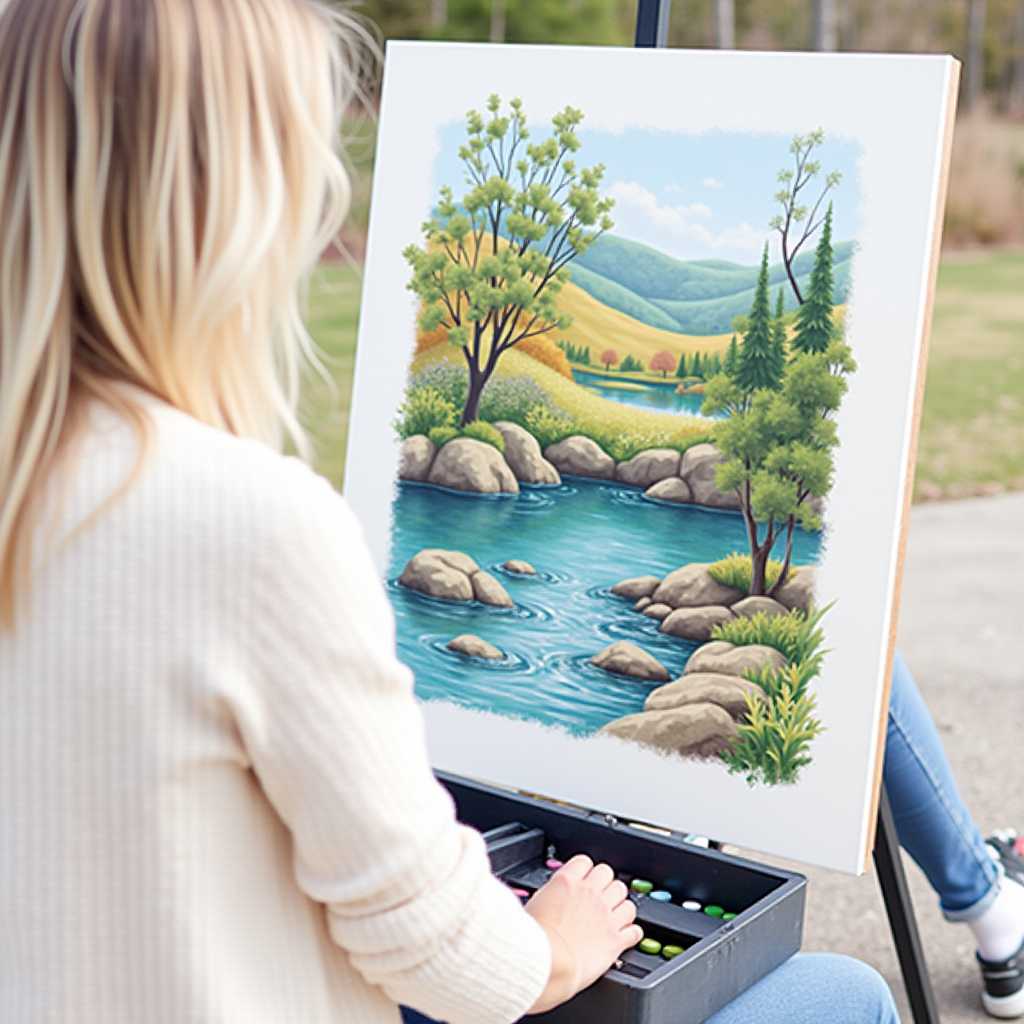} & \includegraphics[width=\imgwidth]{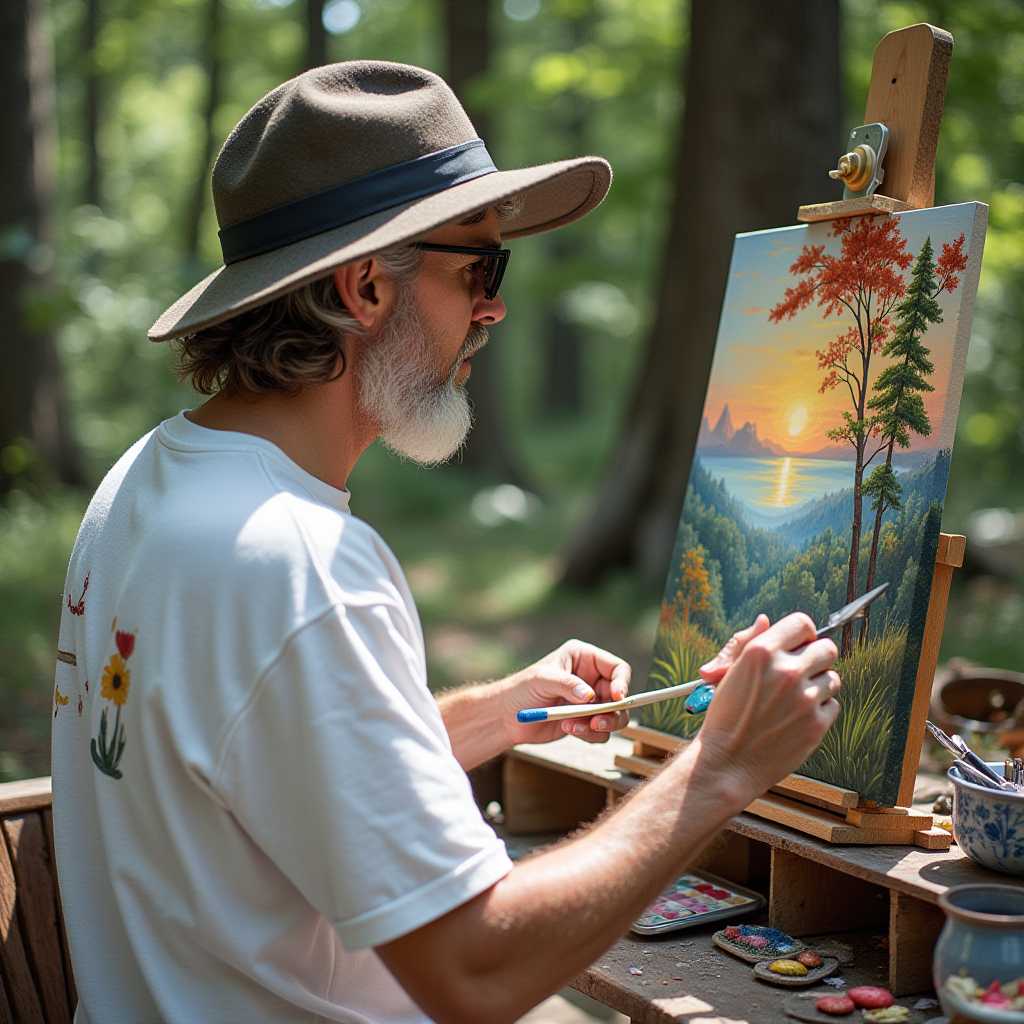} & \includegraphics[width=\imgwidth]{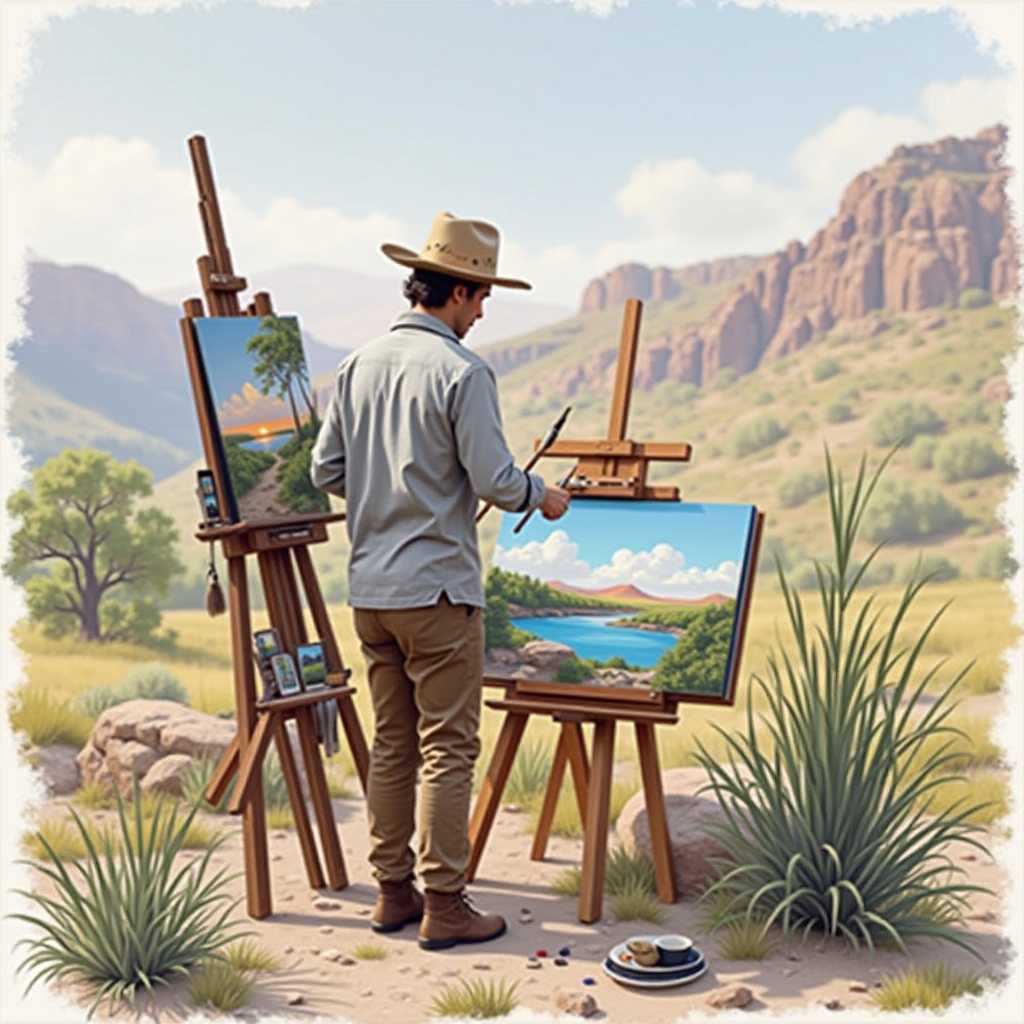} & \includegraphics[width=\imgwidth]{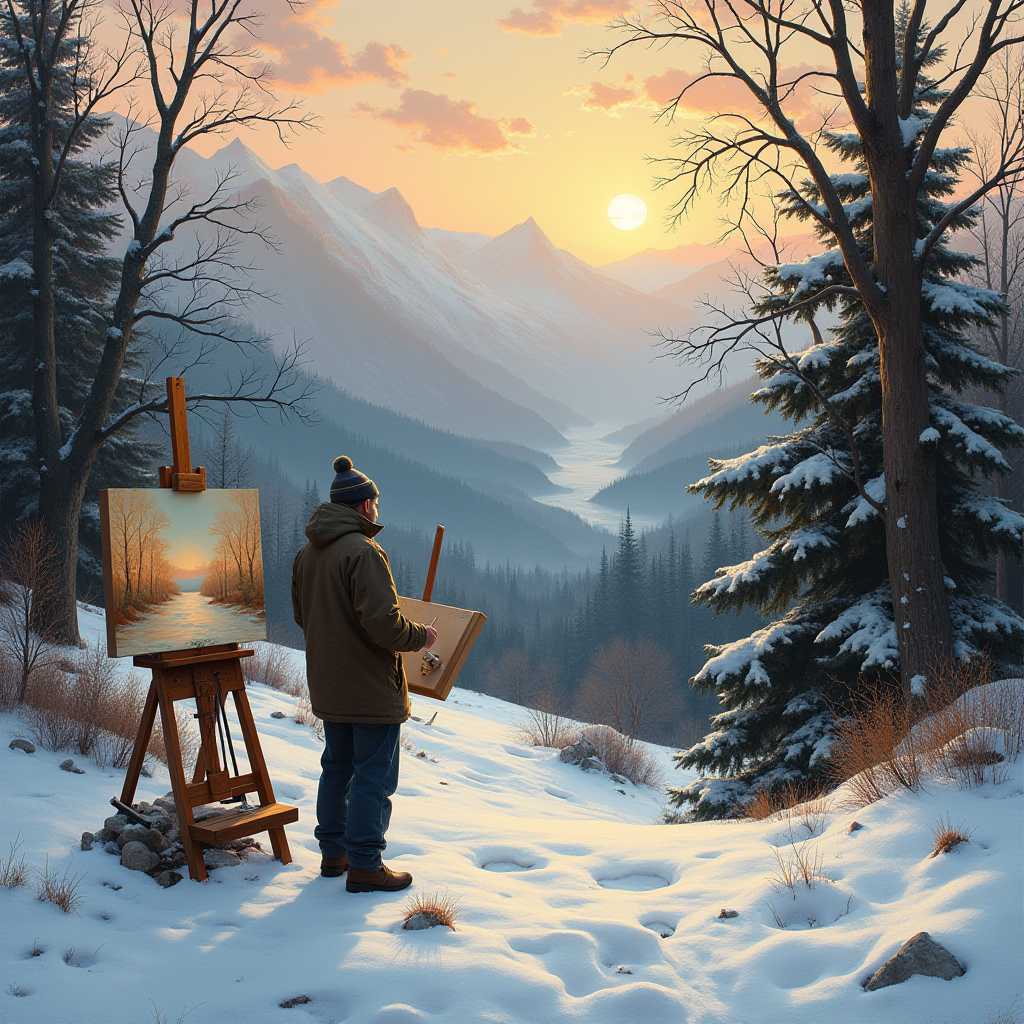} \\
        \multicolumn{9}{c}{\vspace{2pt}\small ``An artist painting a landscape in an outdoor studio'' \vspace{8pt}} \\

        \vertlabel{Flux} & \includegraphics[width=\imgwidth]{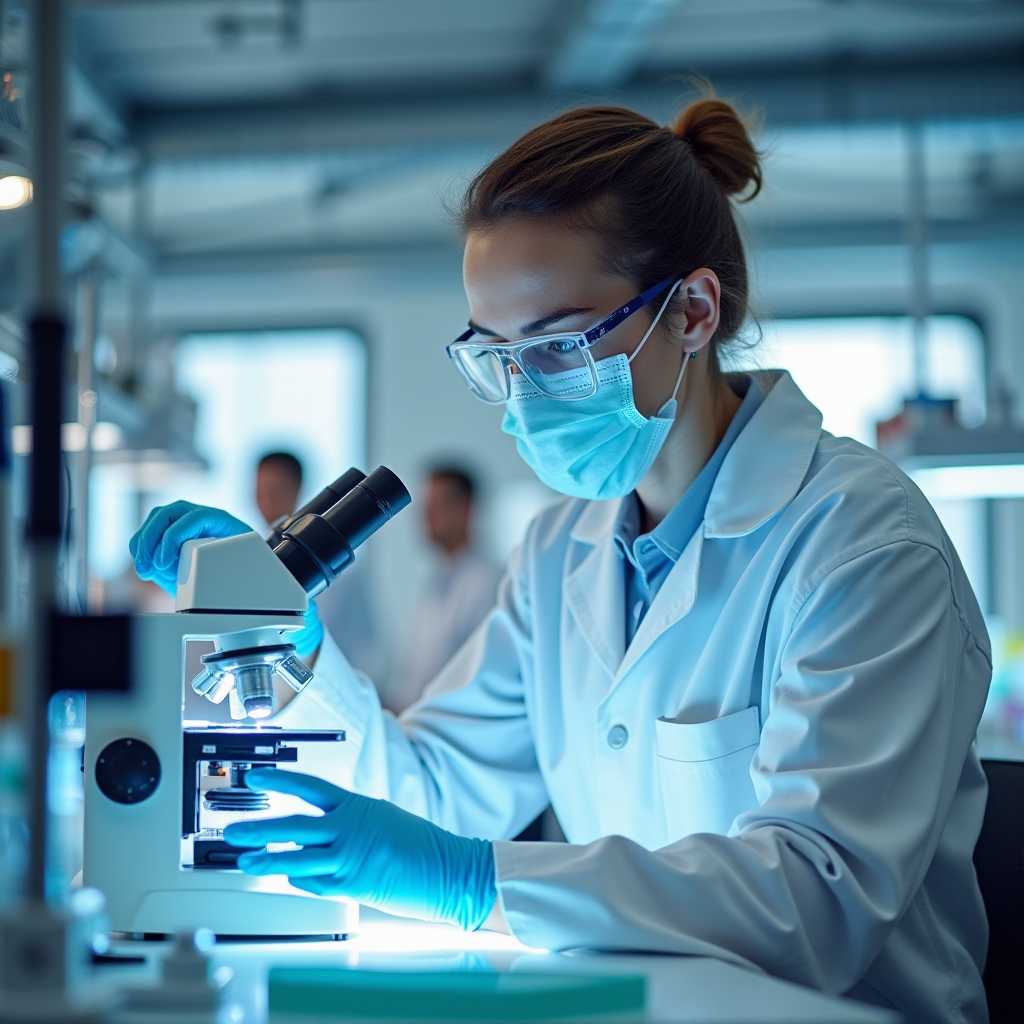} & \includegraphics[width=\imgwidth]{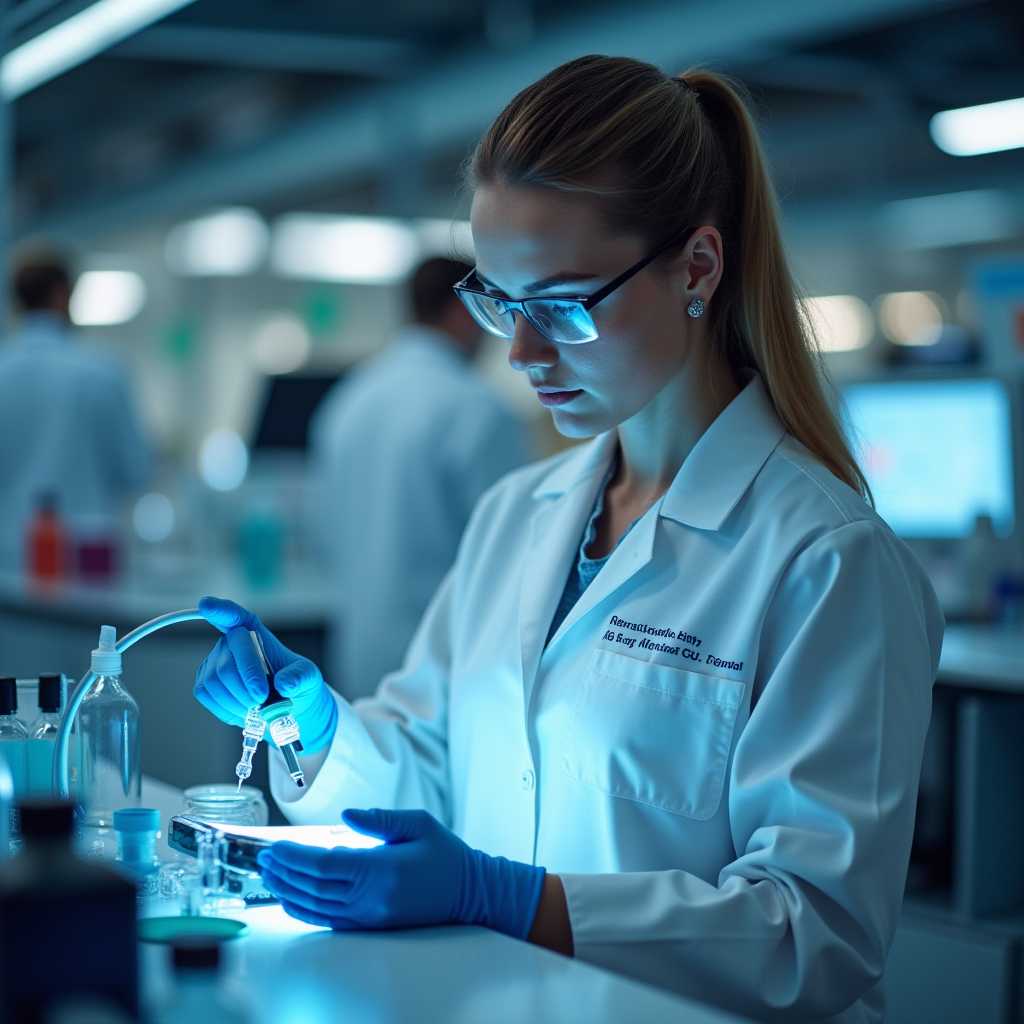} & \includegraphics[width=\imgwidth]{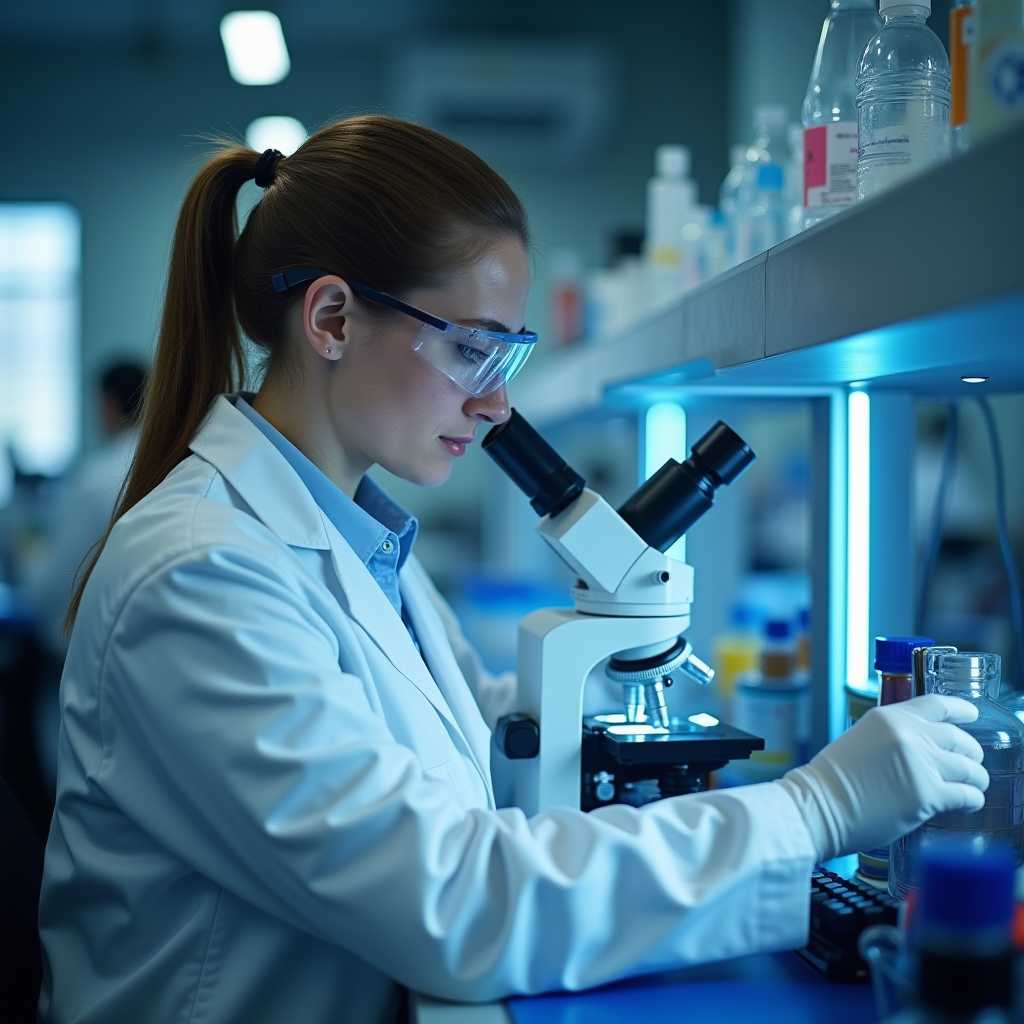} & \includegraphics[width=\imgwidth]{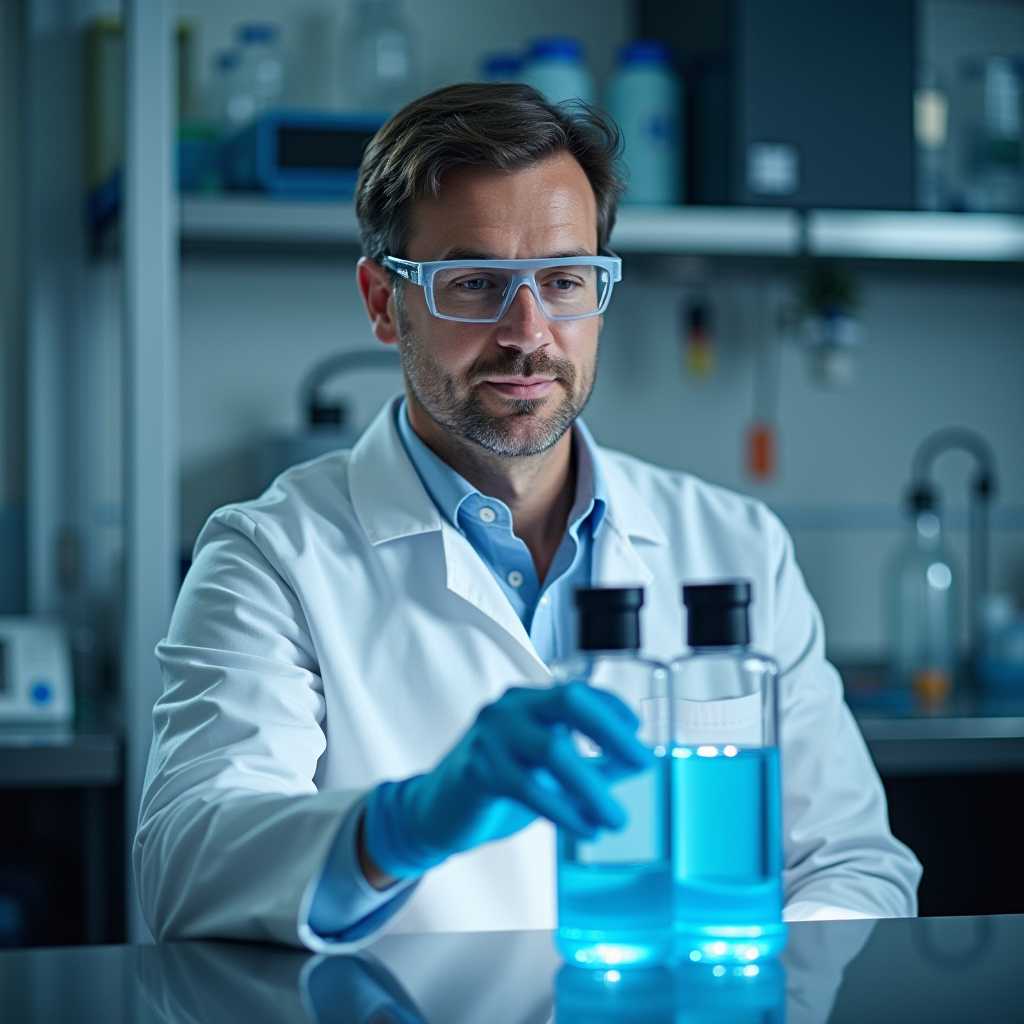} & \includegraphics[width=\imgwidth]{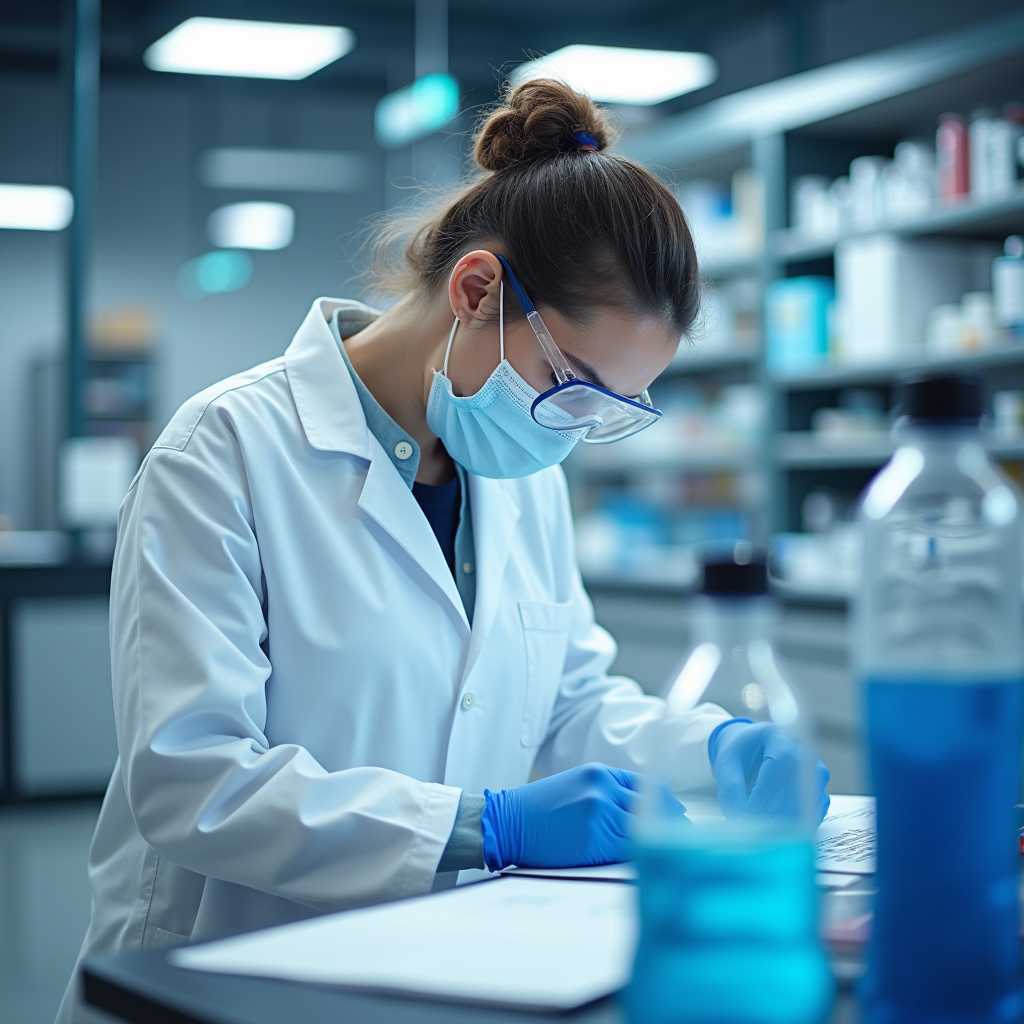} & \includegraphics[width=\imgwidth]{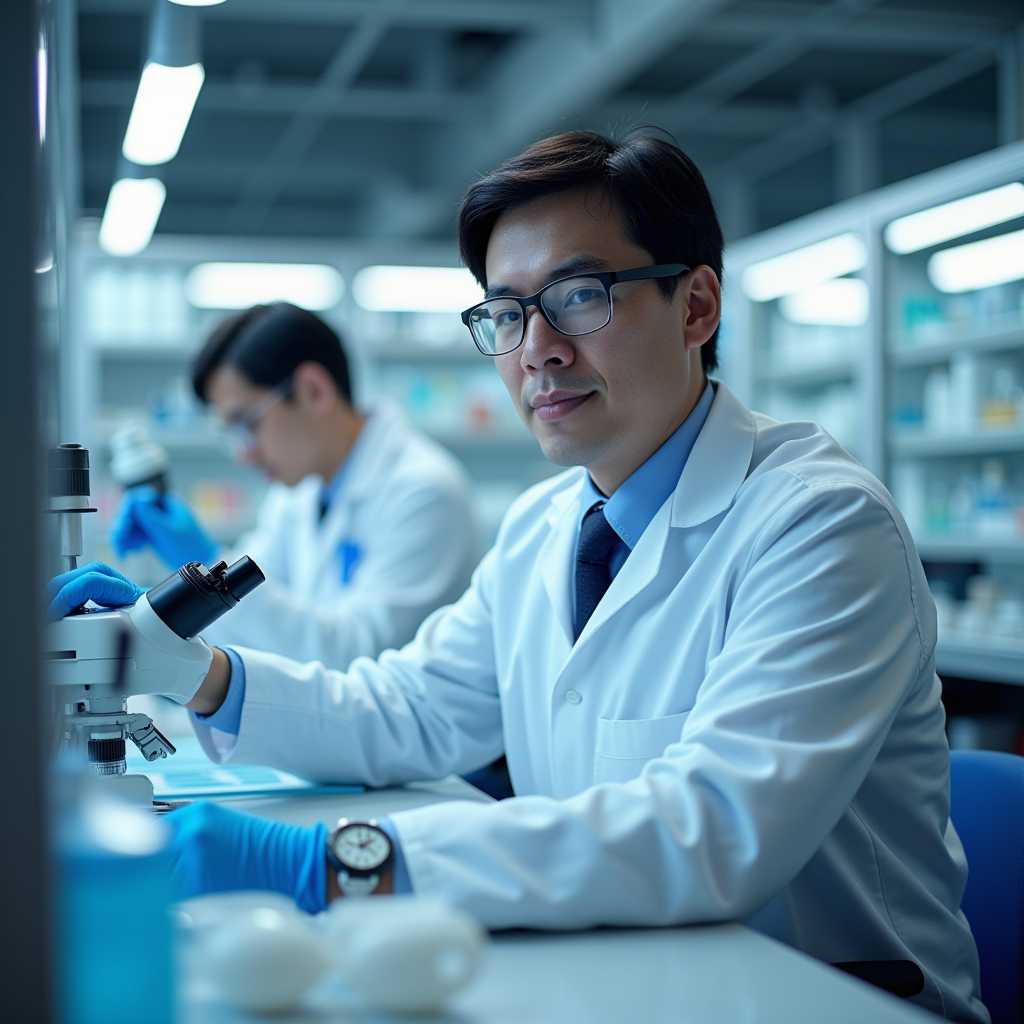} & \includegraphics[width=\imgwidth]{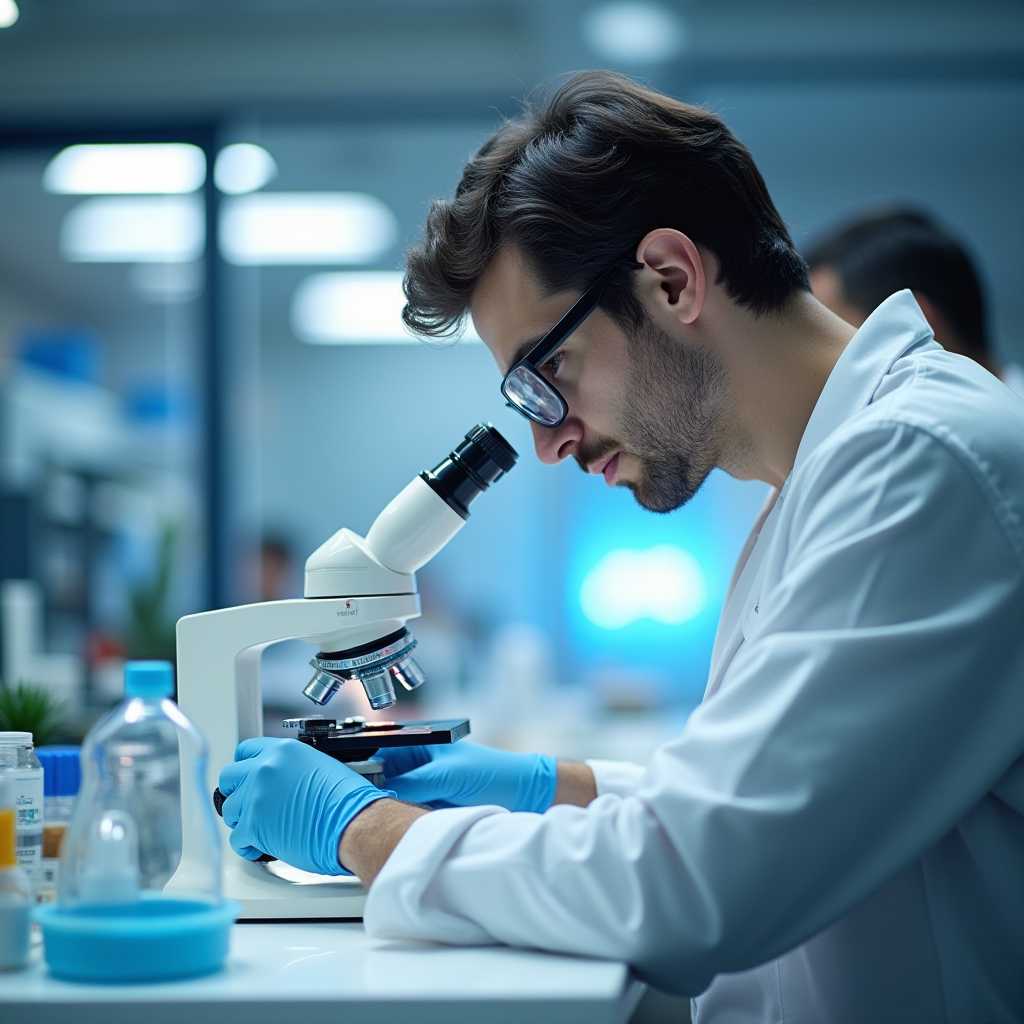} & \includegraphics[width=\imgwidth]{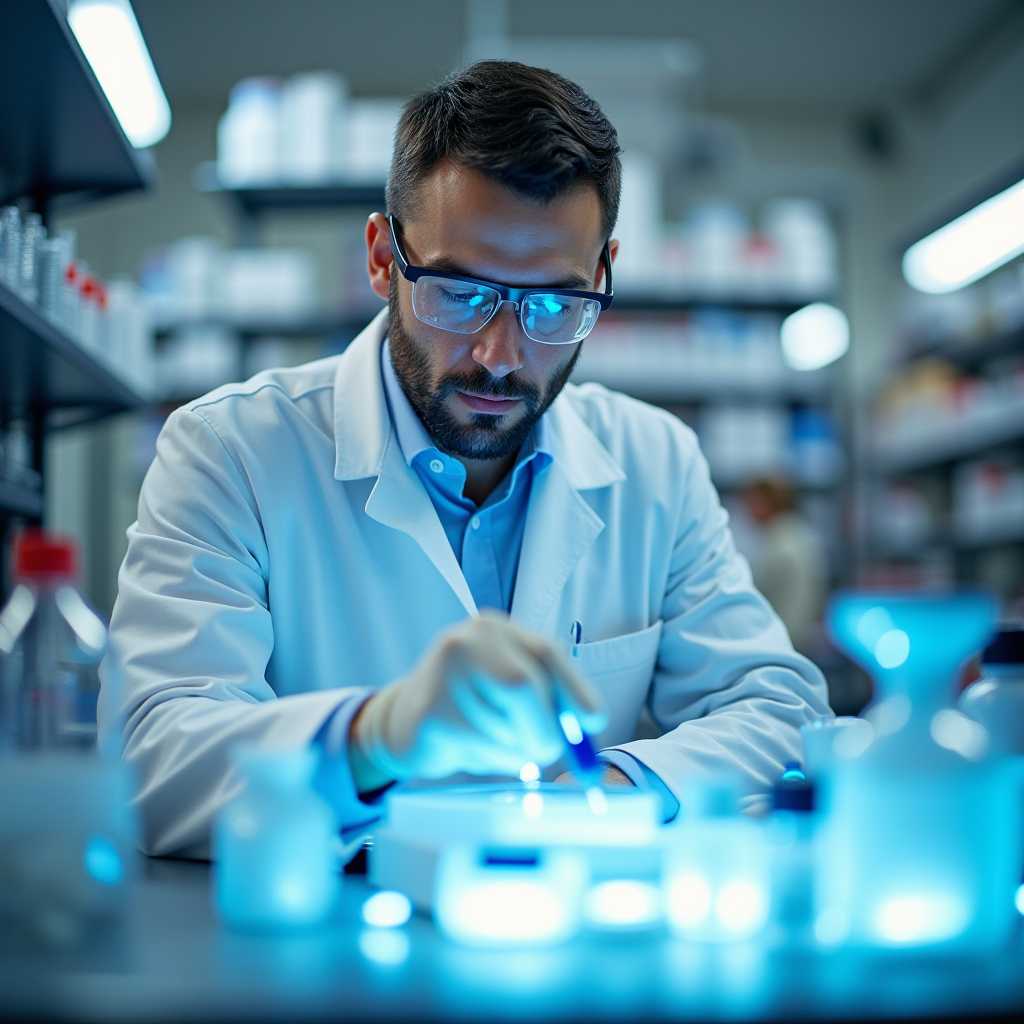} \\[-1pt]
        \vertlabel{Ours} & \includegraphics[width=\imgwidth]{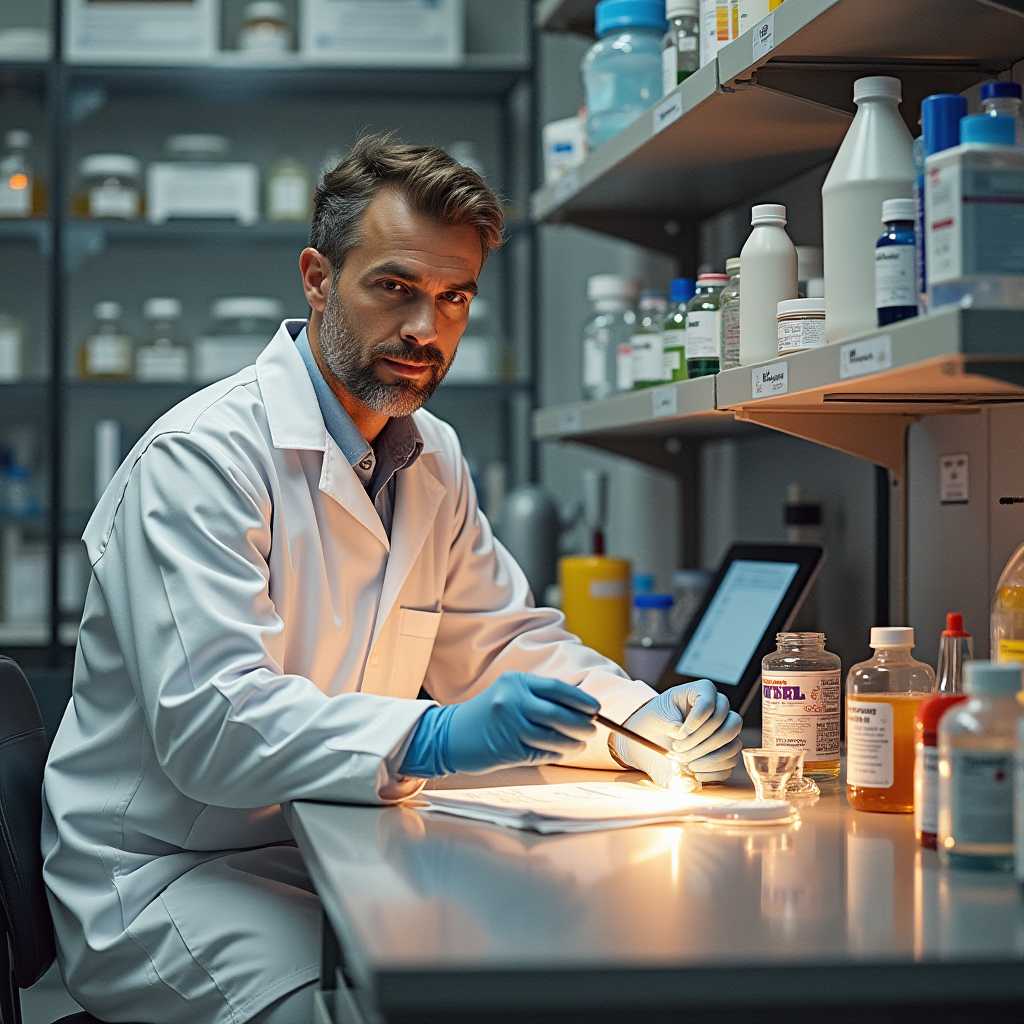} & \includegraphics[width=\imgwidth]{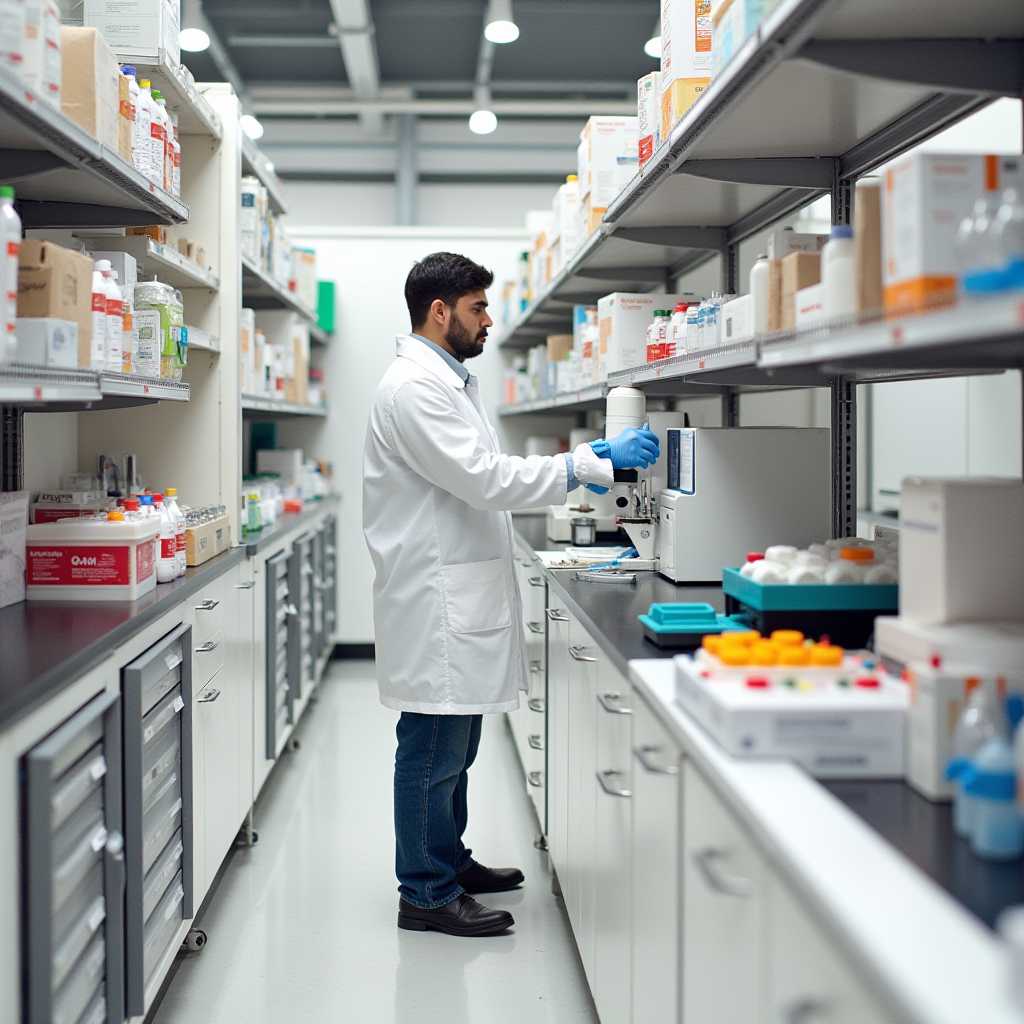} & \includegraphics[width=\imgwidth]{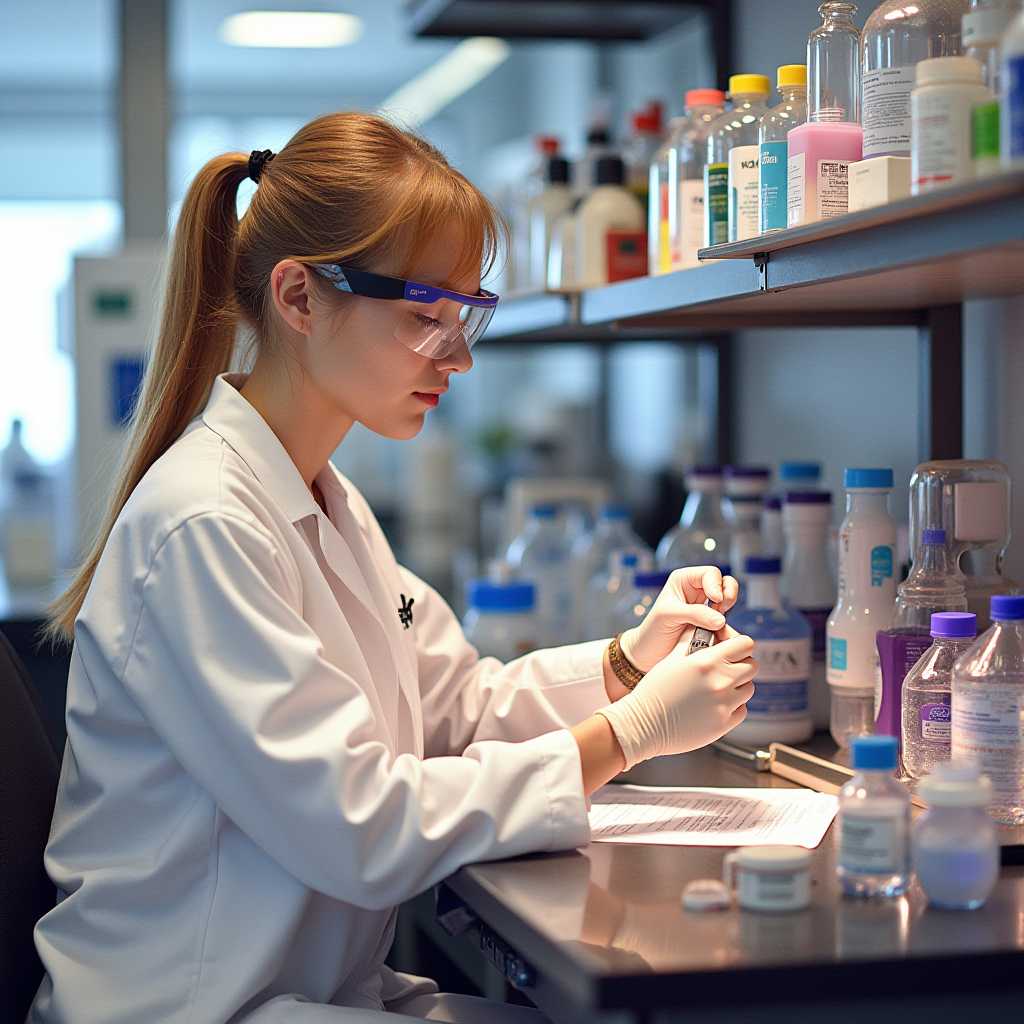} & \includegraphics[width=\imgwidth]{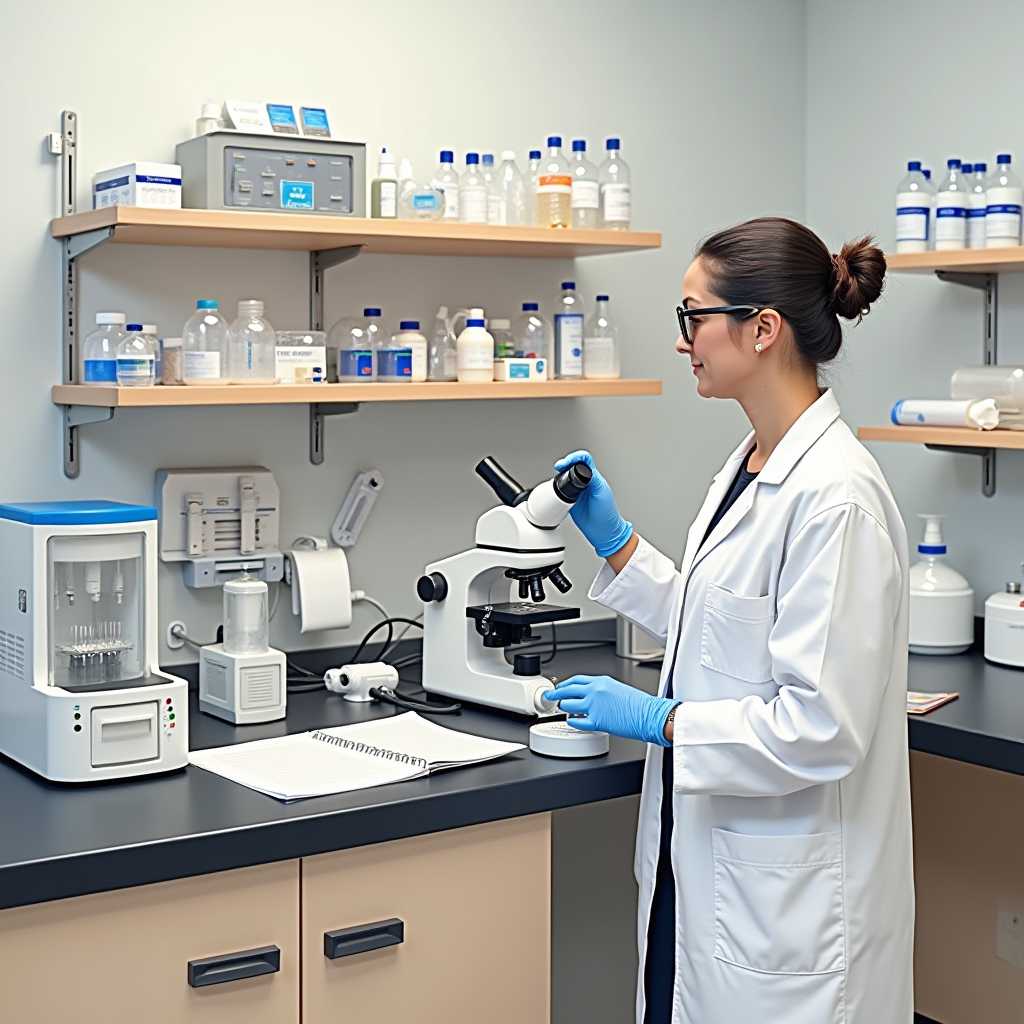} & \includegraphics[width=\imgwidth]{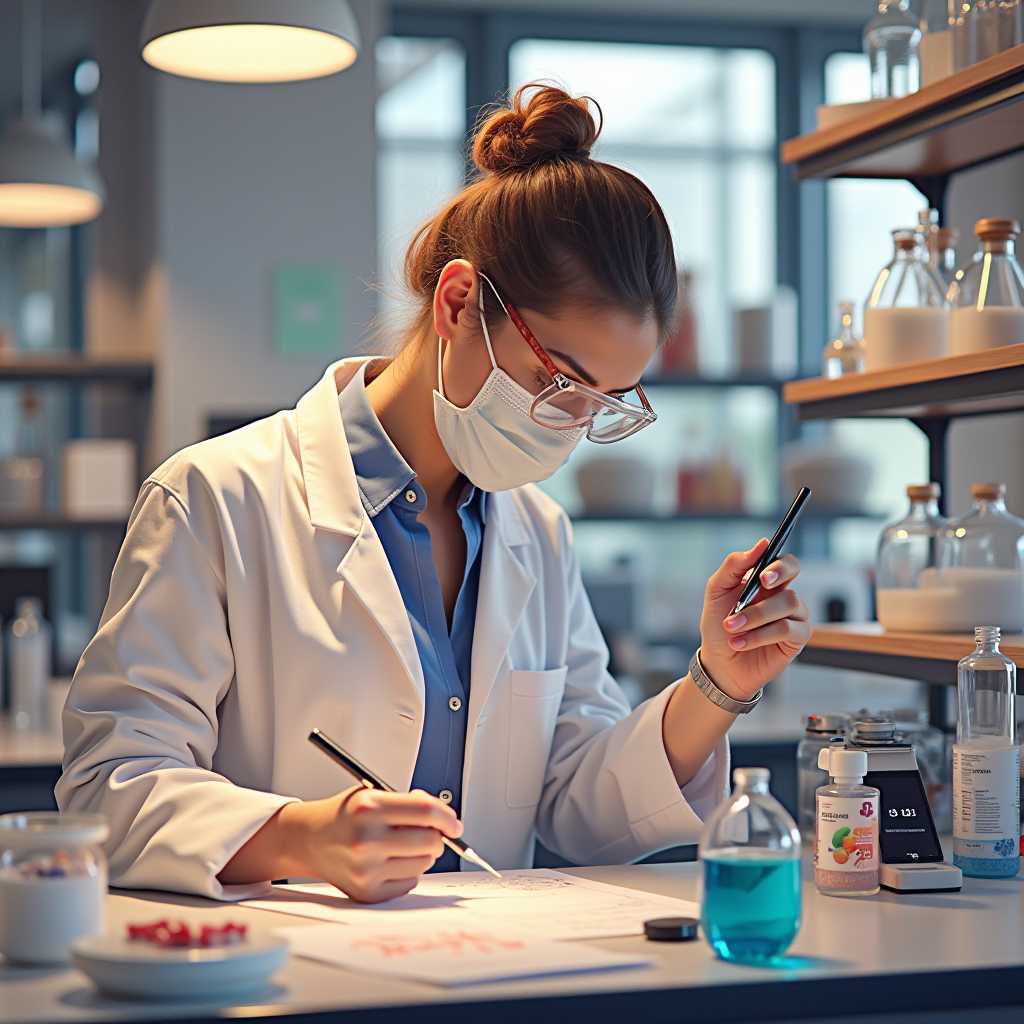} & \includegraphics[width=\imgwidth]{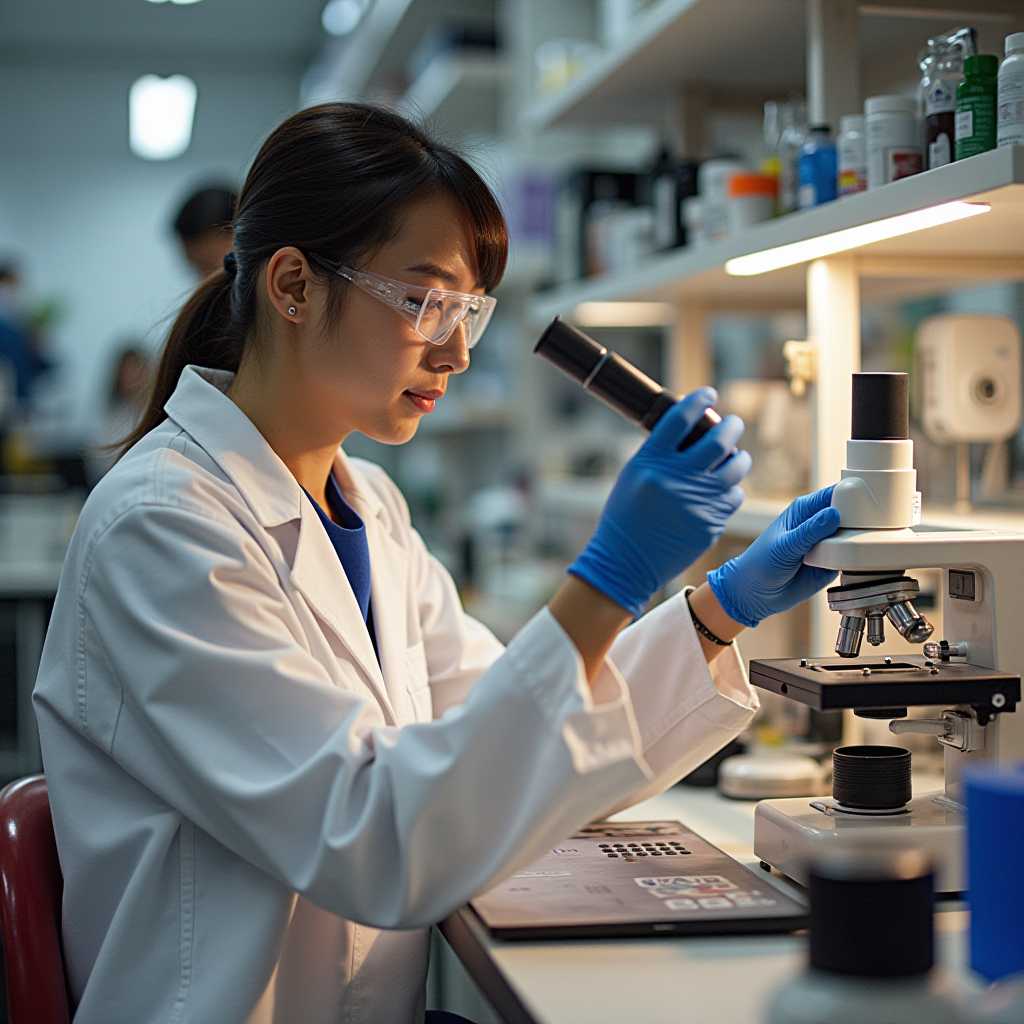} & \includegraphics[width=\imgwidth]{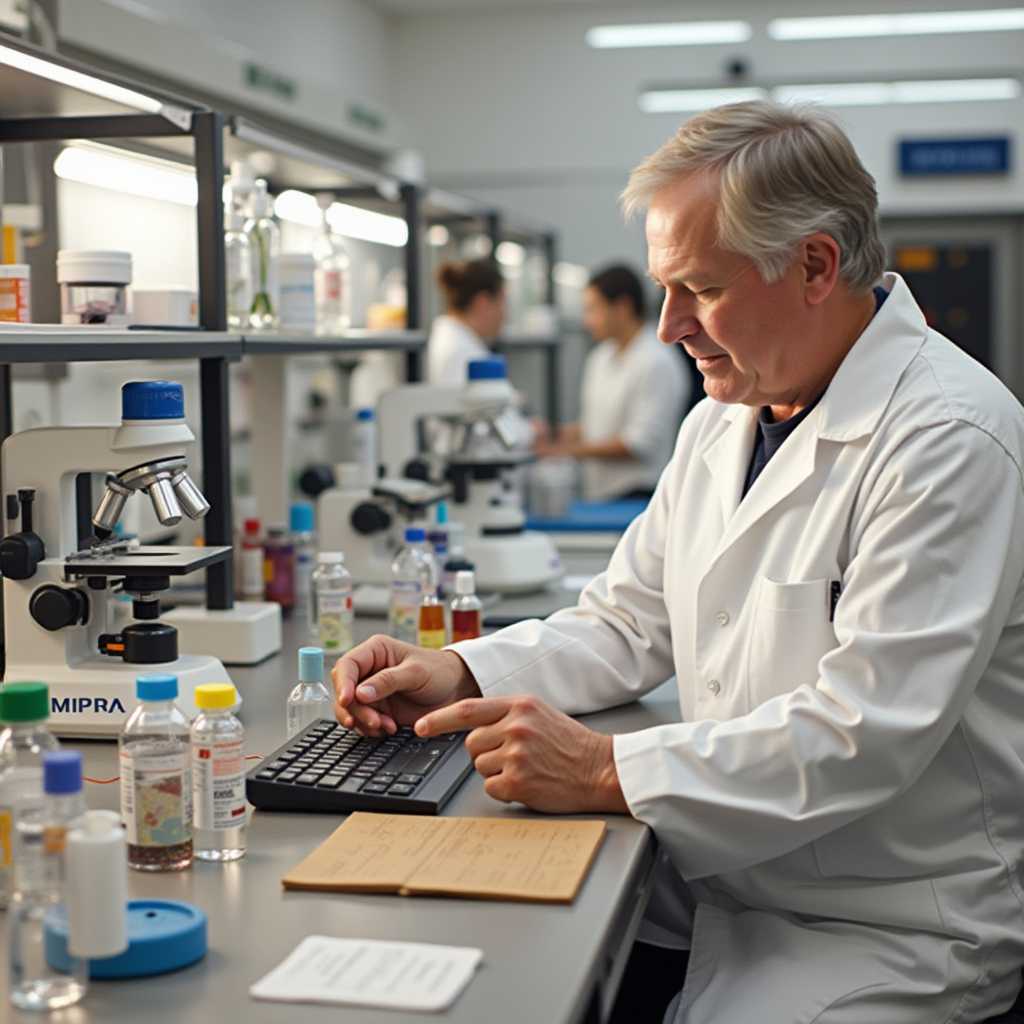} & \includegraphics[width=\imgwidth]{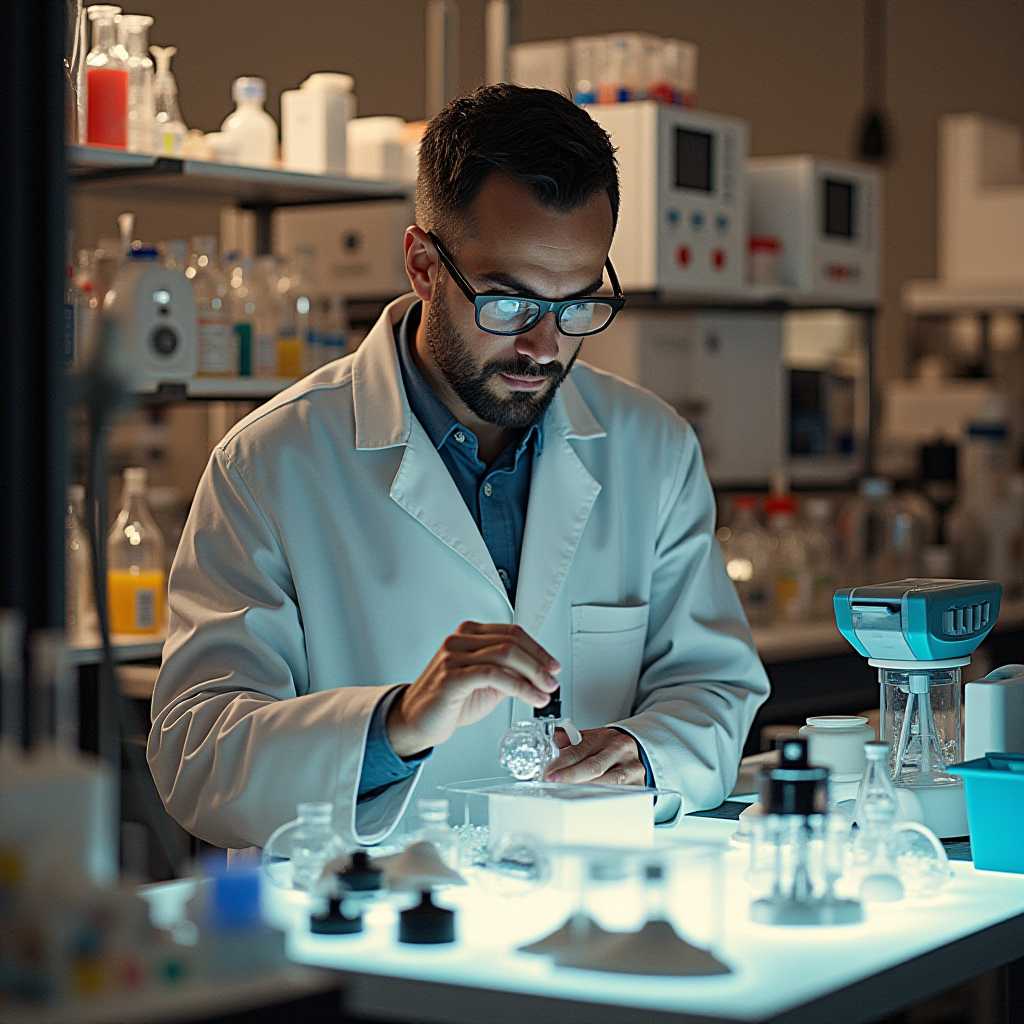} \\
        \multicolumn{9}{c}{\vspace{2pt}\small ``A scientist in a modern laboratory
'' \vspace{8pt}} \\

        \vertlabel{Flux} & \includegraphics[width=\imgwidth]{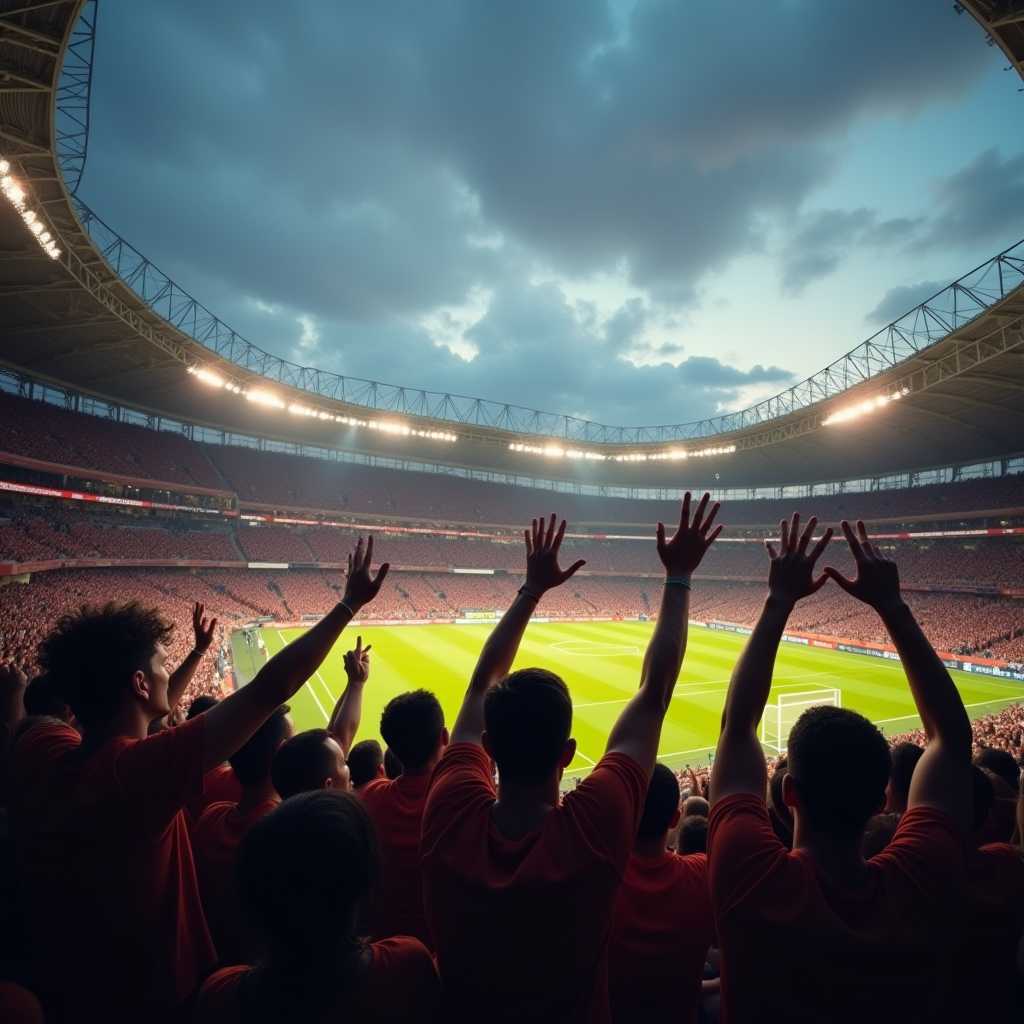} & \includegraphics[width=\imgwidth]{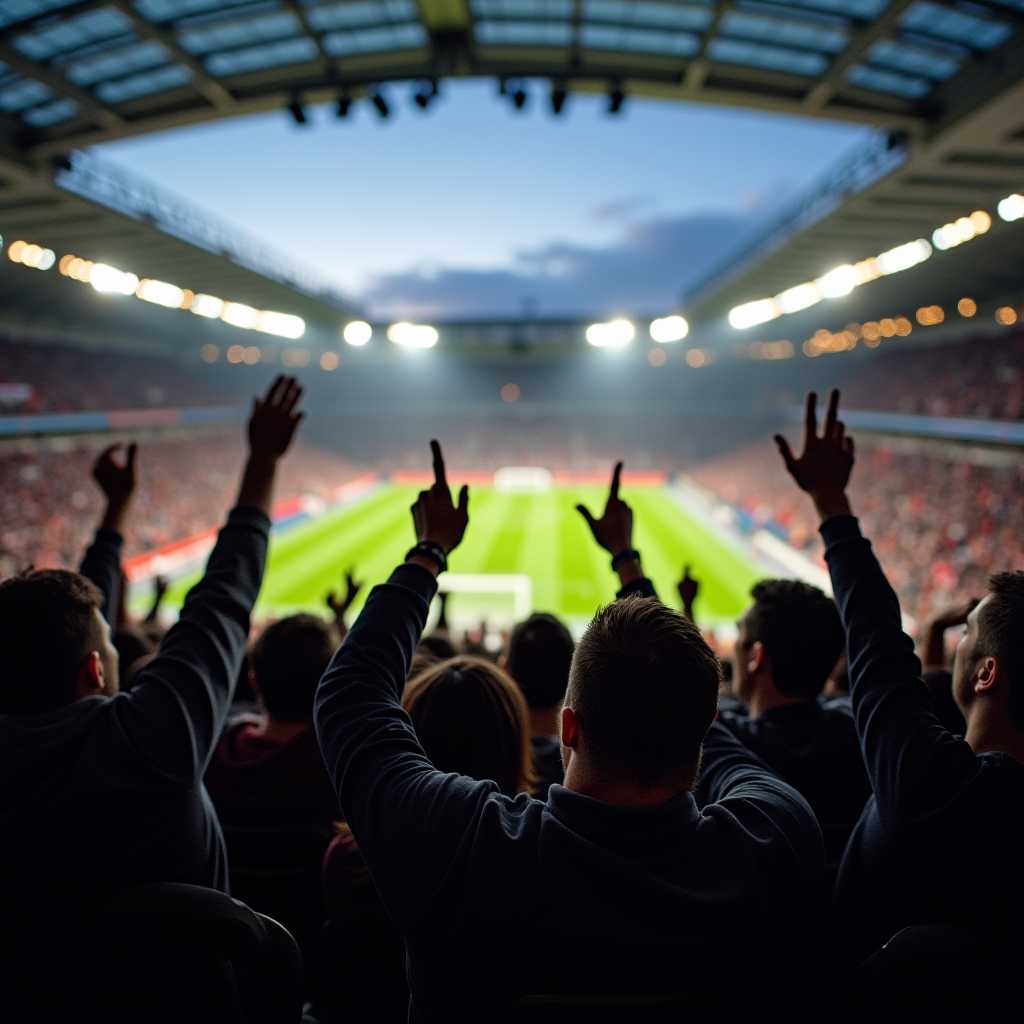} & \includegraphics[width=\imgwidth]{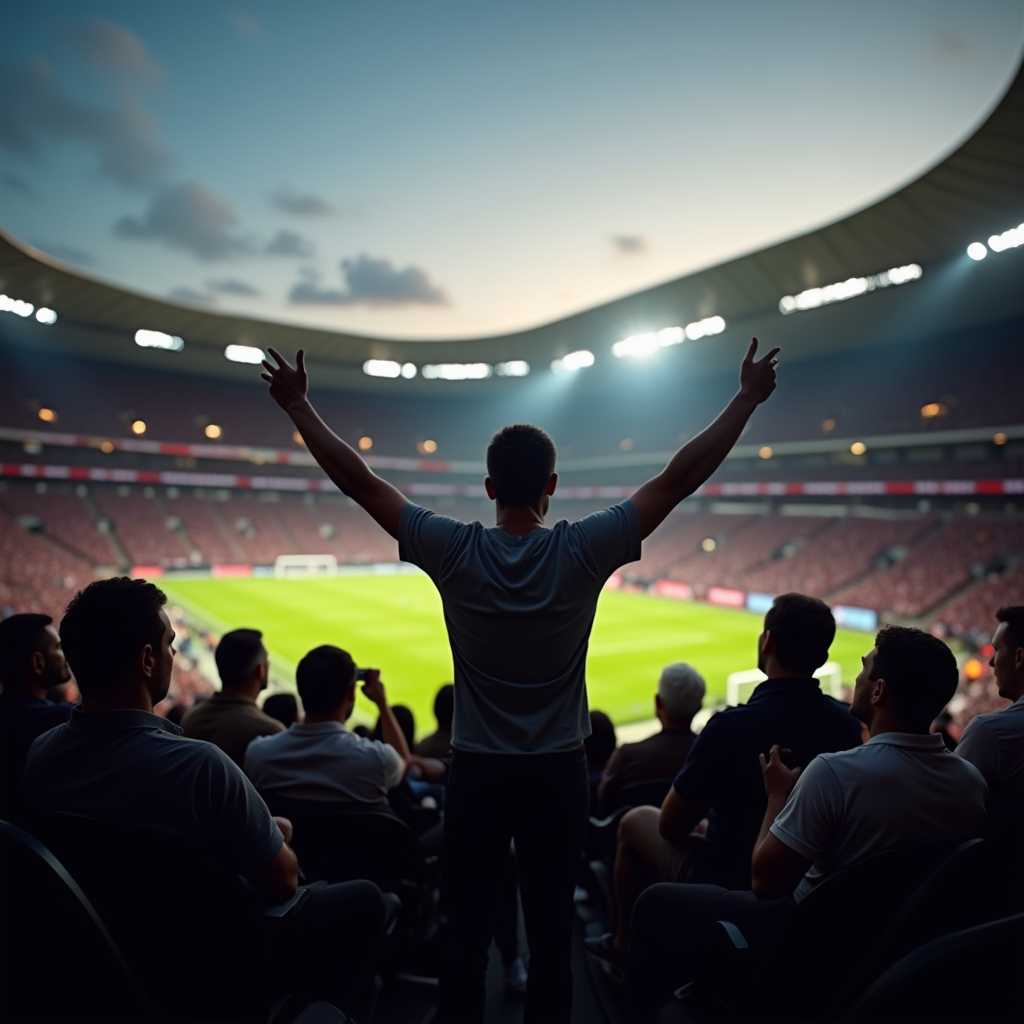} & \includegraphics[width=\imgwidth]{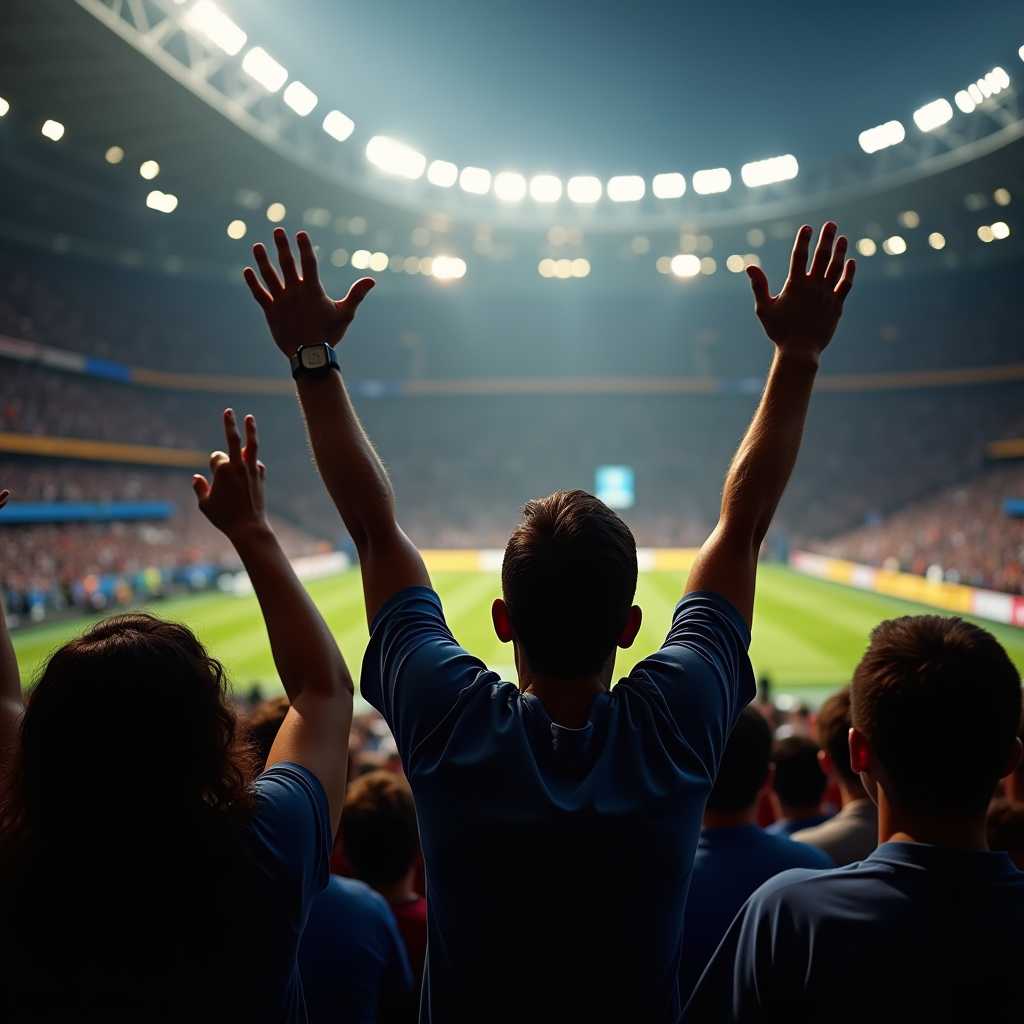} & \includegraphics[width=\imgwidth]{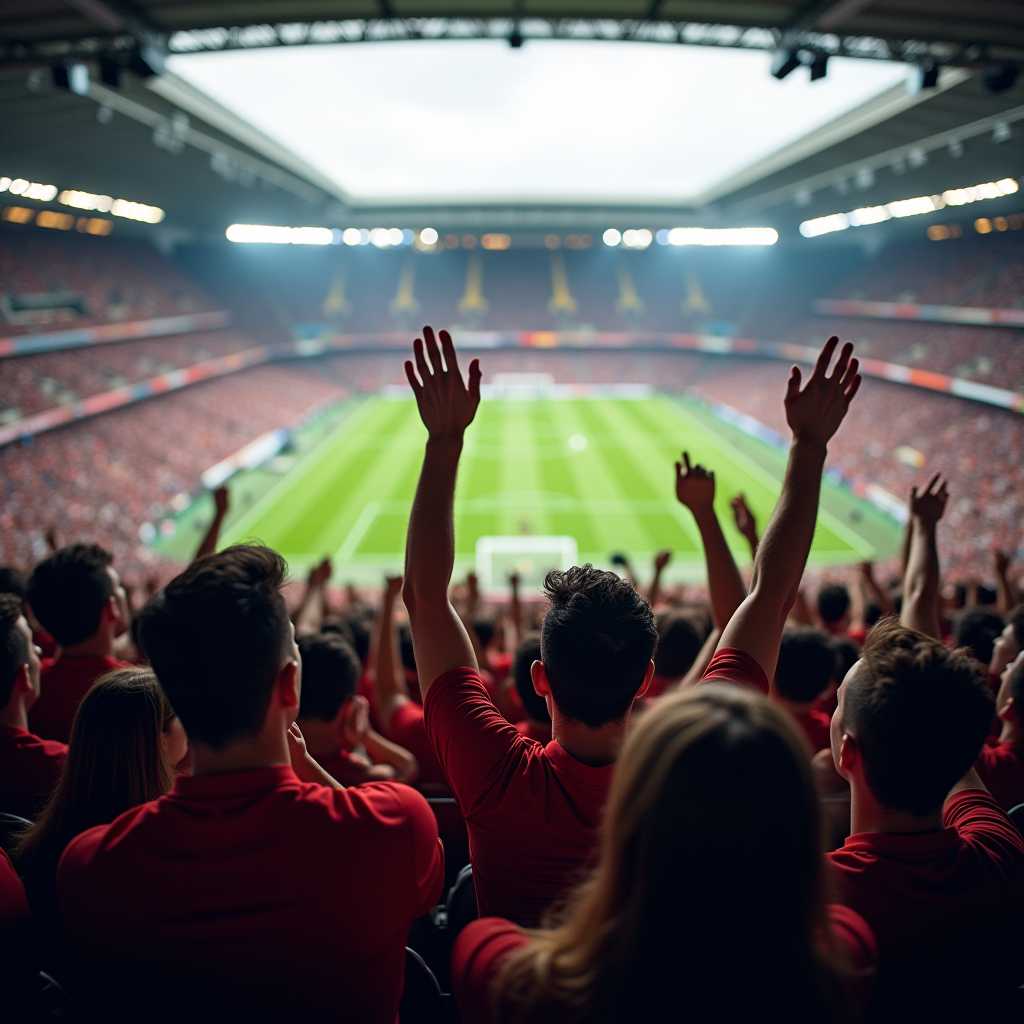} & \includegraphics[width=\imgwidth]{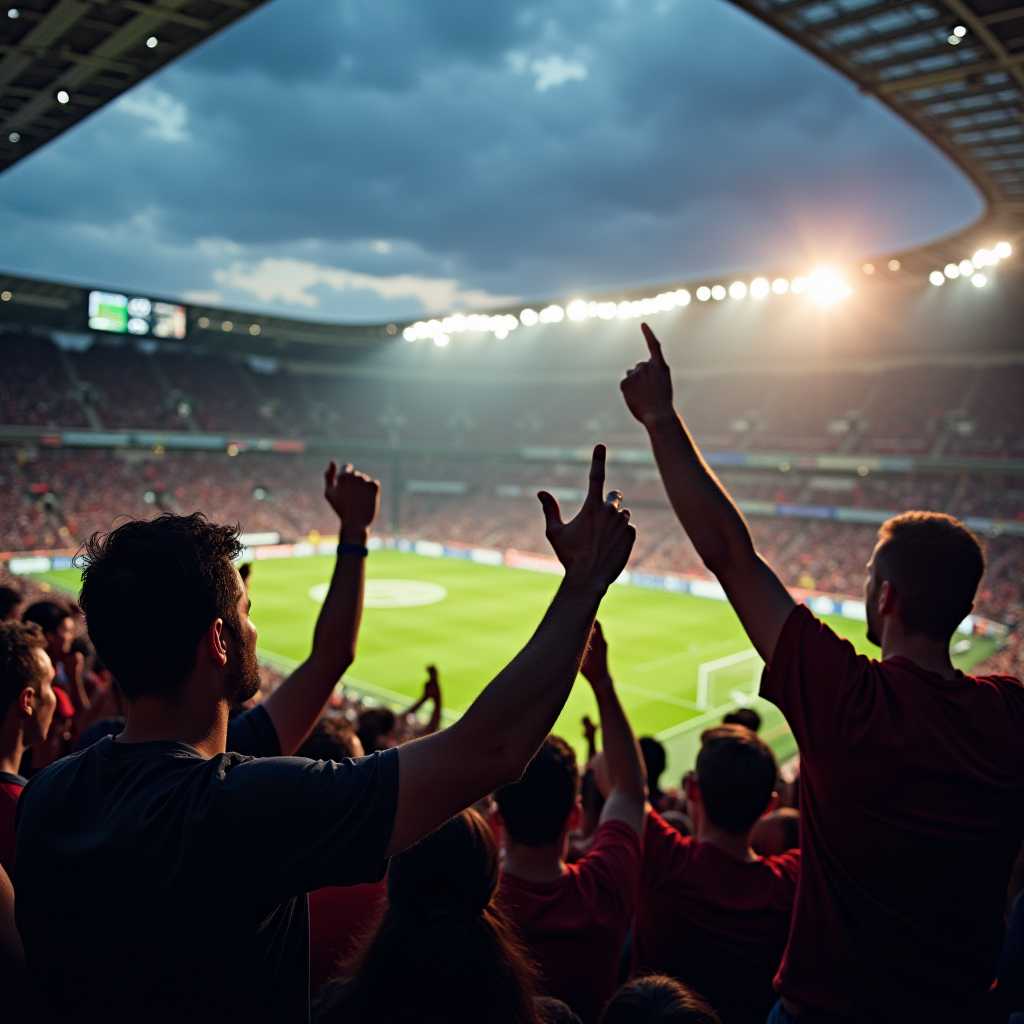} & \includegraphics[width=\imgwidth]{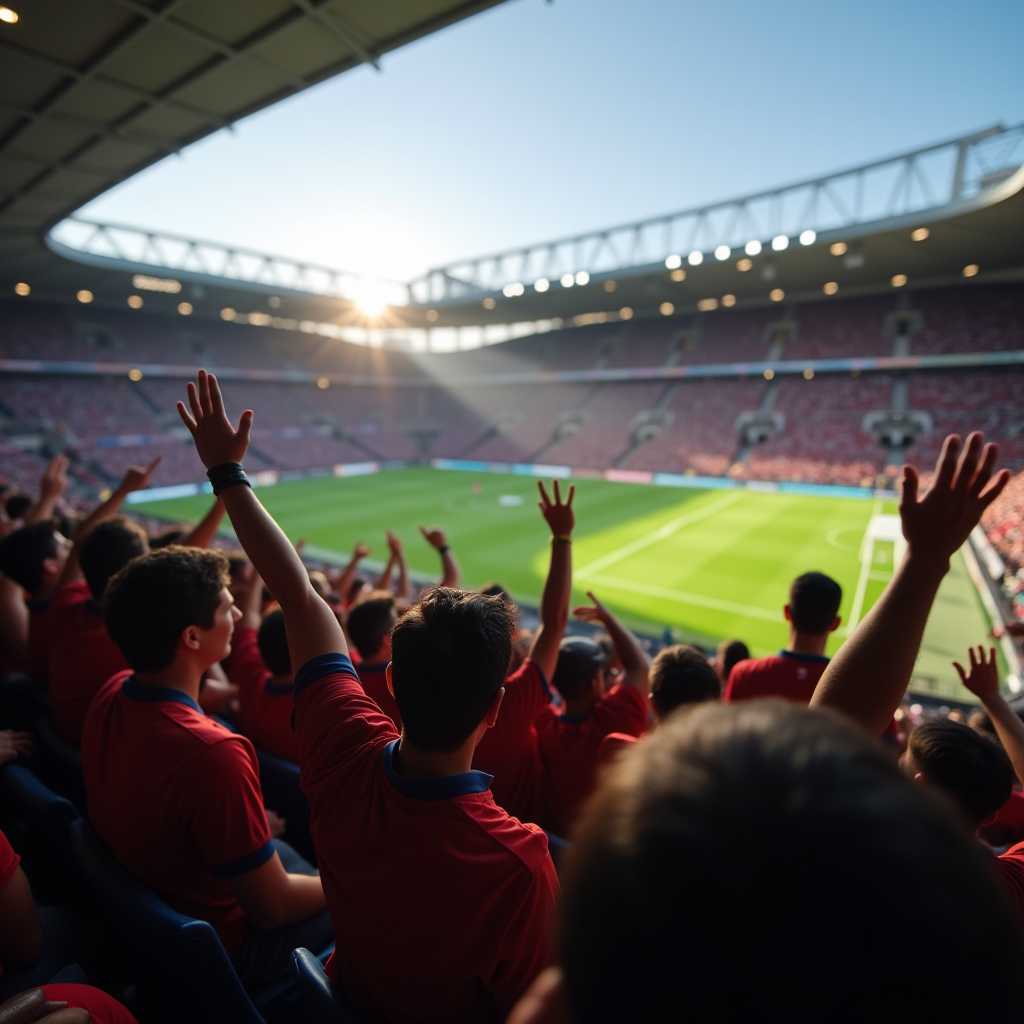} & \includegraphics[width=\imgwidth]{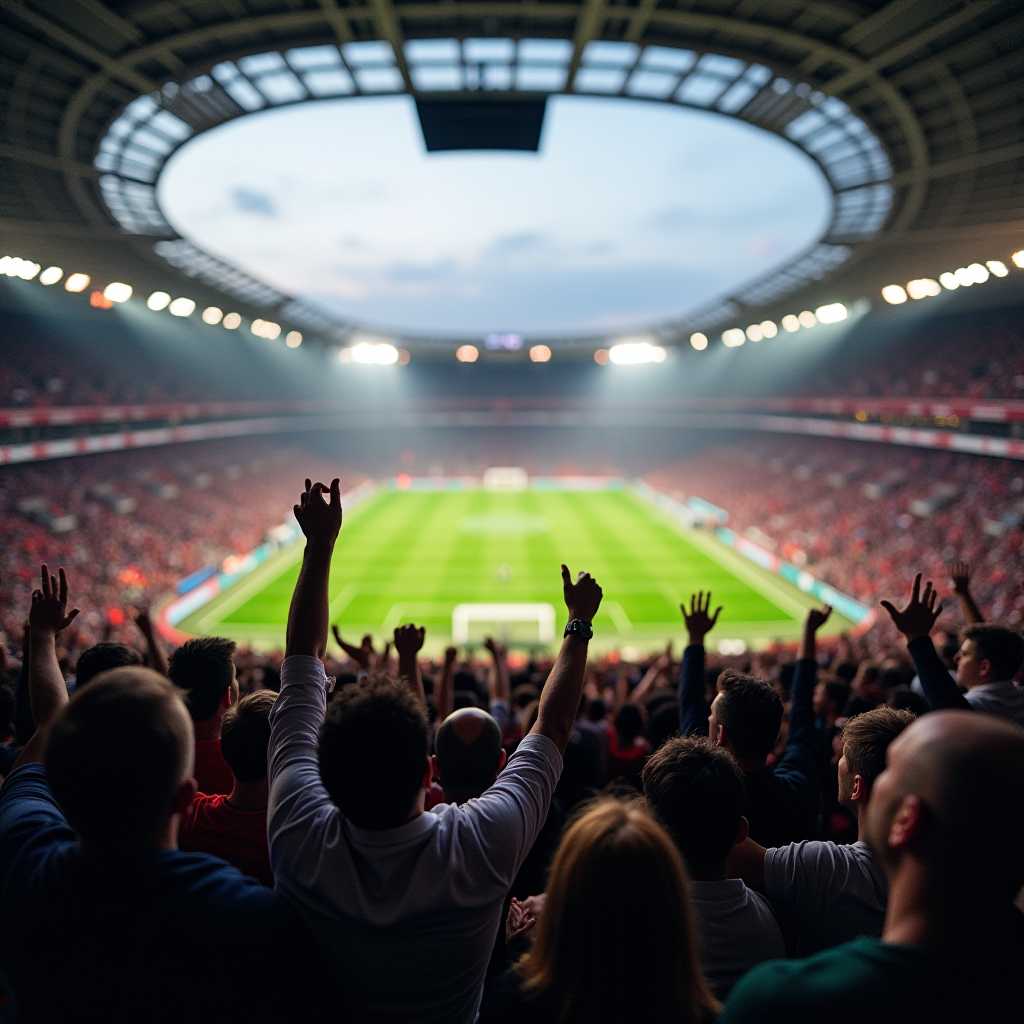} \\[-1pt]
        \vertlabel{Ours} & \includegraphics[width=\imgwidth]{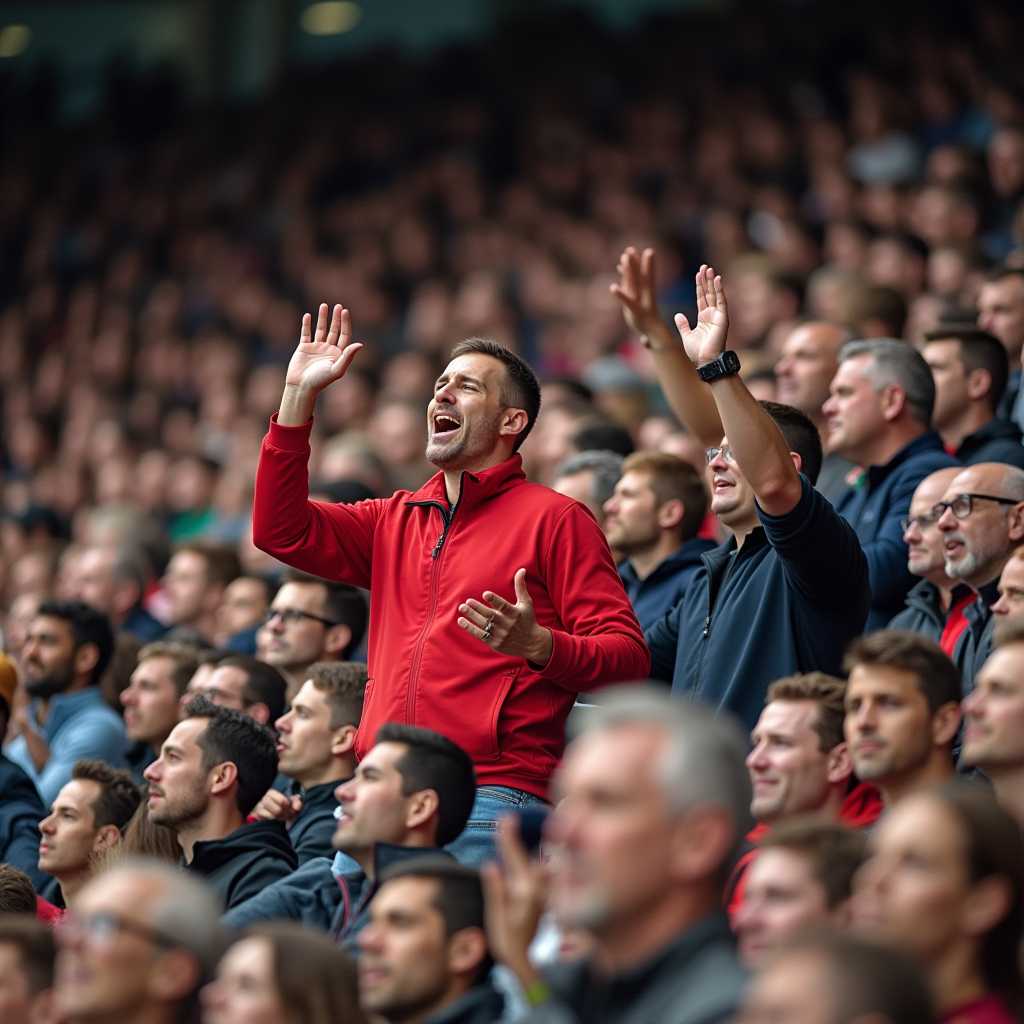} & \includegraphics[width=\imgwidth]{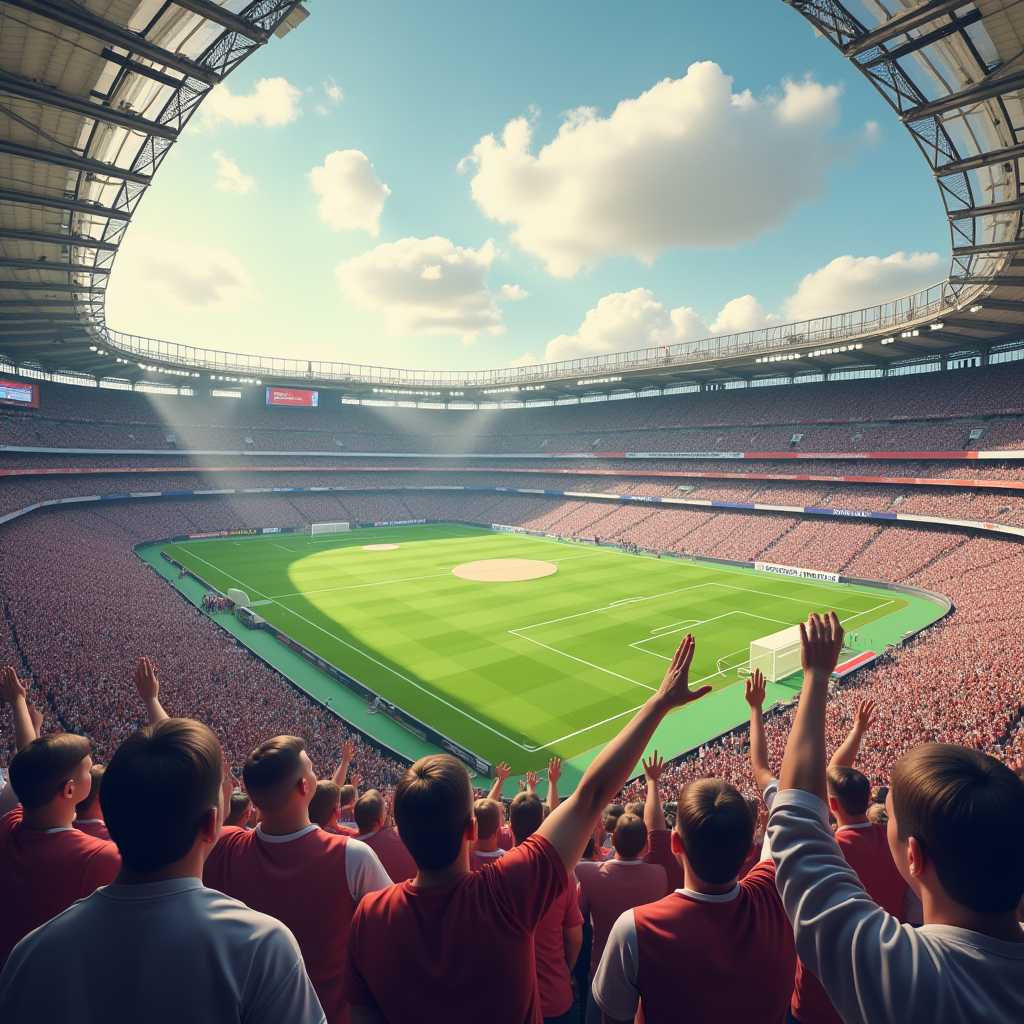} & \includegraphics[width=\imgwidth]{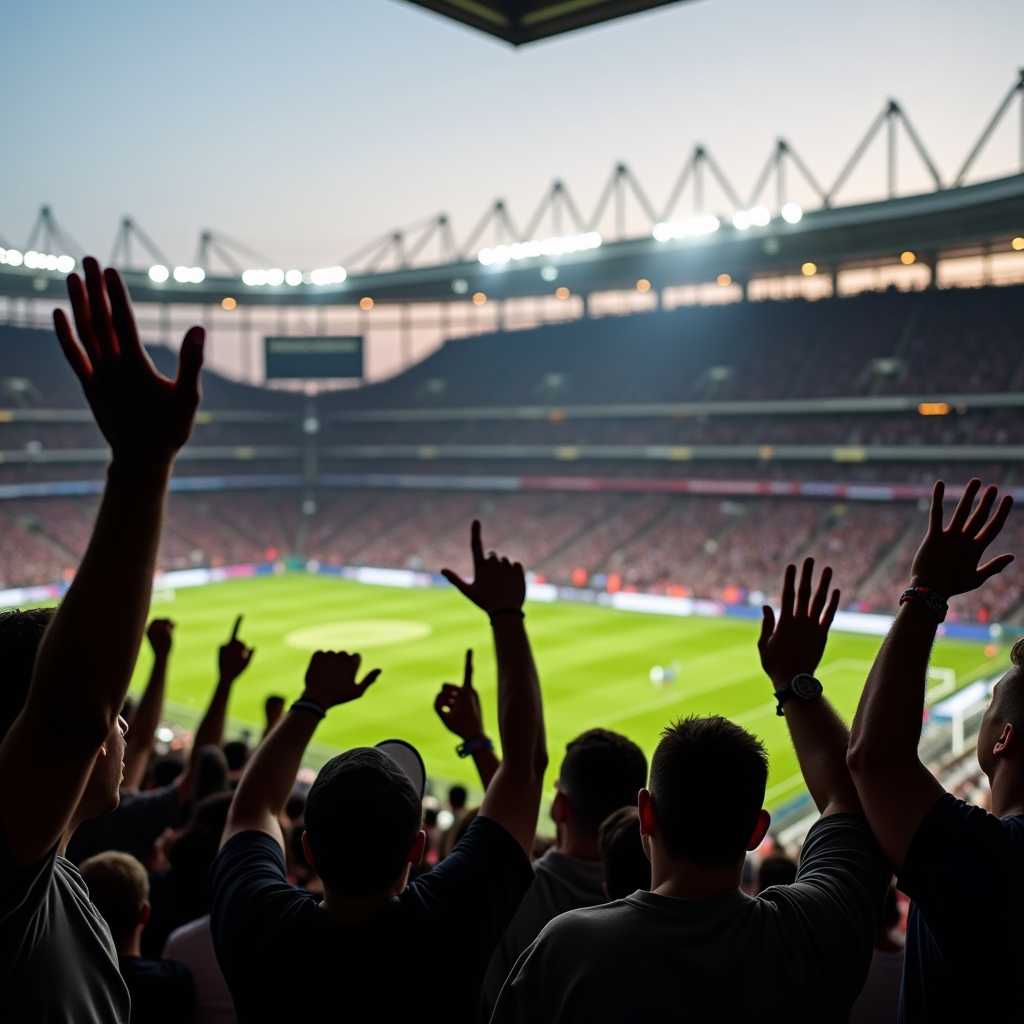} & \includegraphics[width=\imgwidth]{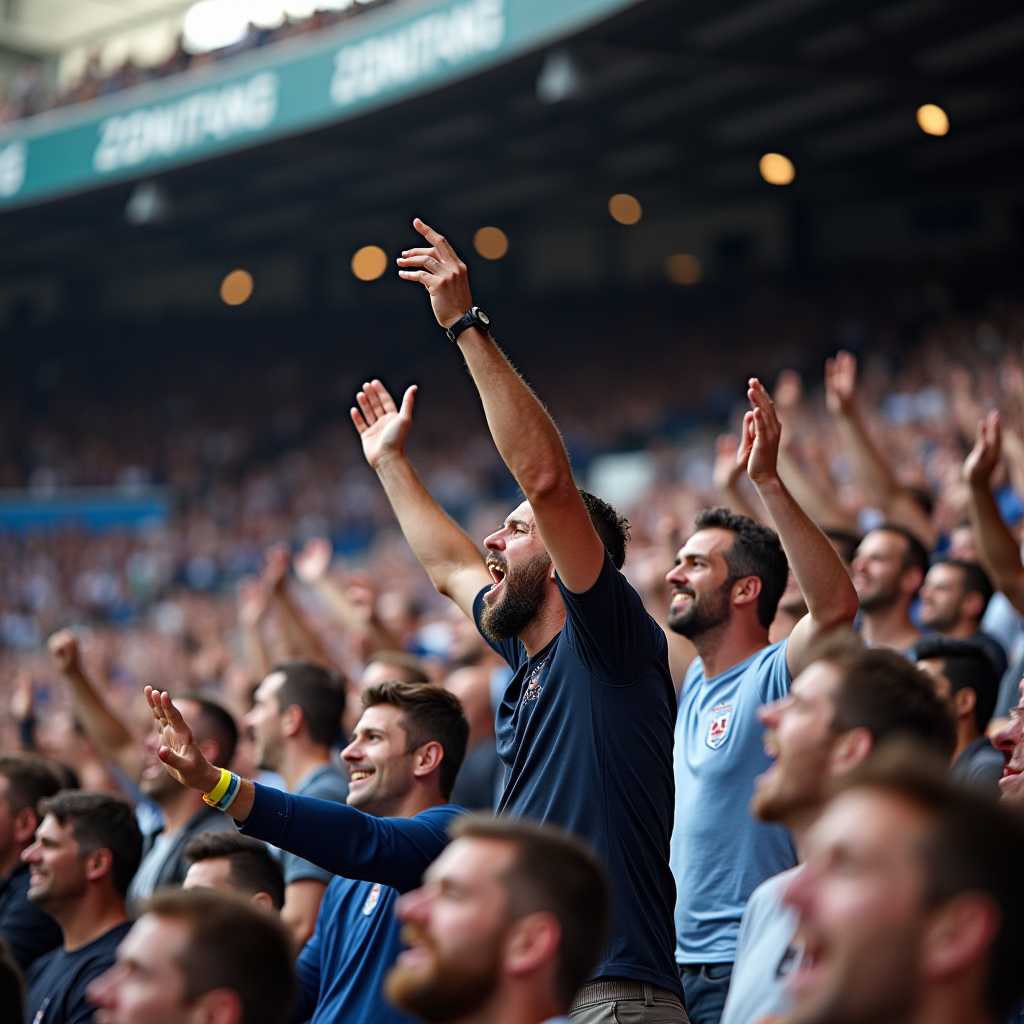} & \includegraphics[width=\imgwidth]{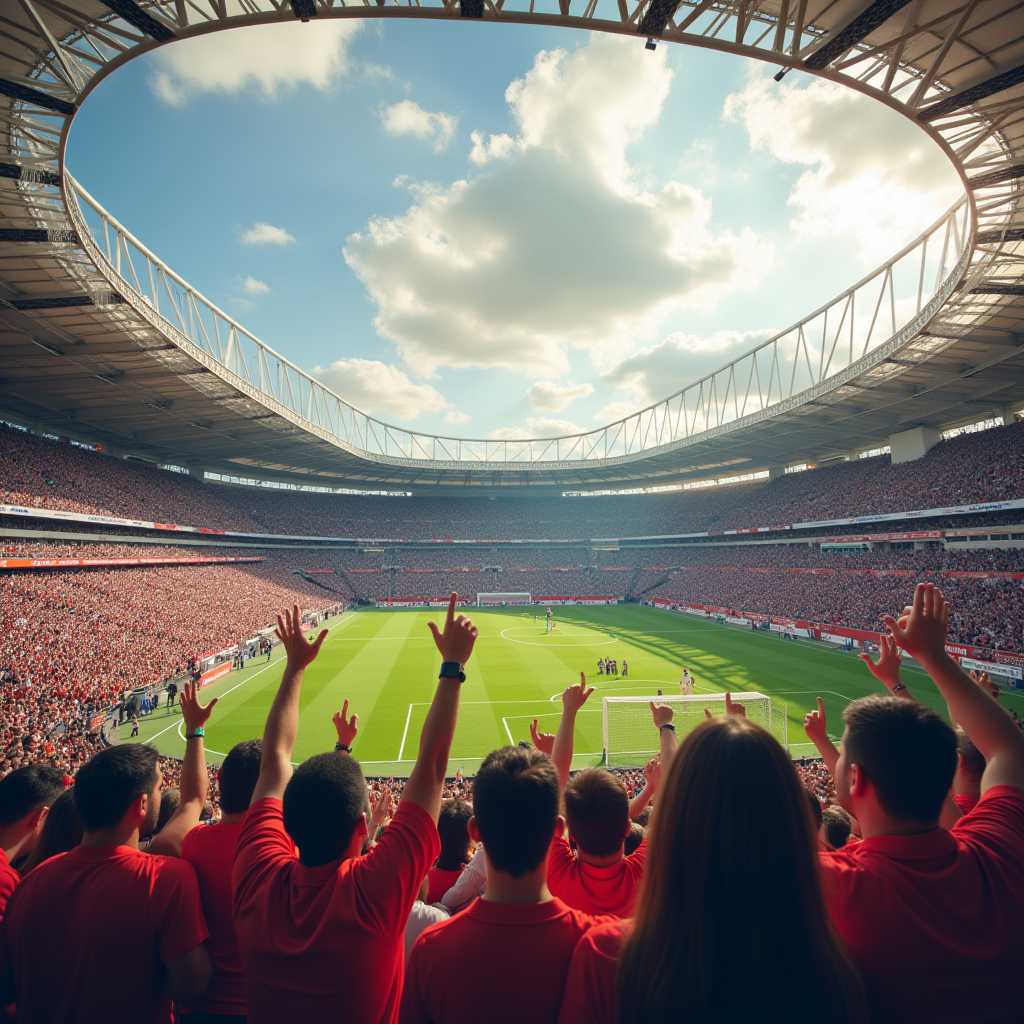} & \includegraphics[width=\imgwidth]{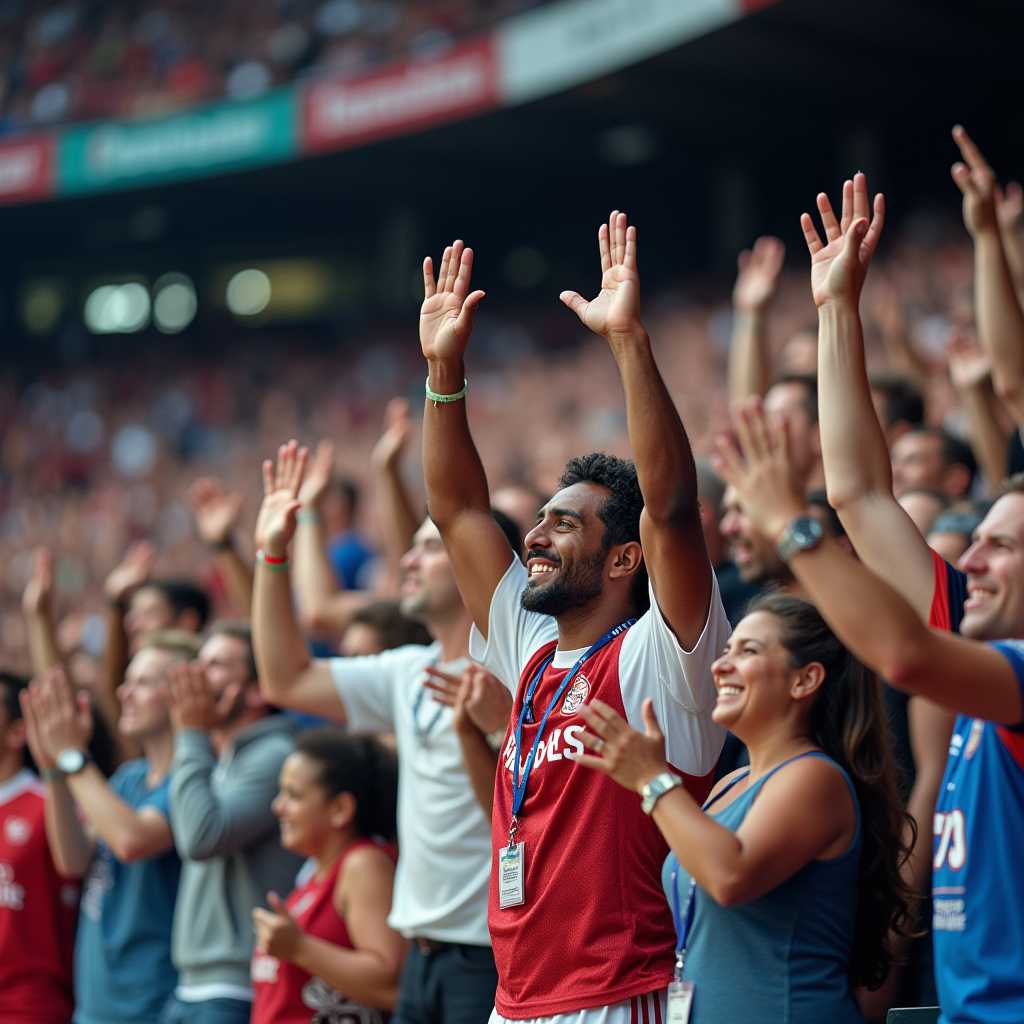} & \includegraphics[width=\imgwidth]{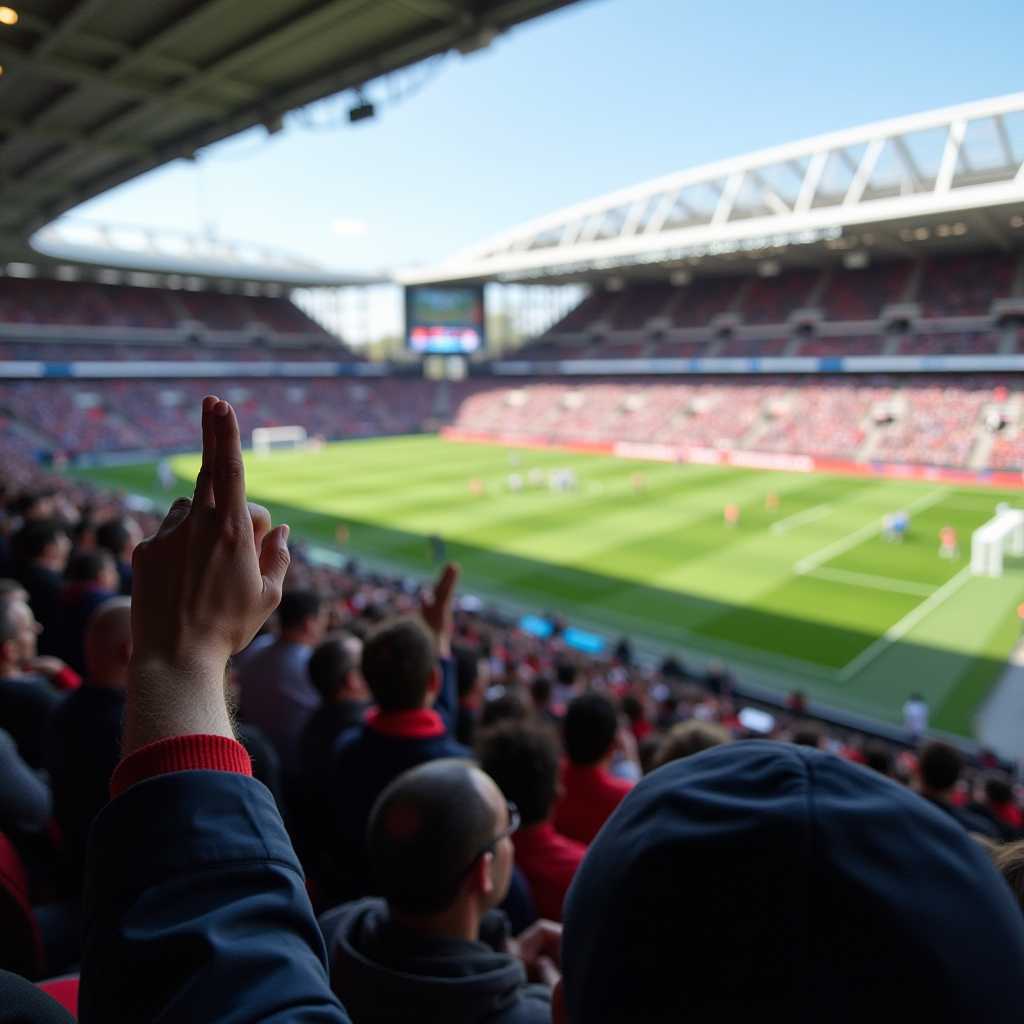} & \includegraphics[width=\imgwidth]{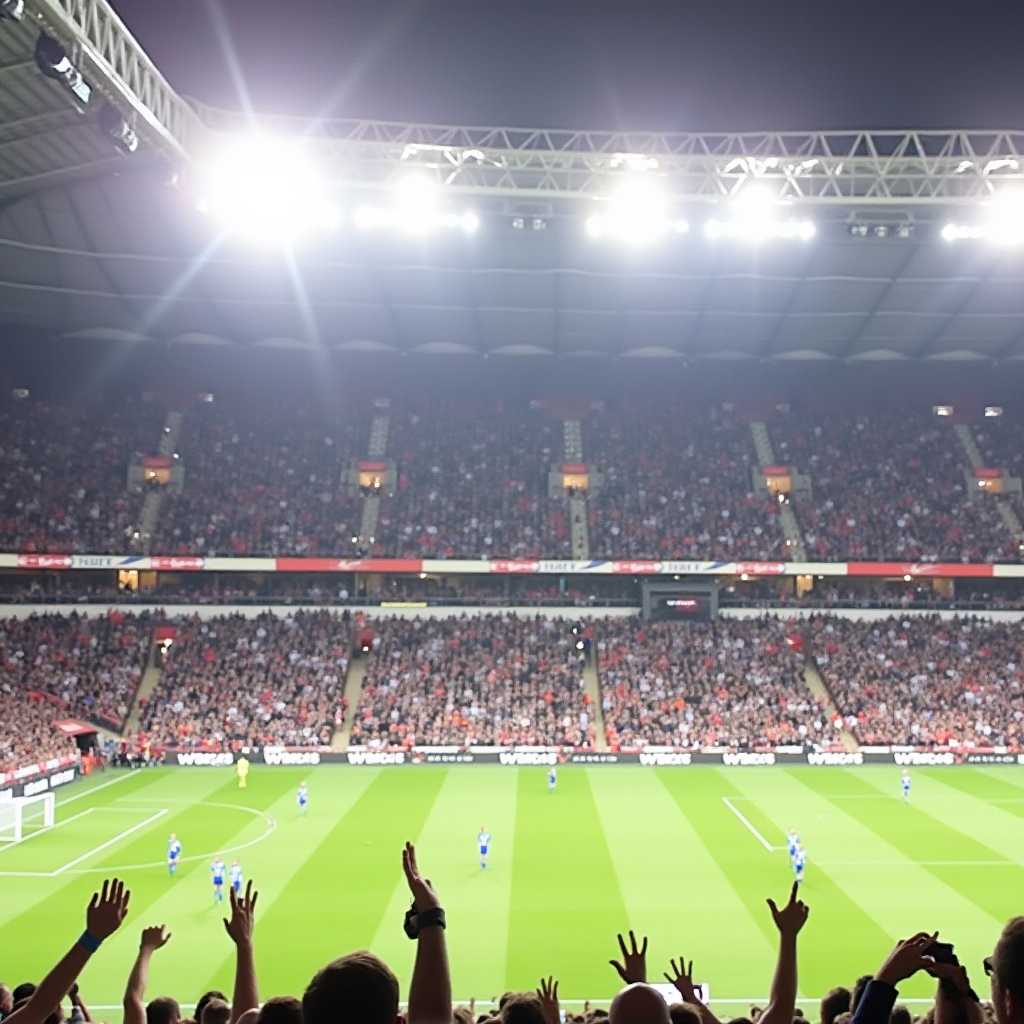} \\
        \multicolumn{9}{c}{\vspace{2pt}\small ``A crowd cheering at a sports stadium'' \vspace{8pt}} \\

    \end{tabular}
    \caption{\textbf{Qualitative results.} For each prompt, we compare the base model results (top) to our results (bottom). All batches were generated using the same random seed initialization. Additional results are provided in
    Appendix~\ref{sec:additional_qualitative_results}.}
    \label{fig:eights}
\end{figure*}

\clearpage

\begin{appendices}
\section*{\LARGE Appendix}
\global\csname @topnum\endcsname 0
\section{Implementation Details}
\label{sec:implementation_details}

All experiments were conducted on an NVIDIA A100 GPU. Quantitative metrics and runtime evaluations were performed by generating groups of 4 images.
Diversity metrics were calculated within each 4-image group and subsequently averaged across all groups.

The number of denoising steps was chosen based on the model architecture: 4 steps for SD3.5-Turbo~\cite{esser2024scaling}, 20 steps for Flux-dev~\cite{labs2025flux}, and 28 steps for SD3.5-Large~\cite{esser2024scaling}. The guidance scale was set to 3.5 for both Flux-dev and SD3.5-Large, and 0.0 for SD3.5-Turbo.

For our proposed method, we employed $M=100$ gradient steps for the Stable Diffusion models and $M=50$ for Flux-dev. For all models, we apply repulsion to the text tokens in the multimodal attention blocks (dual-stream in Flux). For SD3.5-Large, which is not distilled for classifier-free guidance, the repulsion is applied to both the conditional and unconditional branches. For Flux-dev and Flux-Kontext, we additionally apply it to all tokens in the later single-stream blocks, which are specific to these architectures. The repulsion scale $\eta$ was used to balance the trade-off between diversity and fidelity, with the intervention disabled after a fixed number of timesteps, denoted by $\tau$.
The range of $\eta$ was tuned per model: $\eta \in [5 \cdot 10^2, 5\cdot10^6]$ with $\tau=4$ for SD3.5-Large; $\eta \in [1\cdot10^5, 1 \cdot 10^7]$ with $\tau=1$ for SD3.5-Turbo; and $\eta \in [5 \cdot 10^6, 8\cdot10^{9}]$ with $\tau=1$ for Flux-dev.
For simplicity, $\eta$ remained constant throughout the intervention window.

We utilized official implementations for all baseline methods, where available. For baselines without compatible official implementations, we re-implemented them and tuned their hyperparameters to ensure competitive diversity levels. In addition to the shared guidance and step configurations, the following hyperparameters were used for the baselines:
\begin{itemize}
    \item \textbf{PG~\cite{corso2023particle}:} Repulsion scales were varied between 10 and 100.
    \item \textbf{CADS~\cite{sadat2023cads}:} Scales were varied between 0.1 and 0.7, with $\tau_1=0.3, \tau_2=0.8$, and $\psi=1$.
    \item \textbf{SPARKE~\cite{jalali2025sparke}:} Scales were selected between 0.02 and 0.14, depending on the model.
    \item \textbf{SGI~\cite{parmar2025scaling}:} Evaluated with initial candidate groups of $N \in \{8, 16, 32, 64\}$, utilizing default hyperparameters from the official implementation. All qualitative comparisons and the user study results reported here were conducted with $N=64$.
\end{itemize}

\begin{figure}[ t ]
    \centering
    \scriptsize
    \includegraphics[width=\linewidth]{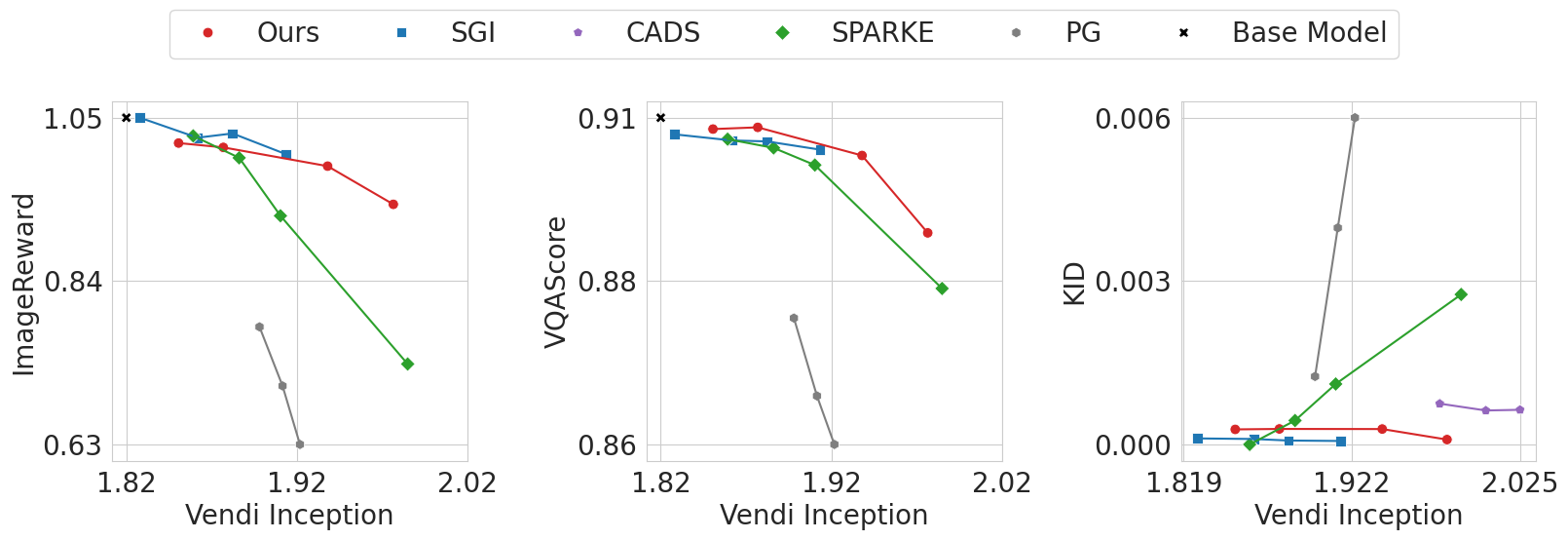}
    \caption{\textbf{Quantitative evaluation on SD3.5-Large.}} \label{fig:sd_quant}
\end{figure}

\begin{figure}[ t ]
    \centering
    \scriptsize
    \includegraphics[width=\linewidth]{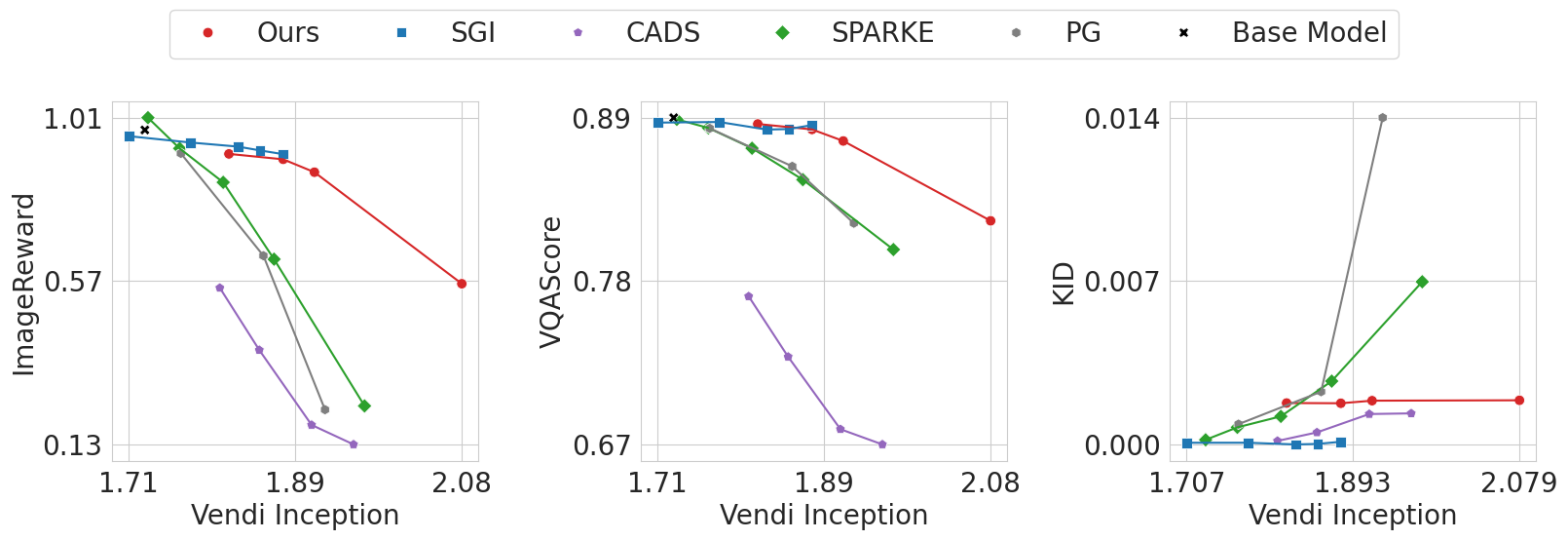}
    \caption{\textbf{Quantitative evaluation on SD3.5-Turbo.}}\label{fig:turbo_quant}
\end{figure}

\section{Additional Qualitative Results} 

\label{sec:additional_qualitative_results}

\begin{figure}
    \centering
    \setlength{\tabcolsep}{0.5pt} \renewcommand{\arraystretch}{0.5} \newcommand{\imgwidth}{0.12\textwidth}
    
    \newcommand{\vertlabel}[2]{\raisebox{#2}{\rotatebox{90}{\scriptsize\textbf{#1}}}}
    
    \newcommand{\largelabel}{\vertlabel{SD3.5-Large}{1.5em}}
    
    \newcommand{\ourslabel}{ \vertlabel{Ours}{2.5em}}
    \begin{tabular}{c c c c c}
        \largelabel & \includegraphics[width=\imgwidth]{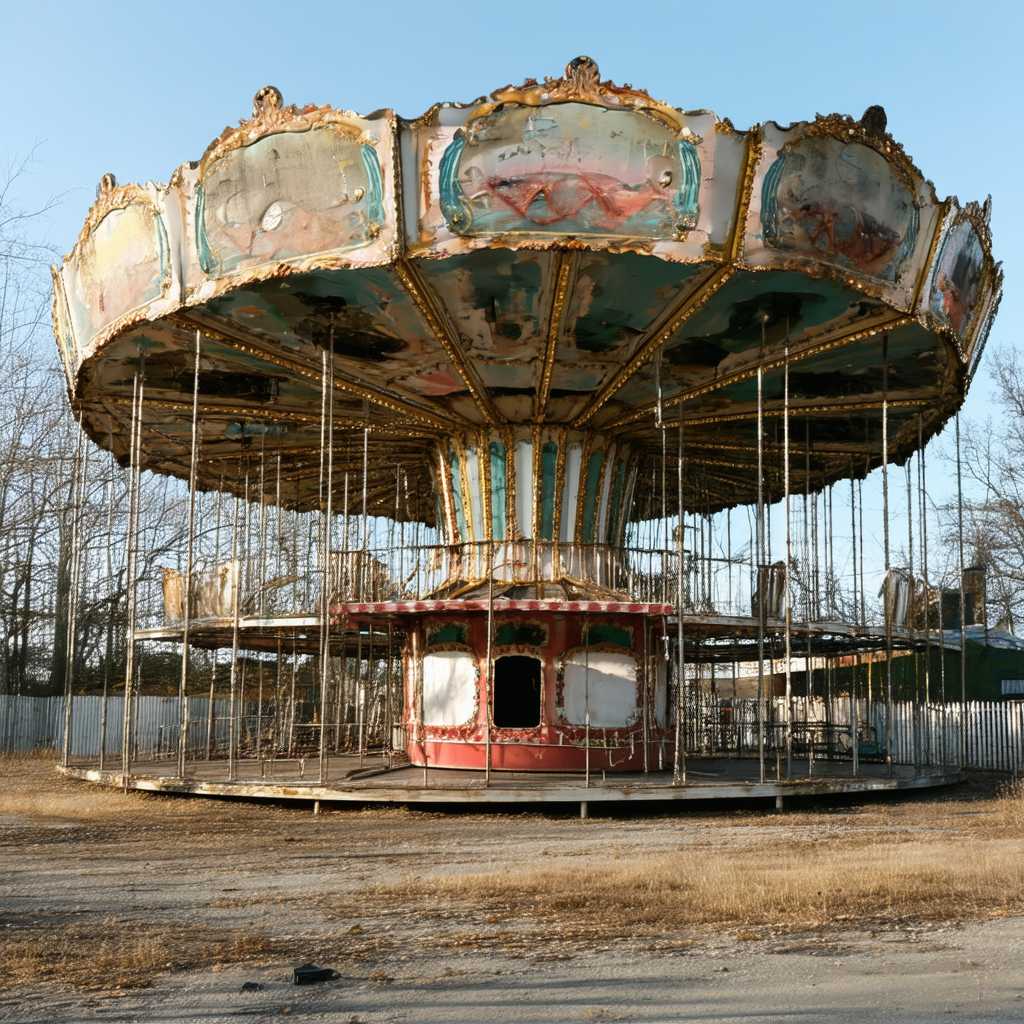} & \includegraphics[width=\imgwidth]{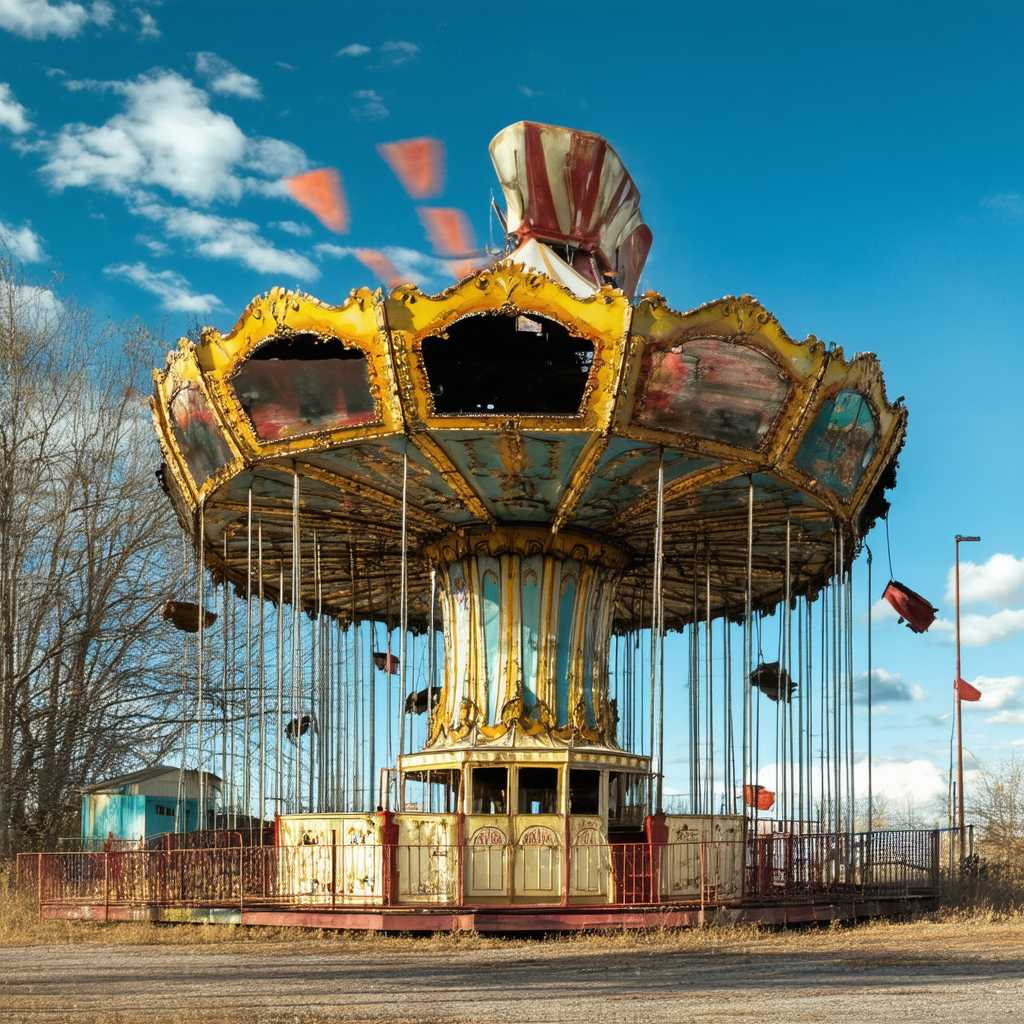} & \includegraphics[width=\imgwidth]{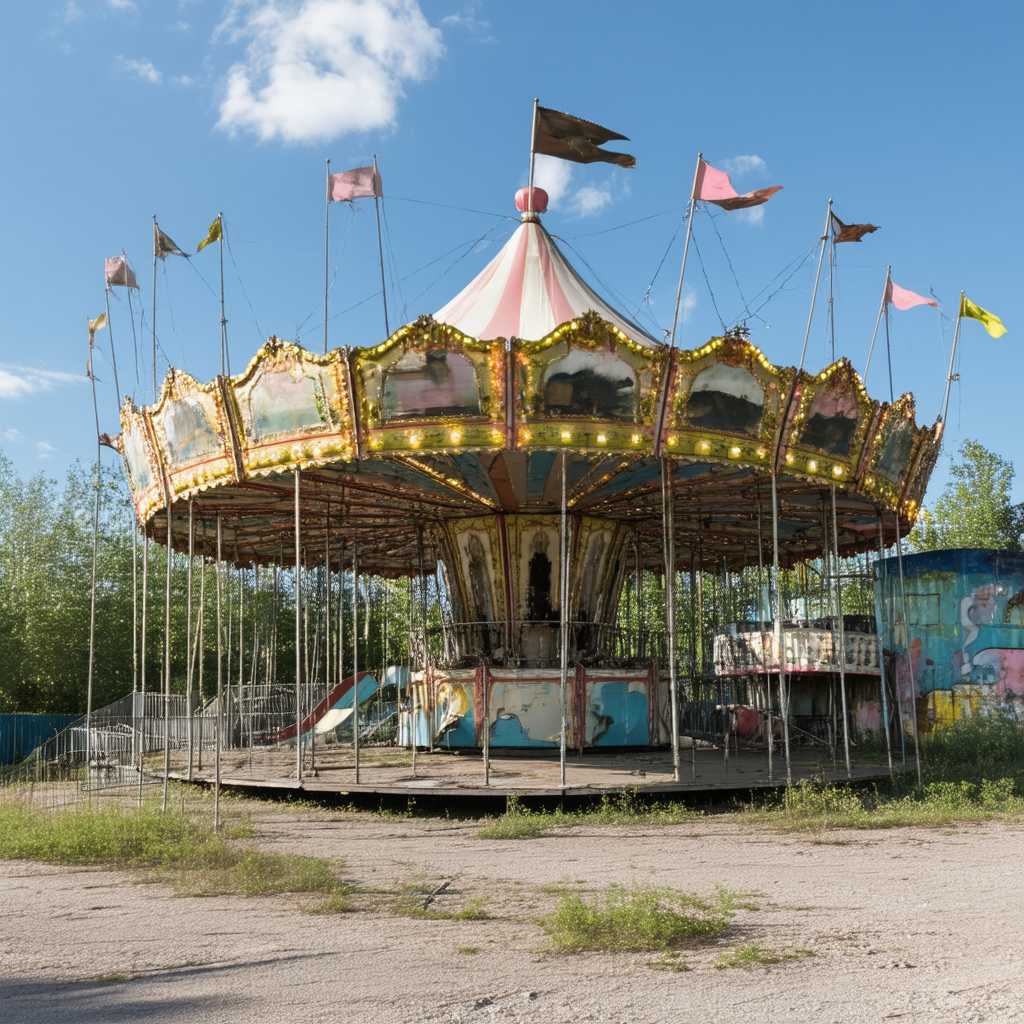} & \includegraphics[width=\imgwidth]{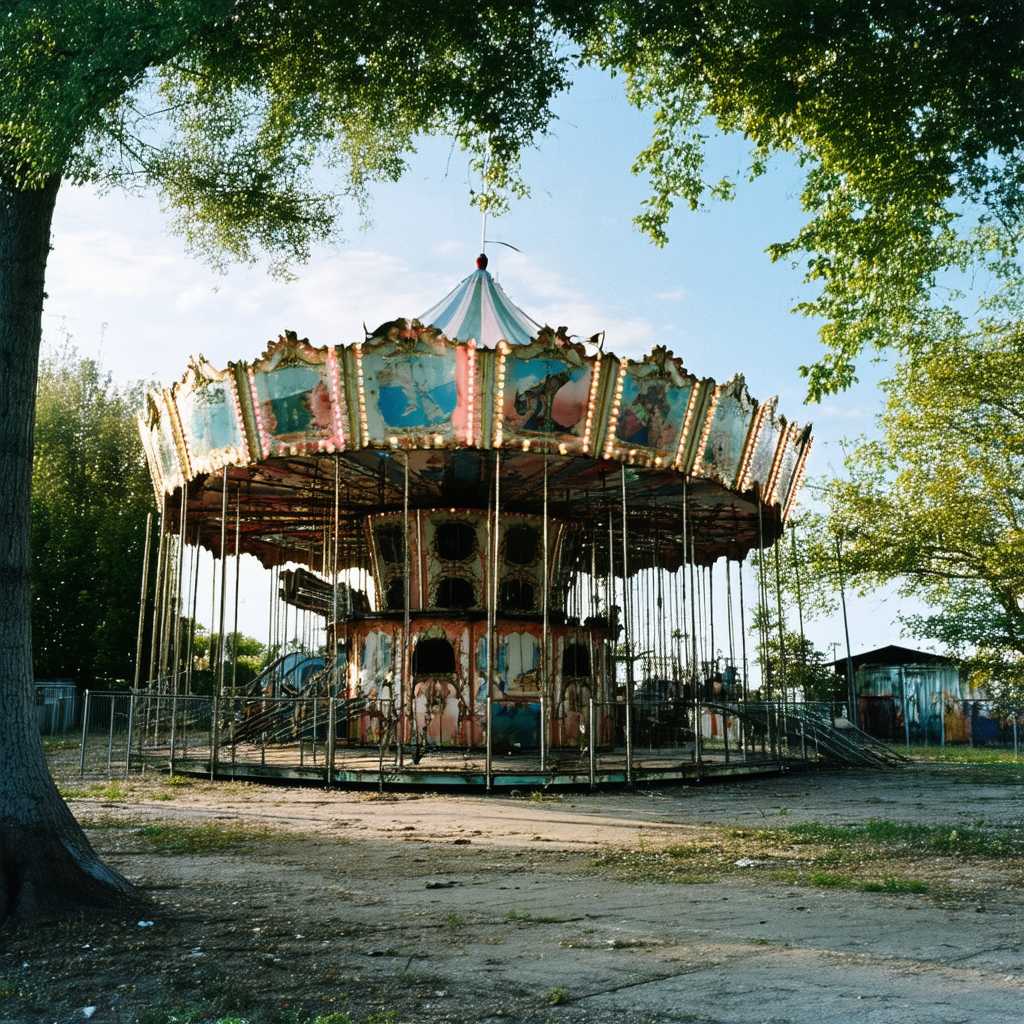} \\[-1pt]
        \ourslabel & \includegraphics[width=\imgwidth]{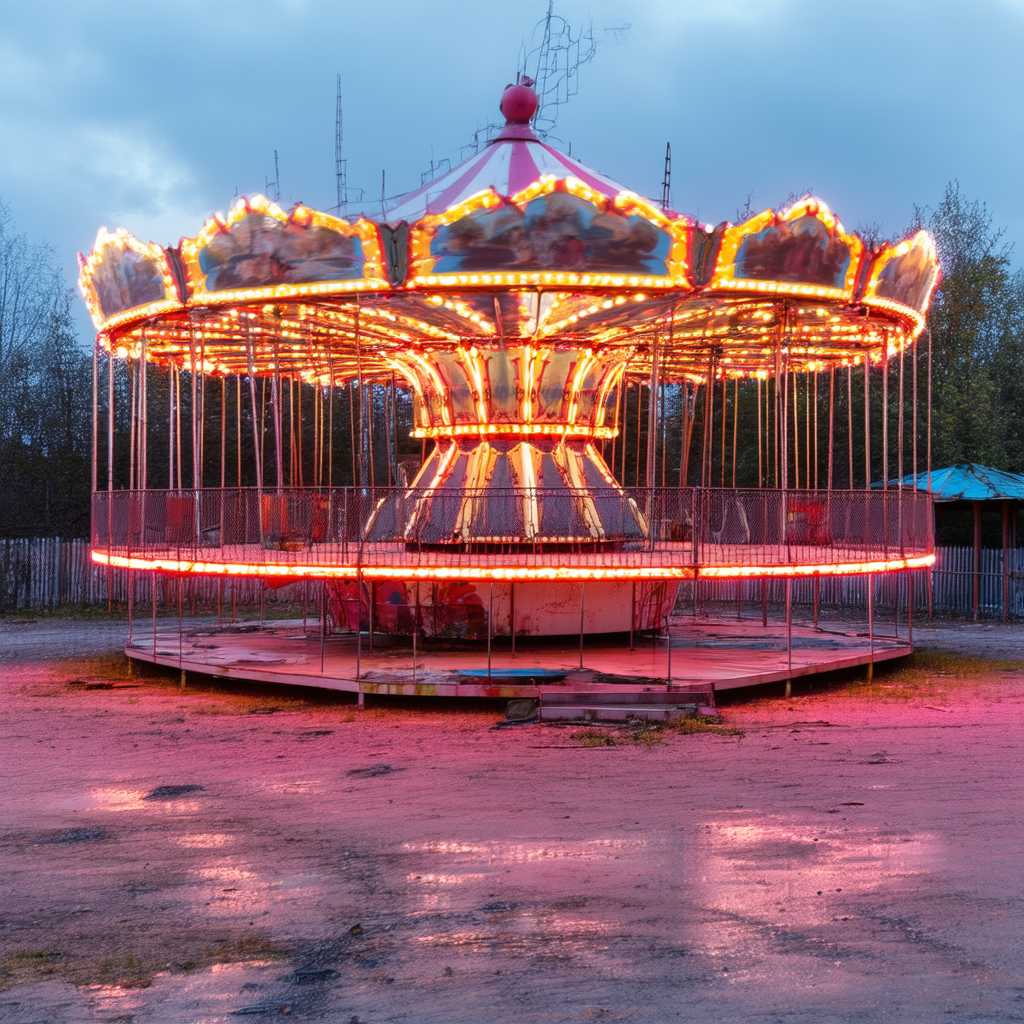} & \includegraphics[width=\imgwidth]{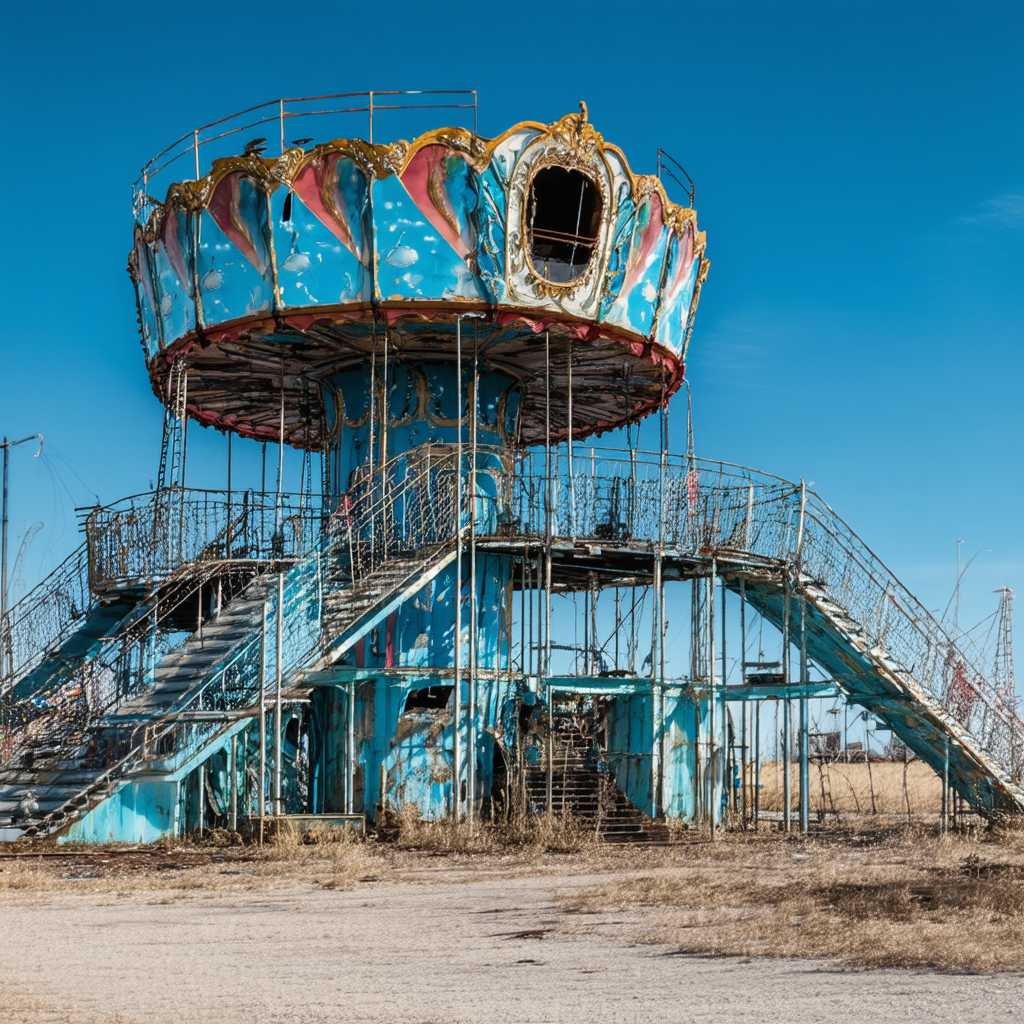} & \includegraphics[width=\imgwidth]{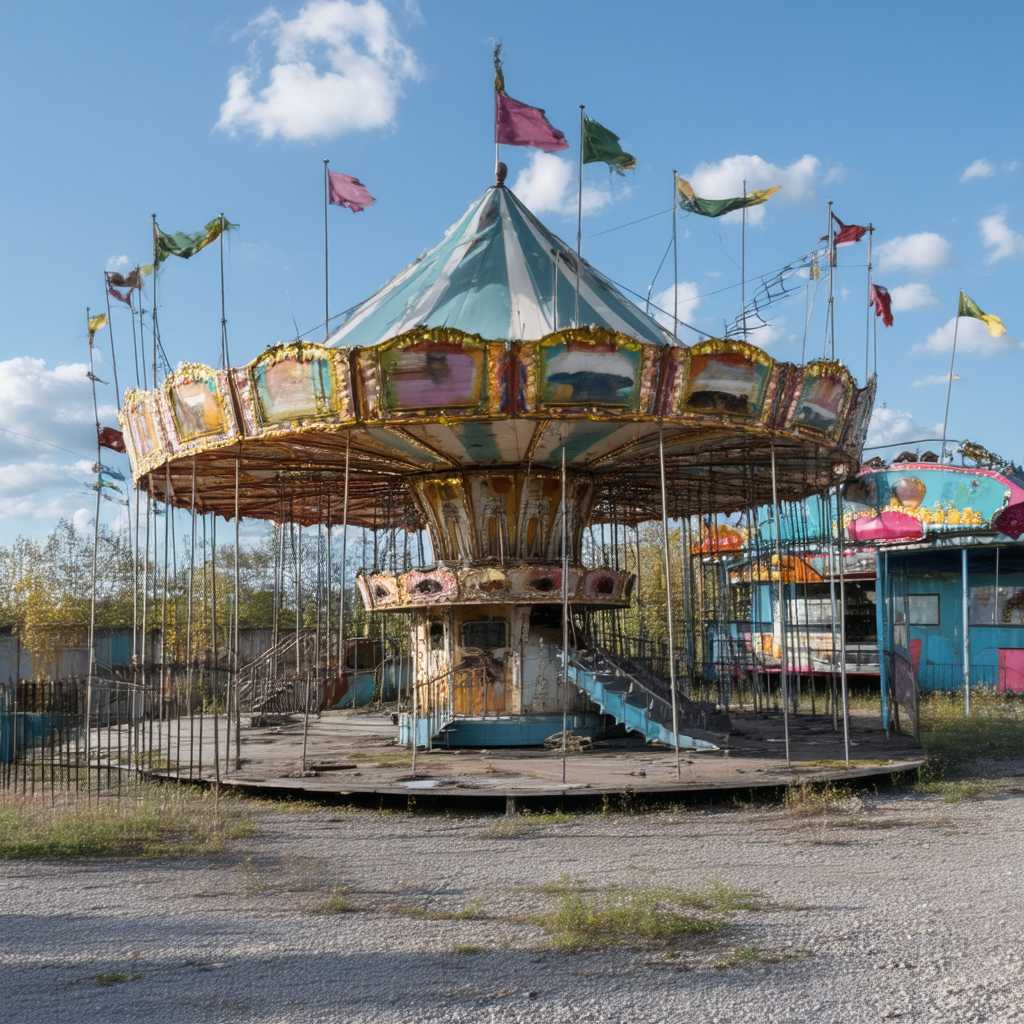} & \includegraphics[width=\imgwidth]{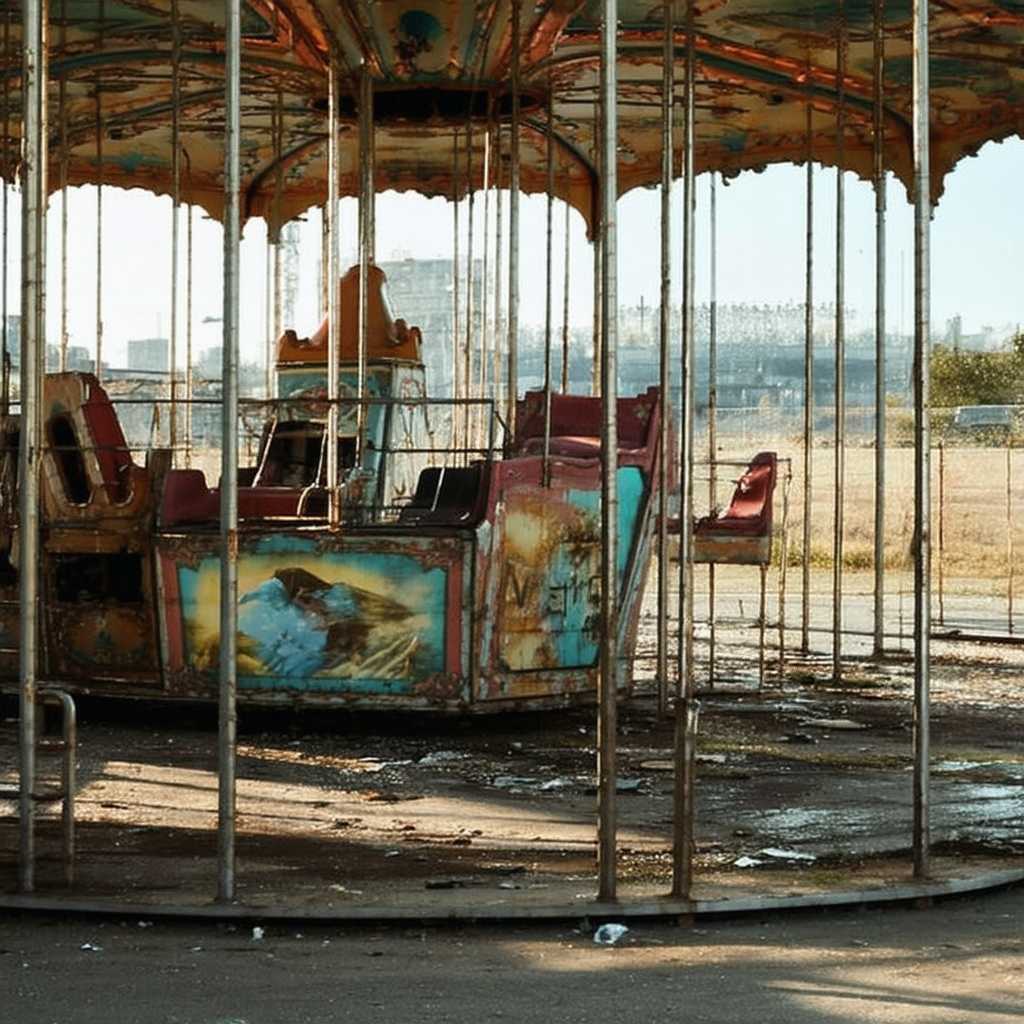} \\
        \multicolumn{5}{c}{\vspace{2pt}\small ``An abandoned carnival'' \vspace{8pt}} \\

        \largelabel & \includegraphics[width=\imgwidth]{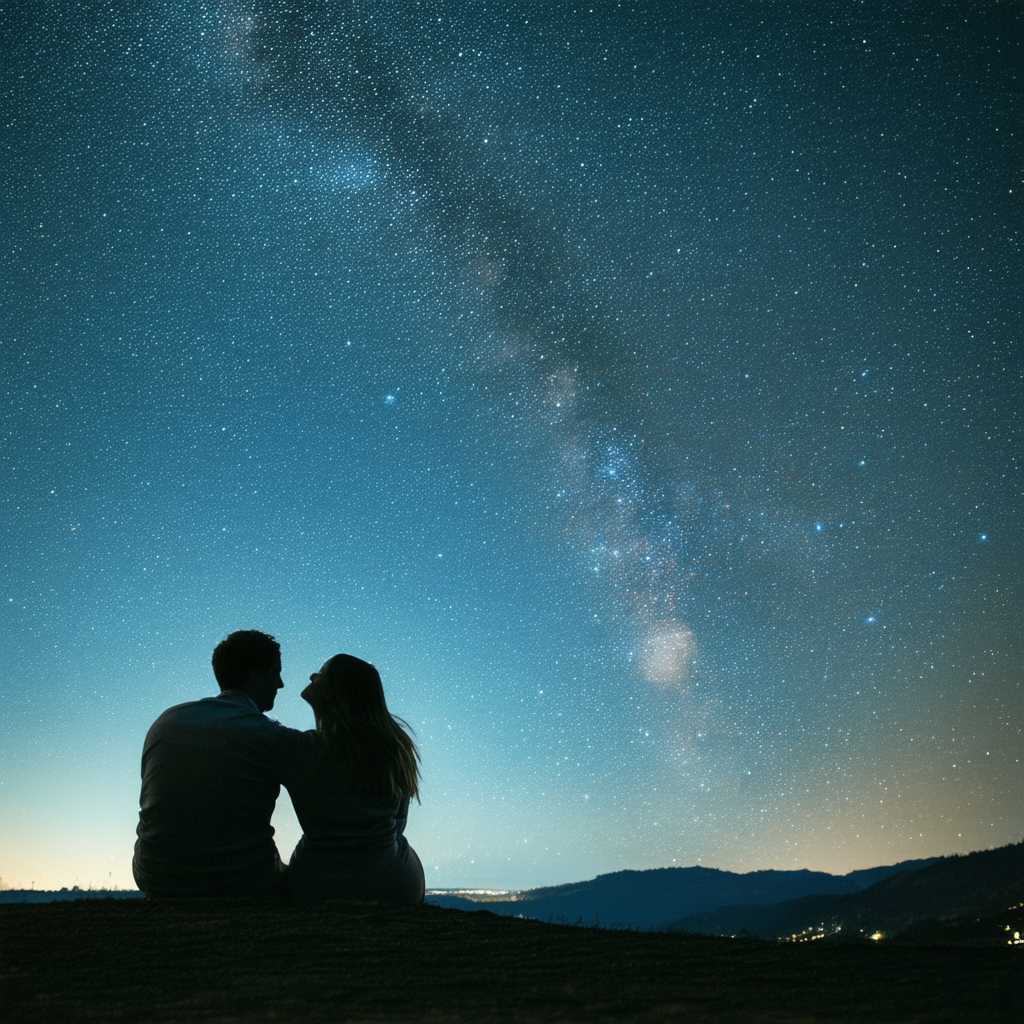} & \includegraphics[width=\imgwidth]{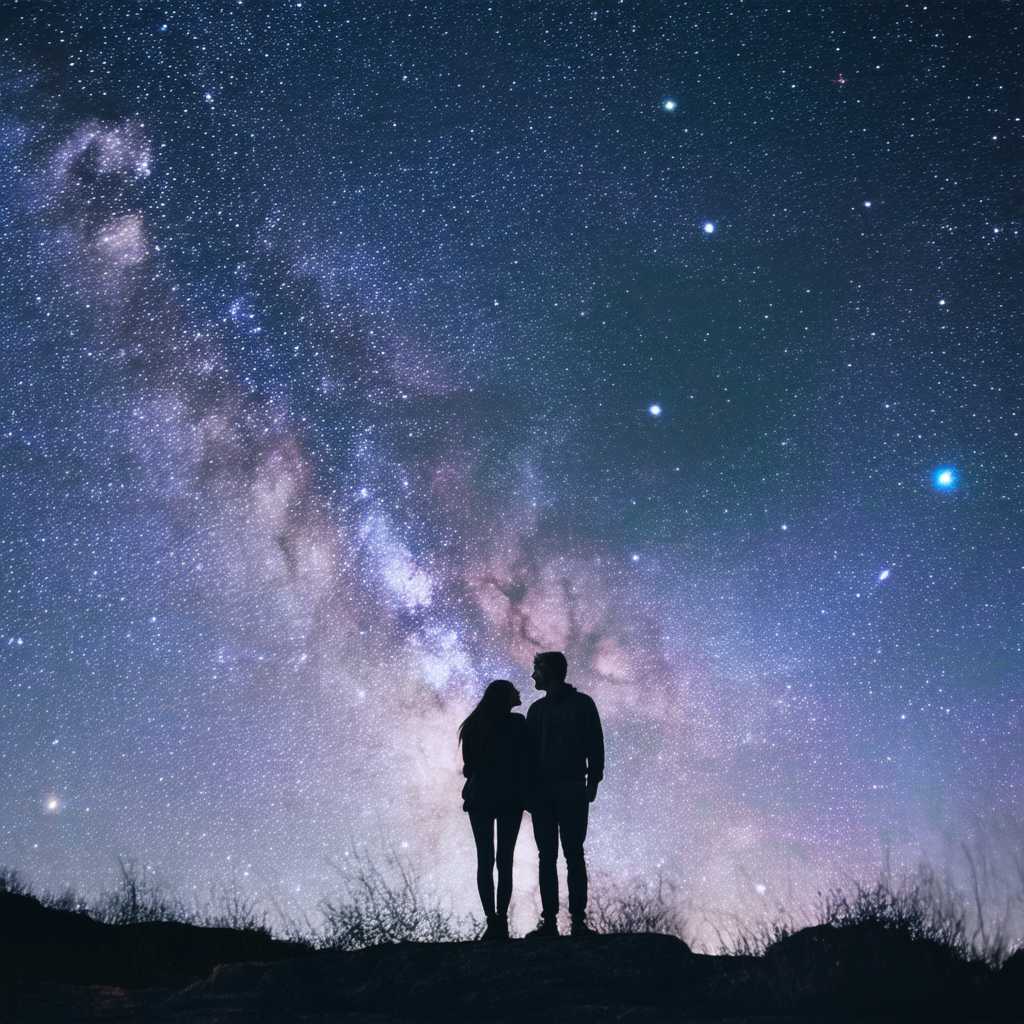} & \includegraphics[width=\imgwidth]{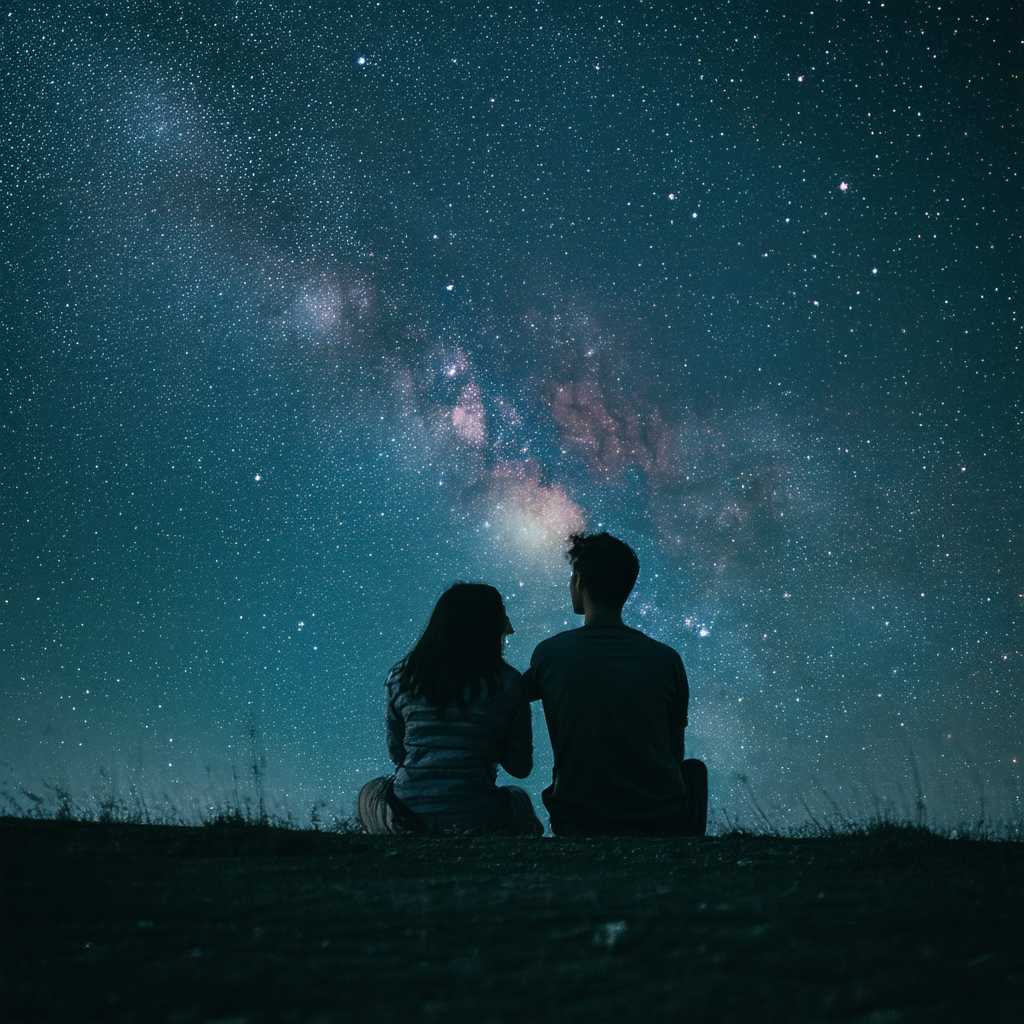} & \includegraphics[width=\imgwidth]{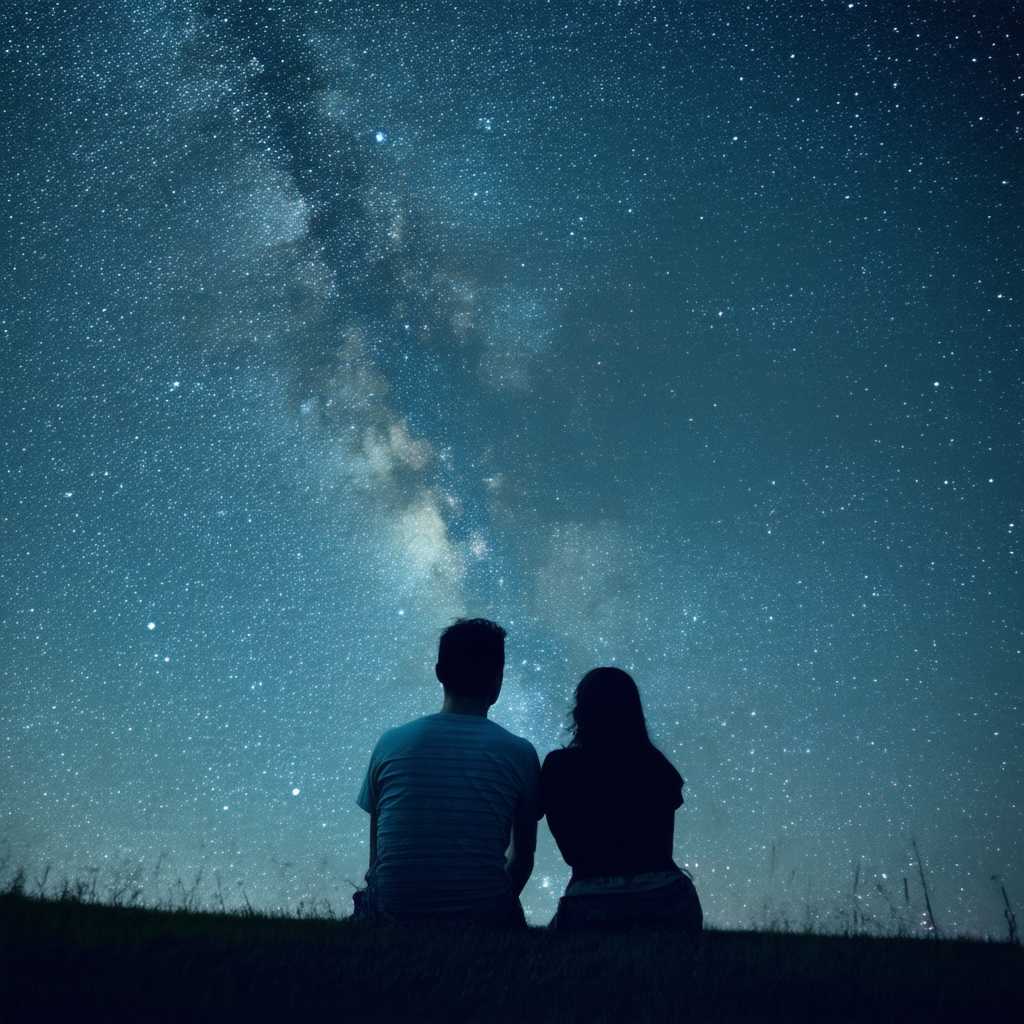} \\[-1pt]
        \ourslabel & \includegraphics[width=\imgwidth]{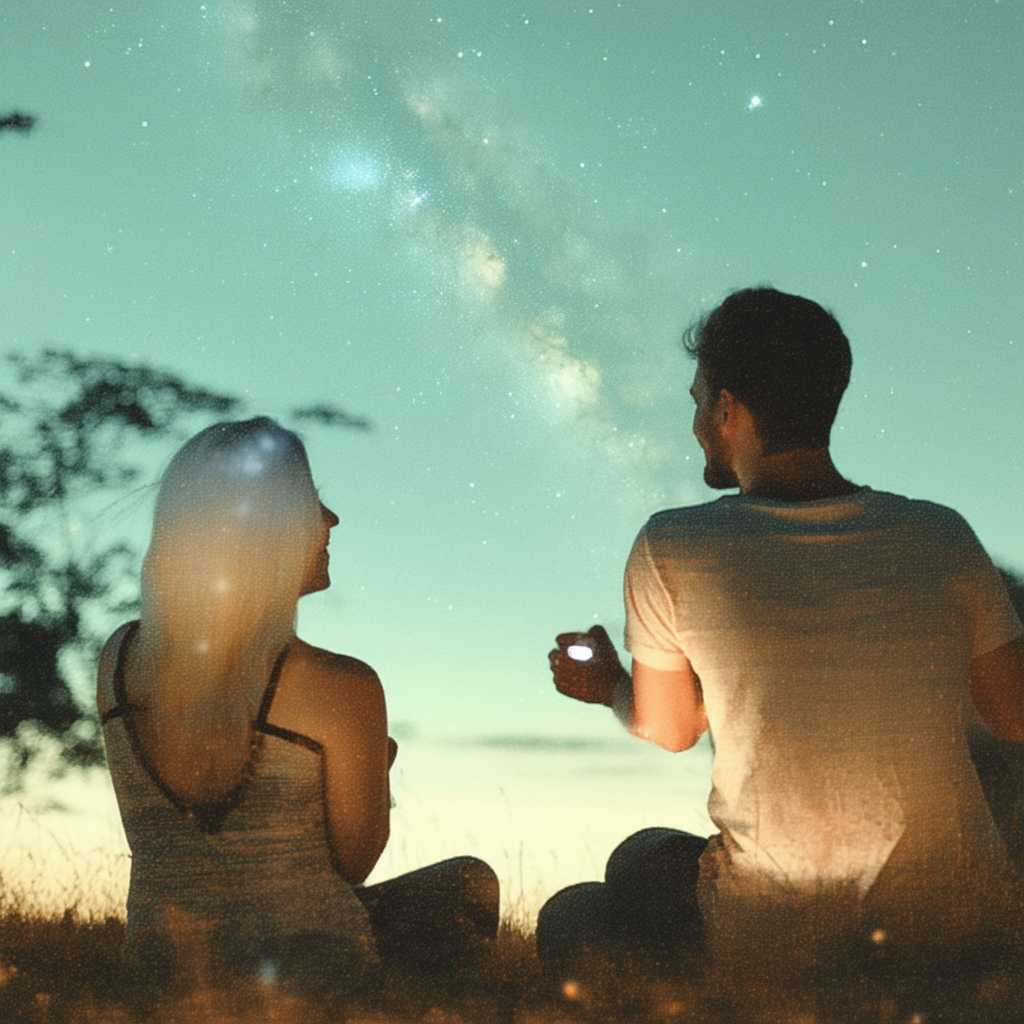} & \includegraphics[width=\imgwidth]{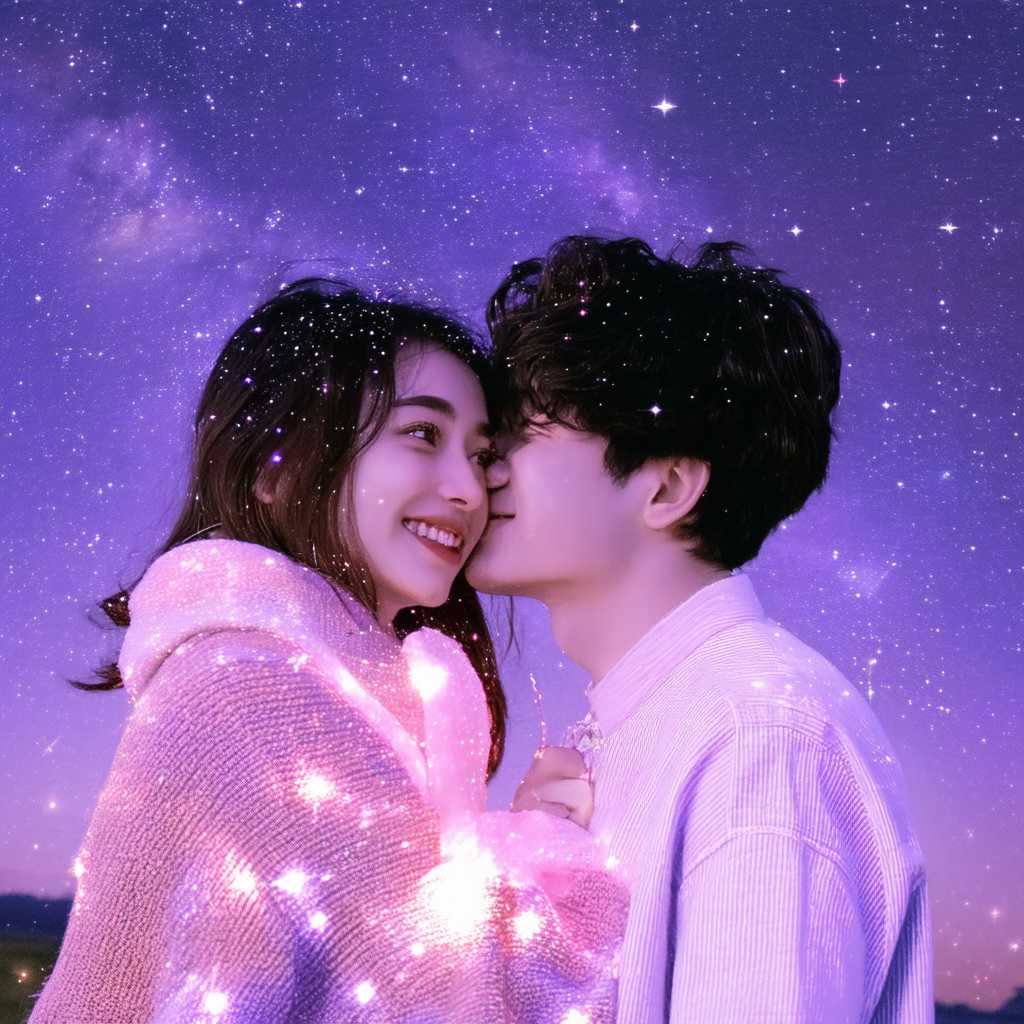} & \includegraphics[width=\imgwidth]{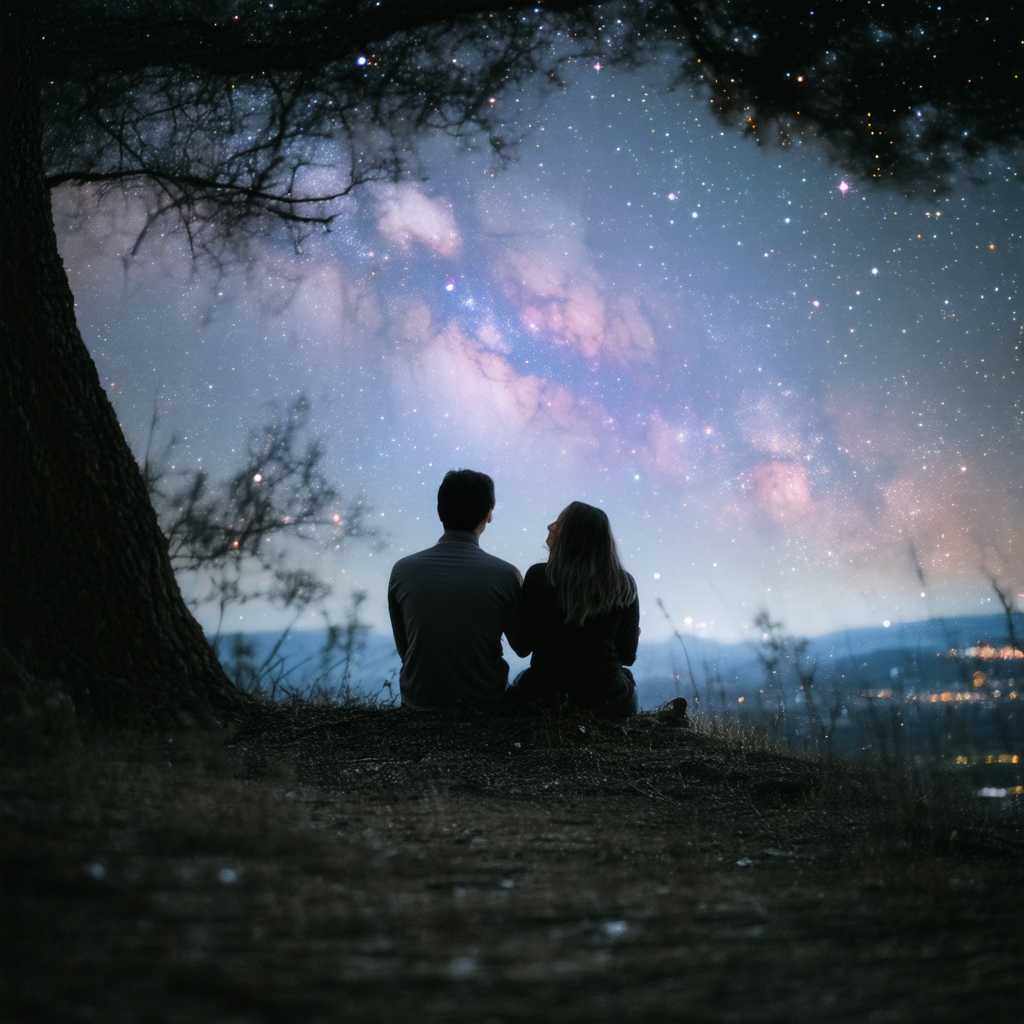} & \includegraphics[width=\imgwidth]{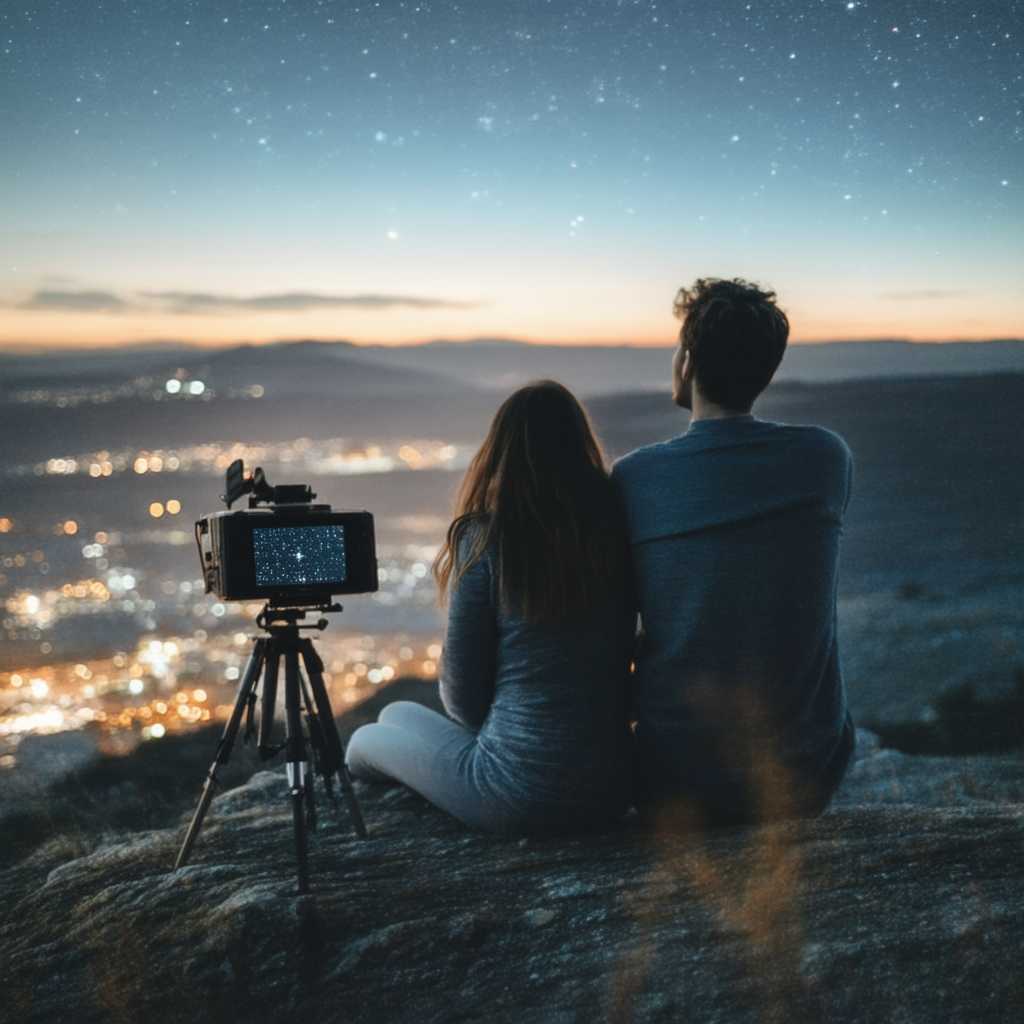} \\
        \multicolumn{5}{c}{\vspace{2pt}\small ``A couple stargazing'' \vspace{8pt}} \\

        \largelabel & \includegraphics[width=\imgwidth]{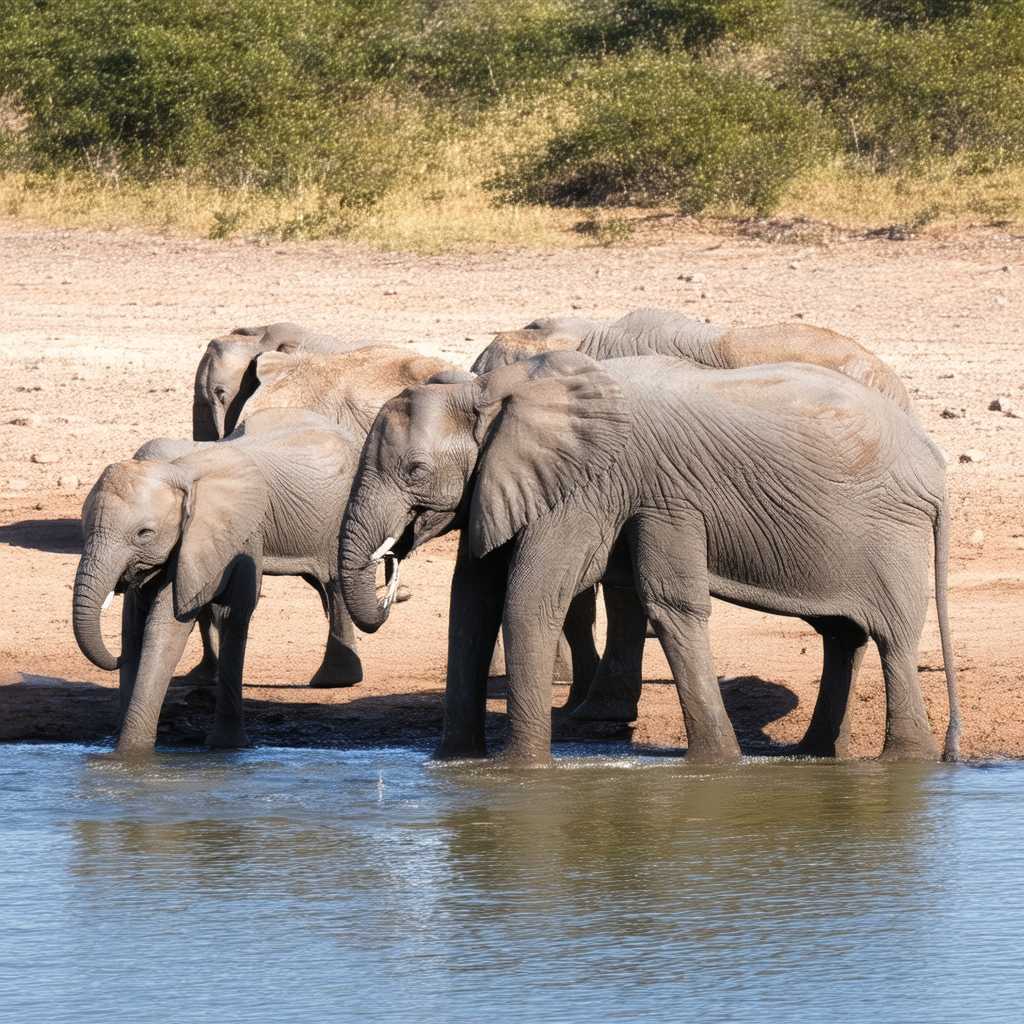} & \includegraphics[width=\imgwidth]{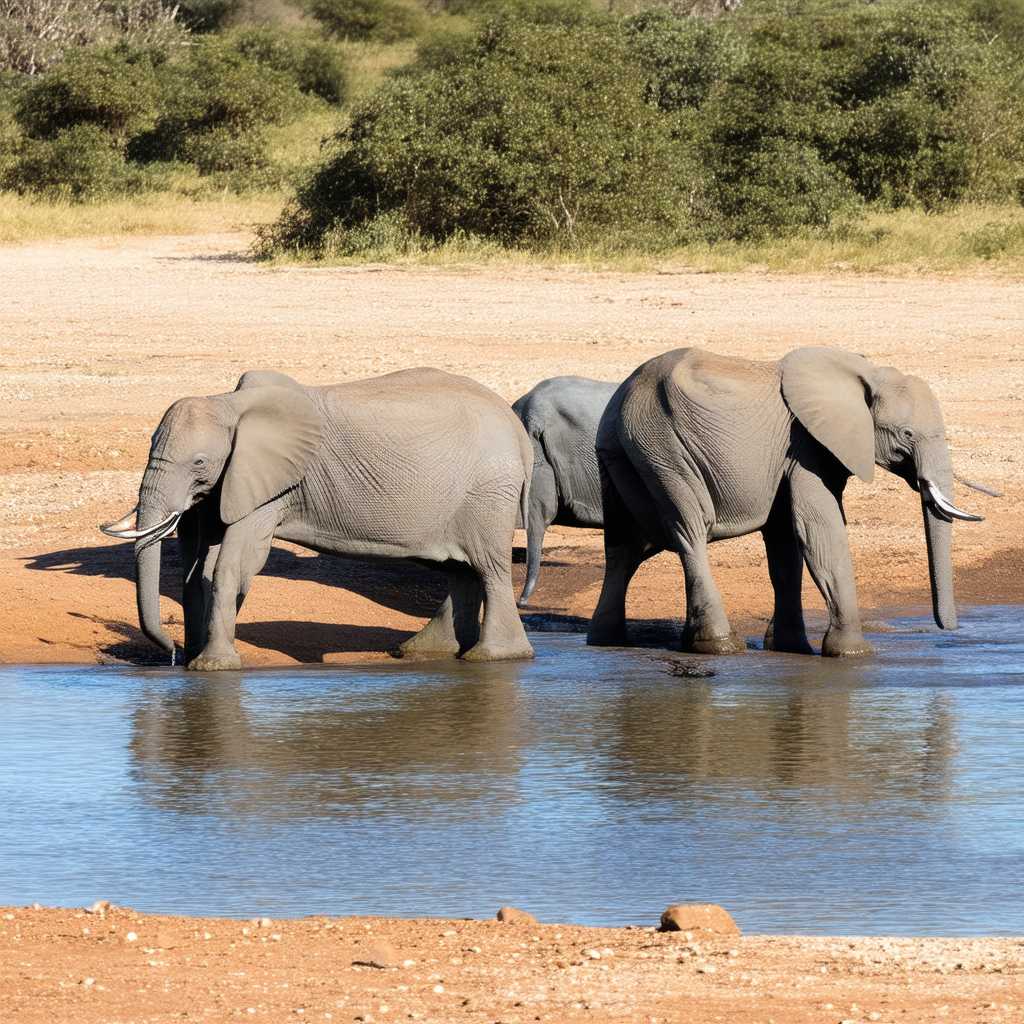} & \includegraphics[width=\imgwidth]{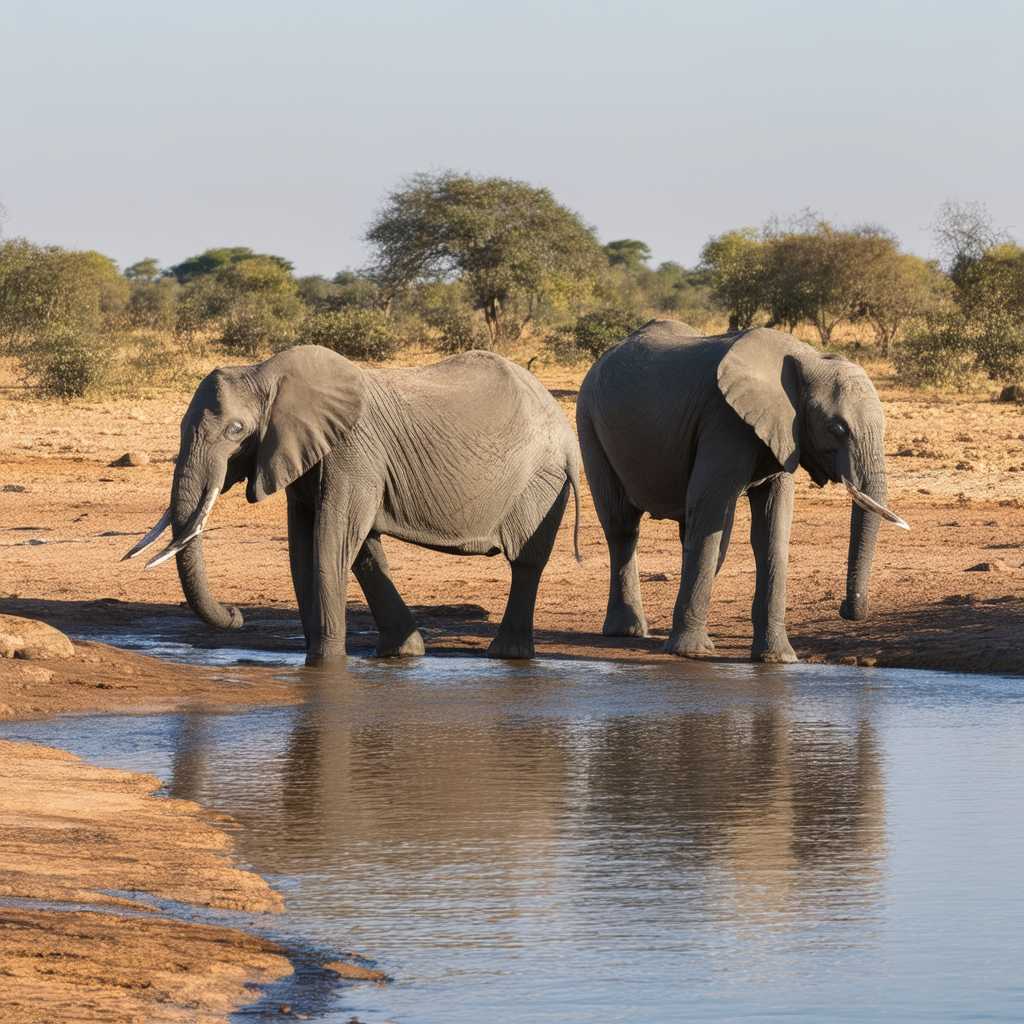} & \includegraphics[width=\imgwidth]{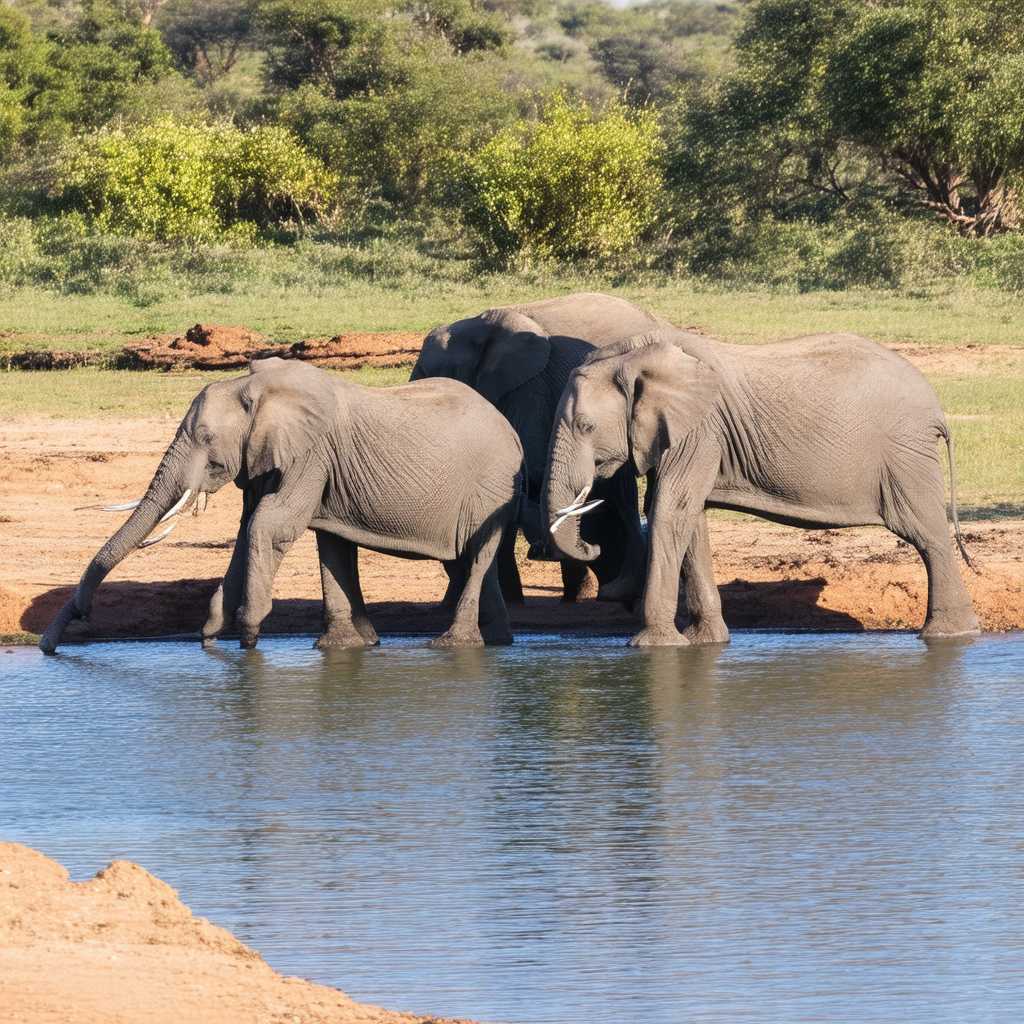} \\[-1pt]
        \ourslabel & \includegraphics[width=\imgwidth]{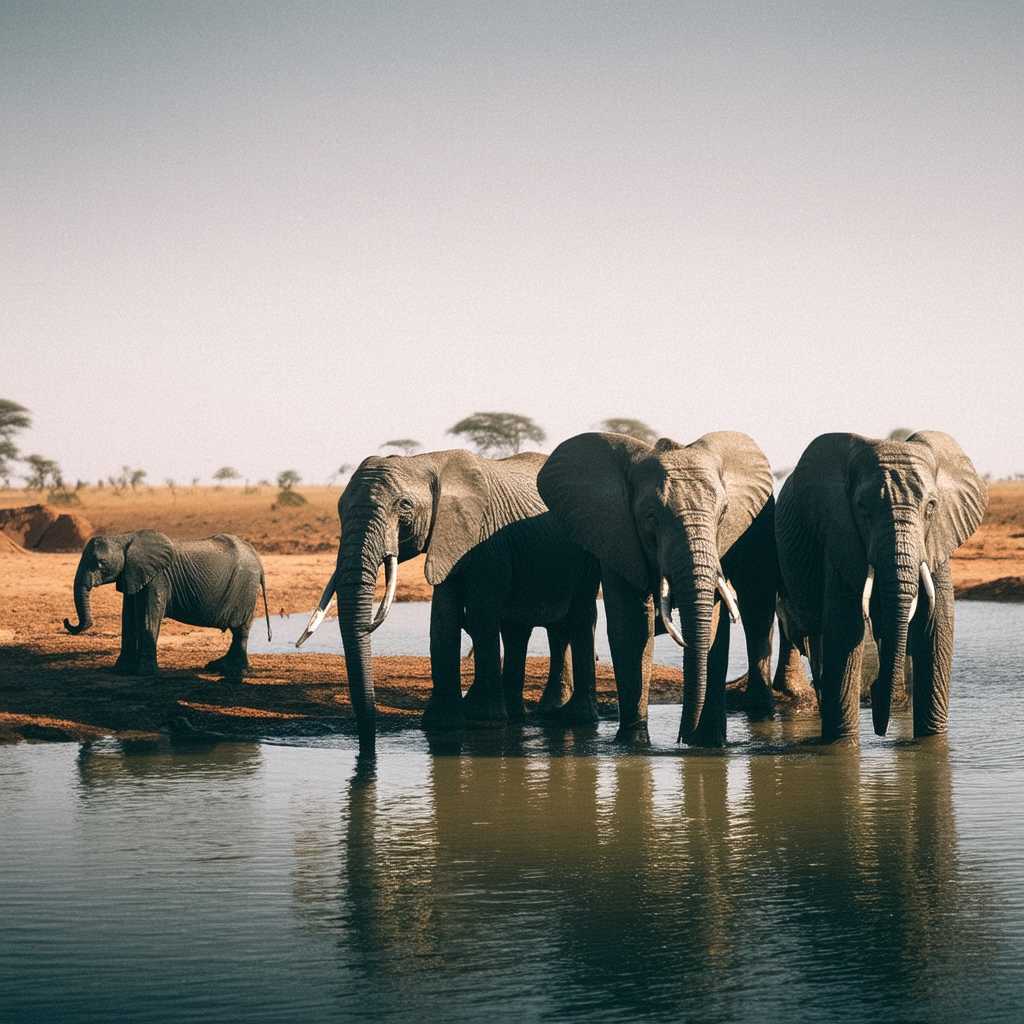} & \includegraphics[width=\imgwidth]{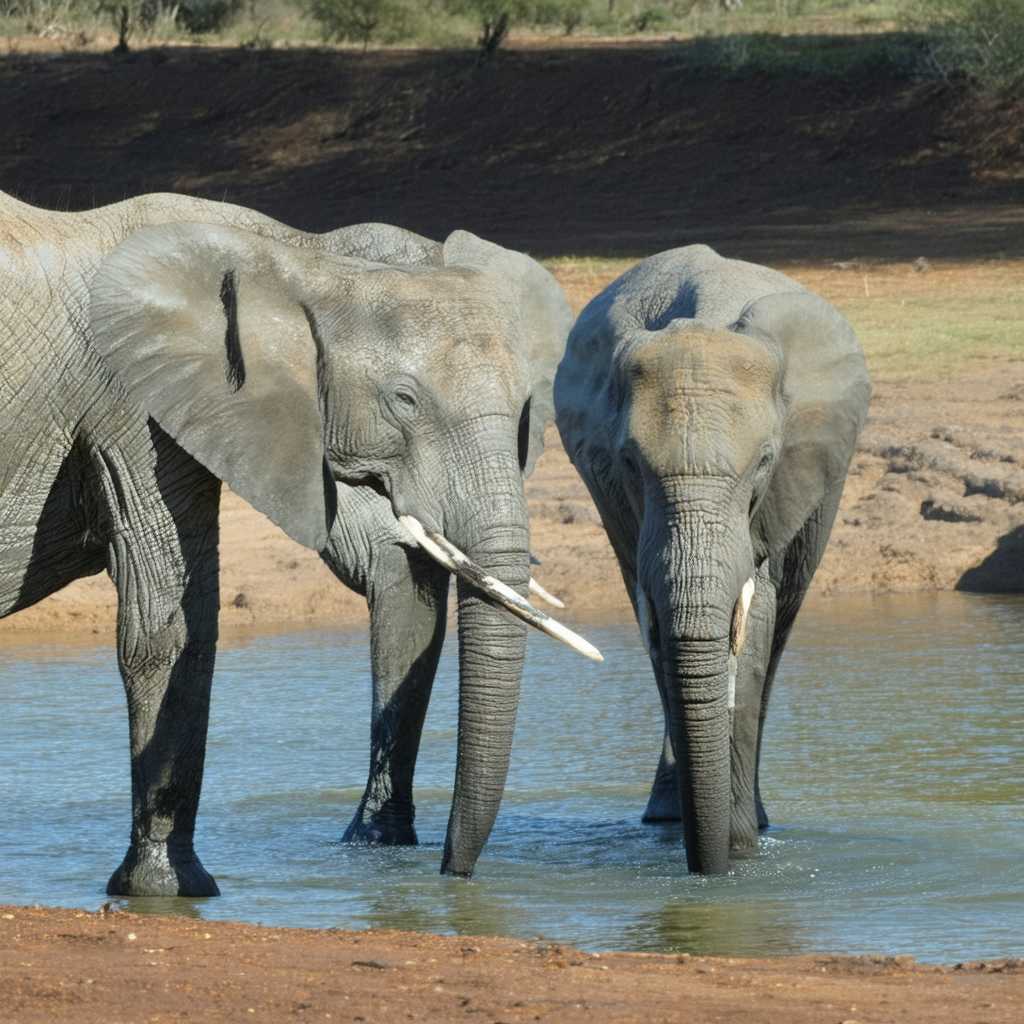} & \includegraphics[width=\imgwidth]{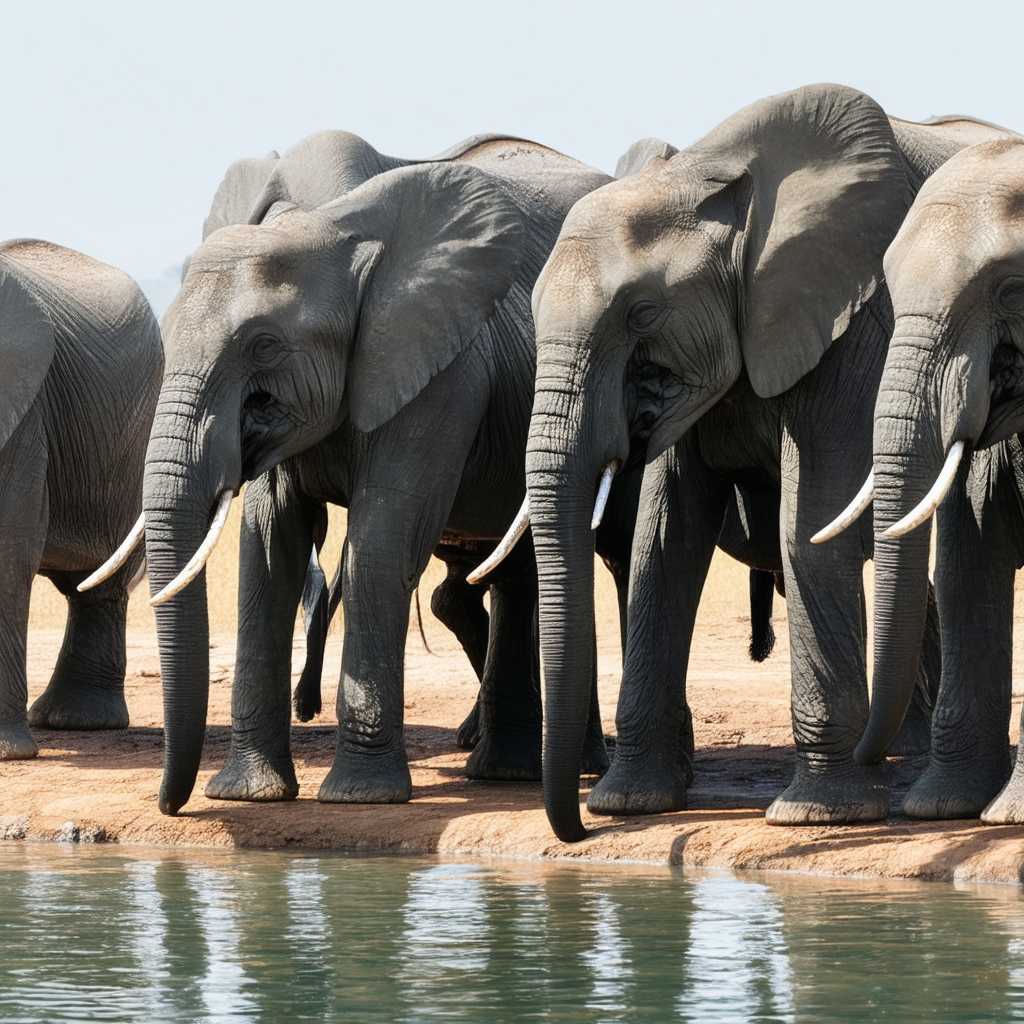} & \includegraphics[width=\imgwidth]{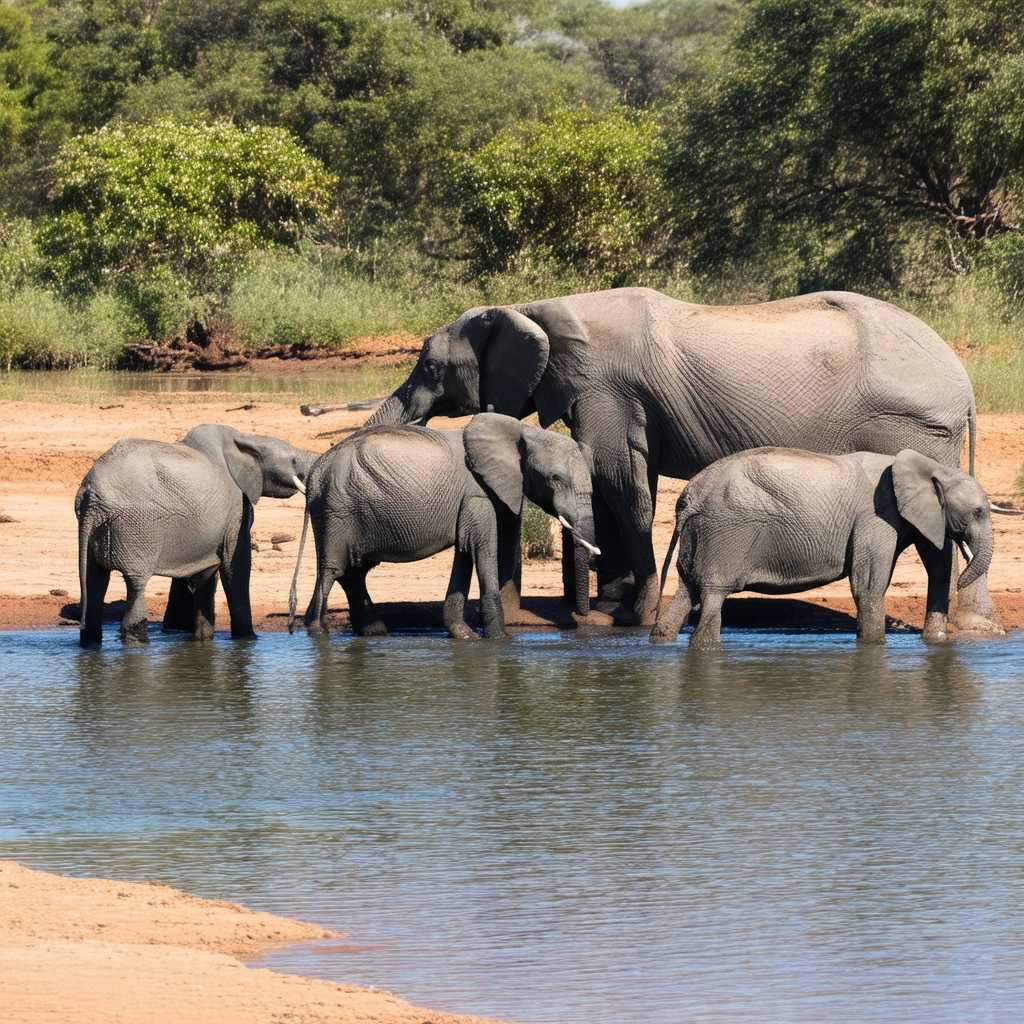} \\
        \multicolumn{5}{c}{\vspace{2pt}\small ``Elephants at a waterhole'' \vspace{8pt}} \\

        \largelabel & \includegraphics[width=\imgwidth]{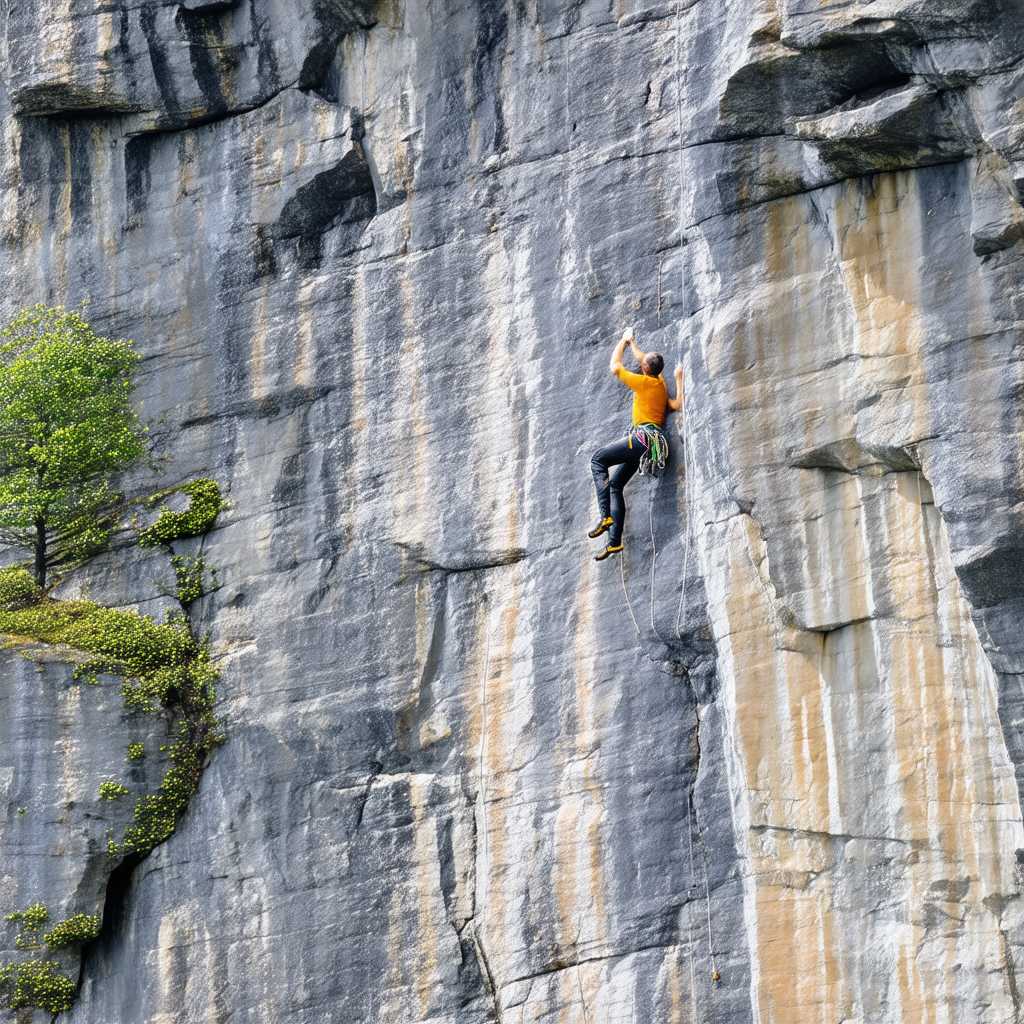} & \includegraphics[width=\imgwidth]{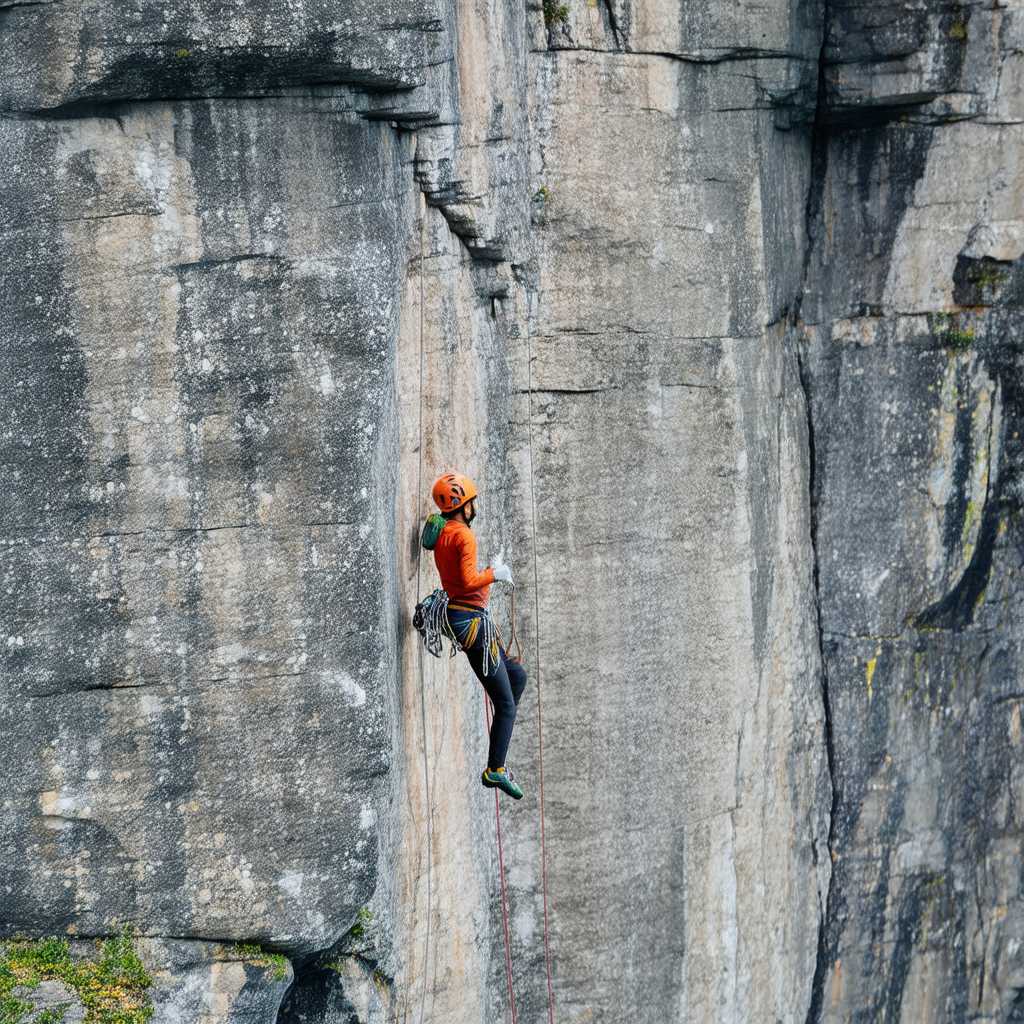} & \includegraphics[width=\imgwidth]{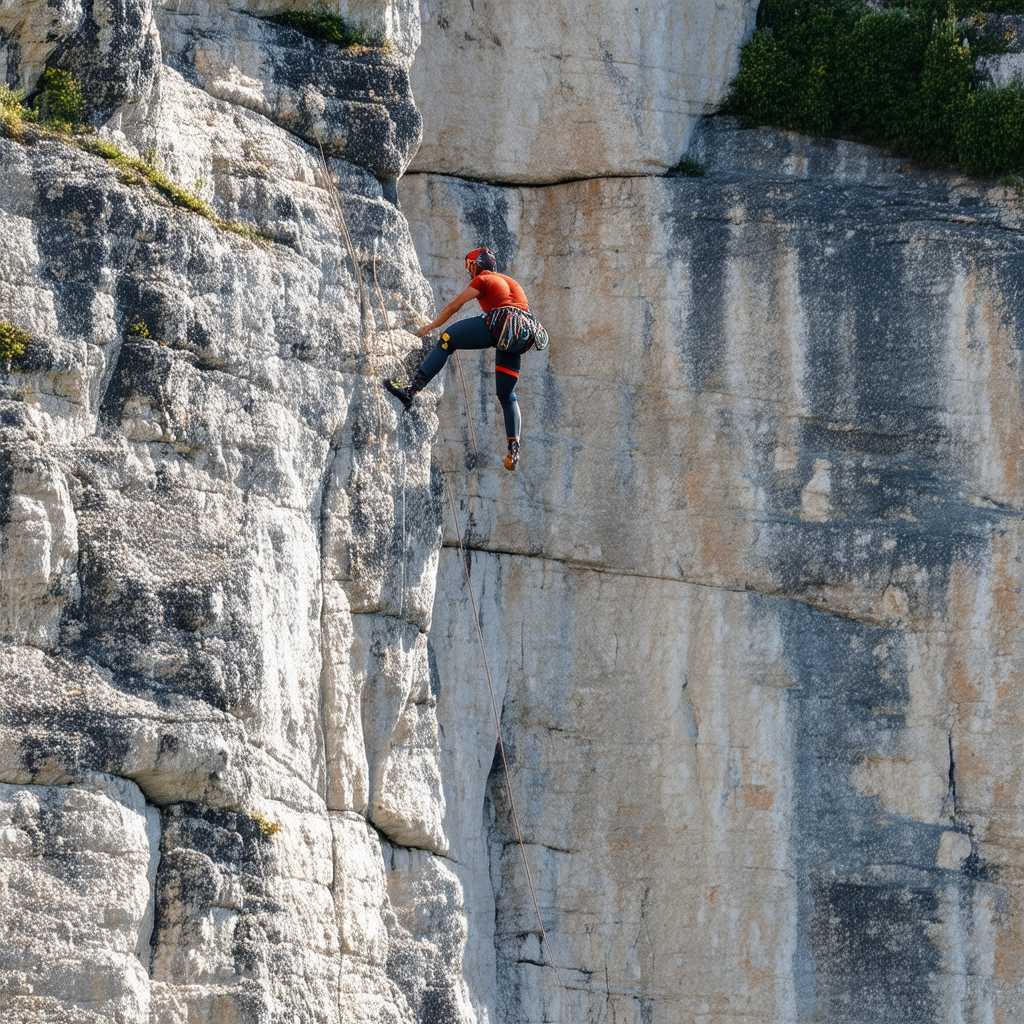} & \includegraphics[width=\imgwidth]{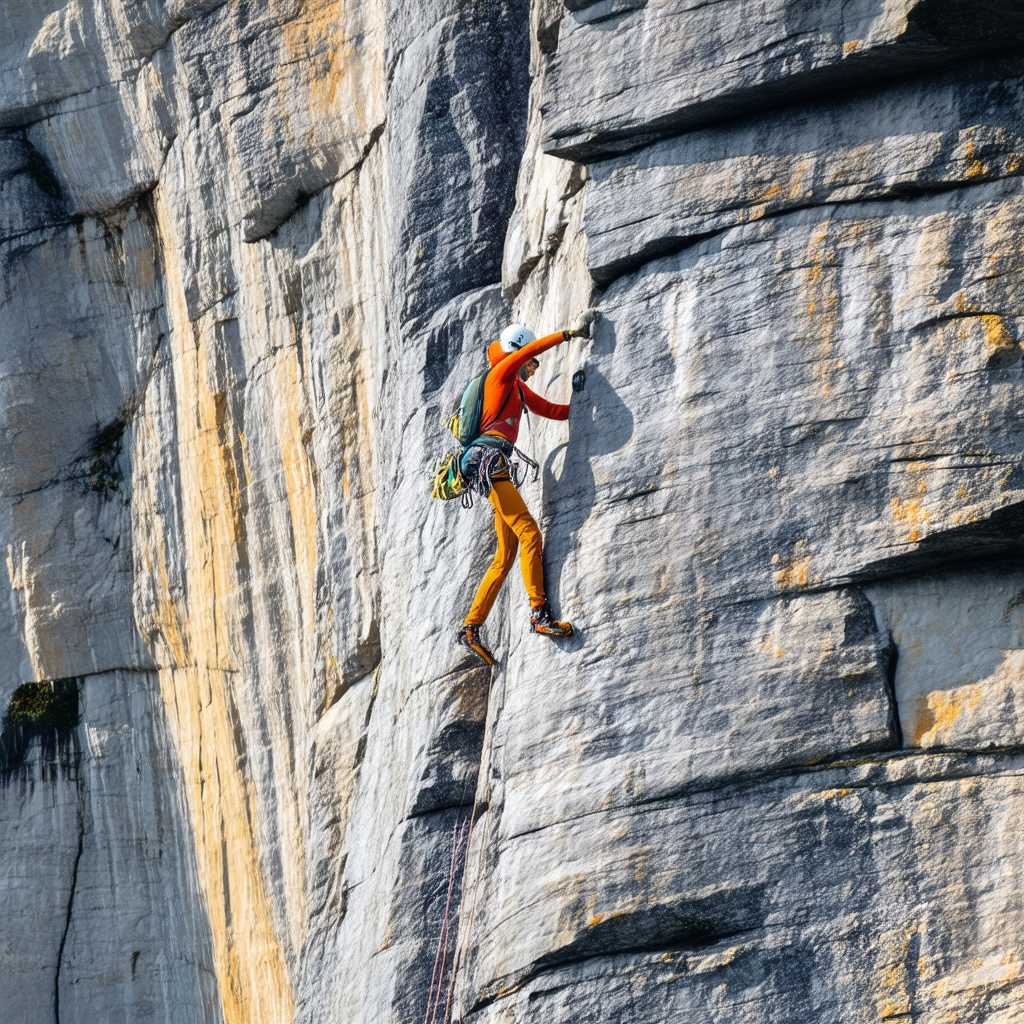} \\[-1pt]
        \ourslabel & \includegraphics[width=\imgwidth]{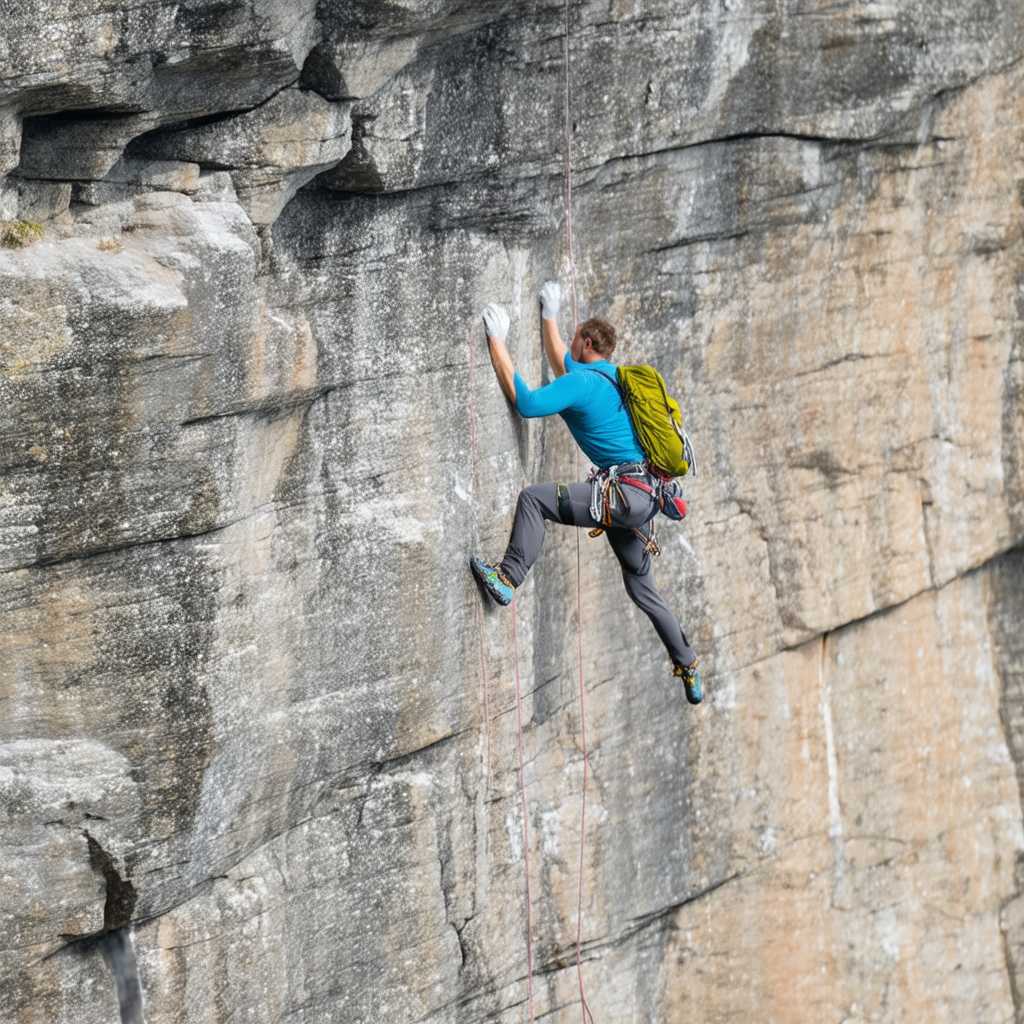} & \includegraphics[width=\imgwidth]{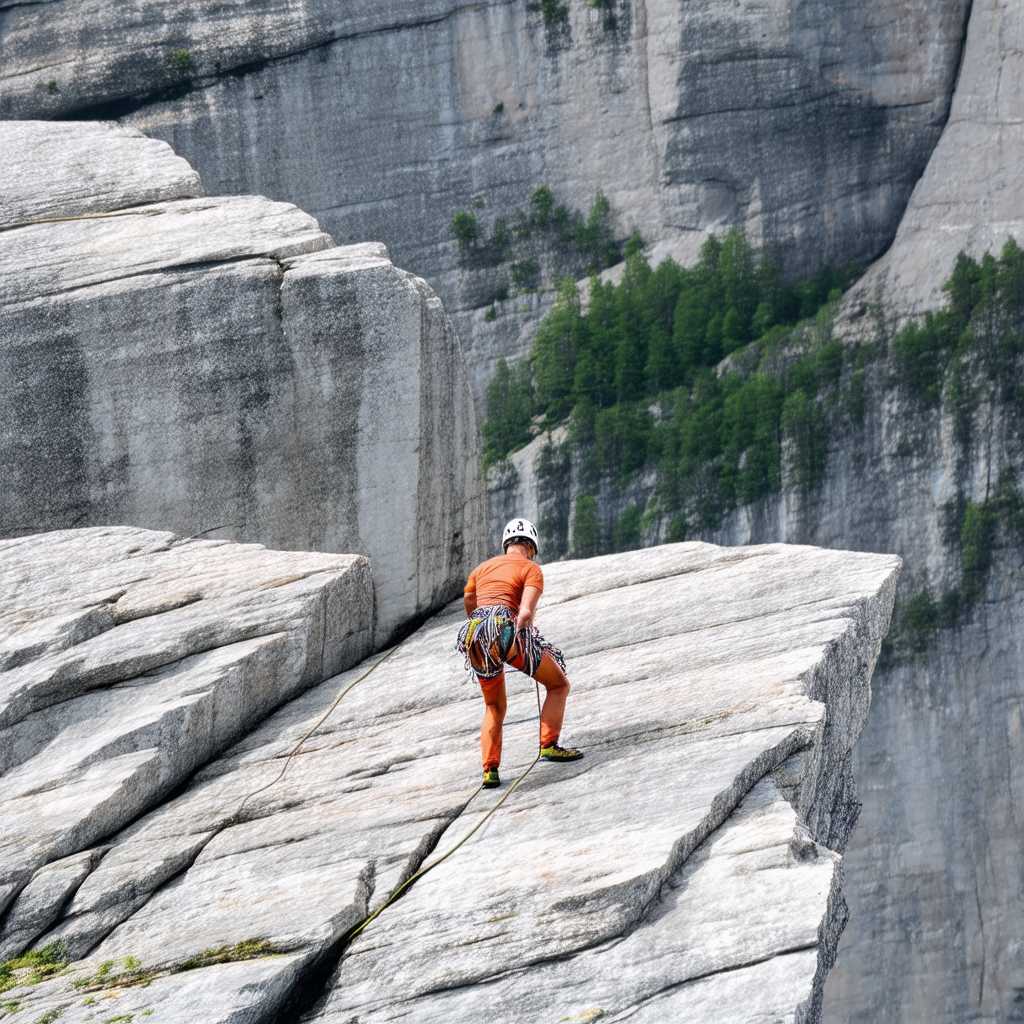} & \includegraphics[width=\imgwidth]{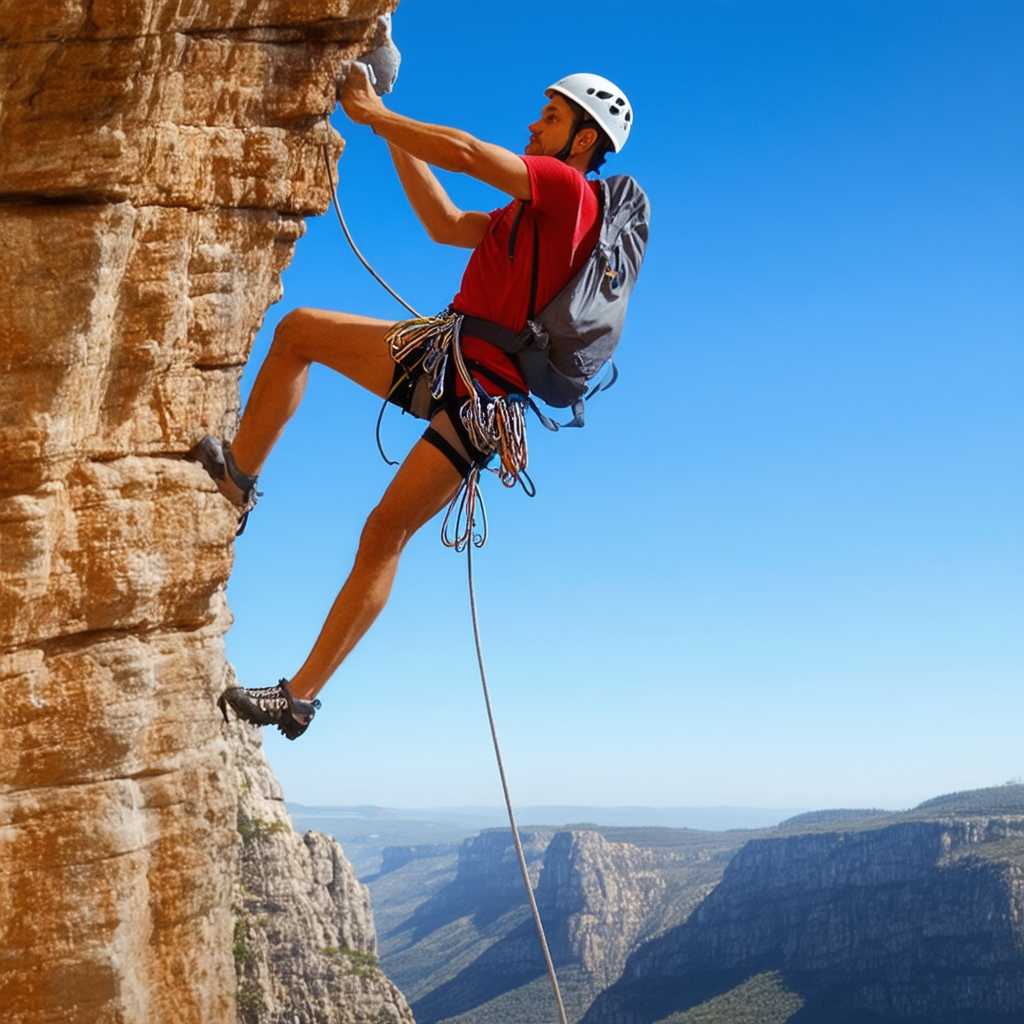} & \includegraphics[width=\imgwidth]{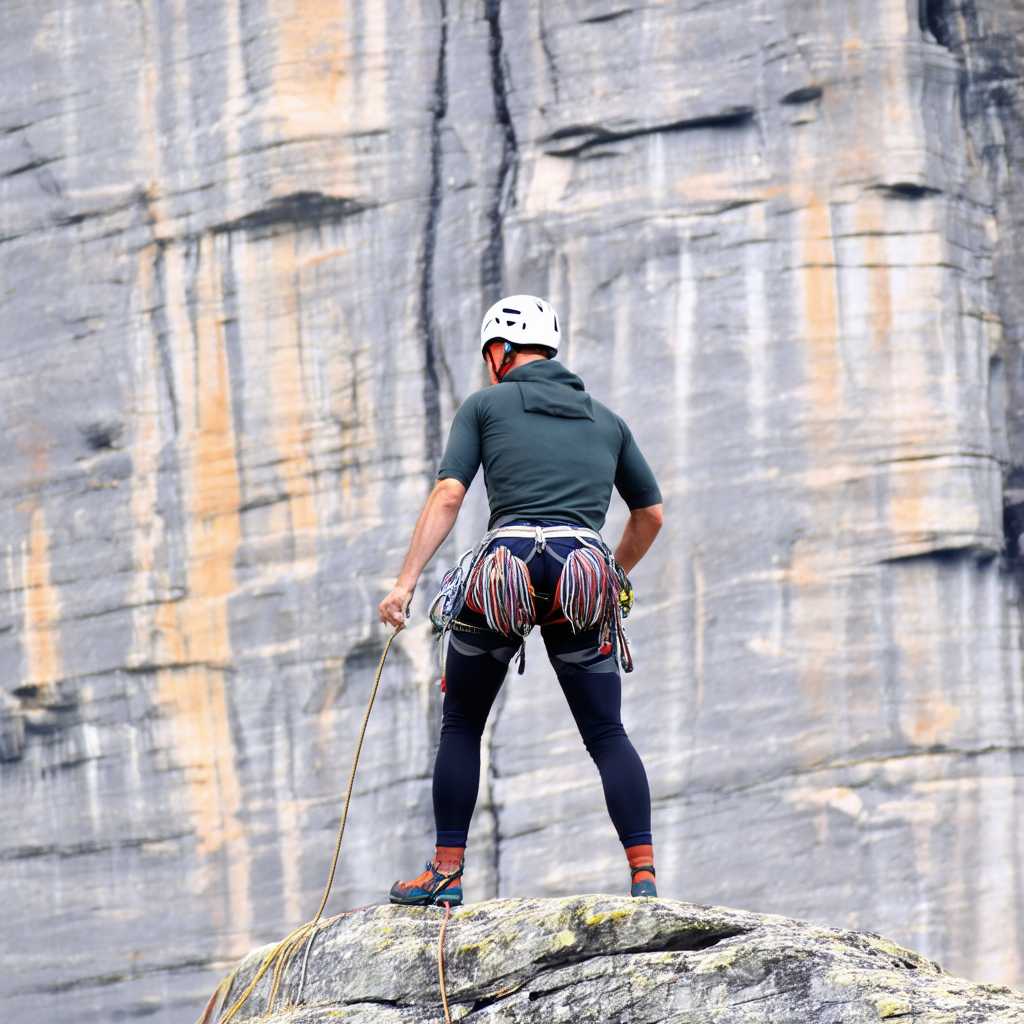} \\
        \multicolumn{5}{c}{\vspace{2pt}\small ``A climber on a cliff'' \vspace{8pt}} \\
    \end{tabular}
    \caption{\textbf{Qualitative results on SD3.5-Large.}}
    \label{fig:large}
\end{figure}

\begin{figure}
    \centering
    \setlength{\tabcolsep}{0.5pt} \renewcommand{\arraystretch}{0.5} \newcommand{\imgwidth}{0.12\textwidth}
    \newcommand{\vertlabel}[2]{\raisebox{#2}{\rotatebox{90}{\scriptsize\textbf{#1}}}}
    
    \newcommand{\turbolabel}{\vertlabel{SD3.5-Turbo}{1.5em}}
    
    \newcommand{\ourslabel}{ \vertlabel{Ours}{2.5em}}

    \begin{tabular}{c c c c c}
        \turbolabel & \includegraphics[width=\imgwidth]{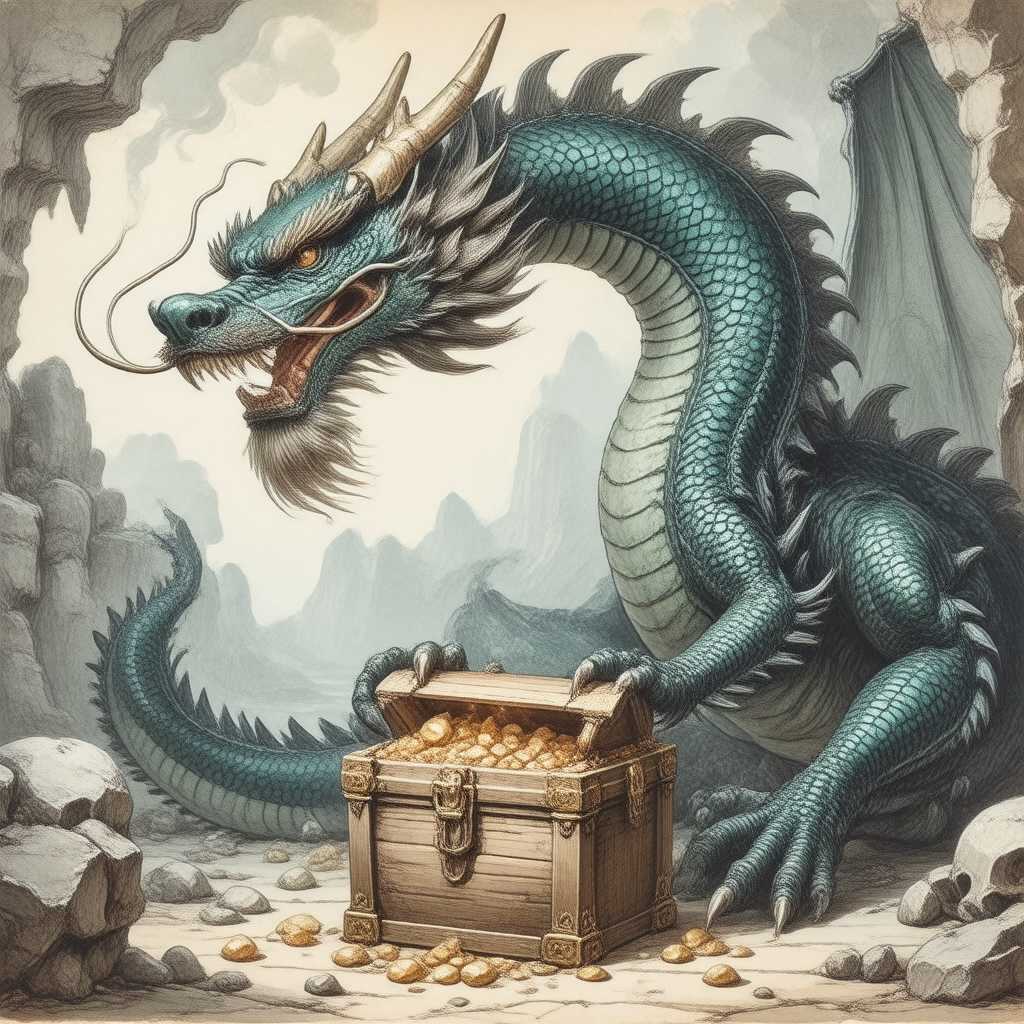} & \includegraphics[width=\imgwidth]{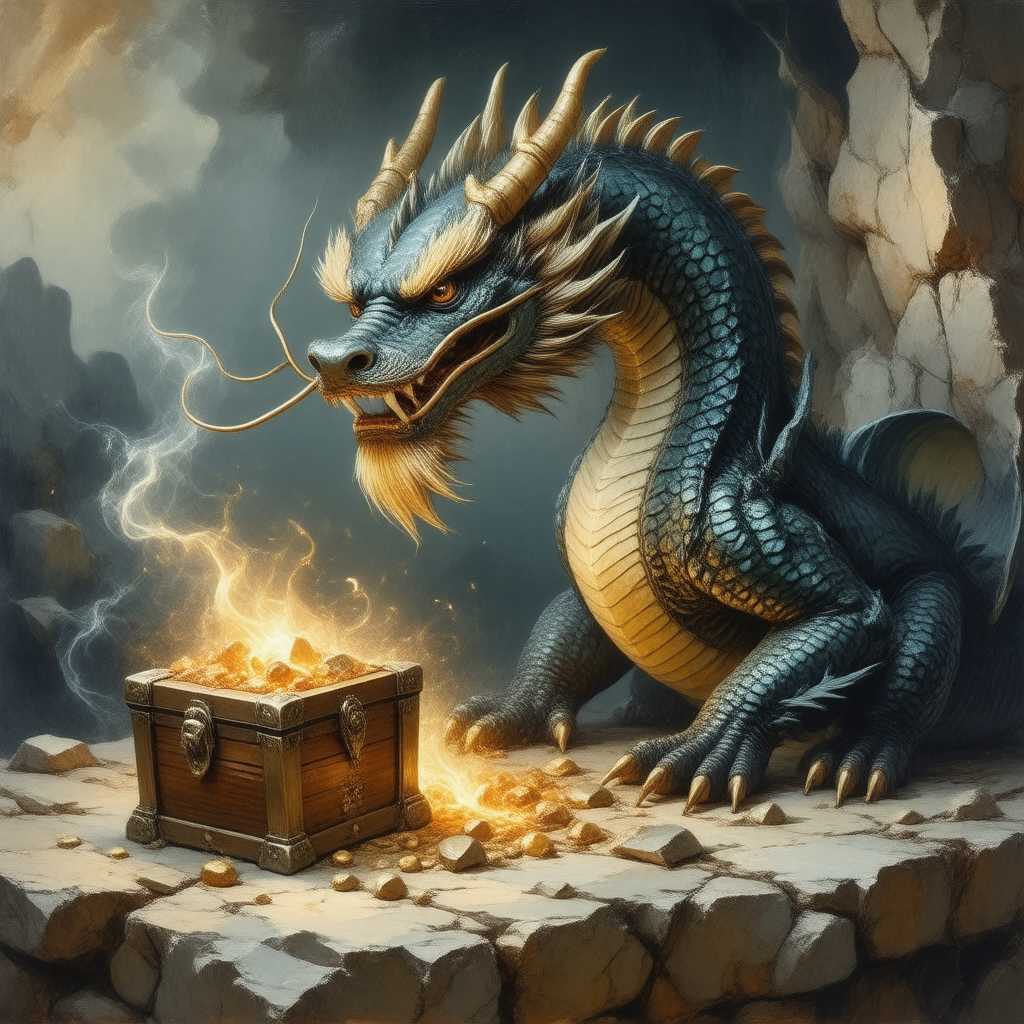} & \includegraphics[width=\imgwidth]{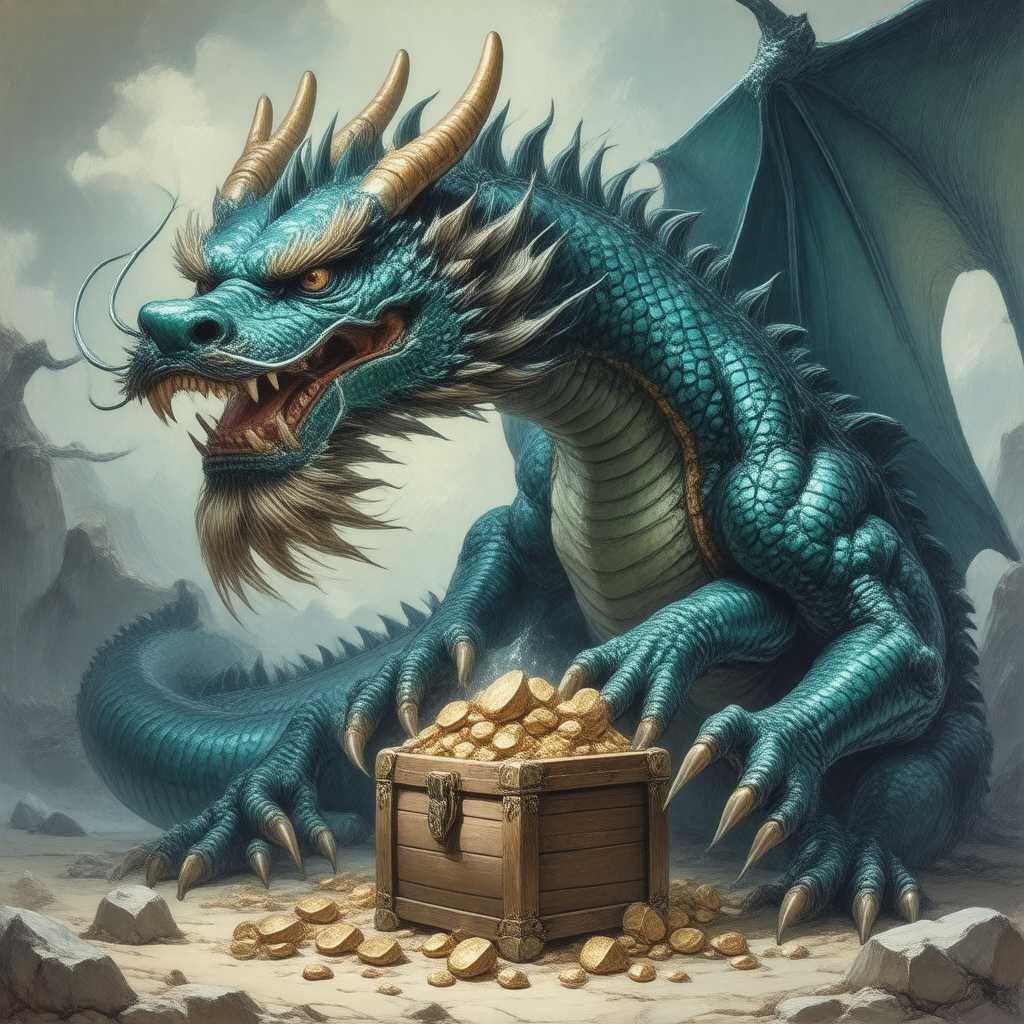} & \includegraphics[width=\imgwidth]{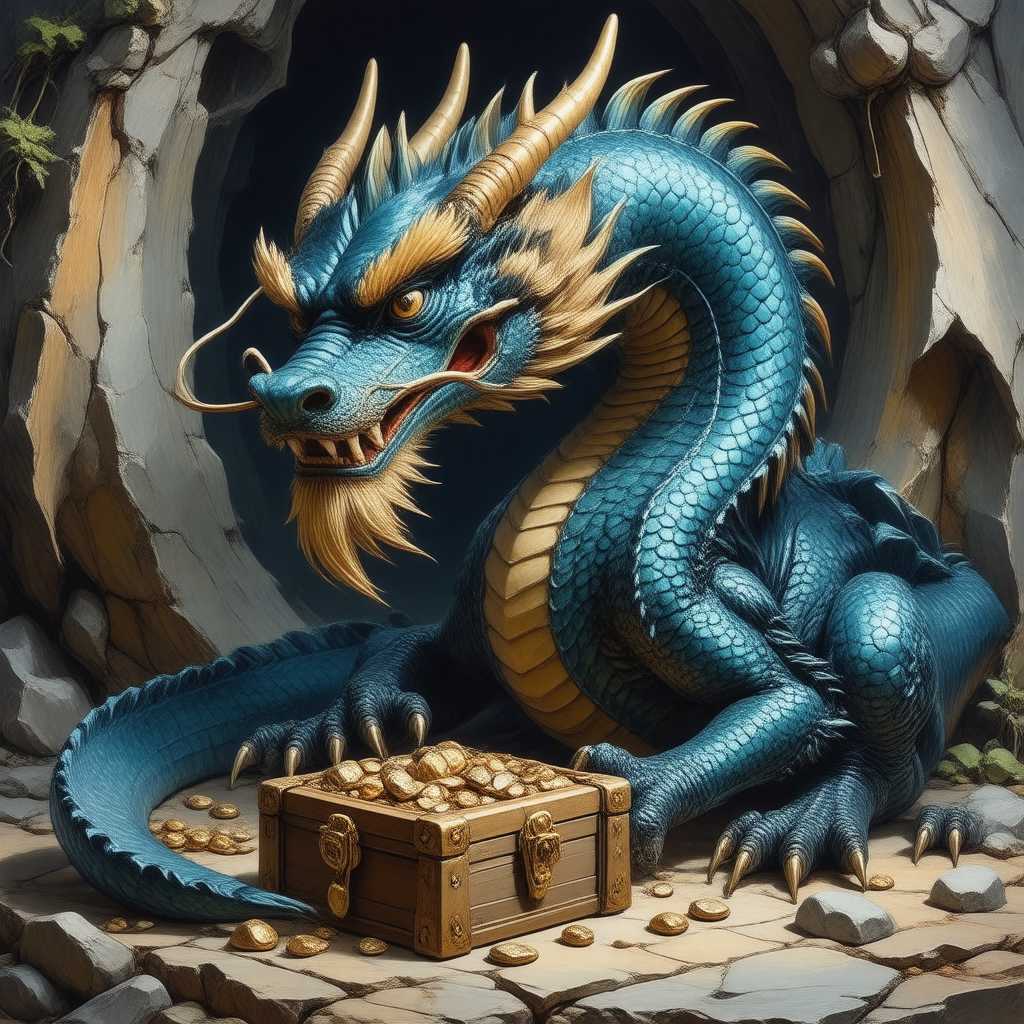} \\[-1pt]
        \ourslabel & \includegraphics[width=\imgwidth]{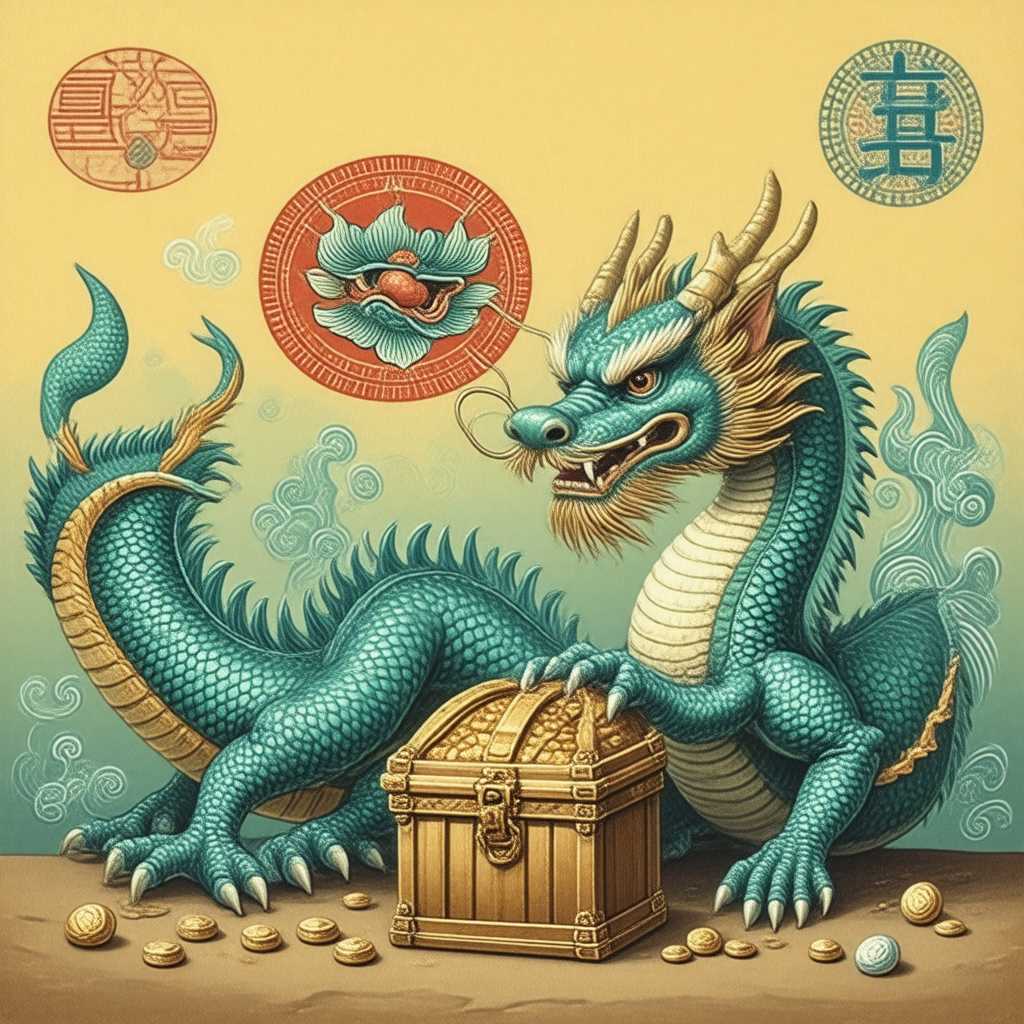} & \includegraphics[width=\imgwidth]{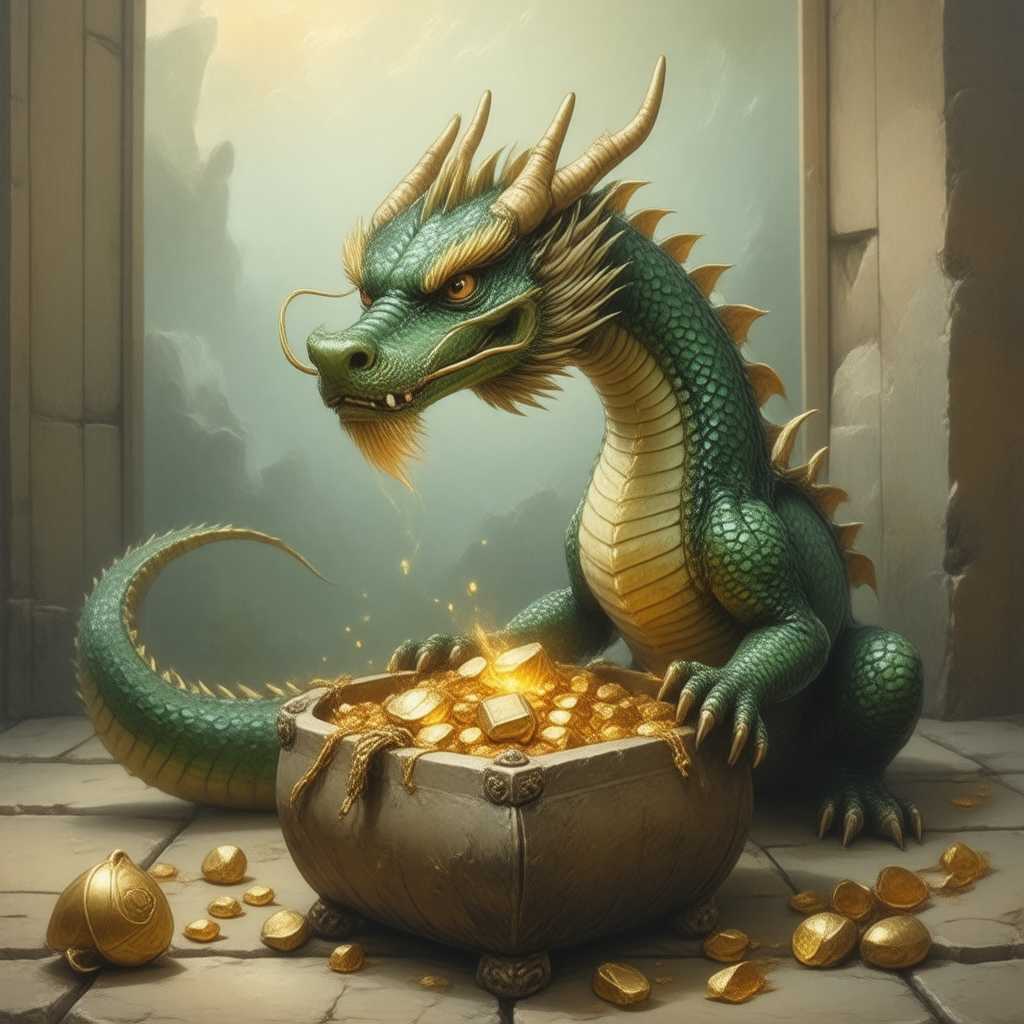} & \includegraphics[width=\imgwidth]{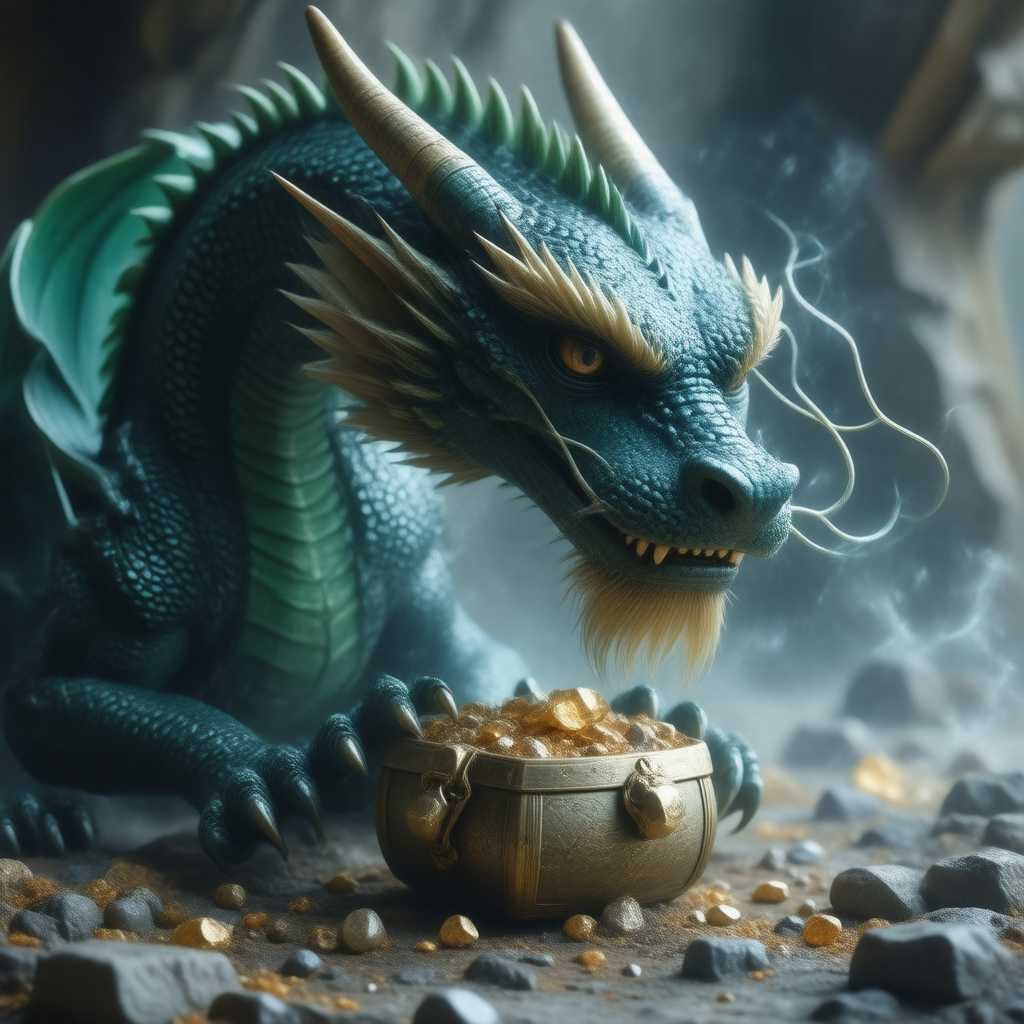} & \includegraphics[width=\imgwidth]{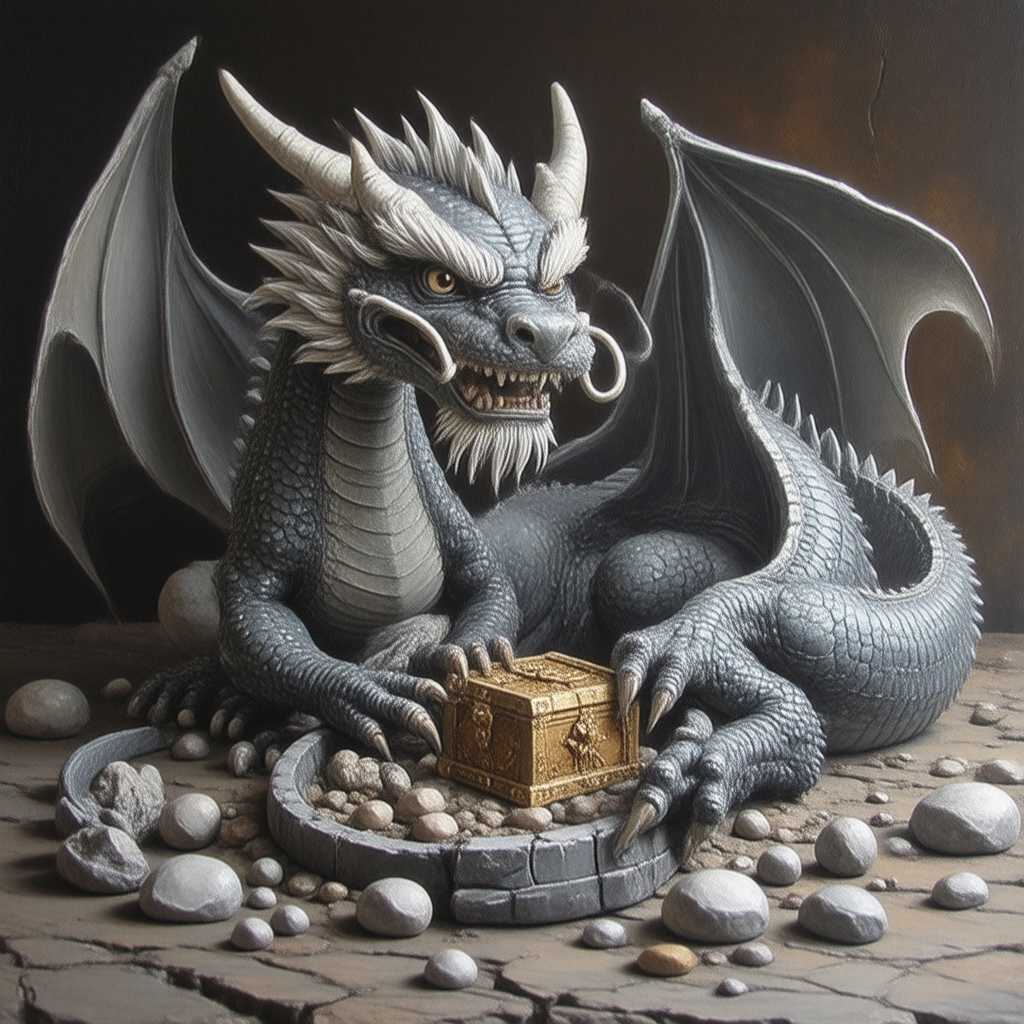} \\
        \multicolumn{5}{c}{\vspace{2pt}\small ``A dragon guarding its treasure'' \vspace{8pt}} \\

        \turbolabel & \includegraphics[width=\imgwidth]{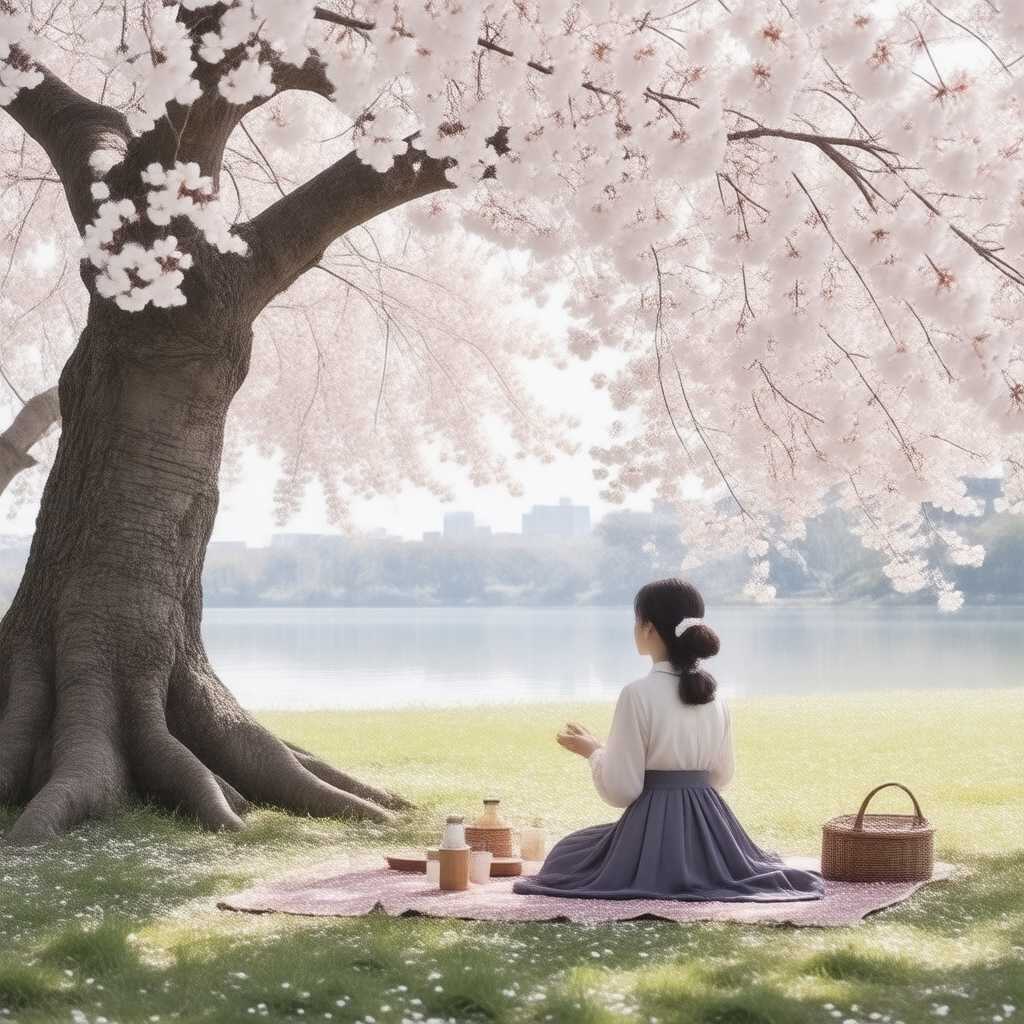} & \includegraphics[width=\imgwidth]{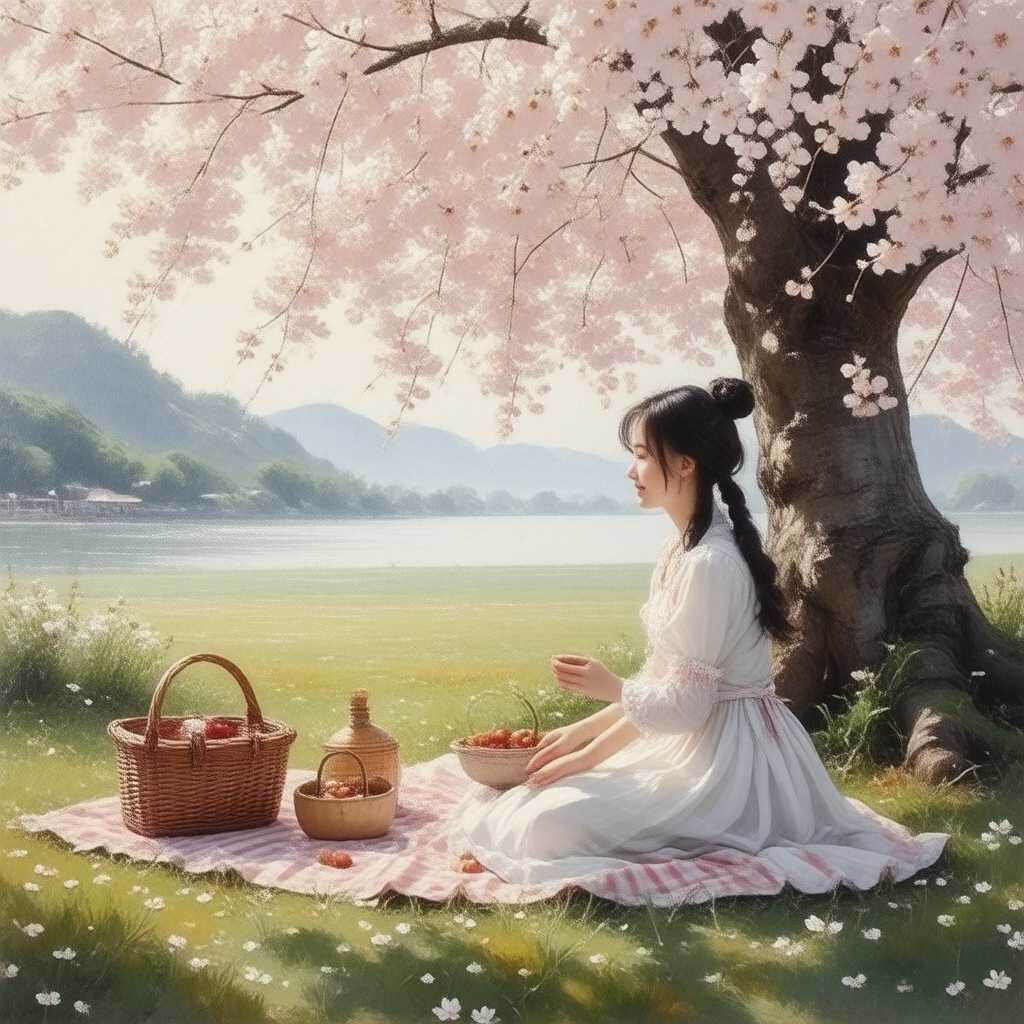} & \includegraphics[width=\imgwidth]{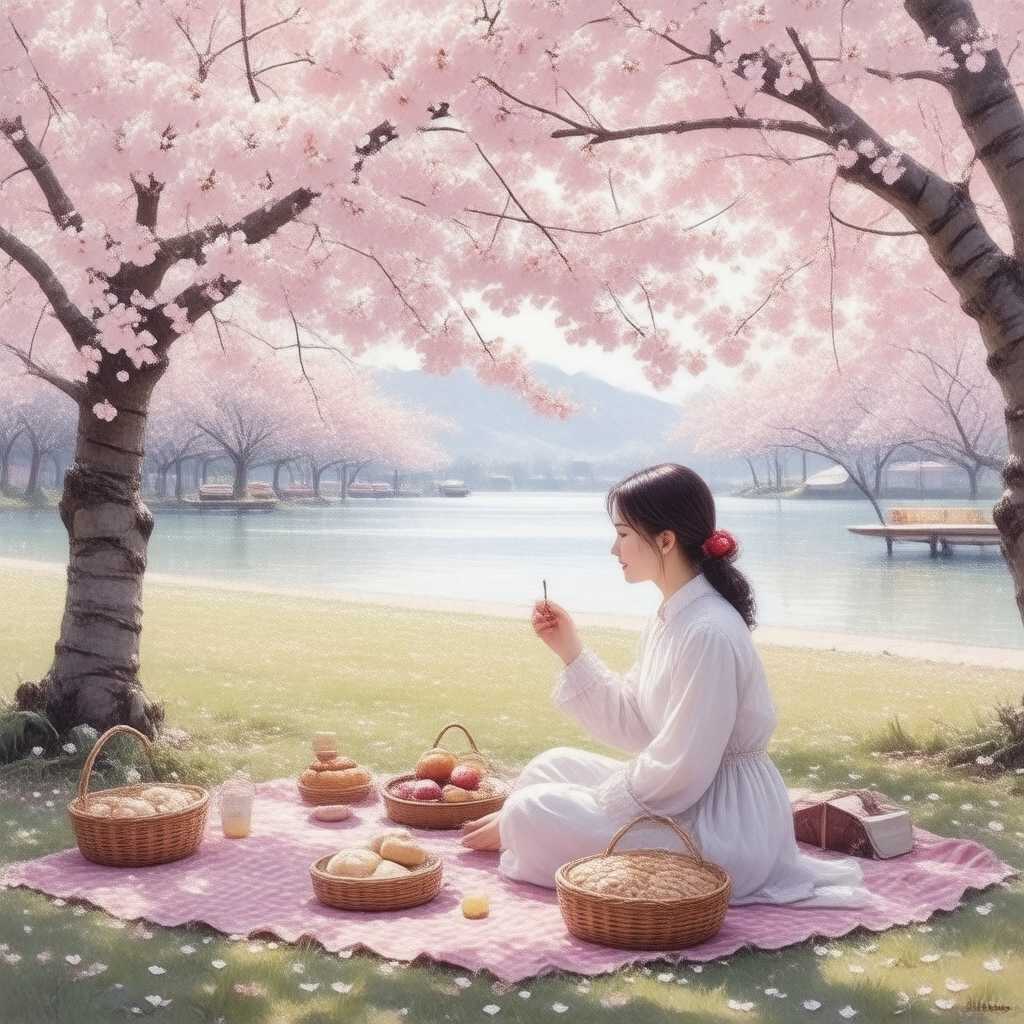} & \includegraphics[width=\imgwidth]{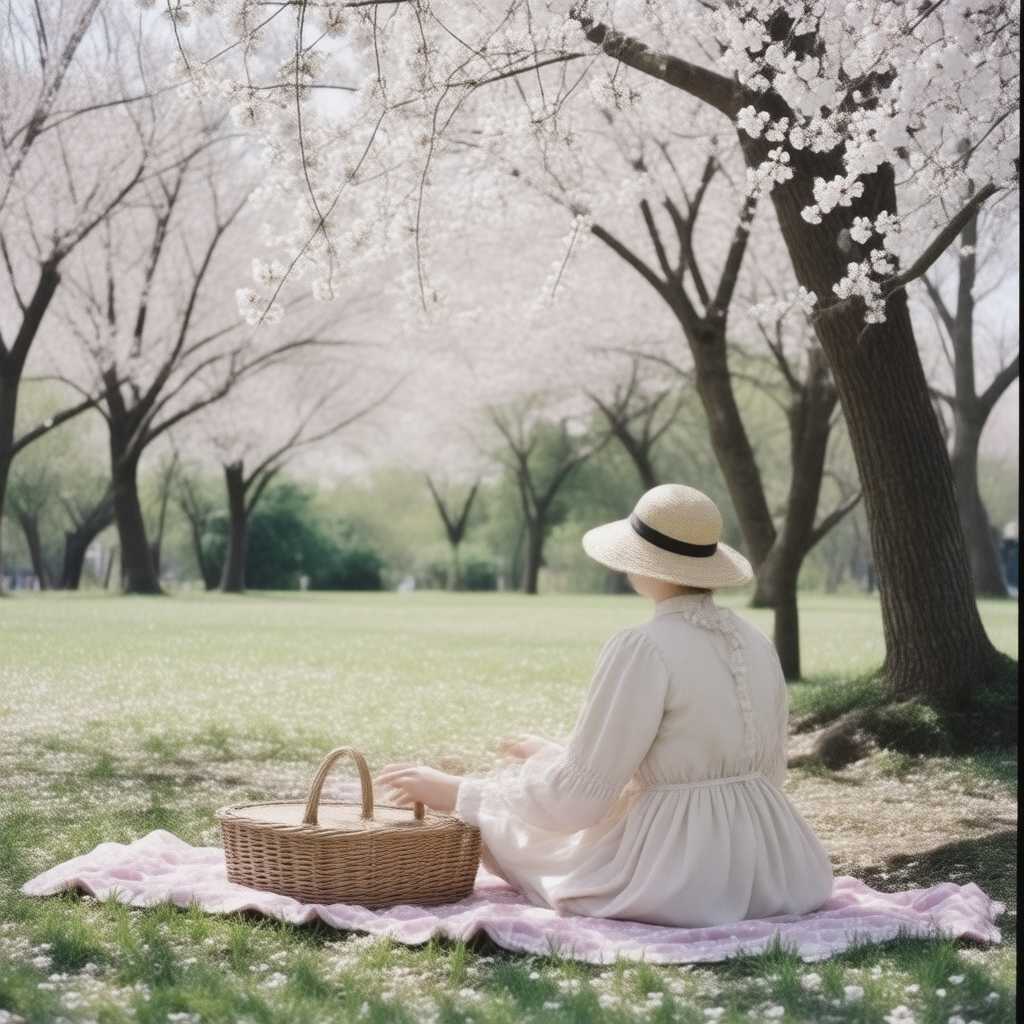} \\[-1pt]
        \ourslabel & \includegraphics[width=\imgwidth]{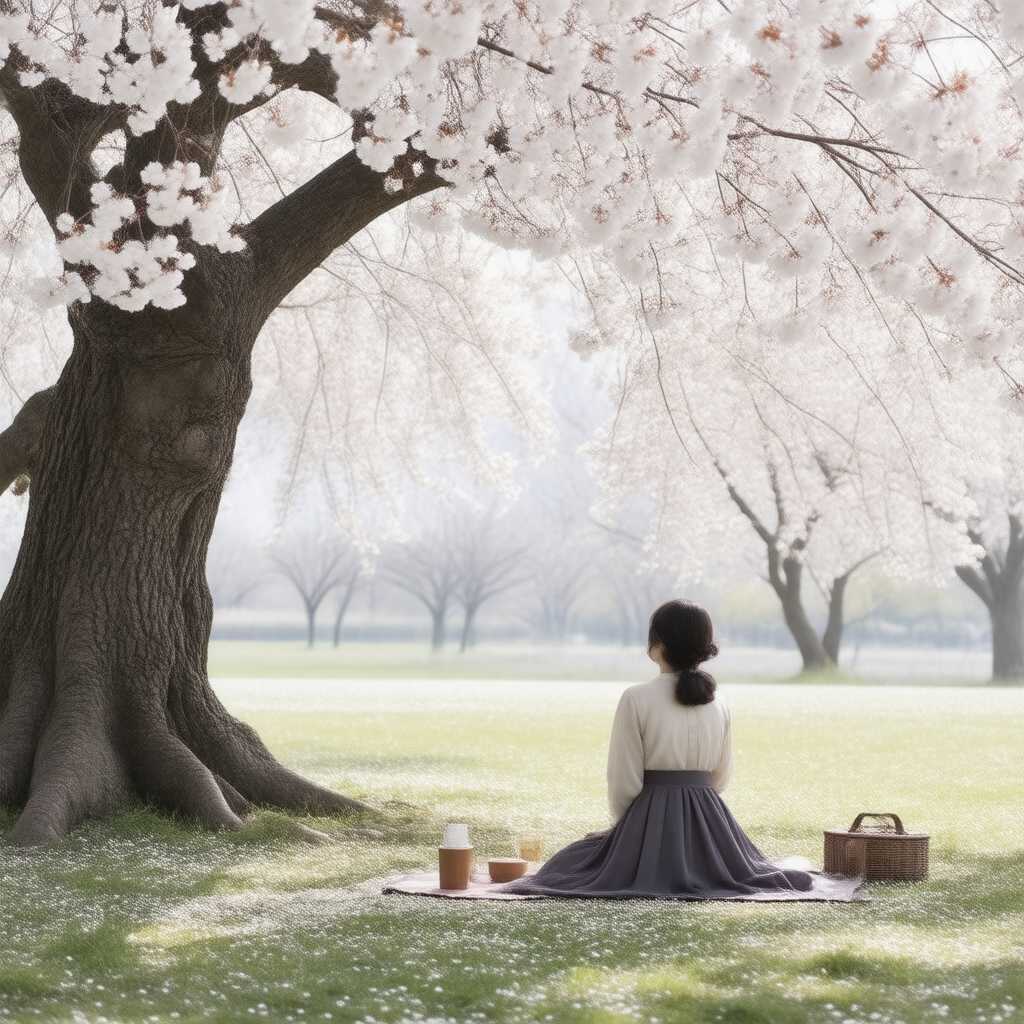} & \includegraphics[width=\imgwidth]{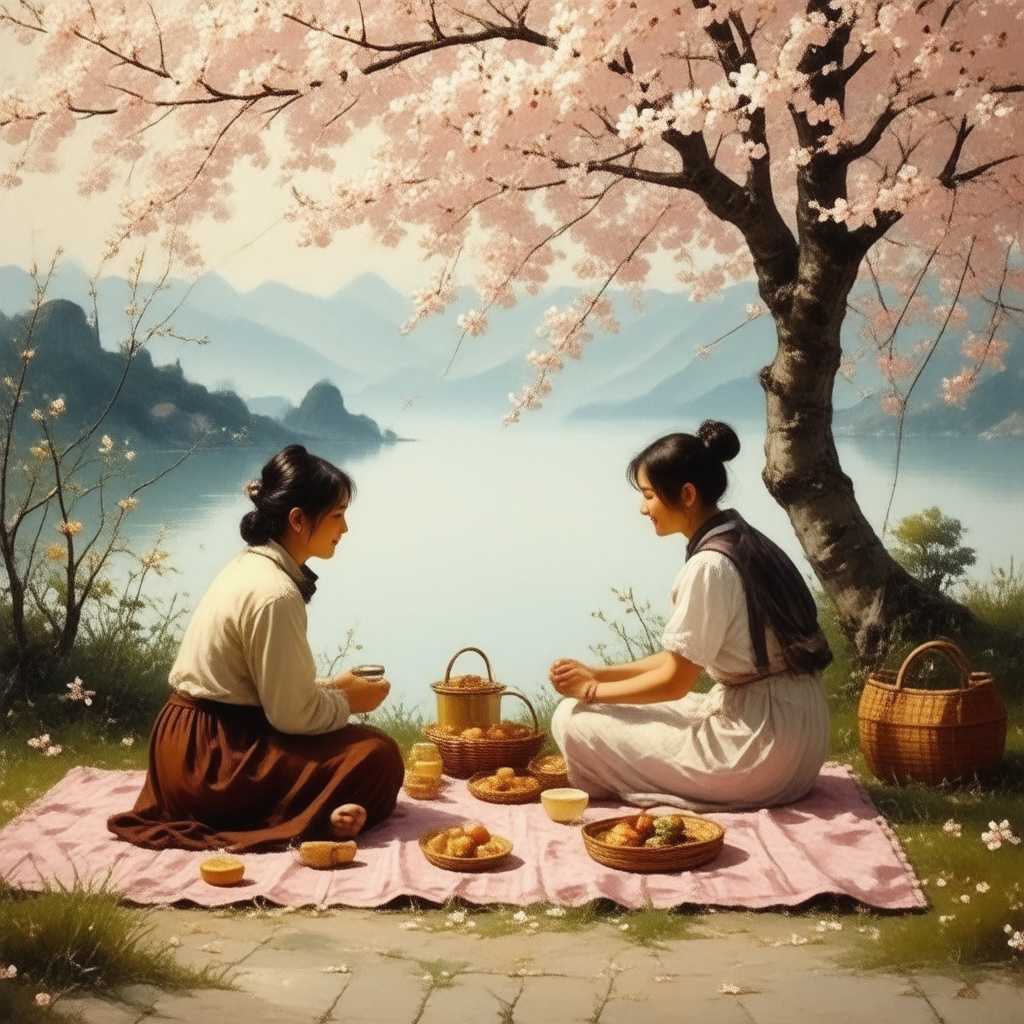} & \includegraphics[width=\imgwidth]{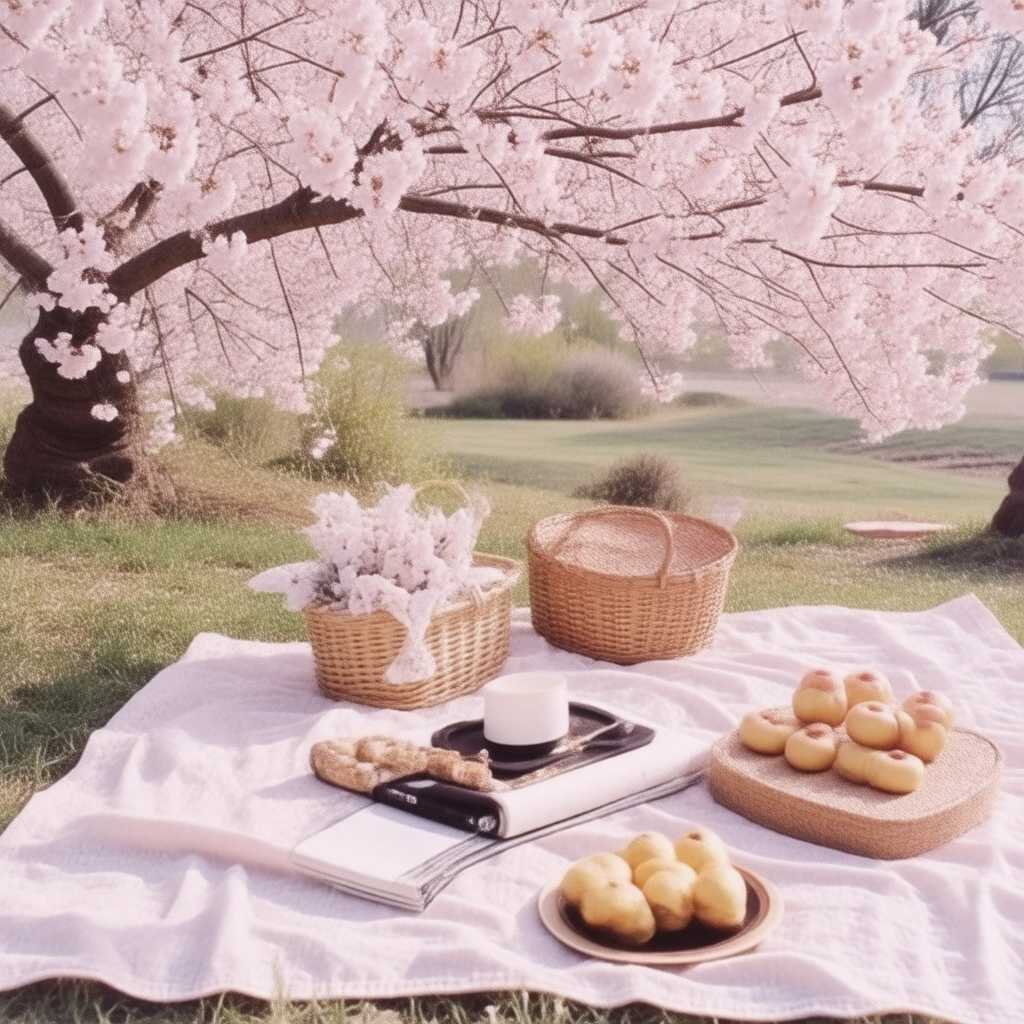} & \includegraphics[width=\imgwidth]{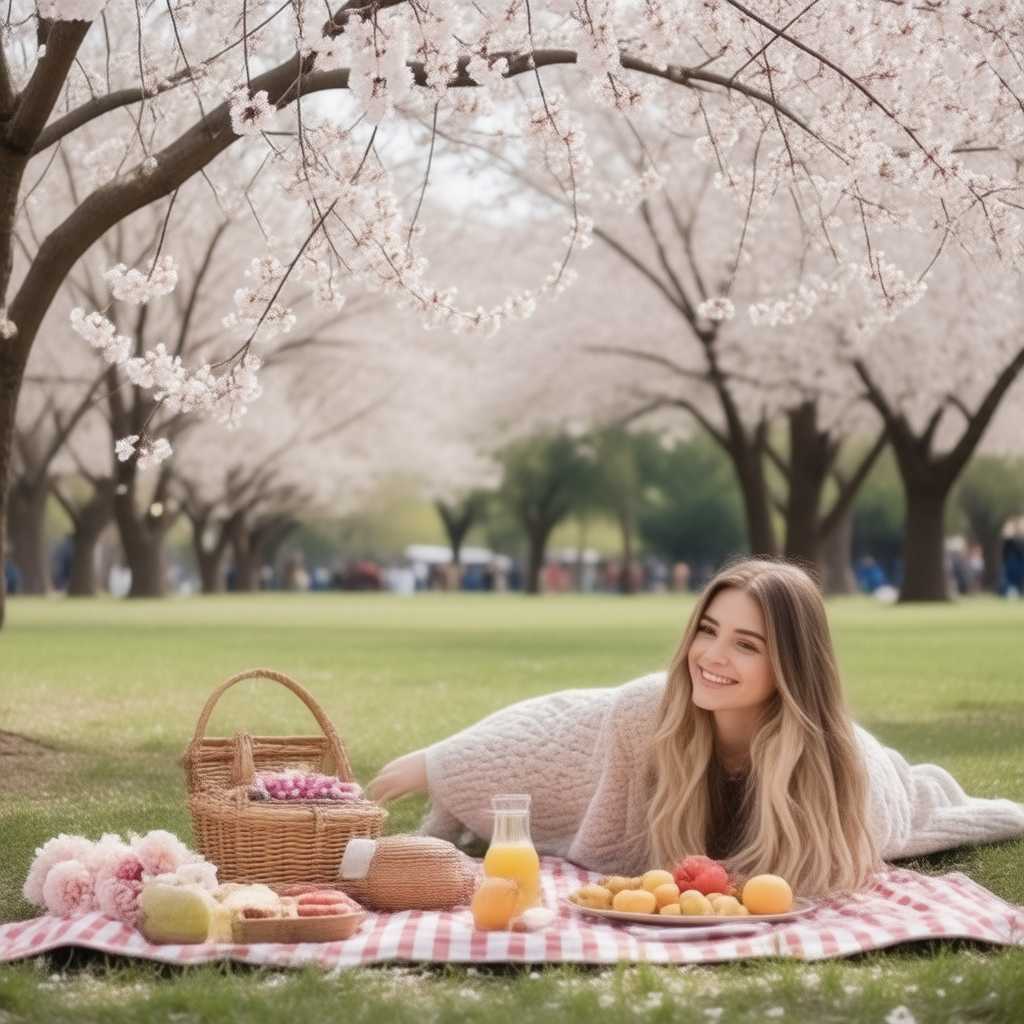} \\
        \multicolumn{5}{c}{\vspace{2pt}\small ``A picnic under cherry blossoms'' \vspace{8pt}} \\

        \turbolabel & \includegraphics[width=\imgwidth]{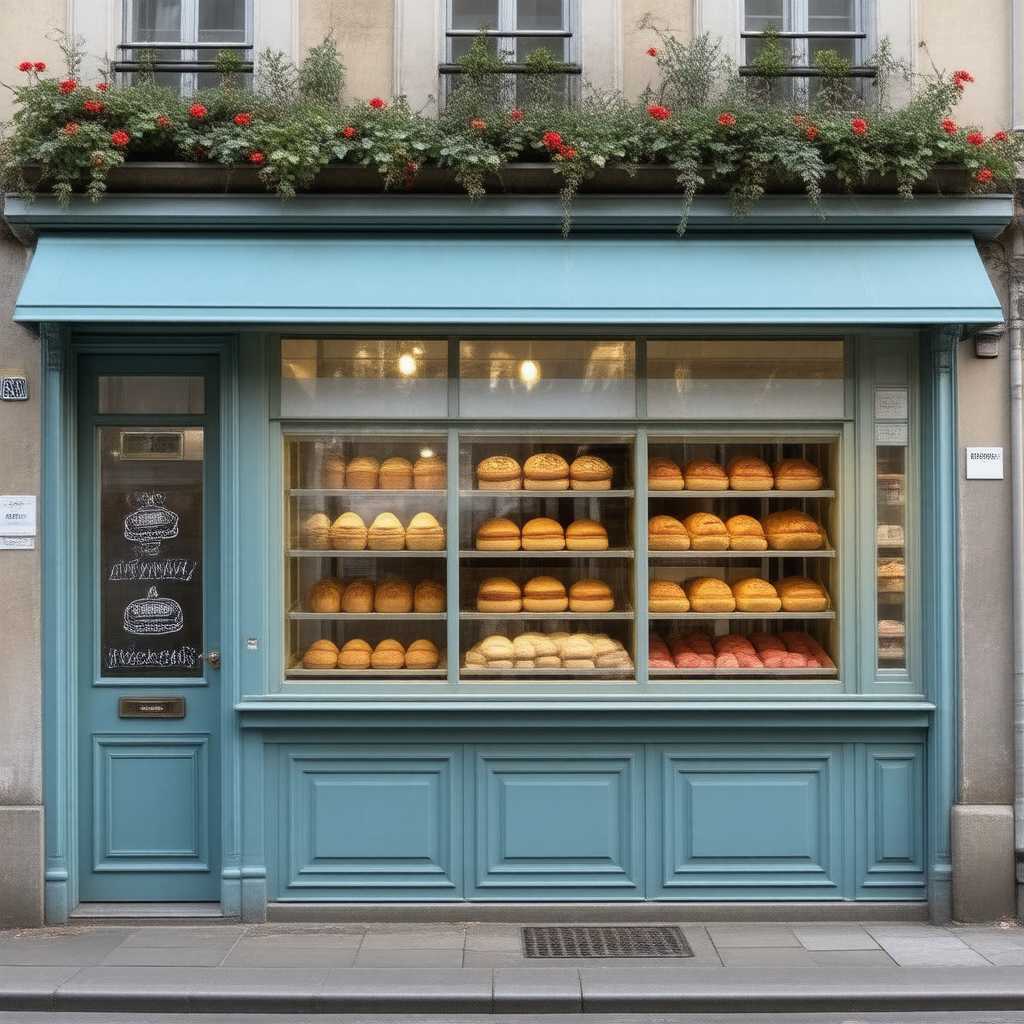} & \includegraphics[width=\imgwidth]{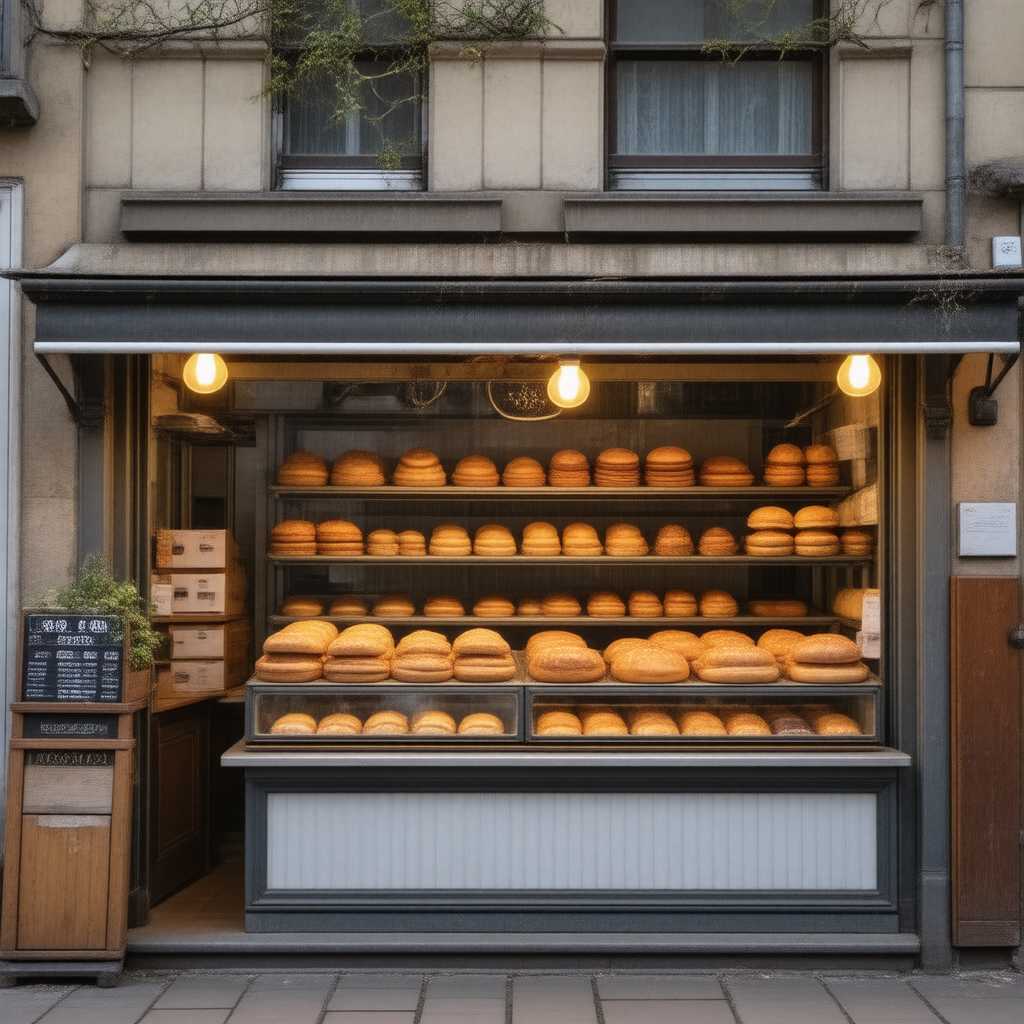} & \includegraphics[width=\imgwidth]{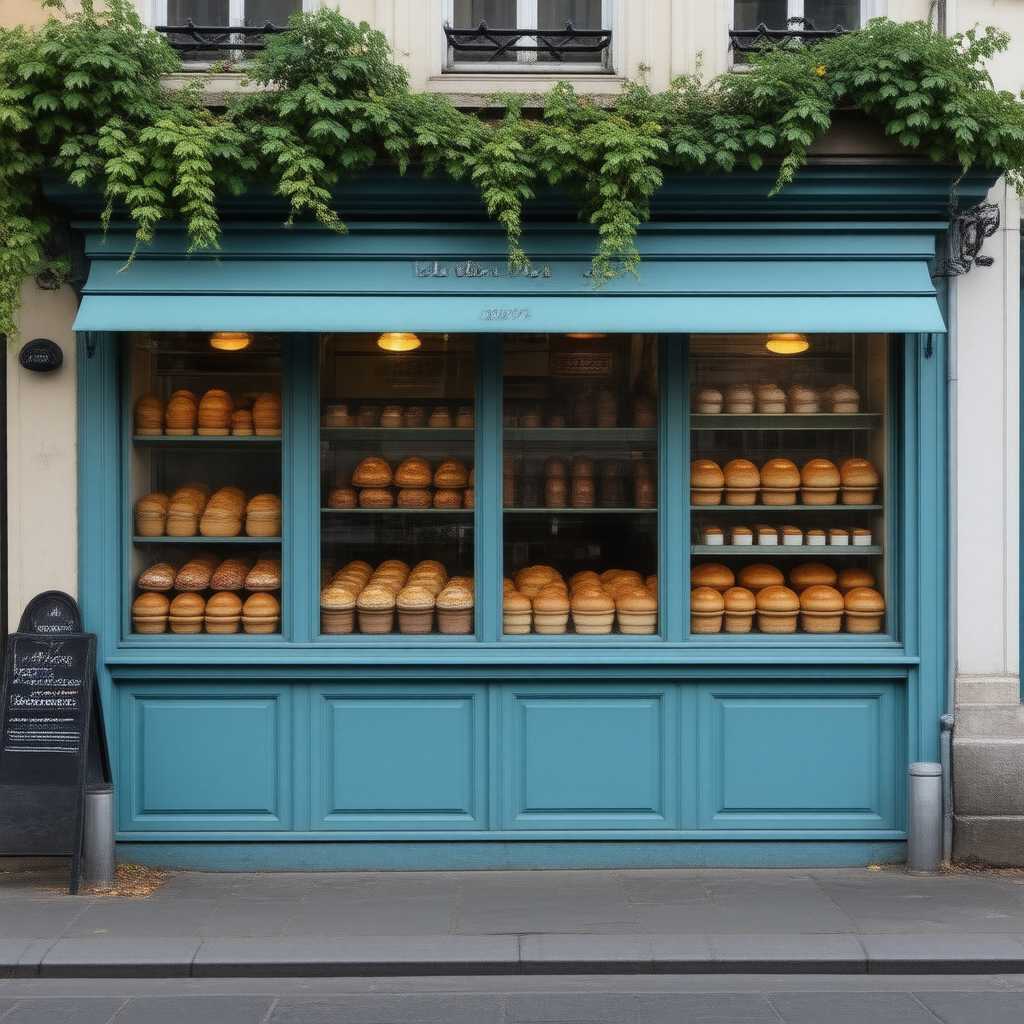} & \includegraphics[width=\imgwidth]{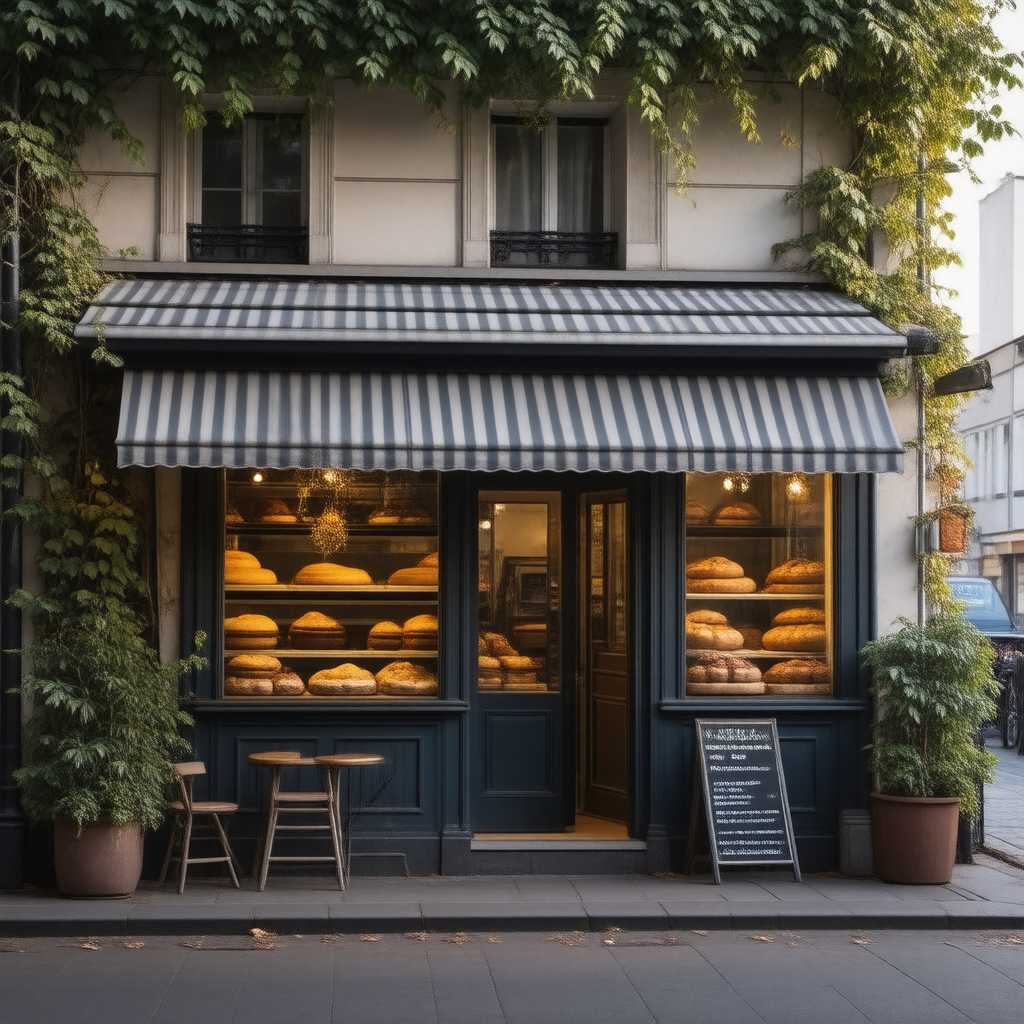} \\[-1pt]
        \ourslabel & \includegraphics[width=\imgwidth]{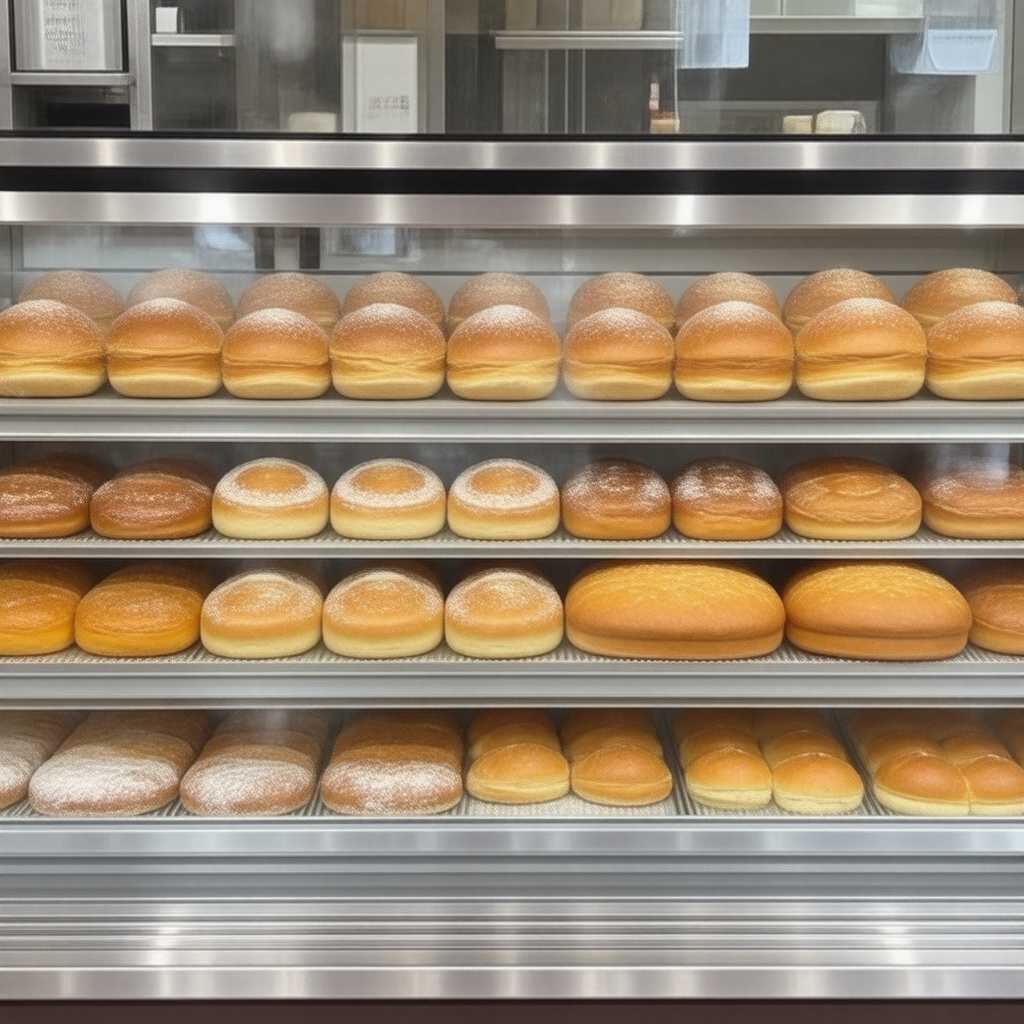} & \includegraphics[width=\imgwidth]{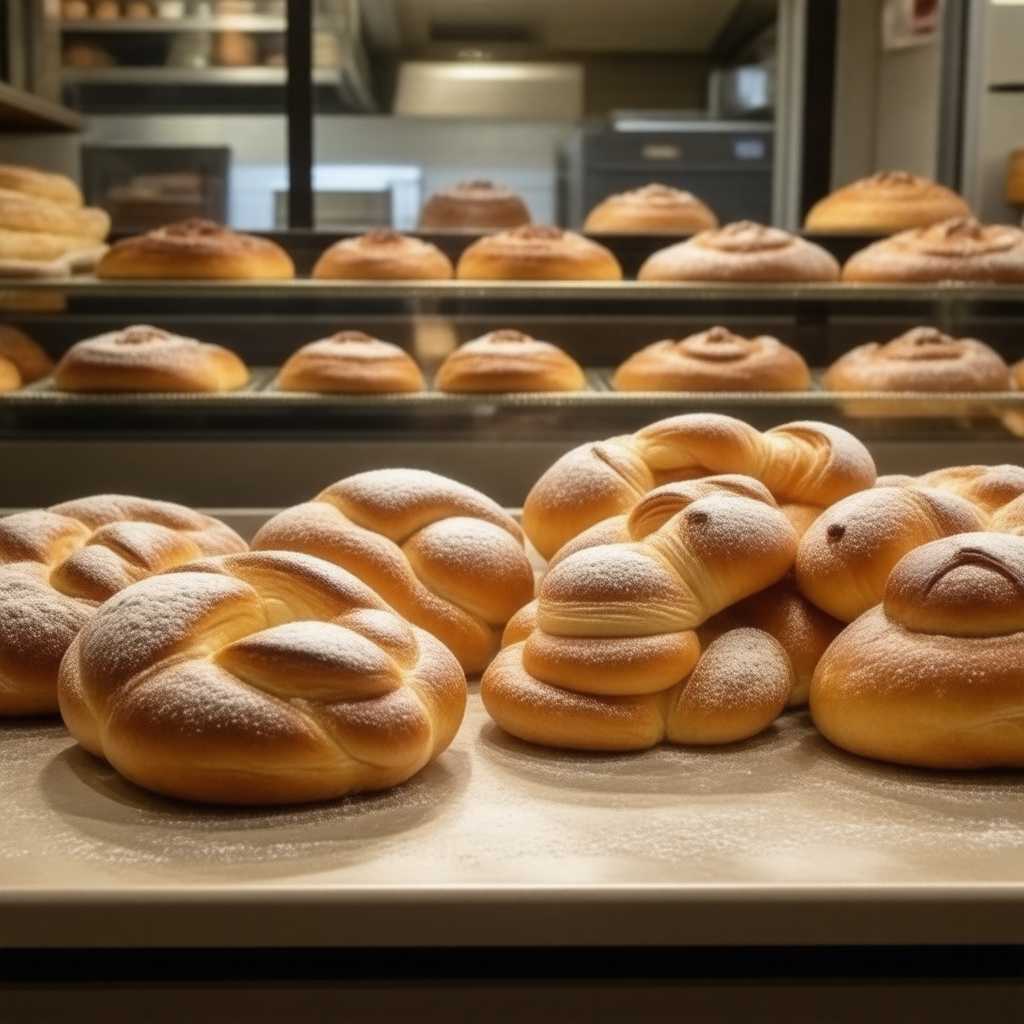} & \includegraphics[width=\imgwidth]{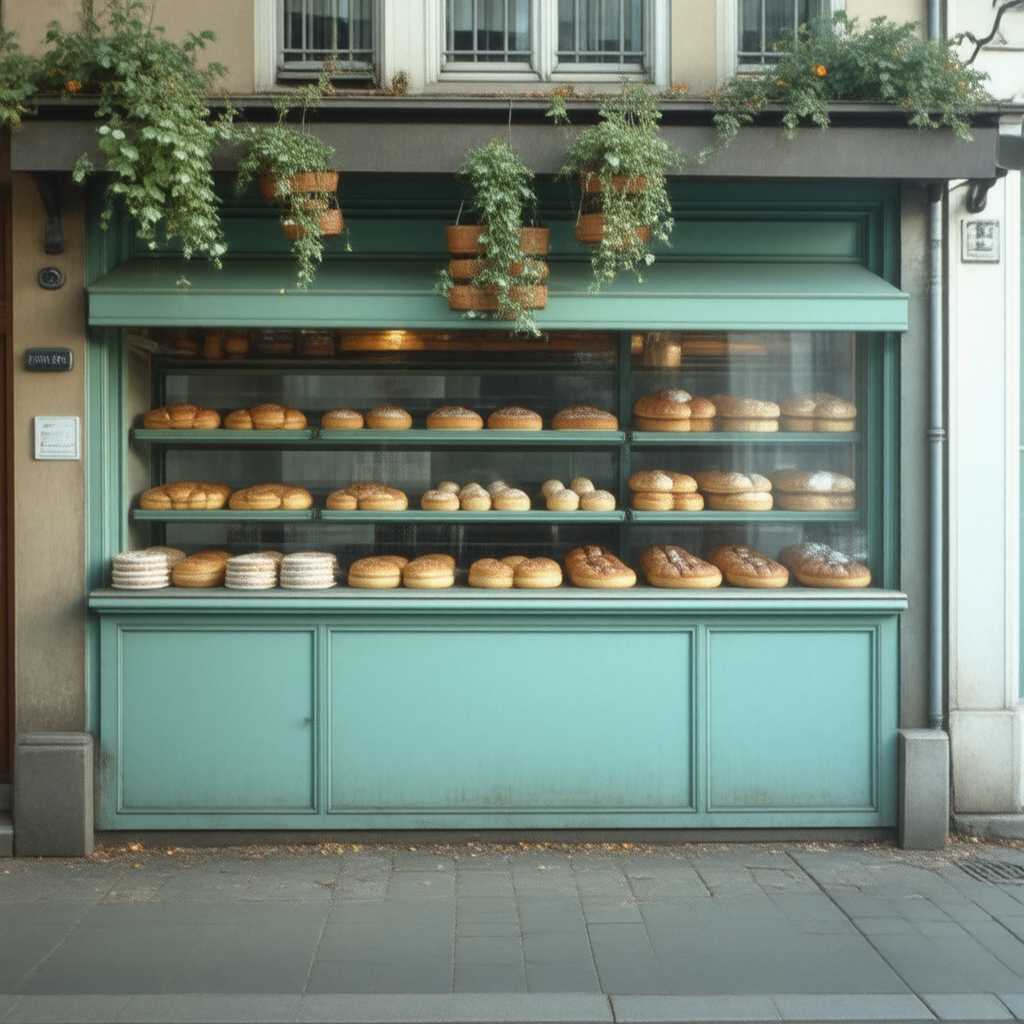} & \includegraphics[width=\imgwidth]{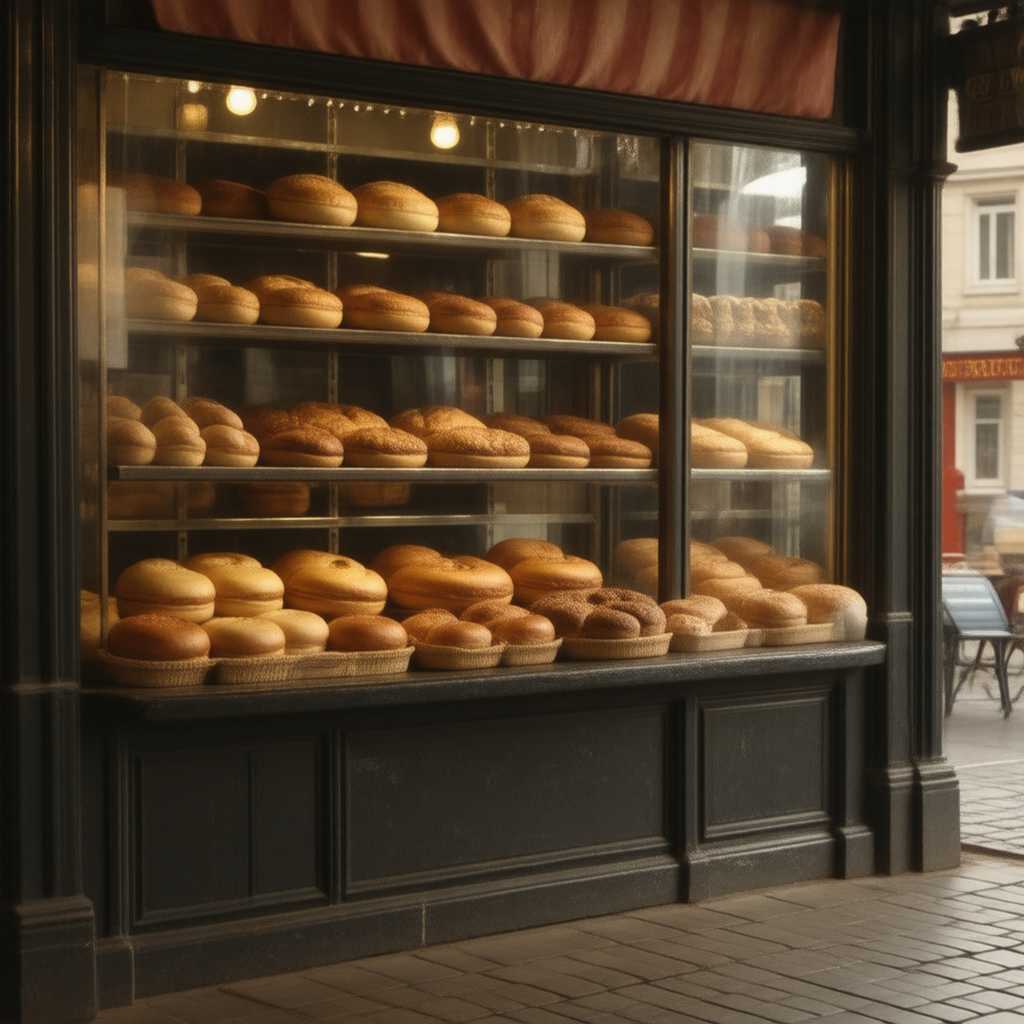} \\
        \multicolumn{5}{c}{\vspace{2pt}\small ``A french bakery at dawn'' \vspace{8pt}} \\

        \turbolabel & \includegraphics[width=\imgwidth]{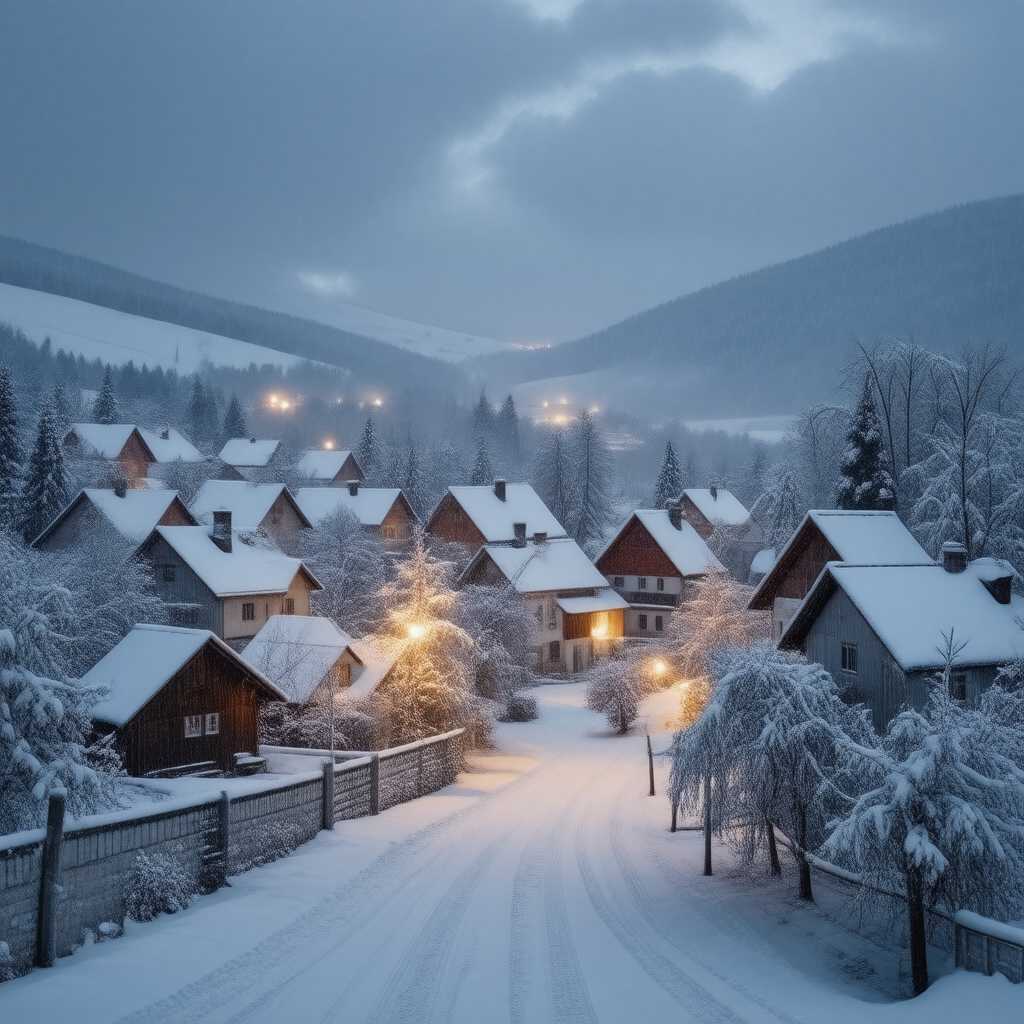} & \includegraphics[width=\imgwidth]{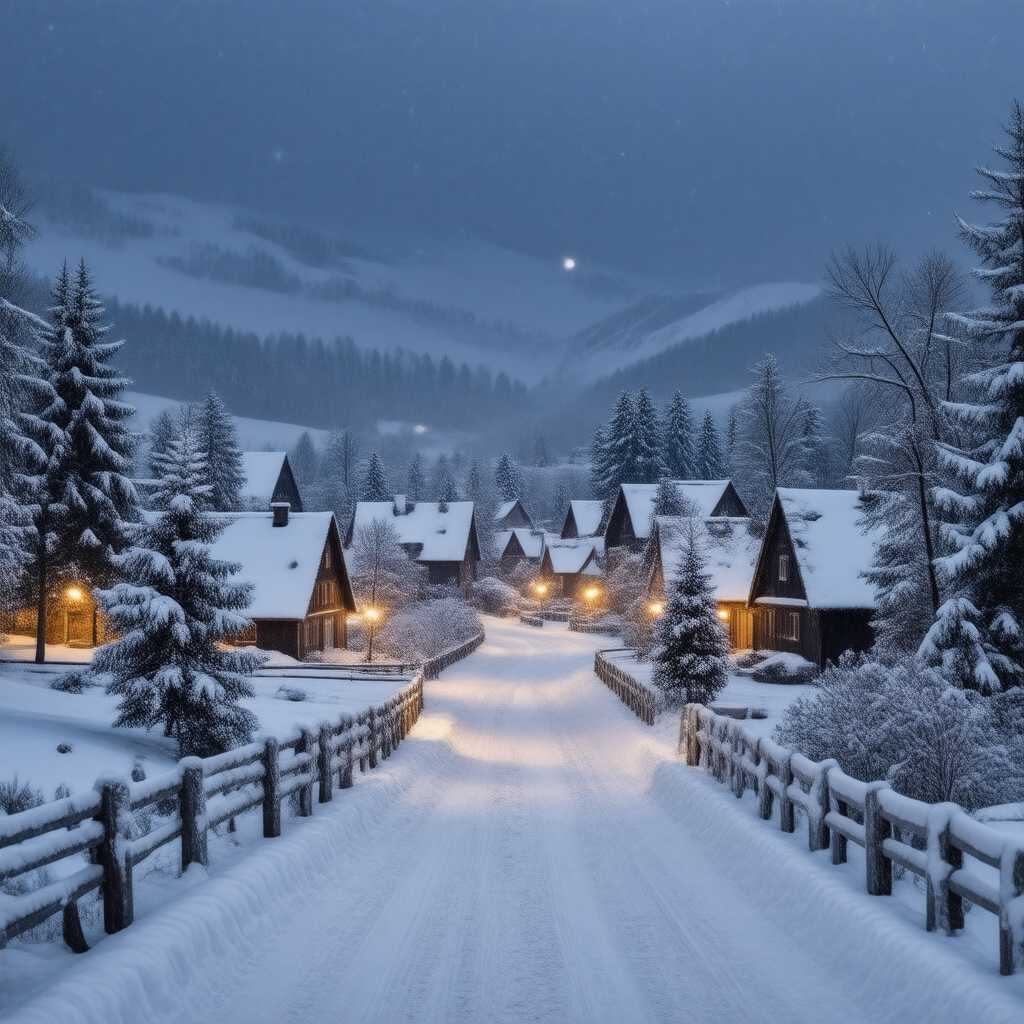} & \includegraphics[width=\imgwidth]{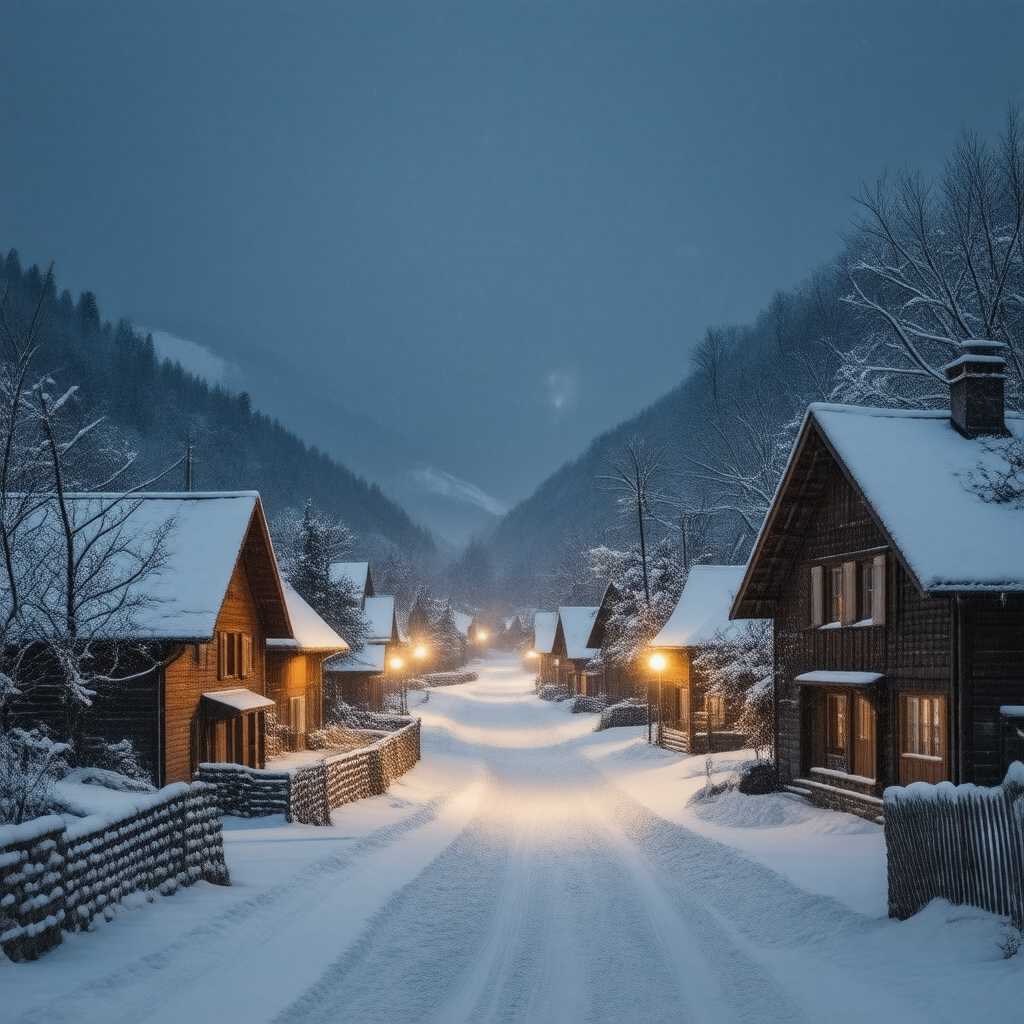} & \includegraphics[width=\imgwidth]{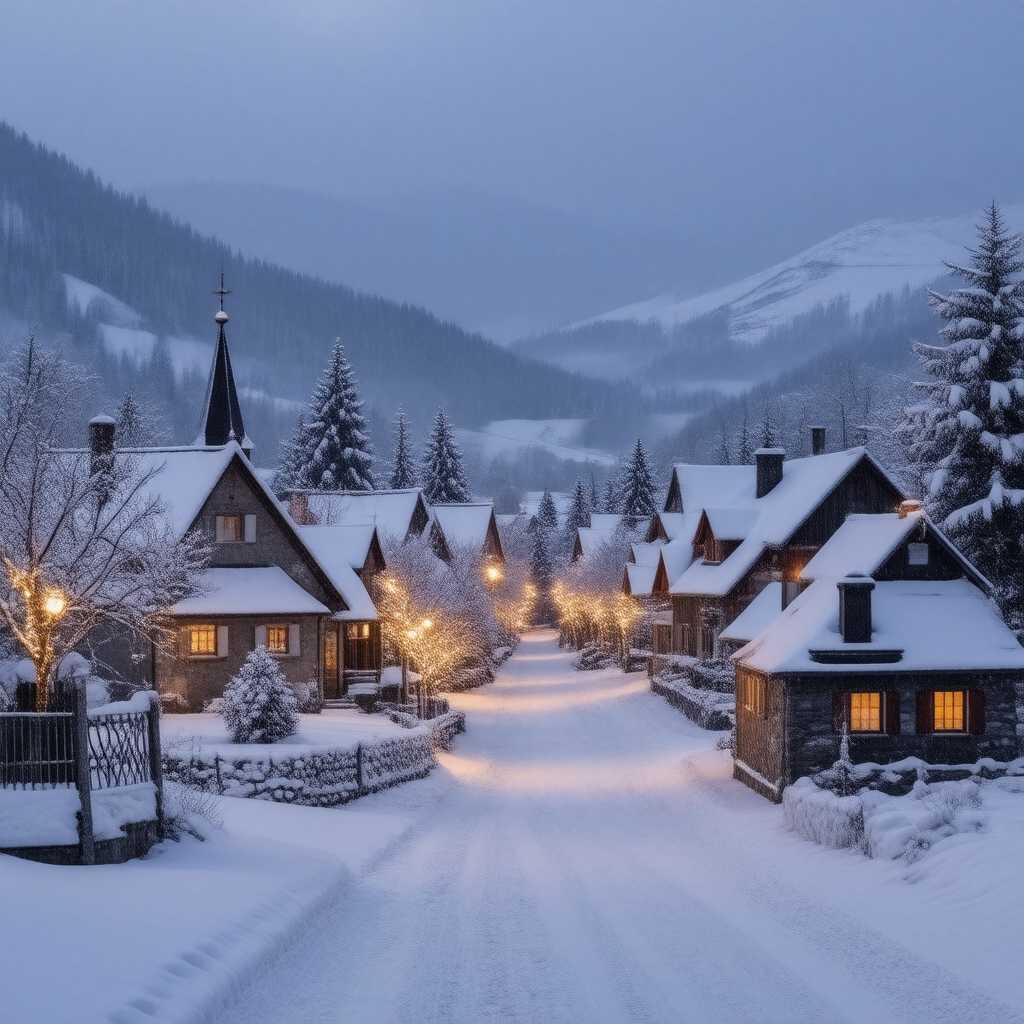} \\[-1pt]
        \ourslabel & \includegraphics[width=\imgwidth]{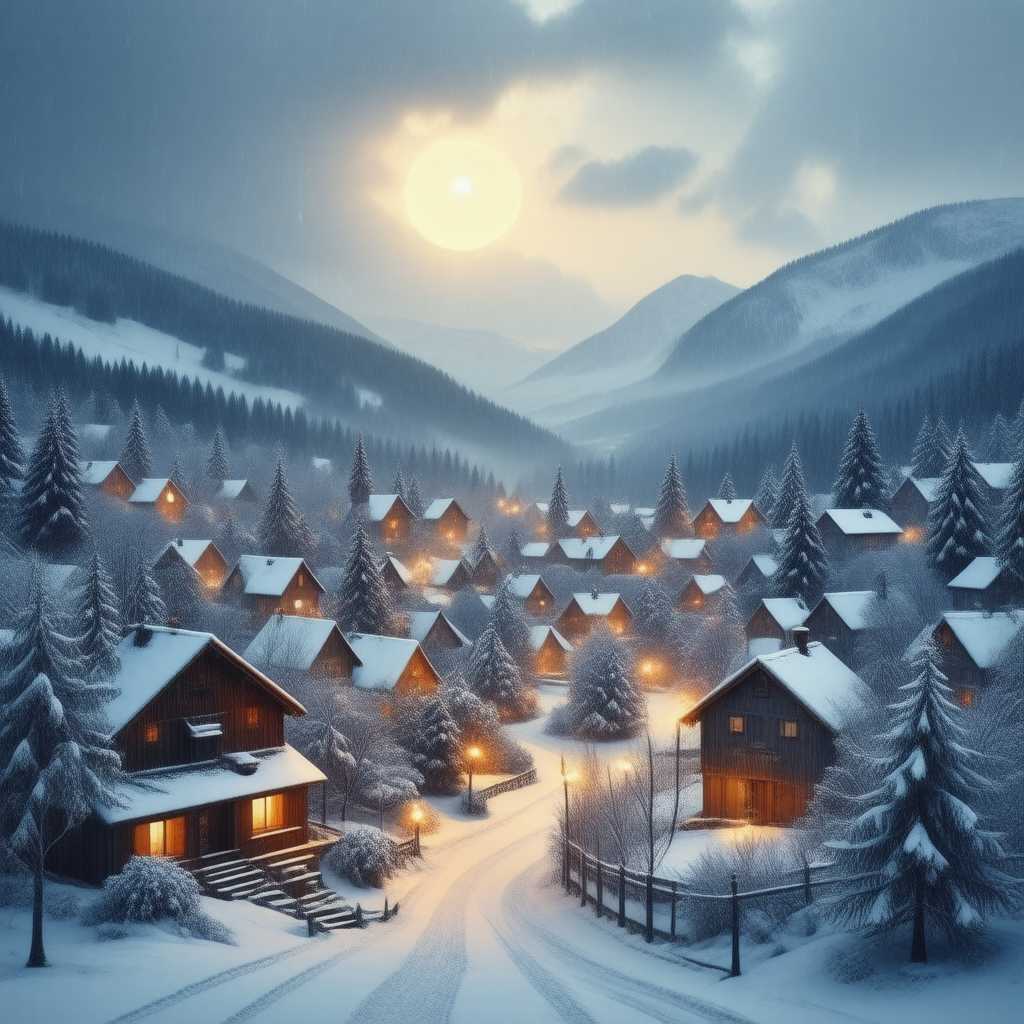} & \includegraphics[width=\imgwidth]{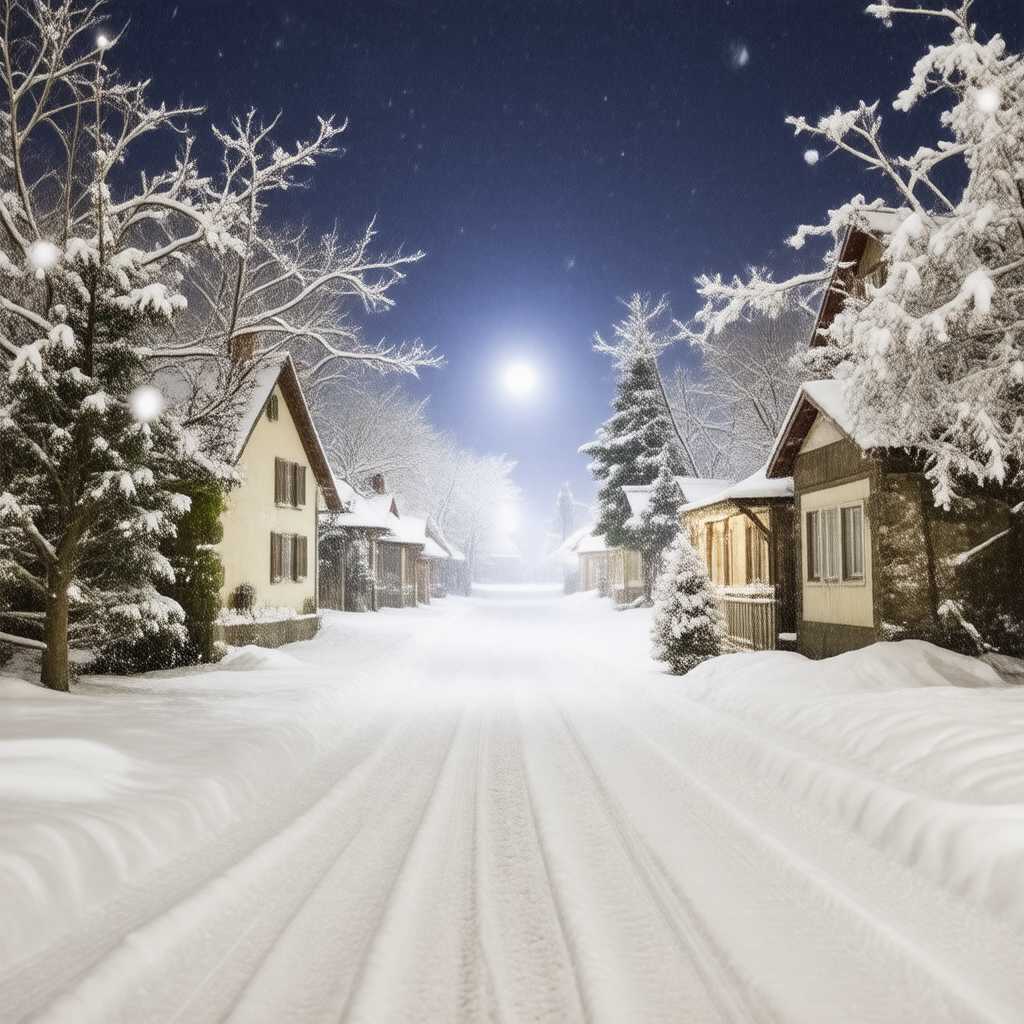} & \includegraphics[width=\imgwidth]{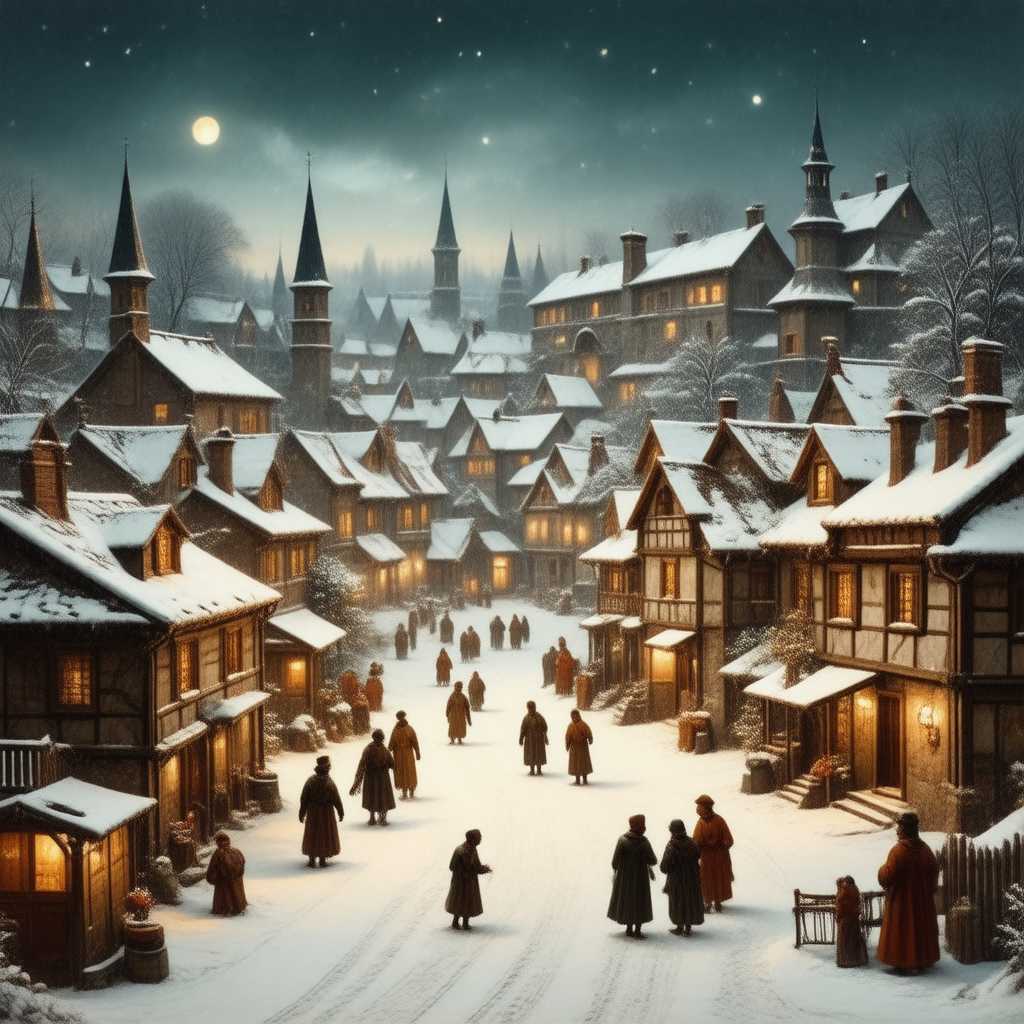} & \includegraphics[width=\imgwidth]{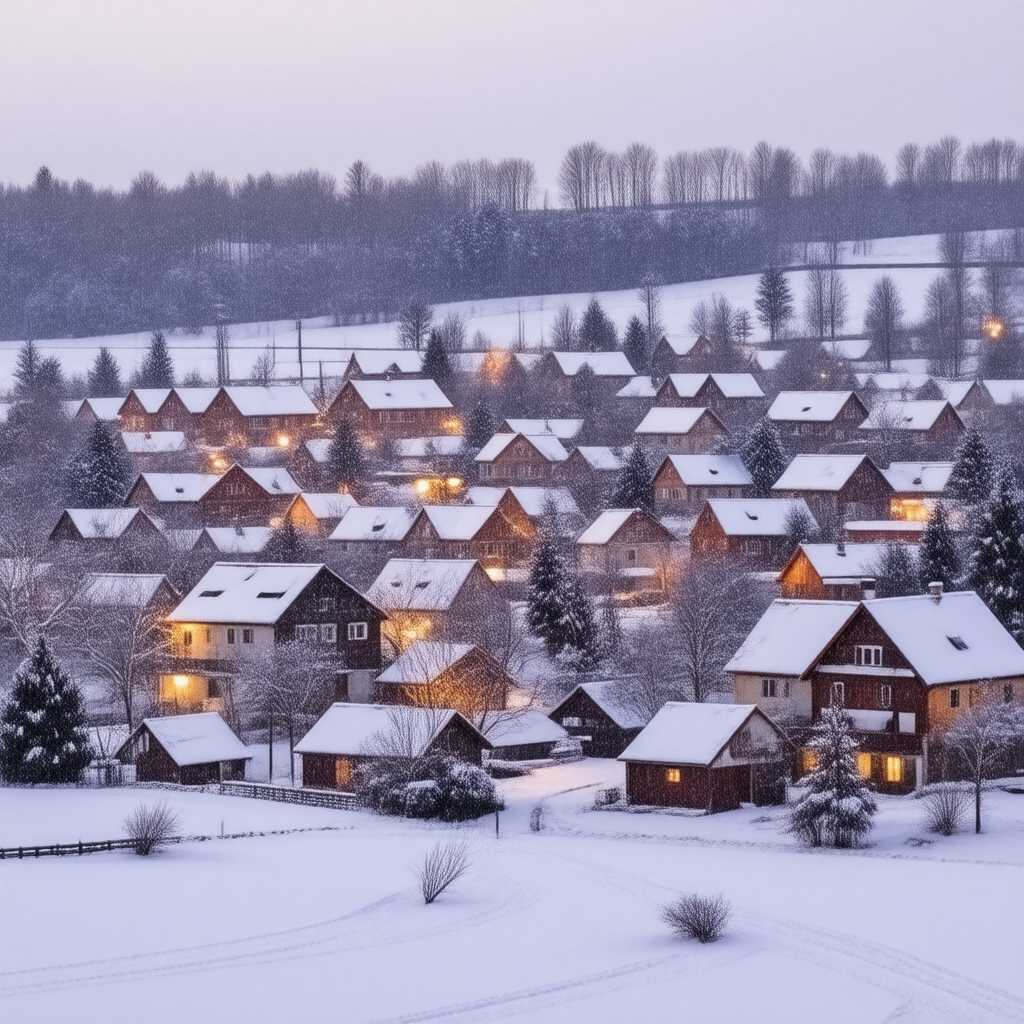} \\
        \multicolumn{5}{c}{\vspace{2pt}\small ``A snowy village at night'' \vspace{8pt}} \\

    \end{tabular}
    \caption{\textbf{Qualitative results on SD3.5-Turbo.}}
    \label{fig:turbo}
\end{figure}

\begin{figure*}
    \centering
    \setlength{\tabcolsep}{0.5pt} \renewcommand{\arraystretch}{0.5} \newcommand{\imgwidth}{0.12\textwidth}
    \newcommand{\vertlabel}[1]{\raisebox{2.5em}{\rotatebox{90}{\scriptsize\textbf{#1}}}}

    \begin{tabular}{c c c c c c c c c}
        
        \vertlabel{Flux} & \includegraphics[width=\imgwidth]{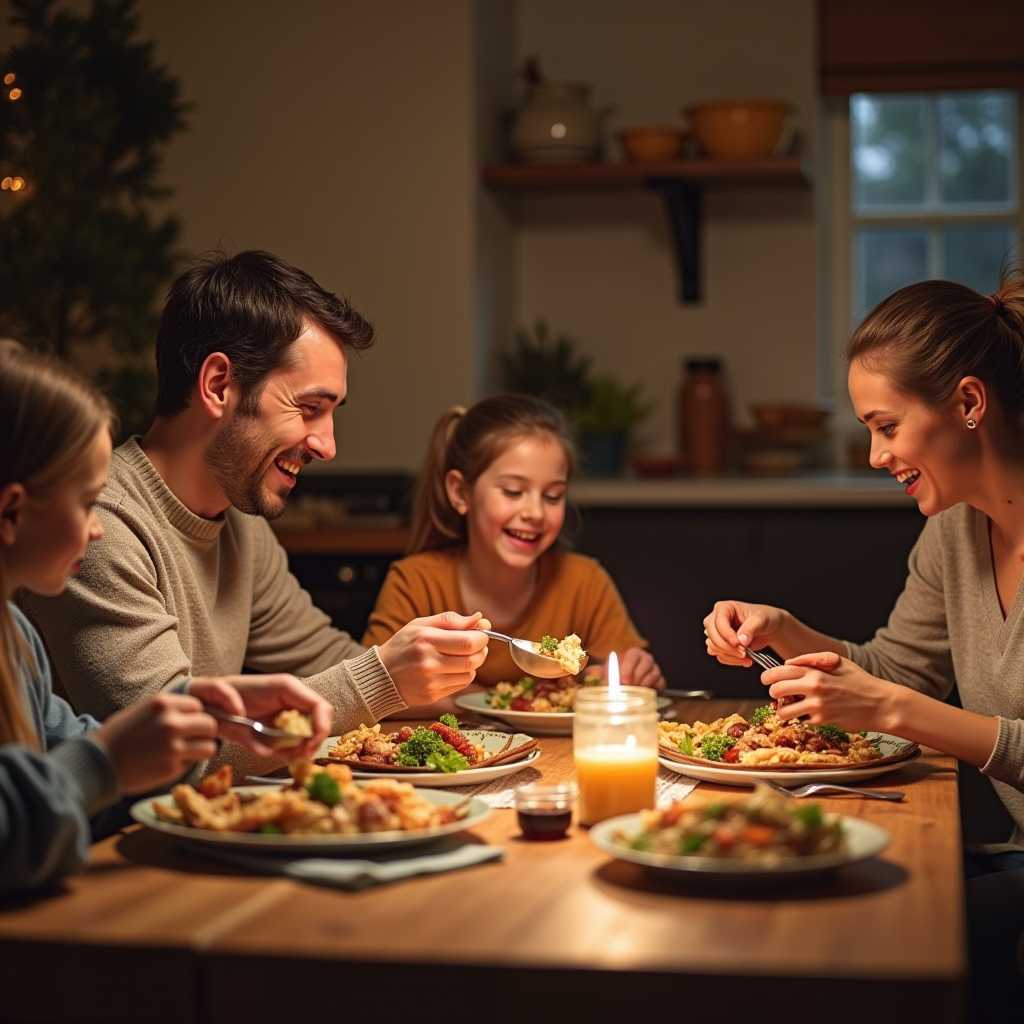} & \includegraphics[width=\imgwidth]{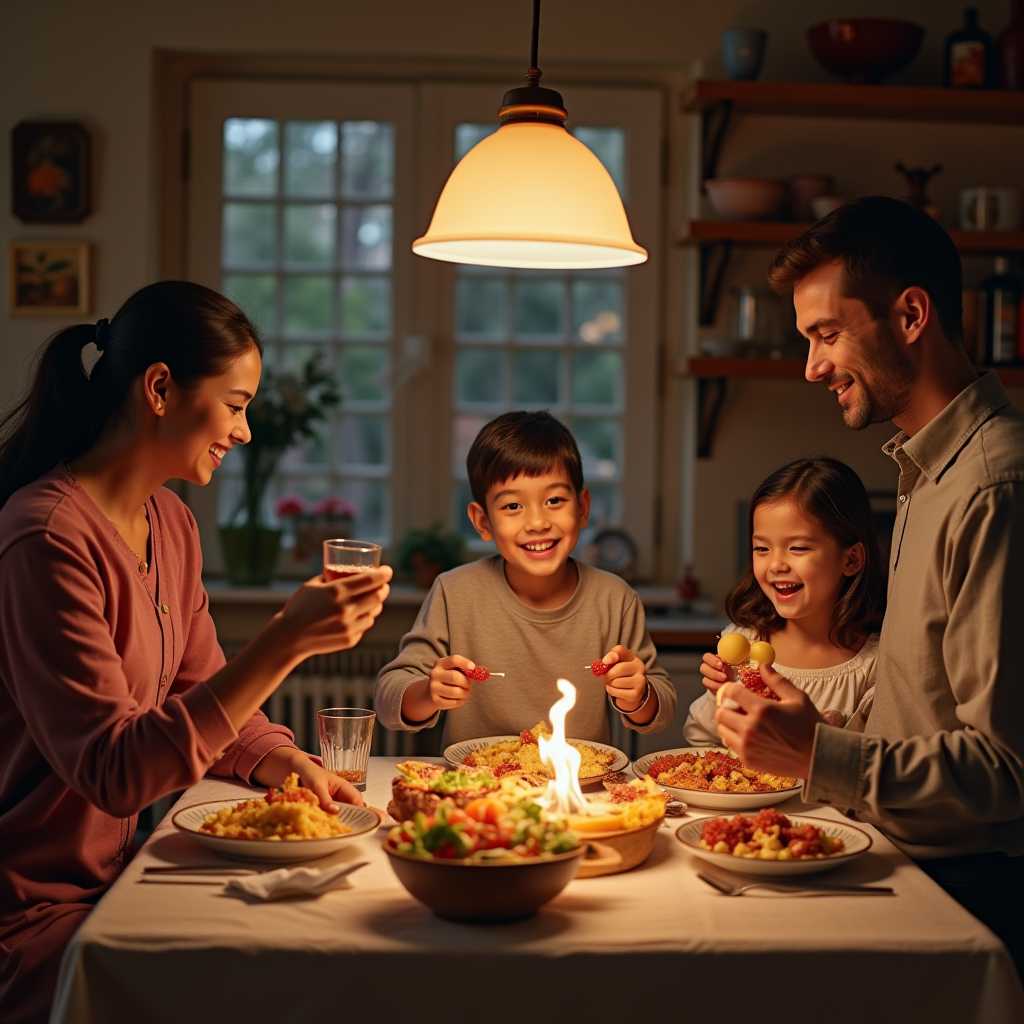} & \includegraphics[width=\imgwidth]{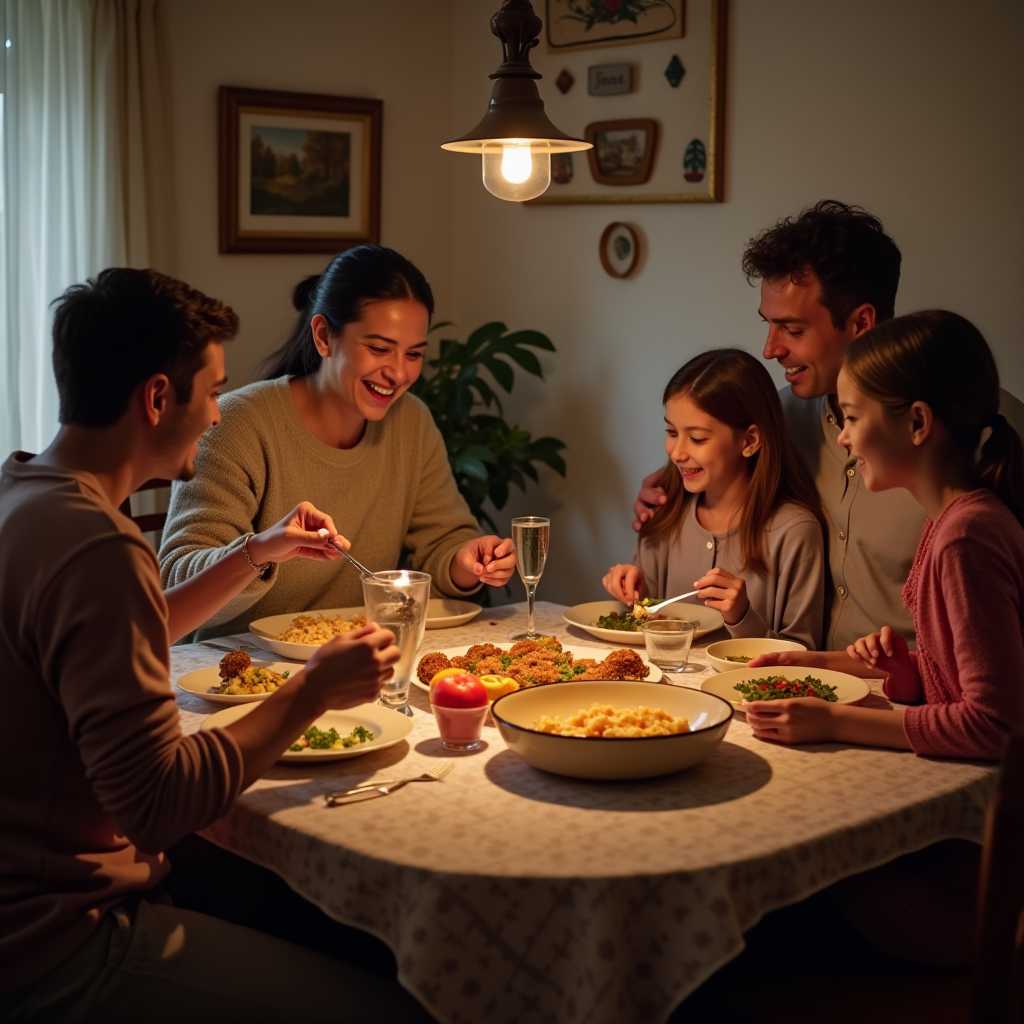} & \includegraphics[width=\imgwidth]{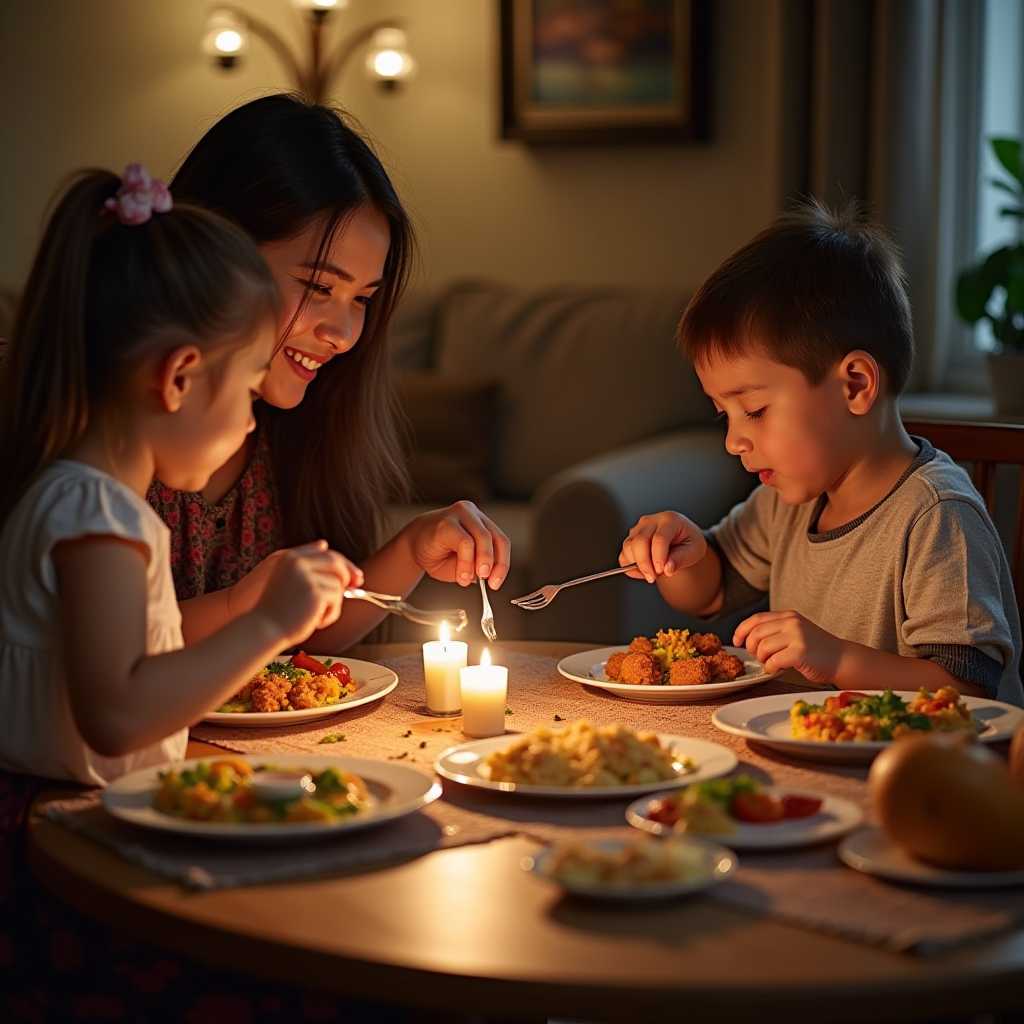} & \includegraphics[width=\imgwidth]{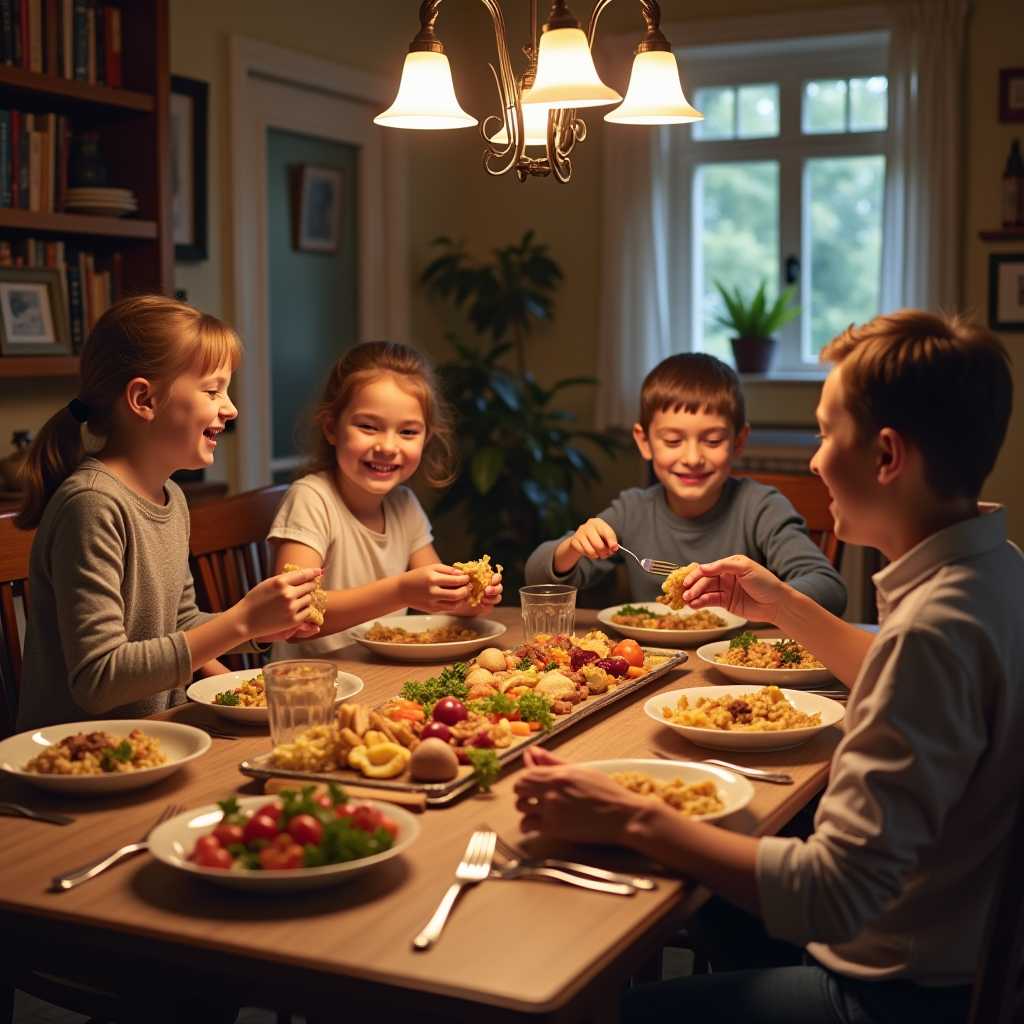} & \includegraphics[width=\imgwidth]{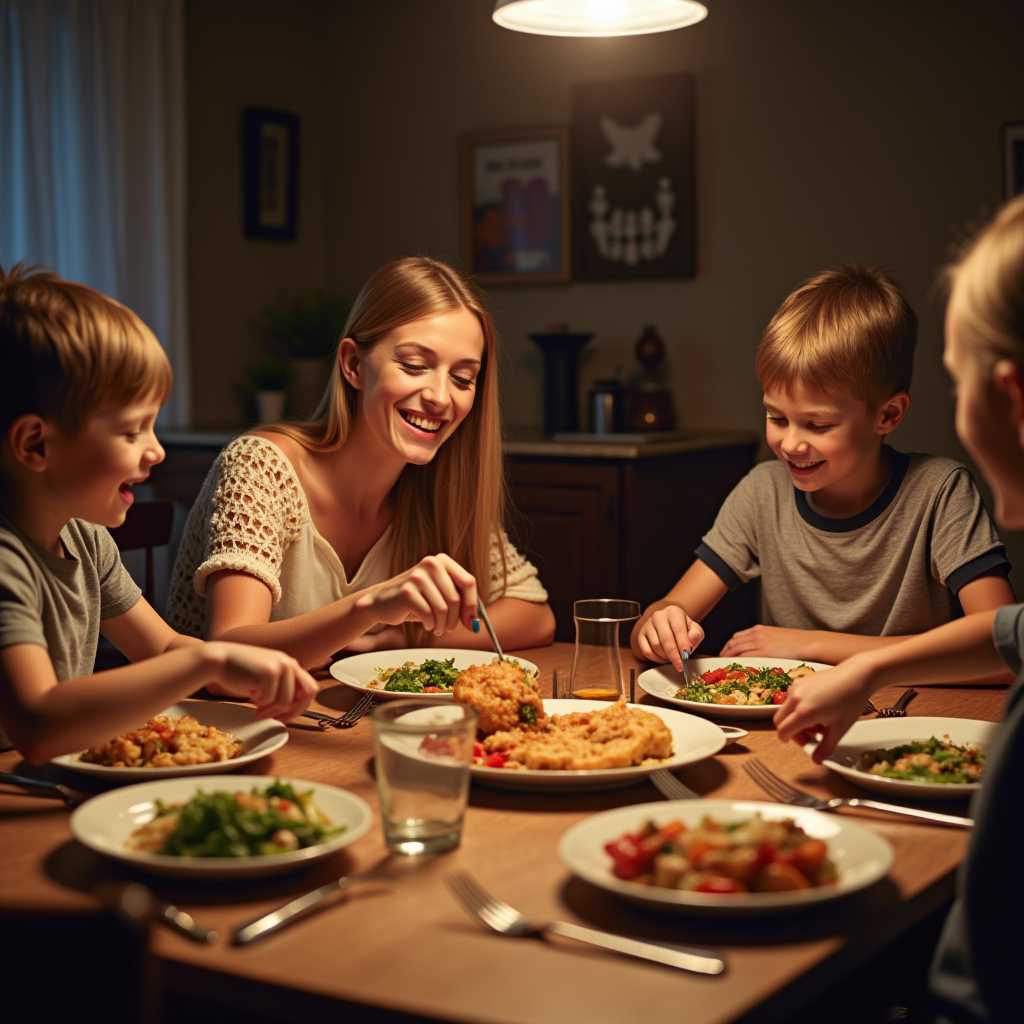} & \includegraphics[width=\imgwidth]{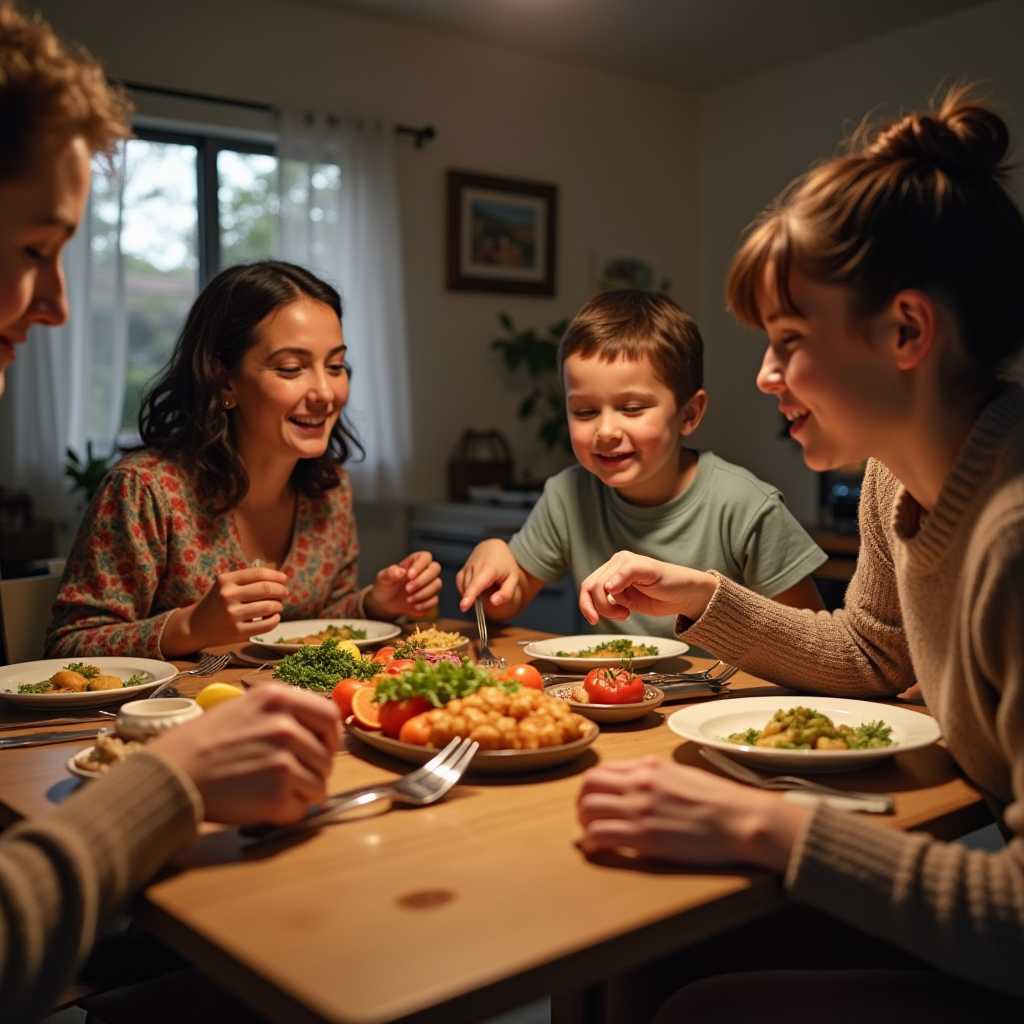} & \includegraphics[width=\imgwidth]{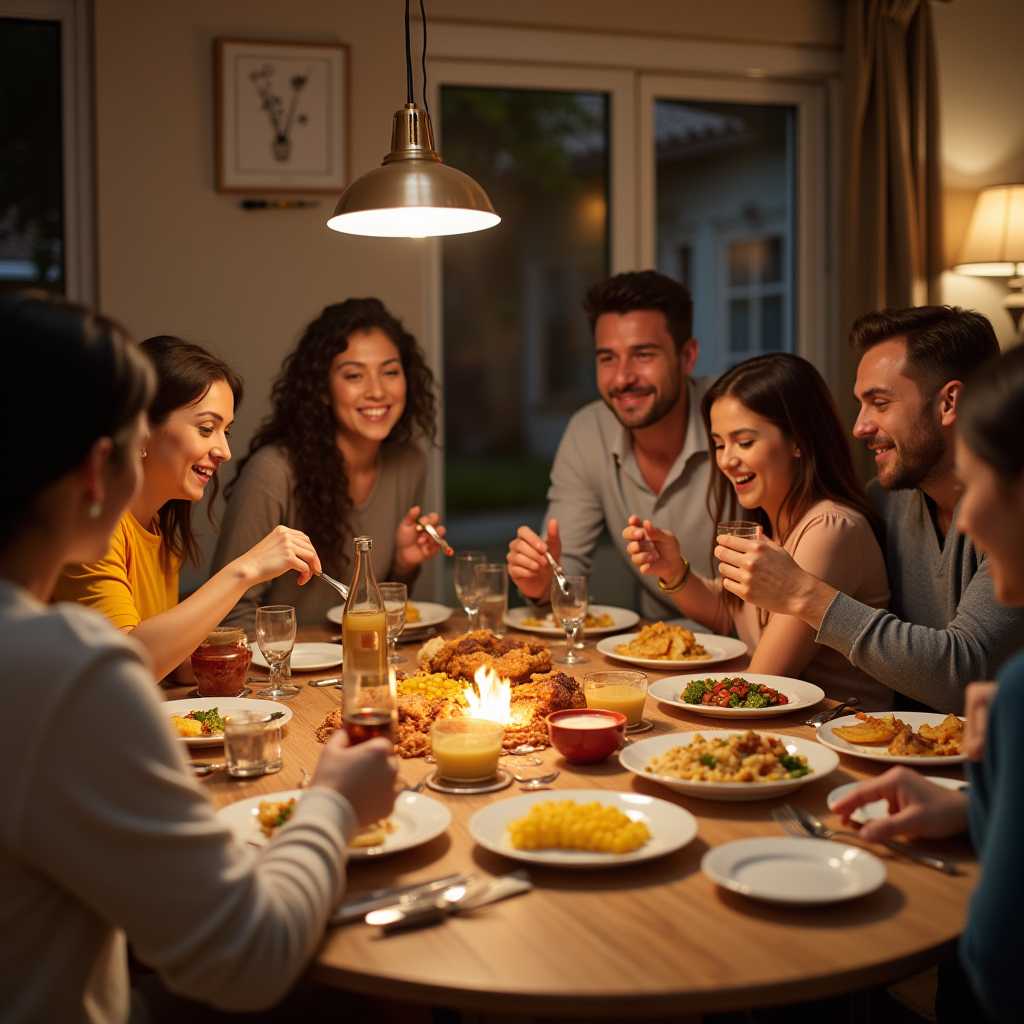} \\[-1pt]
        \vertlabel{Ours} & \includegraphics[width=\imgwidth]{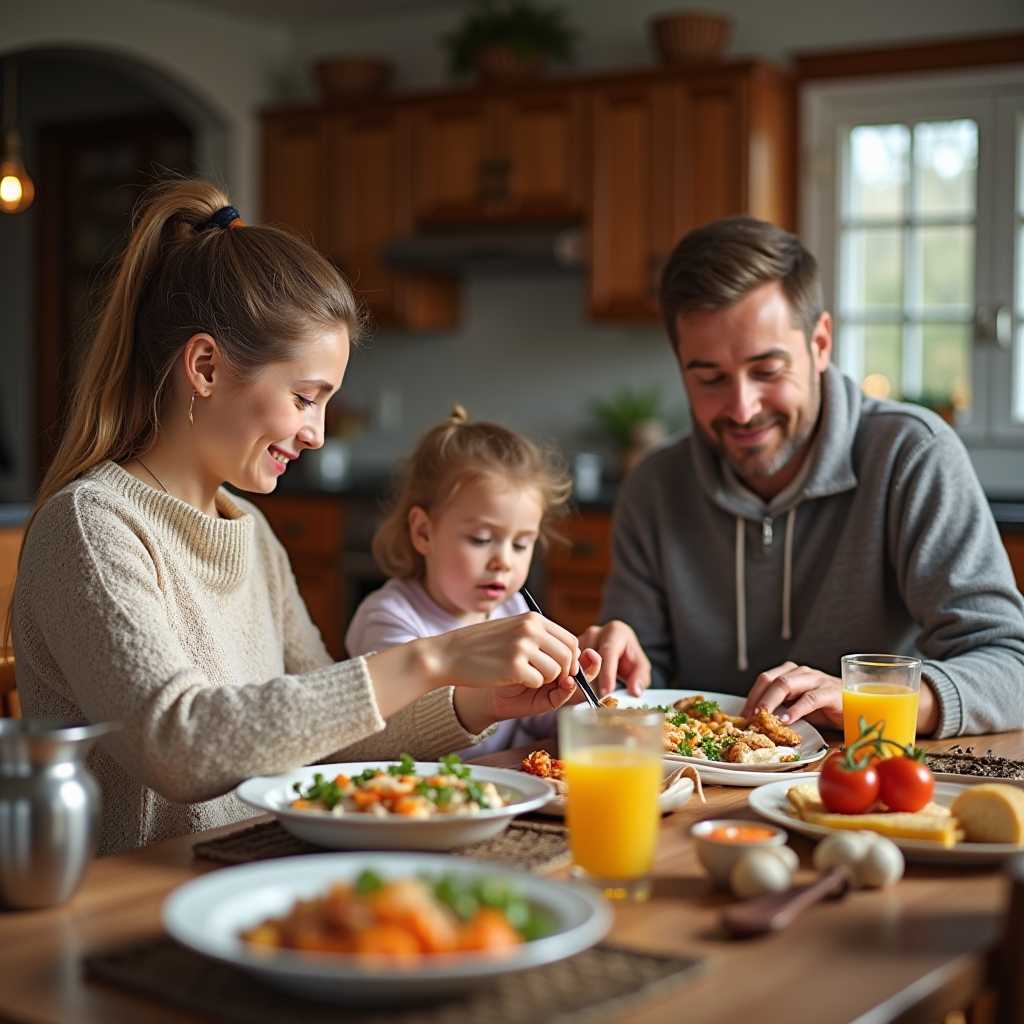} & \includegraphics[width=\imgwidth]{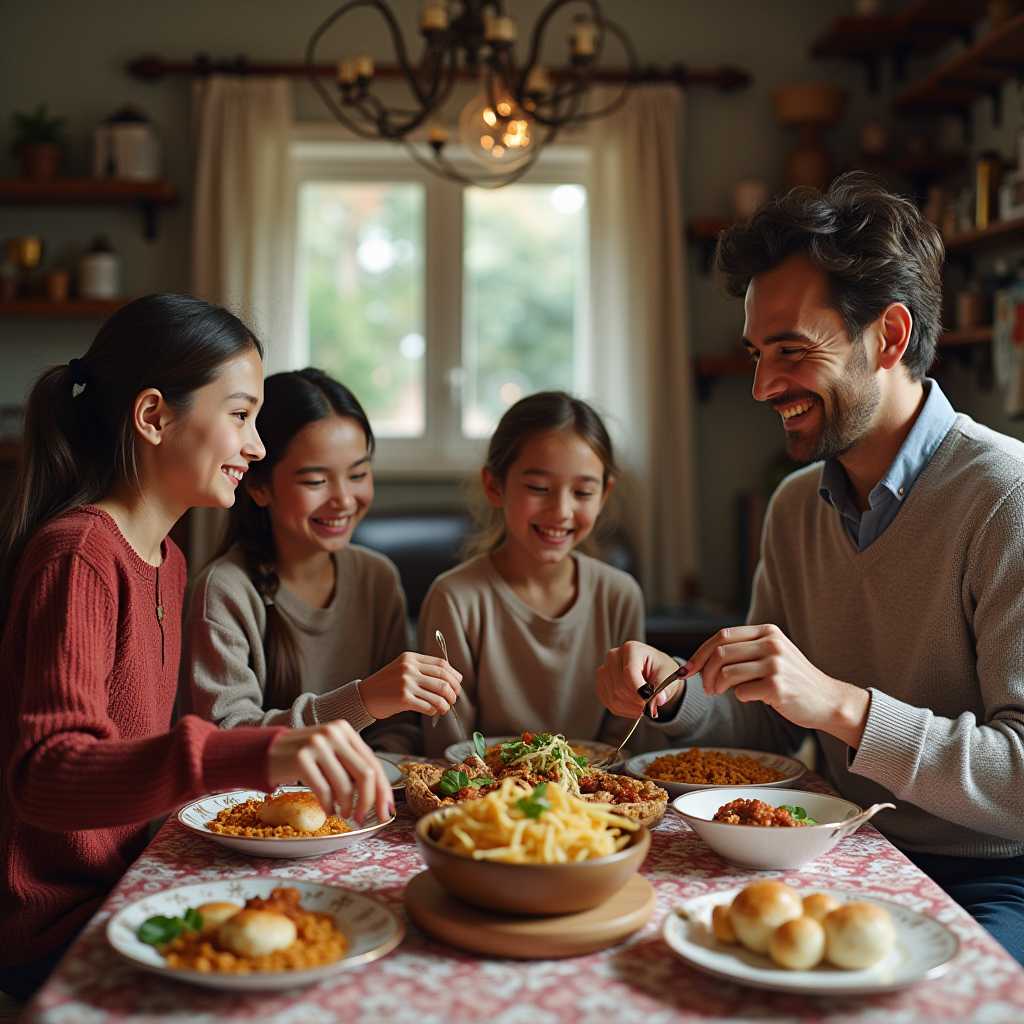} & \includegraphics[width=\imgwidth]{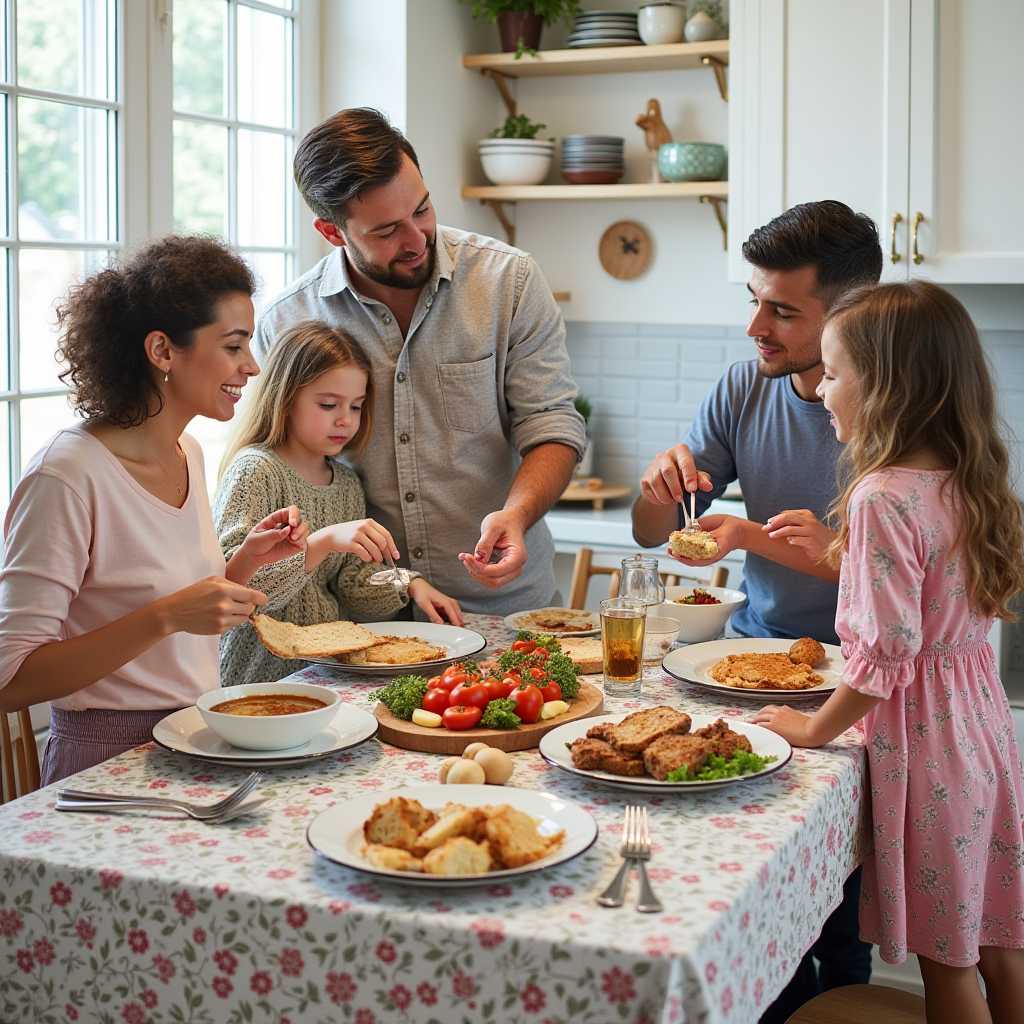} & \includegraphics[width=\imgwidth]{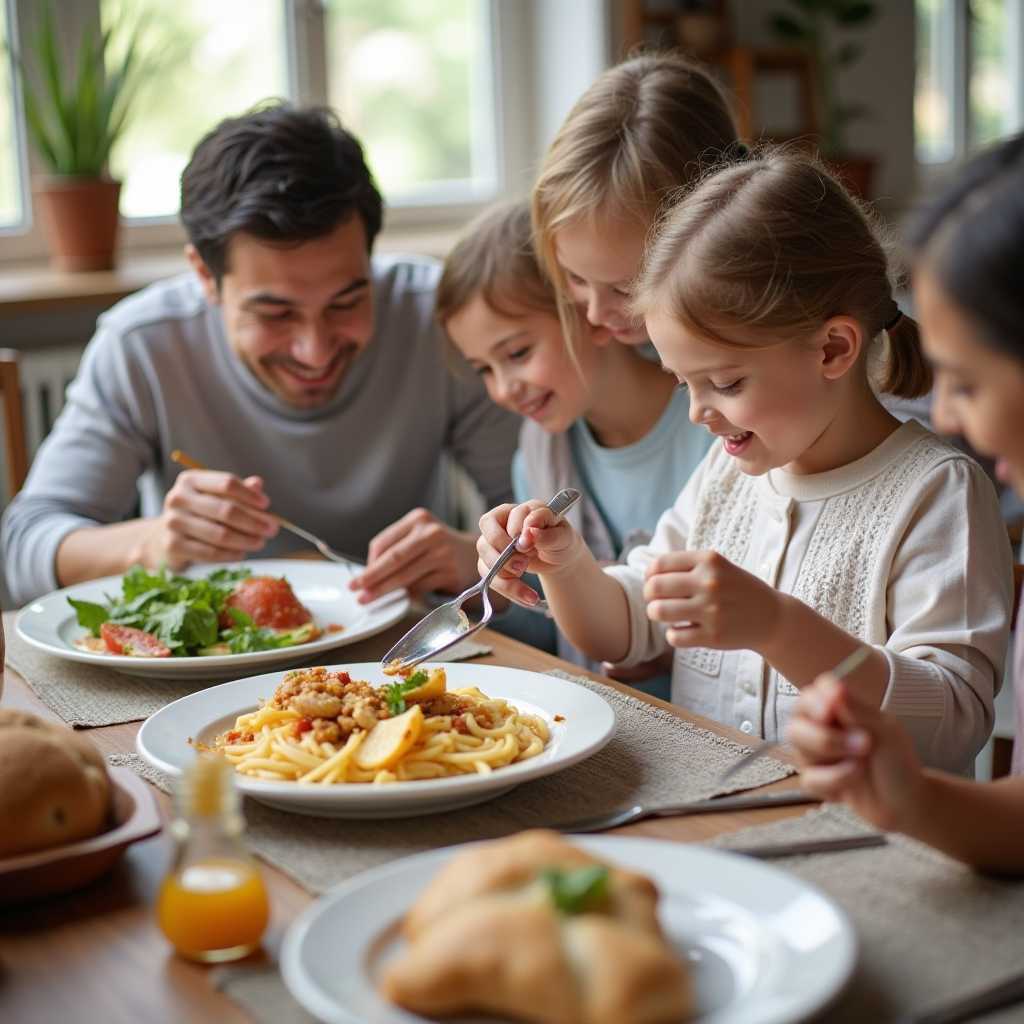} & \includegraphics[width=\imgwidth]{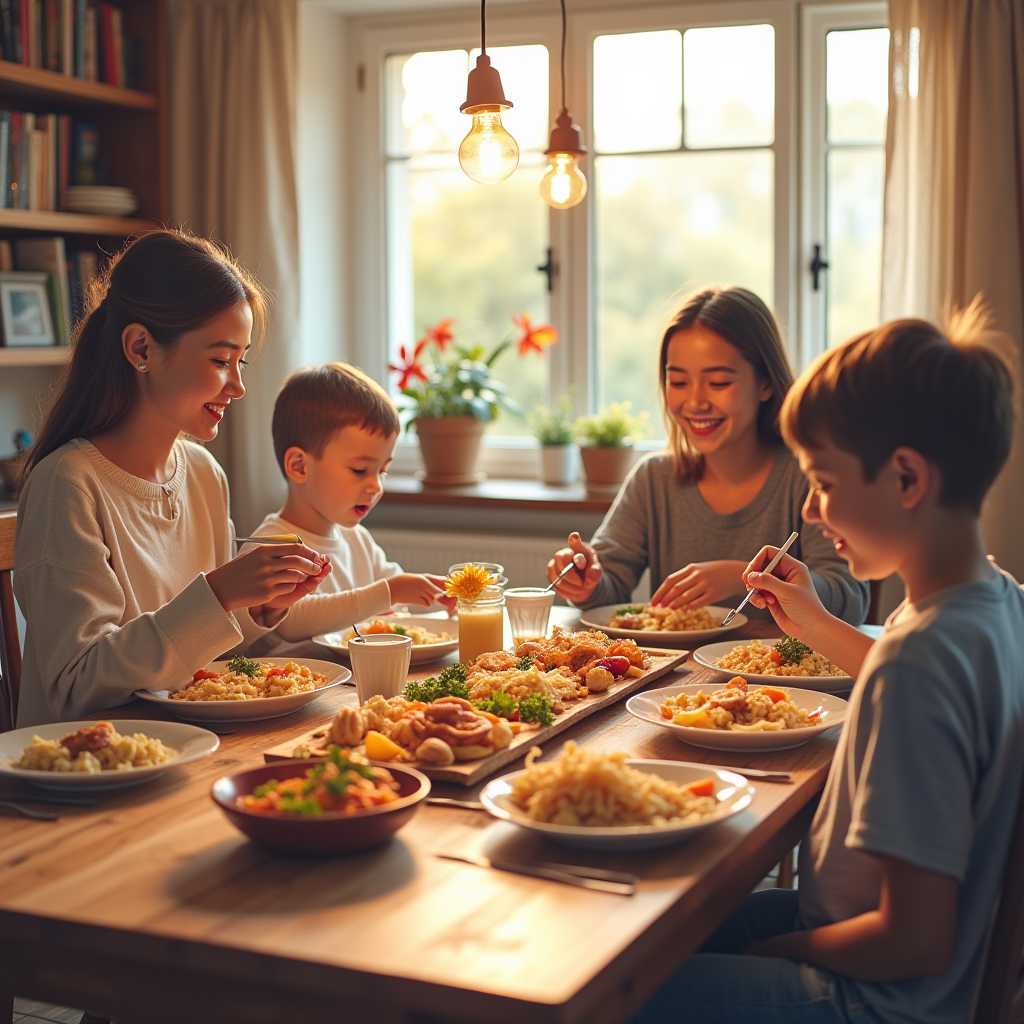} & \includegraphics[width=\imgwidth]{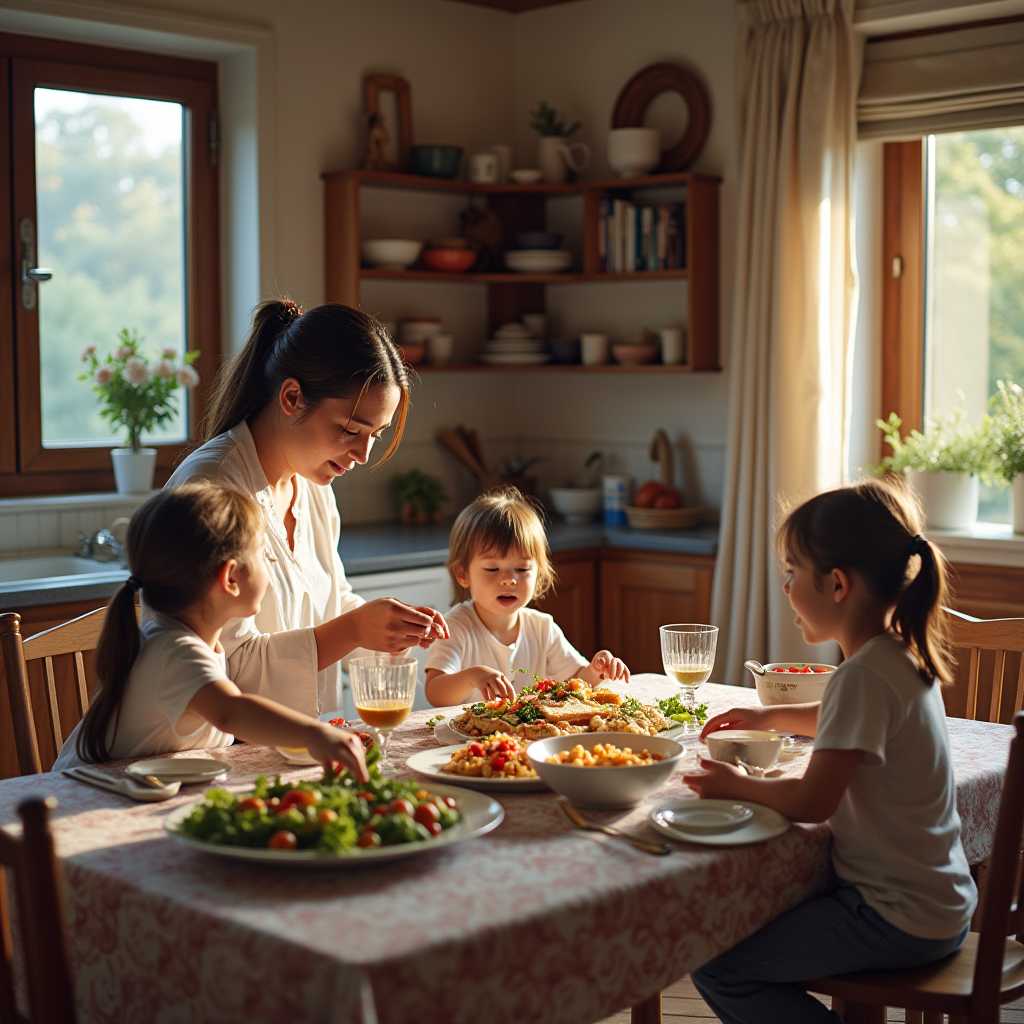} & \includegraphics[width=\imgwidth]{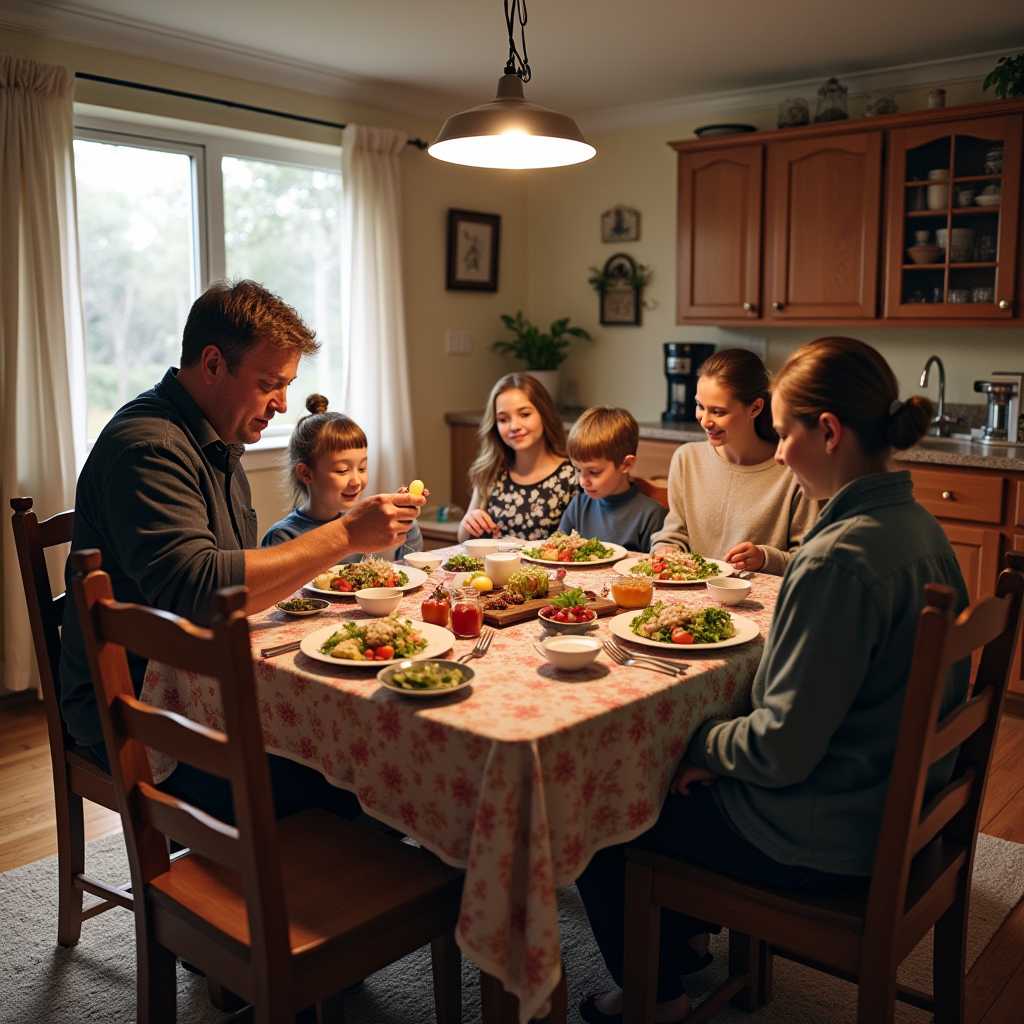} & \includegraphics[width=\imgwidth]{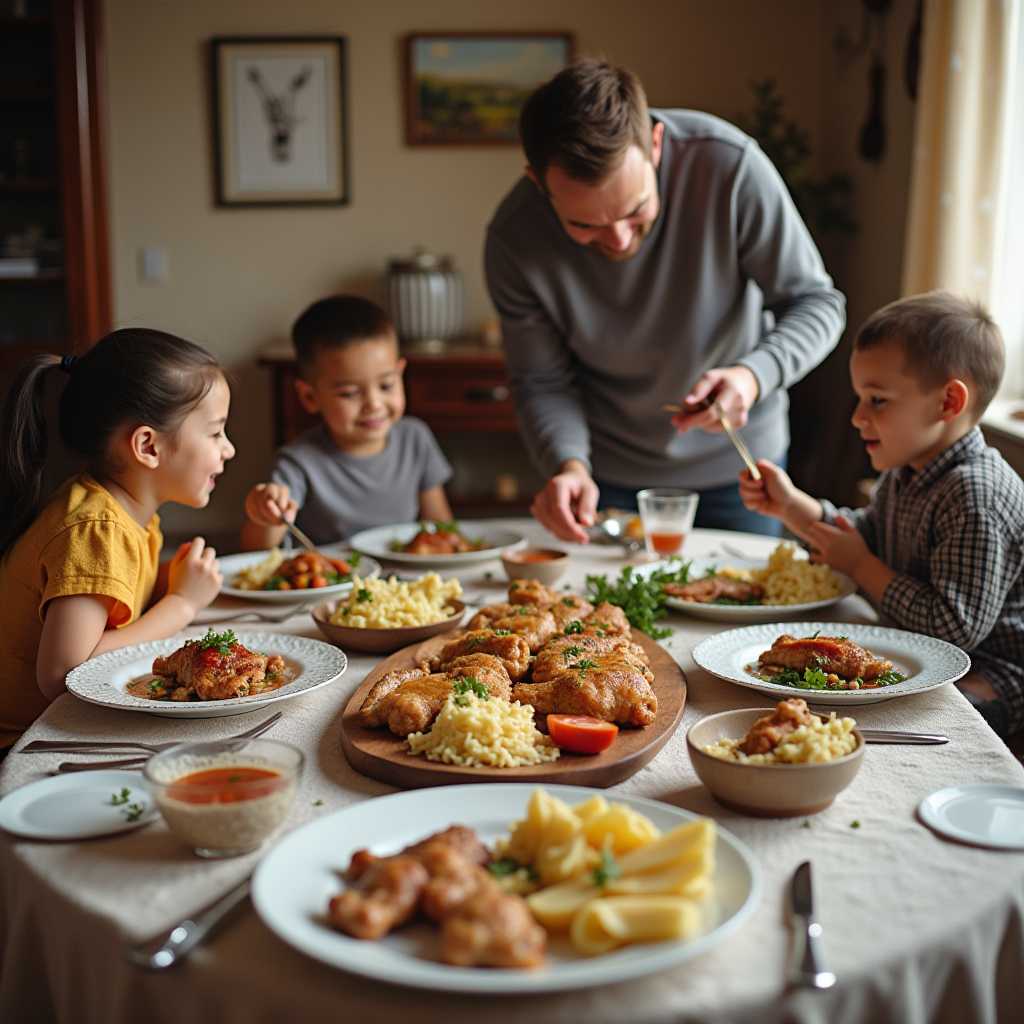} \\
        \multicolumn{9}{c}{\vspace{2pt}\small ``A family enjoying a traditional meal together at home'' \vspace{8pt}} \\

        \vertlabel{Flux} & \includegraphics[width=\imgwidth]{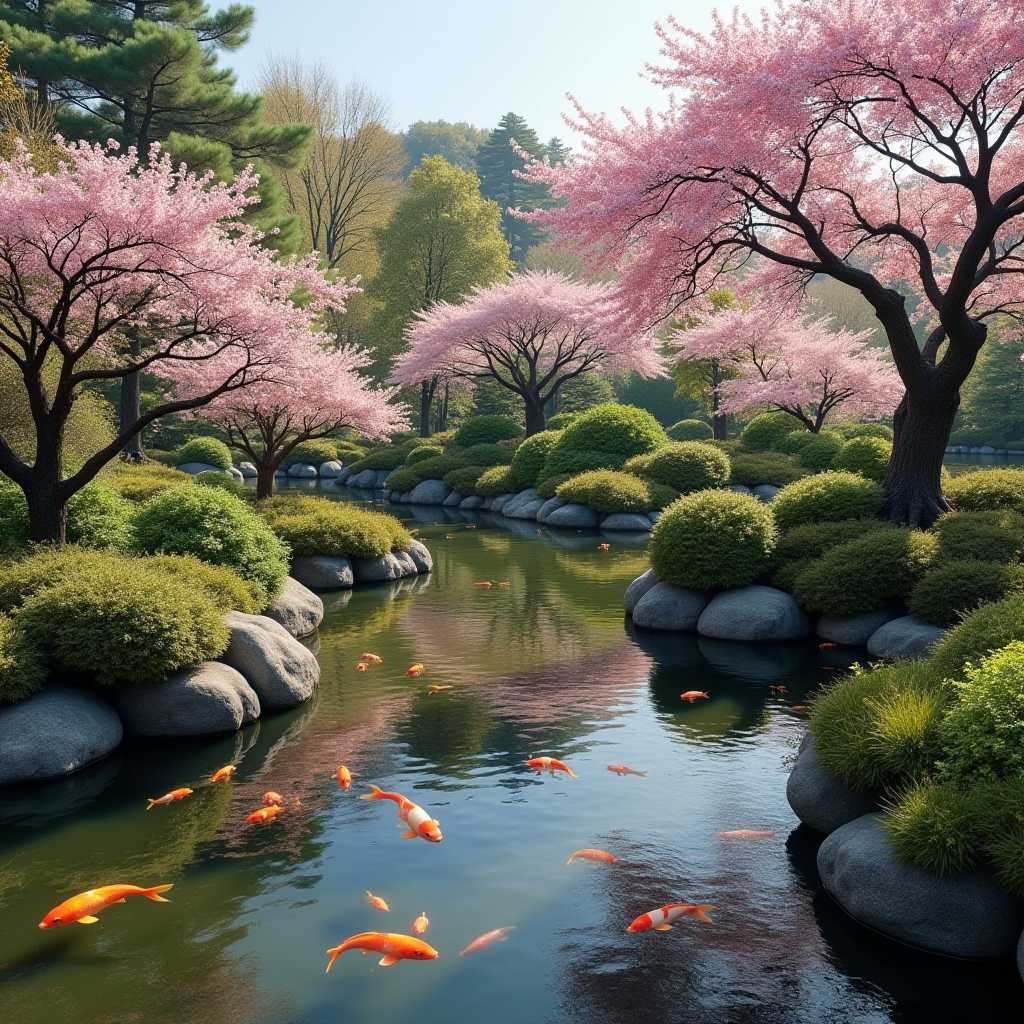} & \includegraphics[width=\imgwidth]{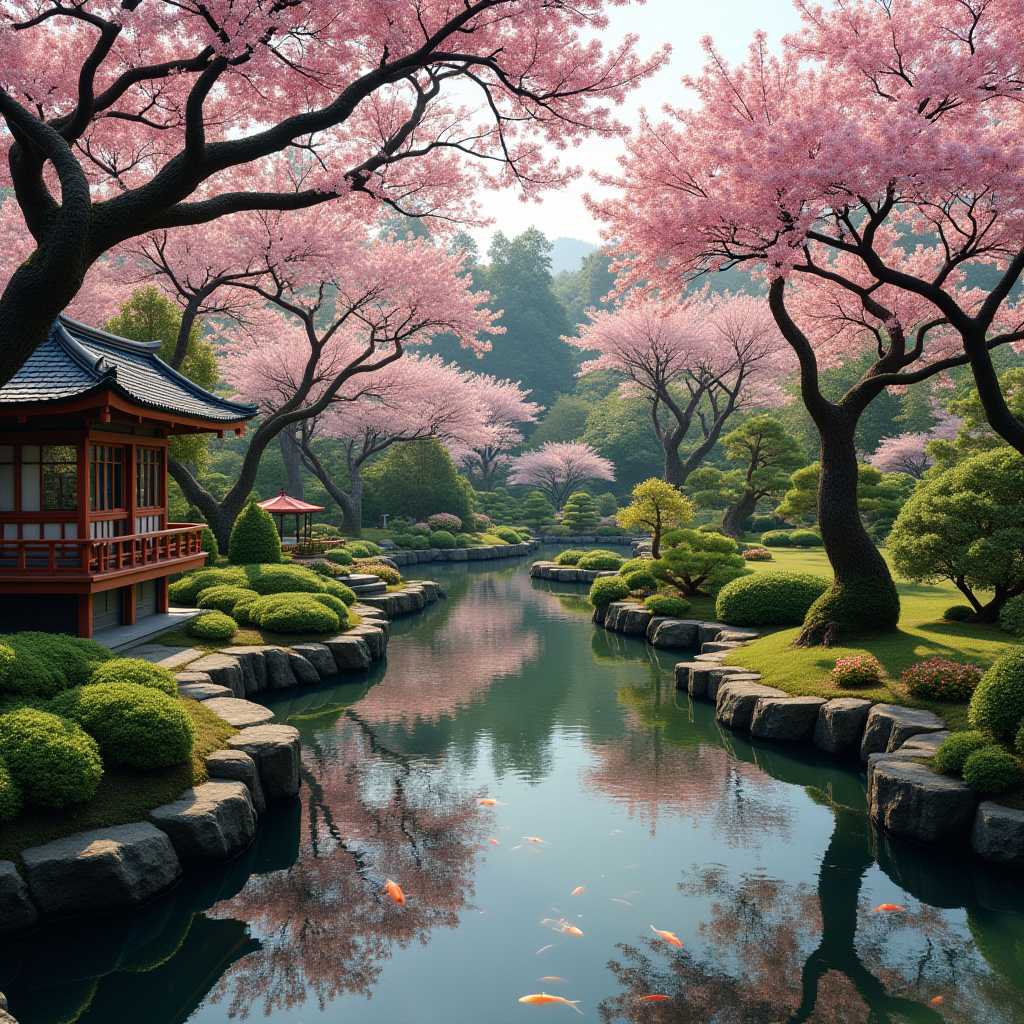} & \includegraphics[width=\imgwidth]{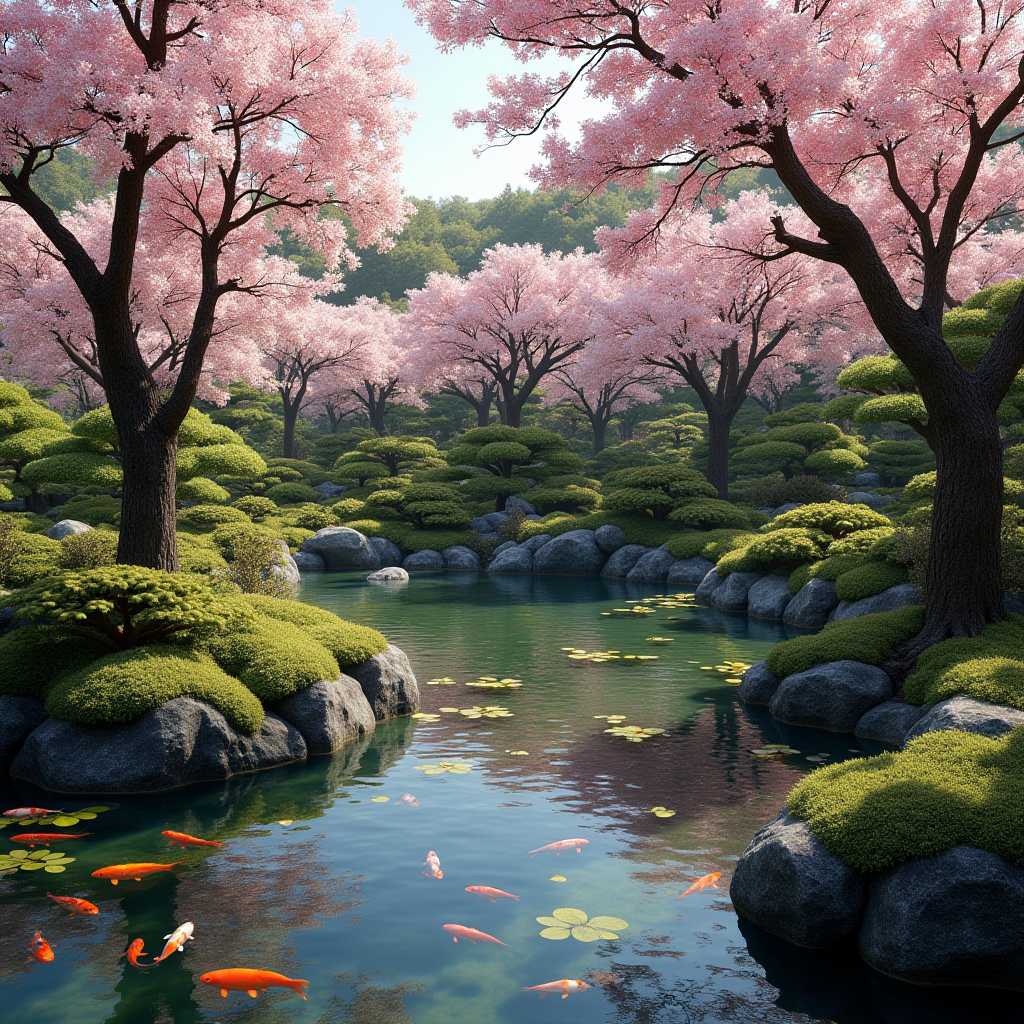} & \includegraphics[width=\imgwidth]{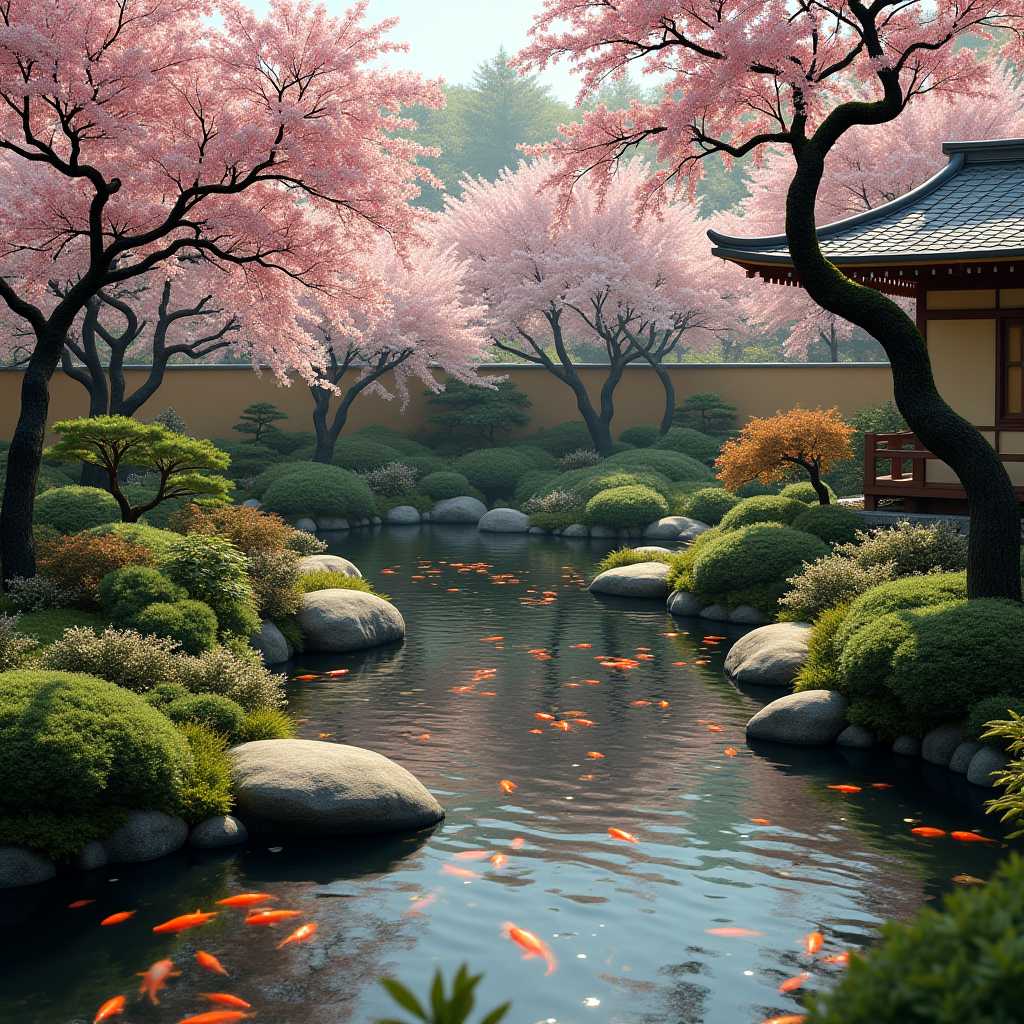} & \includegraphics[width=\imgwidth]{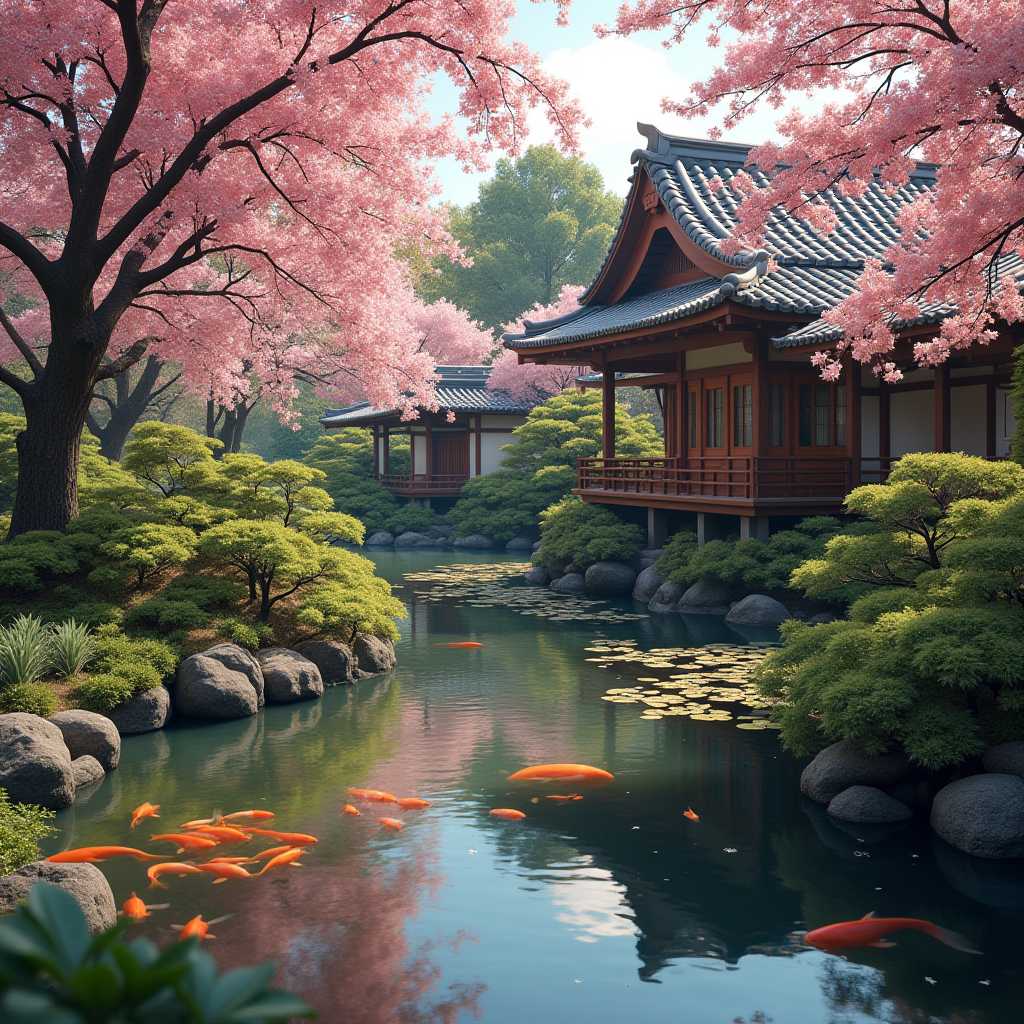} & \includegraphics[width=\imgwidth]{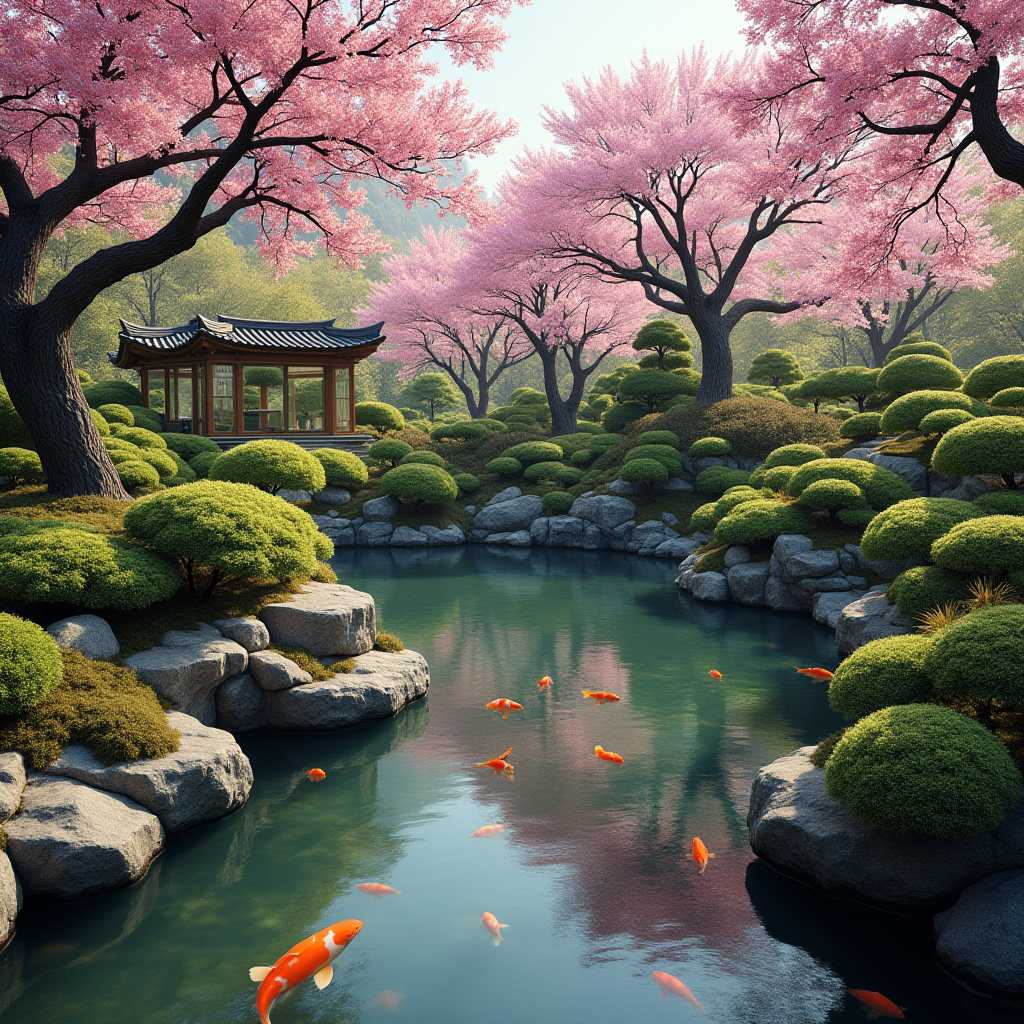} & \includegraphics[width=\imgwidth]{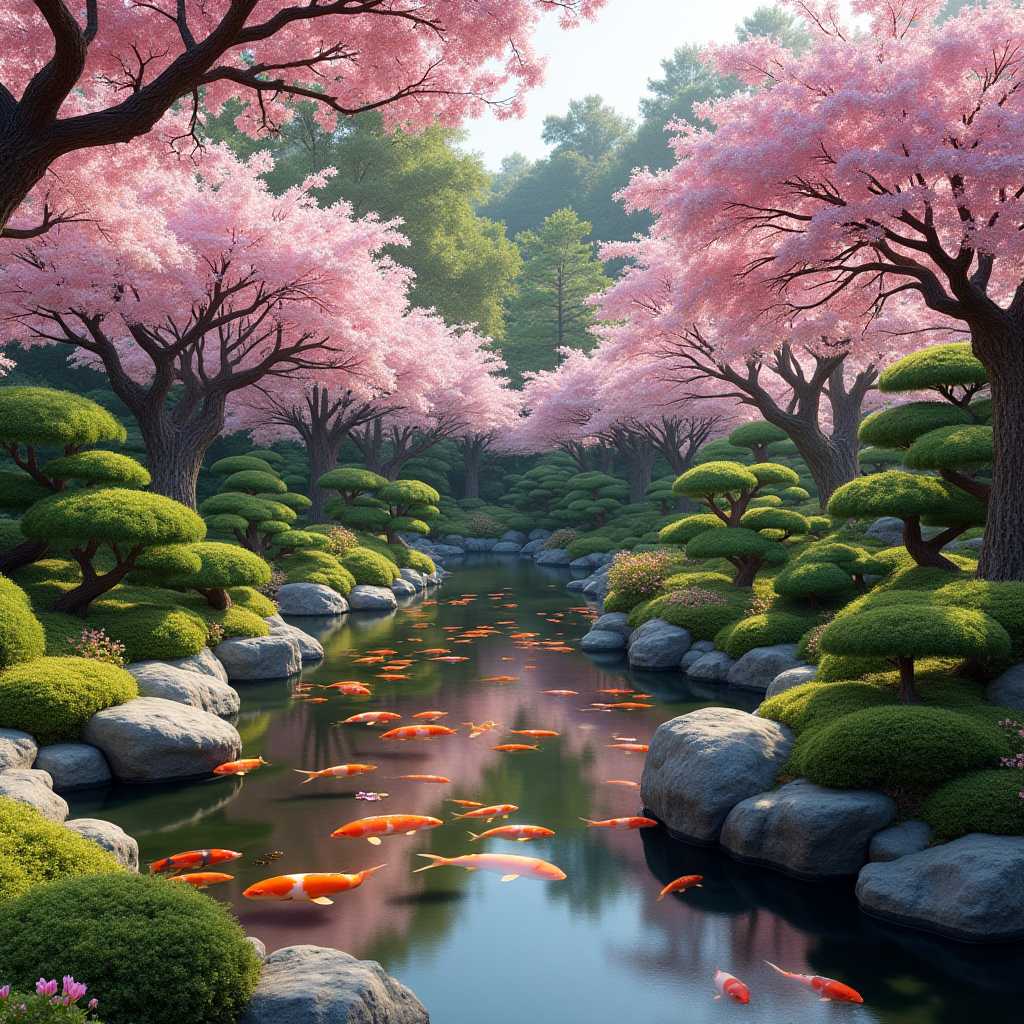} & \includegraphics[width=\imgwidth]{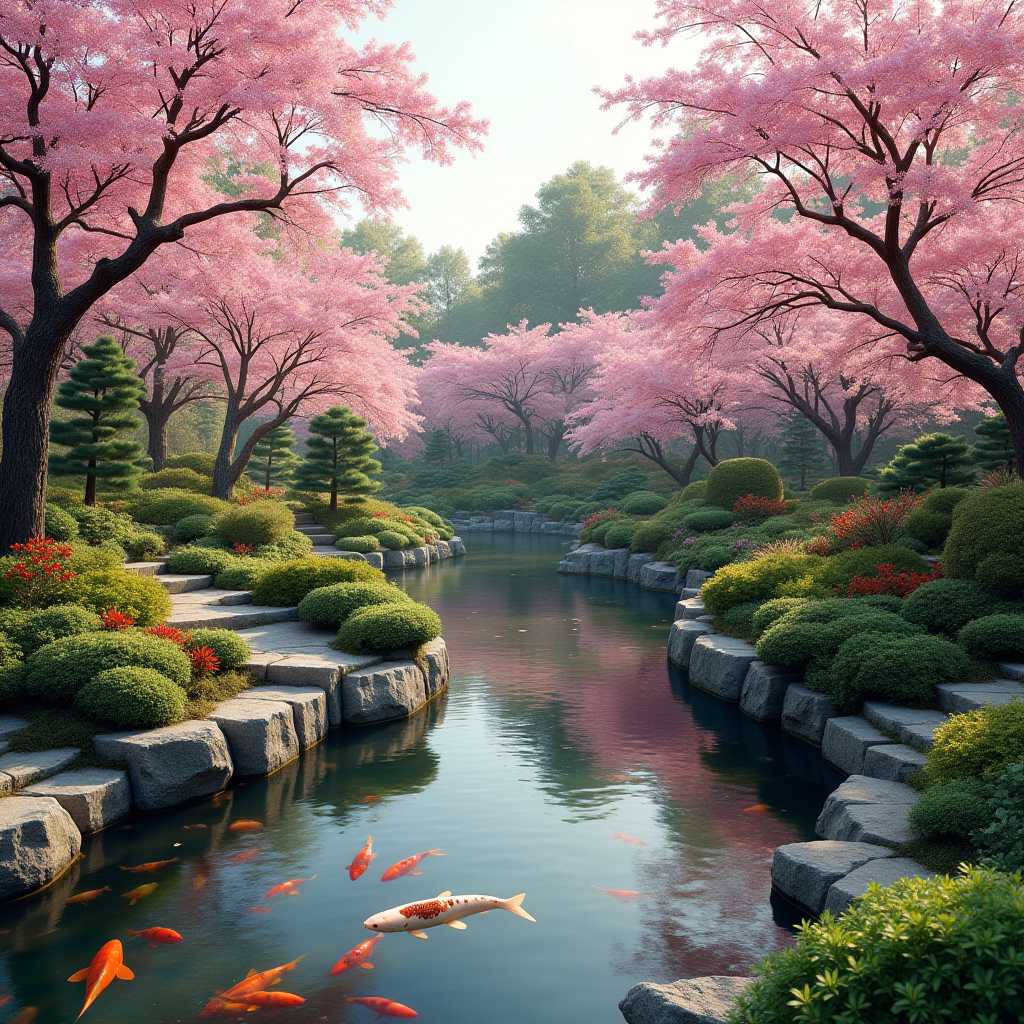} \\[-1pt]
        \vertlabel{Ours} & \includegraphics[width=\imgwidth]{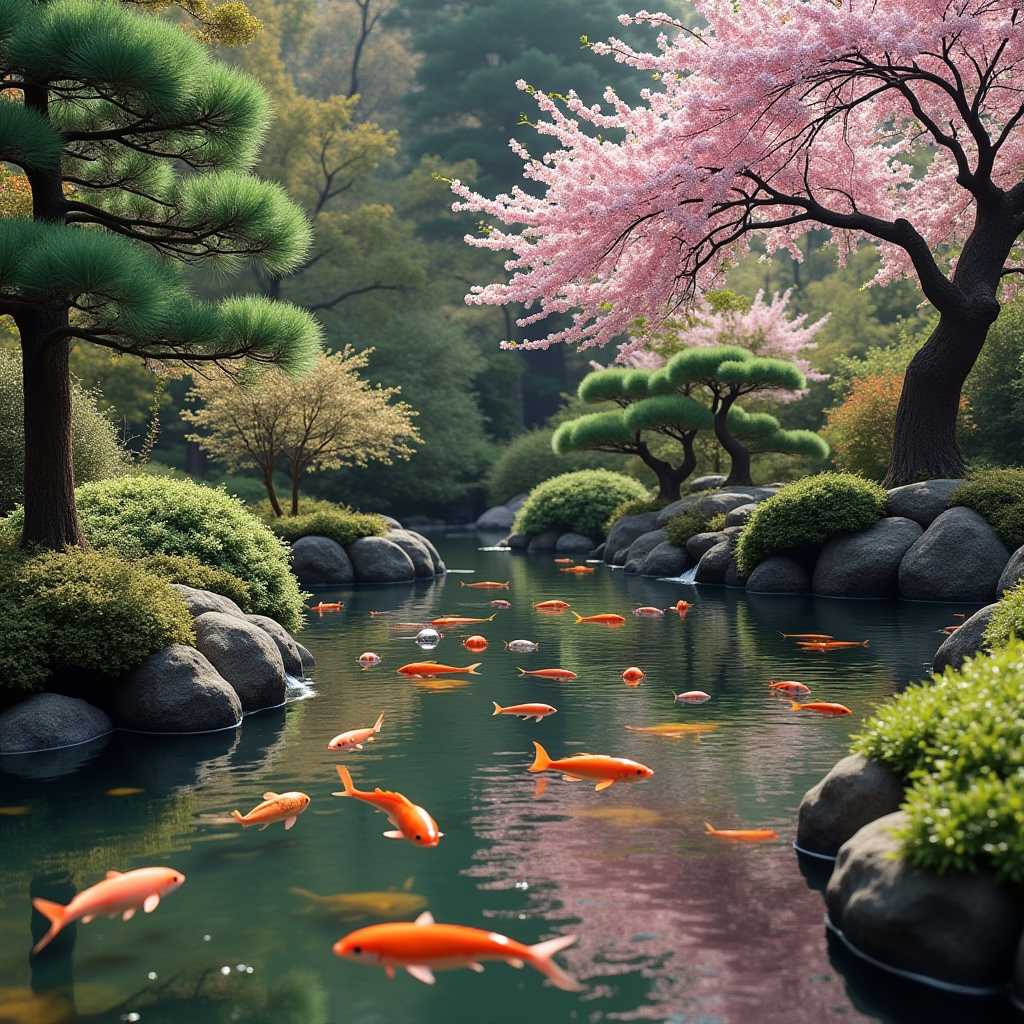} & \includegraphics[width=\imgwidth]{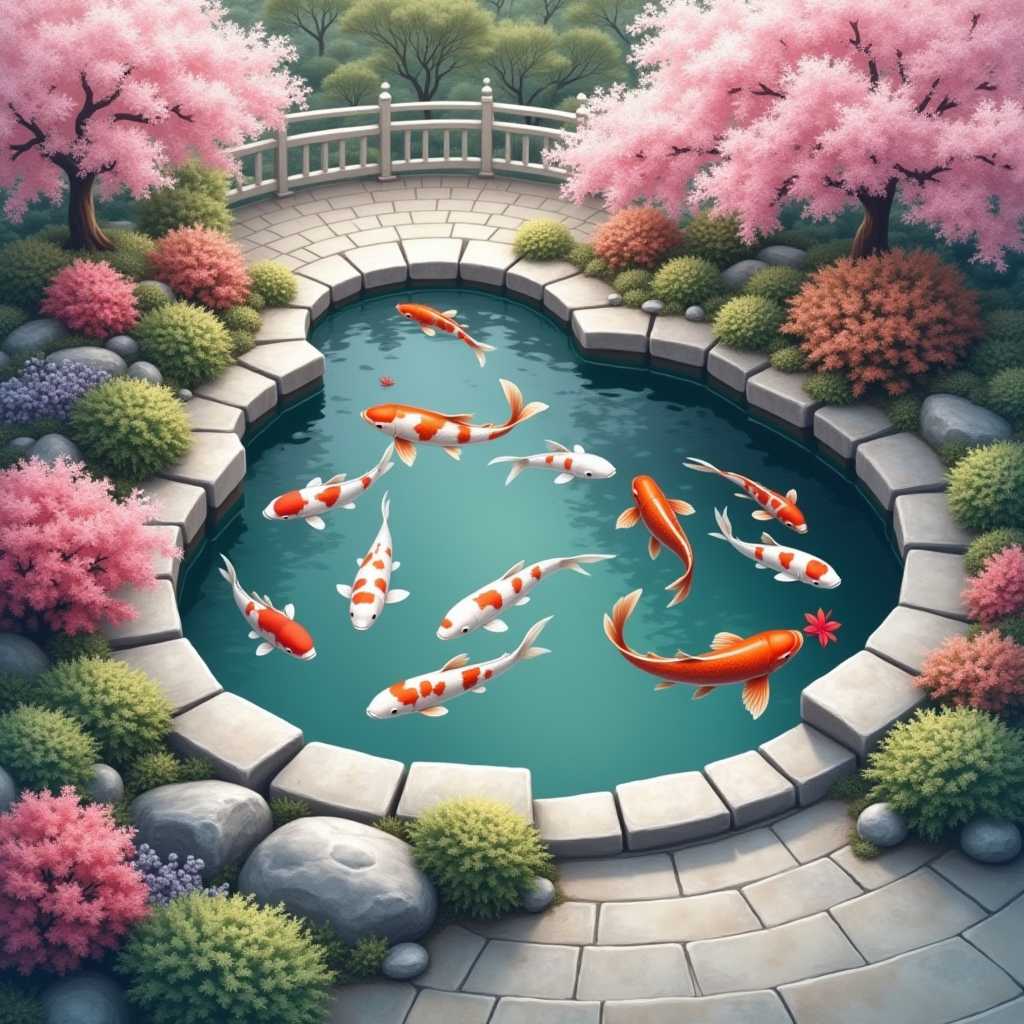} & \includegraphics[width=\imgwidth]{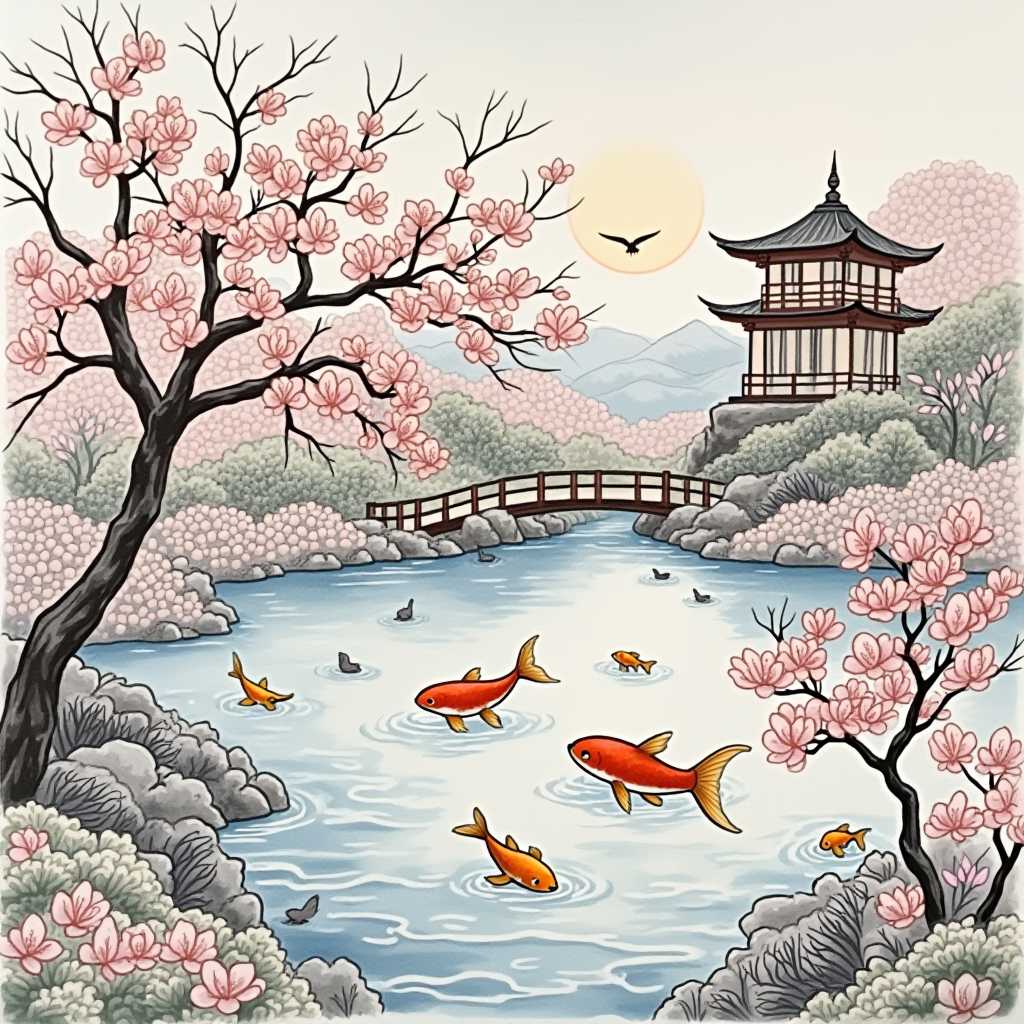} & \includegraphics[width=\imgwidth]{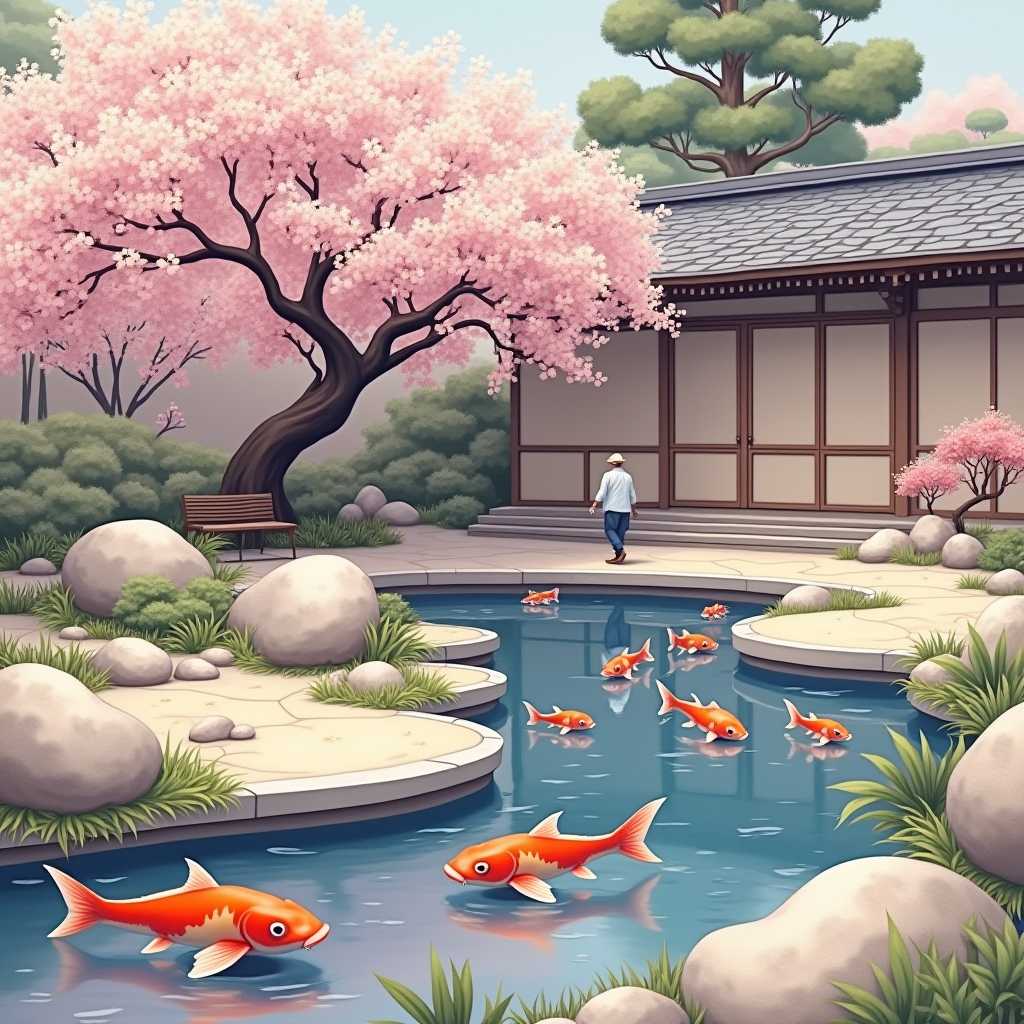} & \includegraphics[width=\imgwidth]{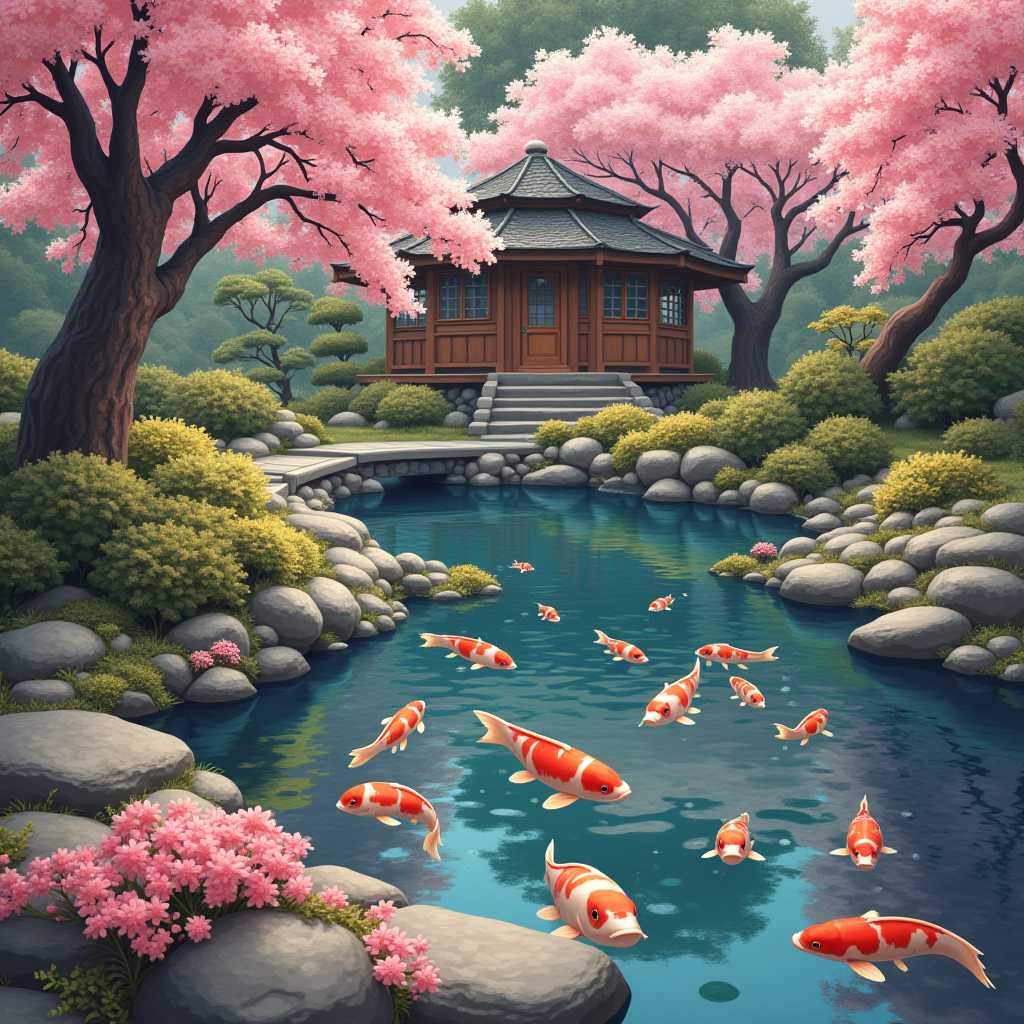} & \includegraphics[width=\imgwidth]{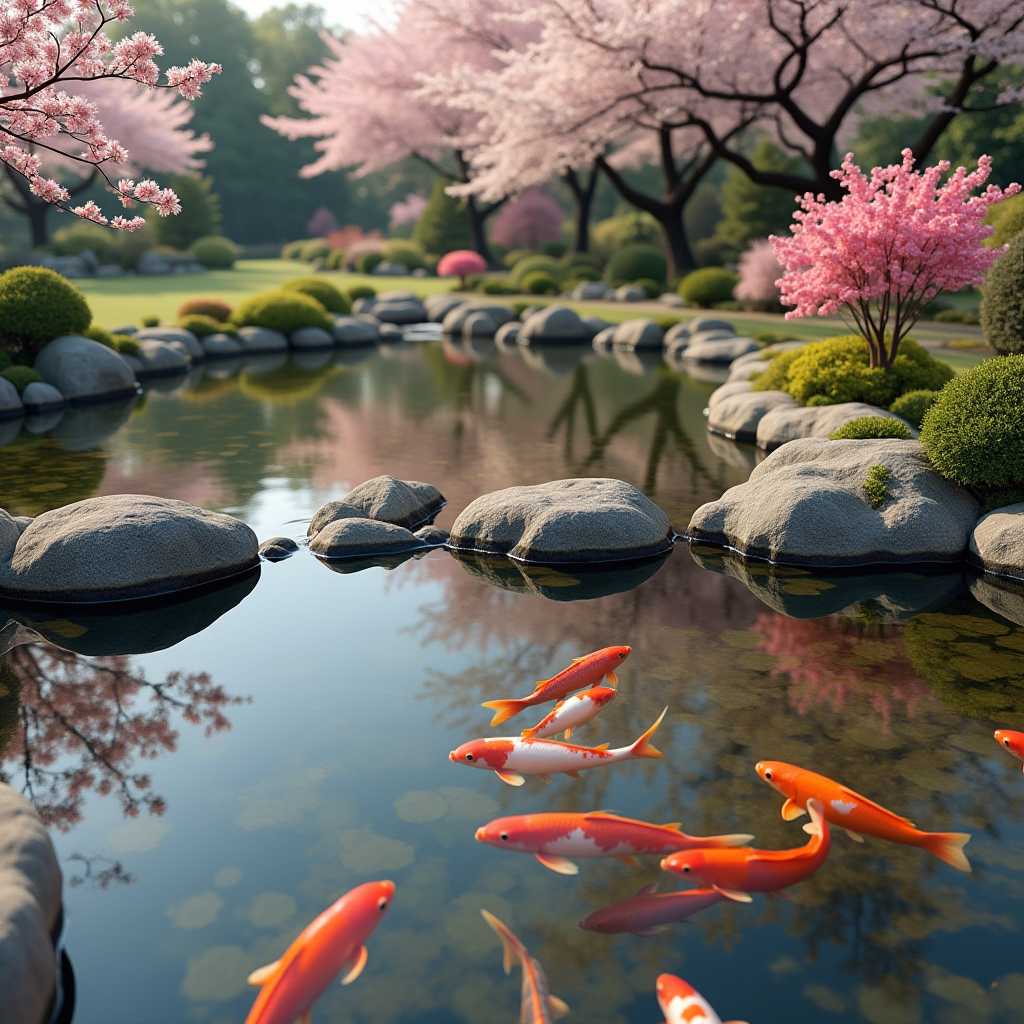} & \includegraphics[width=\imgwidth]{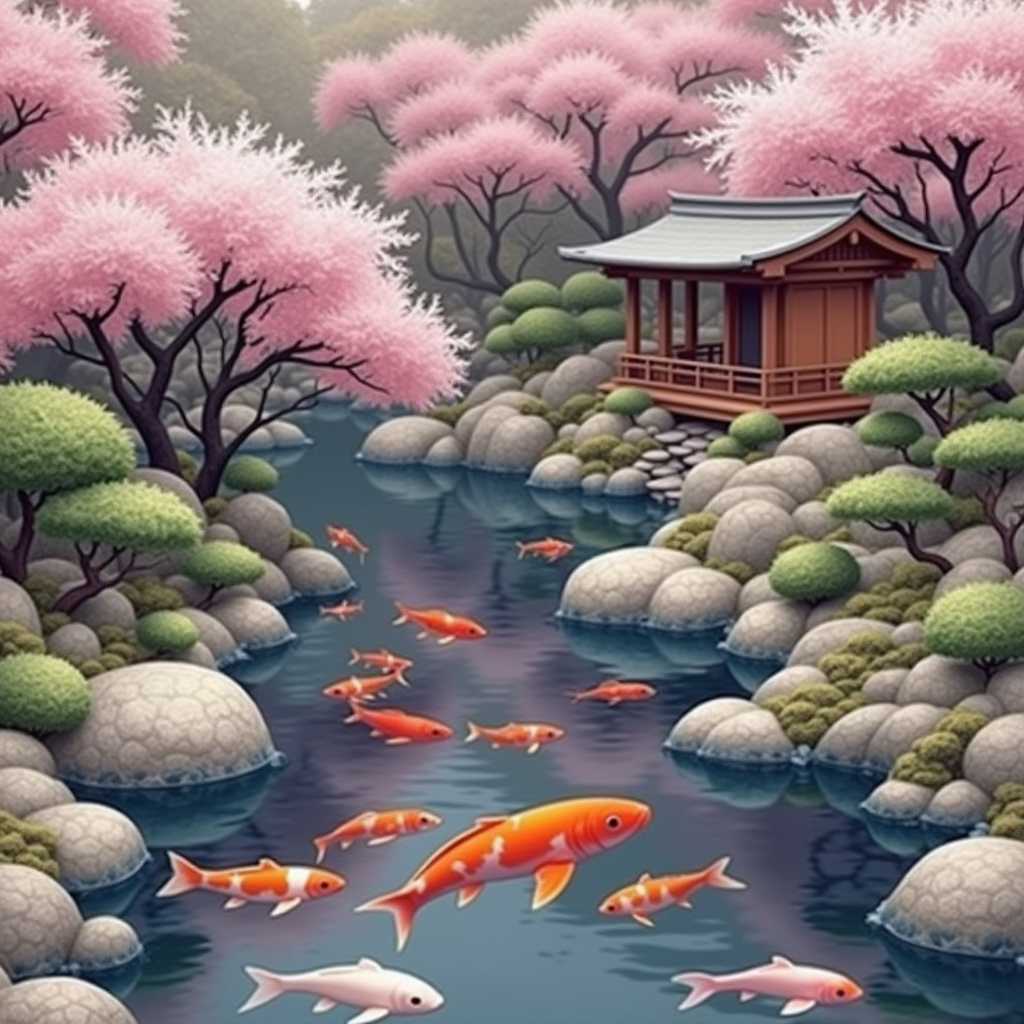} & \includegraphics[width=\imgwidth]{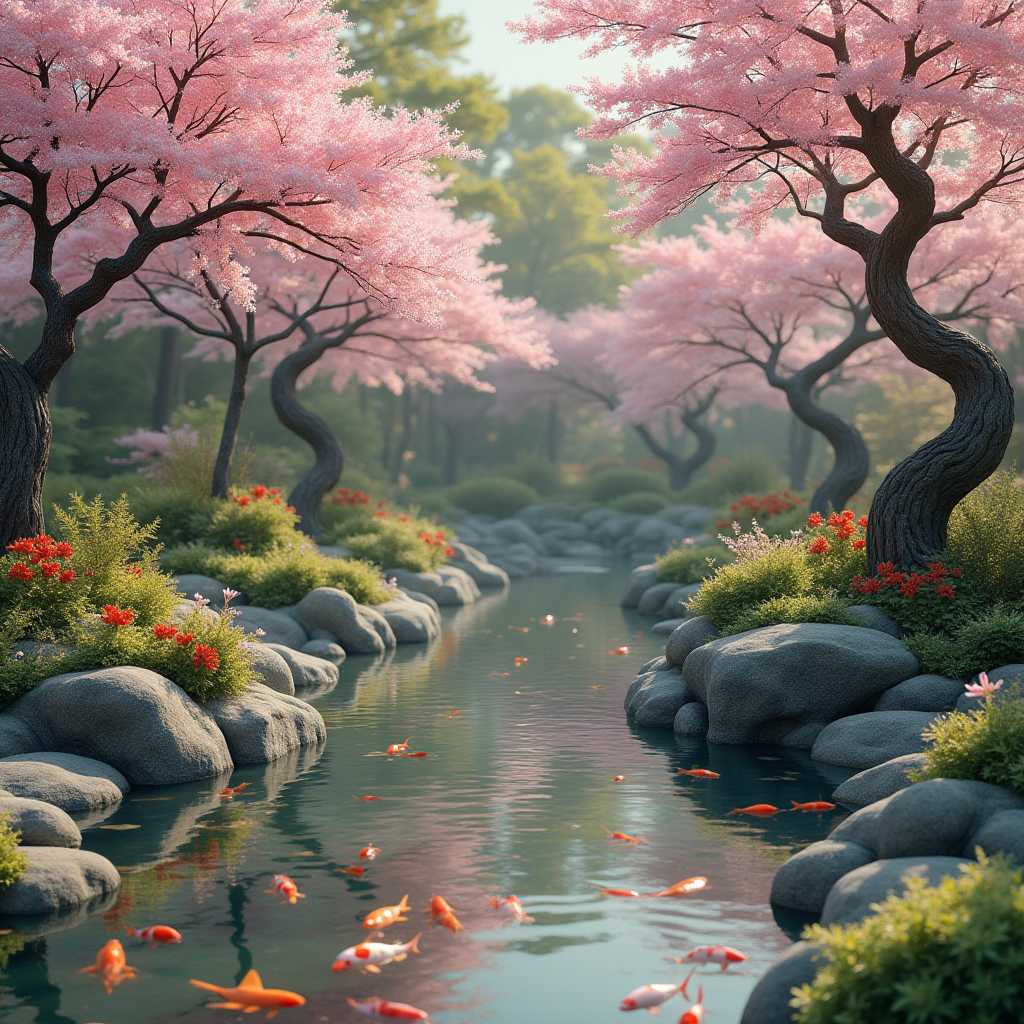} \\
        \multicolumn{9}{c}{\vspace{2pt}\small ``A beautiful Japanese garden with a koi pond and cherry blossoms'' \vspace{8pt}} \\

        \vertlabel{Flux} & \includegraphics[width=\imgwidth]{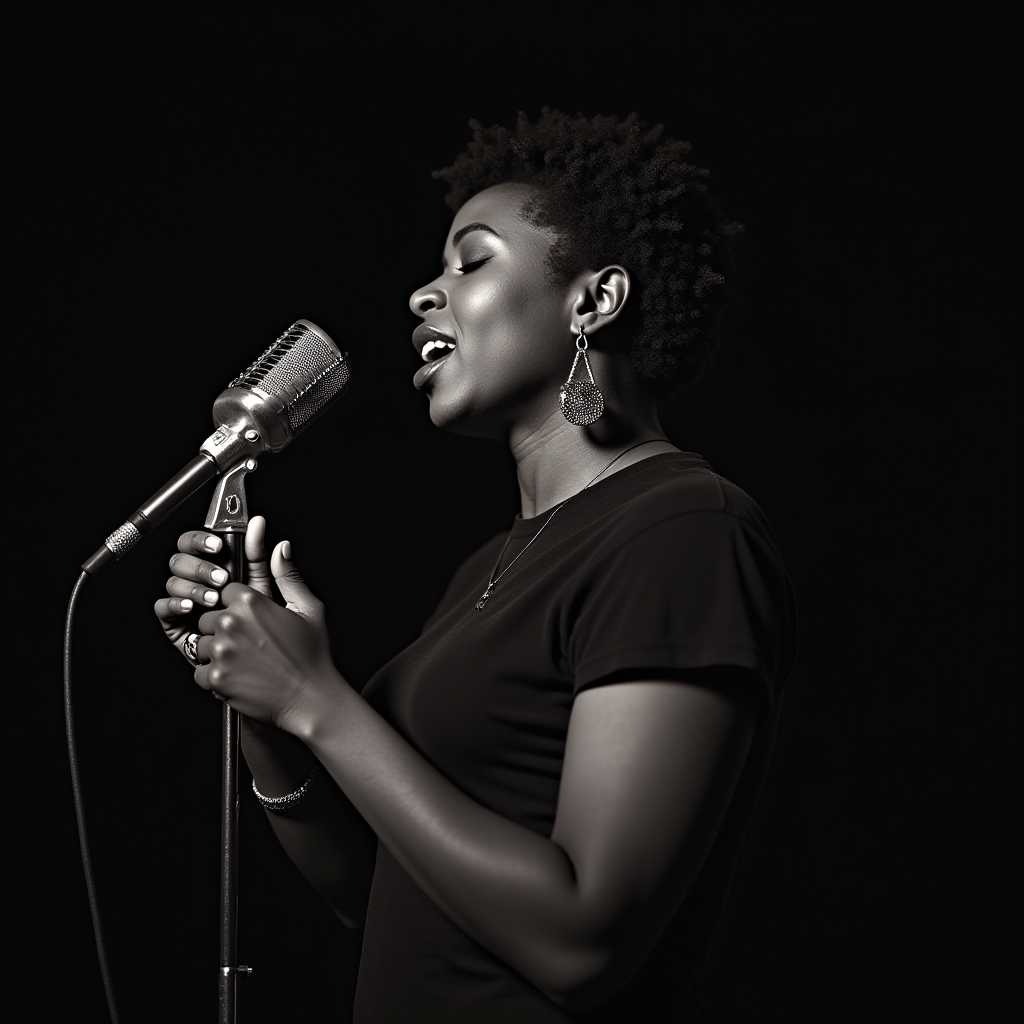} & \includegraphics[width=\imgwidth]{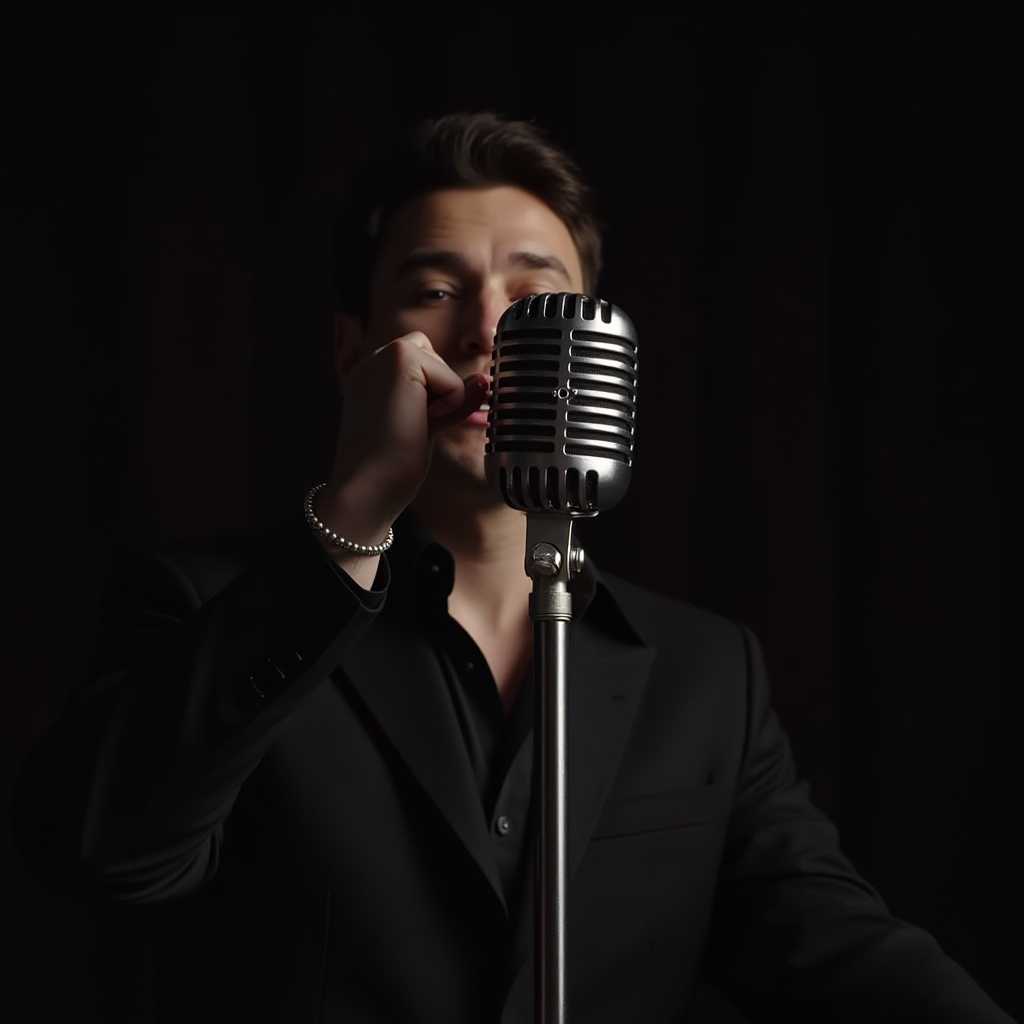} & \includegraphics[width=\imgwidth]{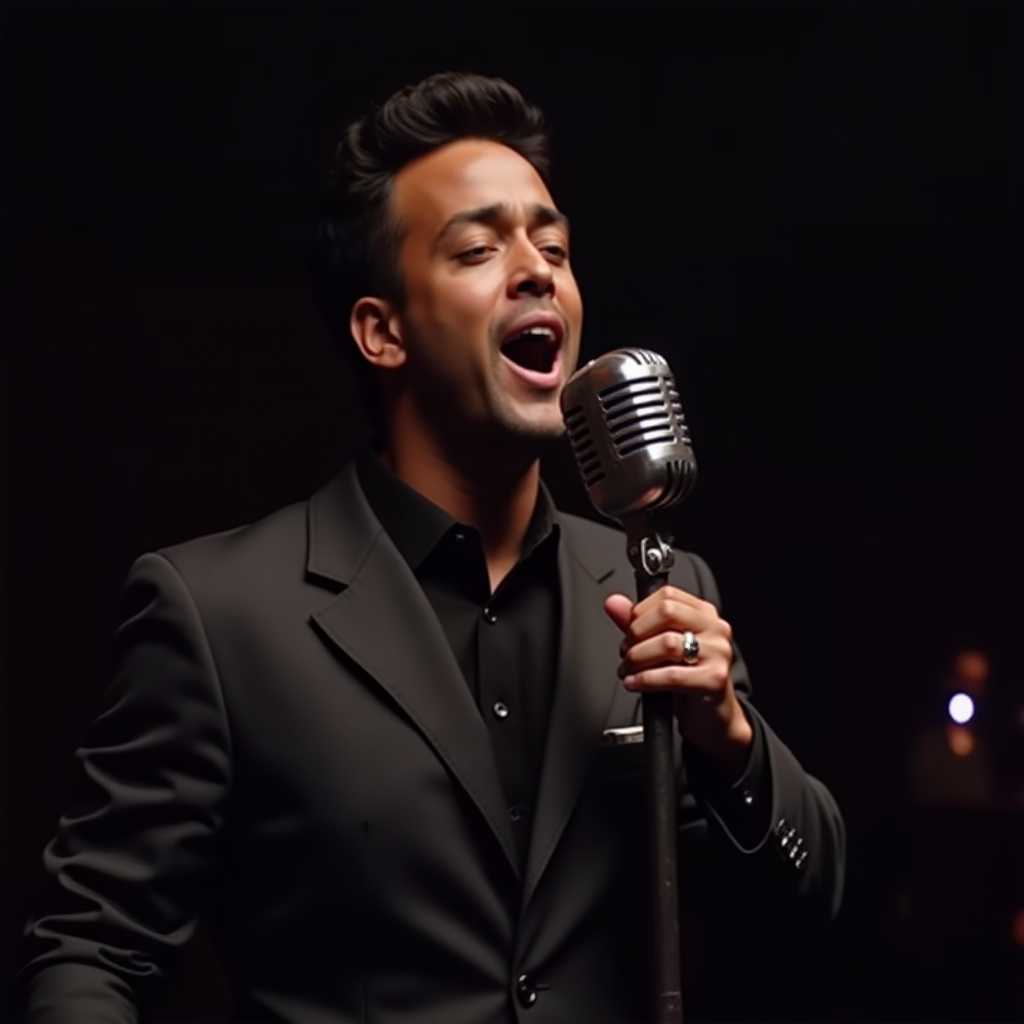} & \includegraphics[width=\imgwidth]{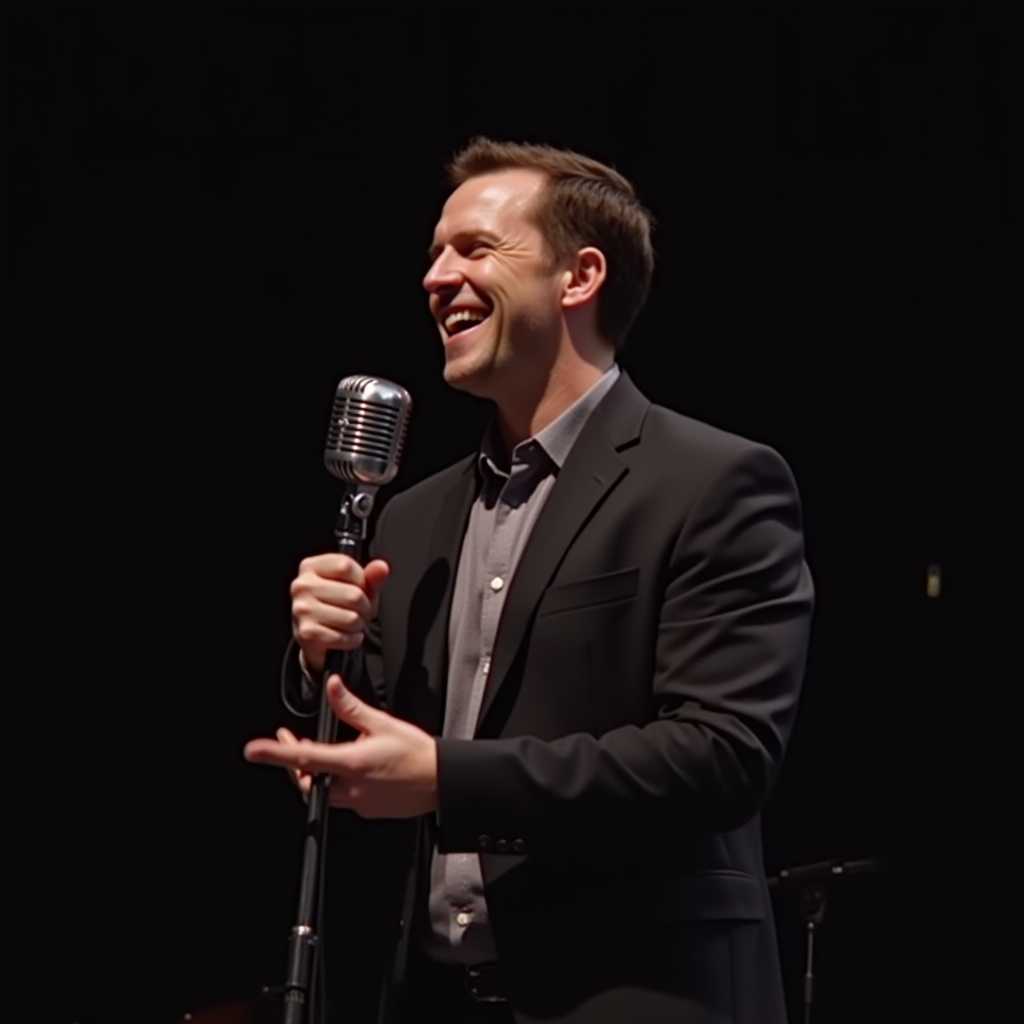} & \includegraphics[width=\imgwidth]{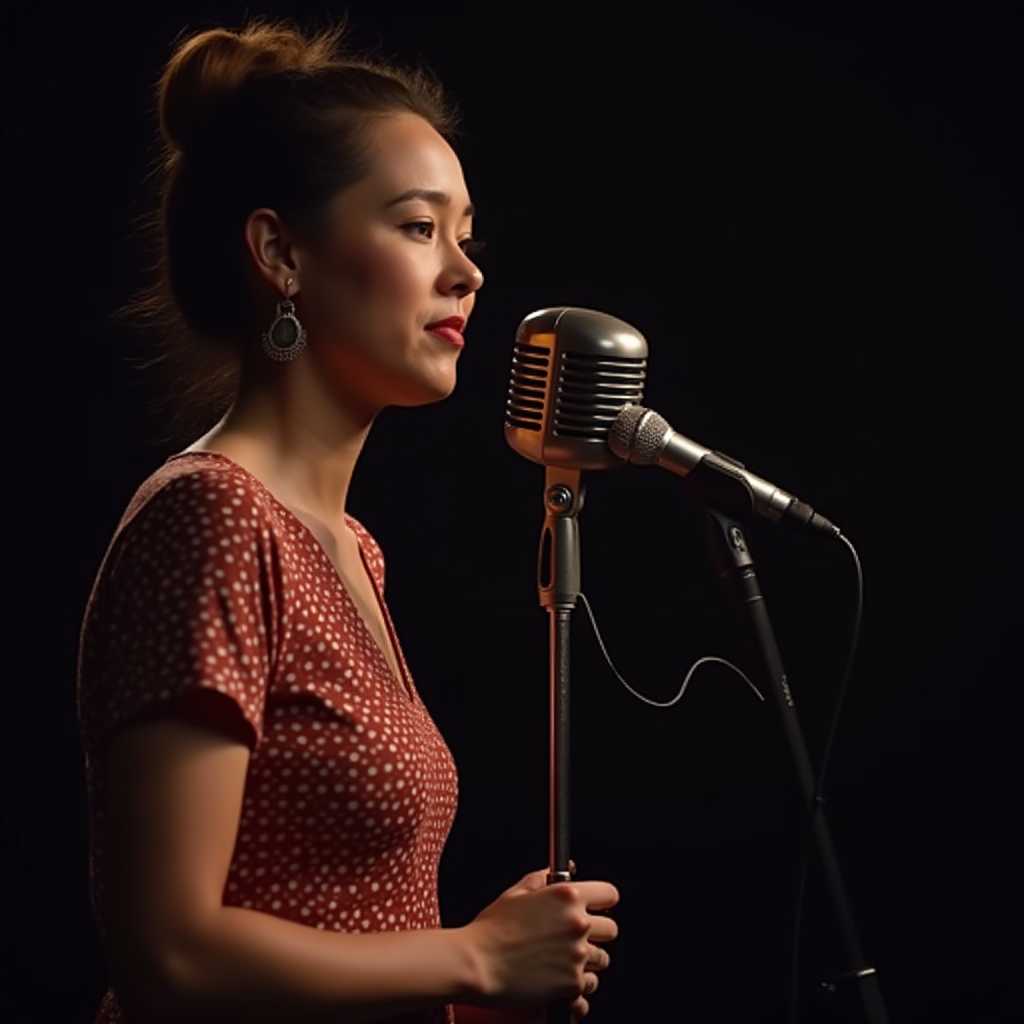} & \includegraphics[width=\imgwidth]{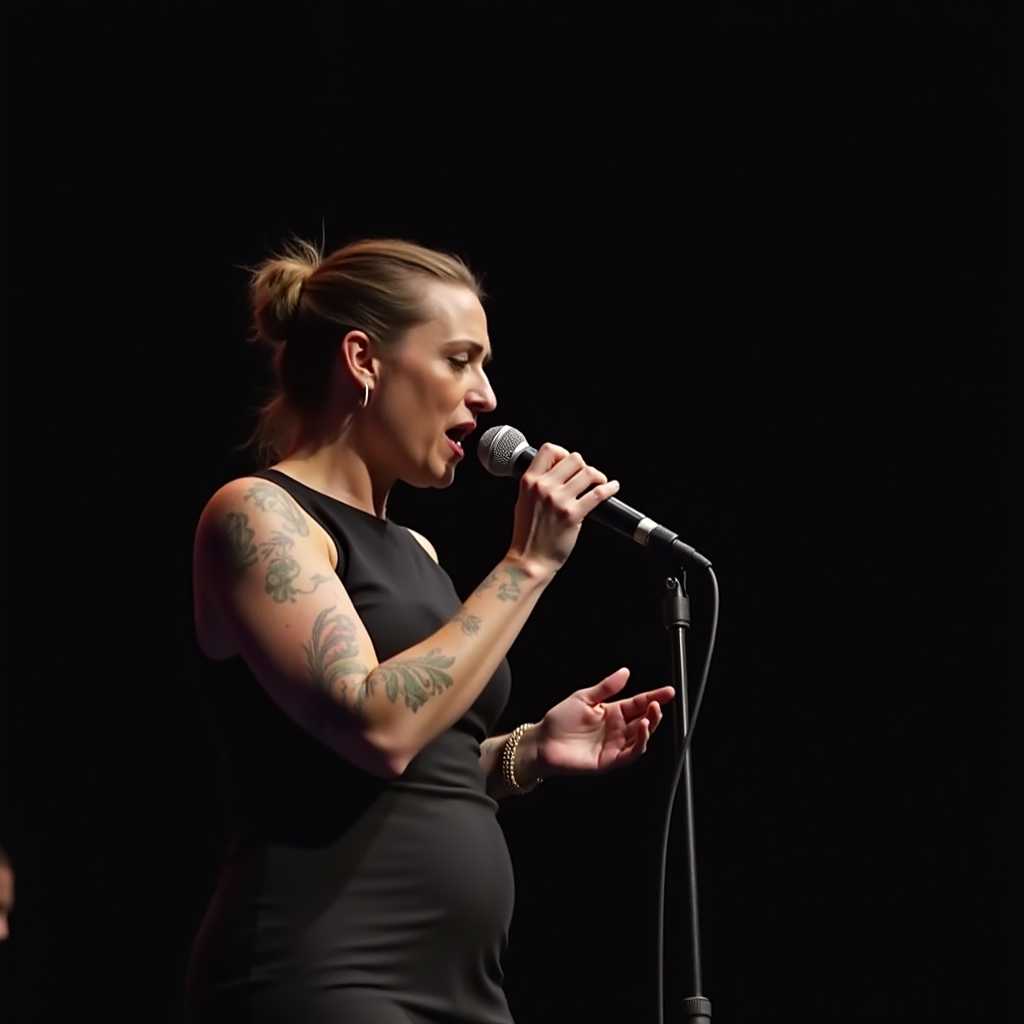} & \includegraphics[width=\imgwidth]{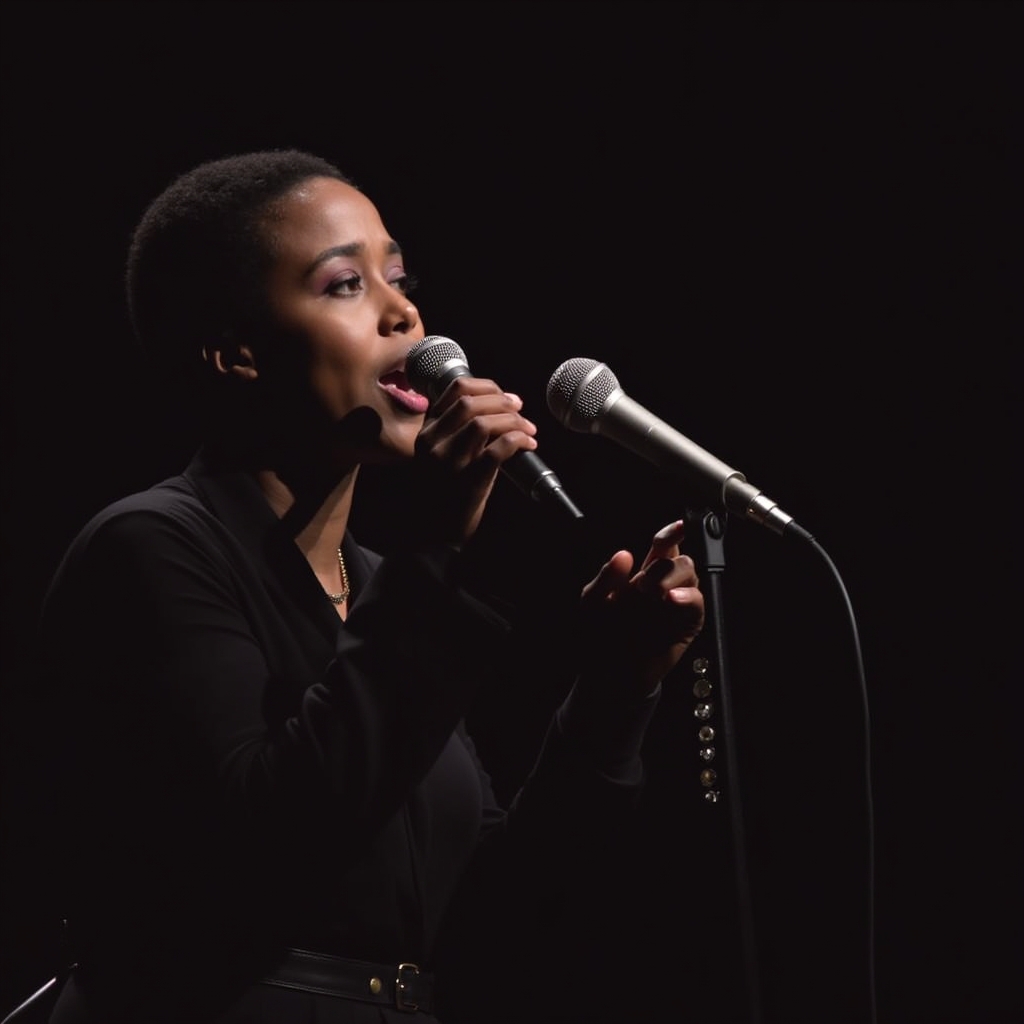} & \includegraphics[width=\imgwidth]{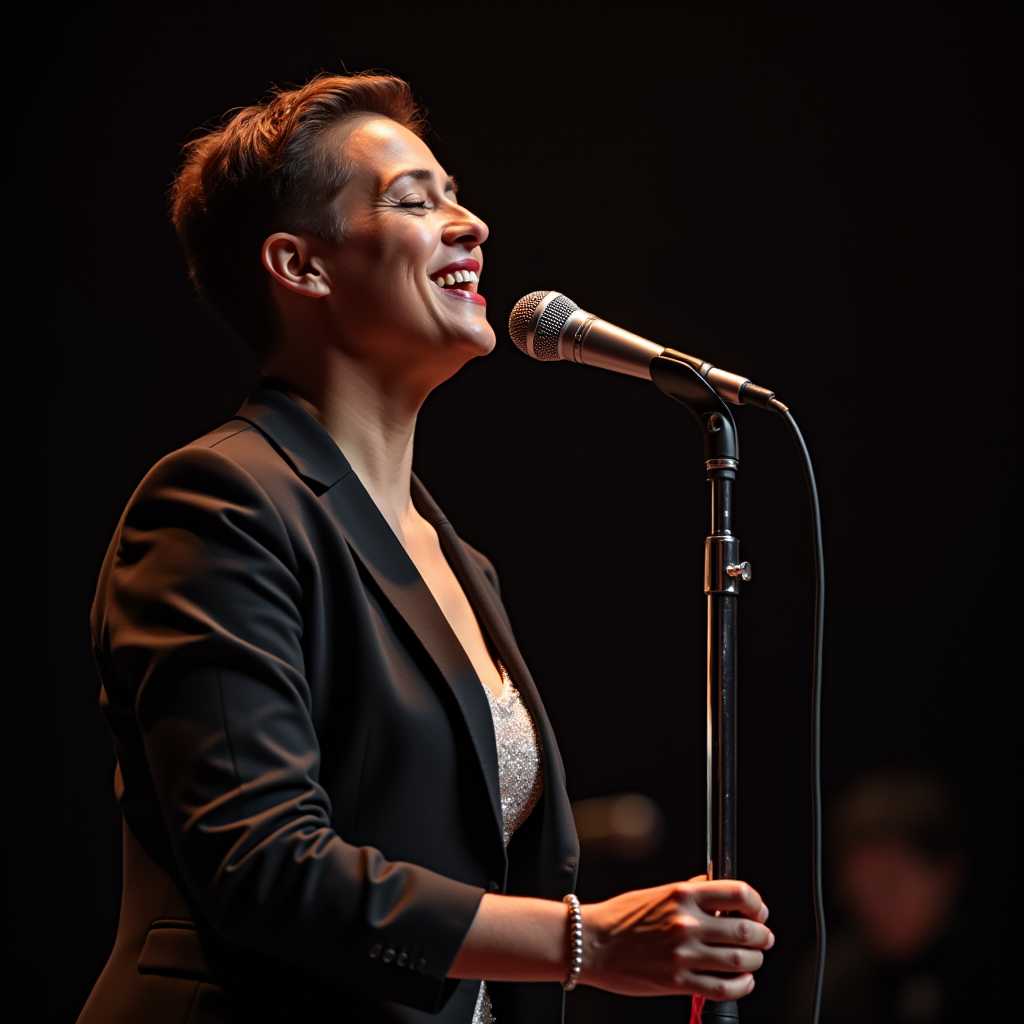} \\[-1pt]
        \vertlabel{Ours} & \includegraphics[width=\imgwidth]{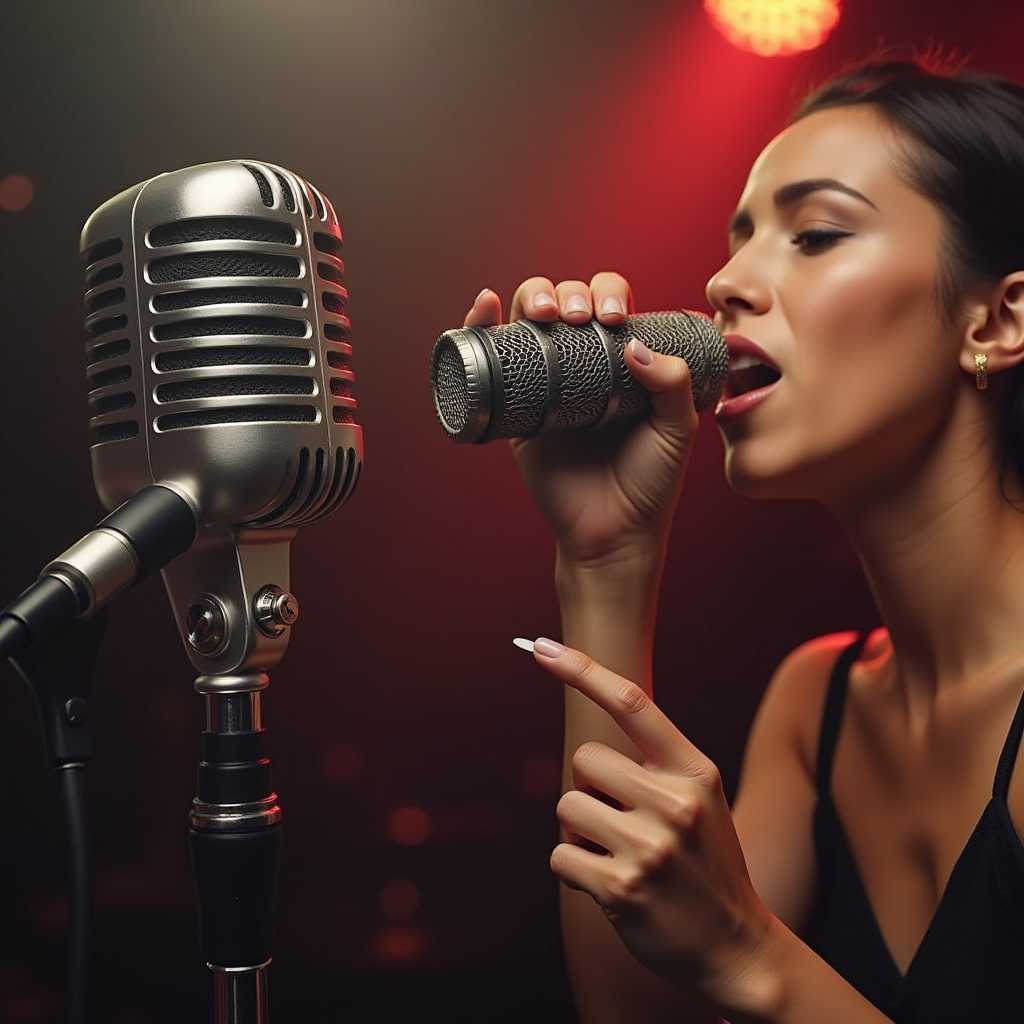} & \includegraphics[width=\imgwidth]{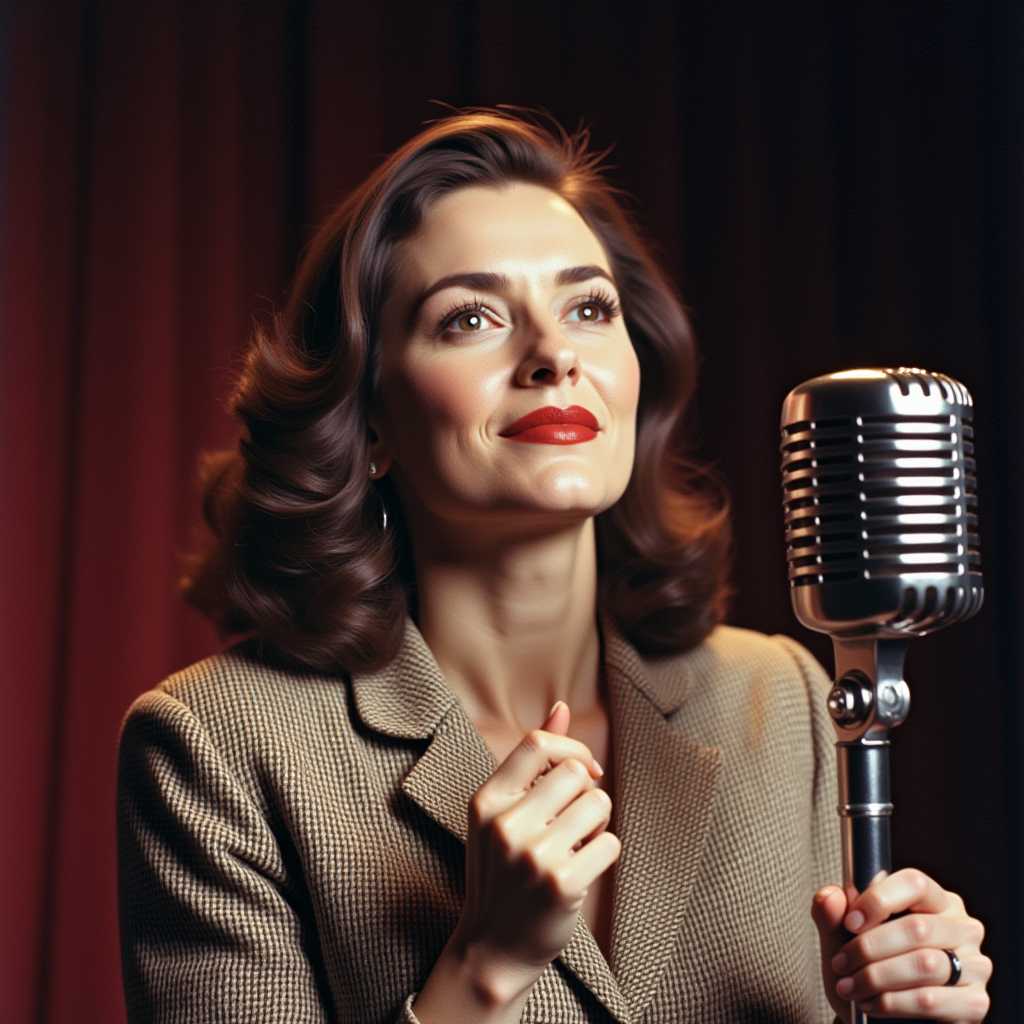} & \includegraphics[width=\imgwidth]{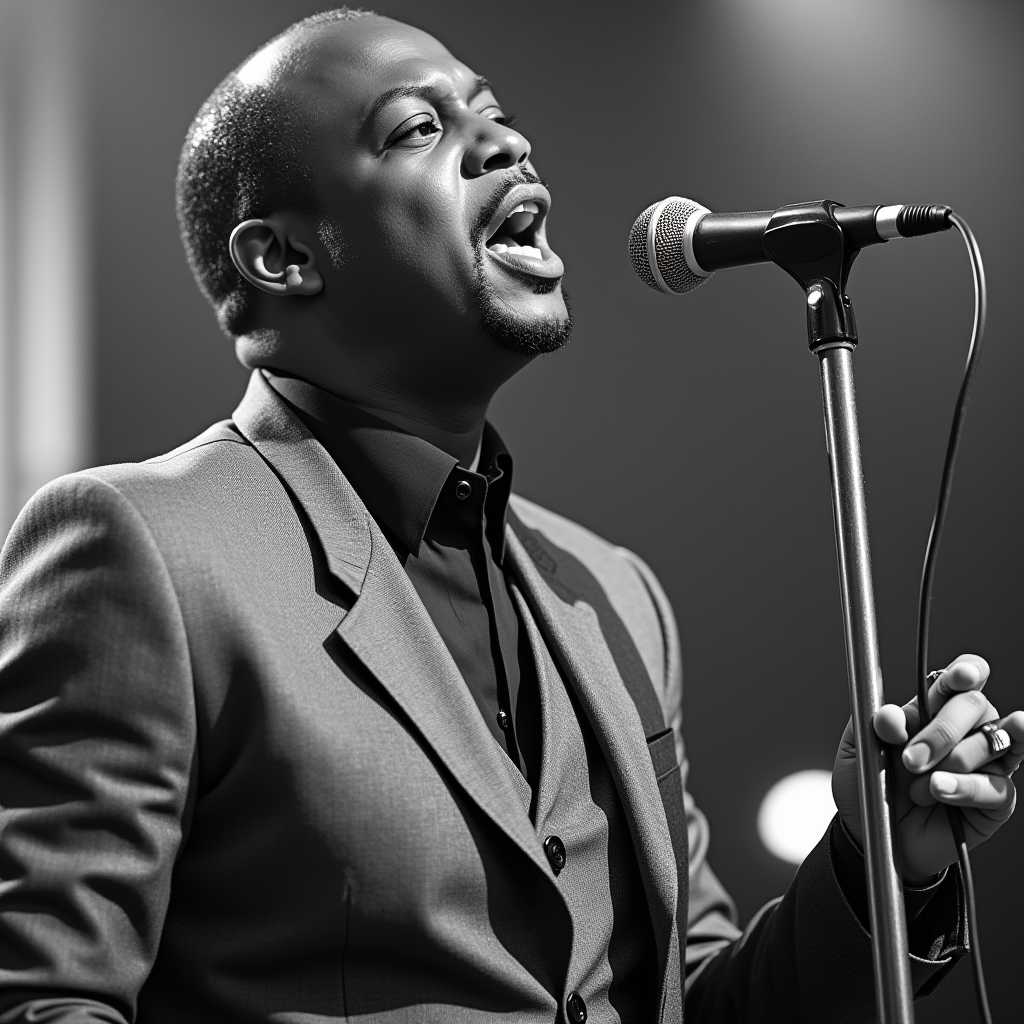} & \includegraphics[width=\imgwidth]{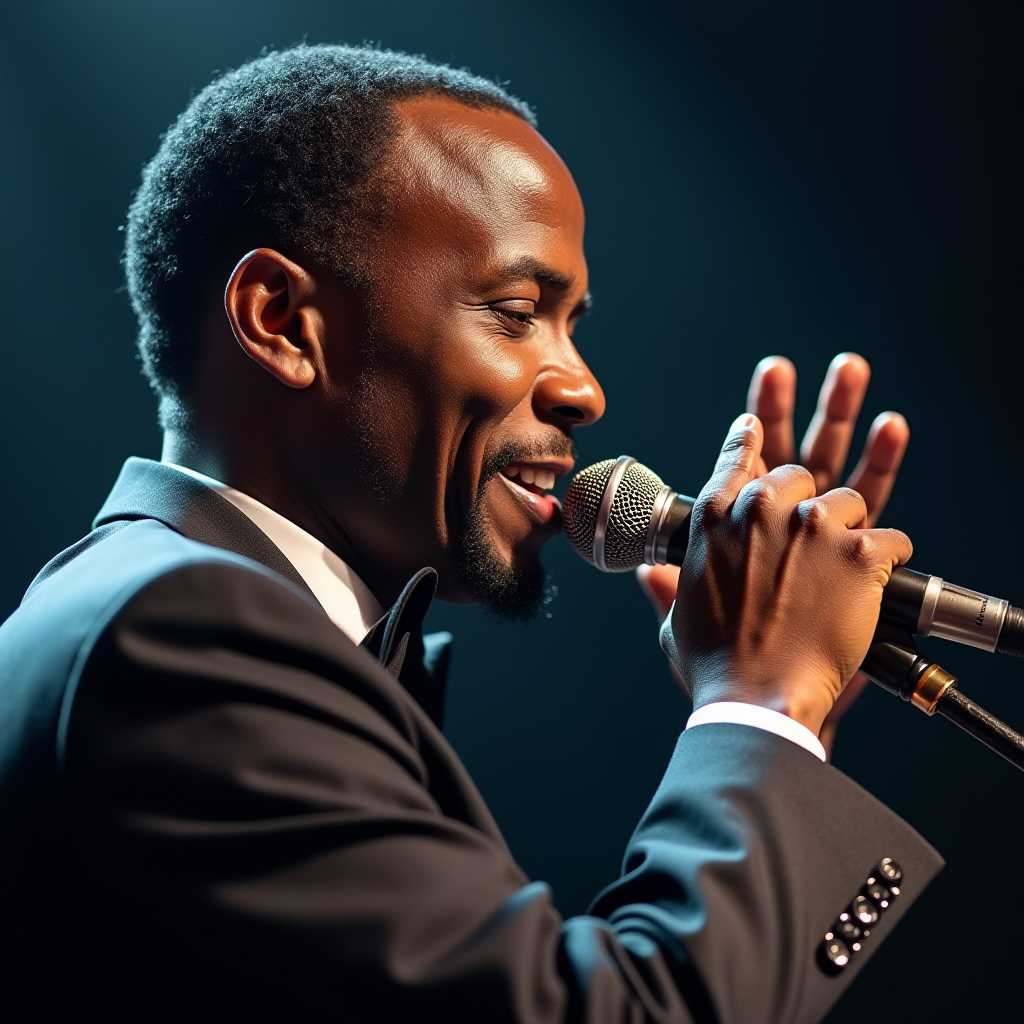} & \includegraphics[width=\imgwidth]{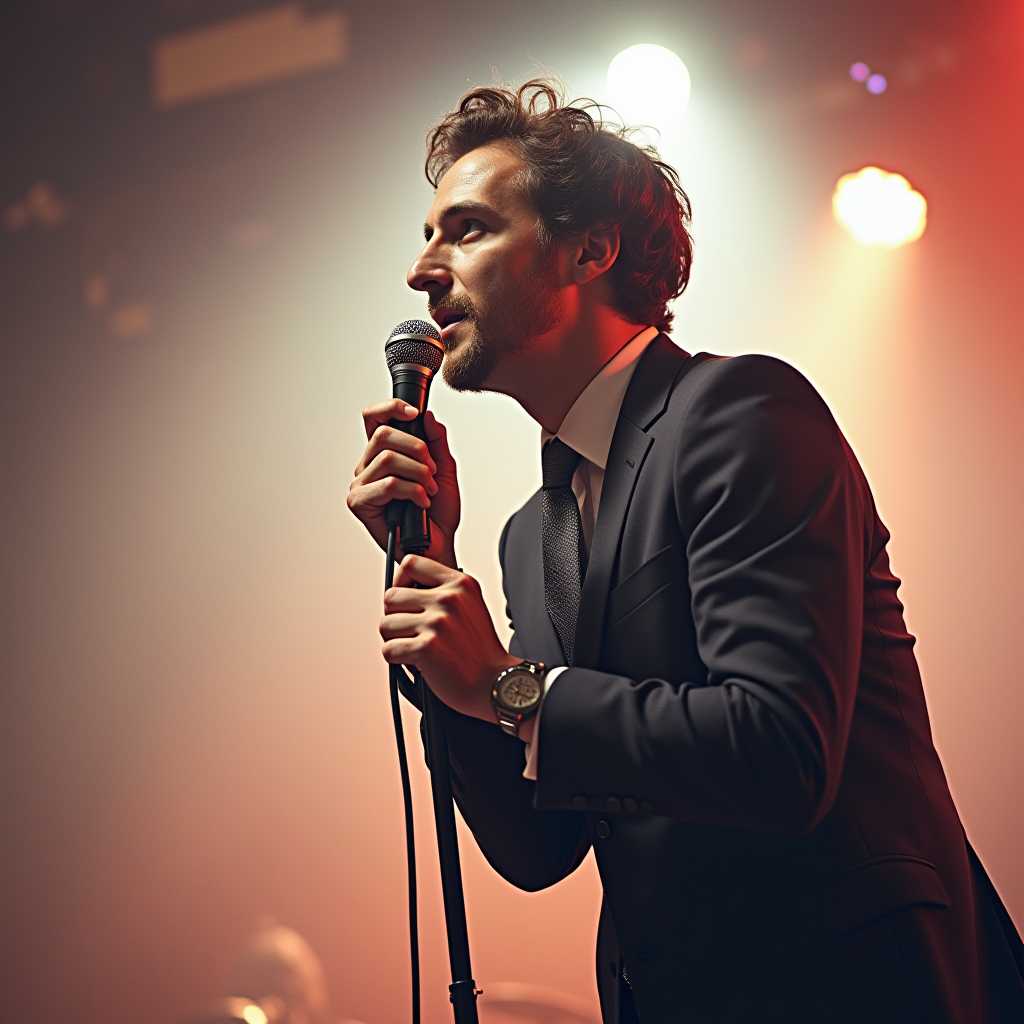} & \includegraphics[width=\imgwidth]{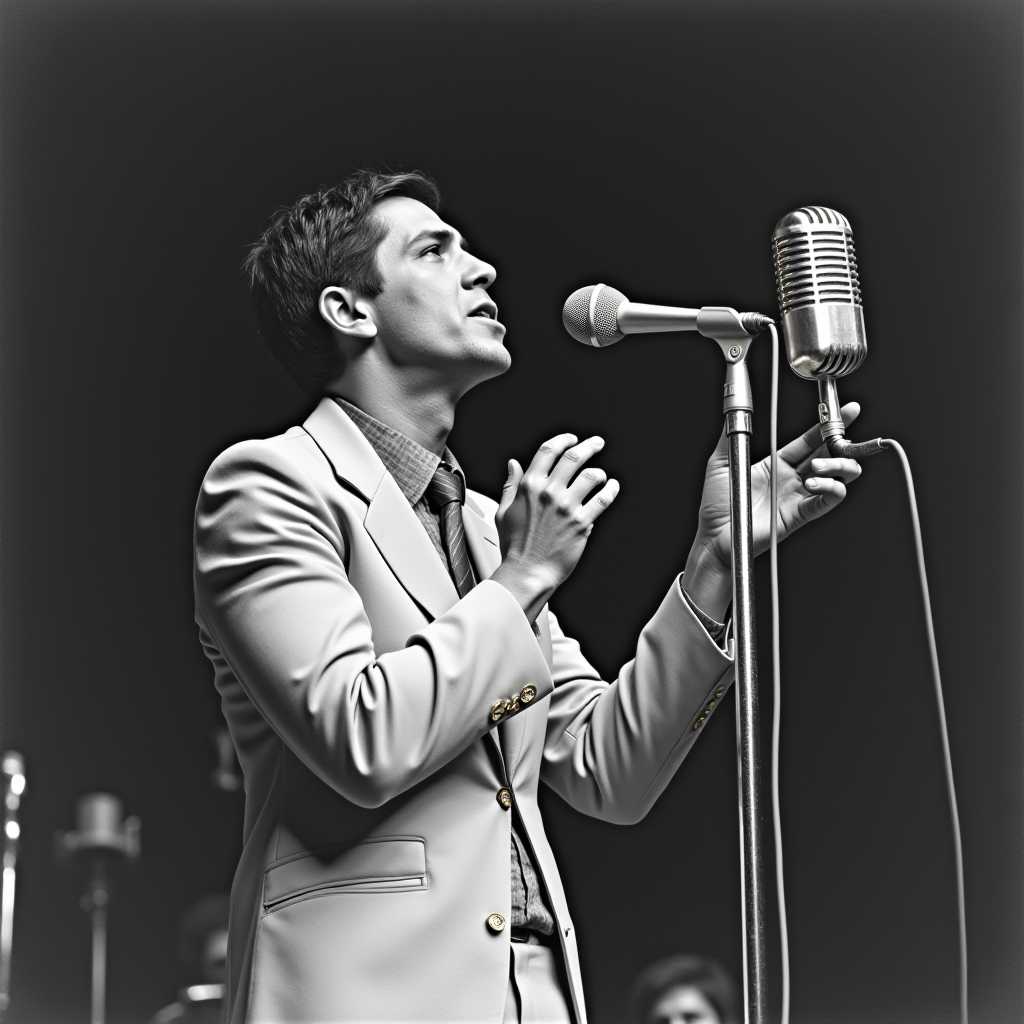} & \includegraphics[width=\imgwidth]{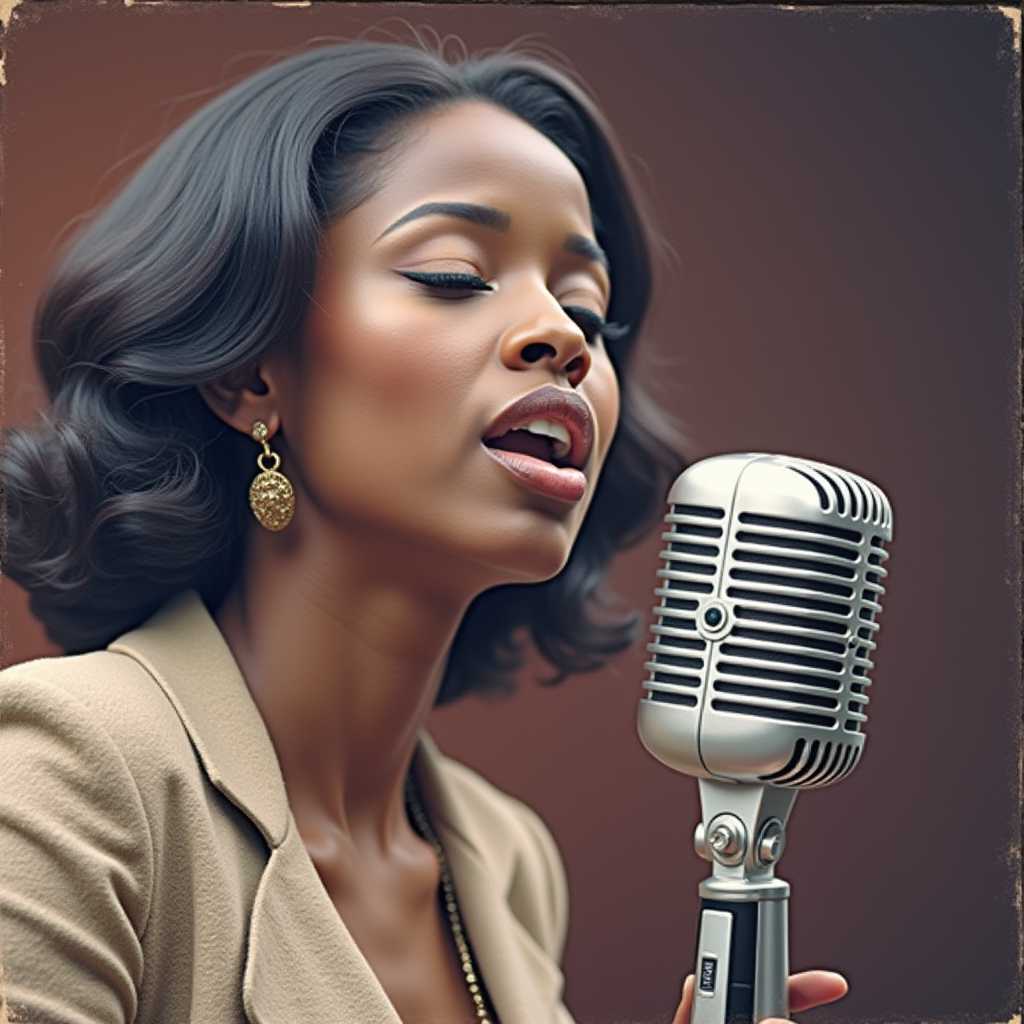} & \includegraphics[width=\imgwidth]{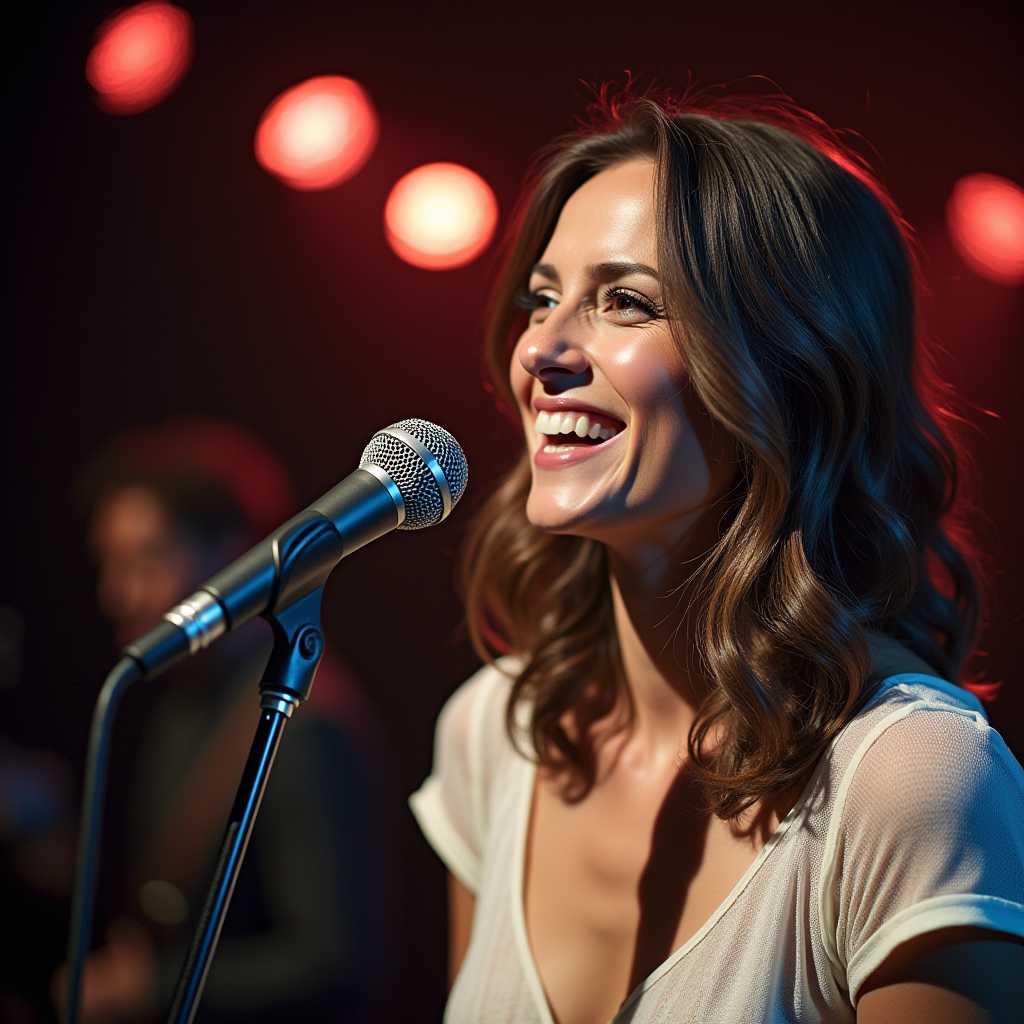} \\
        \multicolumn{9}{c}{\vspace{2pt}\small ``A jazz singer performing on stage with a vintage microphone'' \vspace{8pt}} \\

        \vertlabel{Flux} & \includegraphics[width=\imgwidth]{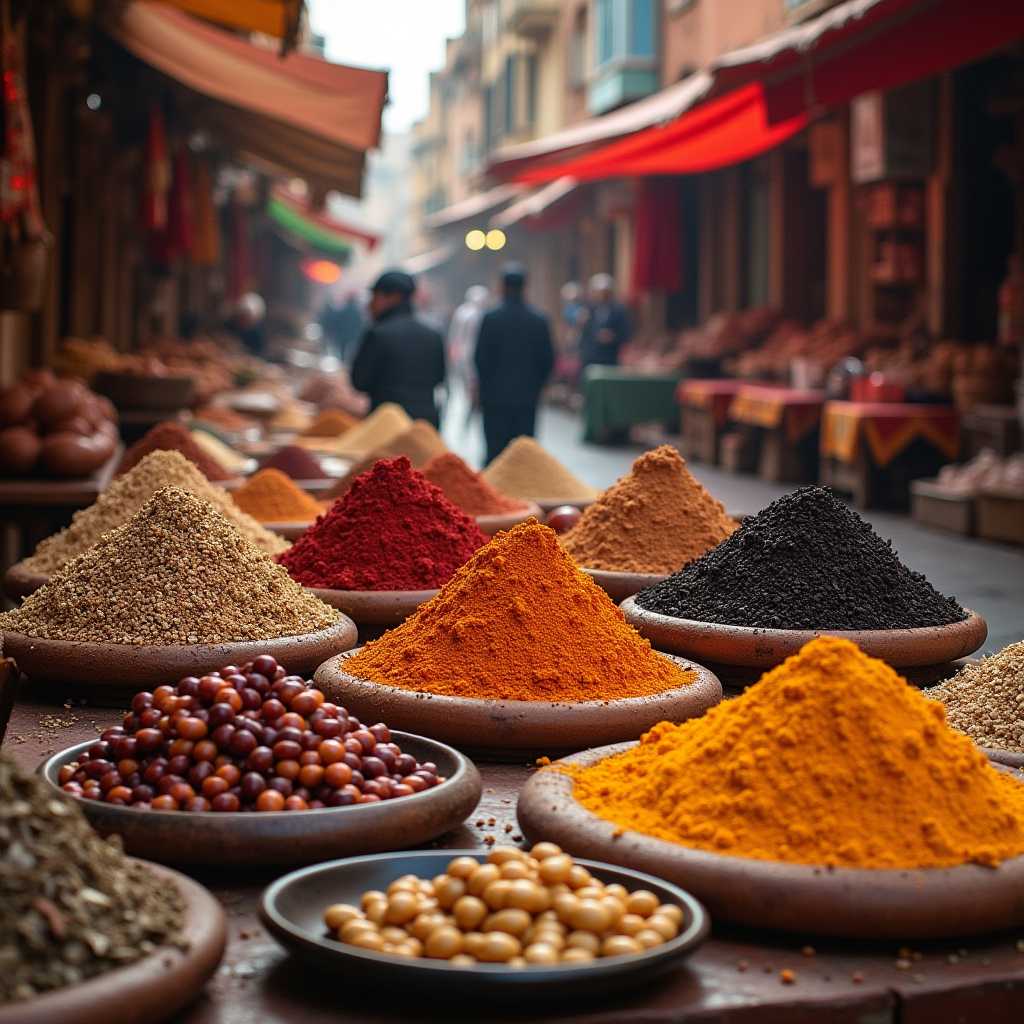} & \includegraphics[width=\imgwidth]{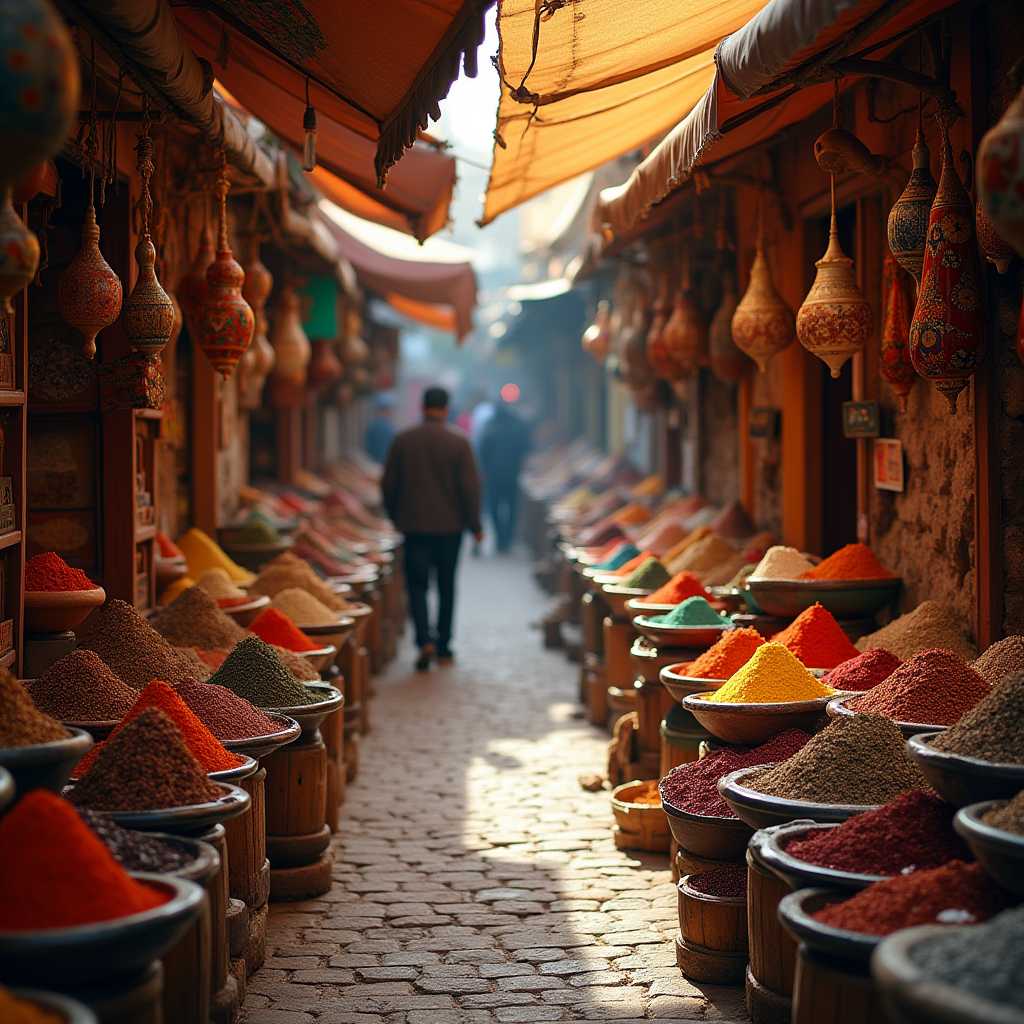} & \includegraphics[width=\imgwidth]{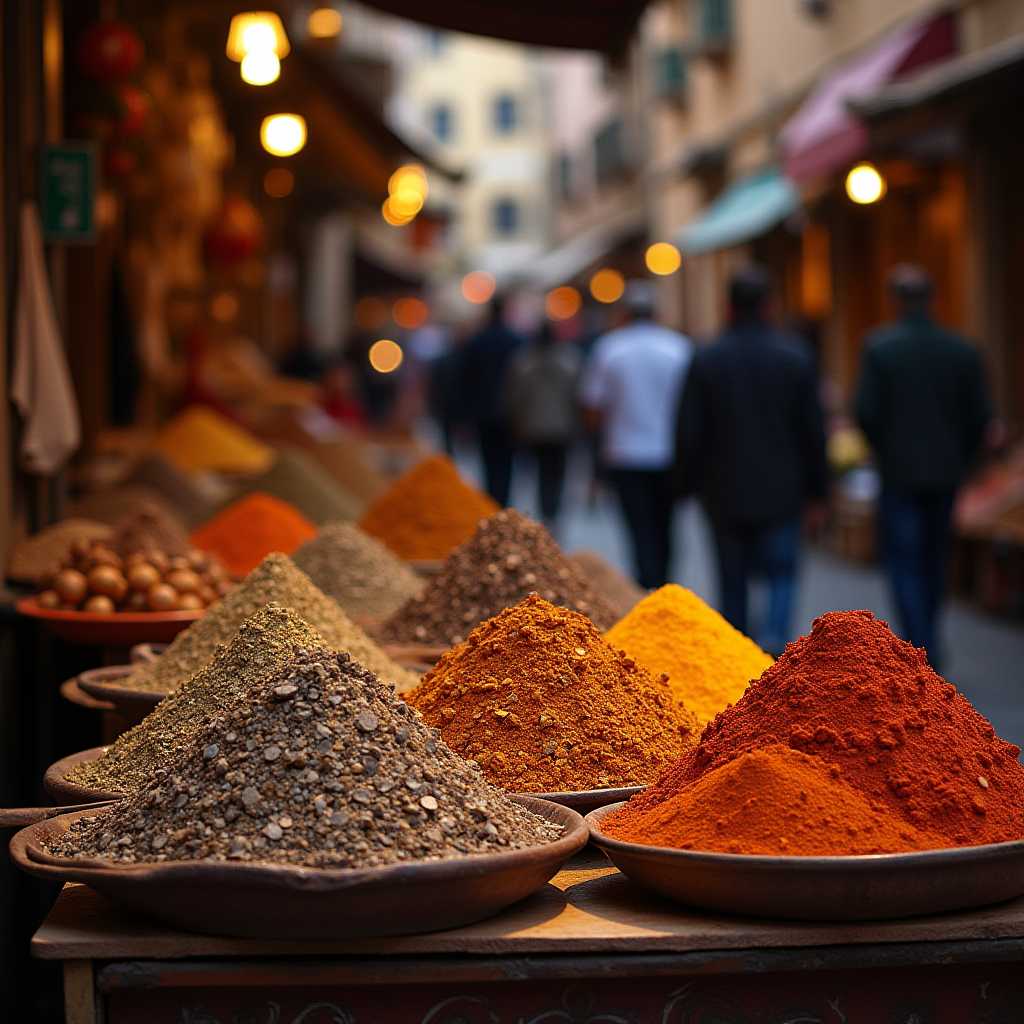} & \includegraphics[width=\imgwidth]{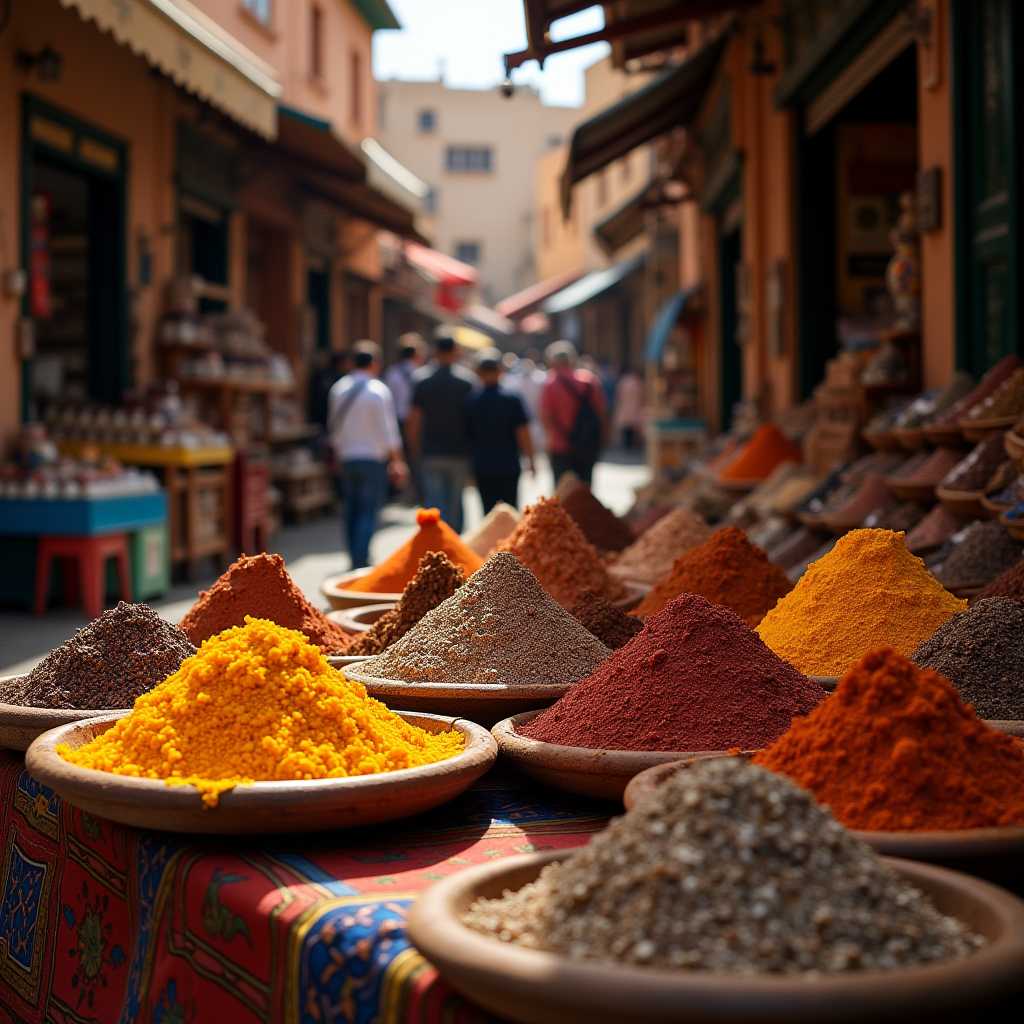} & \includegraphics[width=\imgwidth]{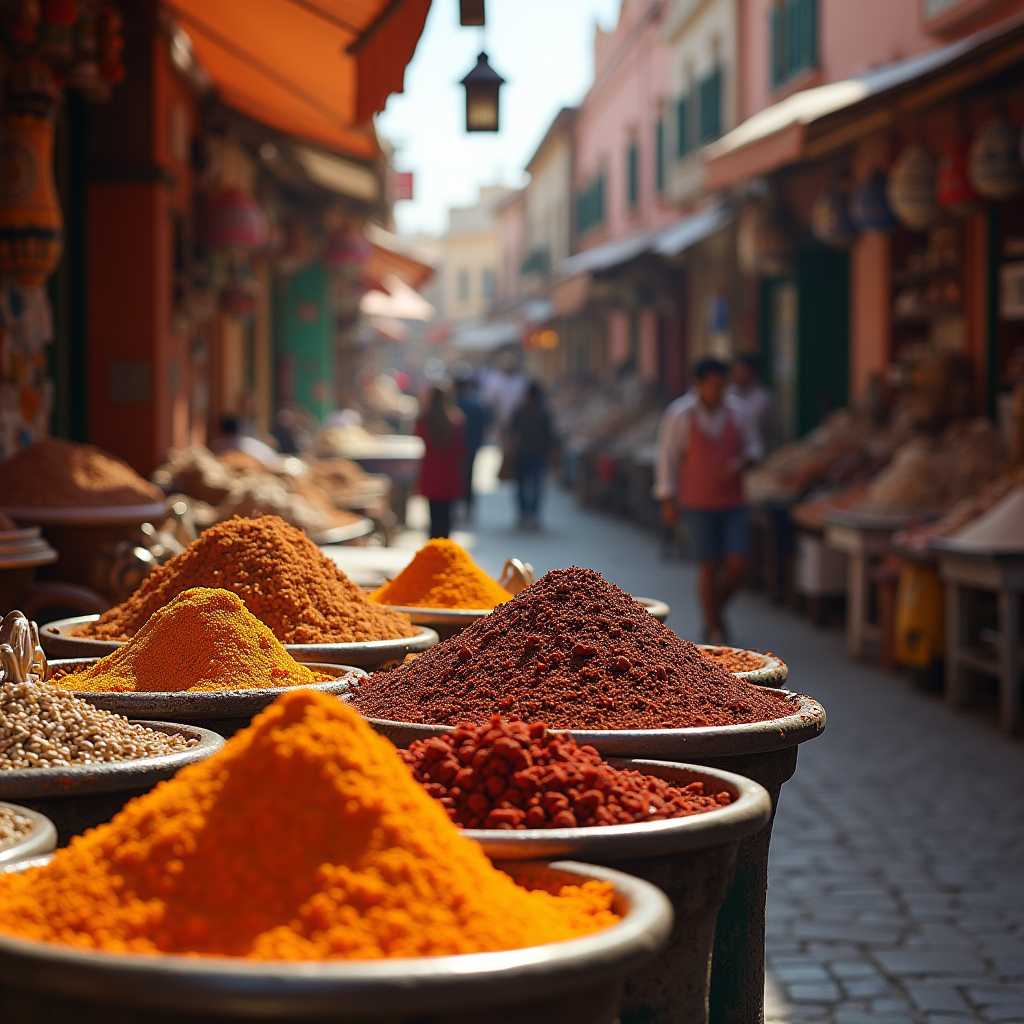} & \includegraphics[width=\imgwidth]{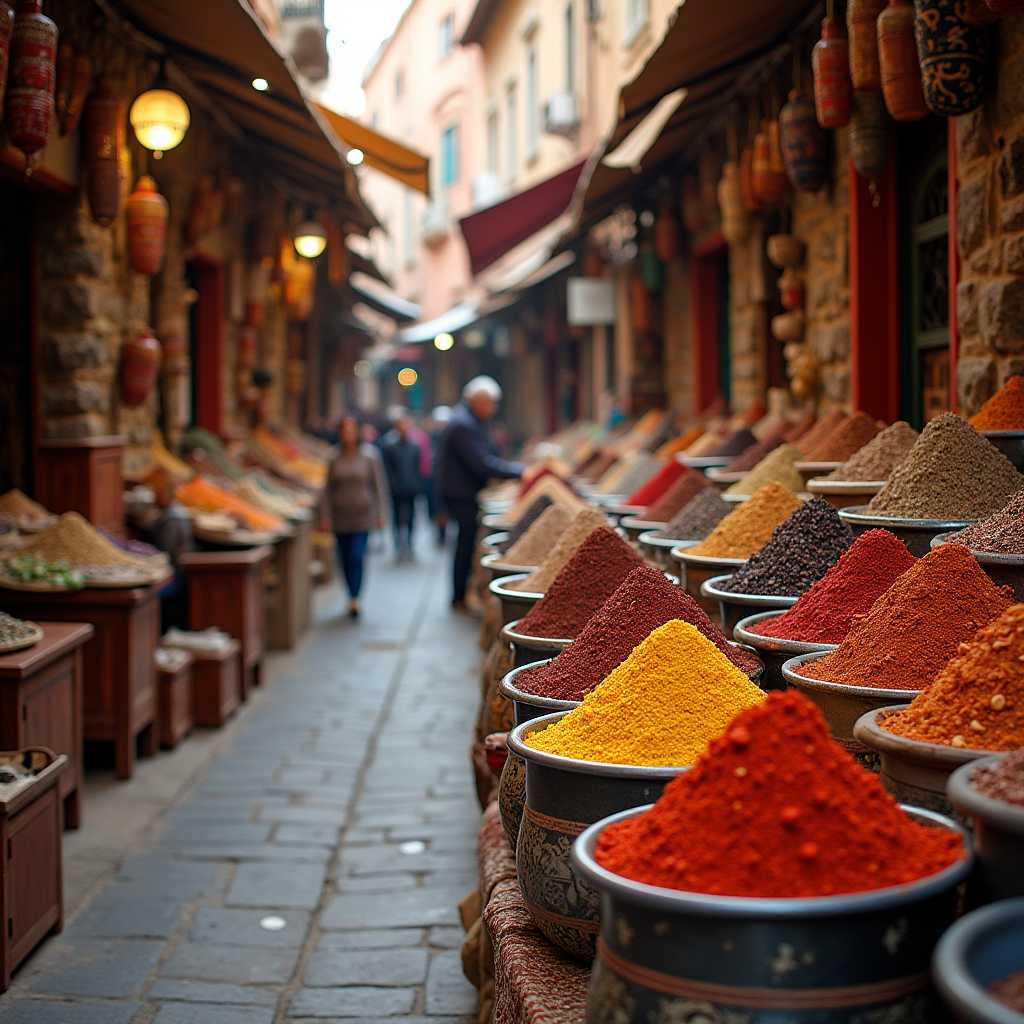} & \includegraphics[width=\imgwidth]{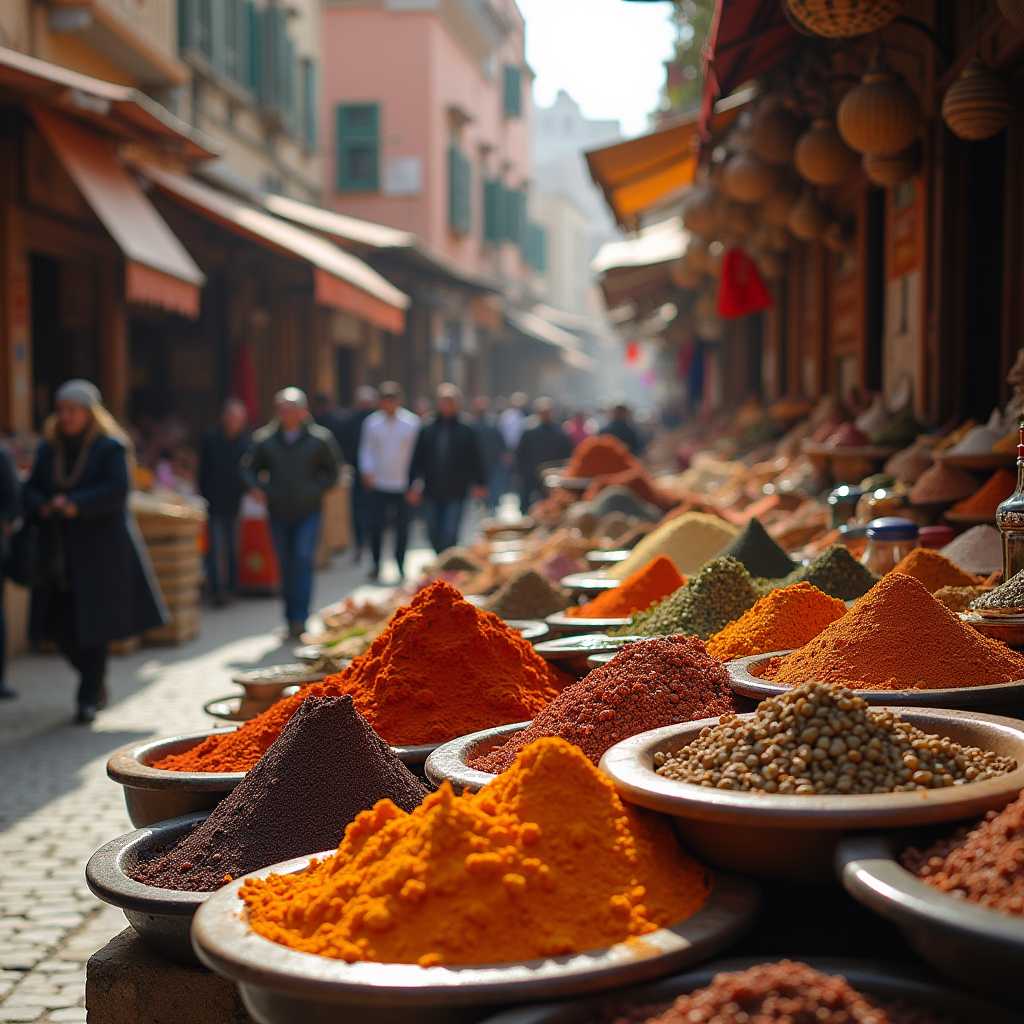} & \includegraphics[width=\imgwidth]{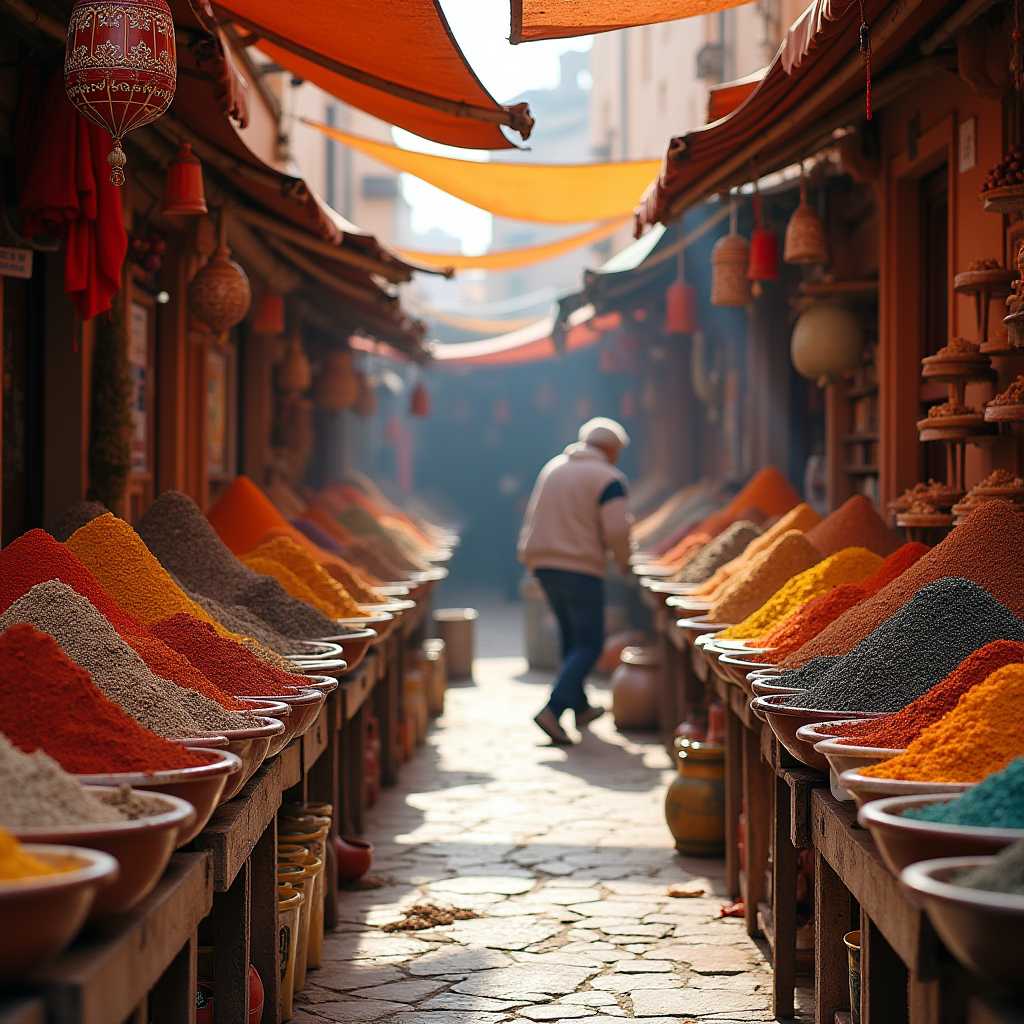} \\[-1pt]
        \vertlabel{Ours} & \includegraphics[width=\imgwidth]{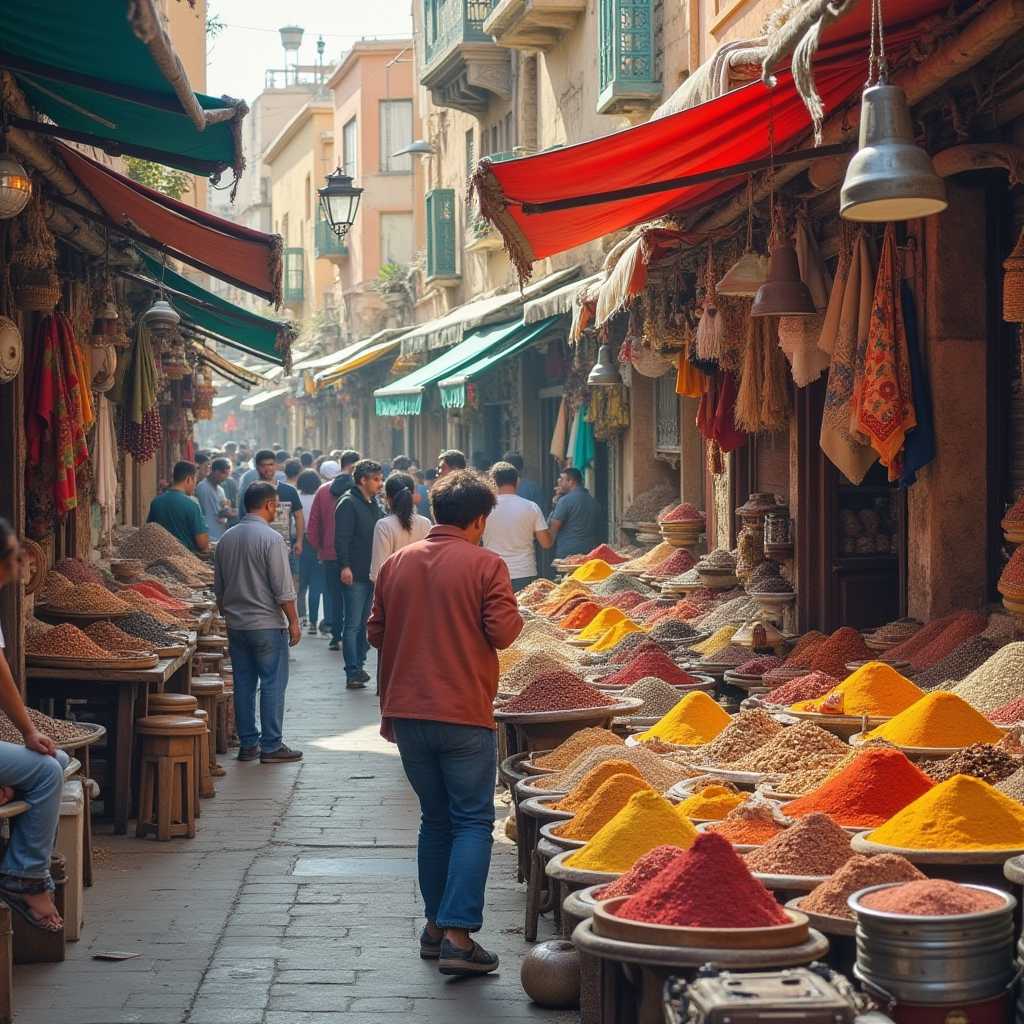} & \includegraphics[width=\imgwidth]{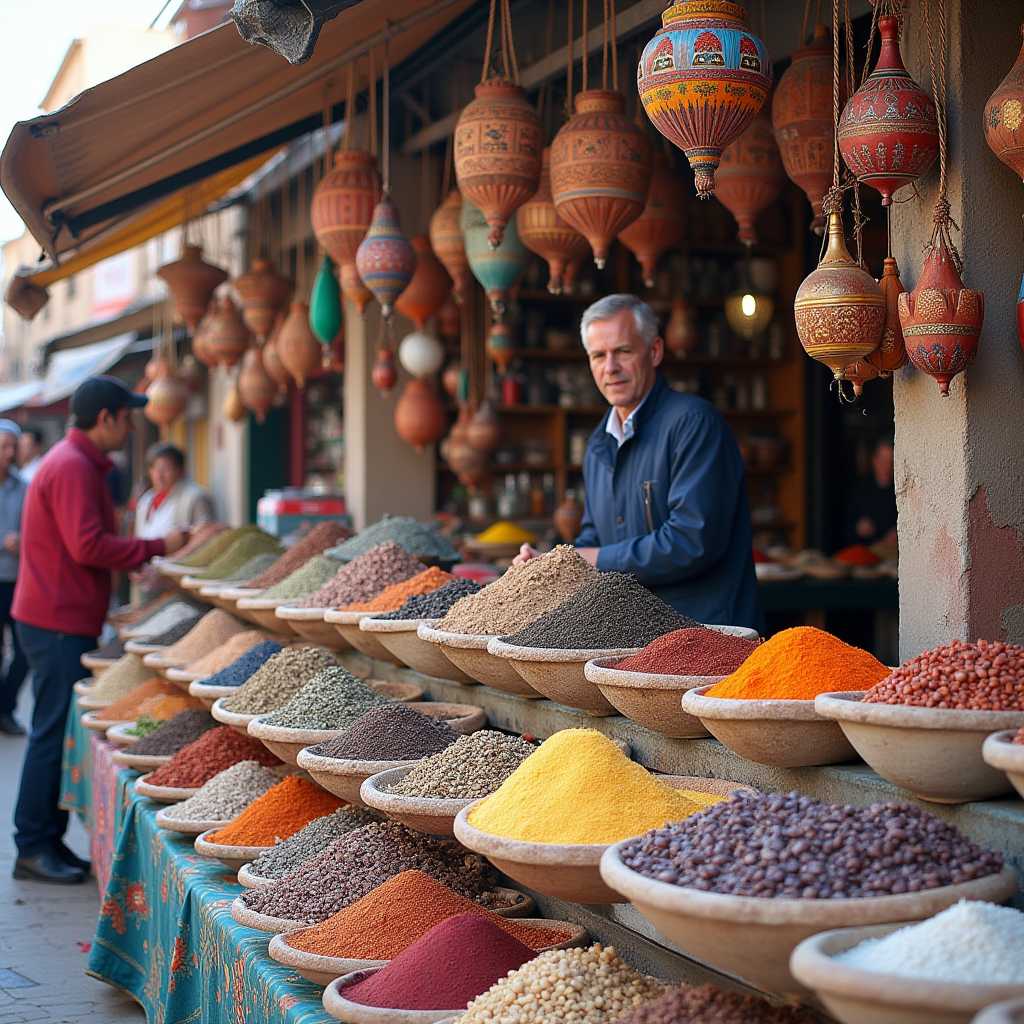} & \includegraphics[width=\imgwidth]{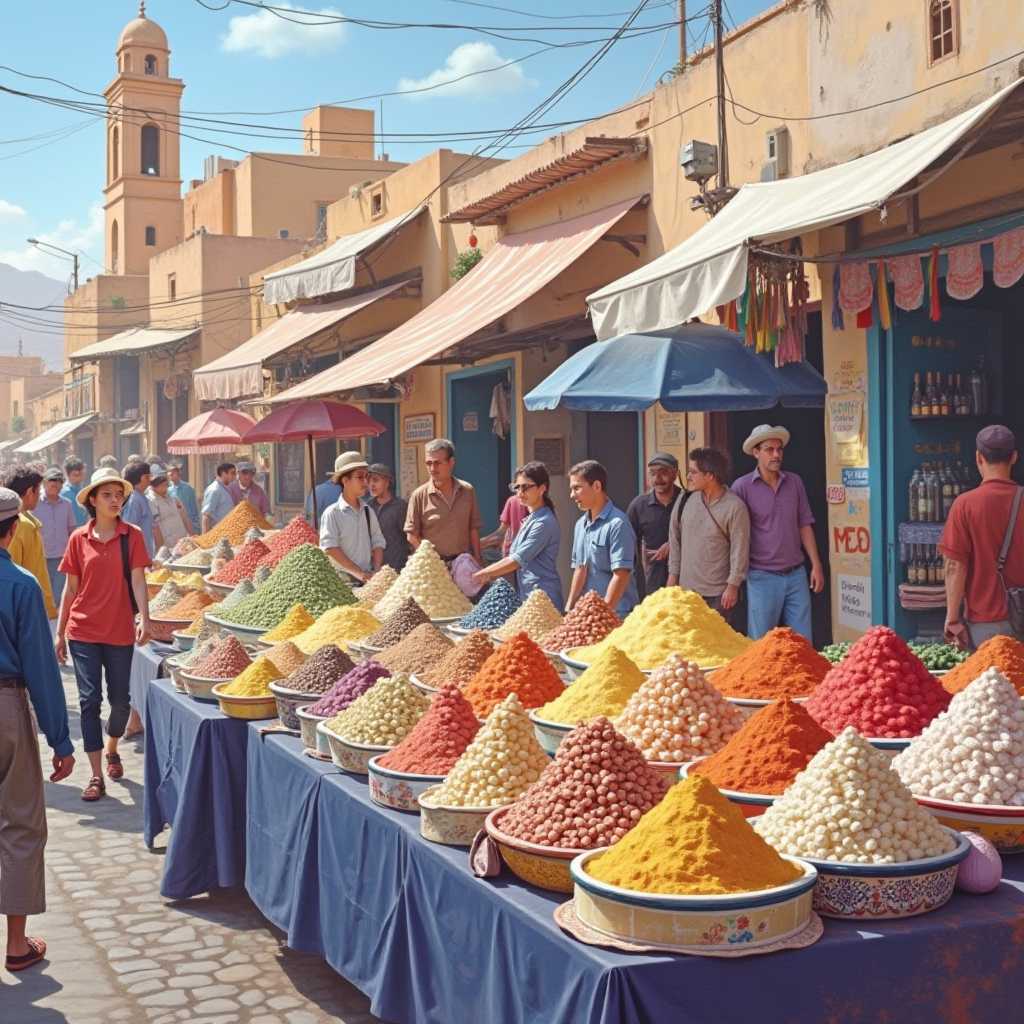} & \includegraphics[width=\imgwidth]{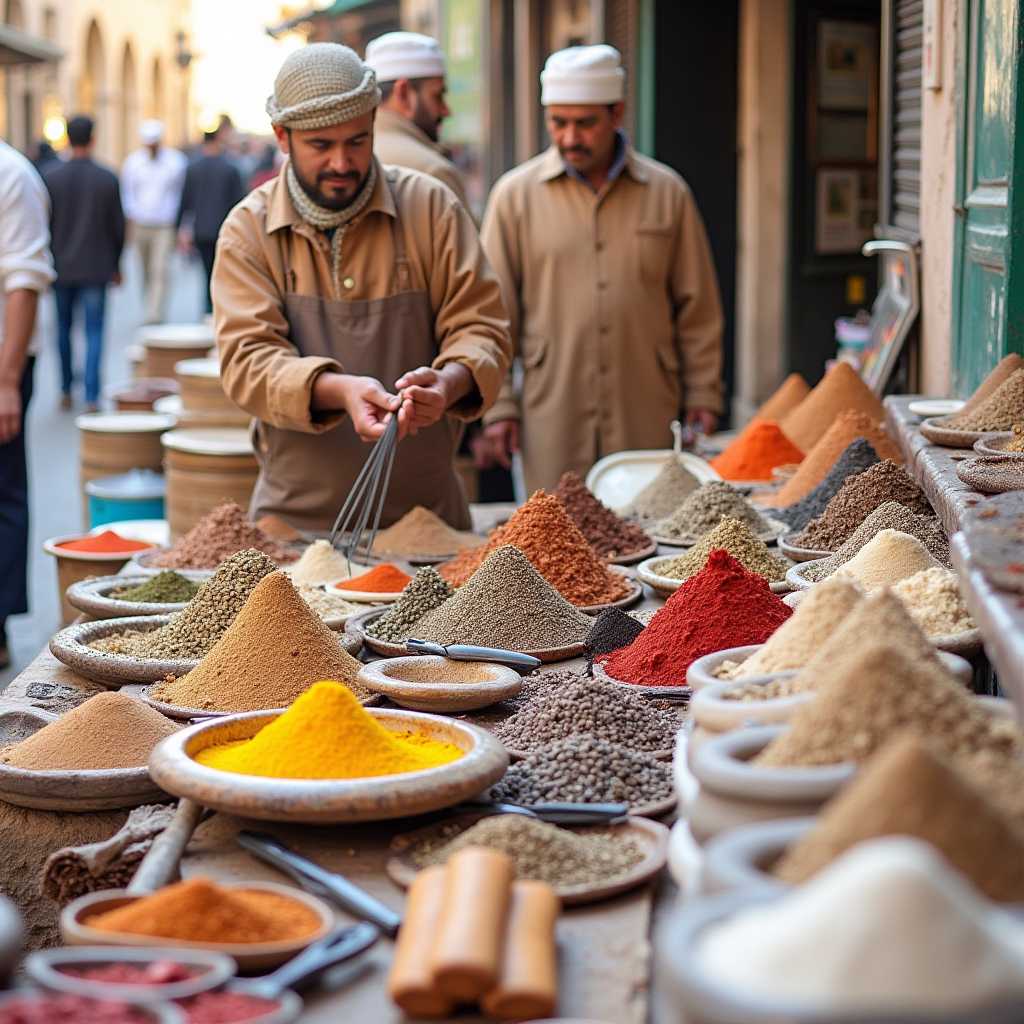} & \includegraphics[width=\imgwidth]{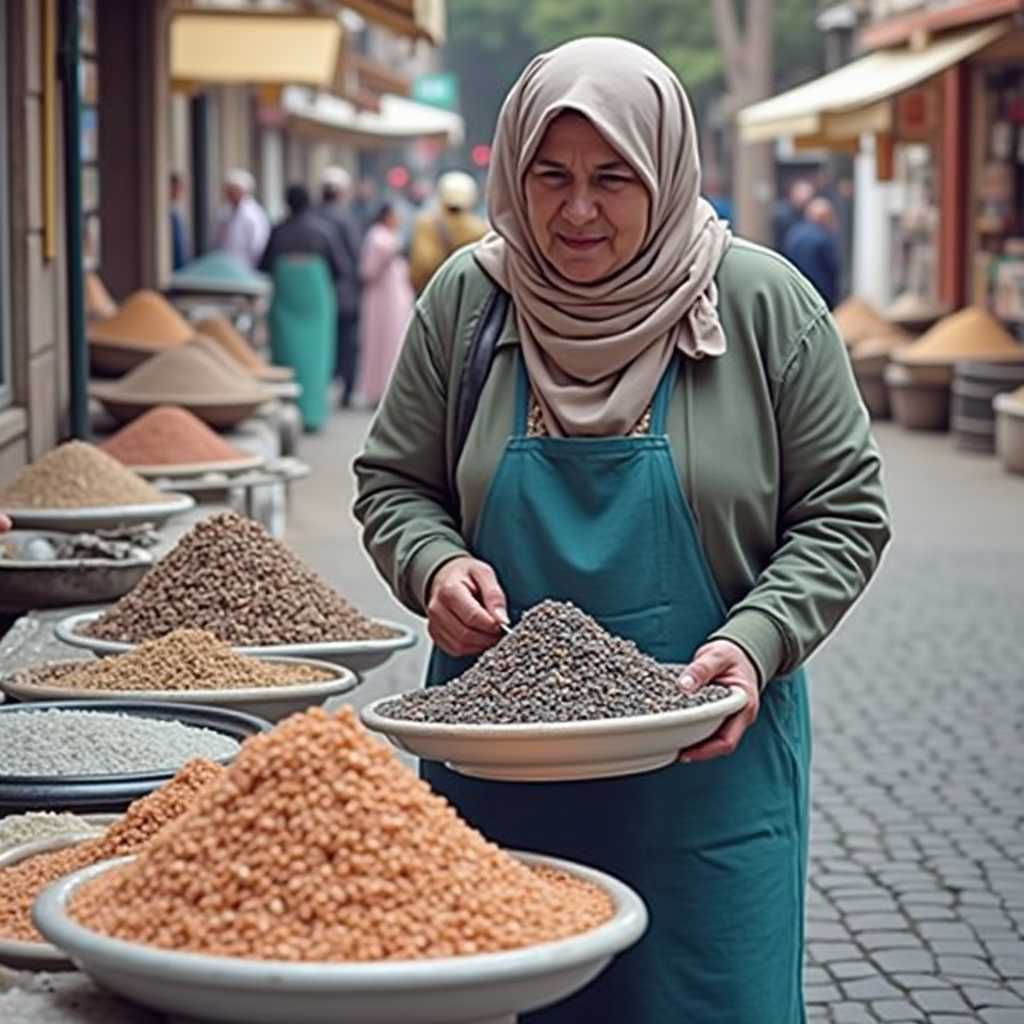} & \includegraphics[width=\imgwidth]{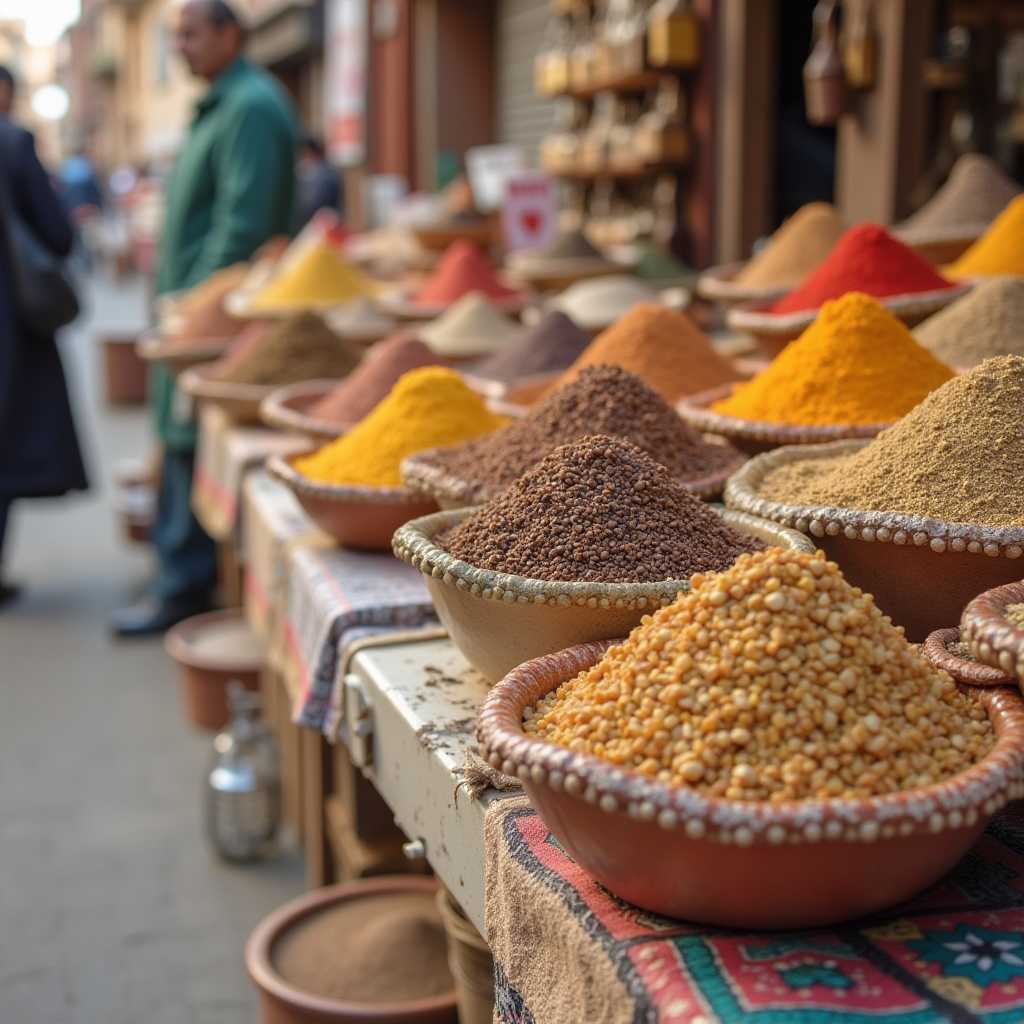} & \includegraphics[width=\imgwidth]{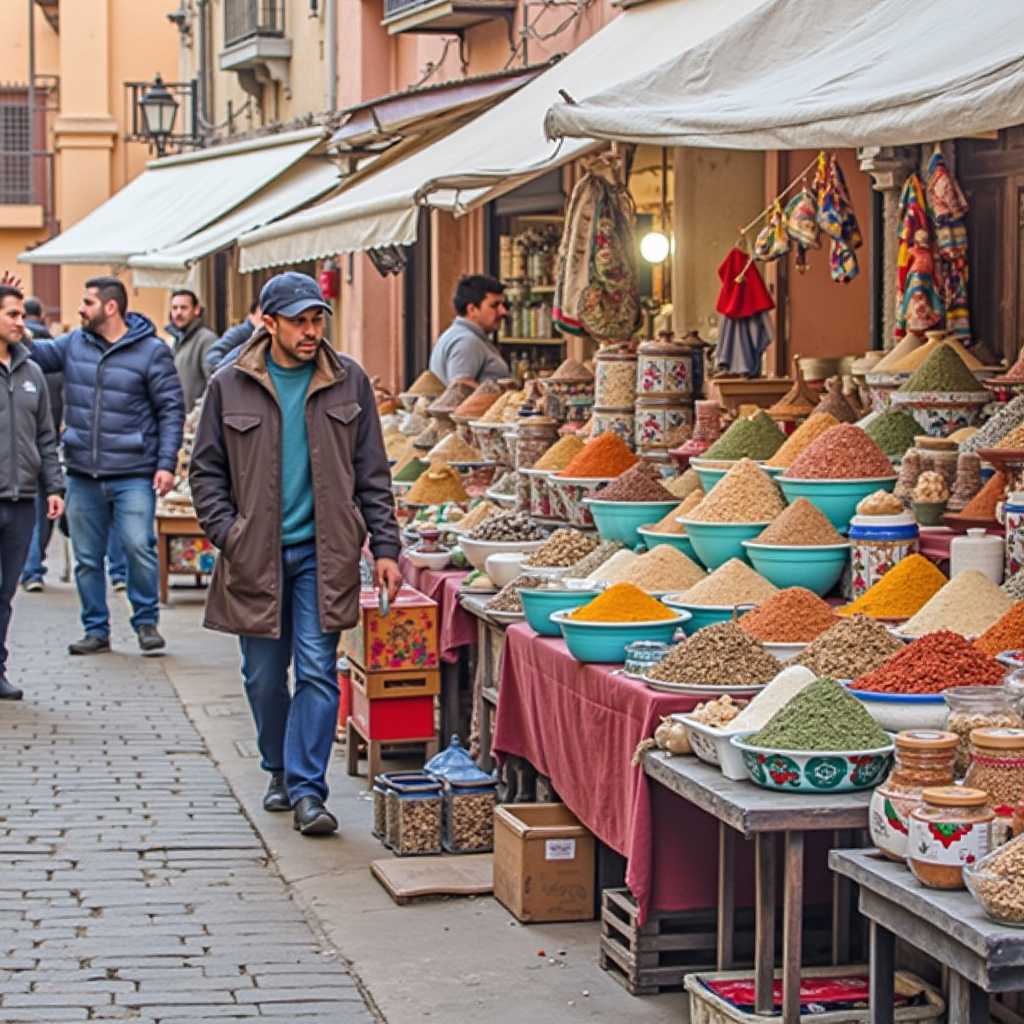} & \includegraphics[width=\imgwidth]{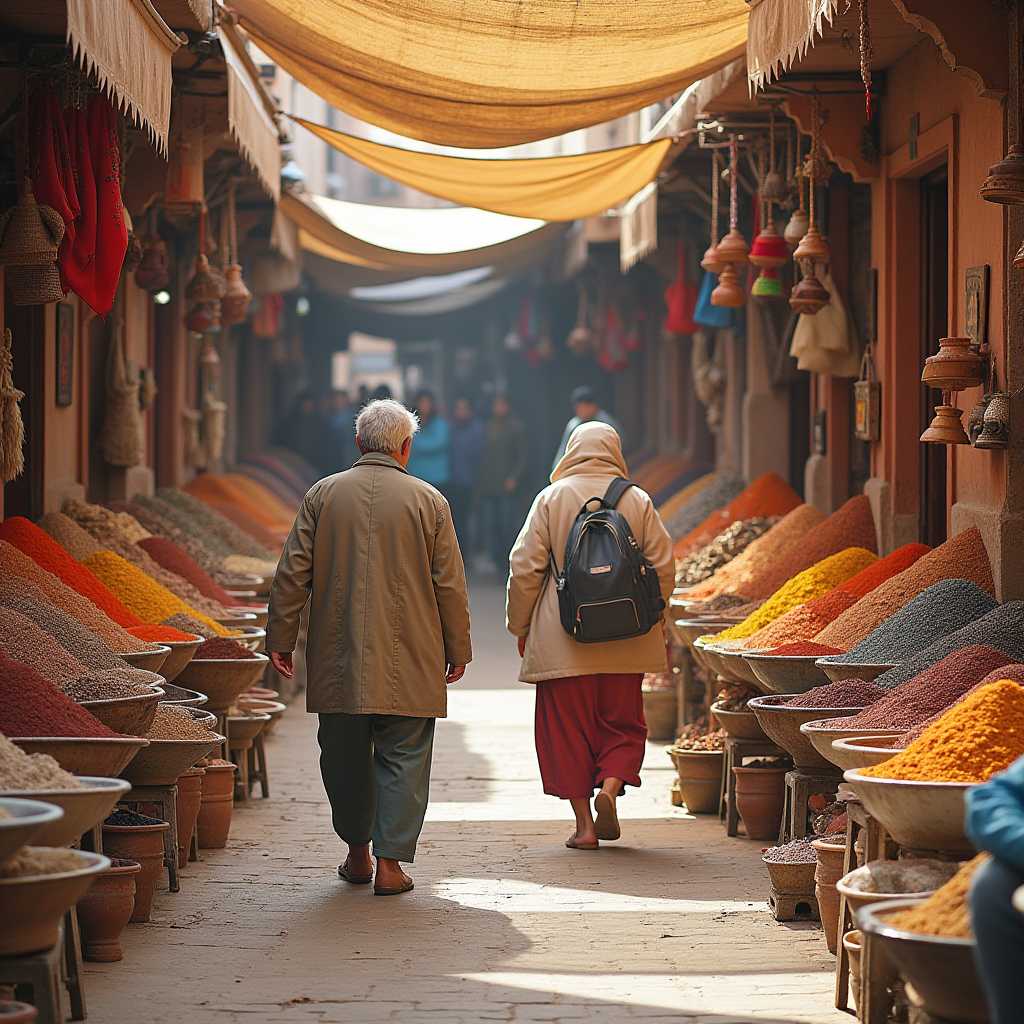} \\
        \multicolumn{9}{c}{\vspace{2pt}\small ``A bustling street market in Morocco with colorful spices'' \vspace{8pt}} \\

    \end{tabular}
    \caption{\textbf{Additional qualitative results on Flux-dev.} All batches were generated using the same random seed initialization.}\label{fig:extra_res1}
\end{figure*}

\begin{figure*}
    \centering
    \setlength{\tabcolsep}{0.5pt} \renewcommand{\arraystretch}{0.5} \newcommand{\imgwidth}{0.12\textwidth}
    \newcommand{\vertlabel}[1]{\raisebox{2.5em}{\rotatebox{90}{\scriptsize\textbf{#1}}}}

    \begin{tabular}{c c c c c c c c c}

        \vertlabel{Flux} & \includegraphics[width=\imgwidth]{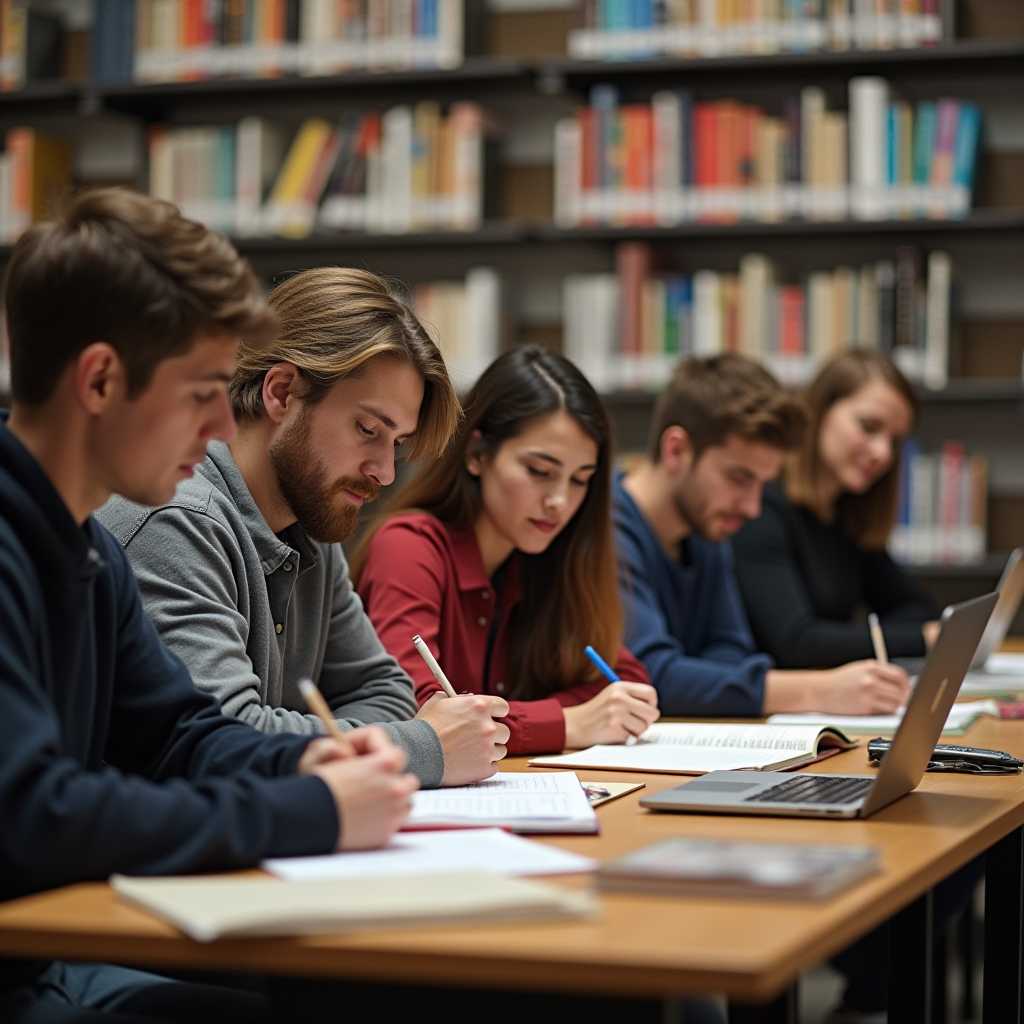} & \includegraphics[width=\imgwidth]{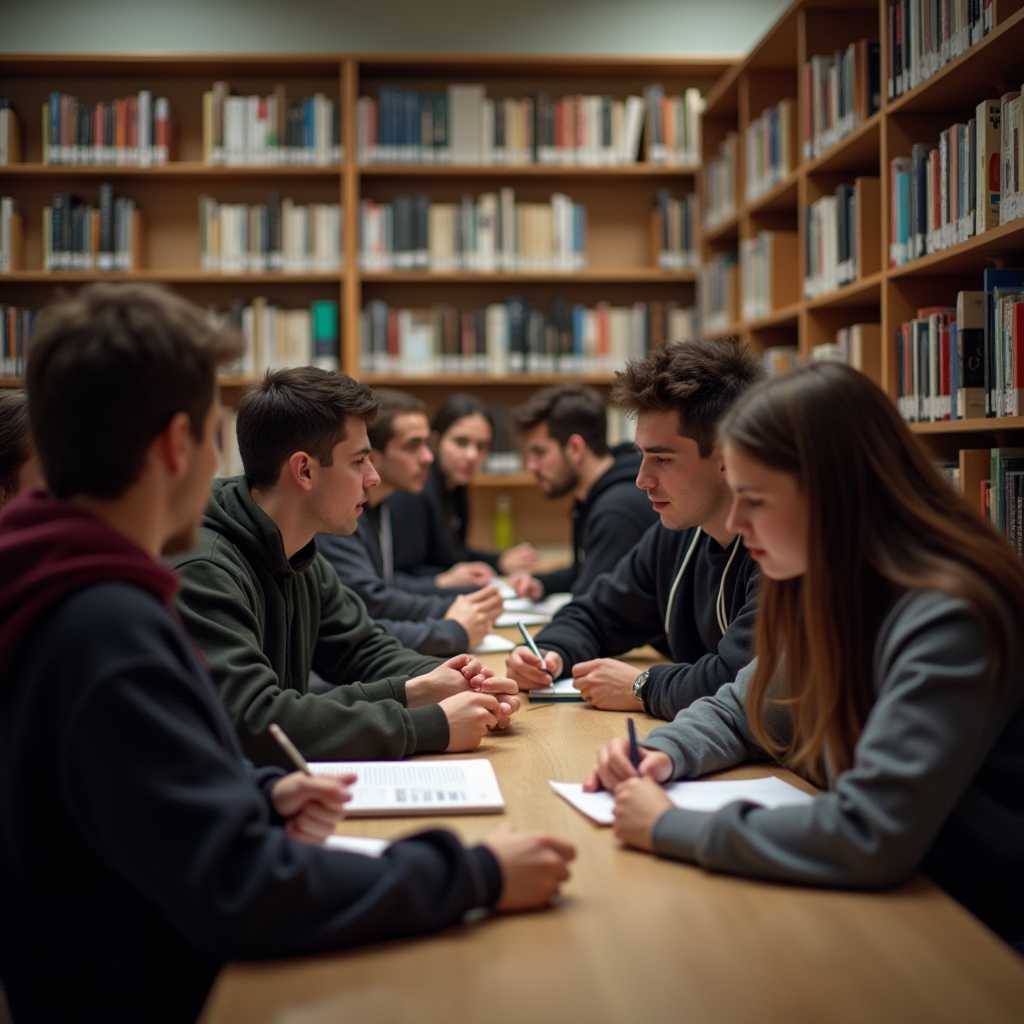} & \includegraphics[width=\imgwidth]{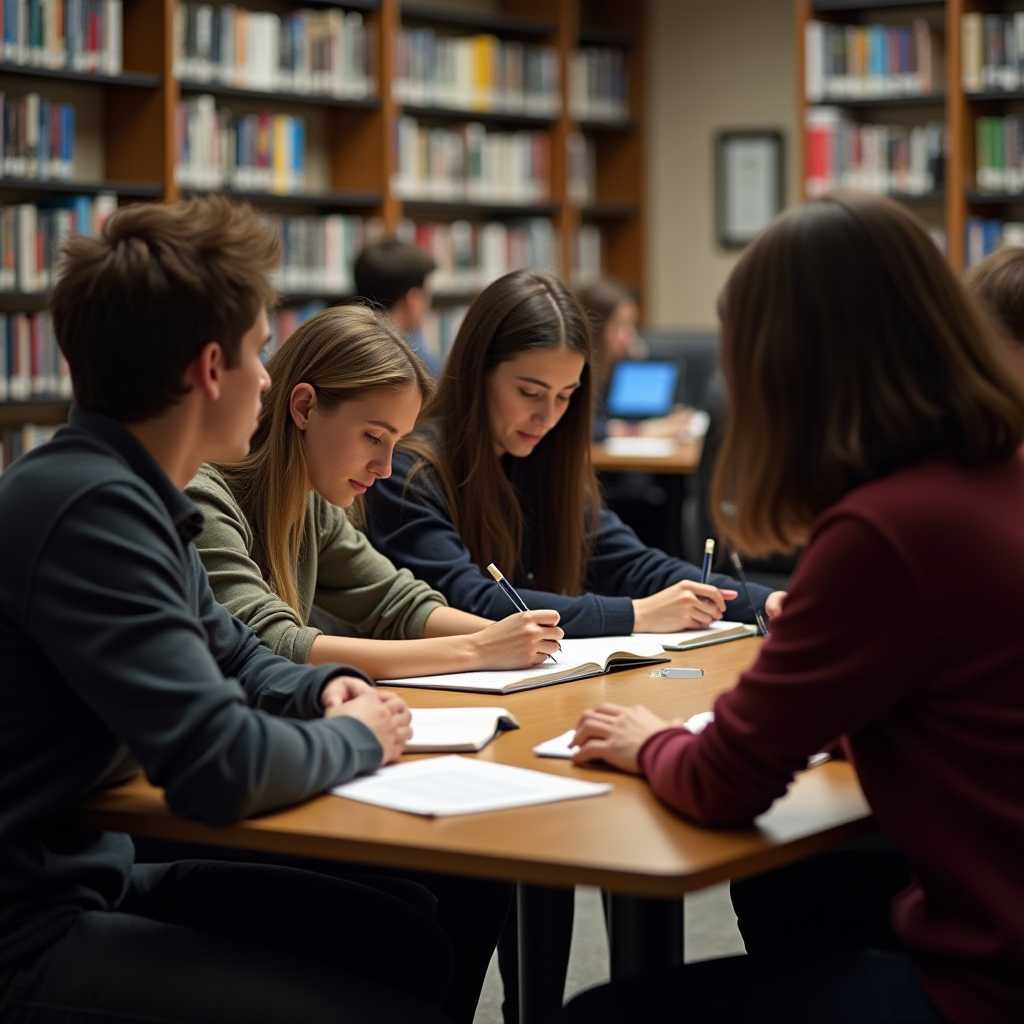} & \includegraphics[width=\imgwidth]{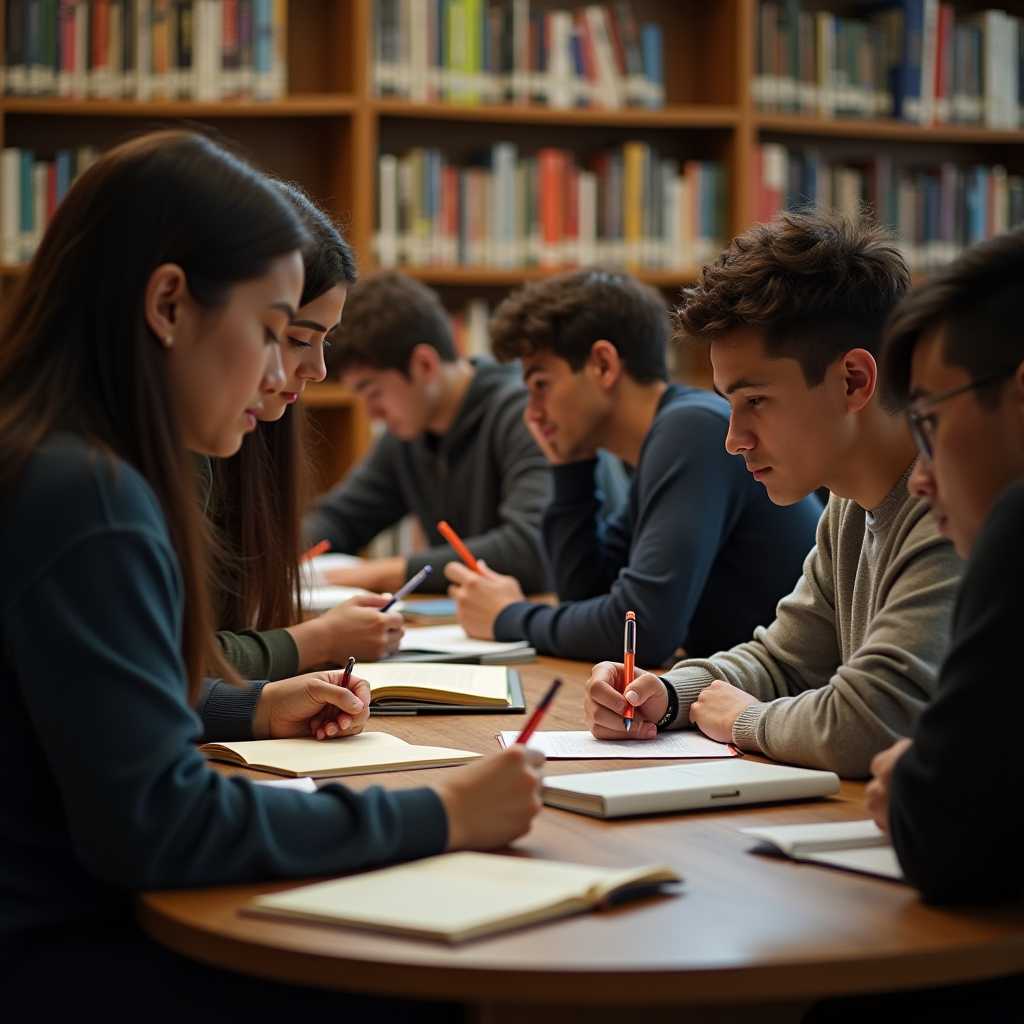} & \includegraphics[width=\imgwidth]{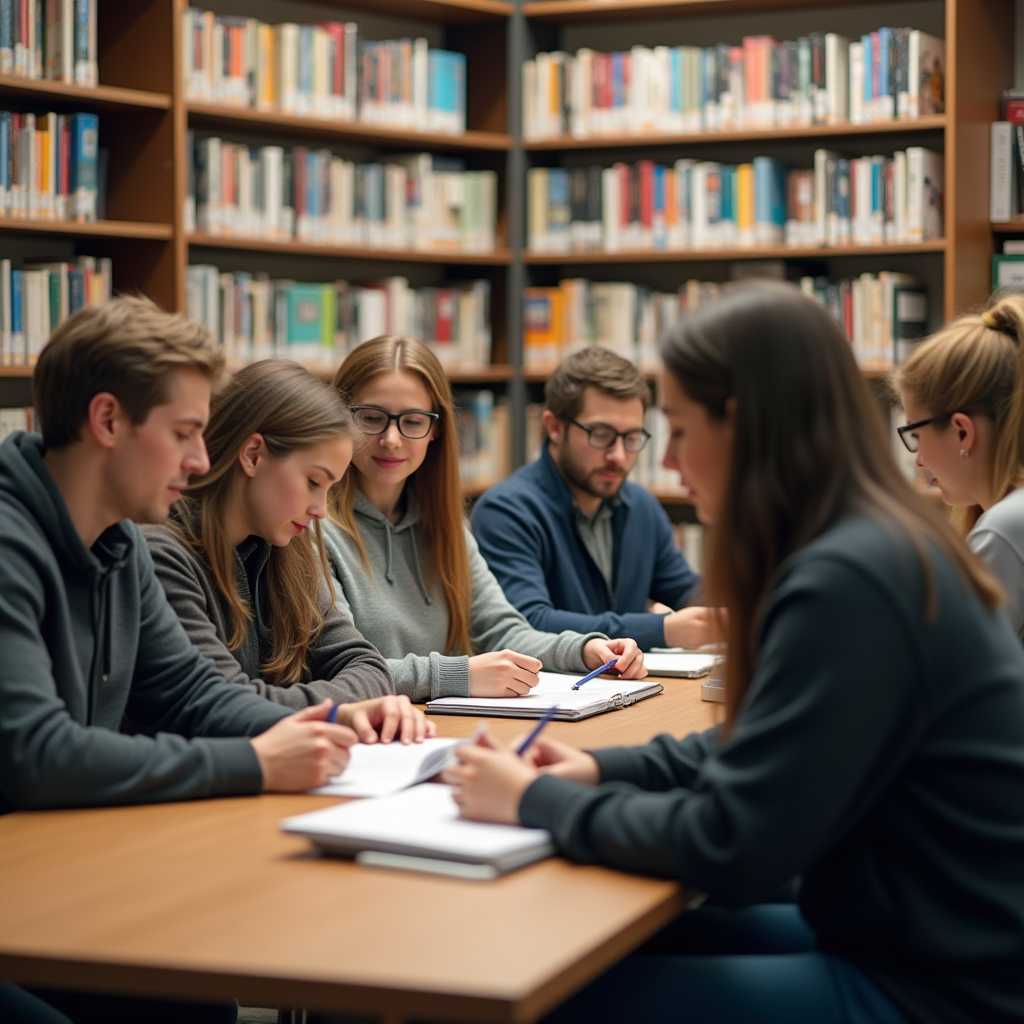} & \includegraphics[width=\imgwidth]{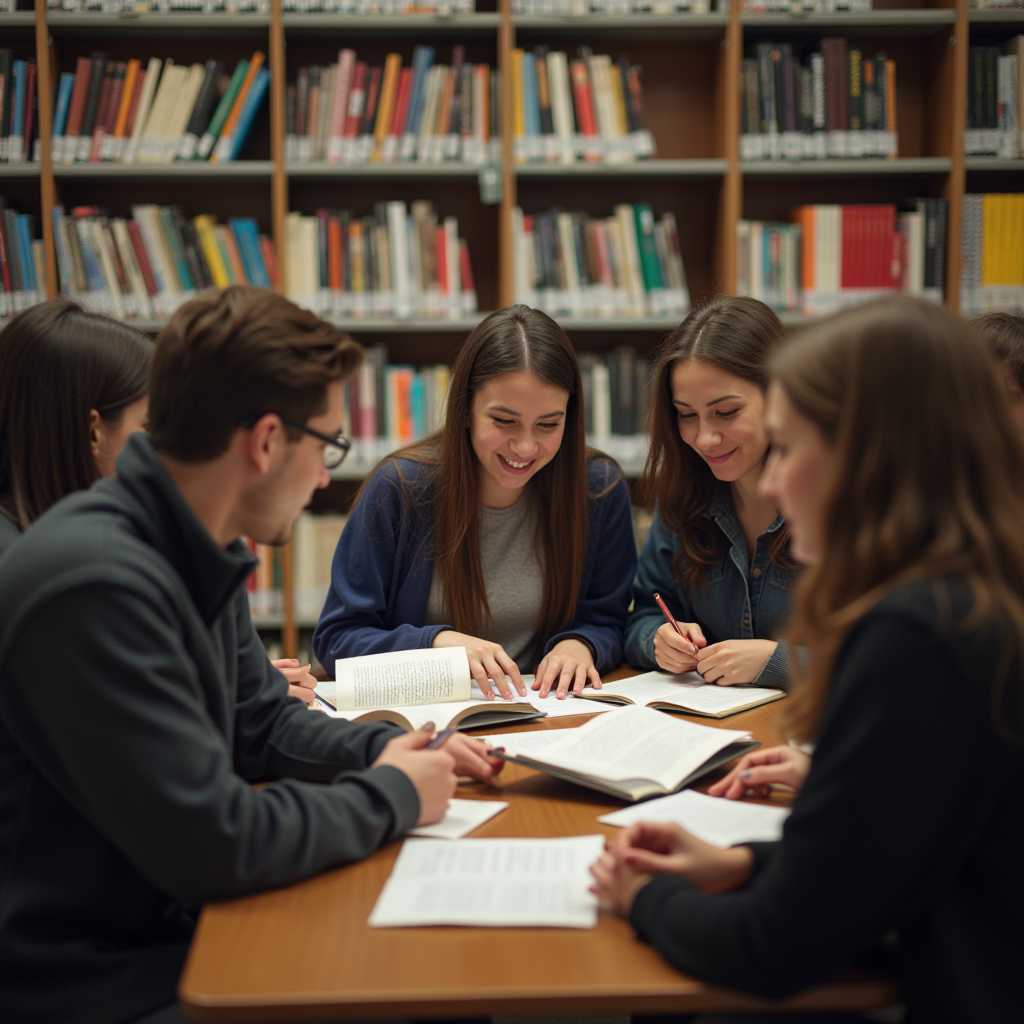} & \includegraphics[width=\imgwidth]{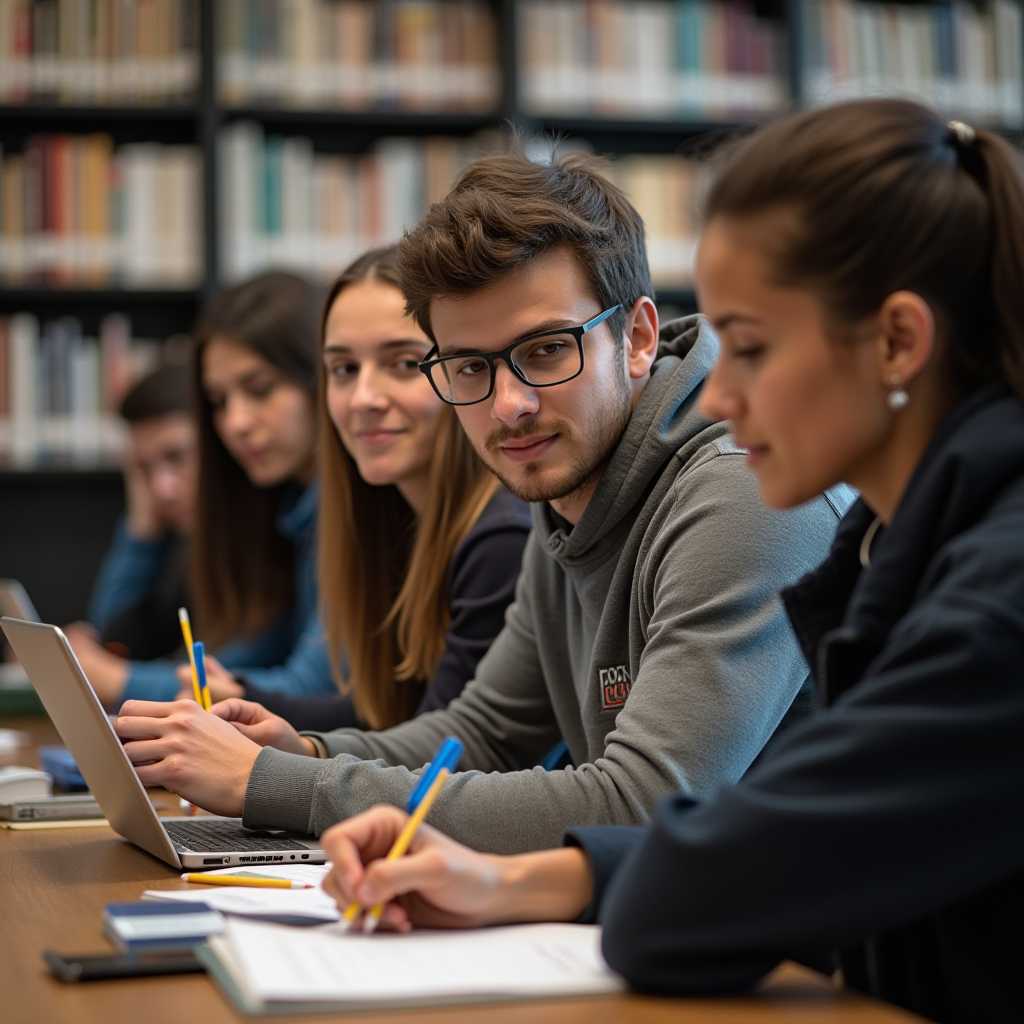} & \includegraphics[width=\imgwidth]{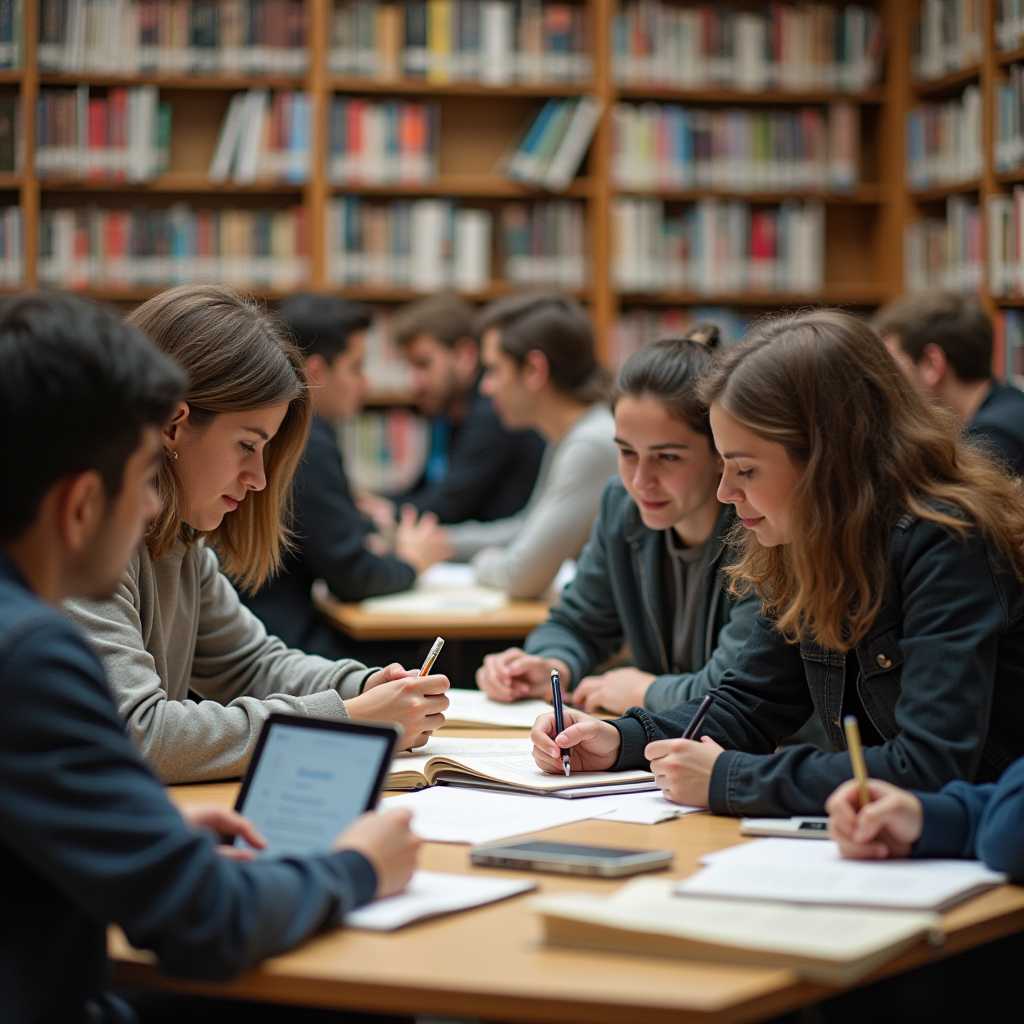} \\[-1pt]
        \vertlabel{Ours} & \includegraphics[width=\imgwidth]{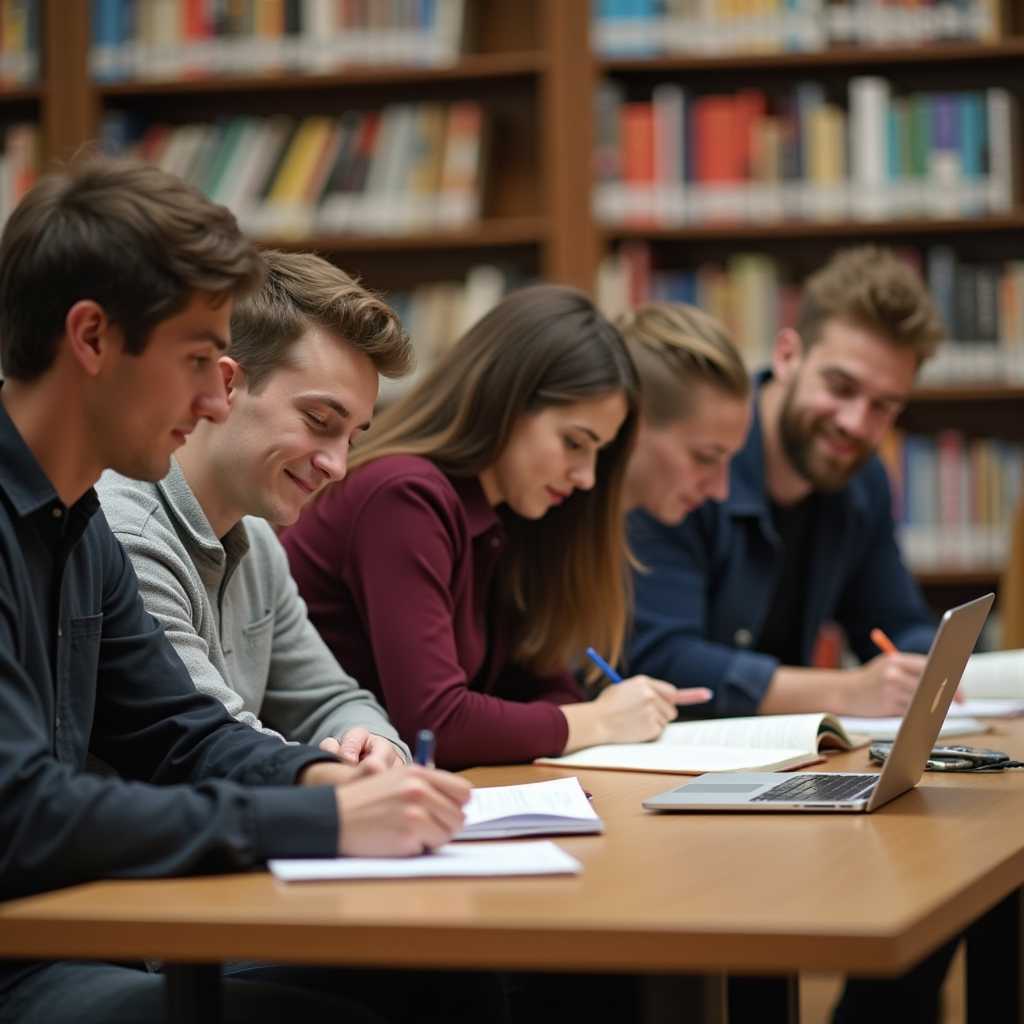} & \includegraphics[width=\imgwidth]{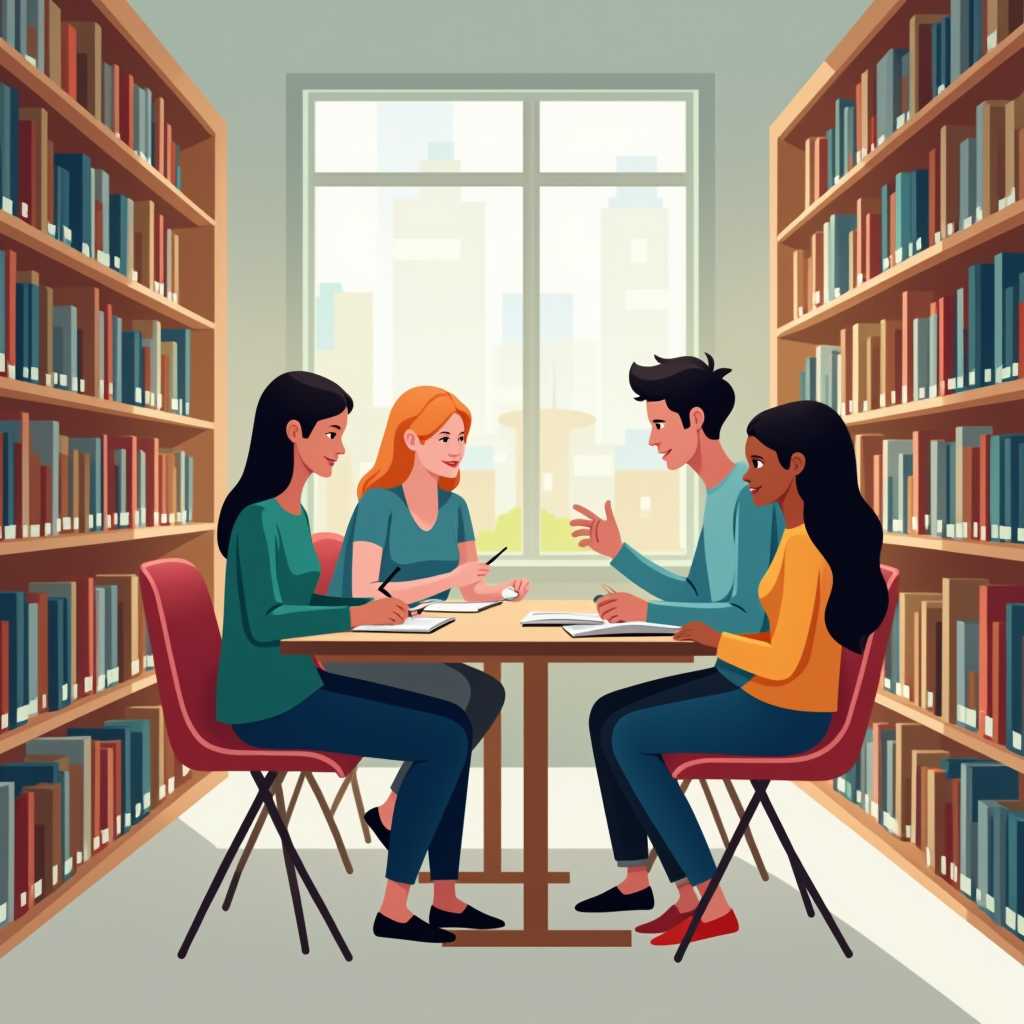} & \includegraphics[width=\imgwidth]{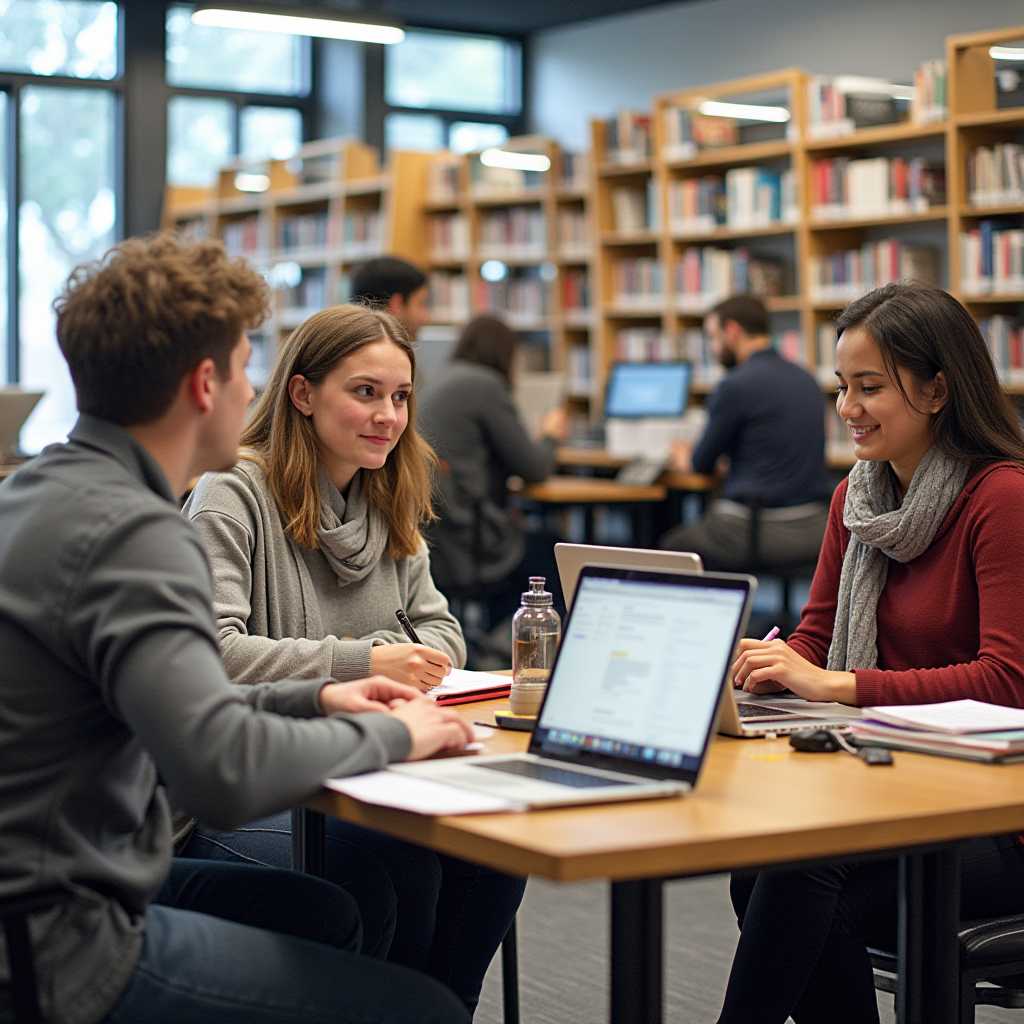} & \includegraphics[width=\imgwidth]{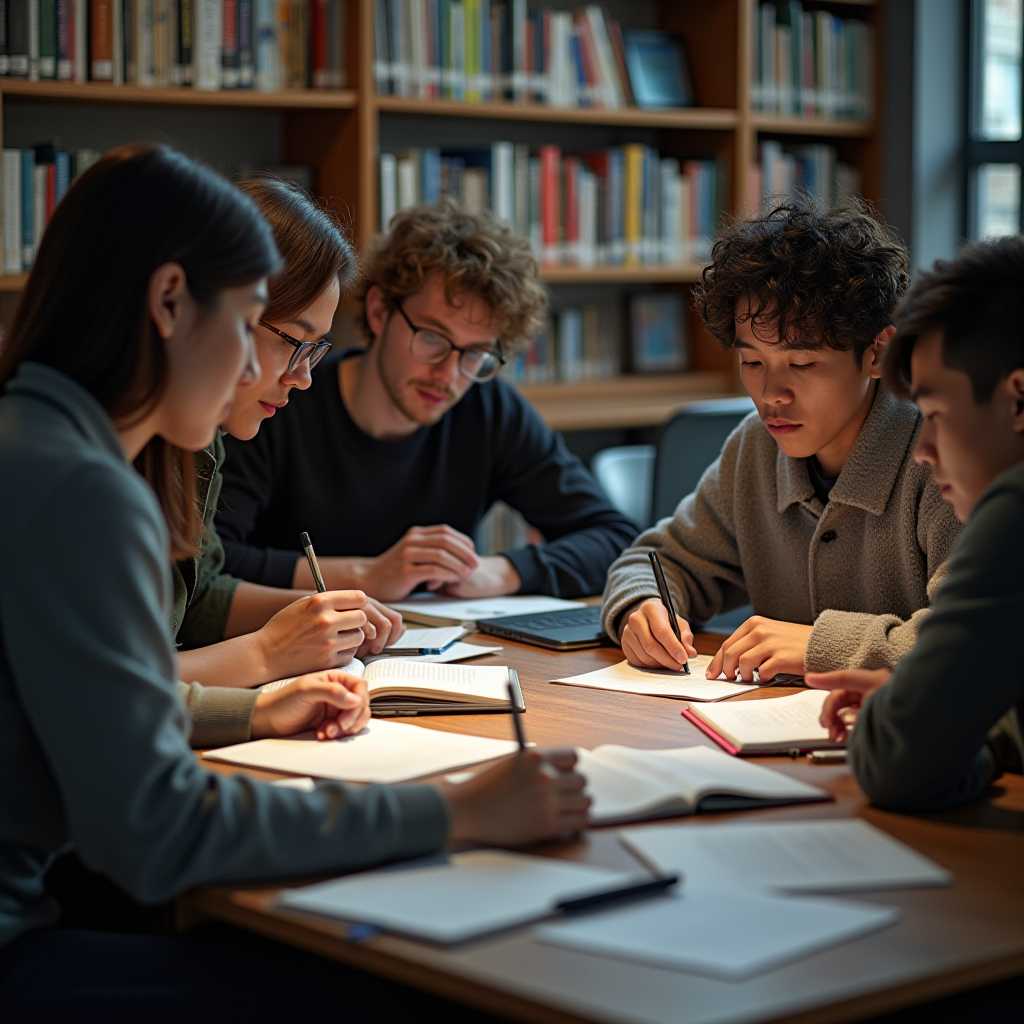} & \includegraphics[width=\imgwidth]{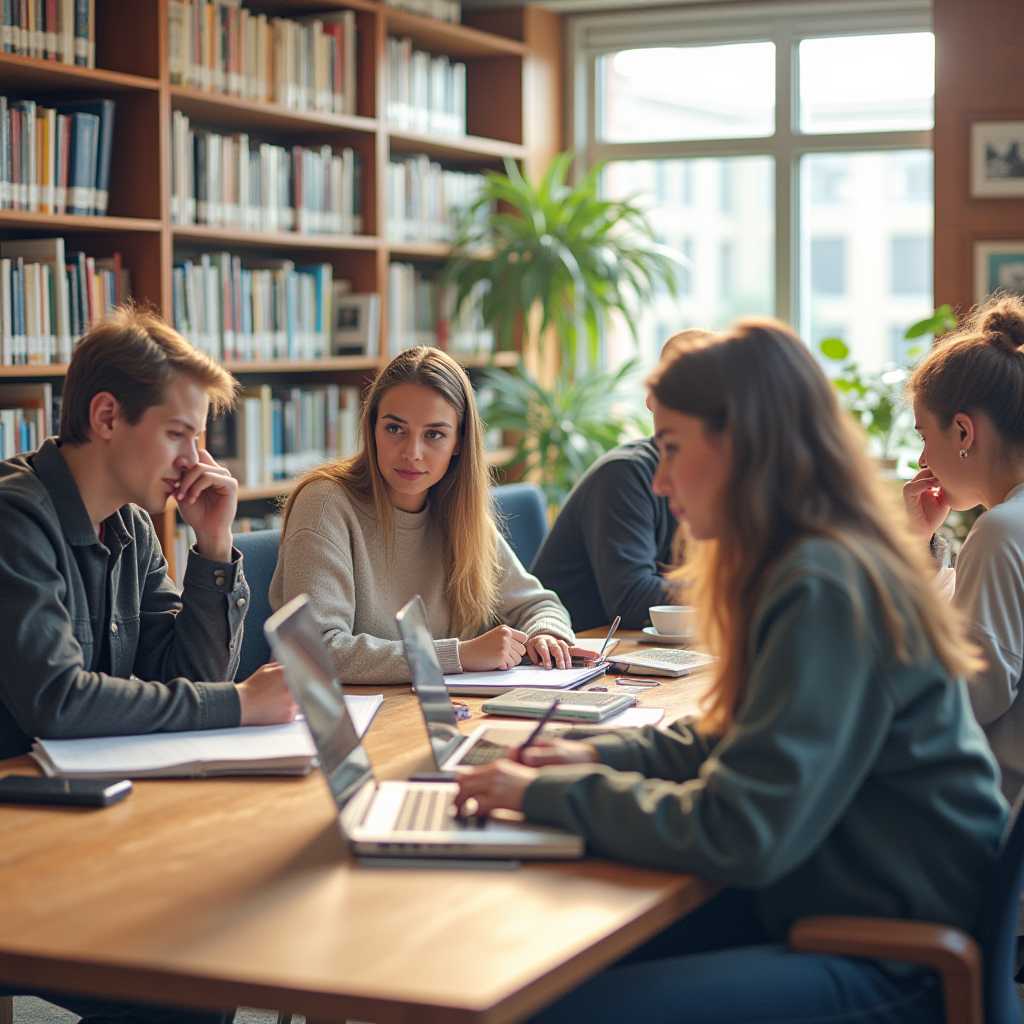} & \includegraphics[width=\imgwidth]{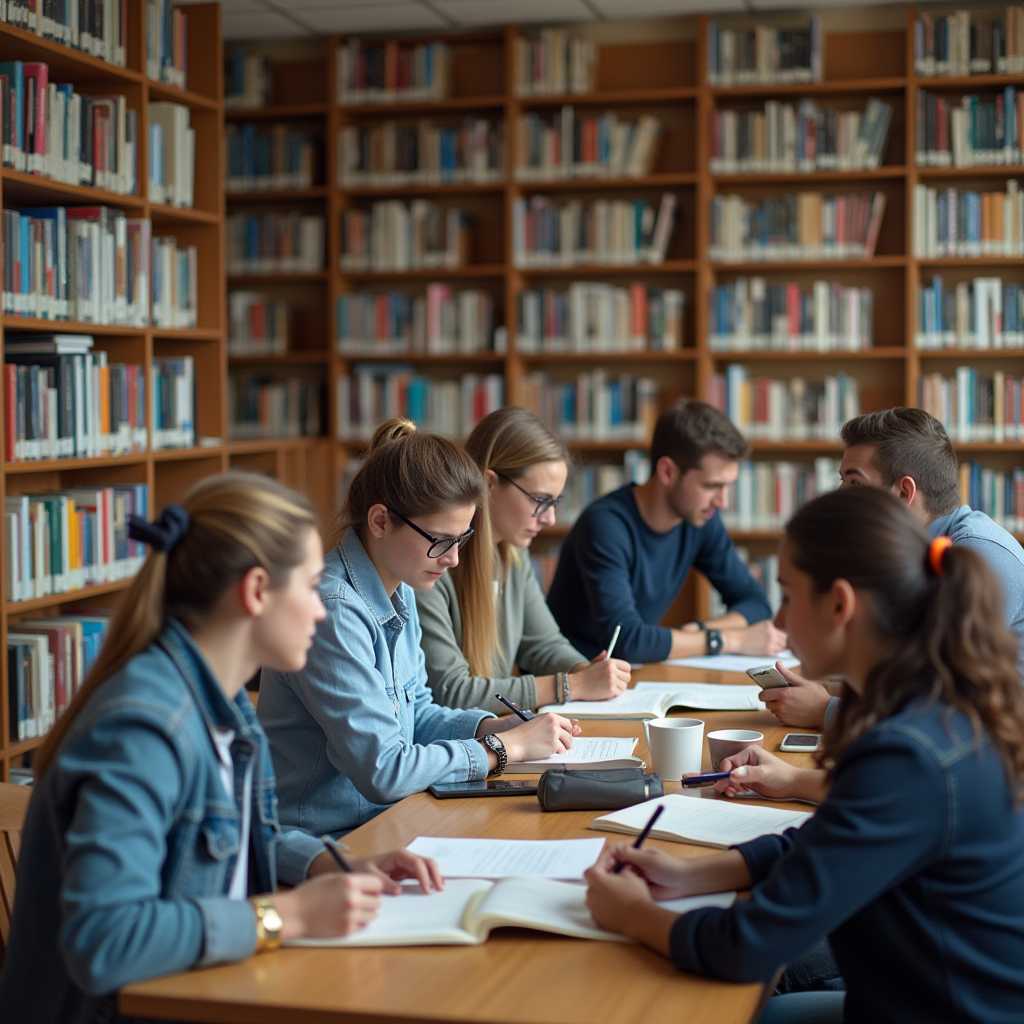} & \includegraphics[width=\imgwidth]{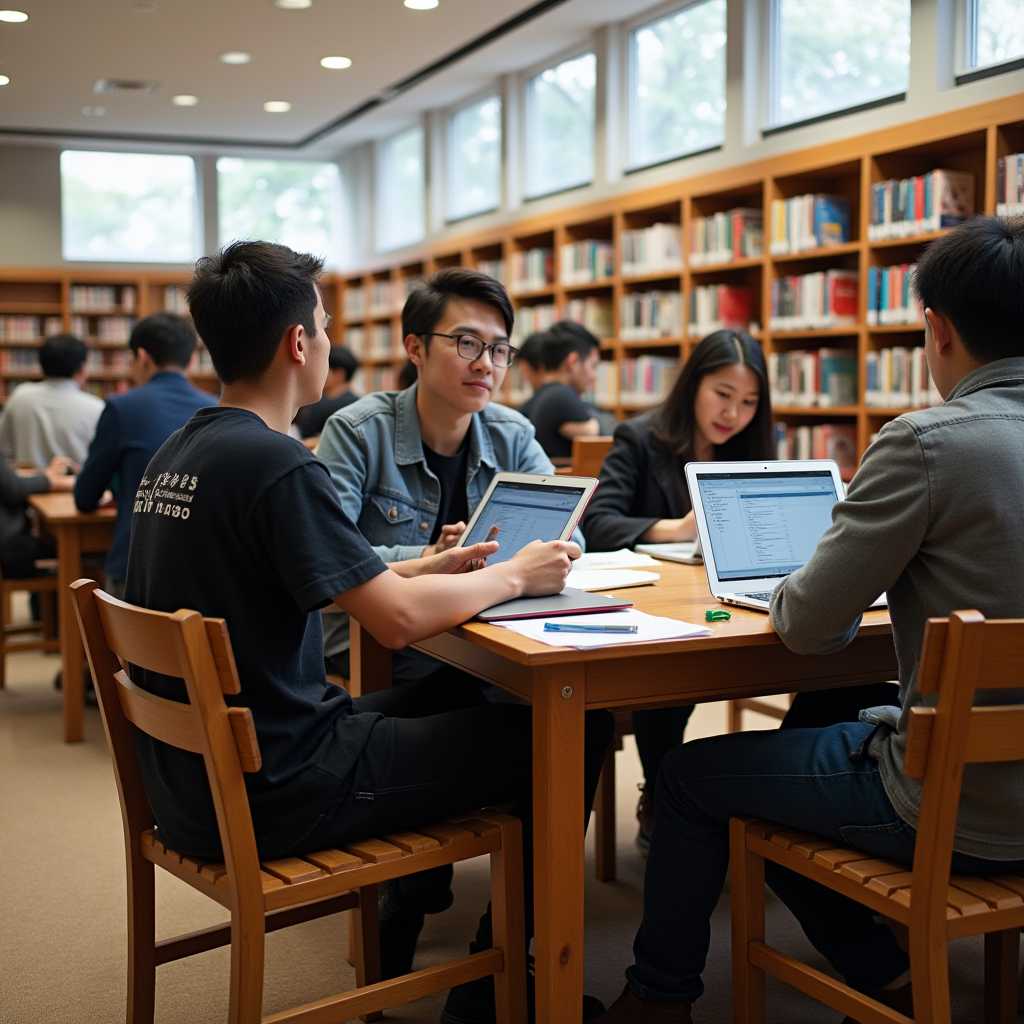} & \includegraphics[width=\imgwidth]{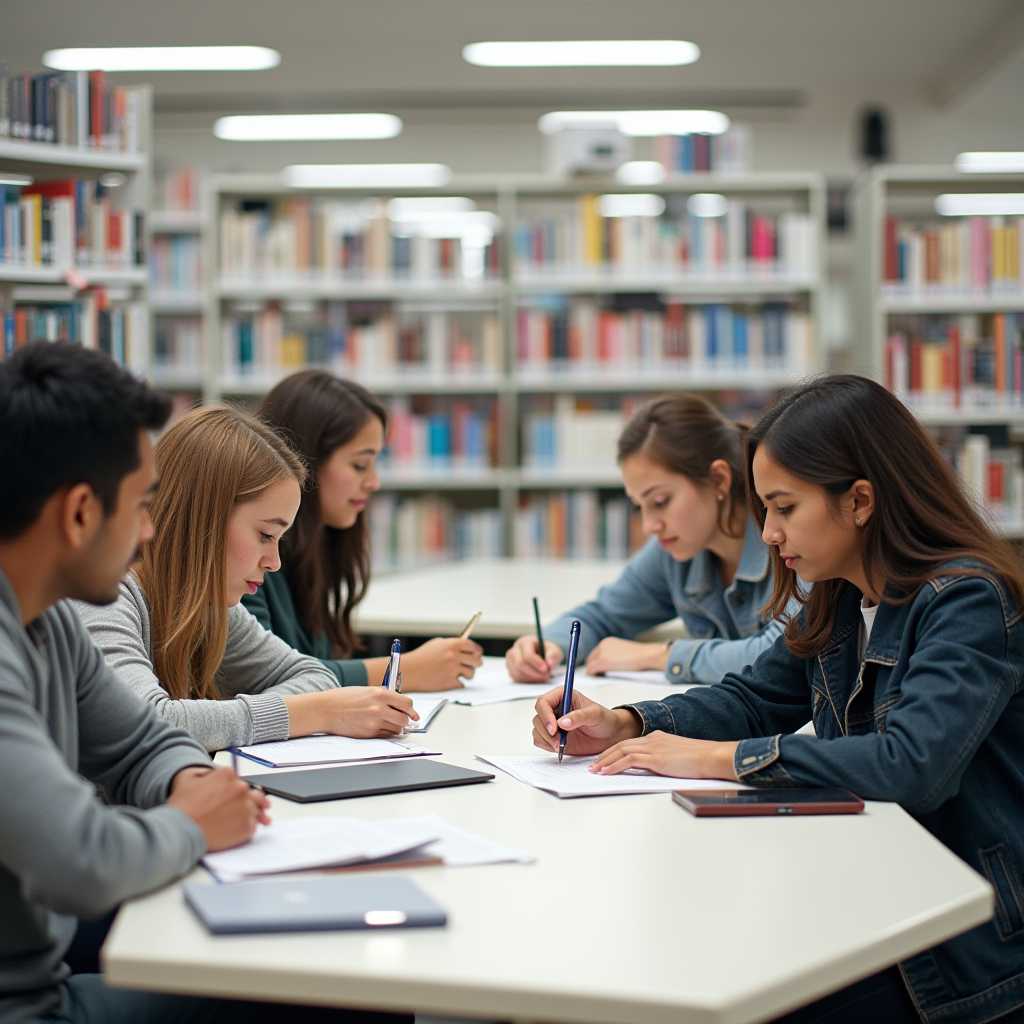} \\
        \multicolumn{9}{c}{\vspace{2pt}\small ``A group of students studying together in a university library'' \vspace{8pt}} \\

        \vertlabel{Flux} & \includegraphics[width=\imgwidth]{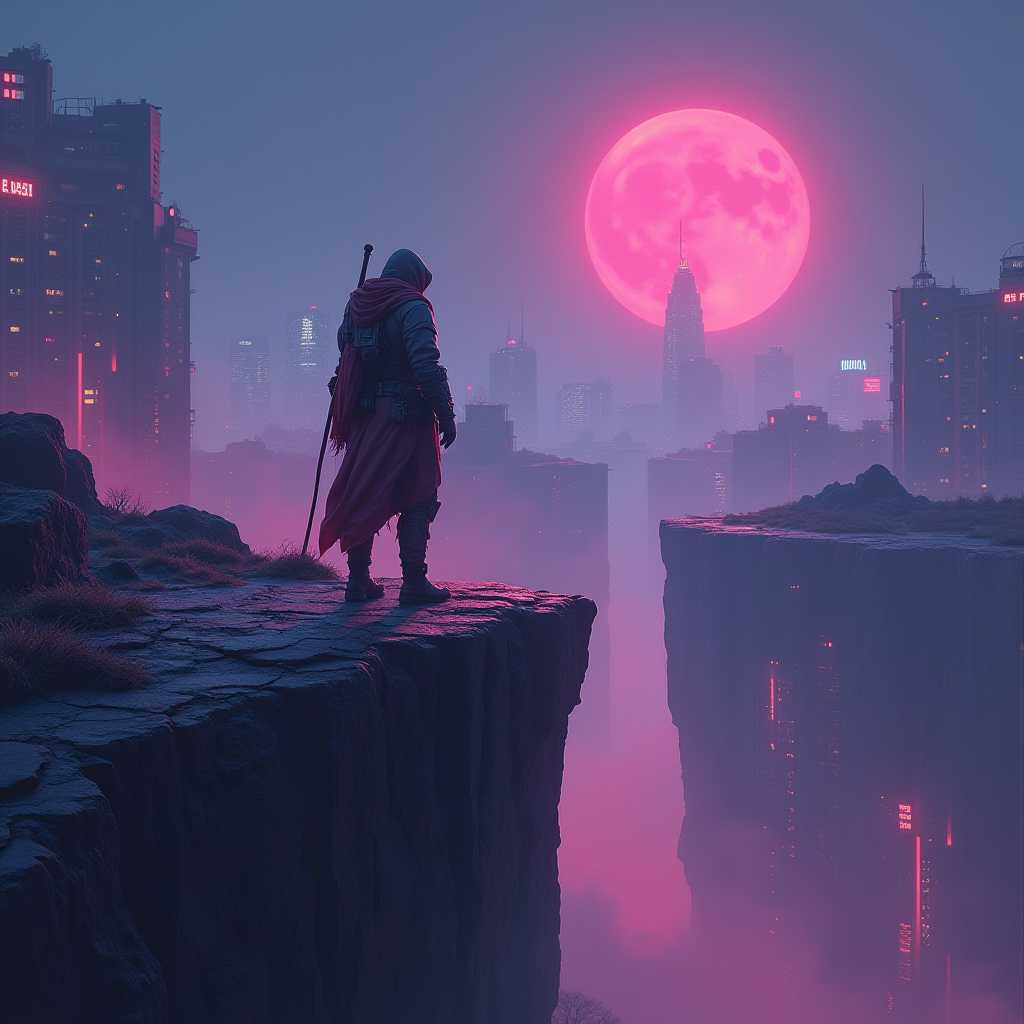} & \includegraphics[width=\imgwidth]{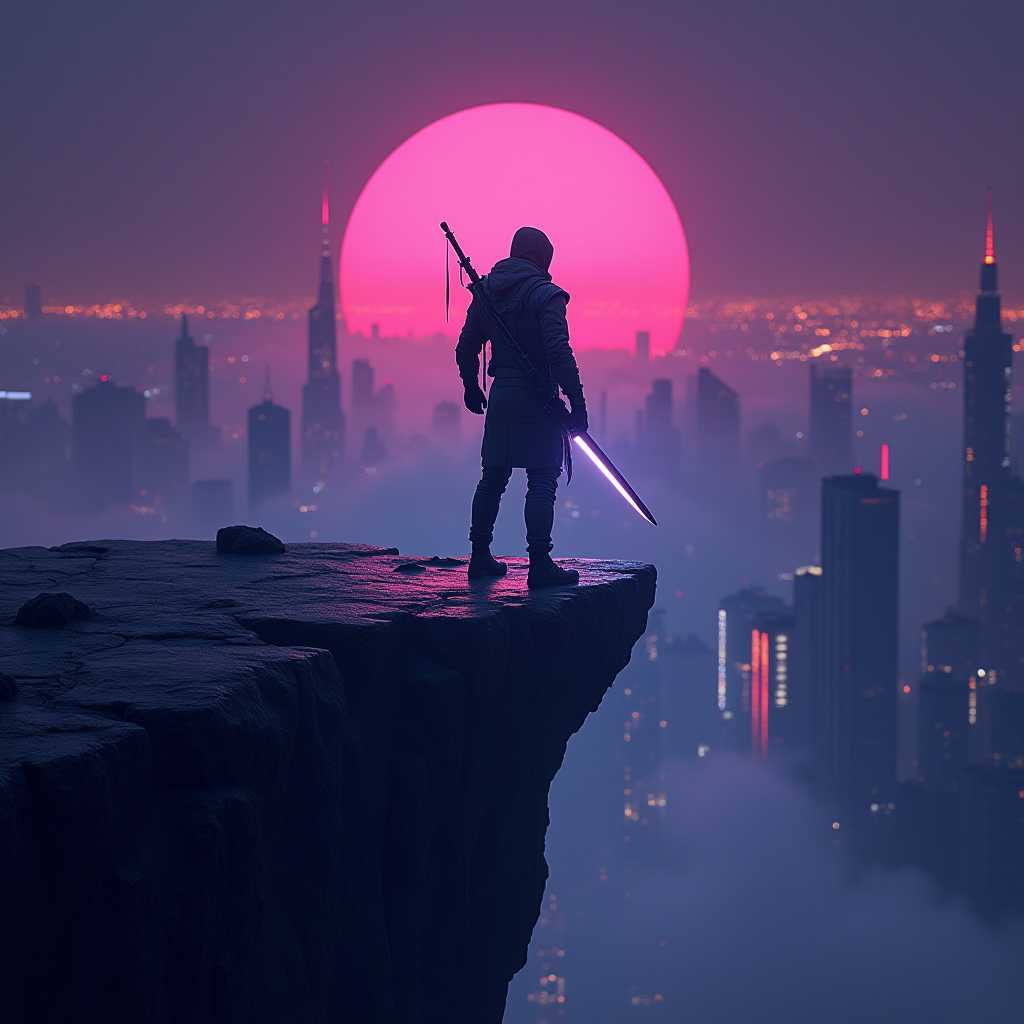} & \includegraphics[width=\imgwidth]{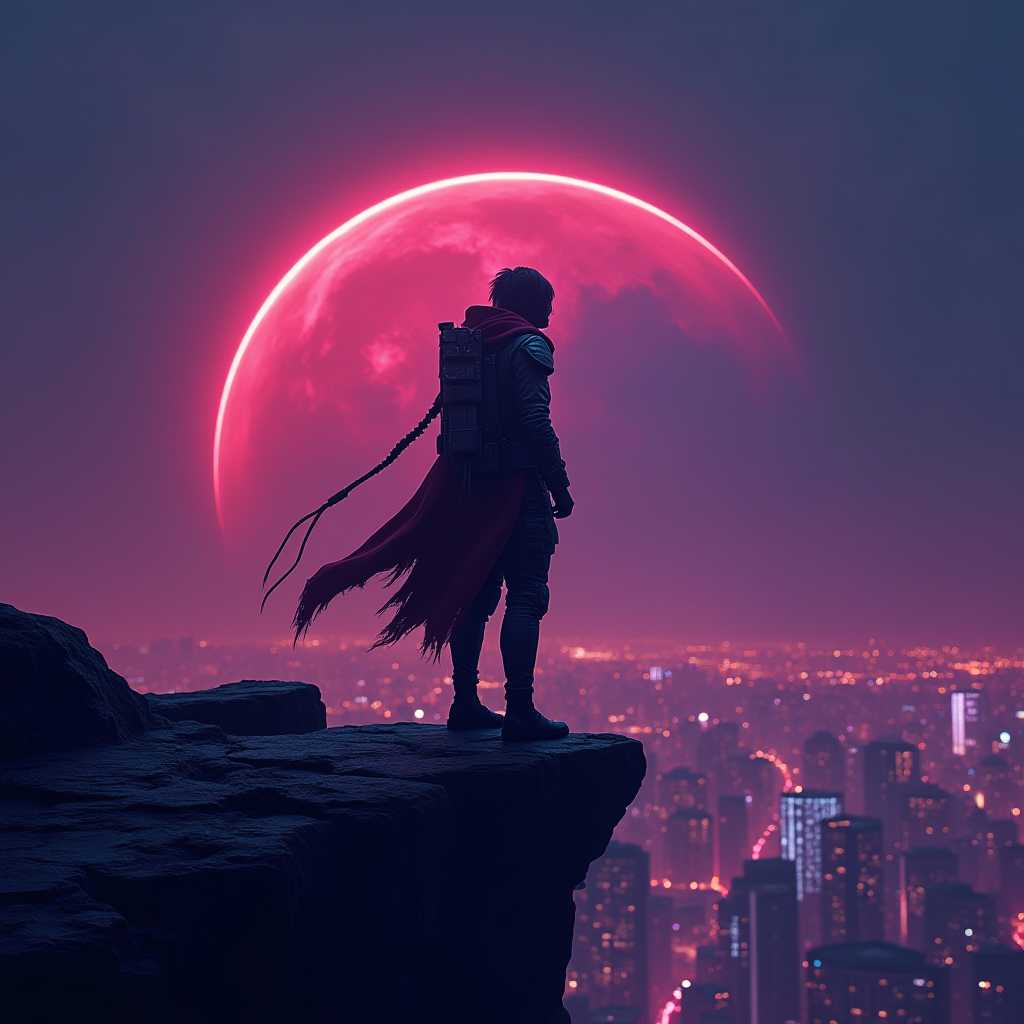} & \includegraphics[width=\imgwidth]{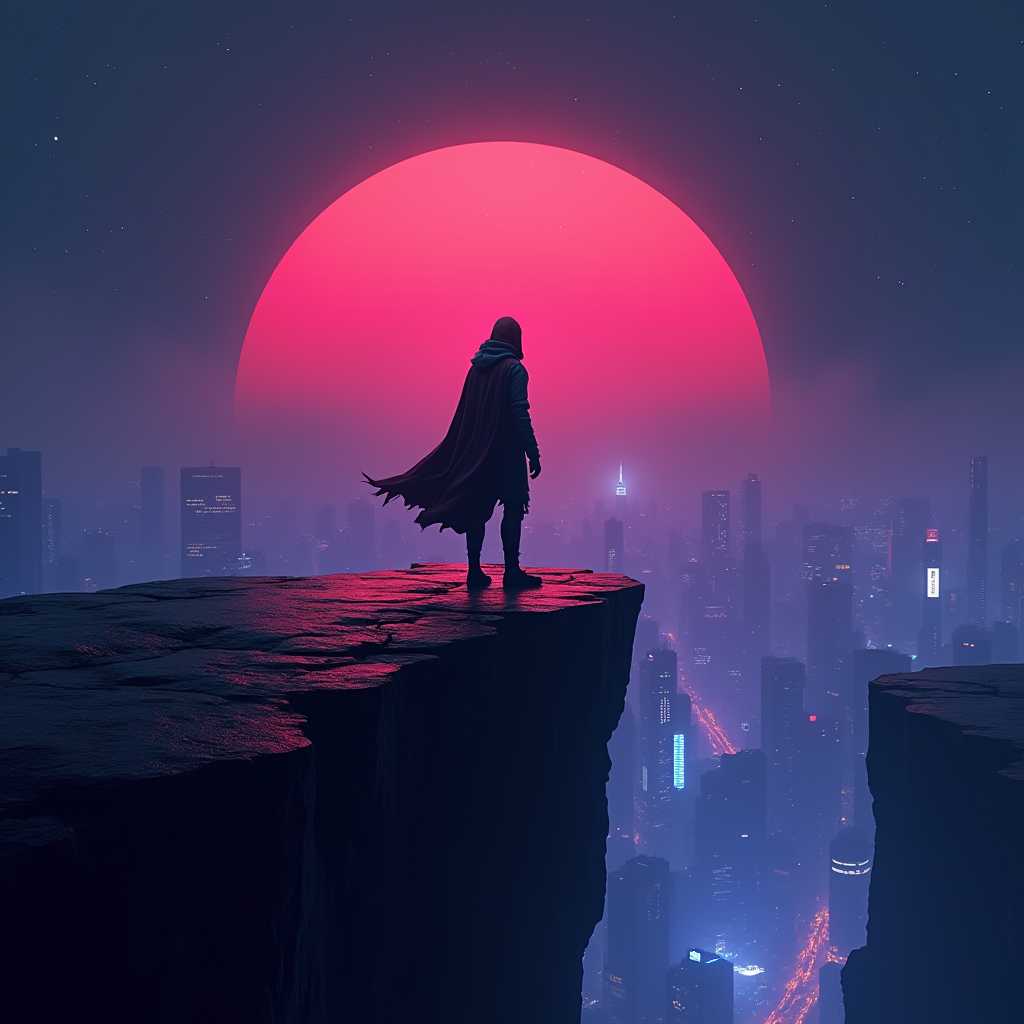} & \includegraphics[width=\imgwidth]{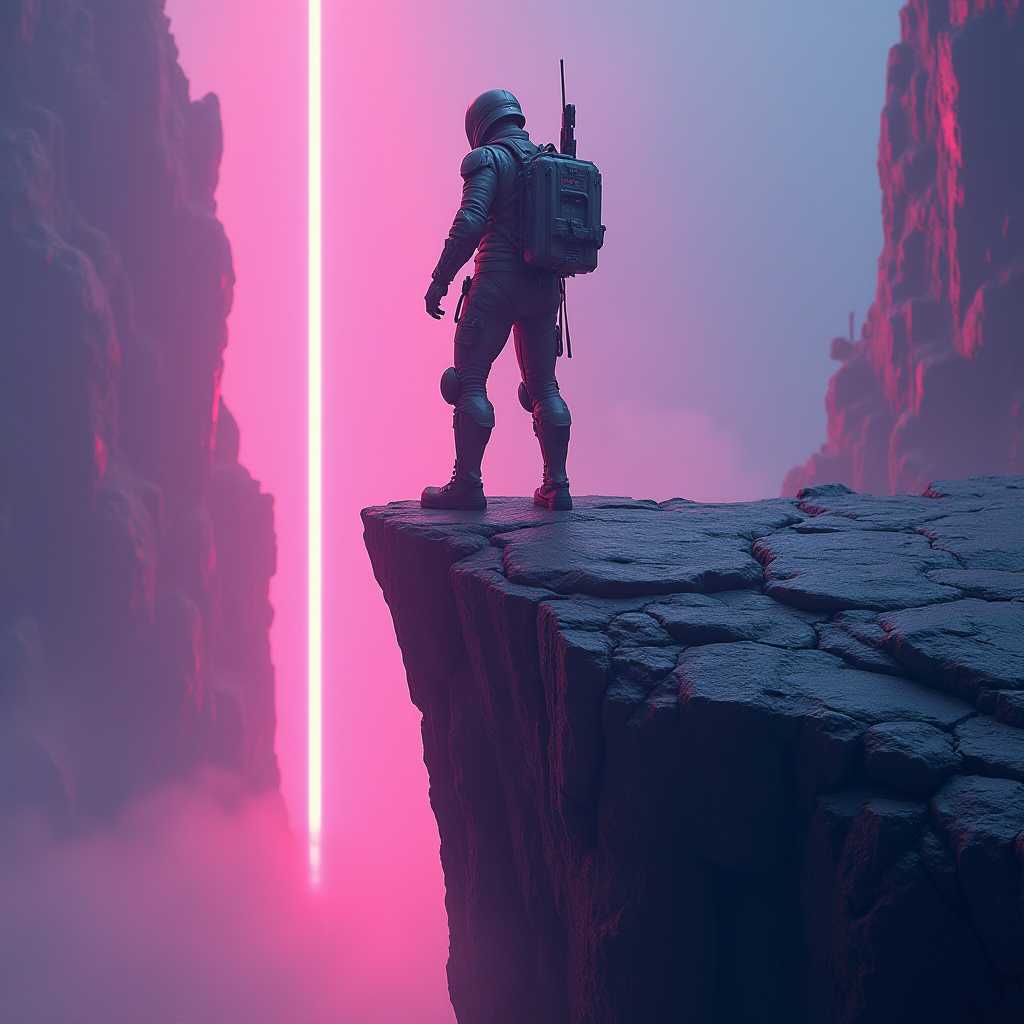} & \includegraphics[width=\imgwidth]{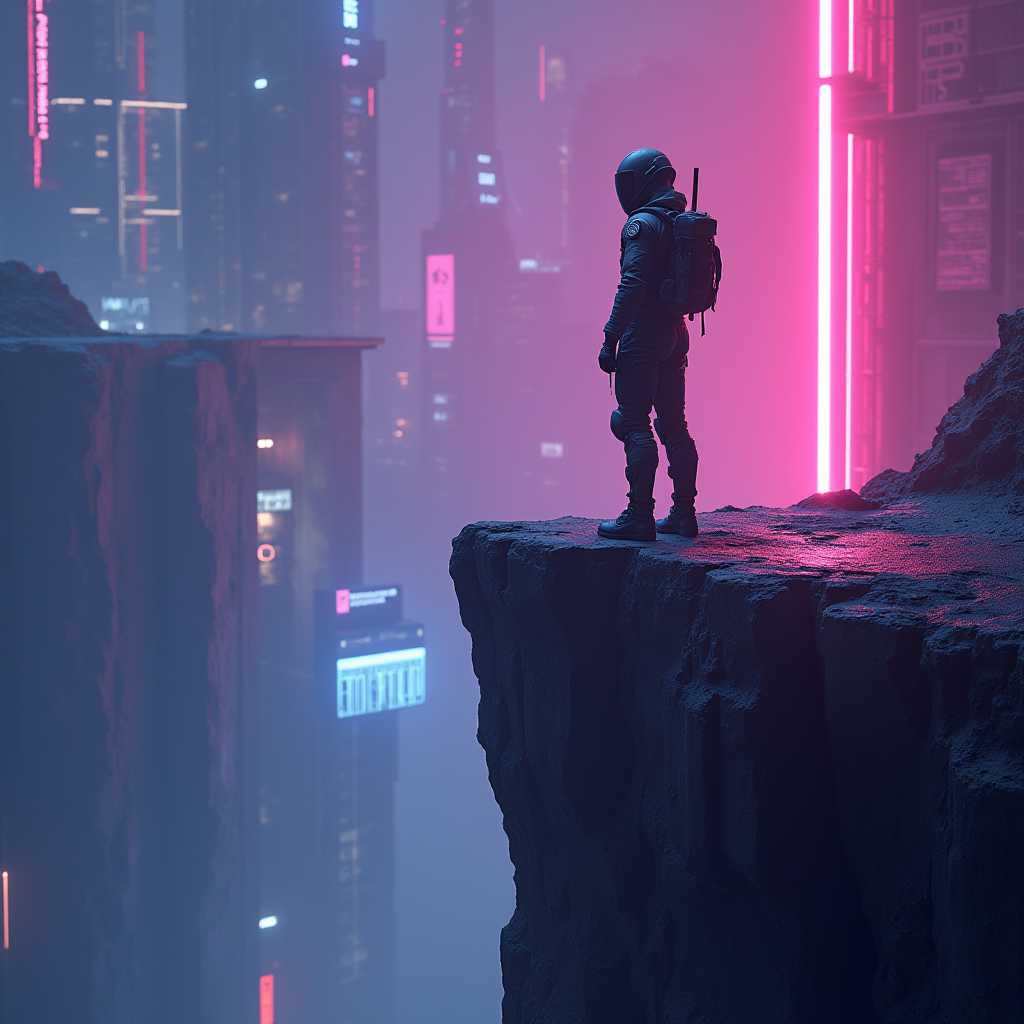} & \includegraphics[width=\imgwidth]{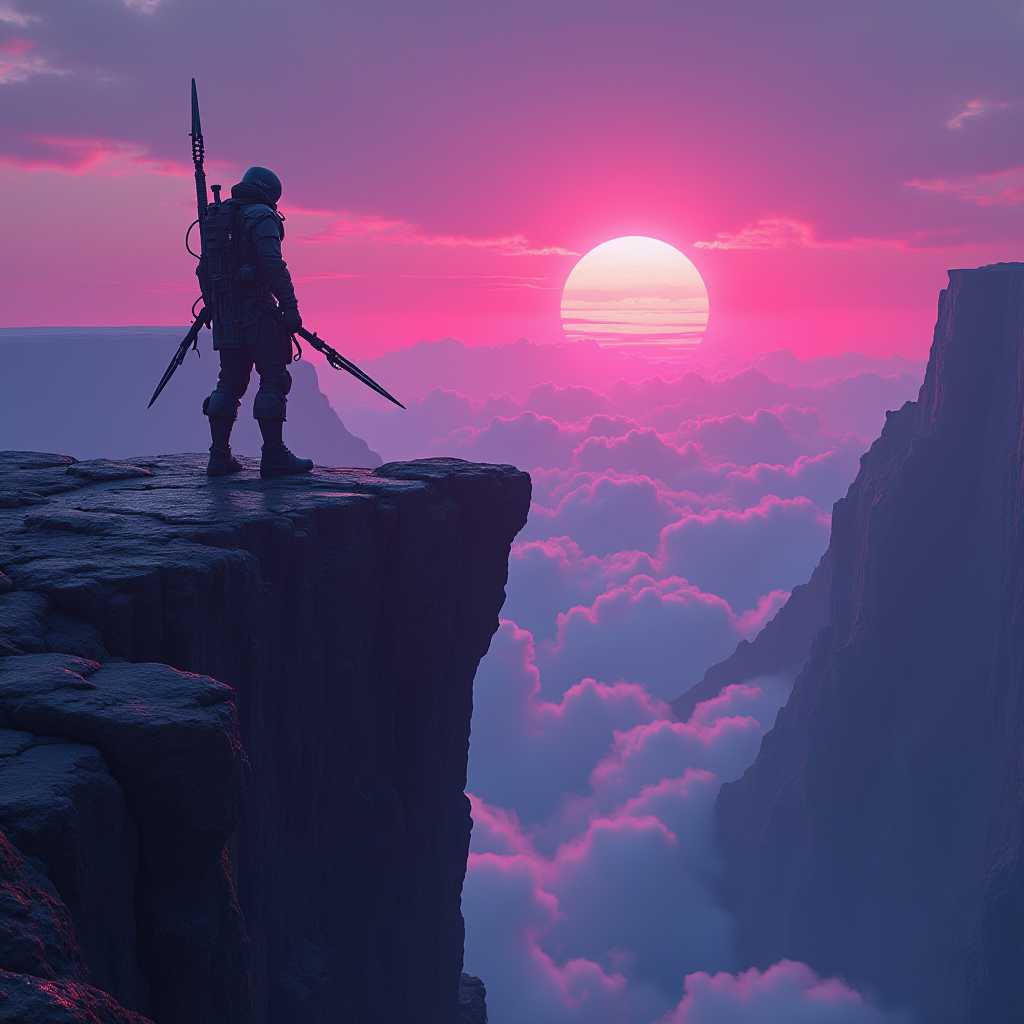} & \includegraphics[width=\imgwidth]{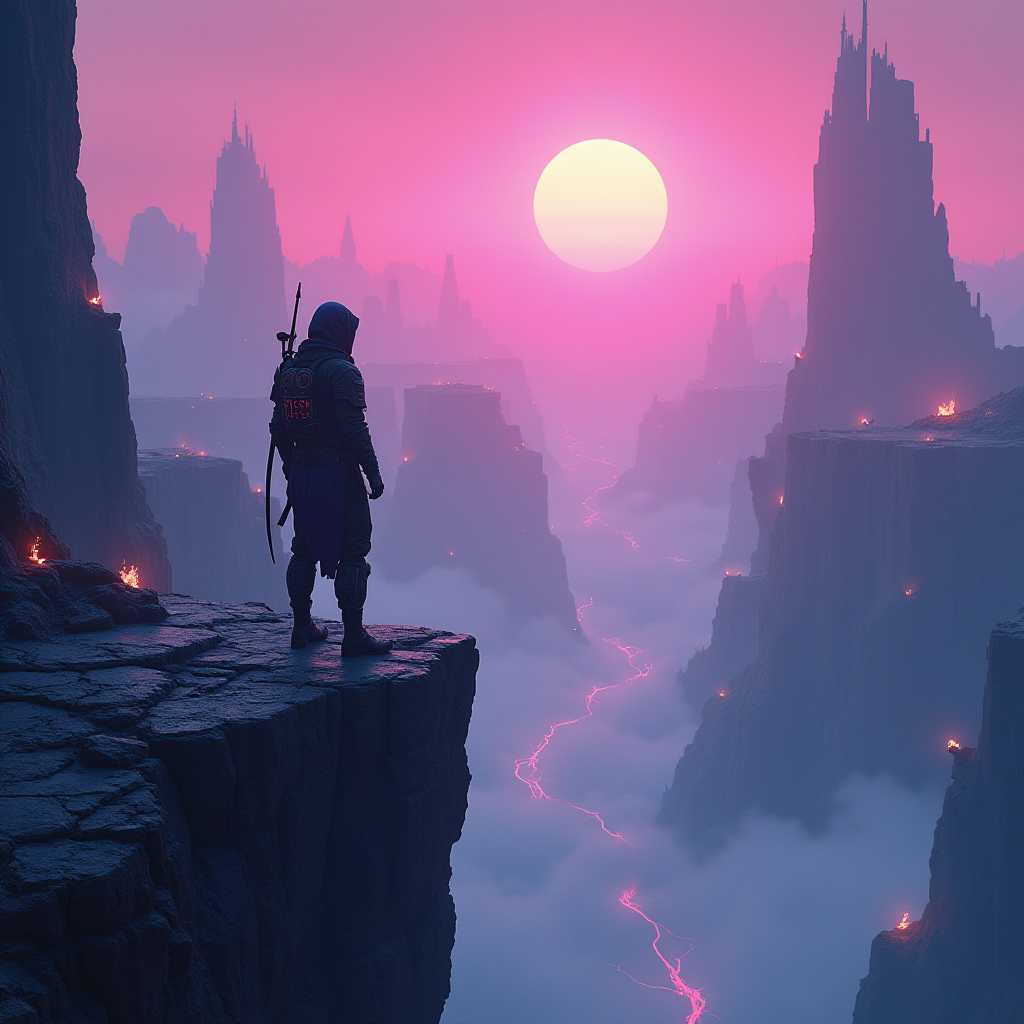} \\[-1pt]
        \vertlabel{Ours} & \includegraphics[width=\imgwidth]{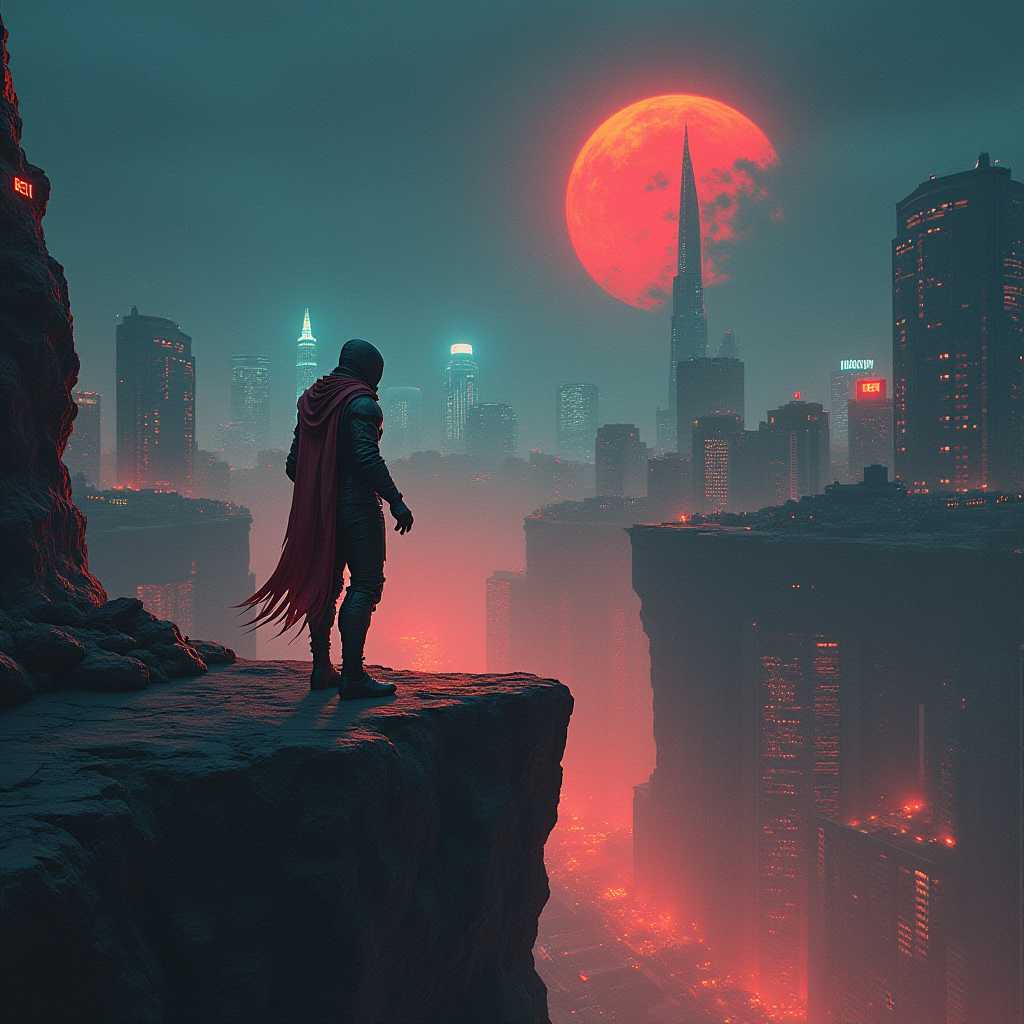} & \includegraphics[width=\imgwidth]{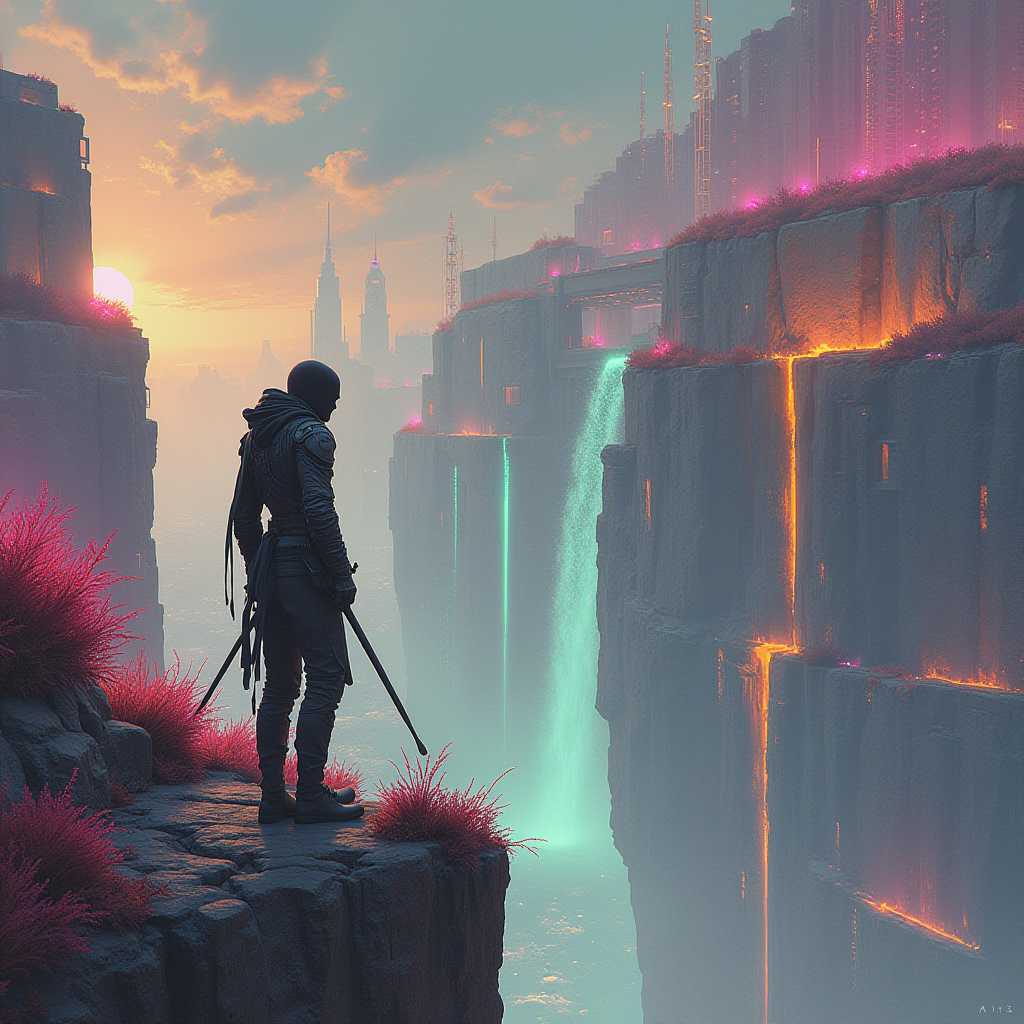} & \includegraphics[width=\imgwidth]{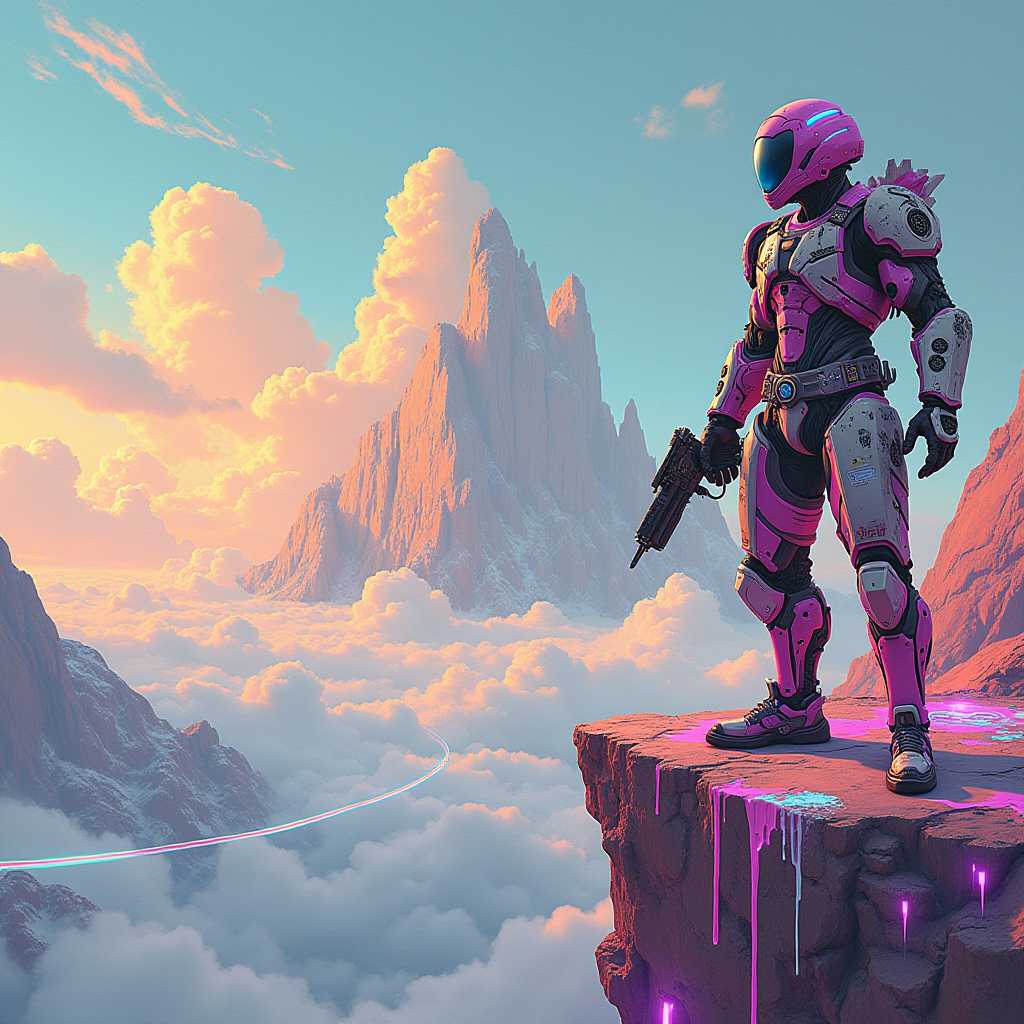} & \includegraphics[width=\imgwidth]{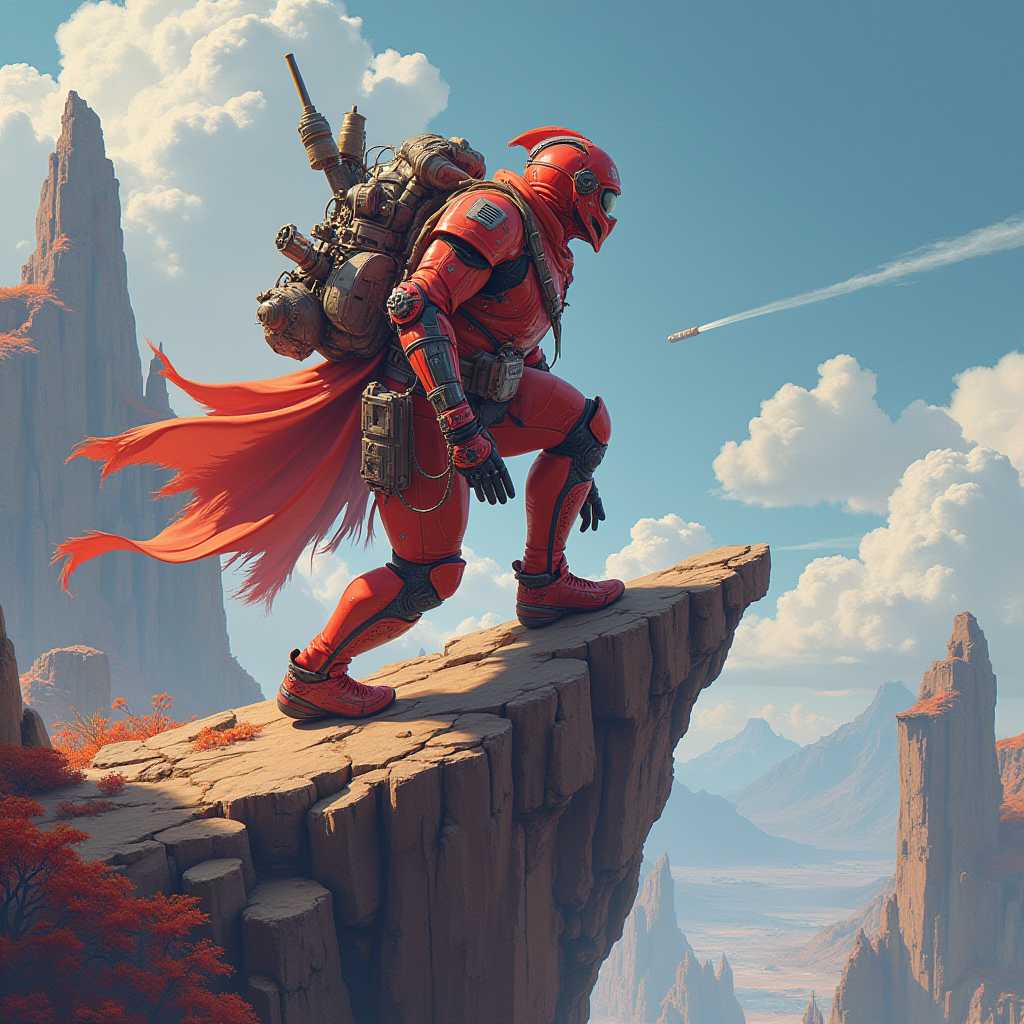} & \includegraphics[width=\imgwidth]{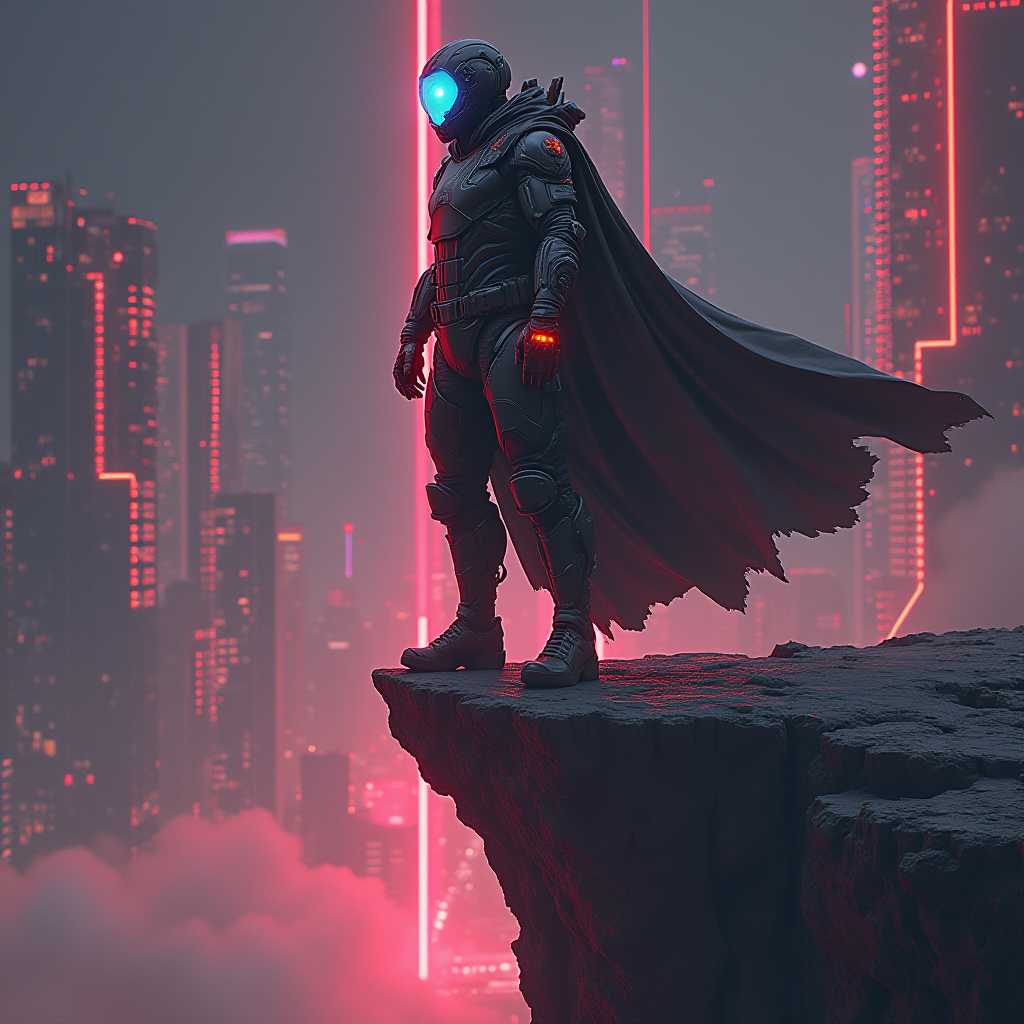} & \includegraphics[width=\imgwidth]{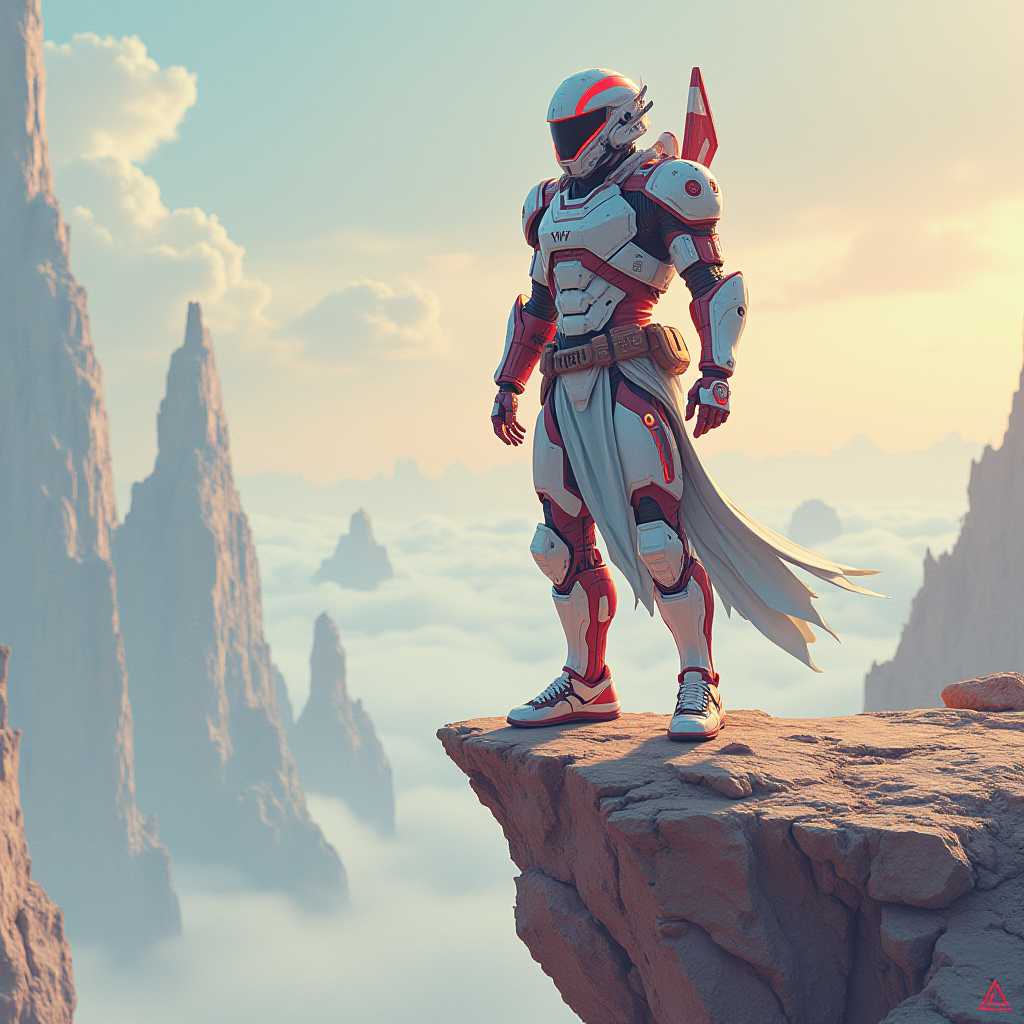} & \includegraphics[width=\imgwidth]{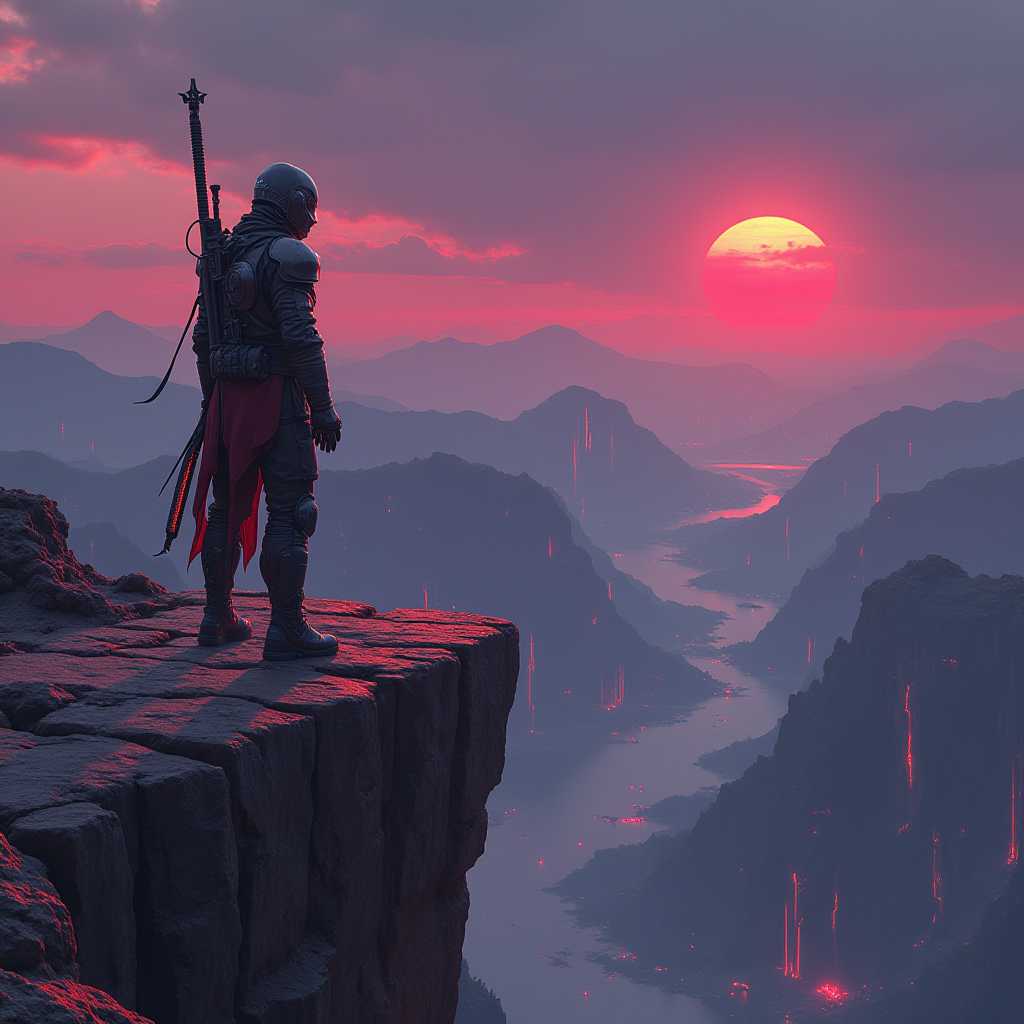} & \includegraphics[width=\imgwidth]{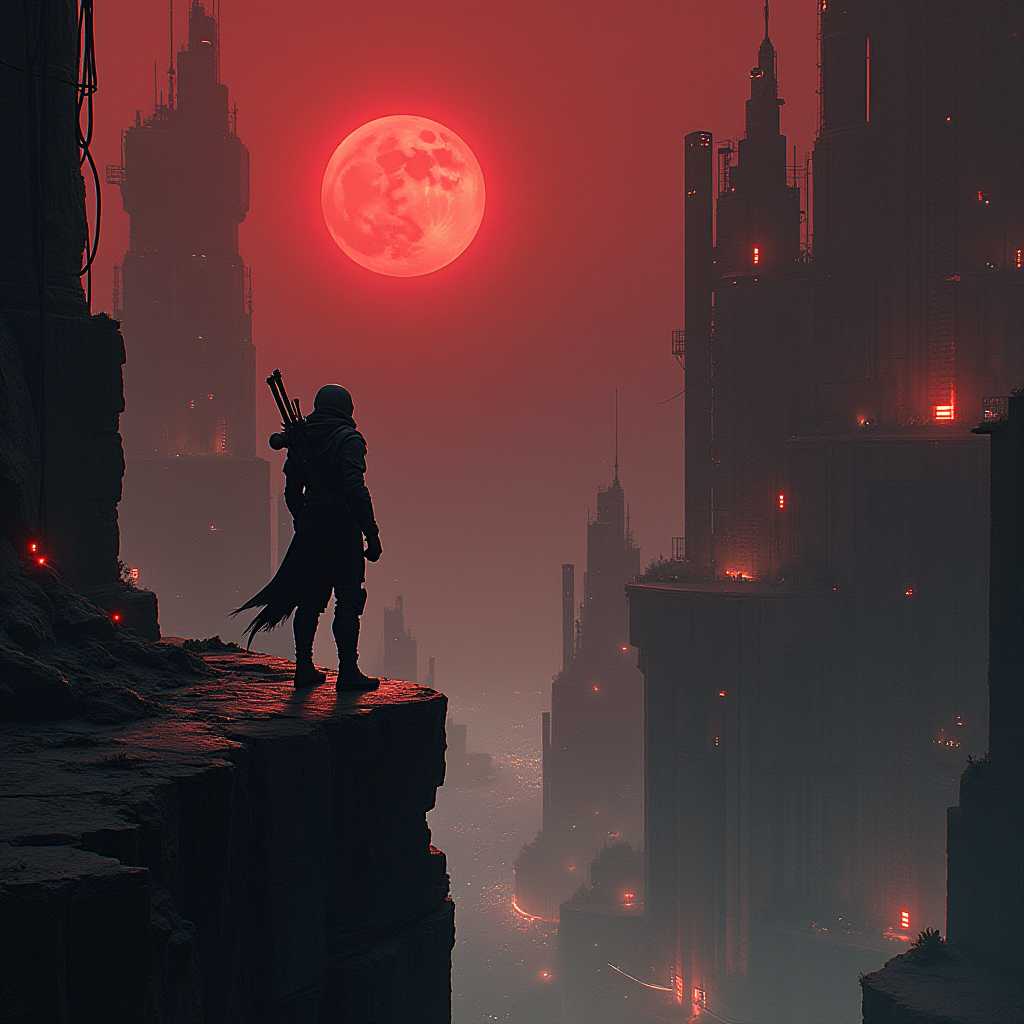} \\
        \multicolumn{9}{c}{\vspace{2pt}\small ``A futuristic warrior standing on the edge of a neon-lit cliff'' \vspace{8pt}} \\

        \vertlabel{Flux} & \includegraphics[width=\imgwidth]{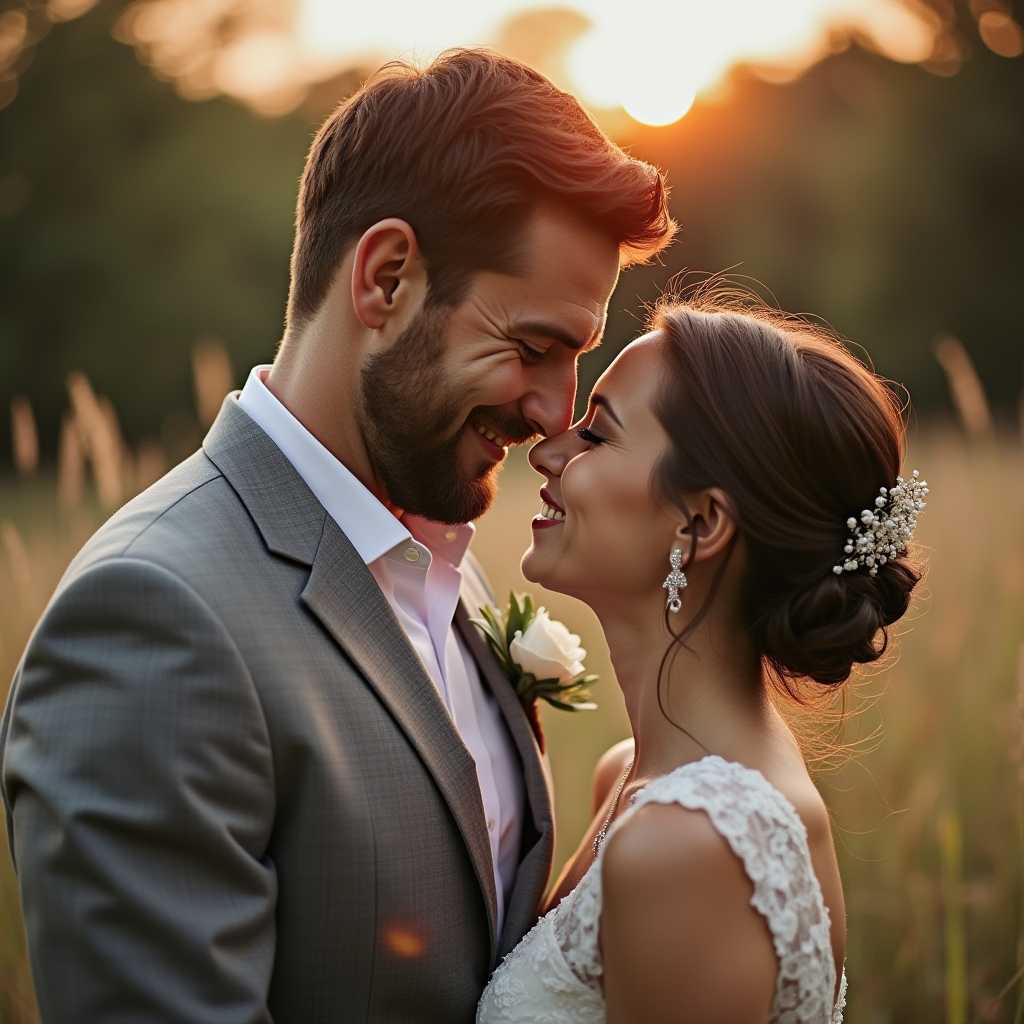} & \includegraphics[width=\imgwidth]{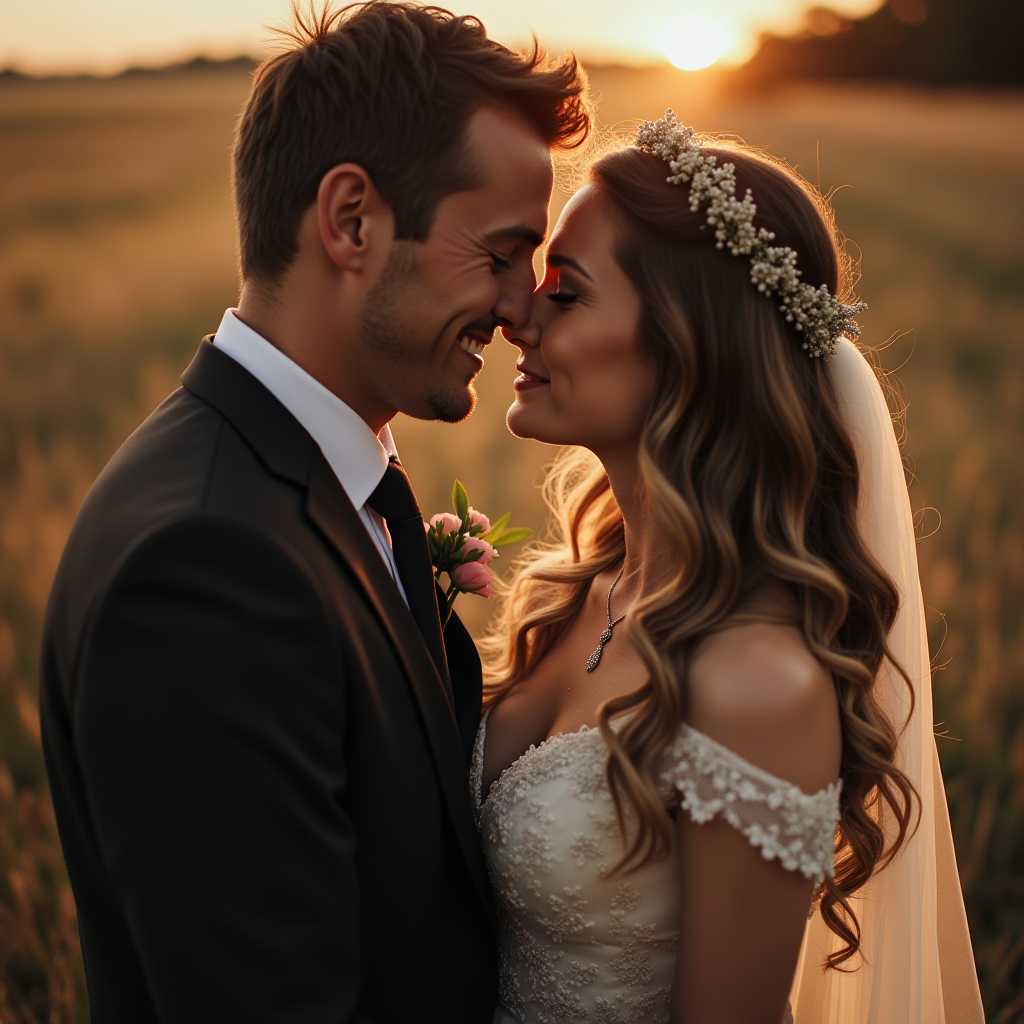} & \includegraphics[width=\imgwidth]{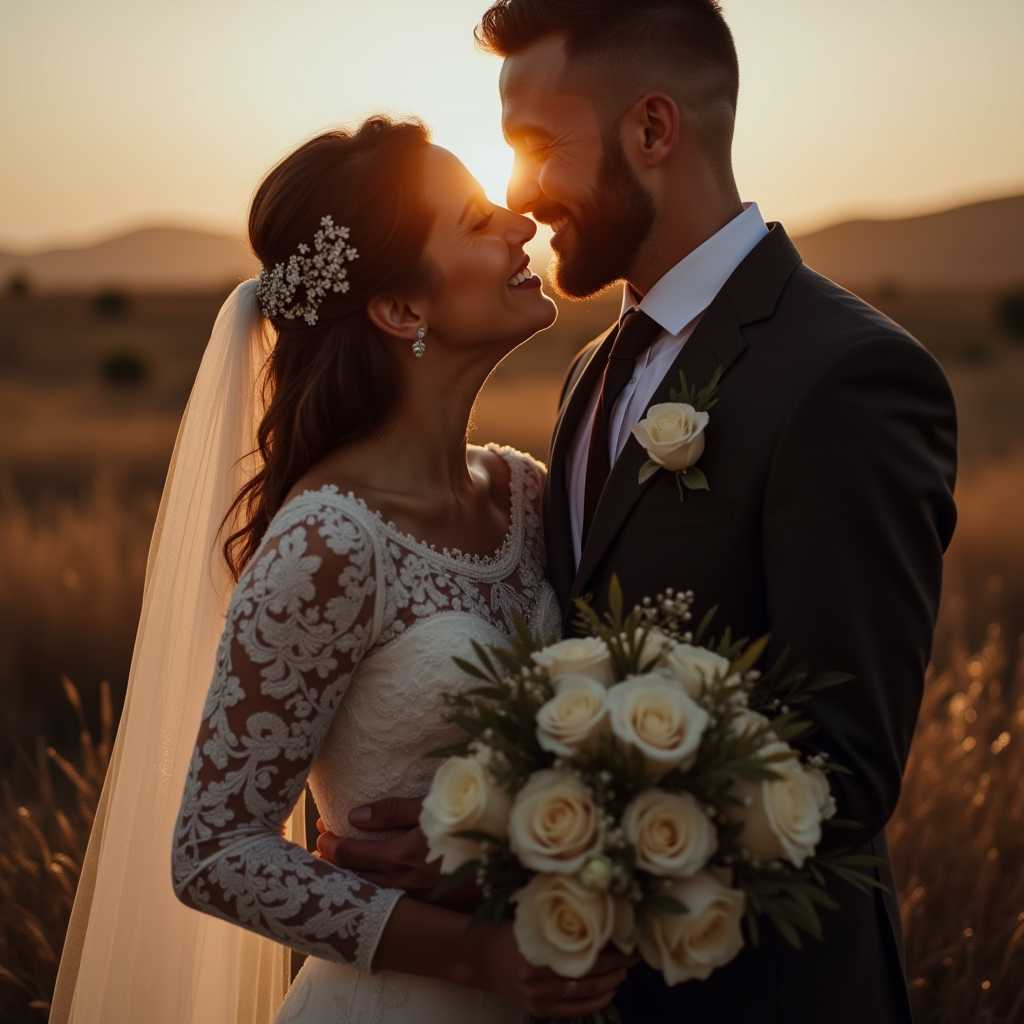} & \includegraphics[width=\imgwidth]{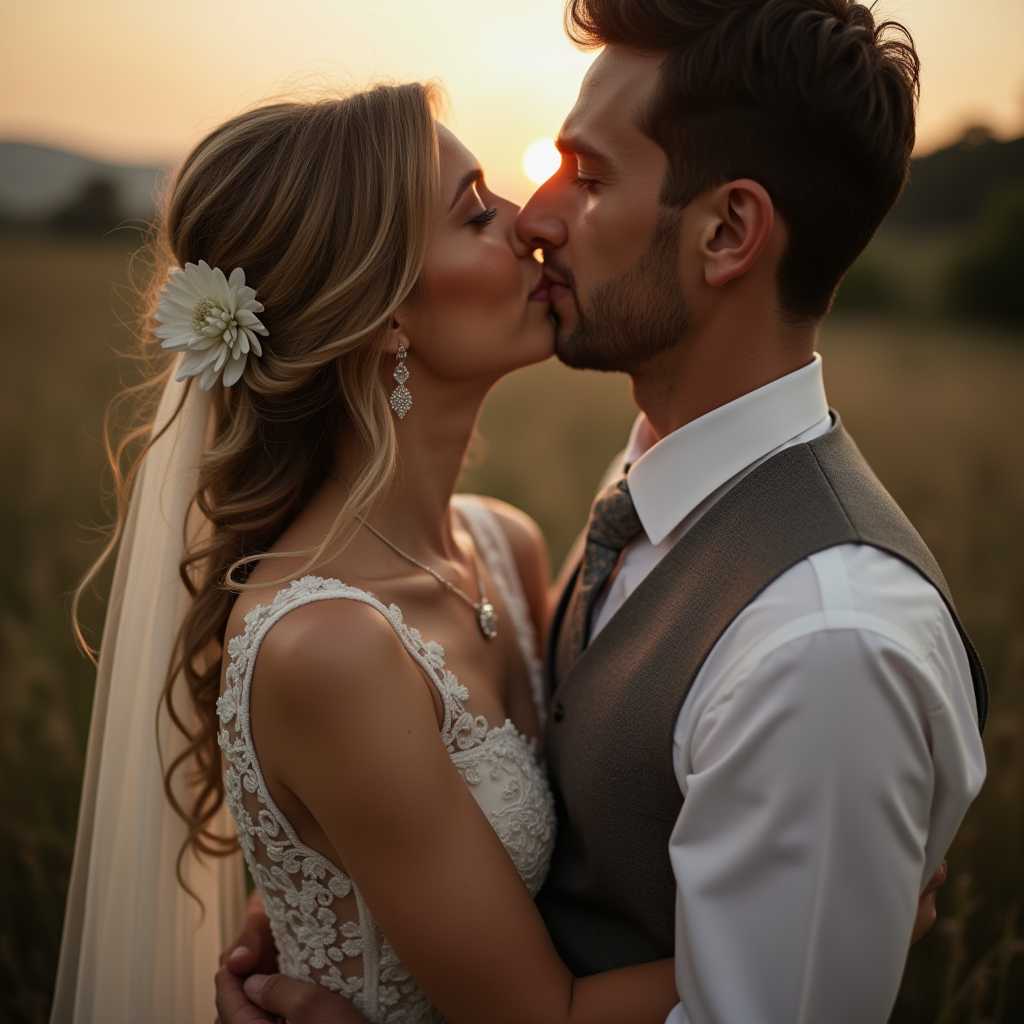} & \includegraphics[width=\imgwidth]{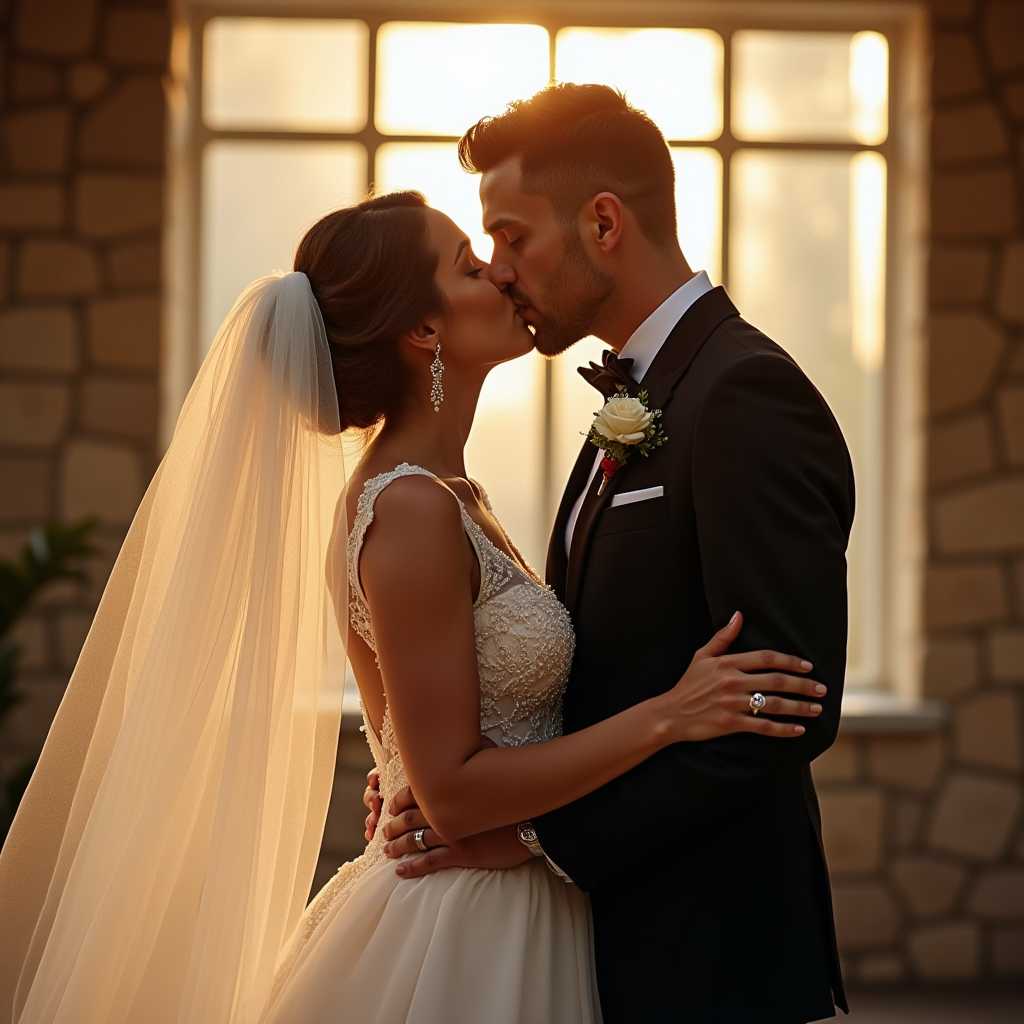} & \includegraphics[width=\imgwidth]{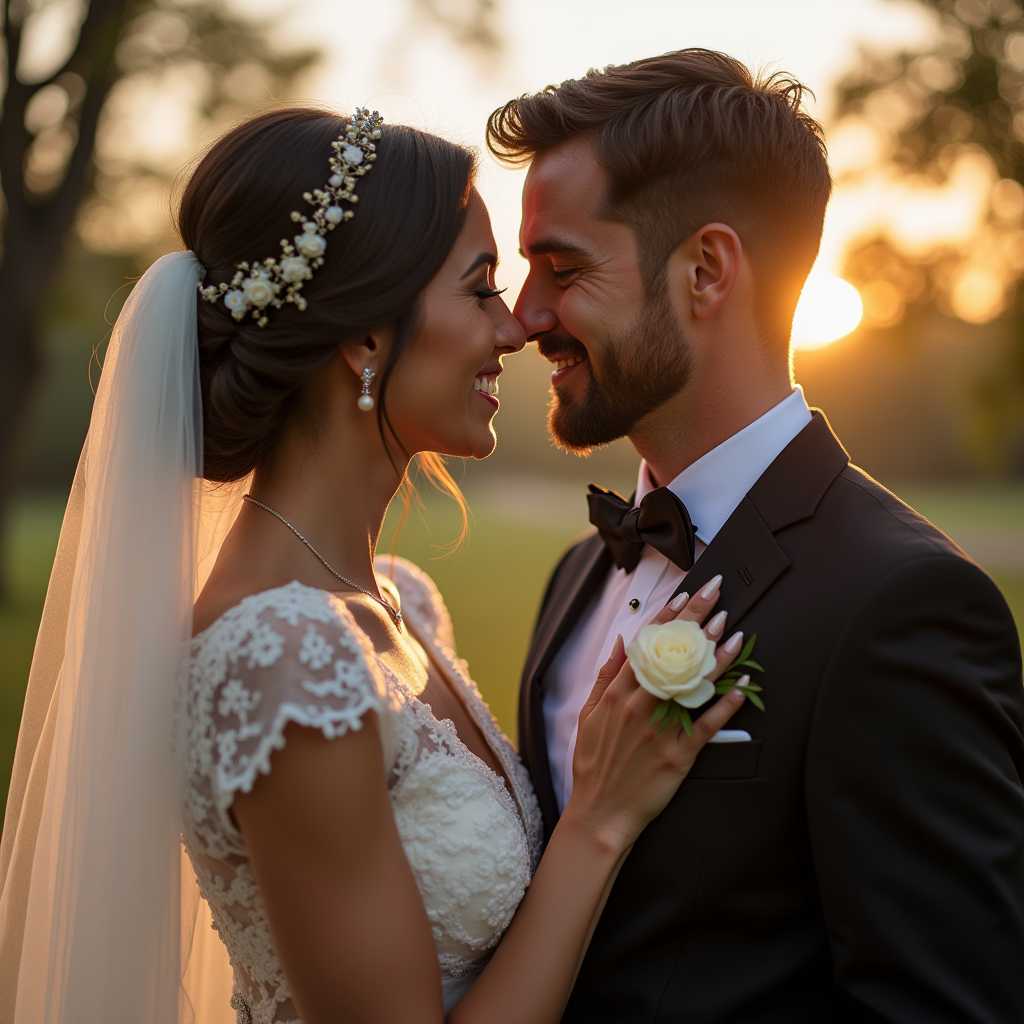} & \includegraphics[width=\imgwidth]{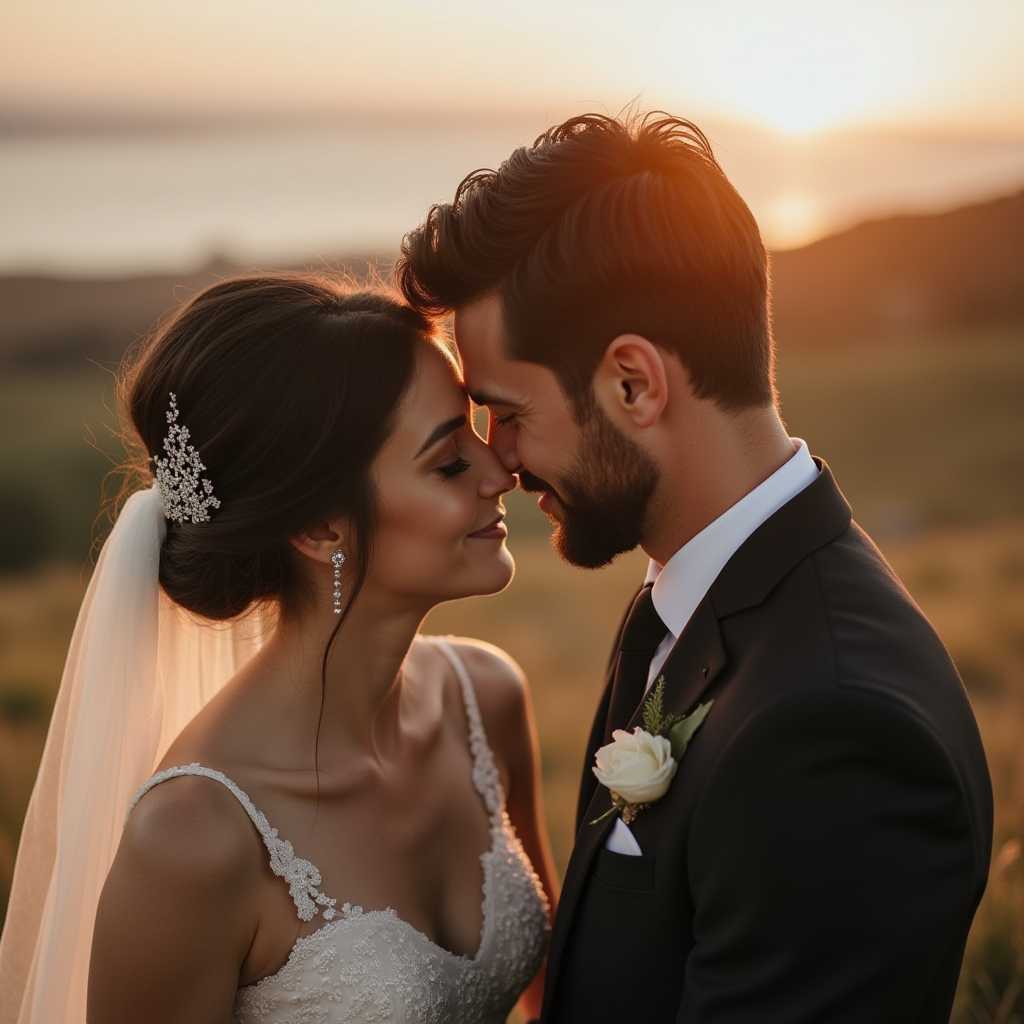} & \includegraphics[width=\imgwidth]{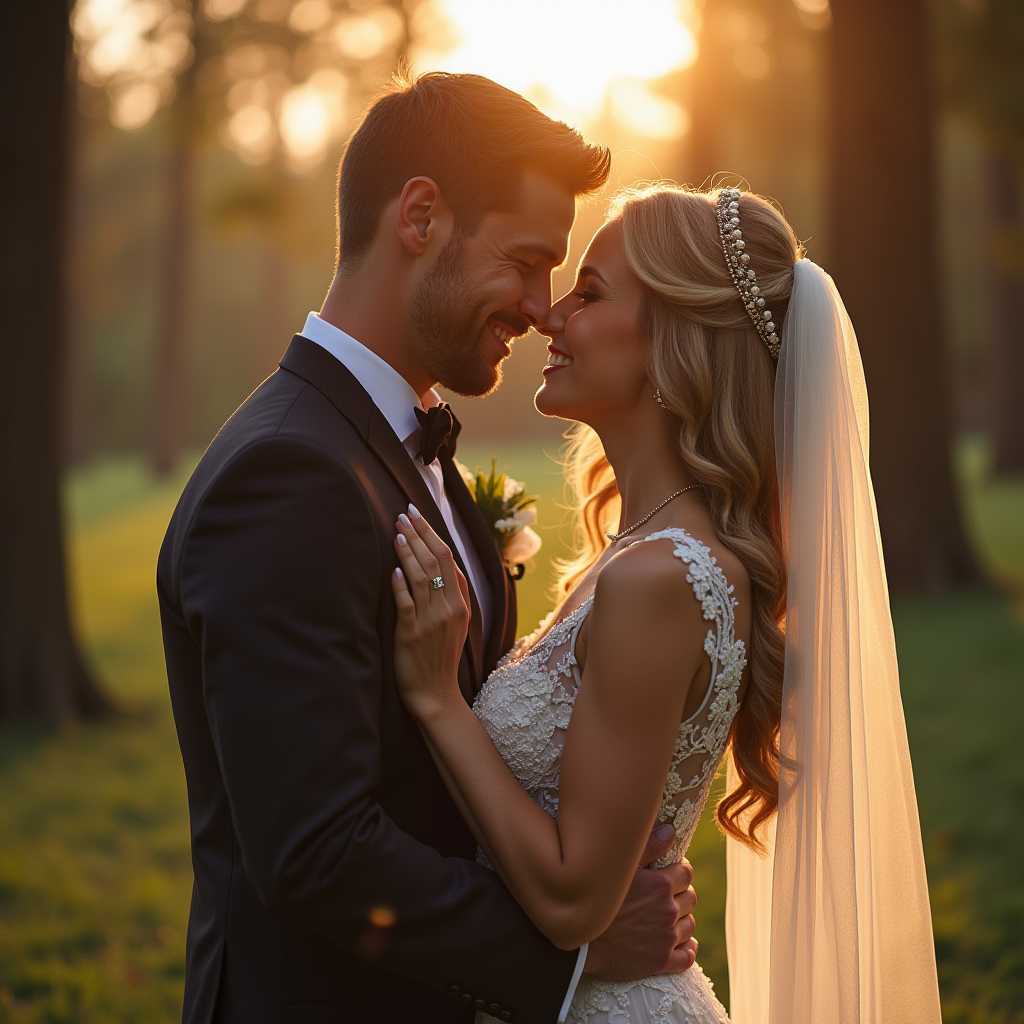} \\[-1pt]
        \vertlabel{Ours} & \includegraphics[width=\imgwidth]{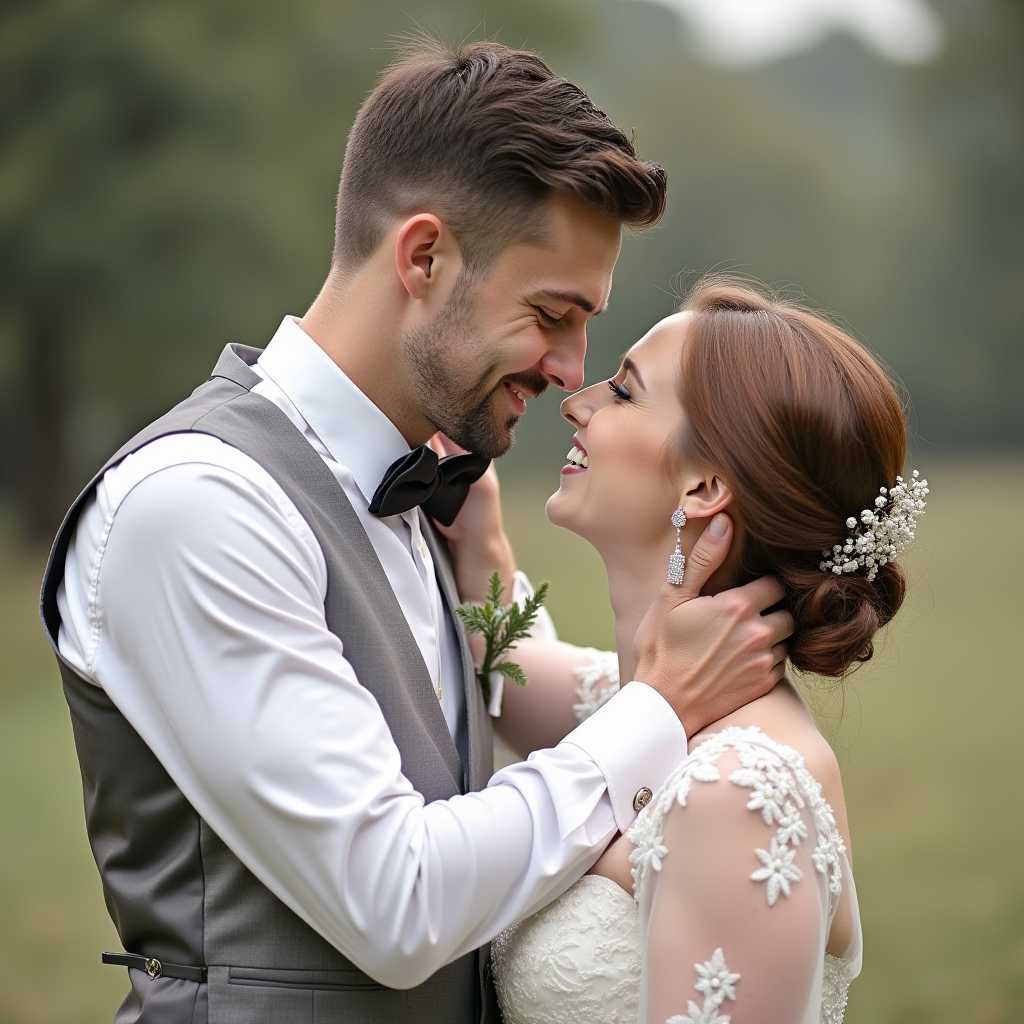} & \includegraphics[width=\imgwidth]{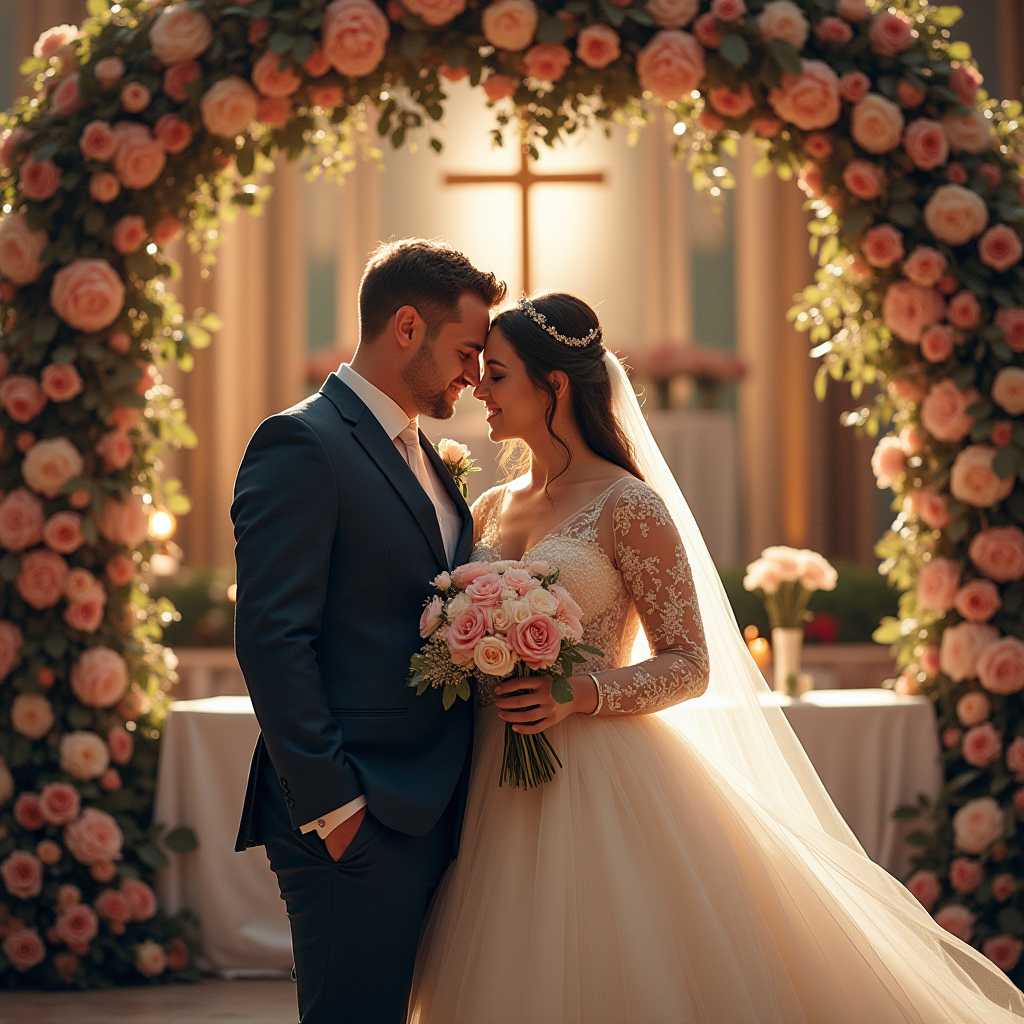} & \includegraphics[width=\imgwidth]{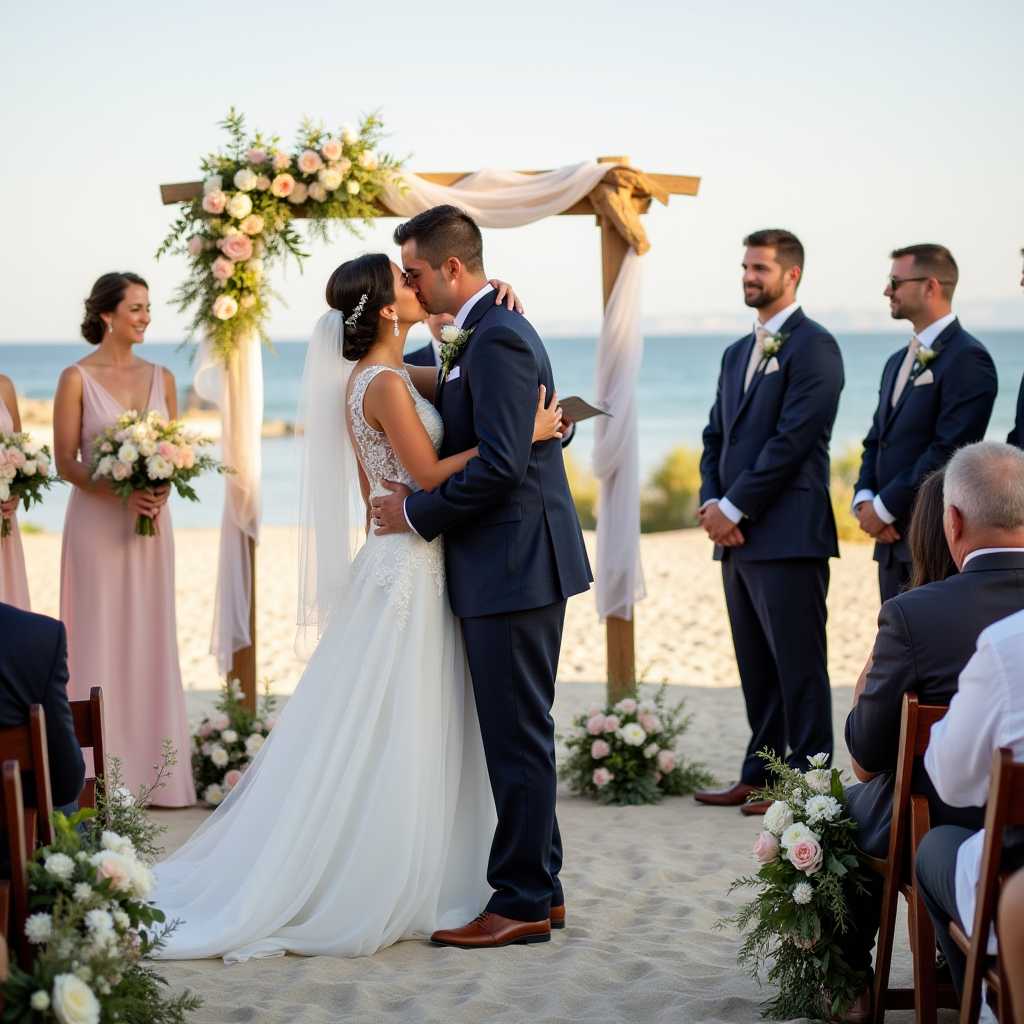} & \includegraphics[width=\imgwidth]{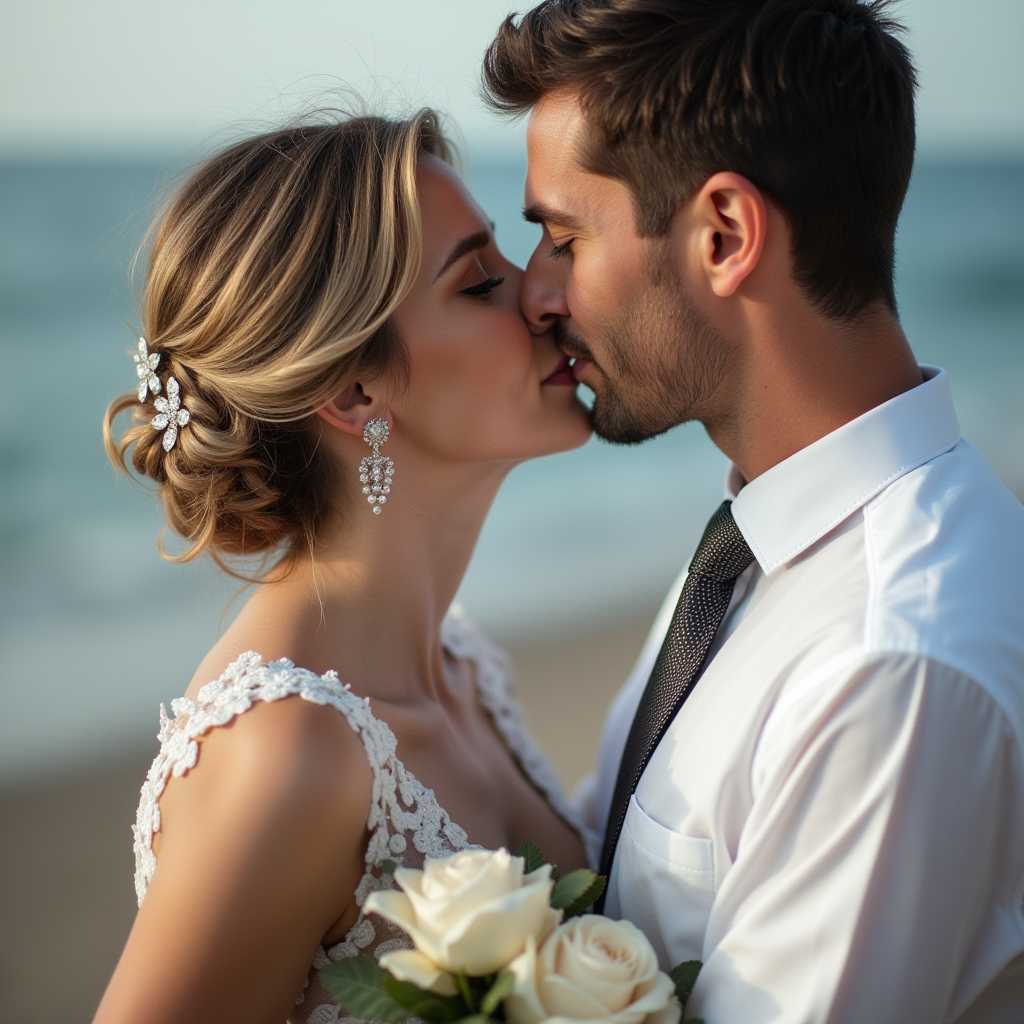} & \includegraphics[width=\imgwidth]{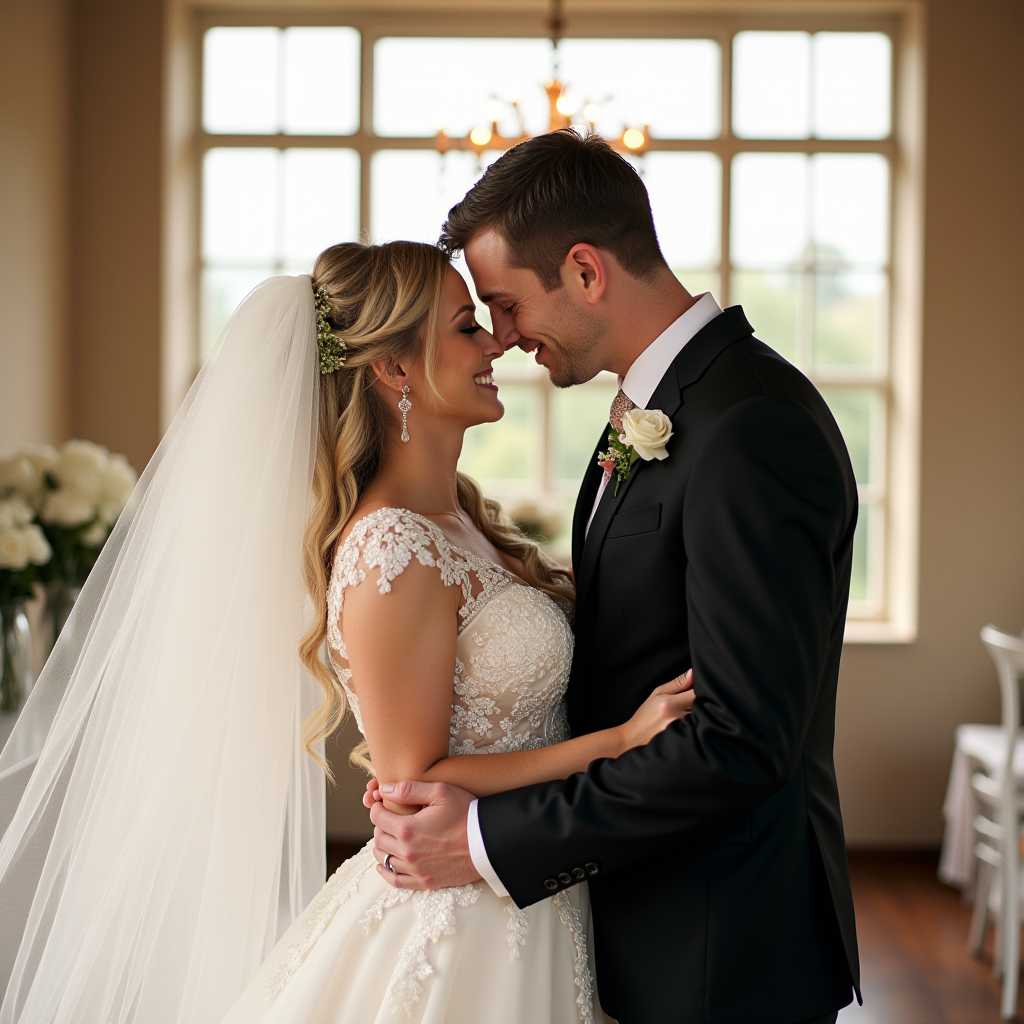} & \includegraphics[width=\imgwidth]{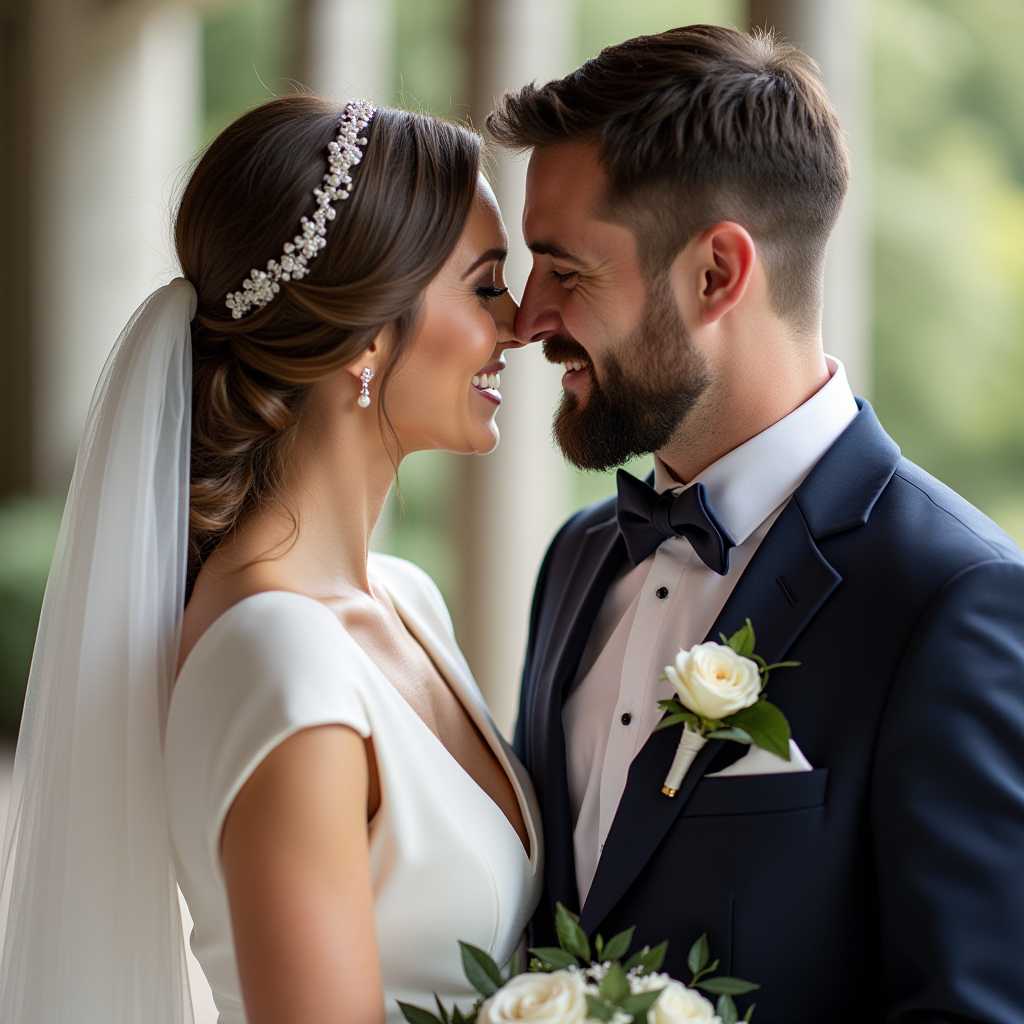} & \includegraphics[width=\imgwidth]{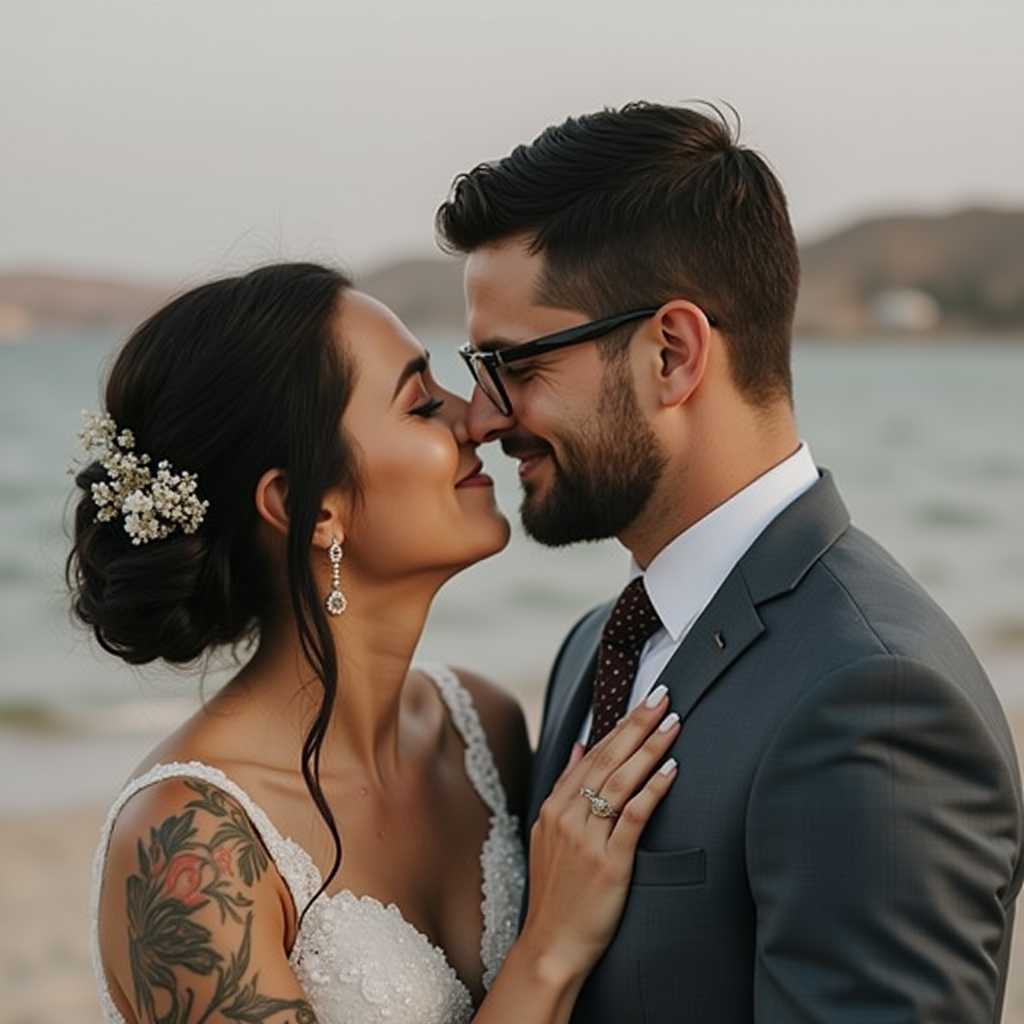} & \includegraphics[width=\imgwidth]{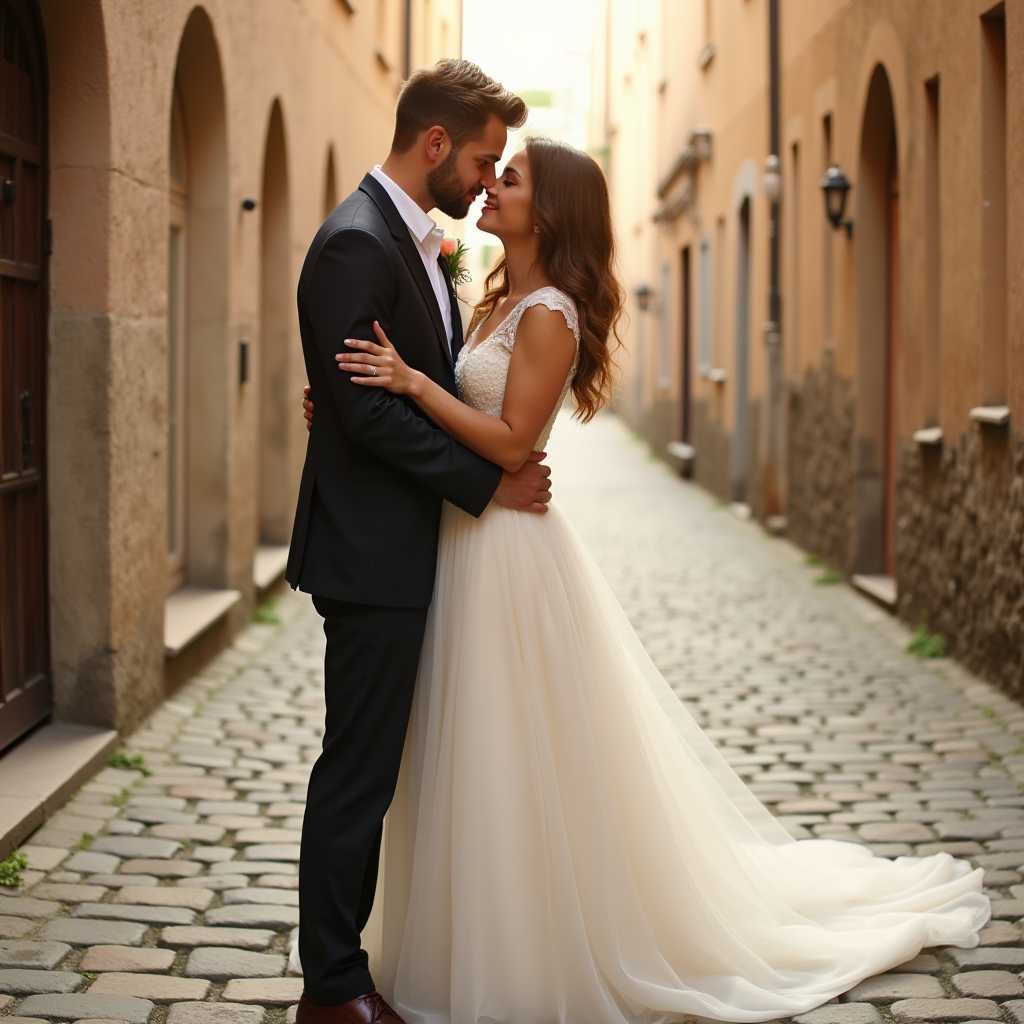} \\
        \multicolumn{9}{c}{\vspace{2pt}\small ``A wedding couple sharing a romantic moment'' \vspace{8pt}} \\

      \vertlabel{Flux} & 
        \includegraphics[width=\imgwidth, height=\imgwidth]{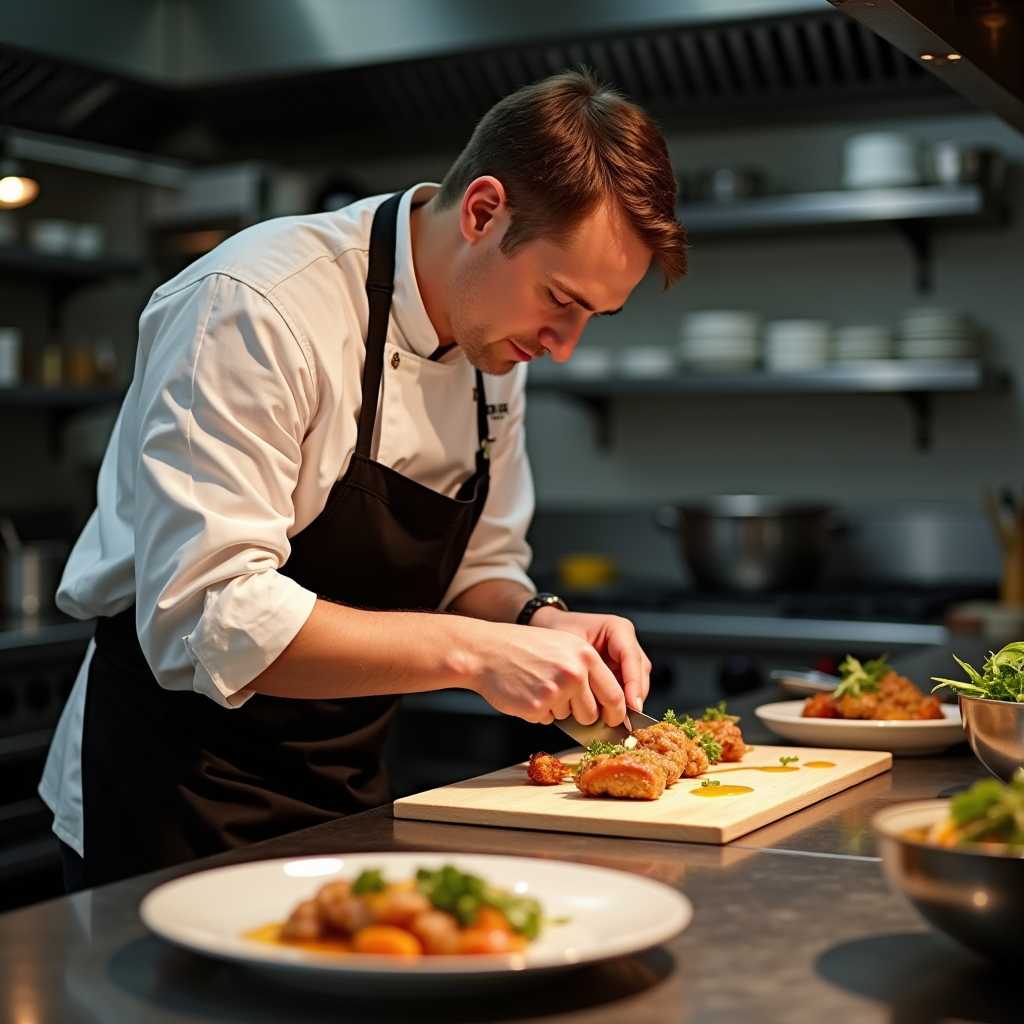} &
        \includegraphics[width=\imgwidth, height=\imgwidth]{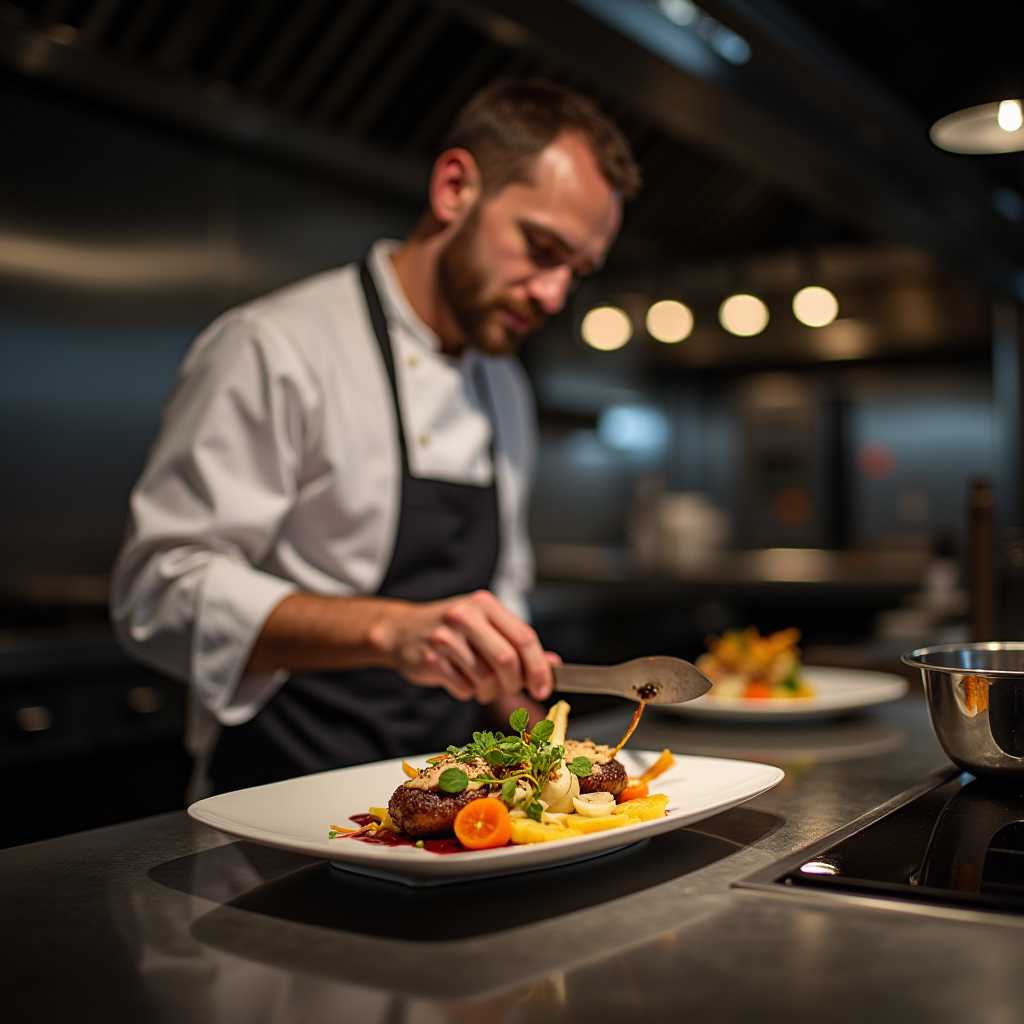} &
        \includegraphics[width=\imgwidth, height=\imgwidth]{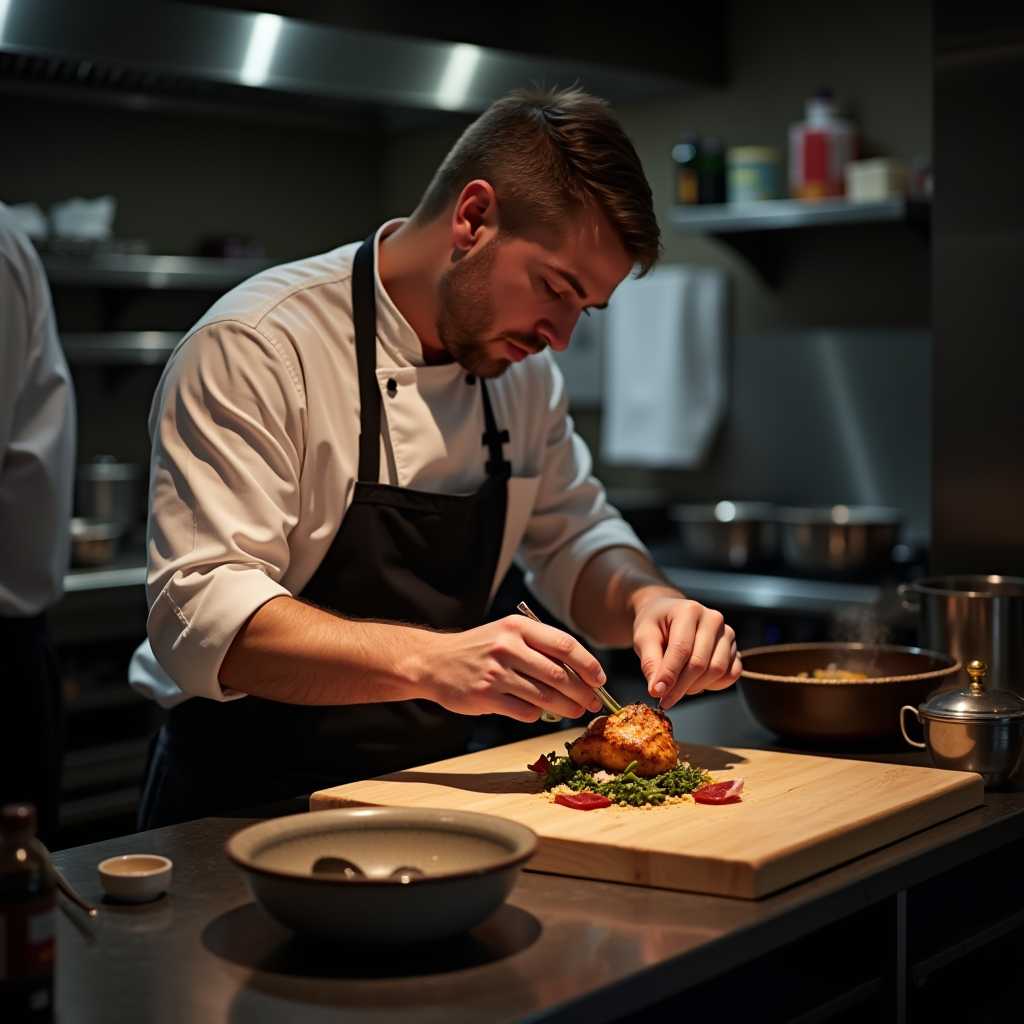} &
        \includegraphics[width=\imgwidth, height=\imgwidth]{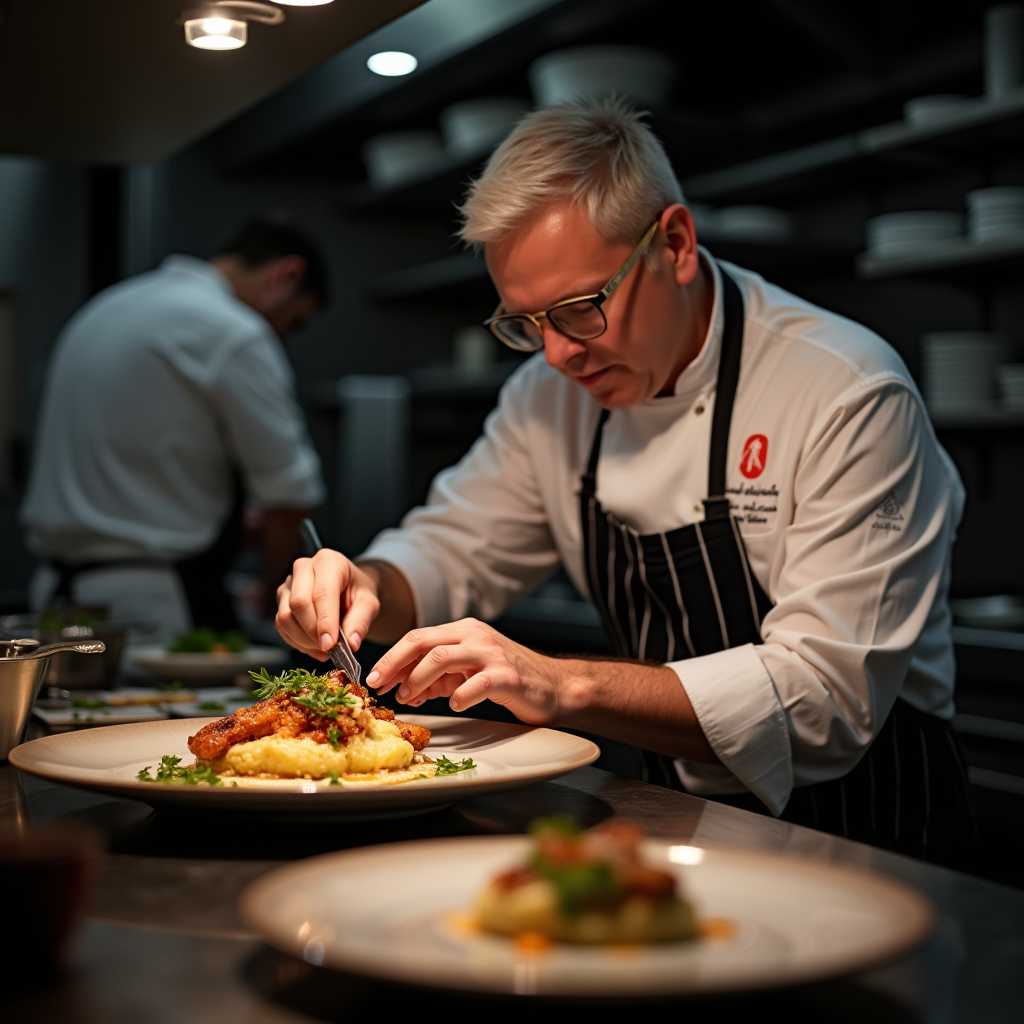} &
        \includegraphics[width=\imgwidth, height=\imgwidth]{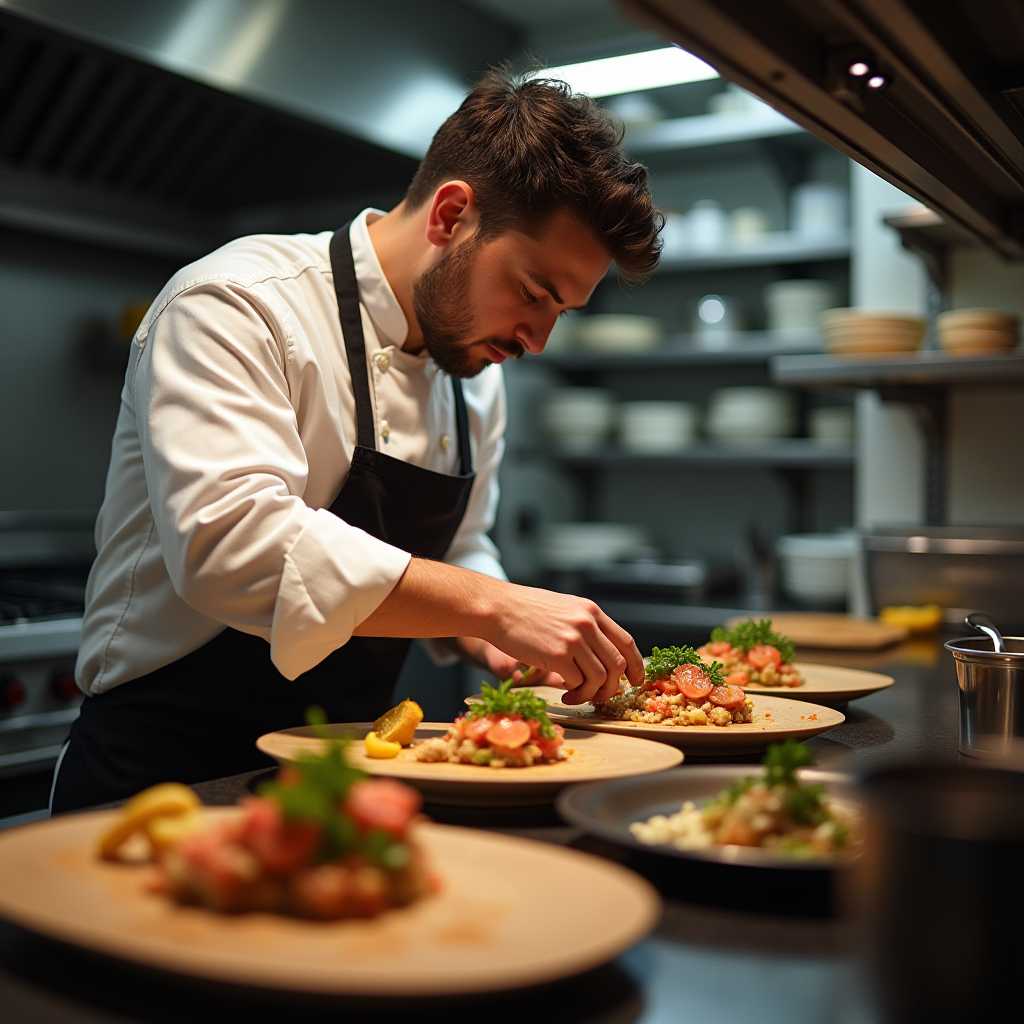} &
        \includegraphics[width=\imgwidth, height=\imgwidth]{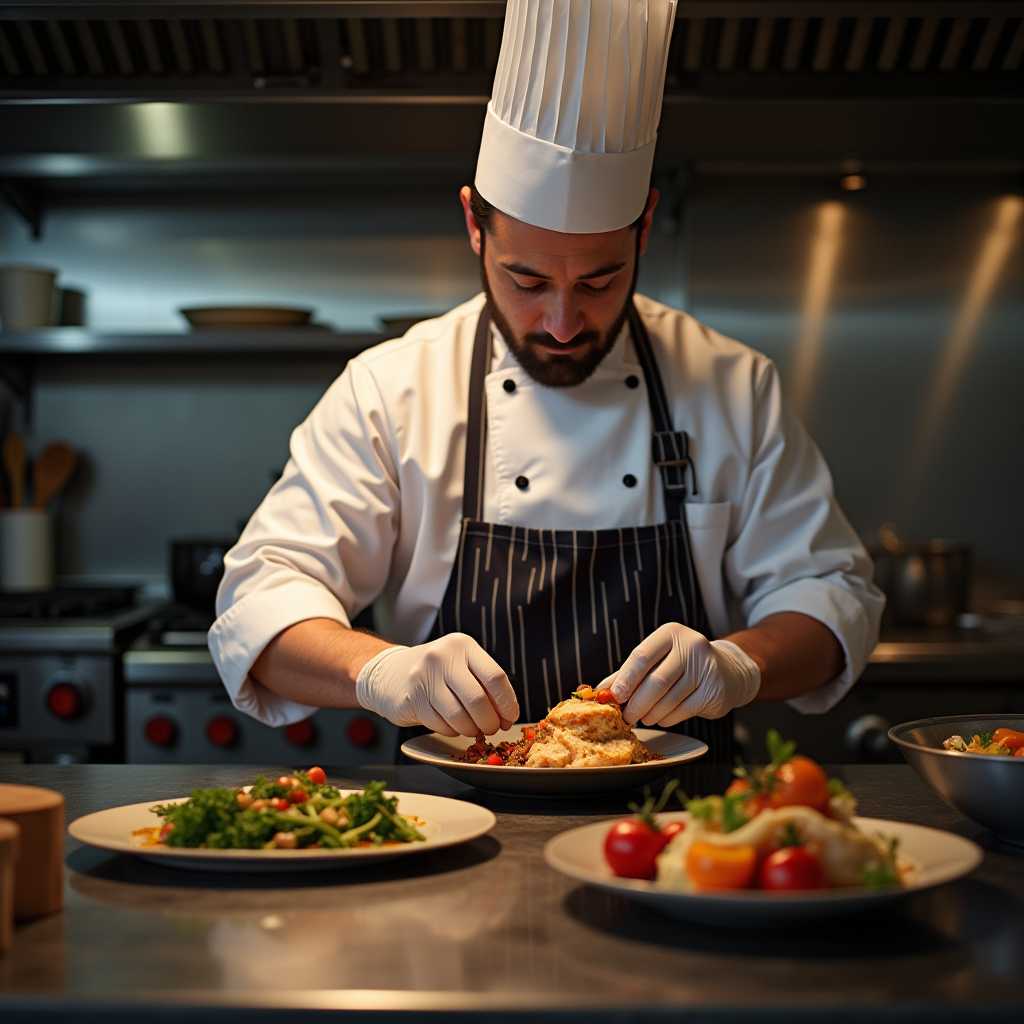} &
        \includegraphics[width=\imgwidth, height=\imgwidth]{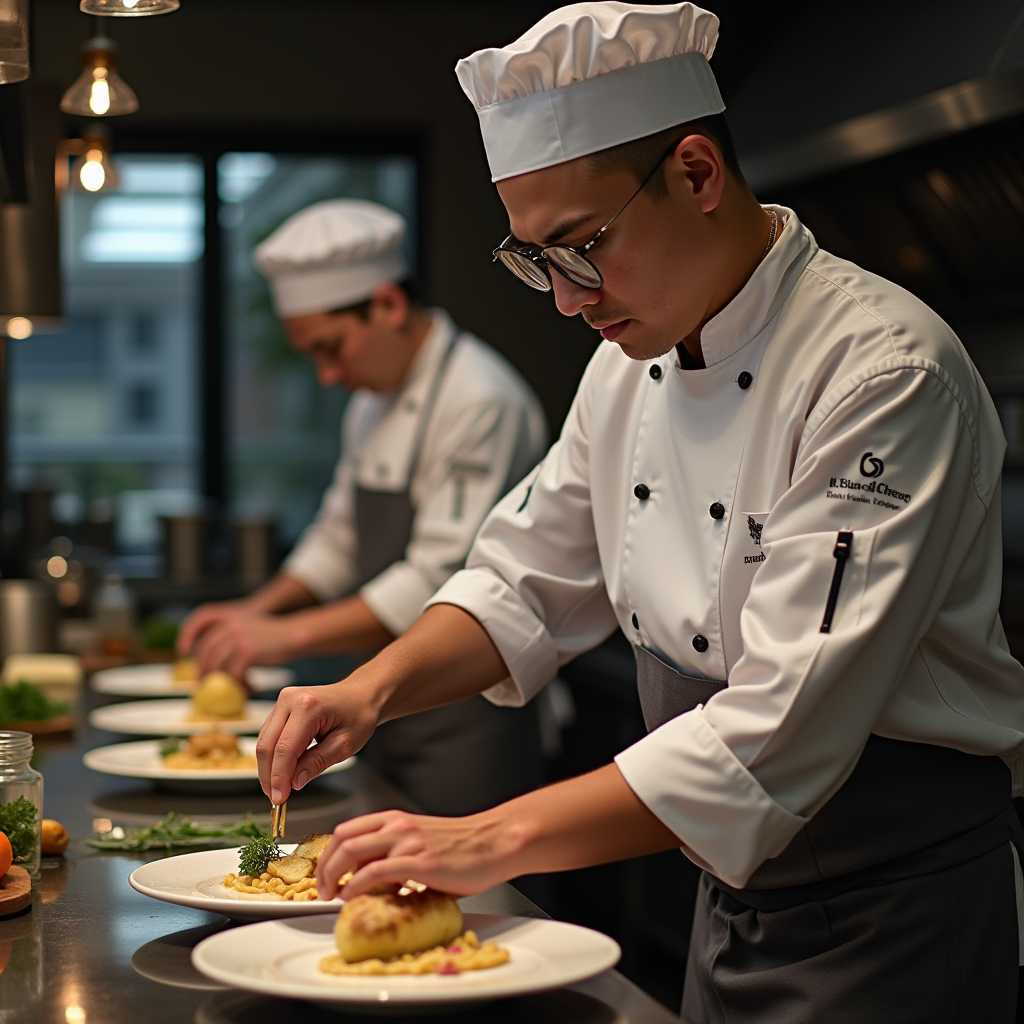} &
        \includegraphics[width=\imgwidth, height=\imgwidth]{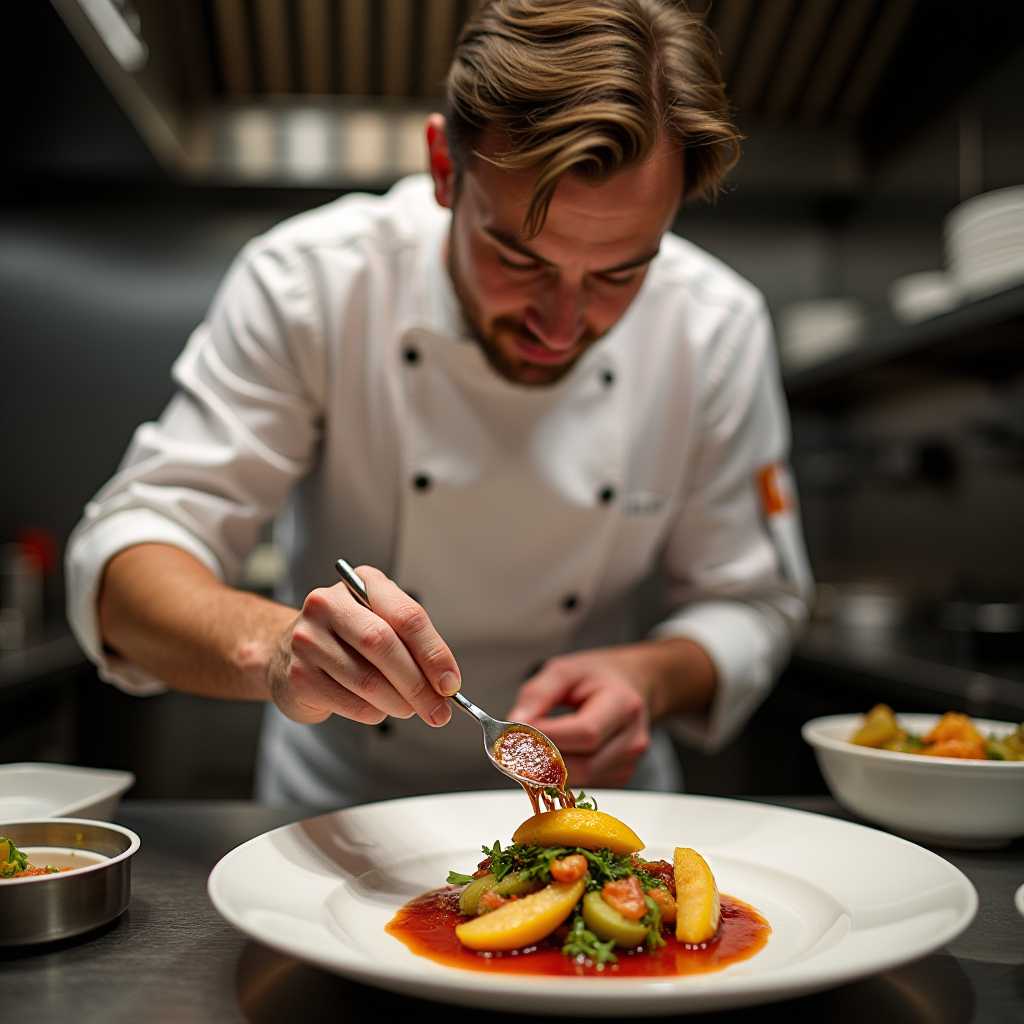} \\[-1pt]
        \vertlabel{Ours} & 
        \includegraphics[width=\imgwidth, height=\imgwidth]{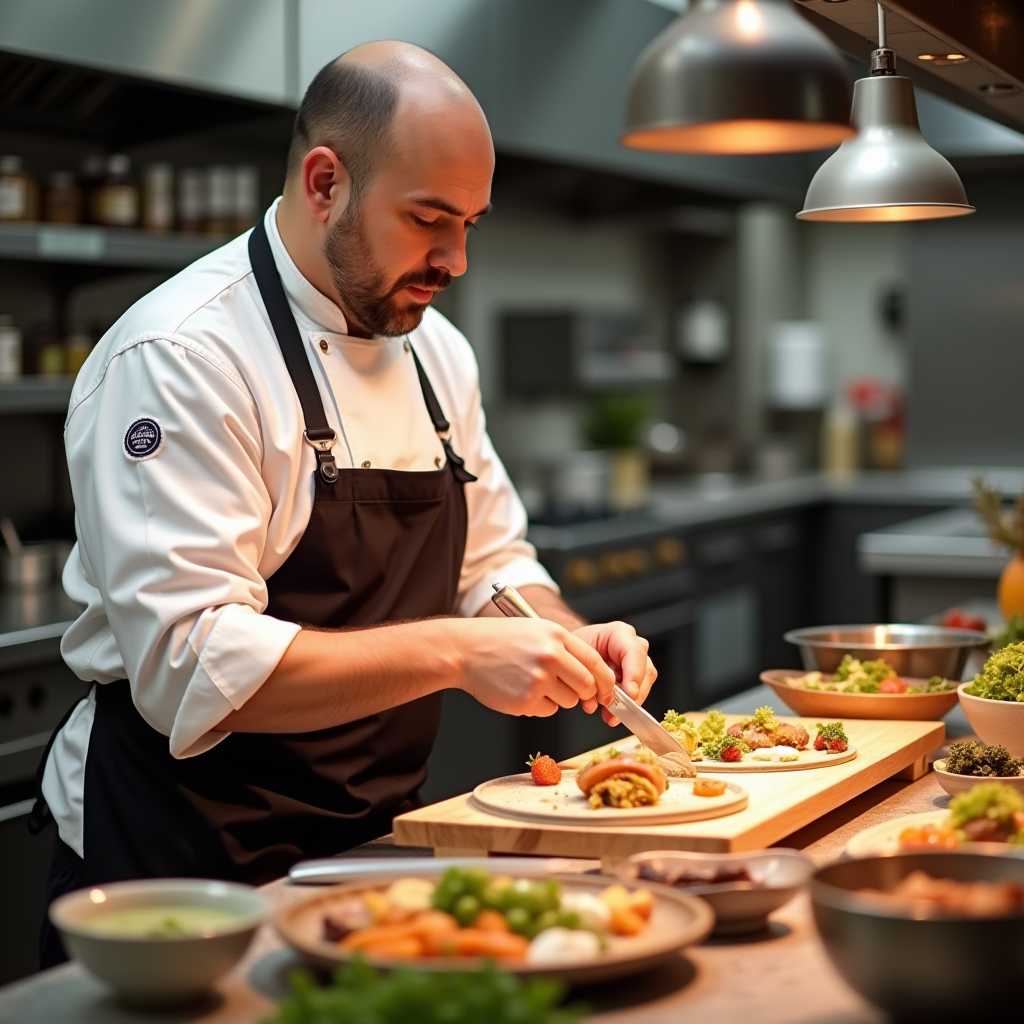} &
        \includegraphics[width=\imgwidth, height=\imgwidth]{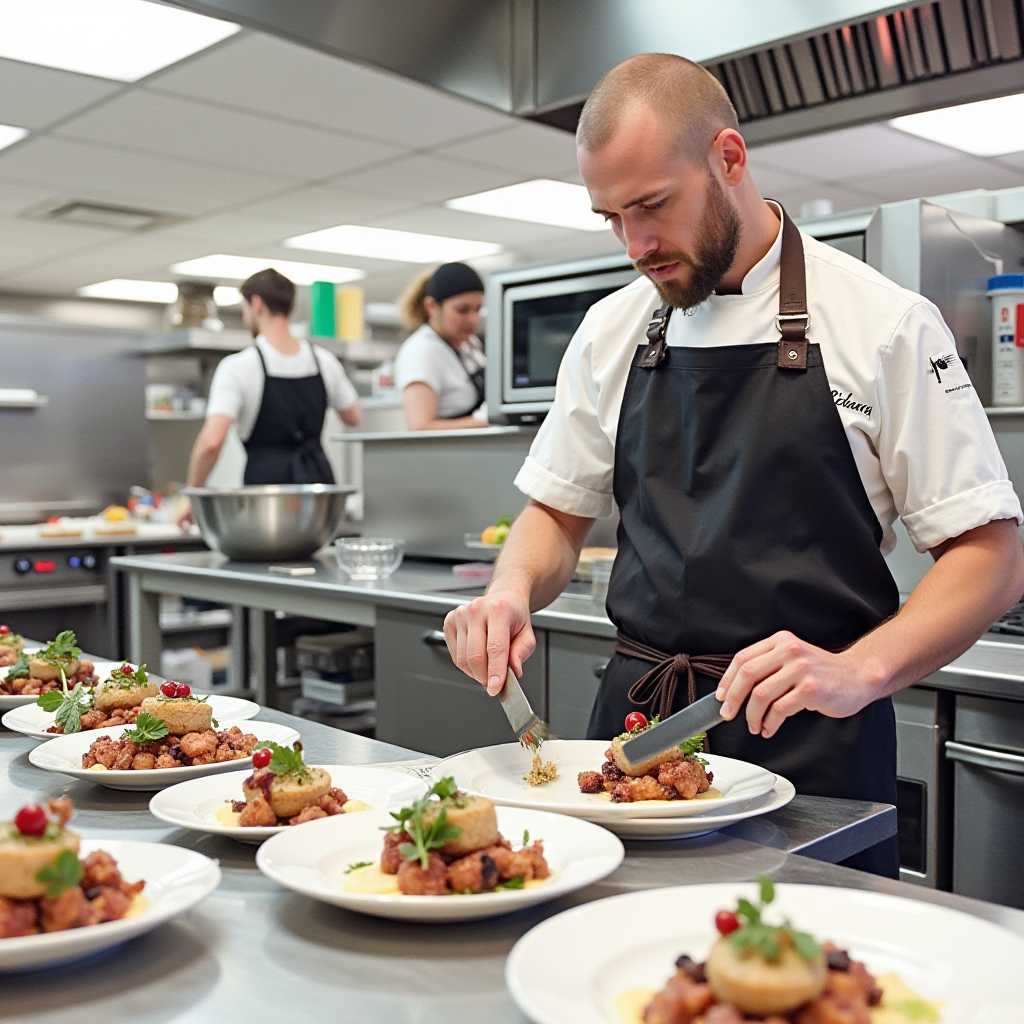} &
        \includegraphics[width=\imgwidth, height=\imgwidth]{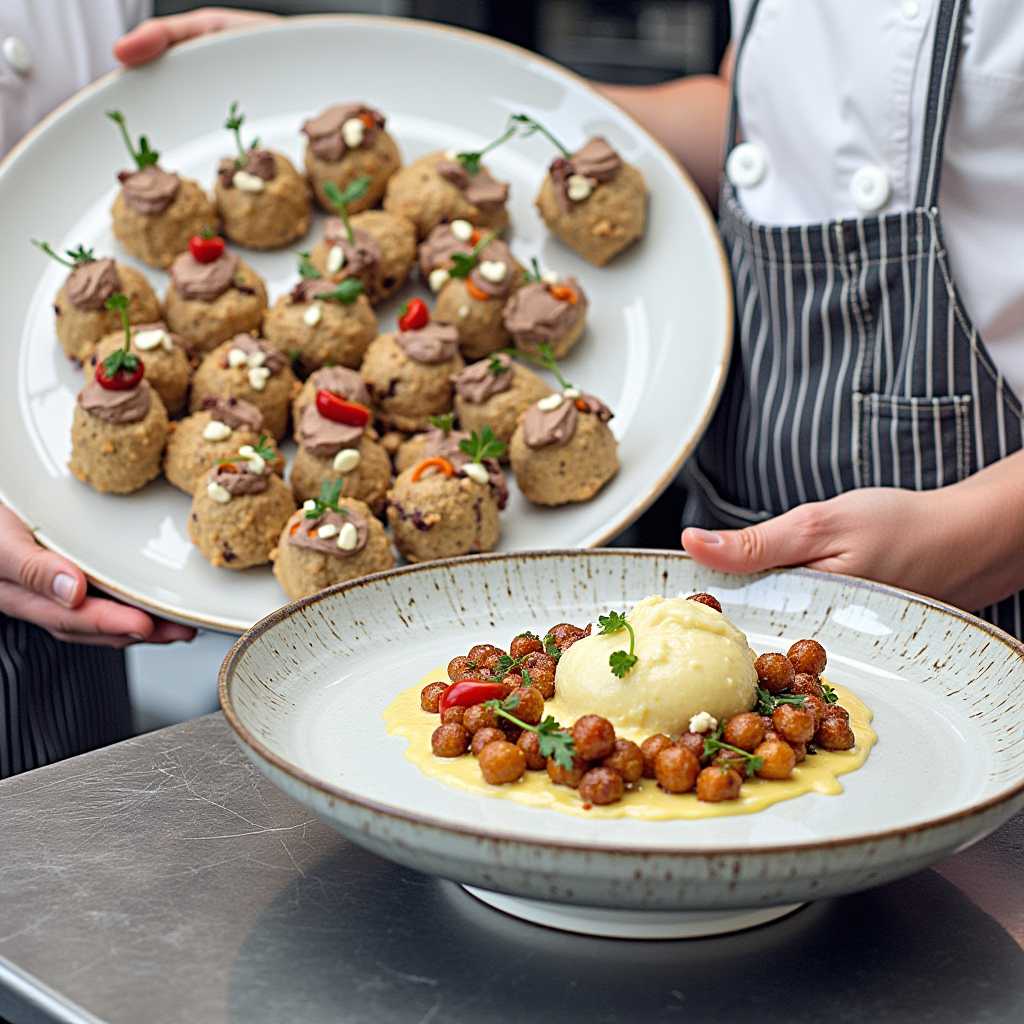} &
        \includegraphics[width=\imgwidth, height=\imgwidth]{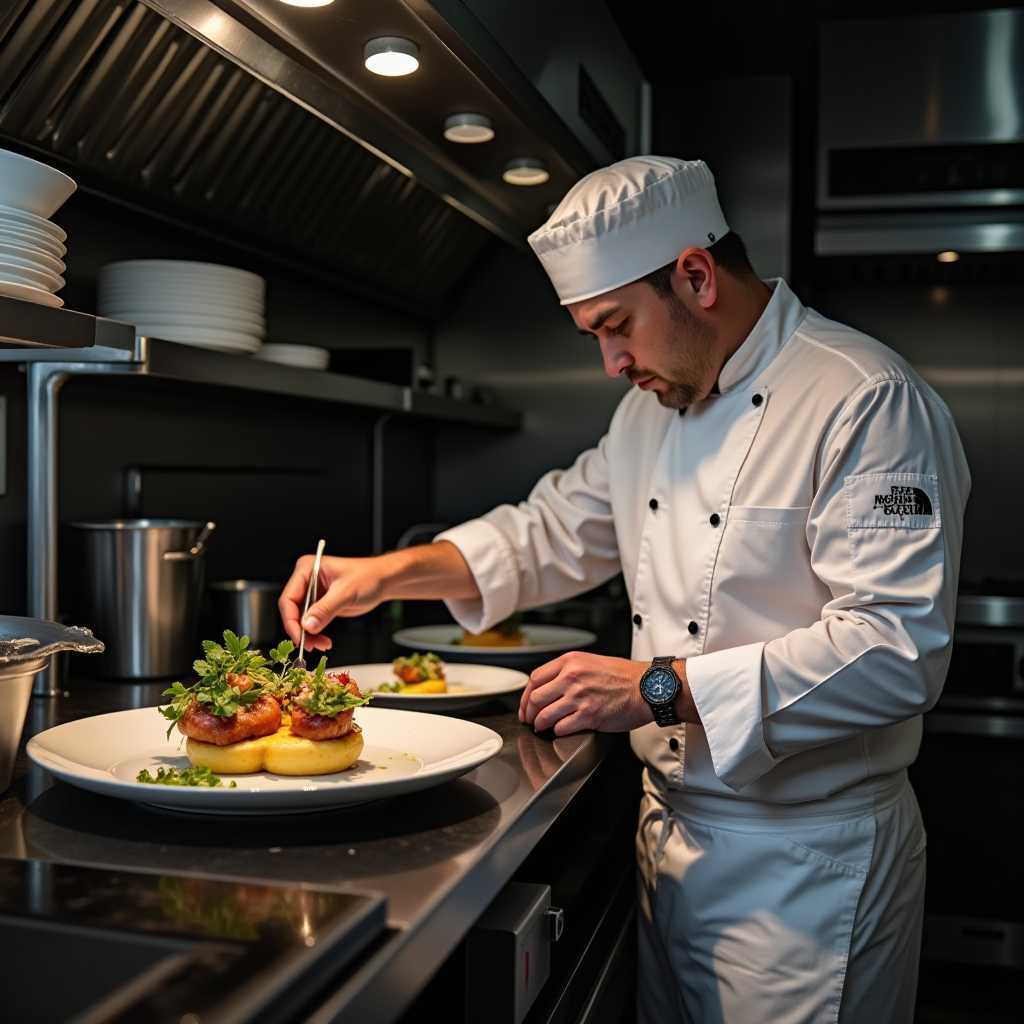} &
        \includegraphics[width=\imgwidth, height=\imgwidth]{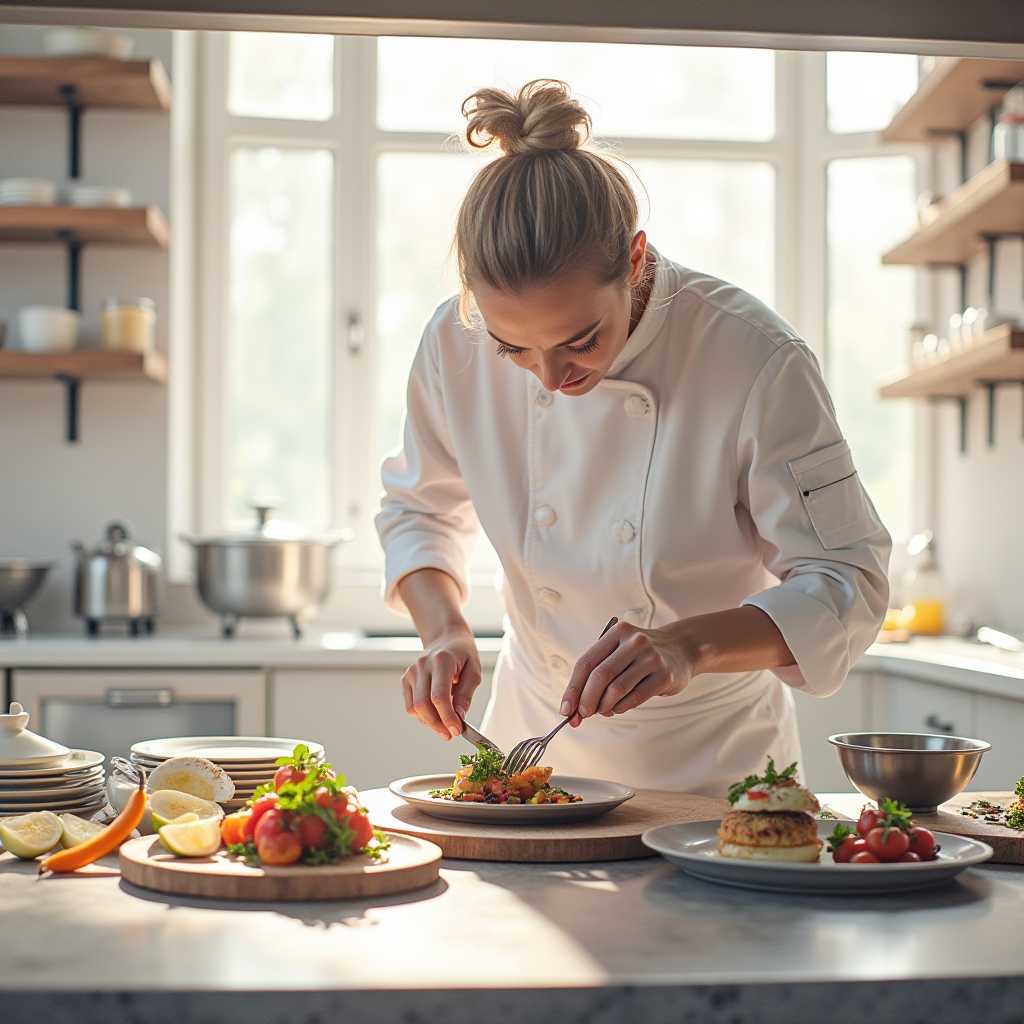} &
        \includegraphics[width=\imgwidth, height=\imgwidth]{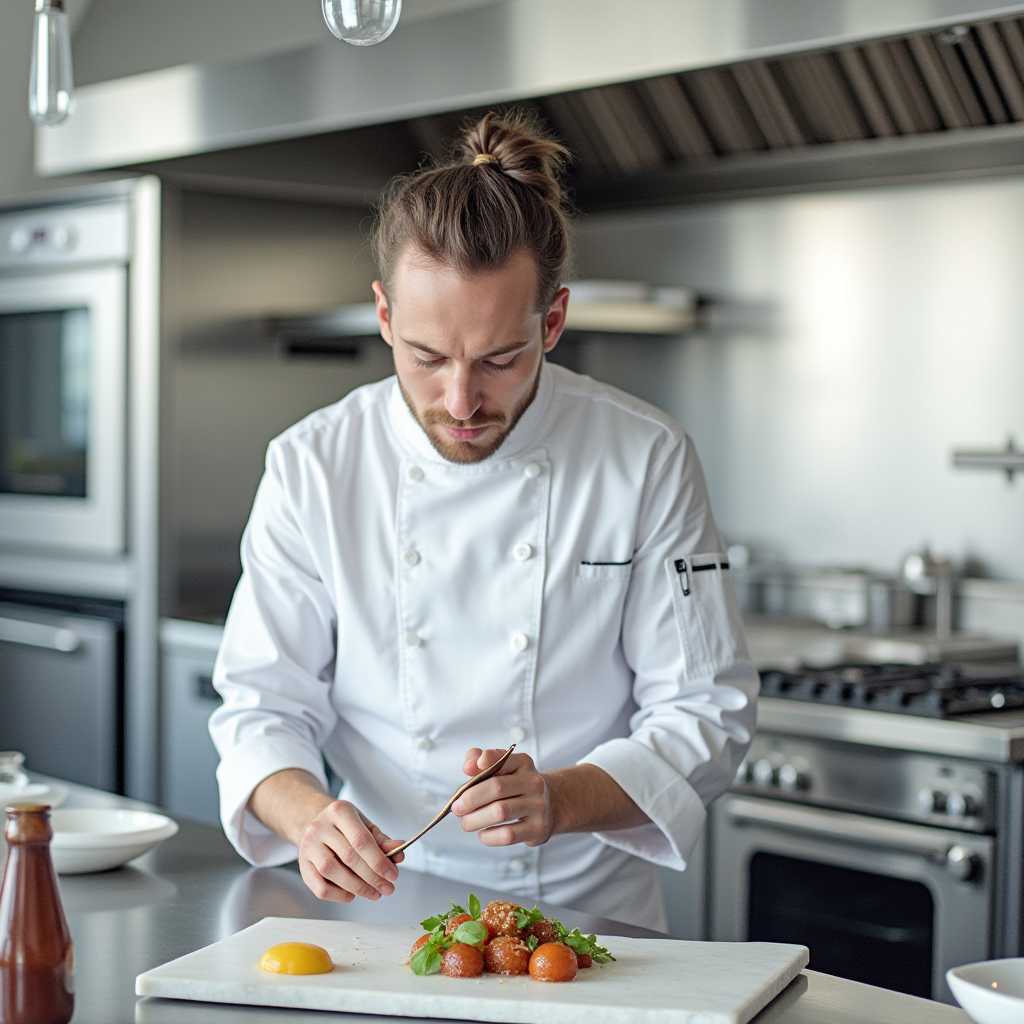} &
        \includegraphics[width=\imgwidth, height=\imgwidth]{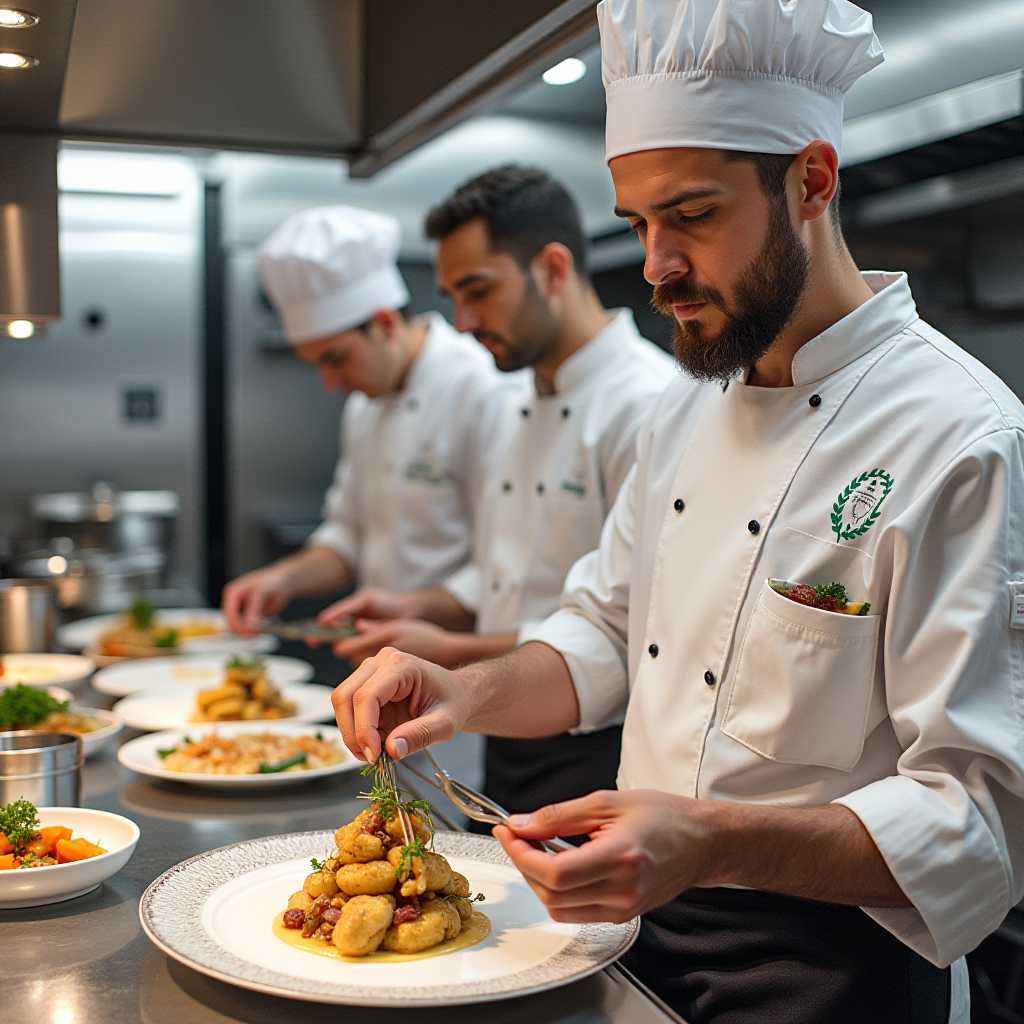} &
        \includegraphics[width=\imgwidth, height=\imgwidth]{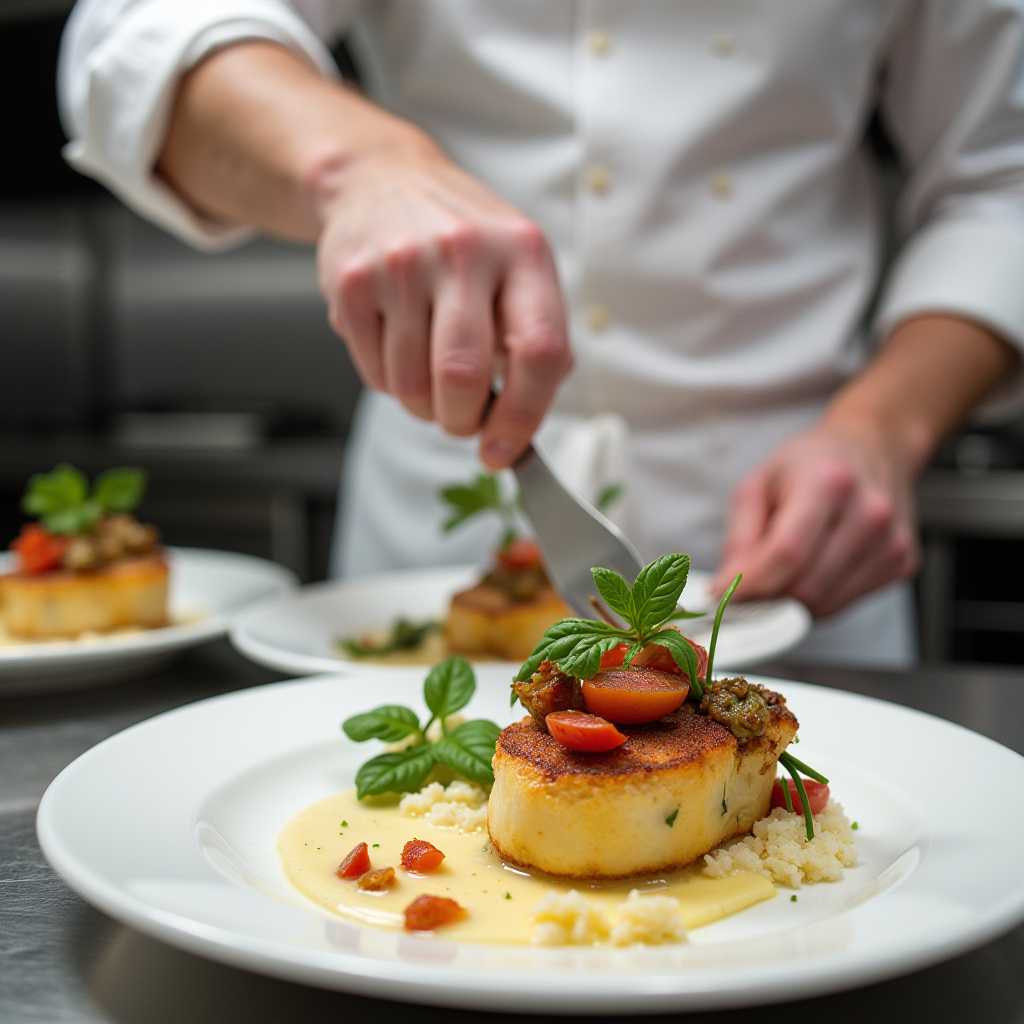} \\
        \multicolumn{9}{c}{\vspace{2pt}\small ``A chef preparing a gourmet meal in a professional kitchen'' \vspace{8pt}} \\
    \end{tabular}
    \caption{\textbf{Additional qualitative results on Flux-dev.} All batches were generated using the same random seed initialization.}\label{fig:extra_res2}
\end{figure*}

\begin{figure*}
    \centering
    \setlength{\tabcolsep}{0.5pt} 
    \renewcommand{\arraystretch}{0.5} 
    \newcommand{\imgwidth}{0.12\textwidth}
    
    \newcommand{\vertlabel}[1]{\raisebox{2.5em}{\rotatebox{90}{\scriptsize\textbf{#1}}}}

    \begin{tabular}{c c c c c c c c c}
        
        \vertlabel{Flux} & 
        \includegraphics[width=\imgwidth, height=\imgwidth]{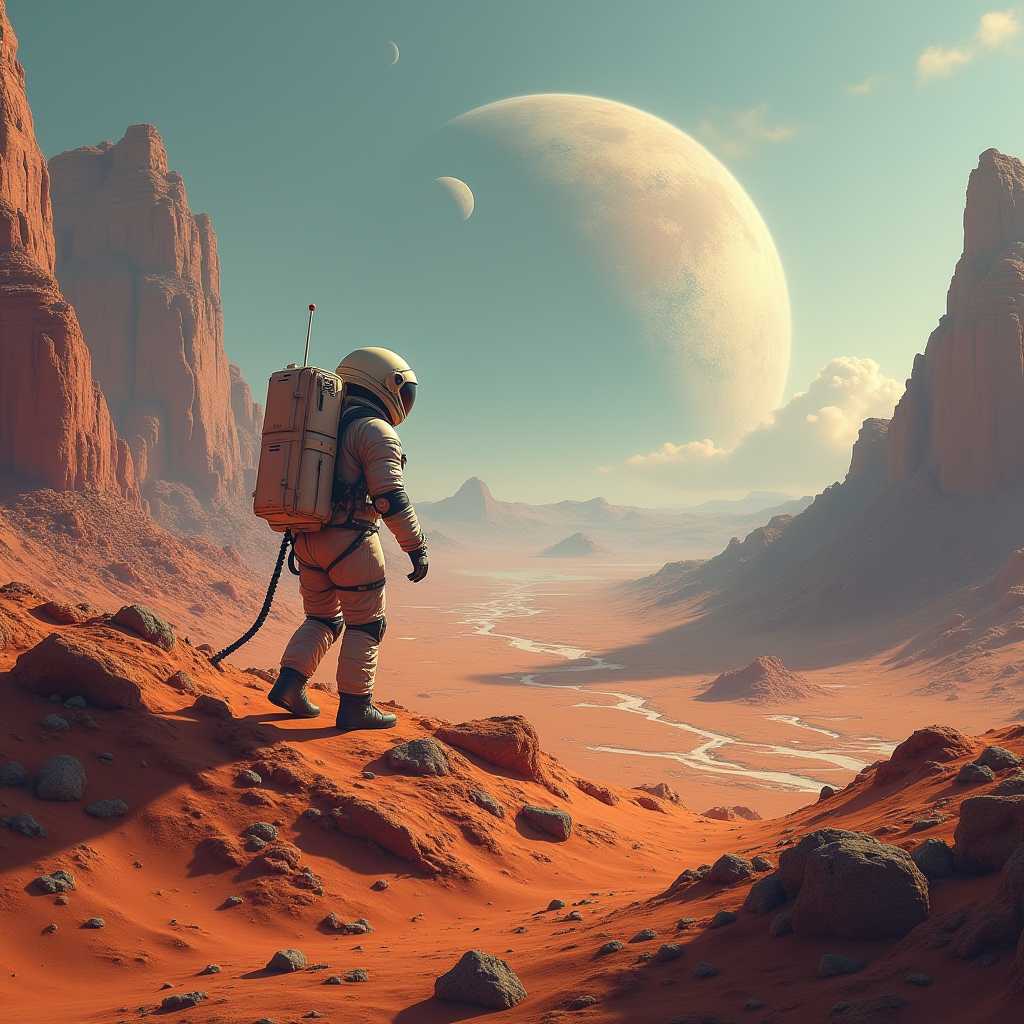} &
        \includegraphics[width=\imgwidth, height=\imgwidth]{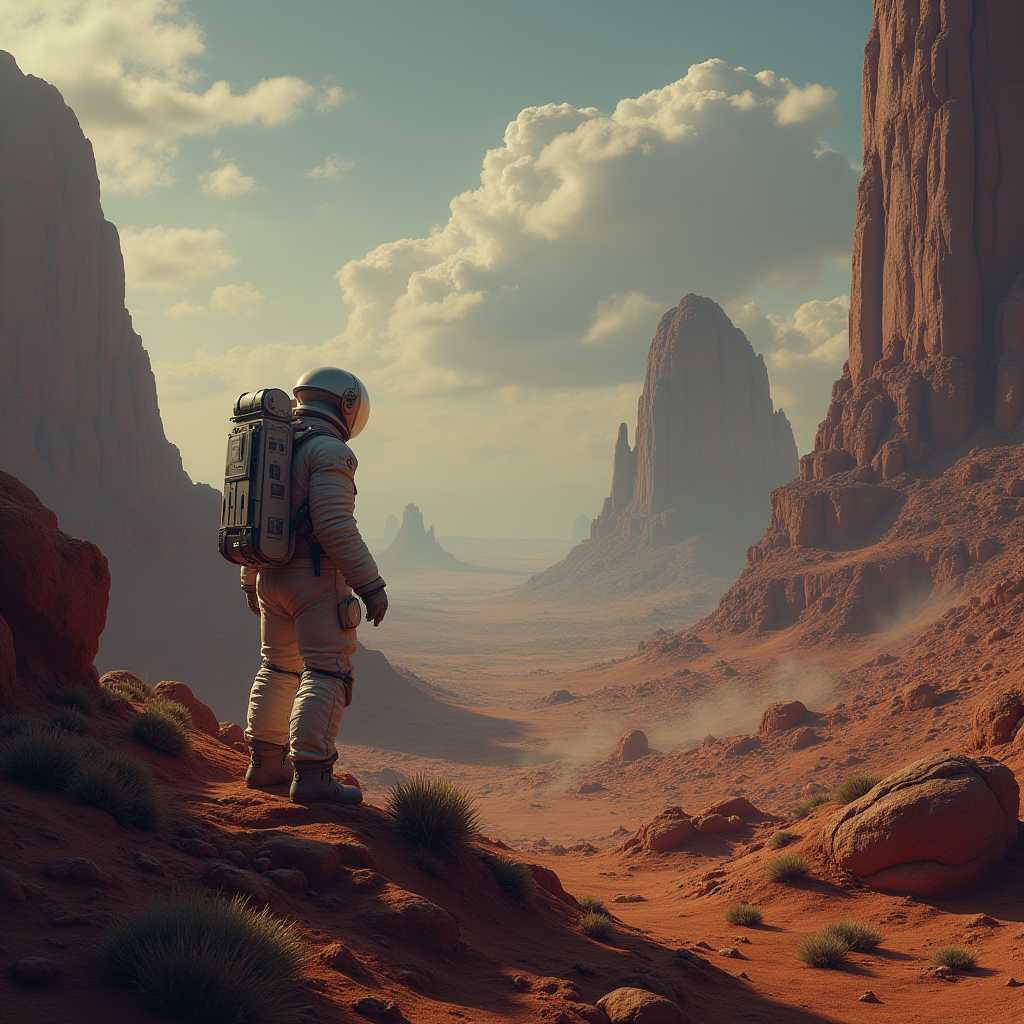} &
        \includegraphics[width=\imgwidth, height=\imgwidth]{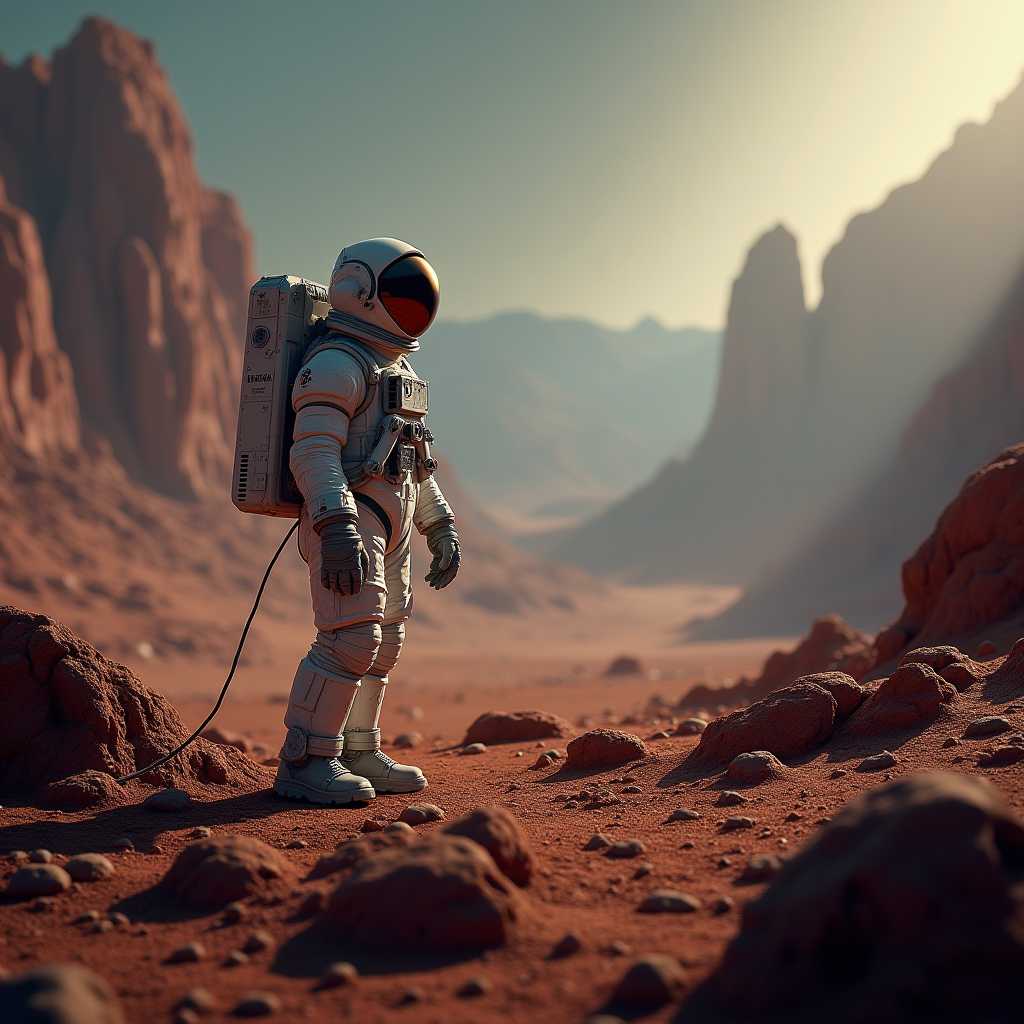} &
        \includegraphics[width=\imgwidth, height=\imgwidth]{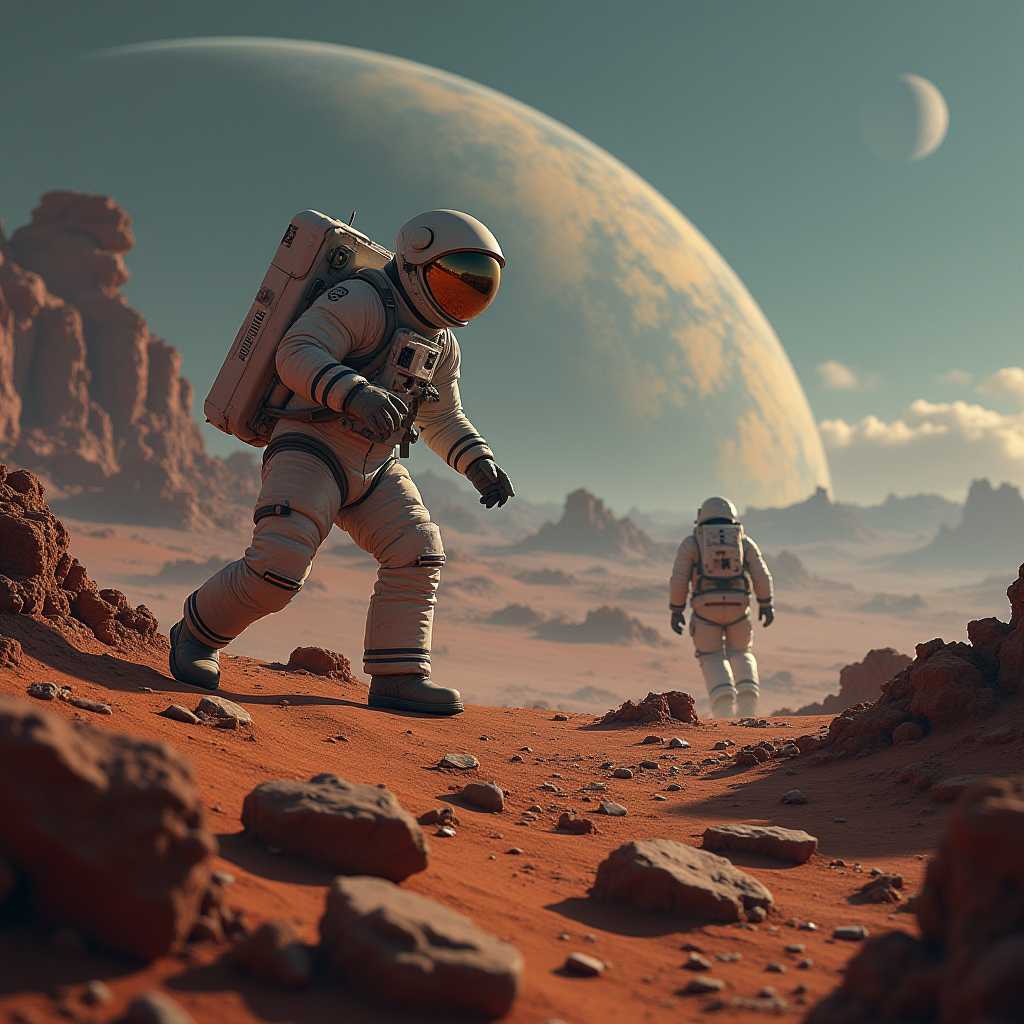} &
        \includegraphics[width=\imgwidth, height=\imgwidth]{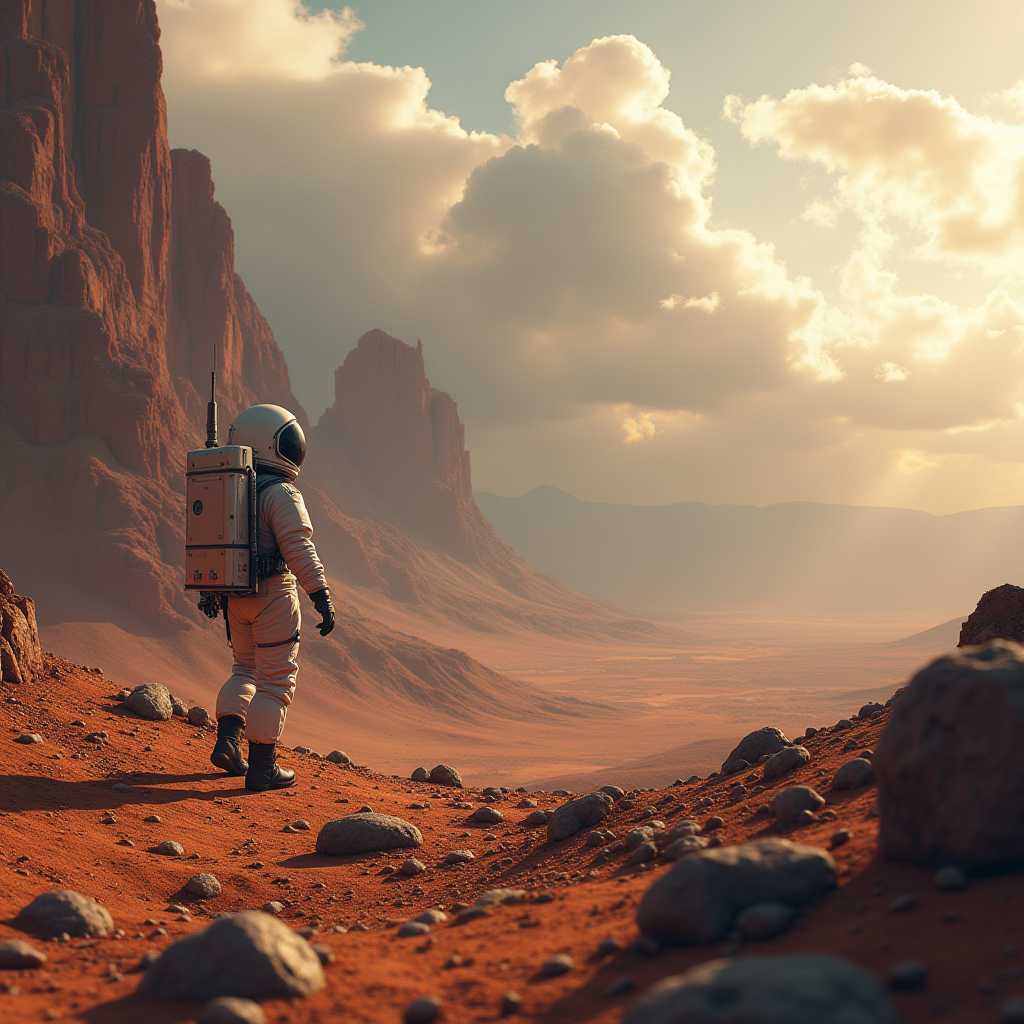} &
        \includegraphics[width=\imgwidth, height=\imgwidth]{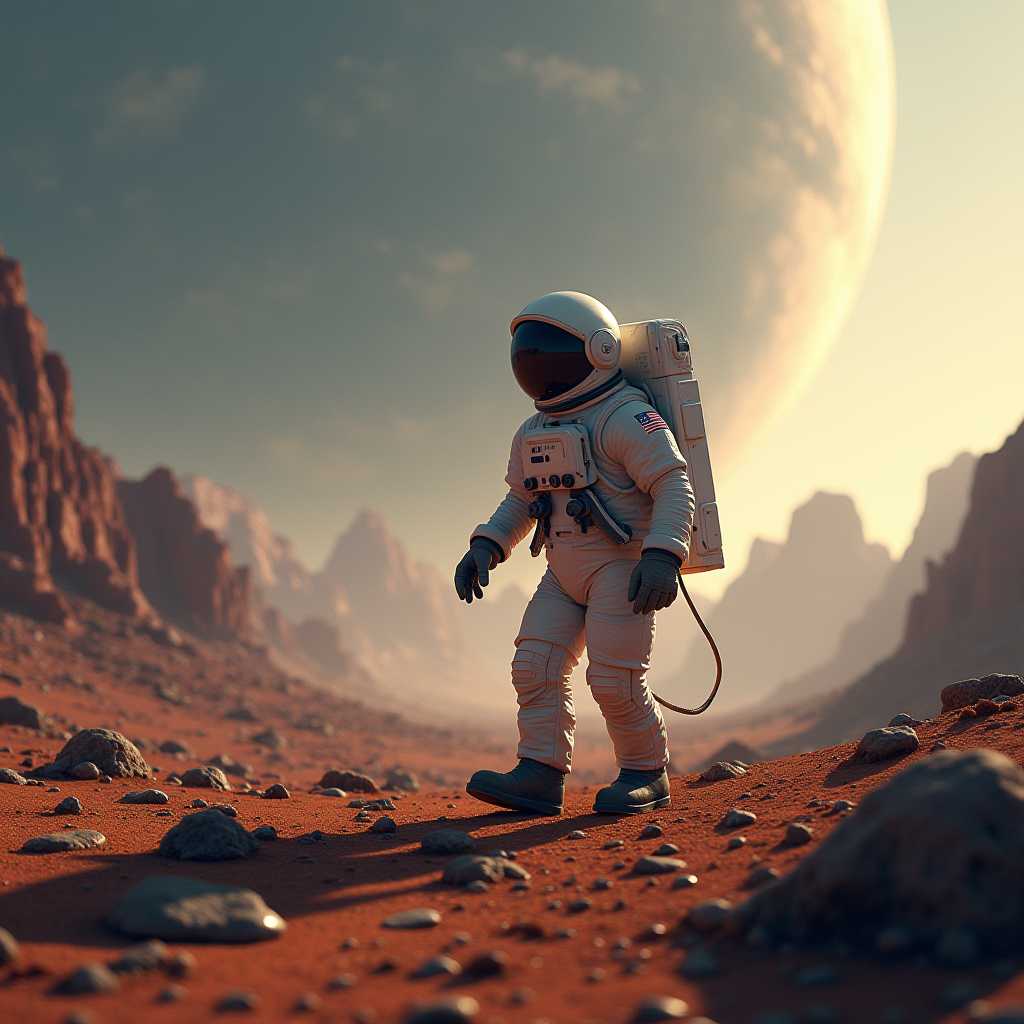} &
        \includegraphics[width=\imgwidth, height=\imgwidth]{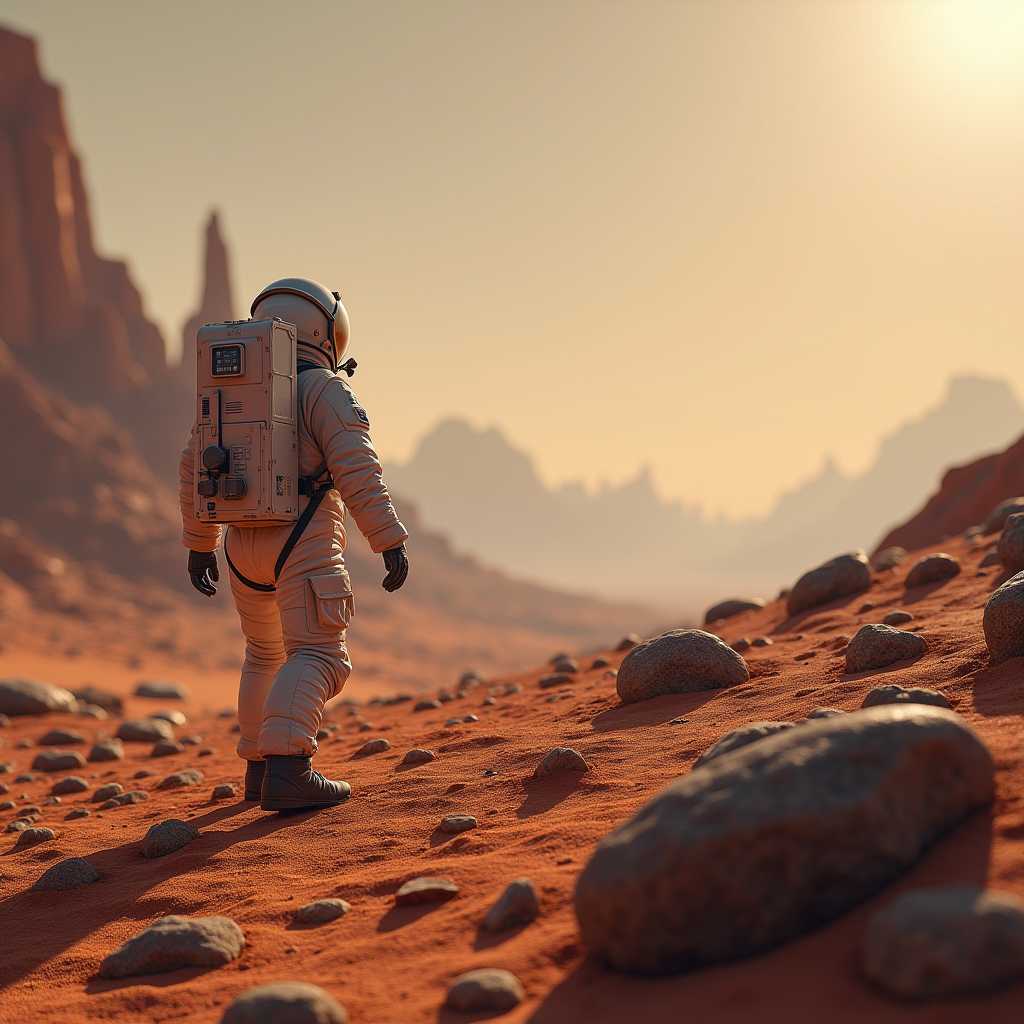} &
        \includegraphics[width=\imgwidth, height=\imgwidth]{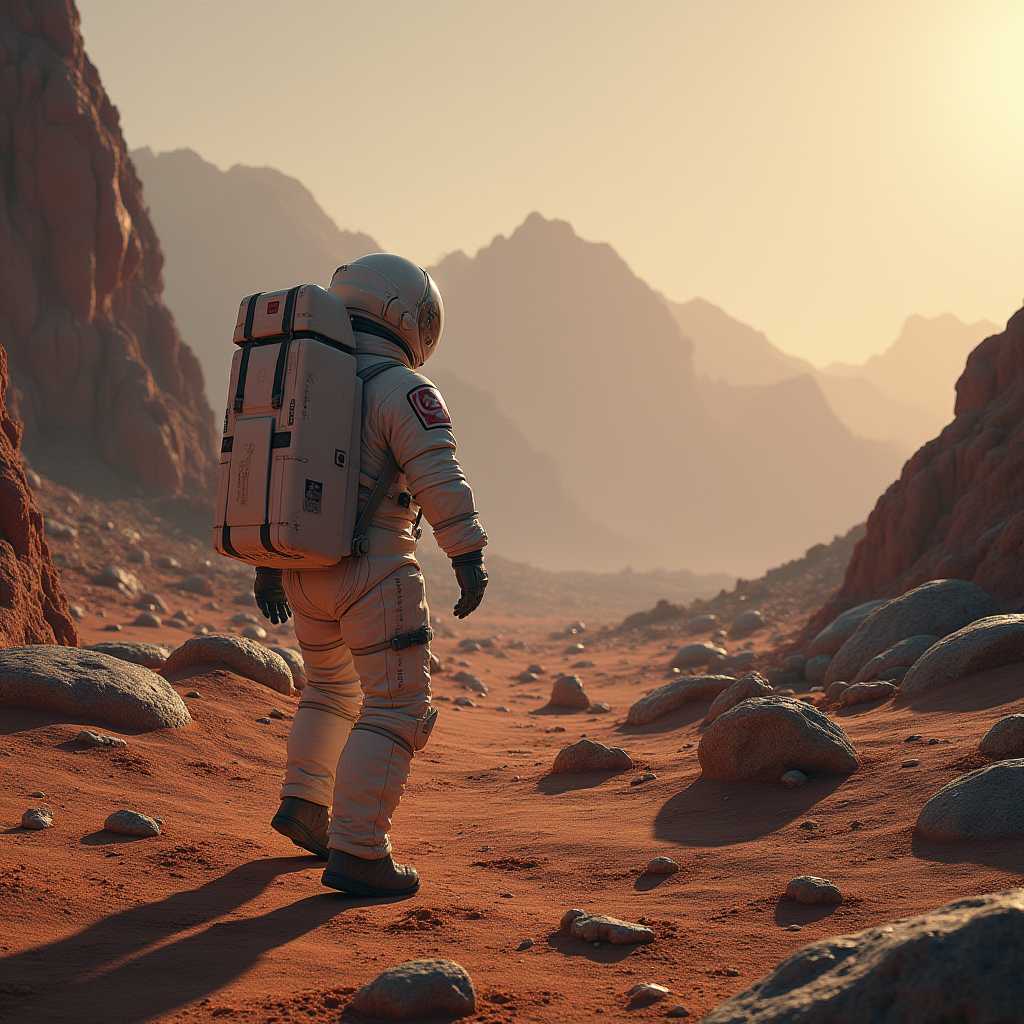} \\[-1pt]

        \vertlabel{Ours} & 
        \includegraphics[width=\imgwidth, height=\imgwidth]{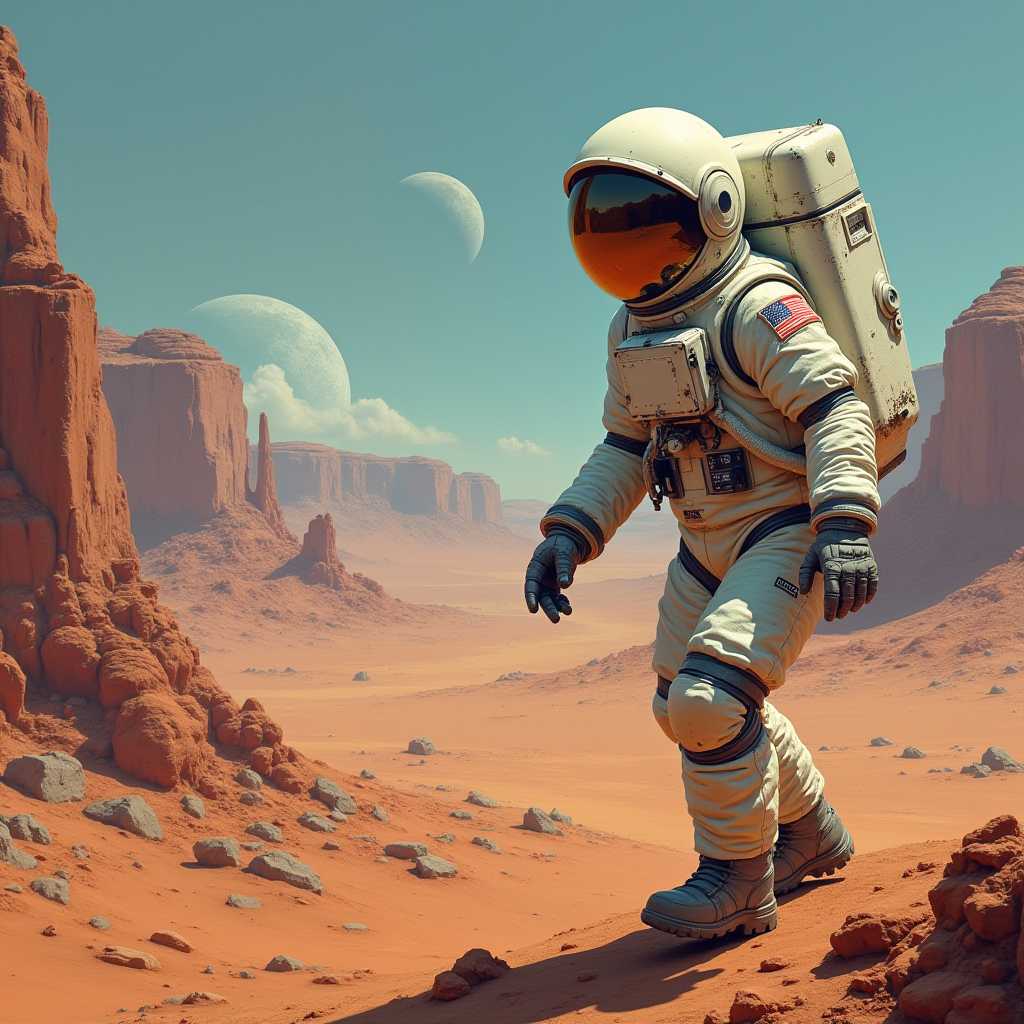} &
        \includegraphics[width=\imgwidth, height=\imgwidth]{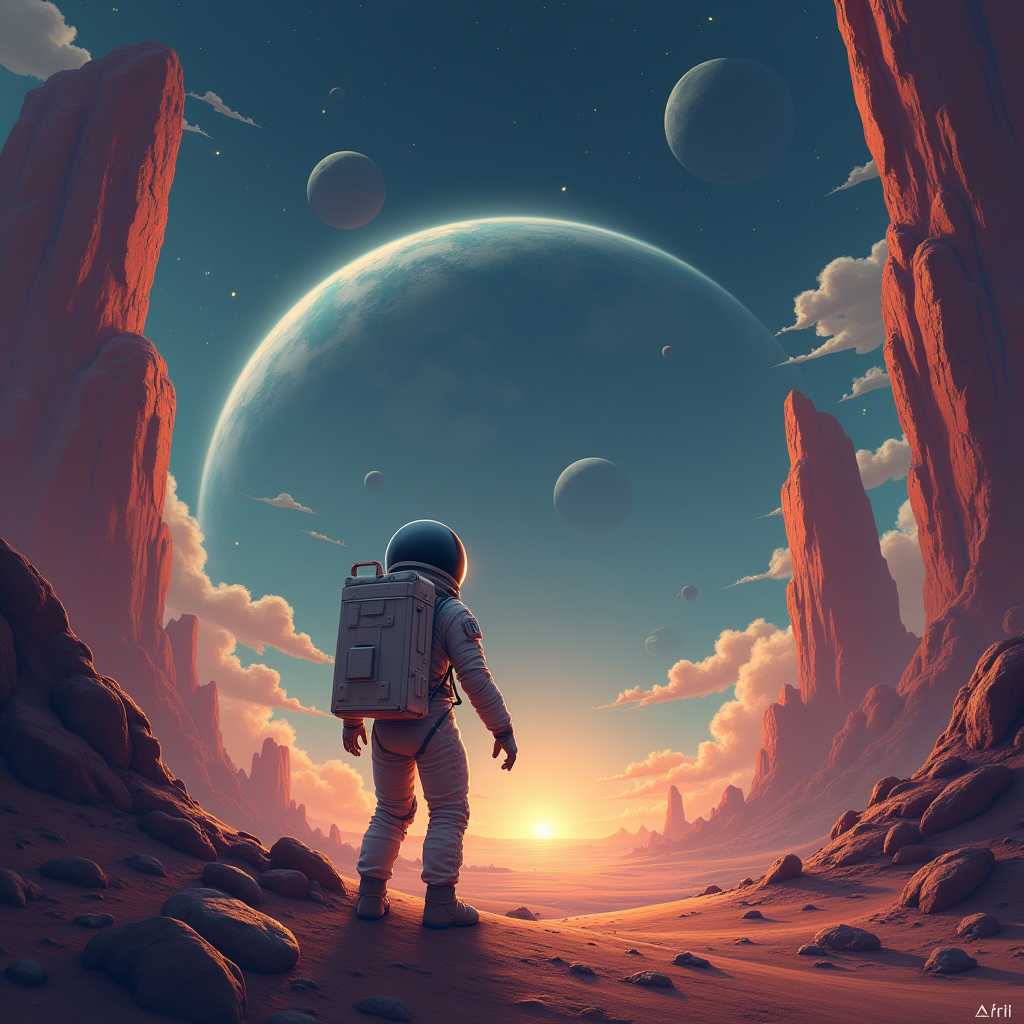} &
        \includegraphics[width=\imgwidth, height=\imgwidth]{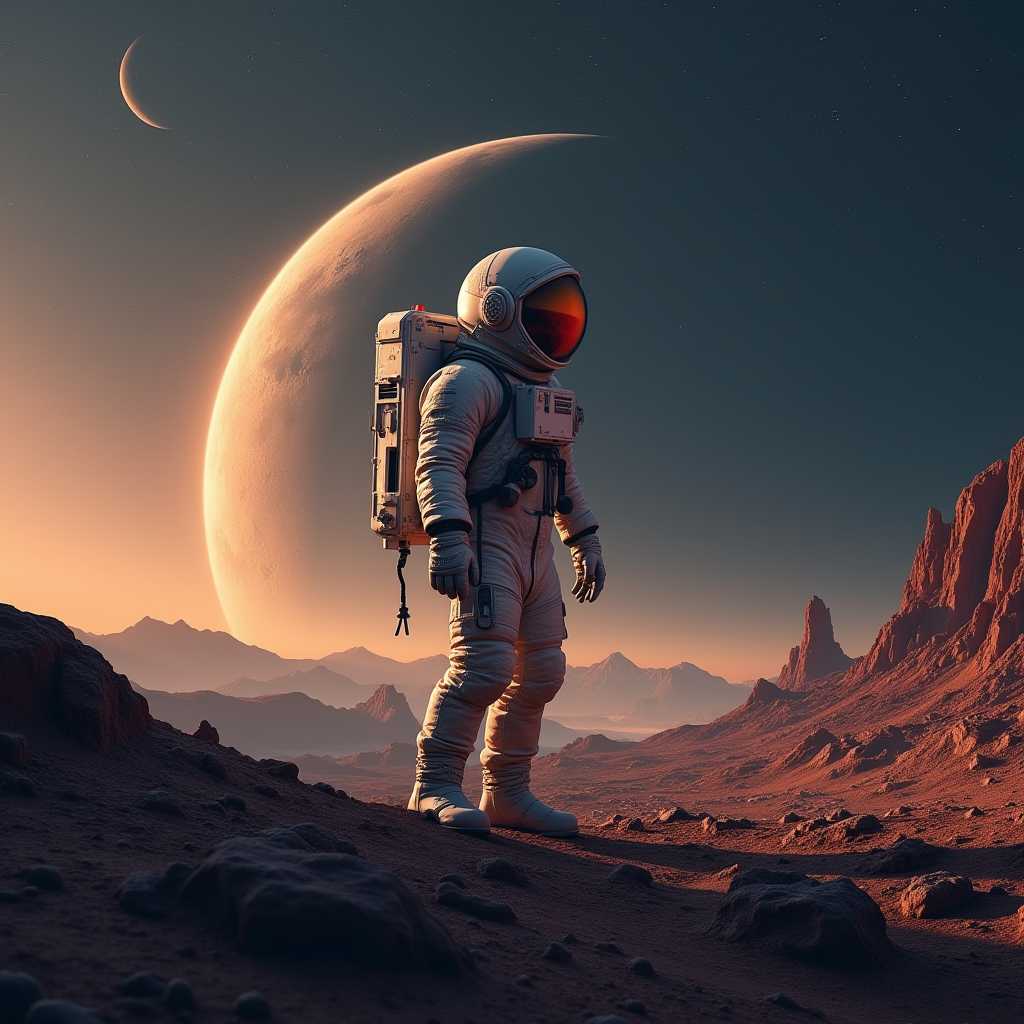} &
        \includegraphics[width=\imgwidth, height=\imgwidth]{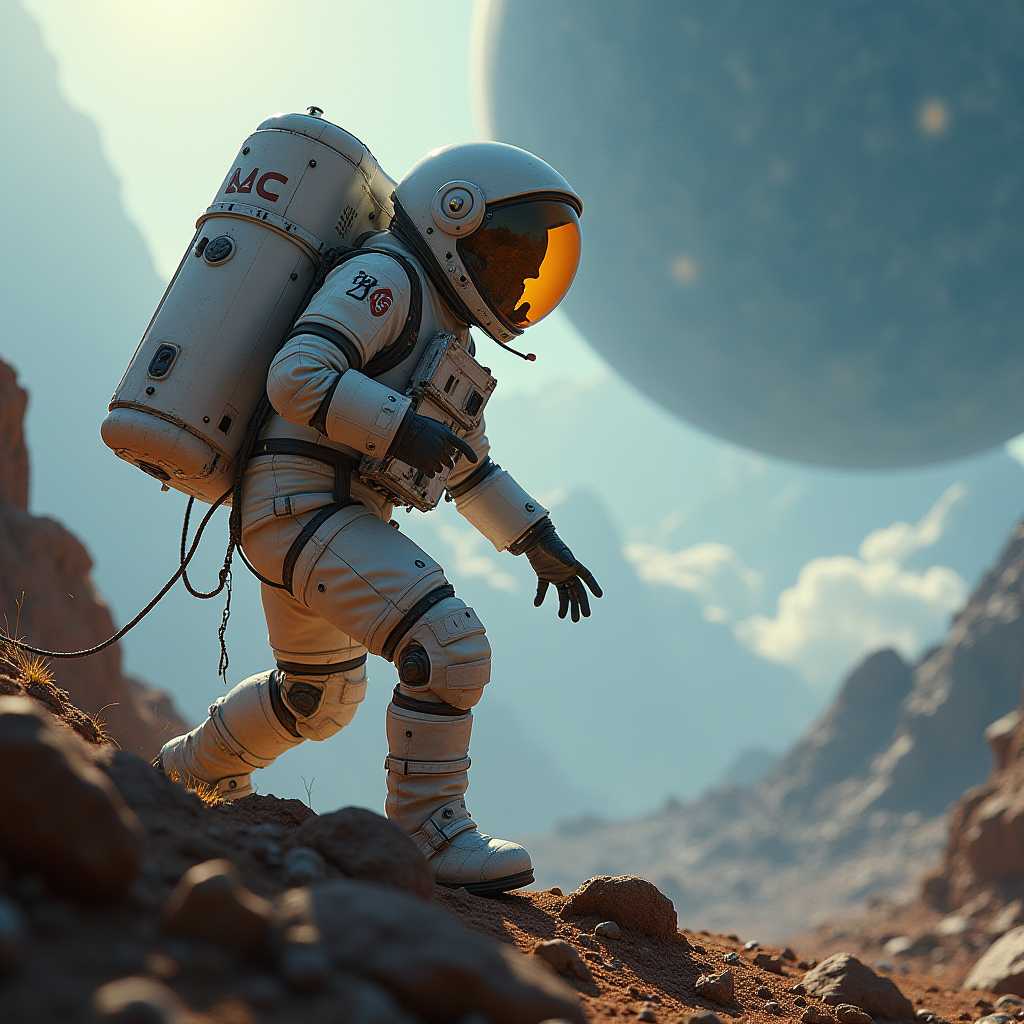} &
        \includegraphics[width=\imgwidth, height=\imgwidth]{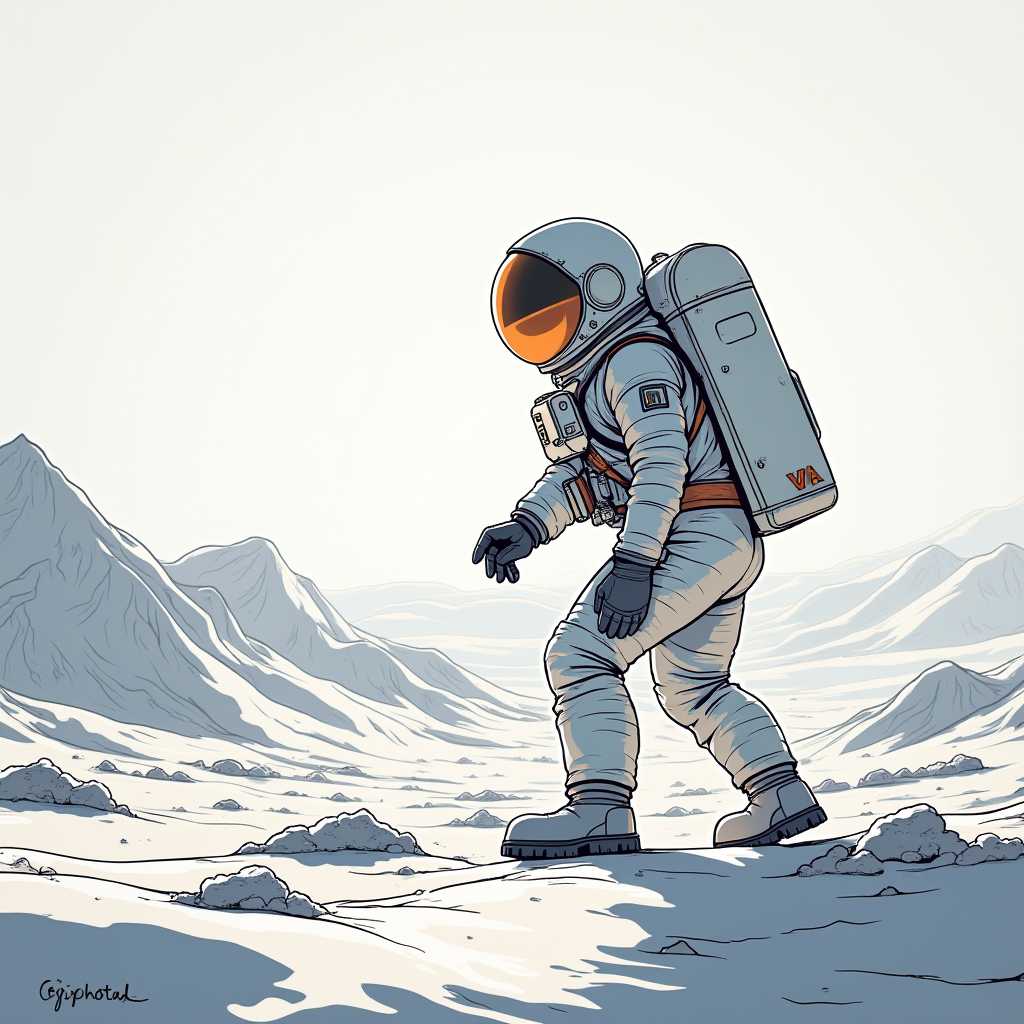} &
        \includegraphics[width=\imgwidth, height=\imgwidth]{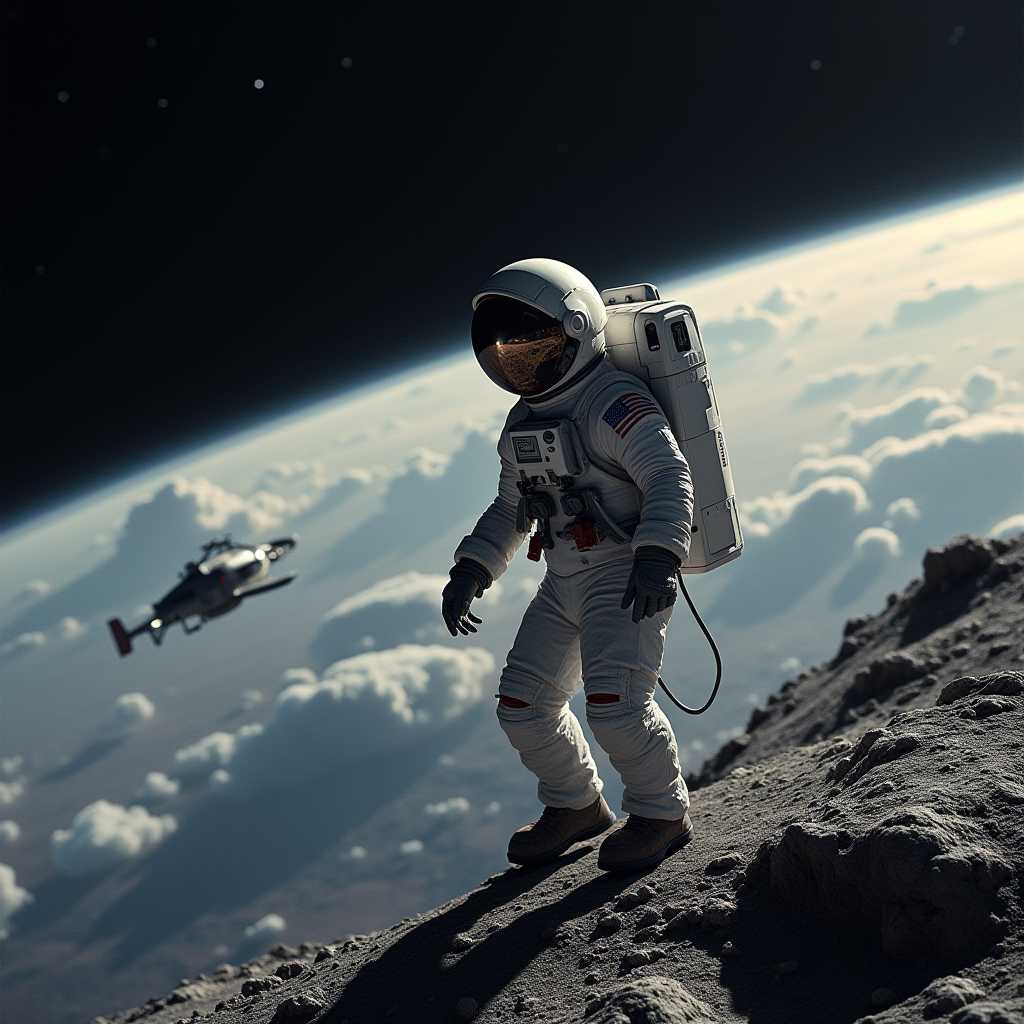} &
        \includegraphics[width=\imgwidth, height=\imgwidth]{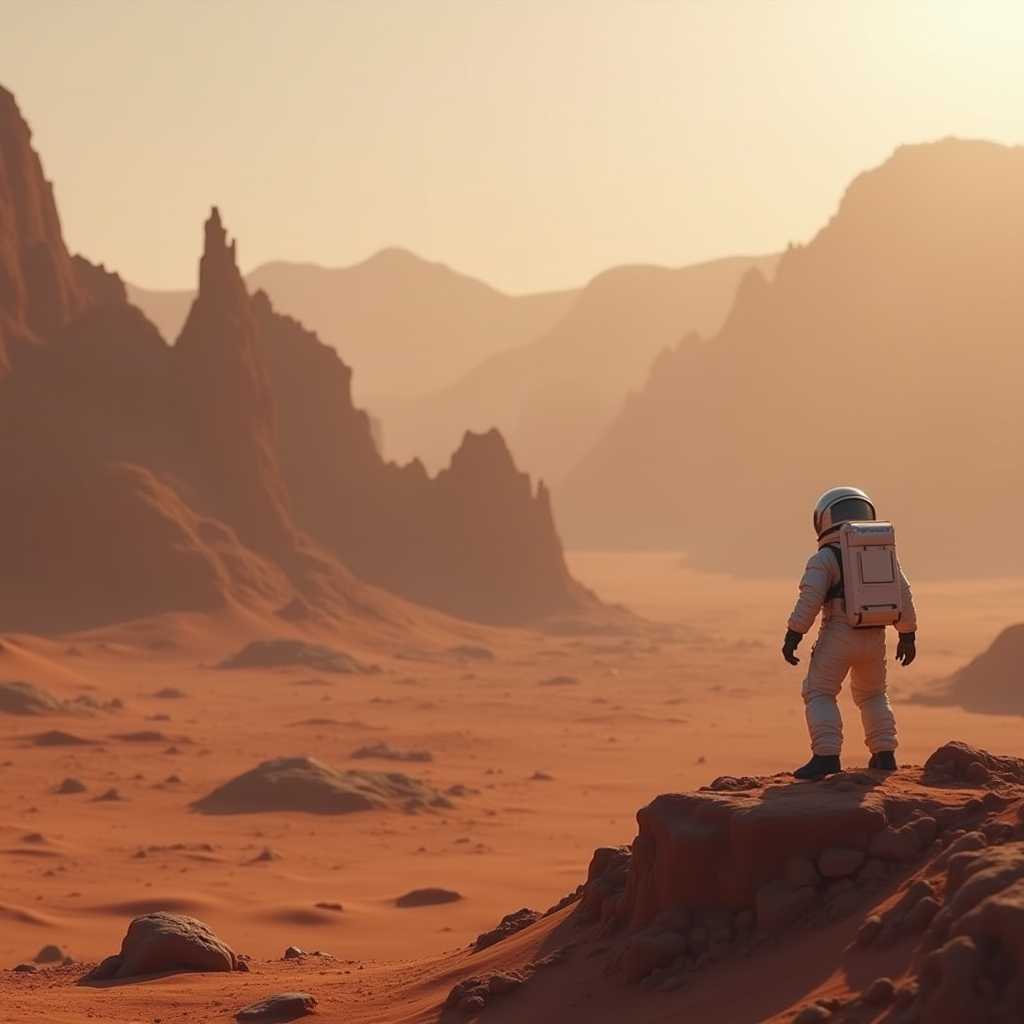} &
        \includegraphics[width=\imgwidth, height=\imgwidth]{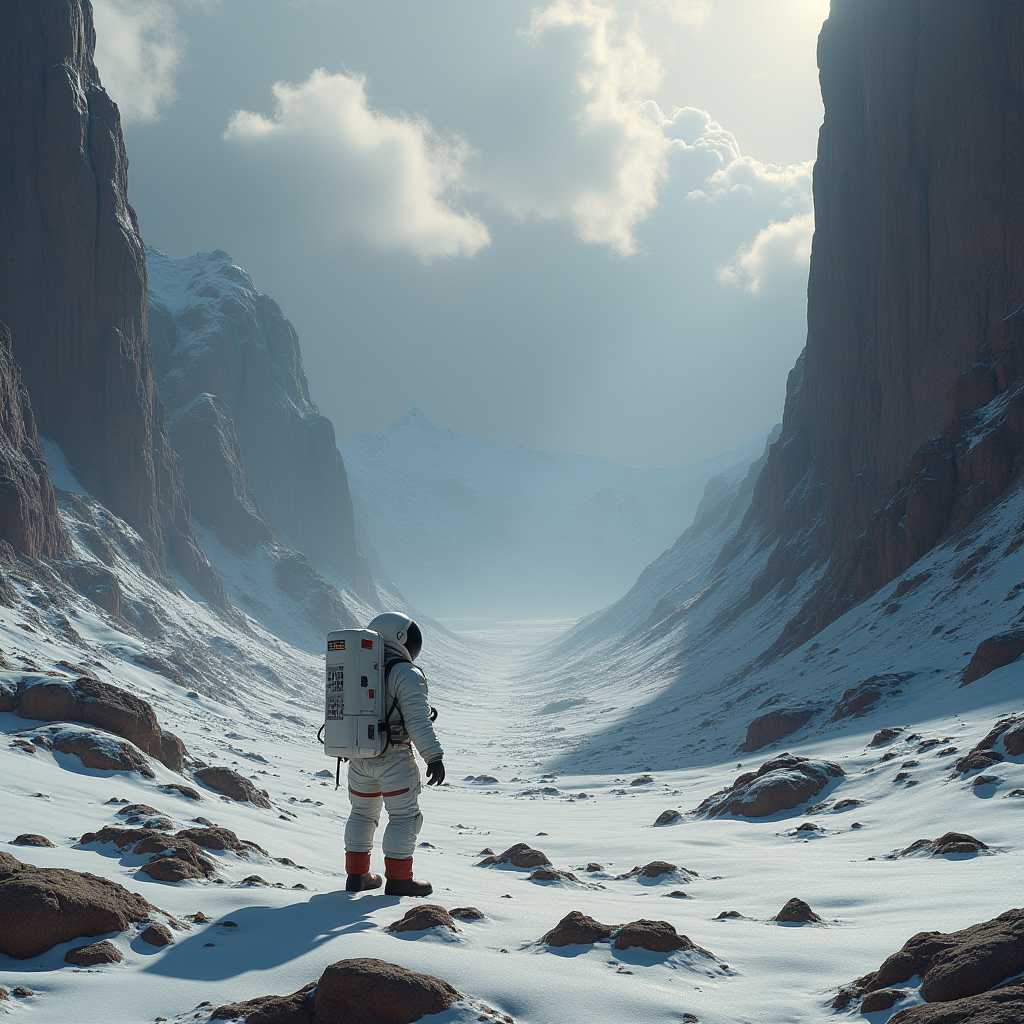} \\
        \multicolumn{9}{c}{\vspace{2pt}\small ``An astronaut exploring the terrain of an alien planet'' \vspace{8pt}} \\

        \vertlabel{Flux} & 
        \includegraphics[width=\imgwidth, height=\imgwidth]{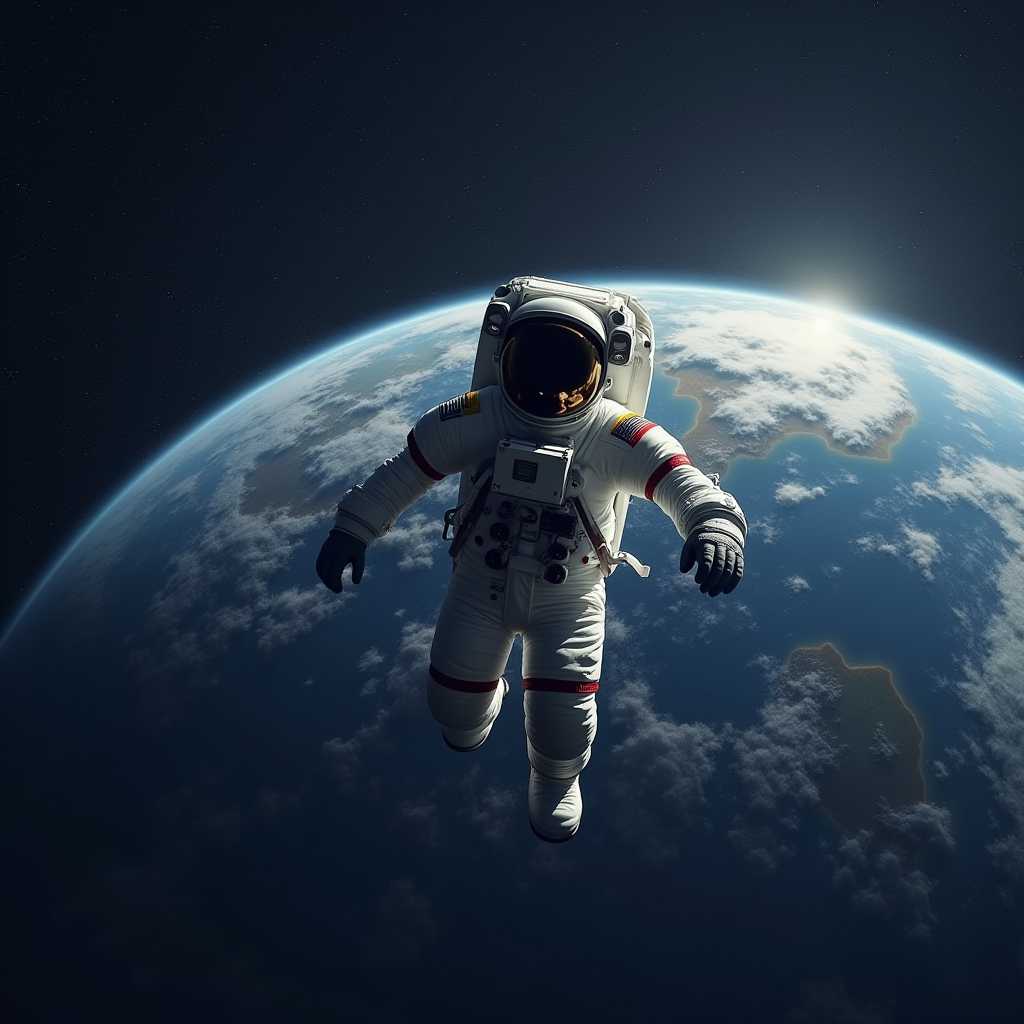} &
        \includegraphics[width=\imgwidth, height=\imgwidth]{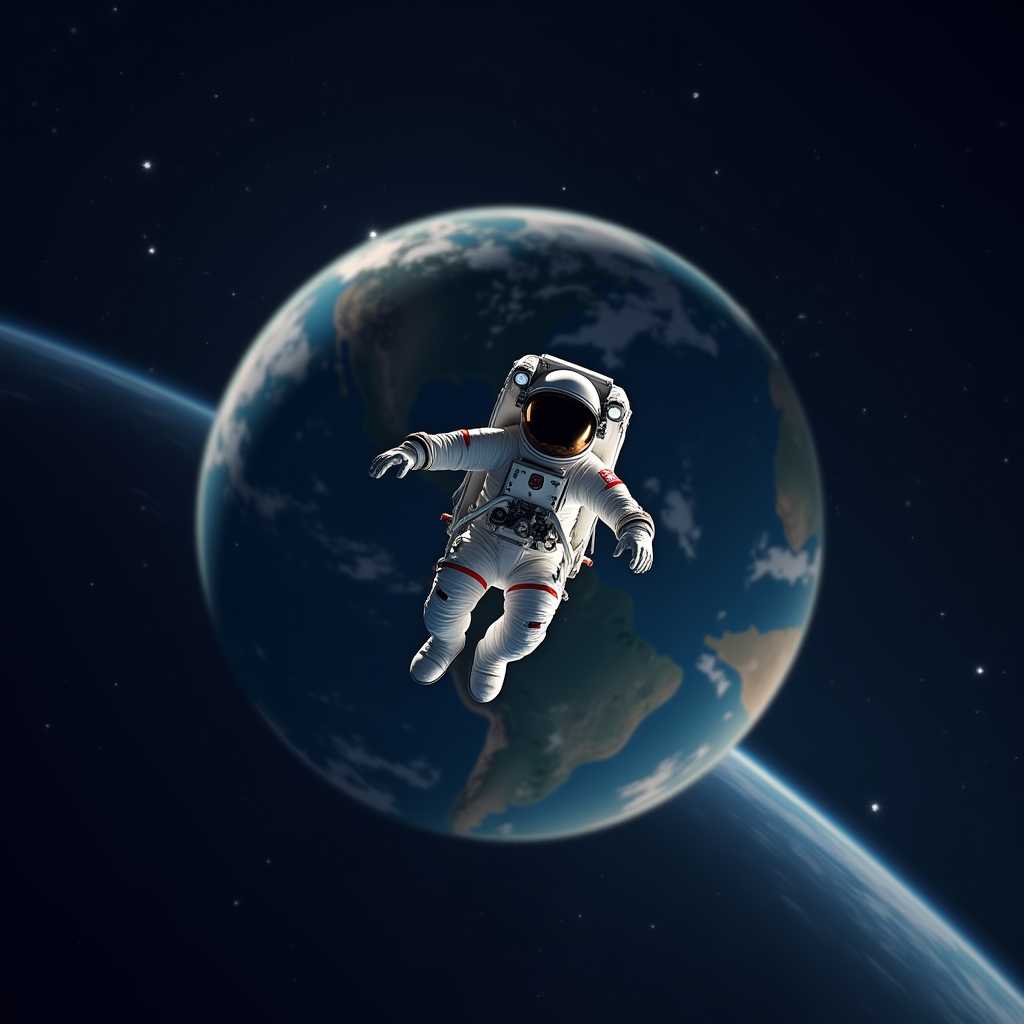} &
        \includegraphics[width=\imgwidth, height=\imgwidth]{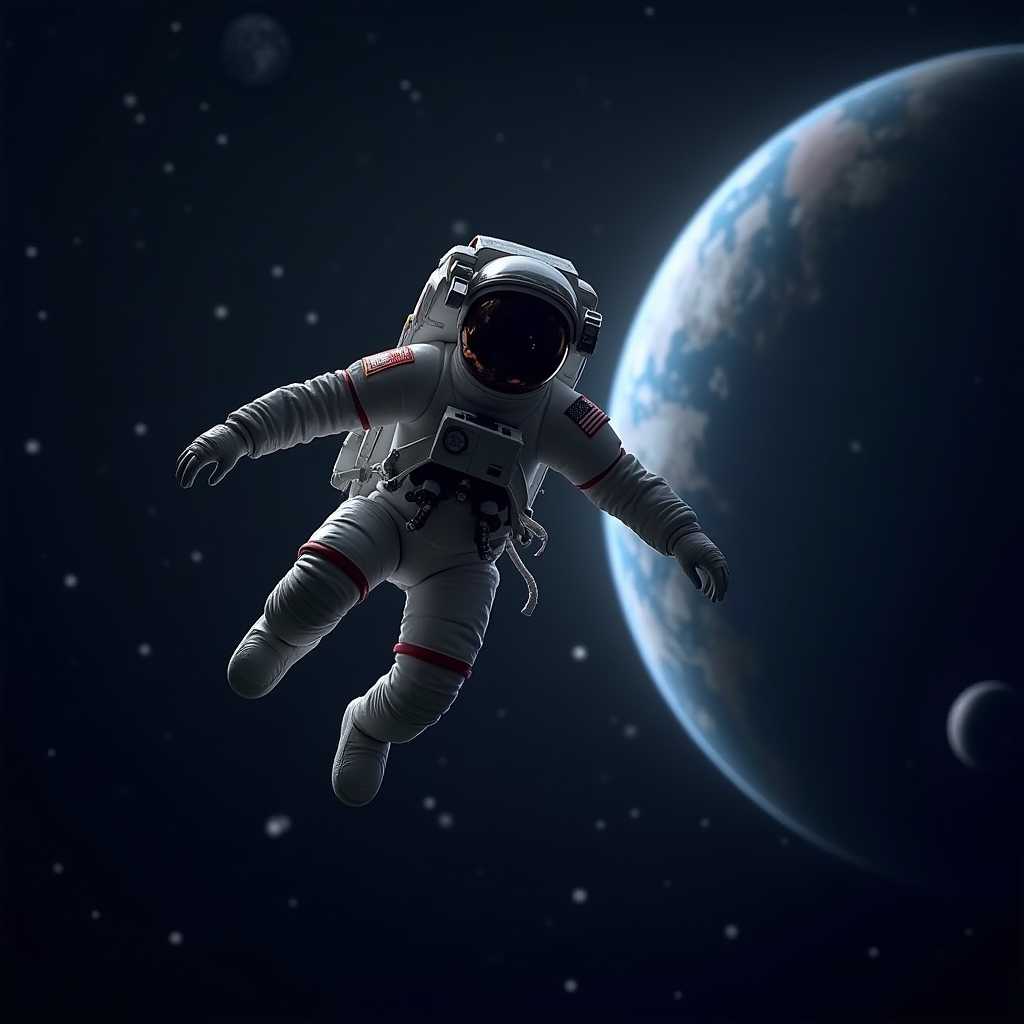} &
        \includegraphics[width=\imgwidth, height=\imgwidth]{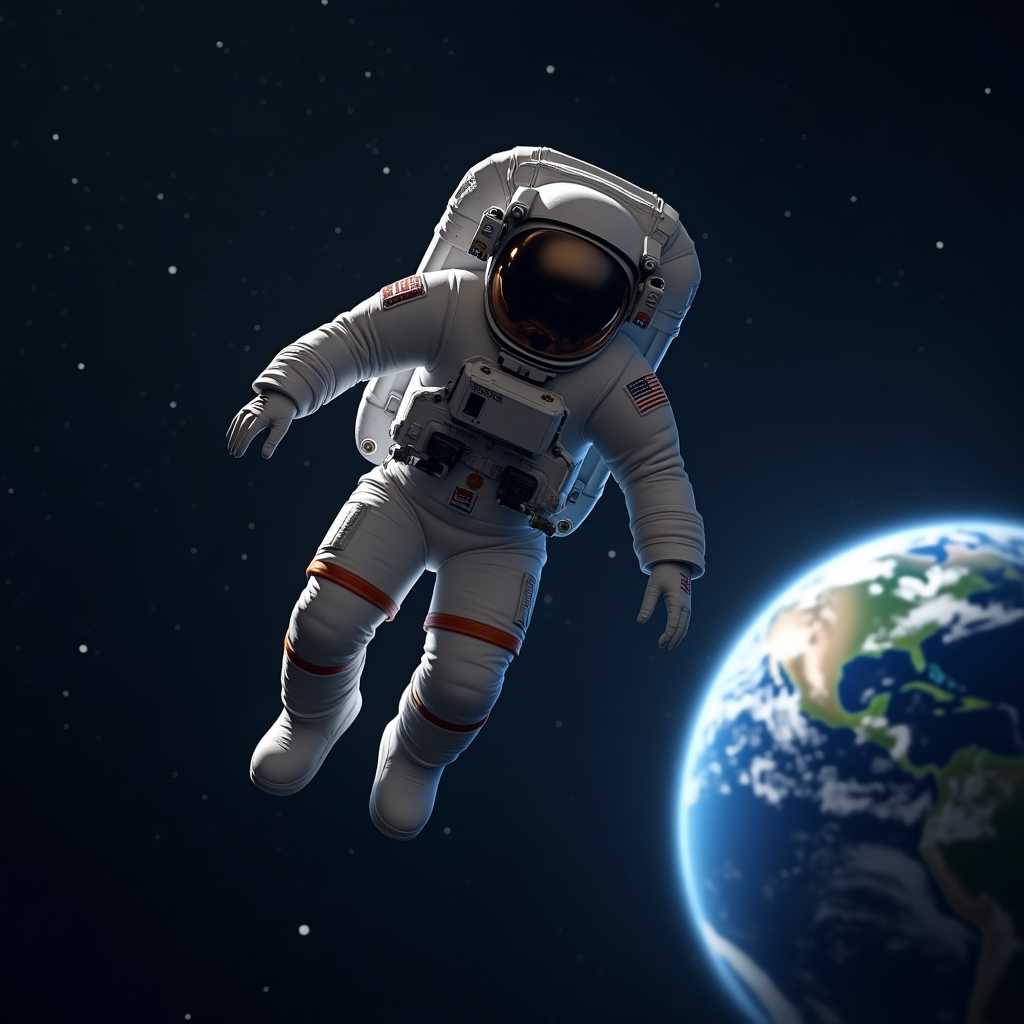} &
        \includegraphics[width=\imgwidth, height=\imgwidth]{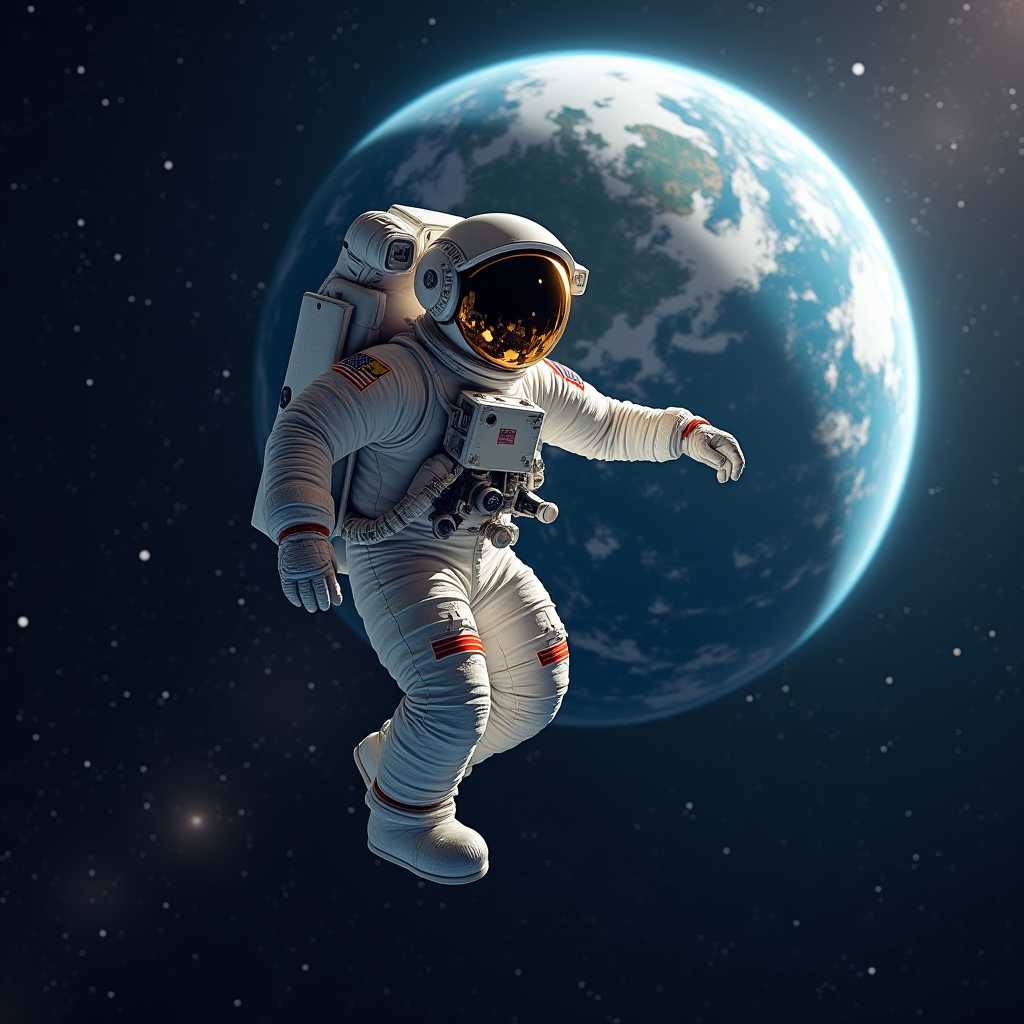} &
        \includegraphics[width=\imgwidth, height=\imgwidth]{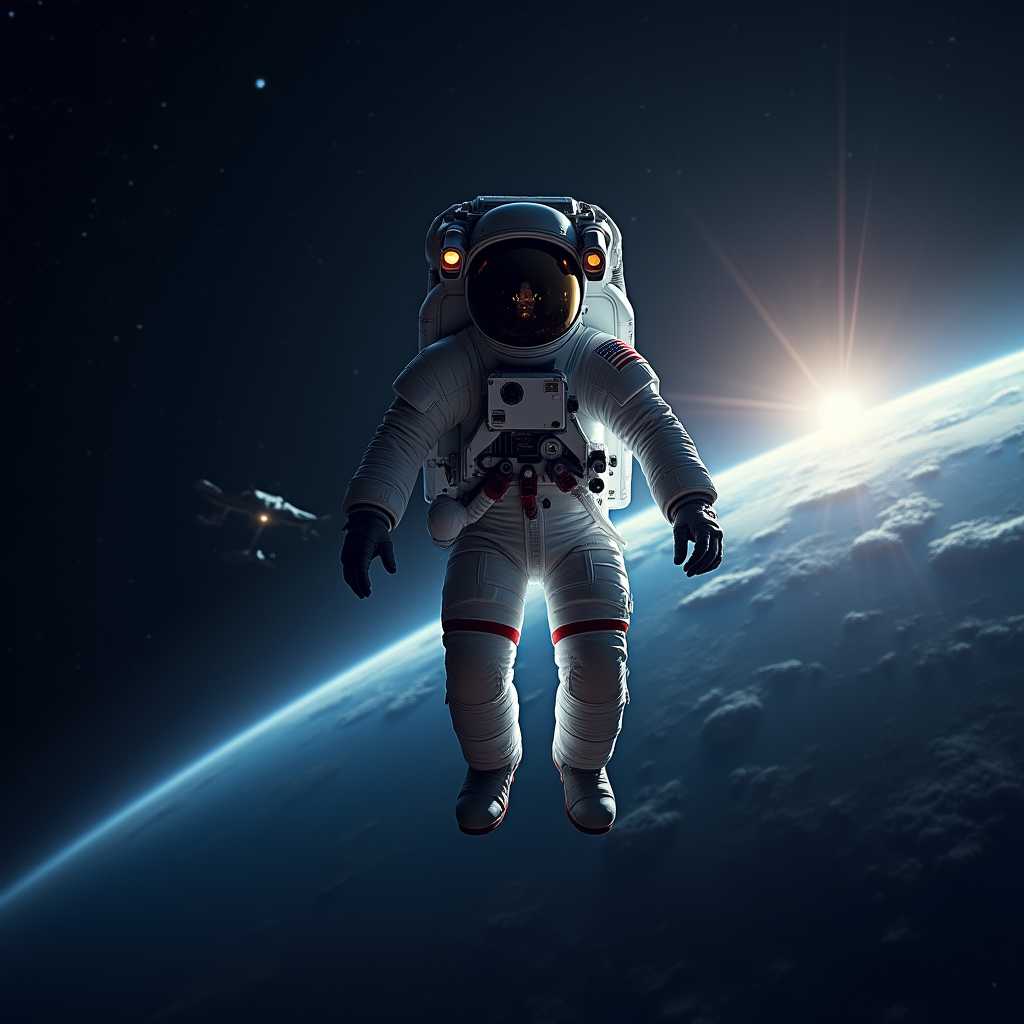} &
        \includegraphics[width=\imgwidth, height=\imgwidth]{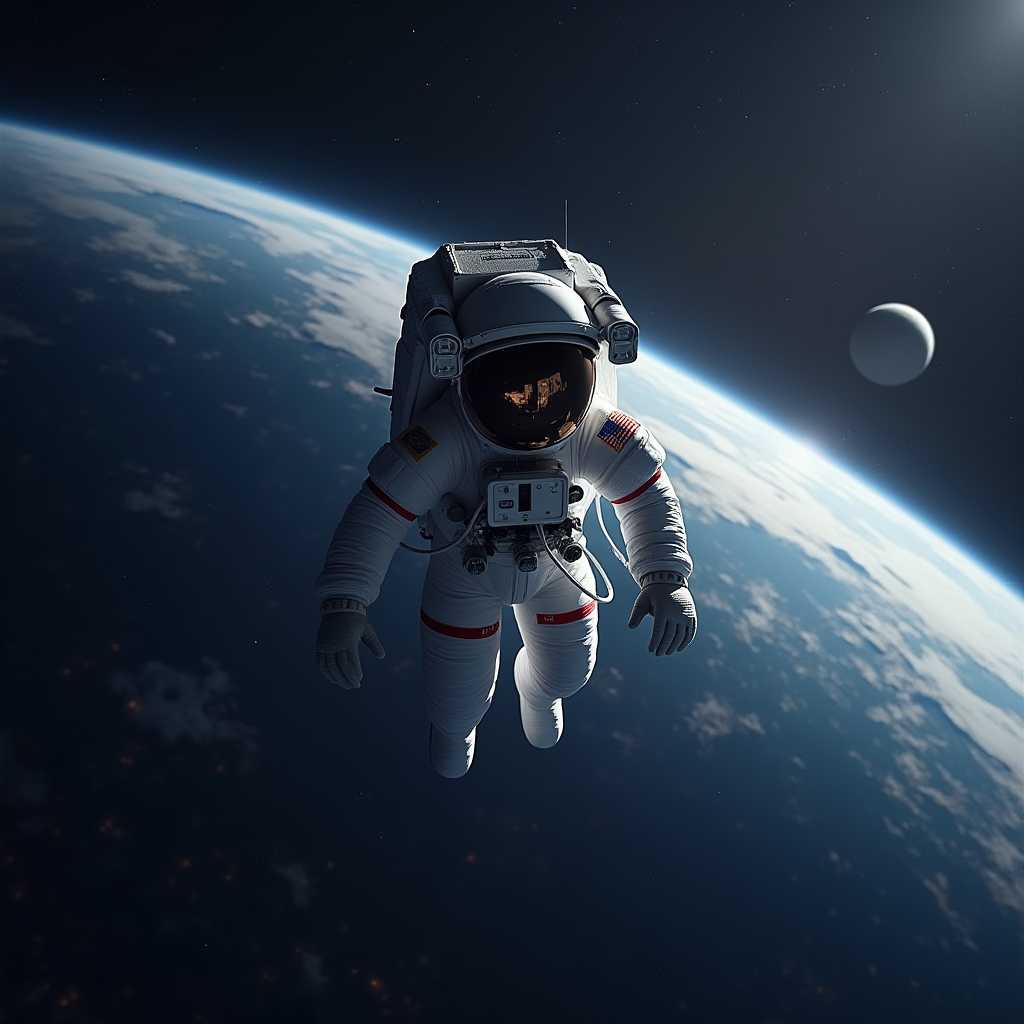} &
        \includegraphics[width=\imgwidth, height=\imgwidth]{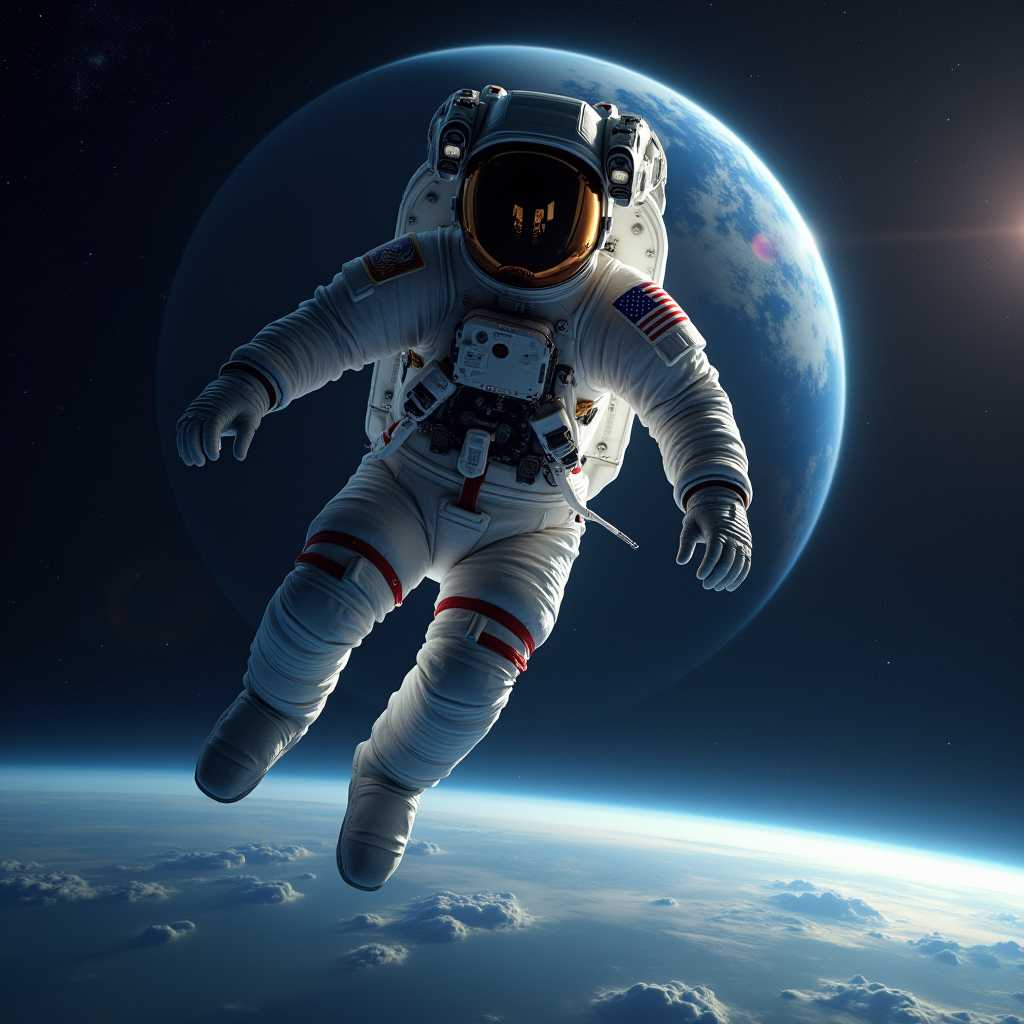} \\[-1pt]

        \vertlabel{Ours} & 
        \includegraphics[width=\imgwidth, height=\imgwidth]{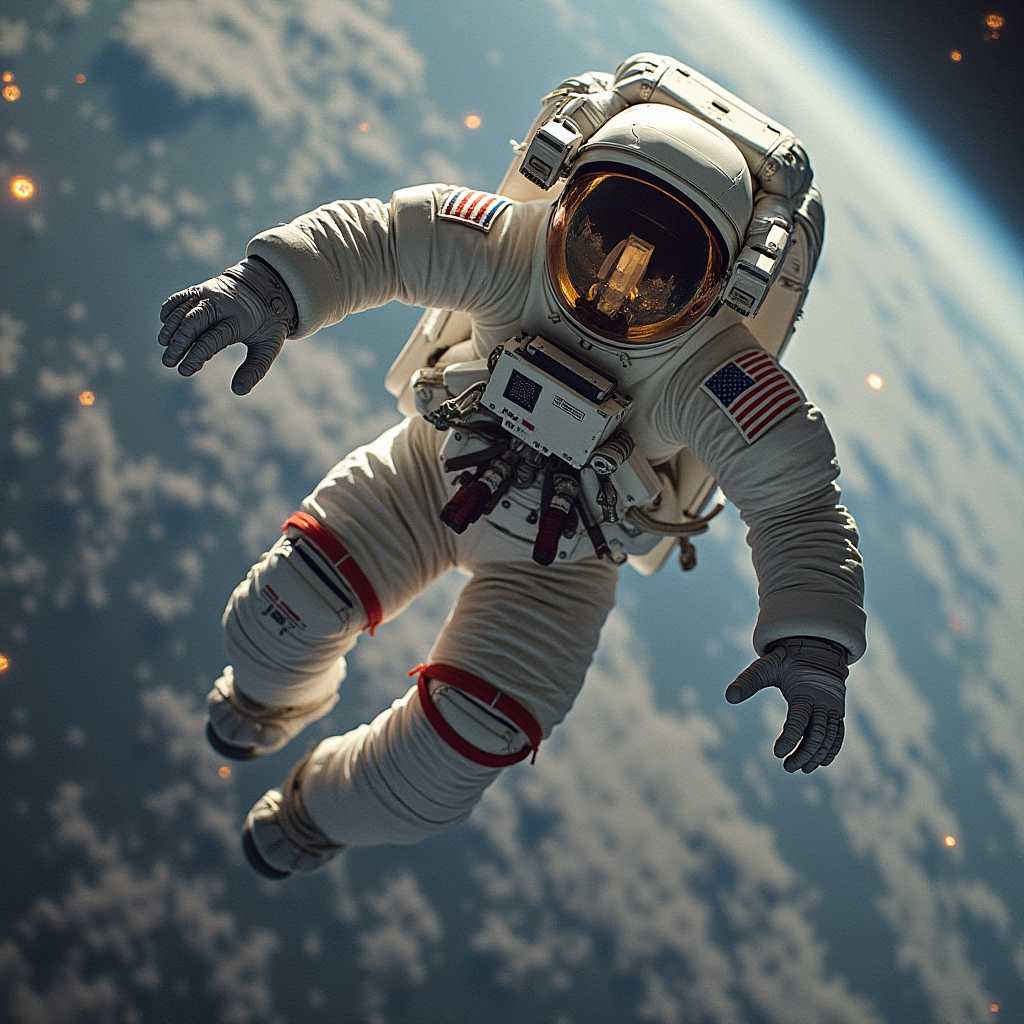} &
        \includegraphics[width=\imgwidth, height=\imgwidth]{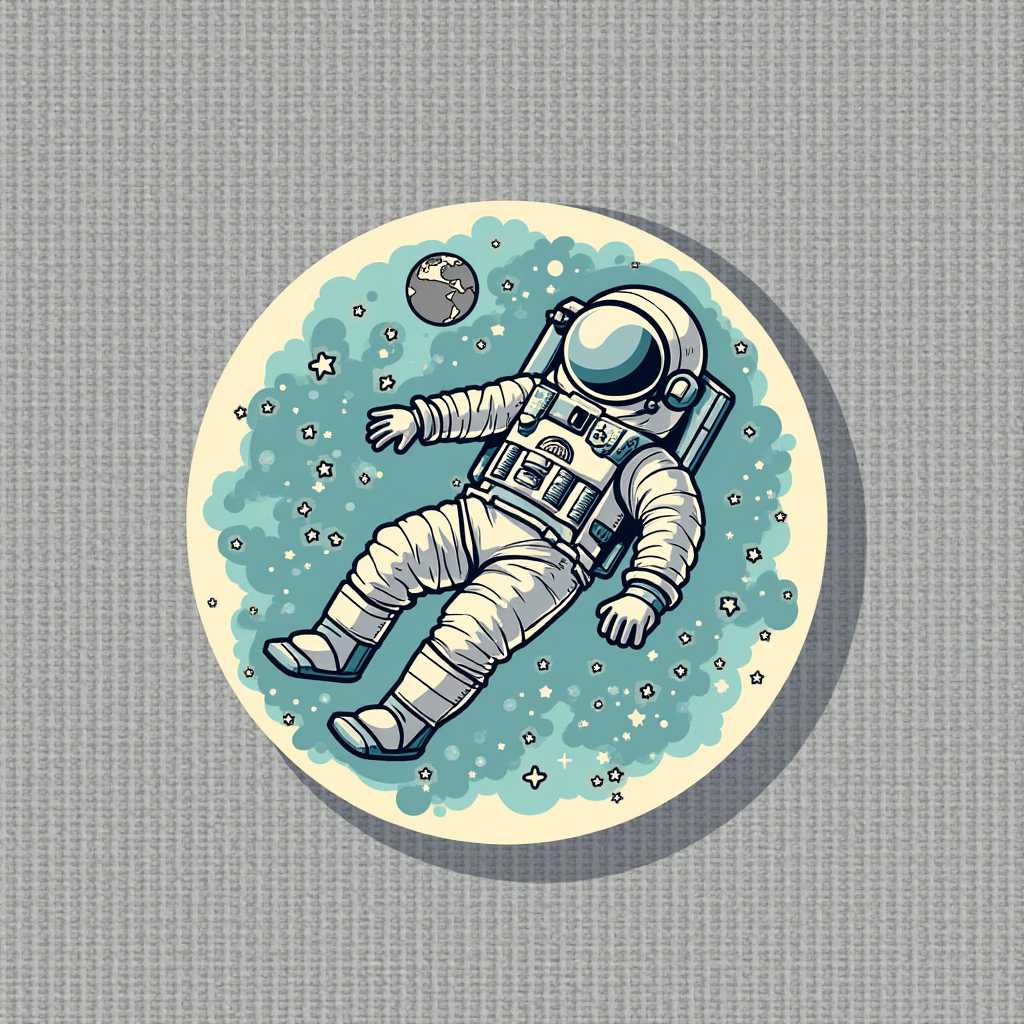} &
        \includegraphics[width=\imgwidth, height=\imgwidth]{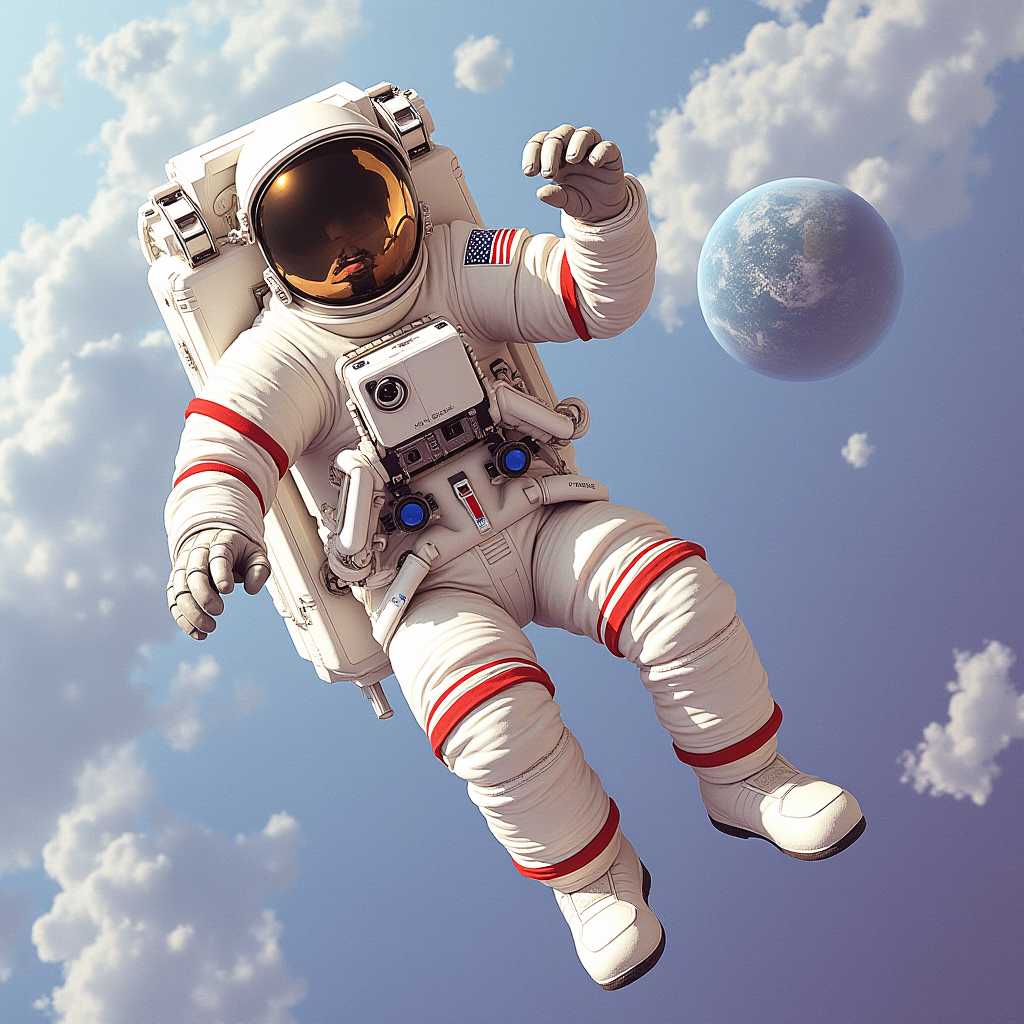} &
        \includegraphics[width=\imgwidth, height=\imgwidth]{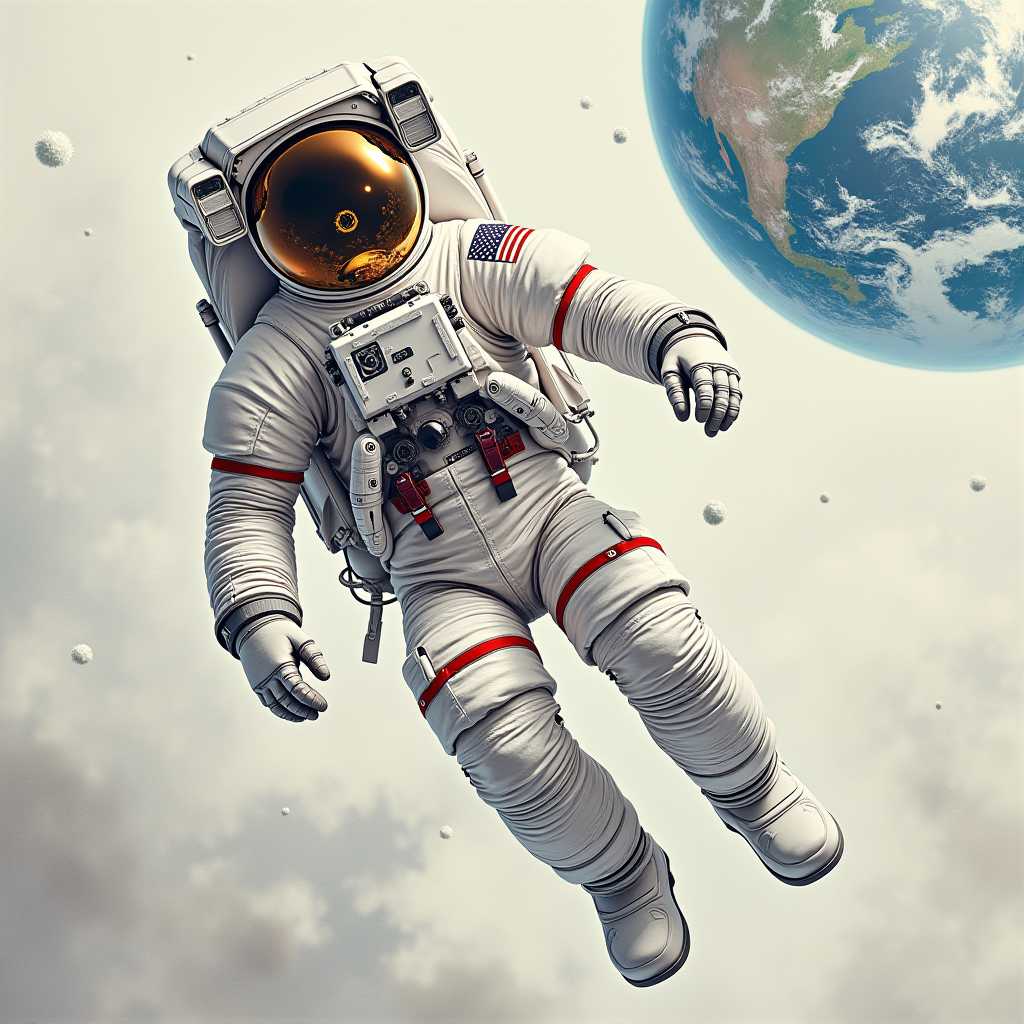} &
        \includegraphics[width=\imgwidth, height=\imgwidth]{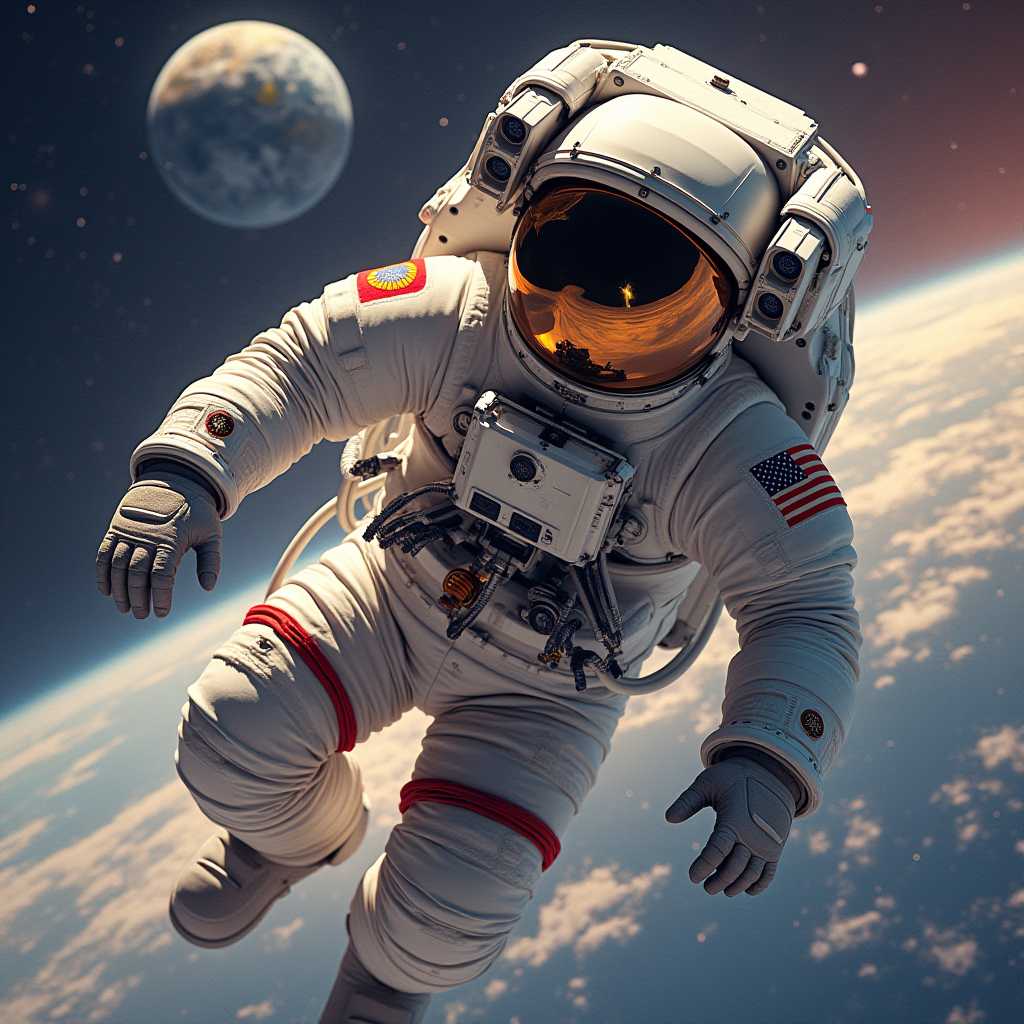} &
        \includegraphics[width=\imgwidth, height=\imgwidth]{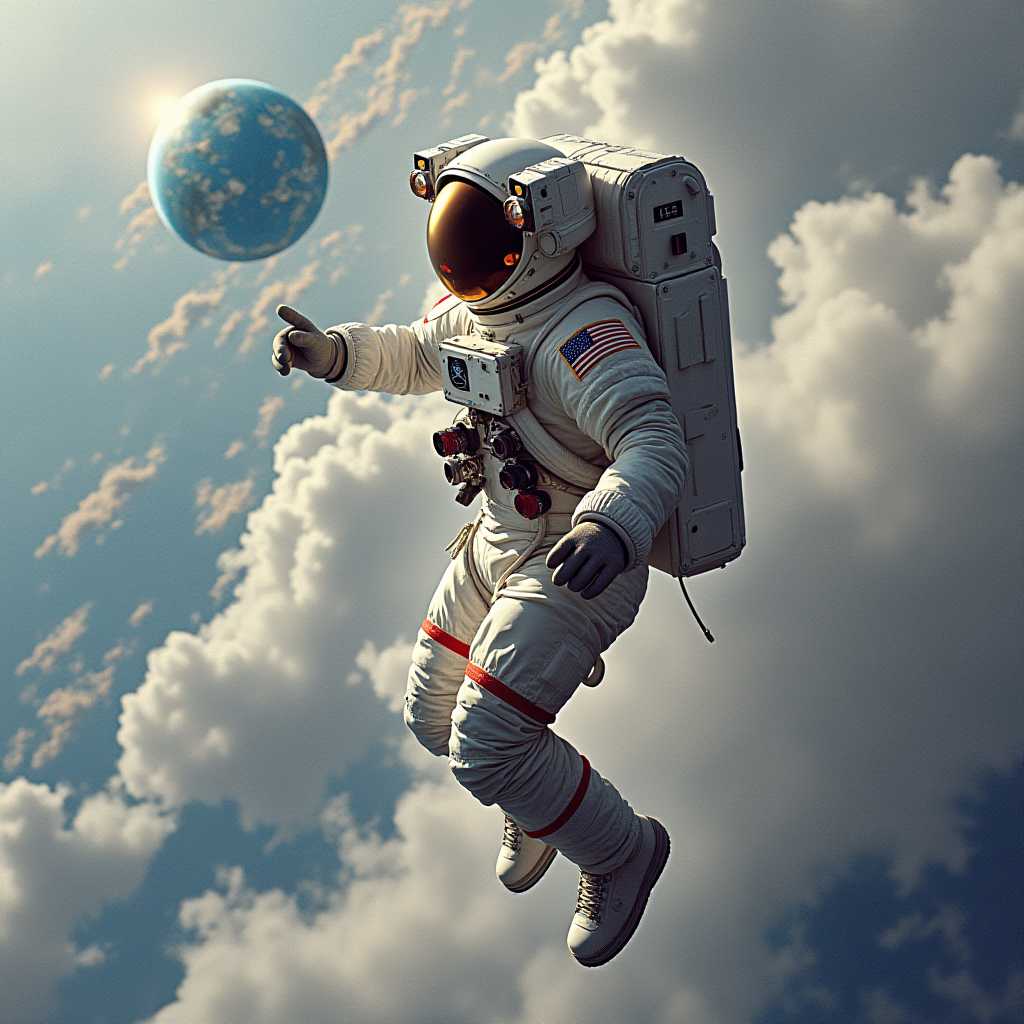} &
        \includegraphics[width=\imgwidth, height=\imgwidth]{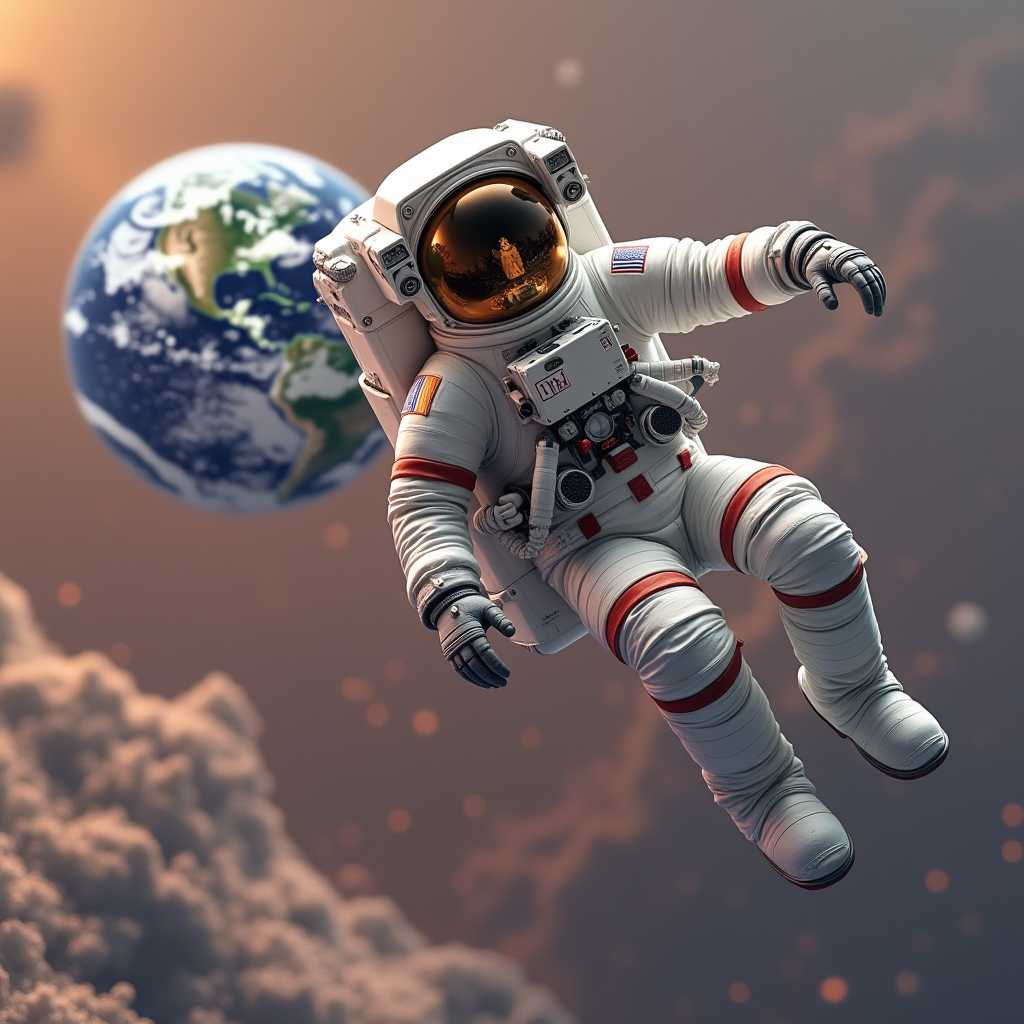} &
        \includegraphics[width=\imgwidth, height=\imgwidth]{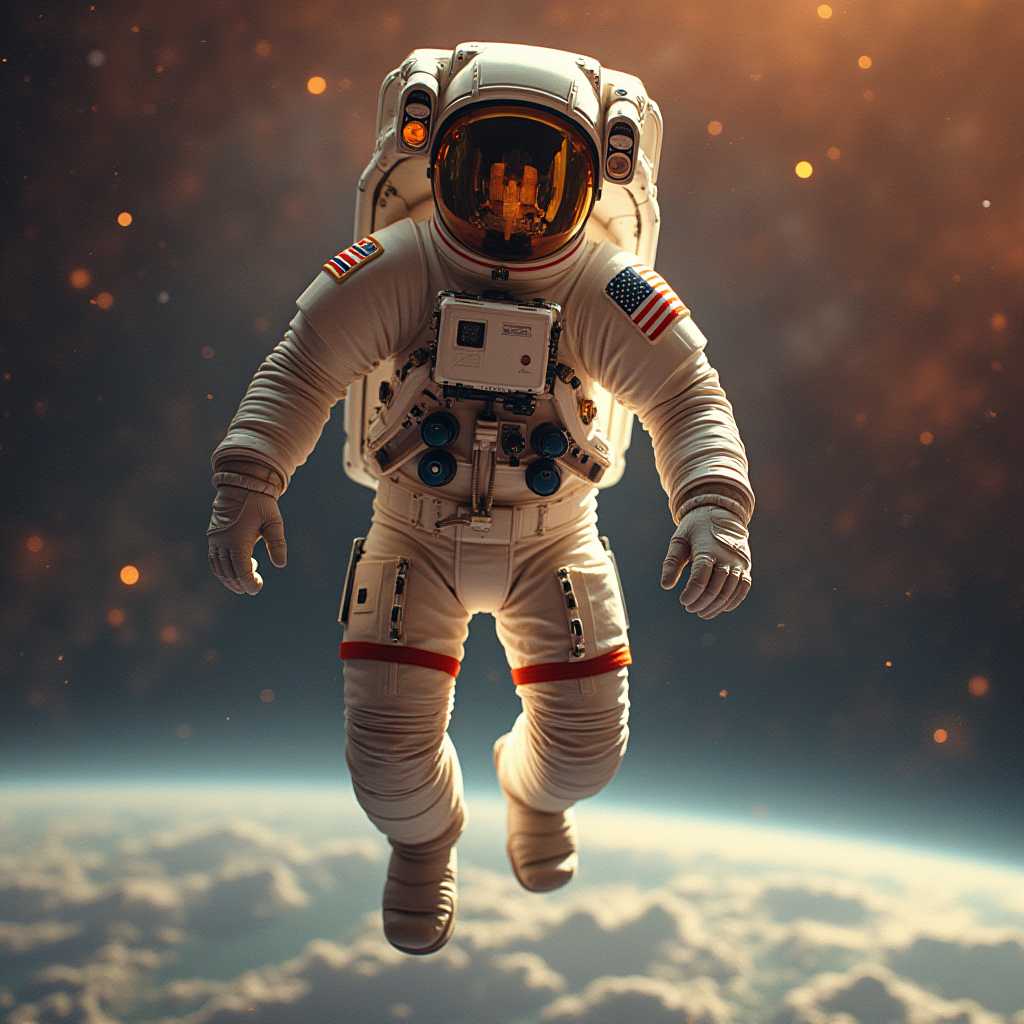} \\
        \multicolumn{9}{c}{\vspace{2pt}\small ``An astronaut floating in space with Earth in the background'' \vspace{8pt}} \\

        \vertlabel{Flux} & 
        \includegraphics[width=\imgwidth, height=\imgwidth]{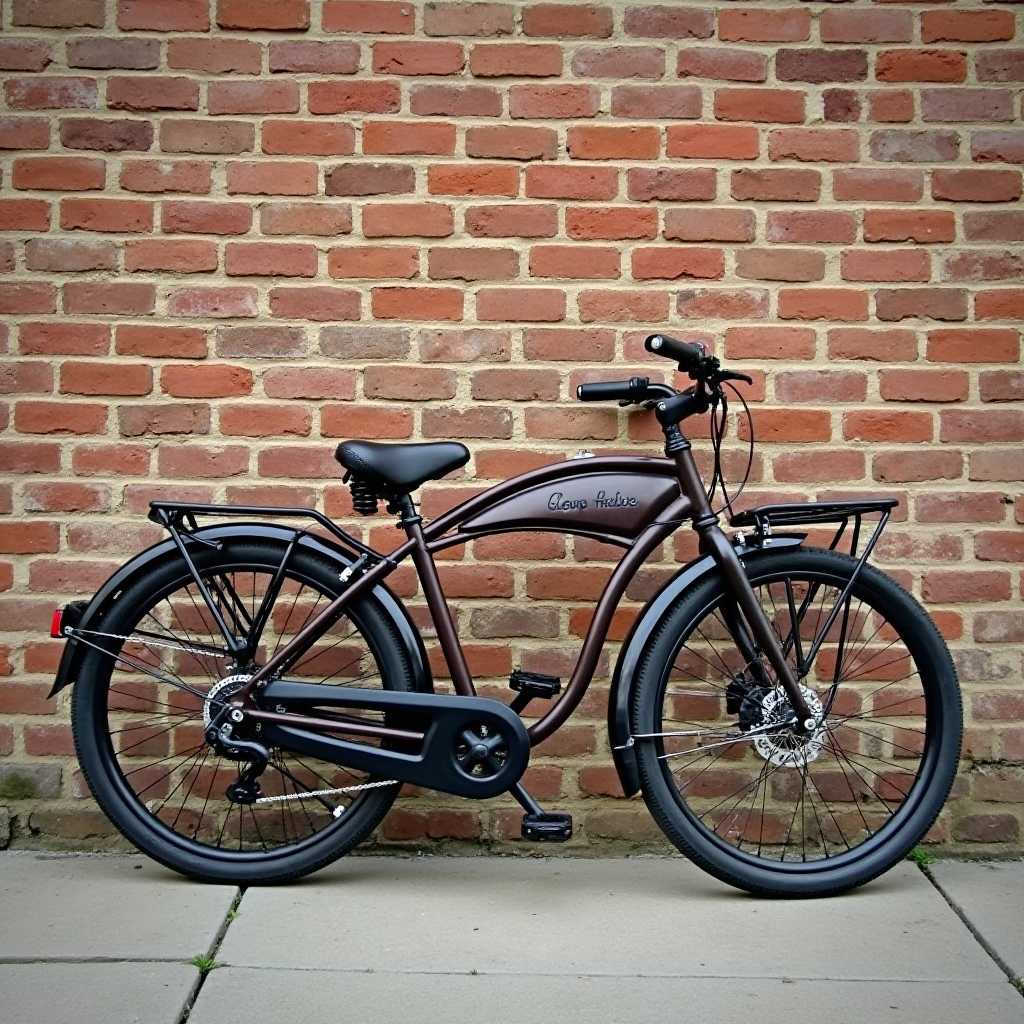} &
        \includegraphics[width=\imgwidth, height=\imgwidth]{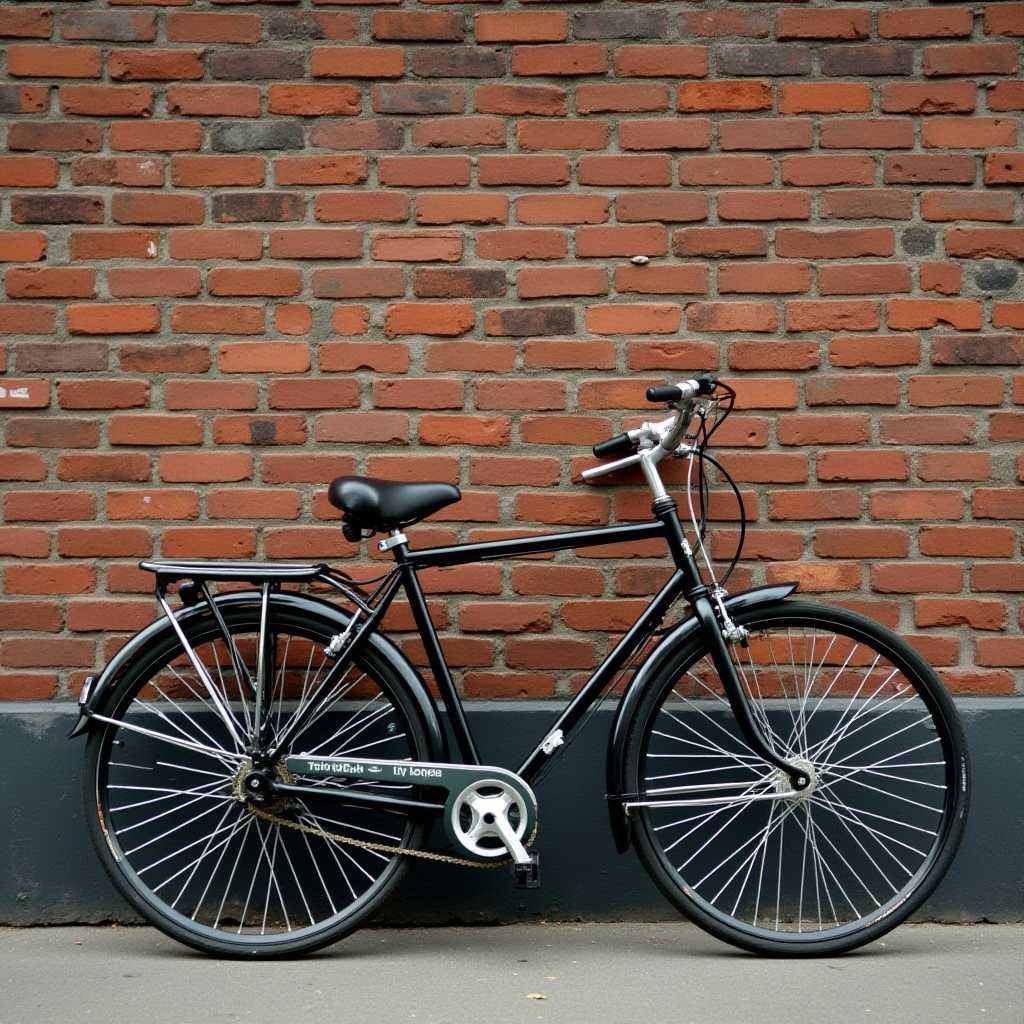} &
        \includegraphics[width=\imgwidth, height=\imgwidth]{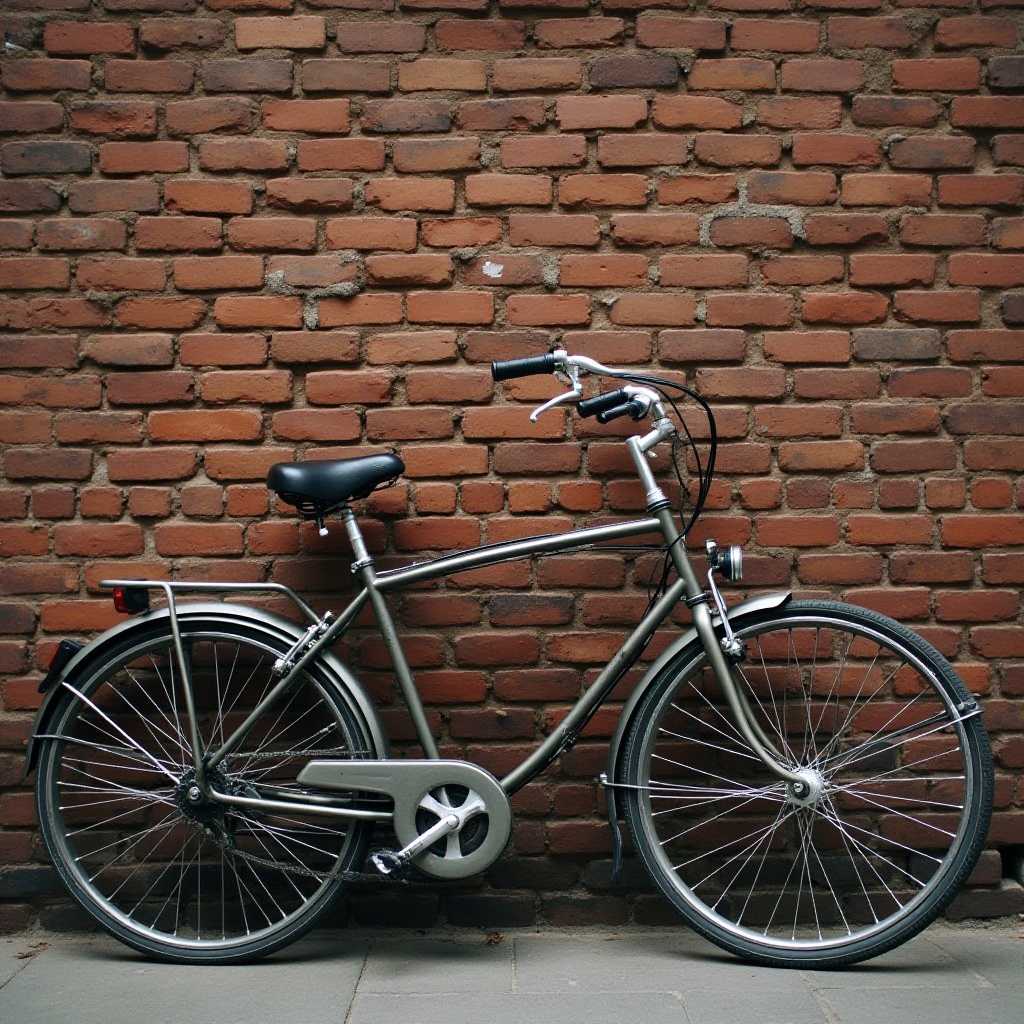} &
        \includegraphics[width=\imgwidth, height=\imgwidth]{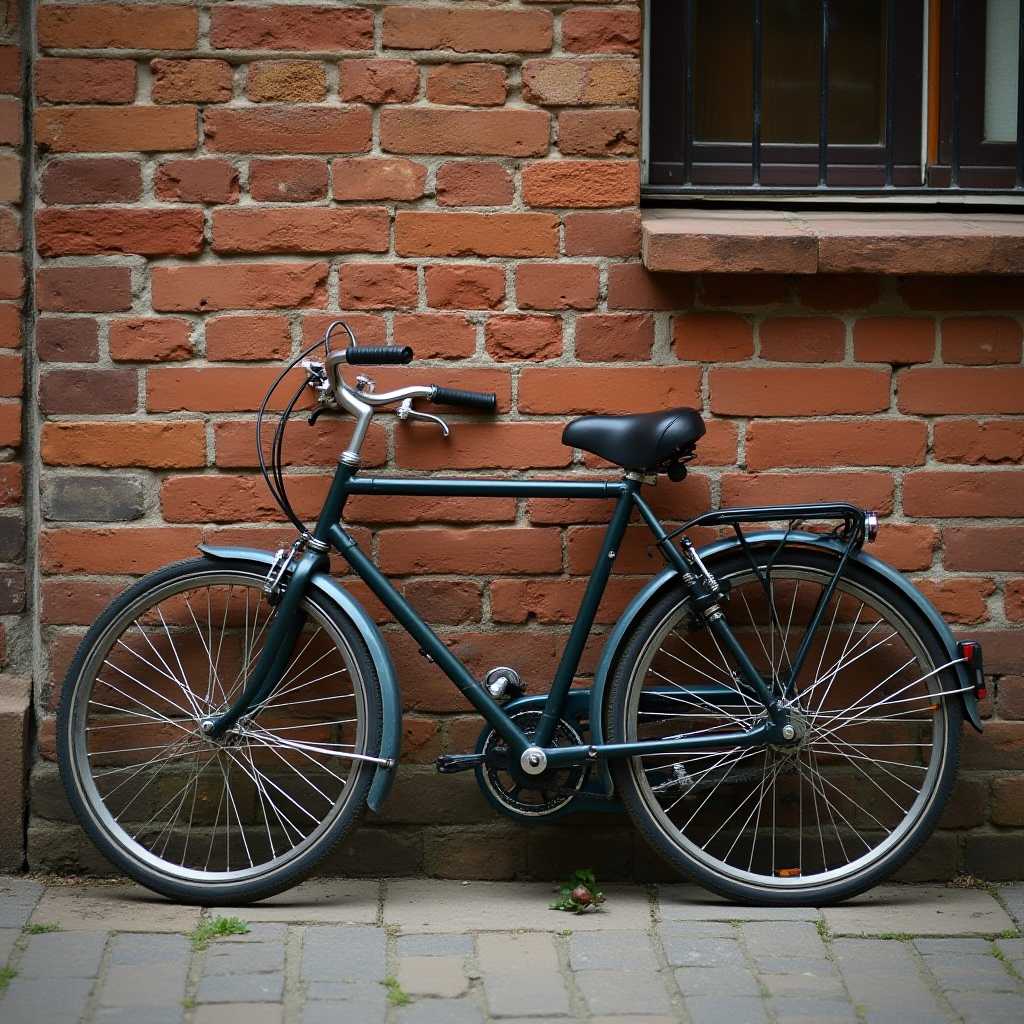} &
        \includegraphics[width=\imgwidth, height=\imgwidth]{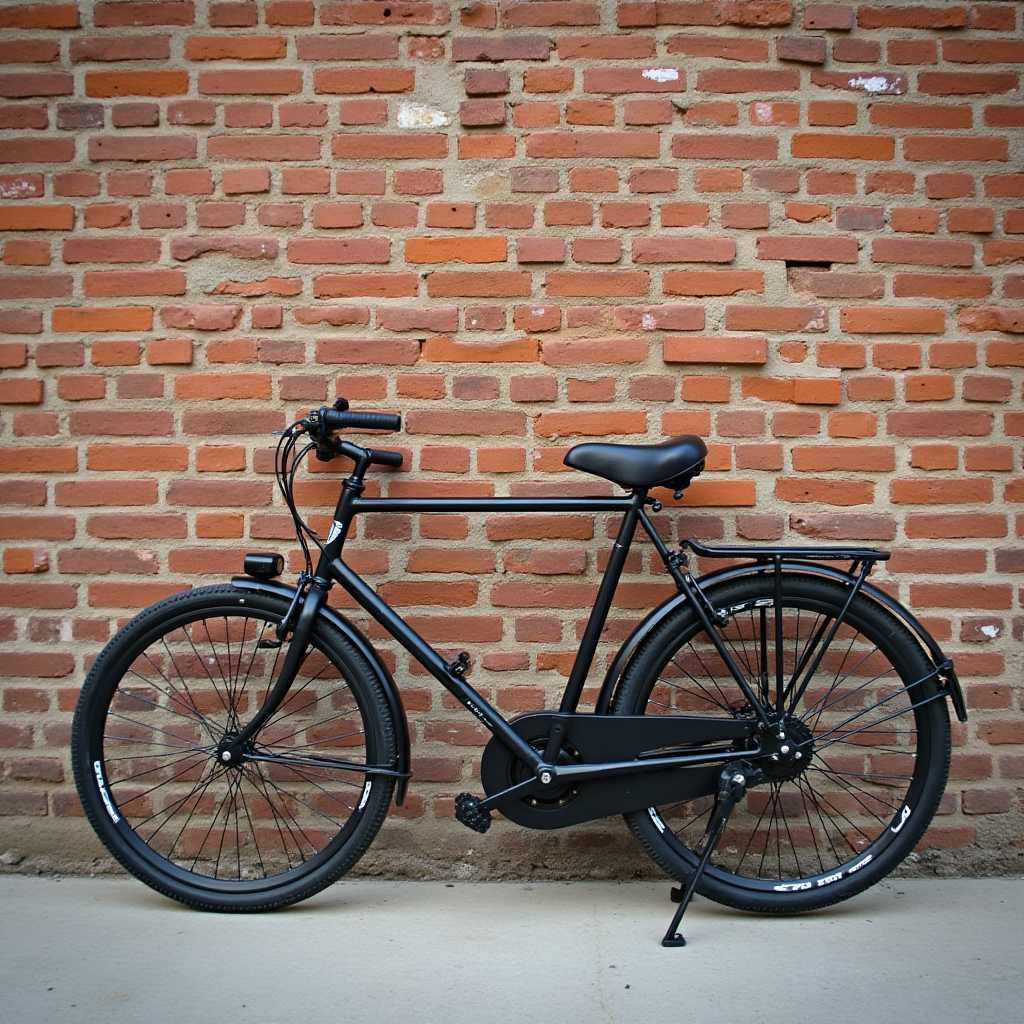} &
        \includegraphics[width=\imgwidth, height=\imgwidth]{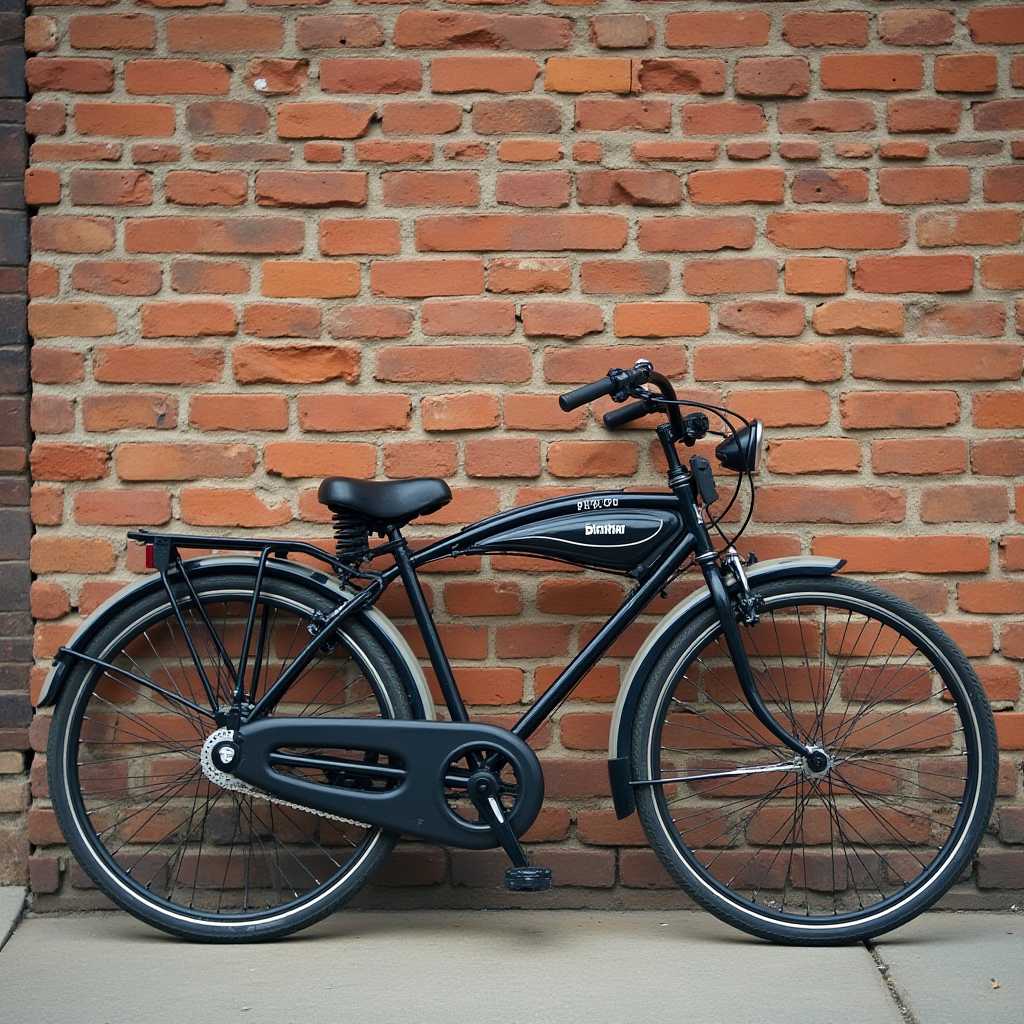} &
        \includegraphics[width=\imgwidth, height=\imgwidth]{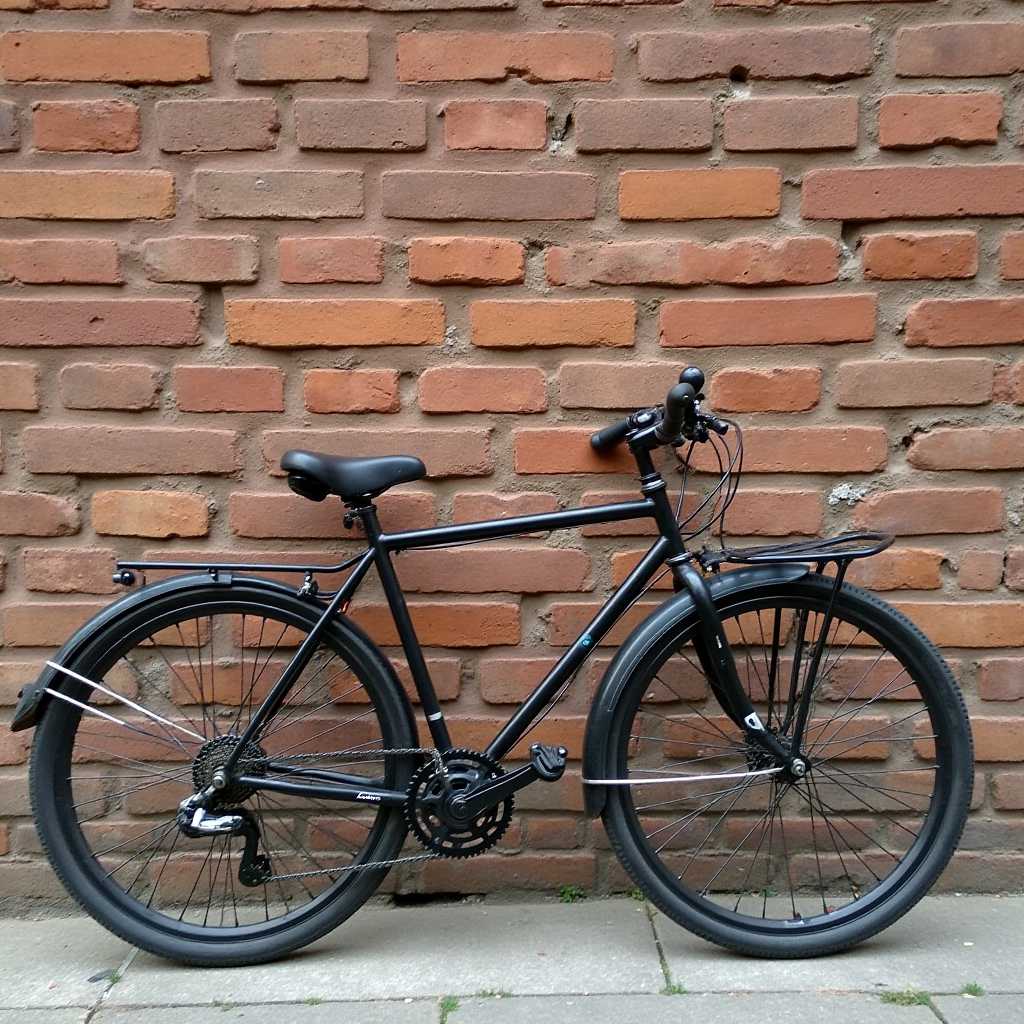} &
        \includegraphics[width=\imgwidth, height=\imgwidth]{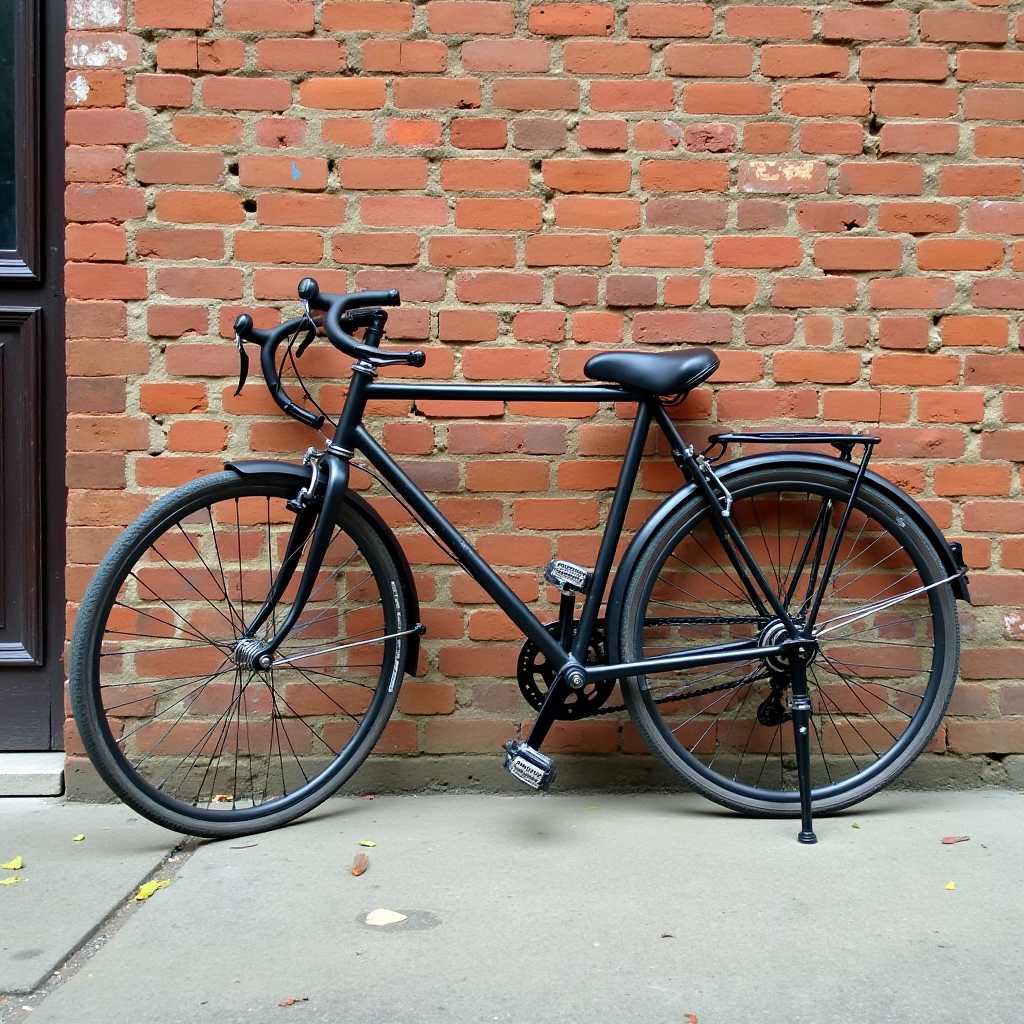} \\[-1pt]

        \vertlabel{Ours} & 
        \includegraphics[width=\imgwidth, height=\imgwidth]{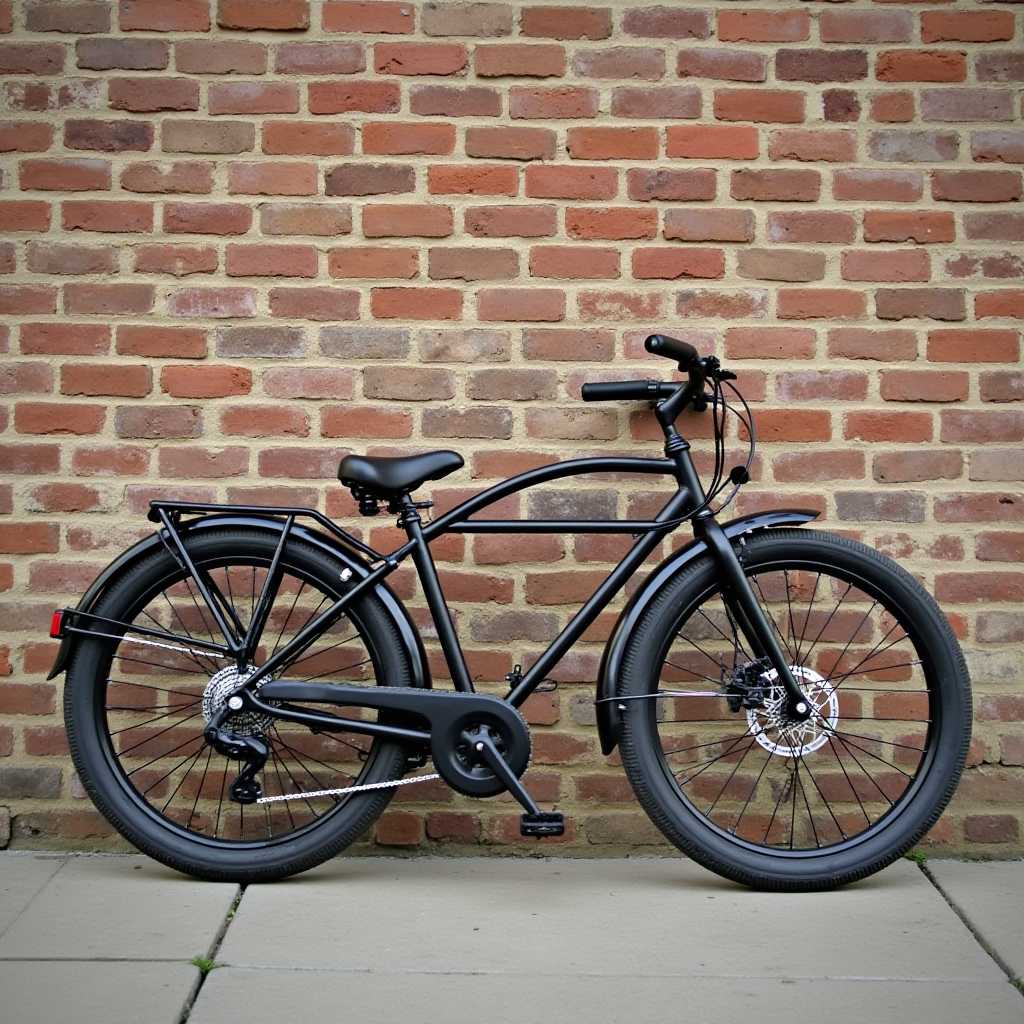} &
        \includegraphics[width=\imgwidth, height=\imgwidth]{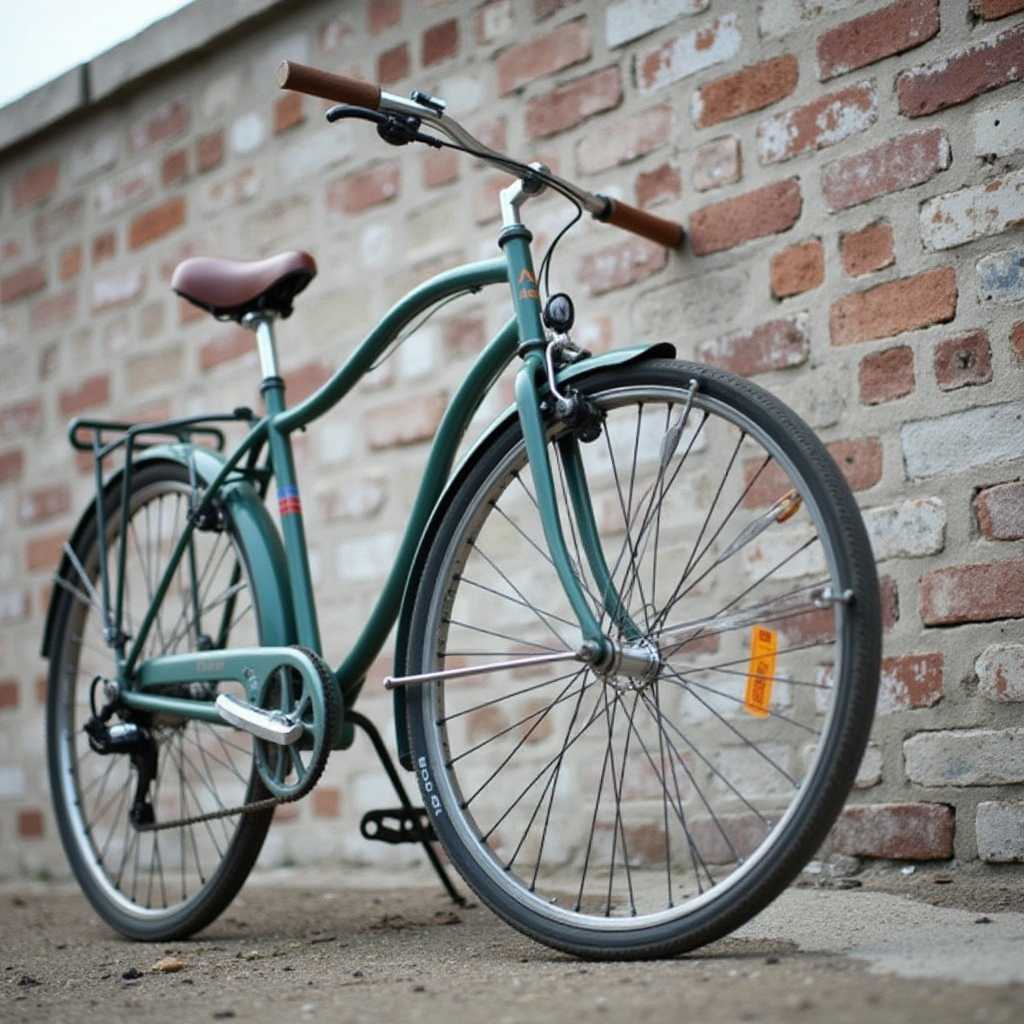} &
        \includegraphics[width=\imgwidth, height=\imgwidth]{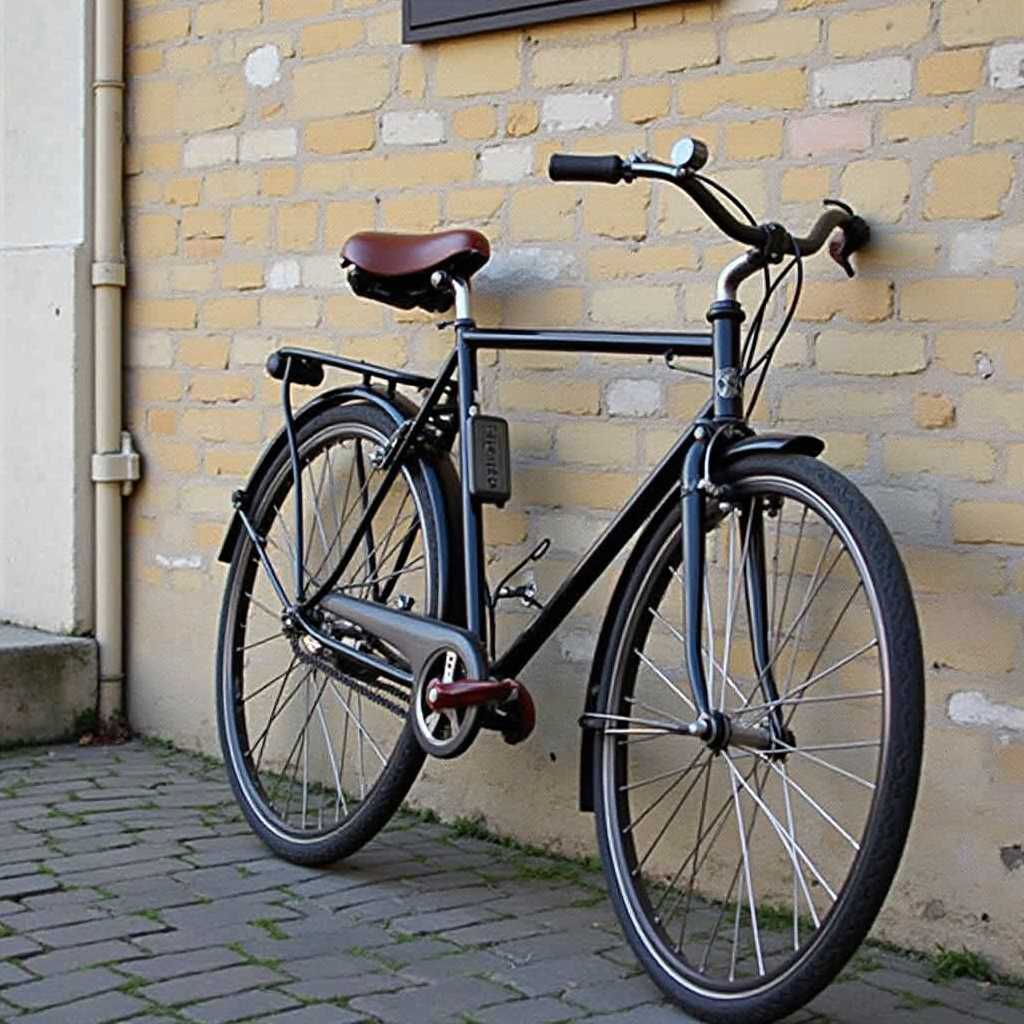} &
        \includegraphics[width=\imgwidth, height=\imgwidth]{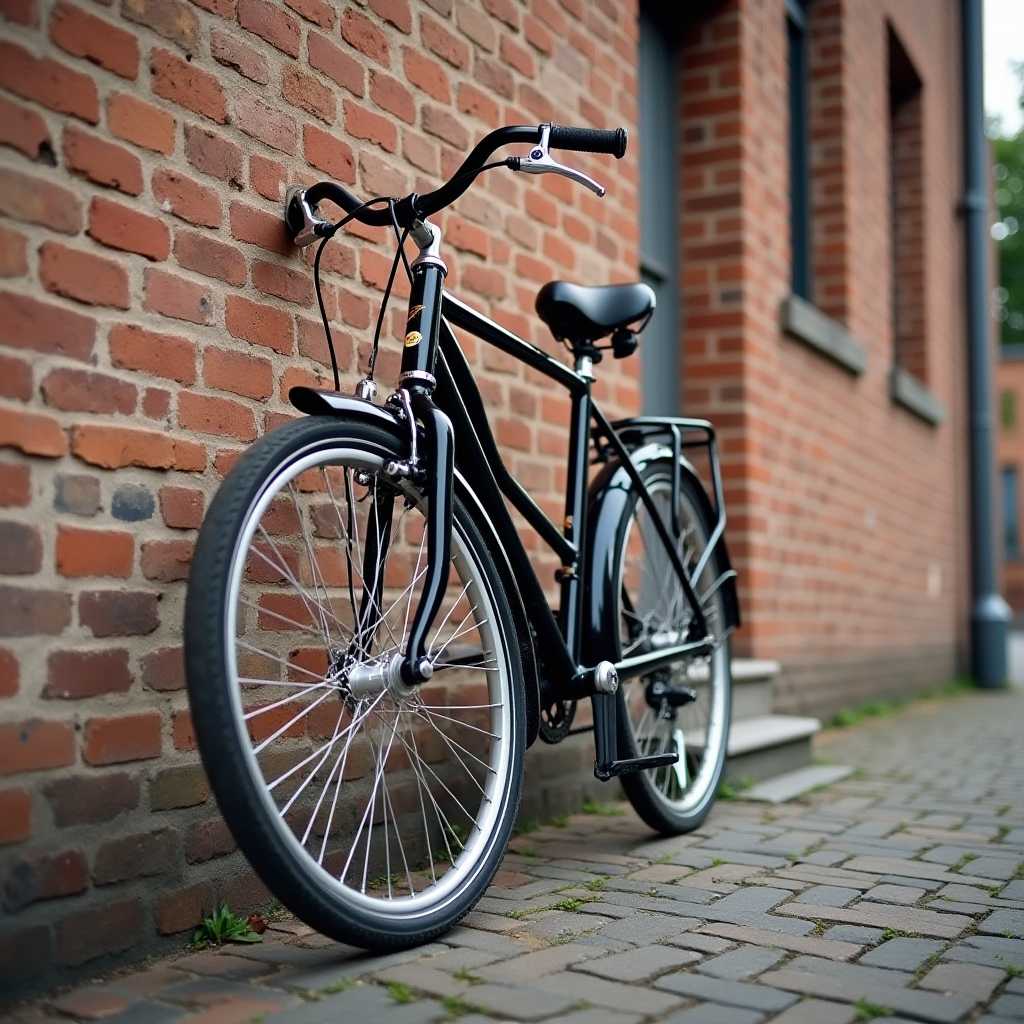} &
        \includegraphics[width=\imgwidth, height=\imgwidth]{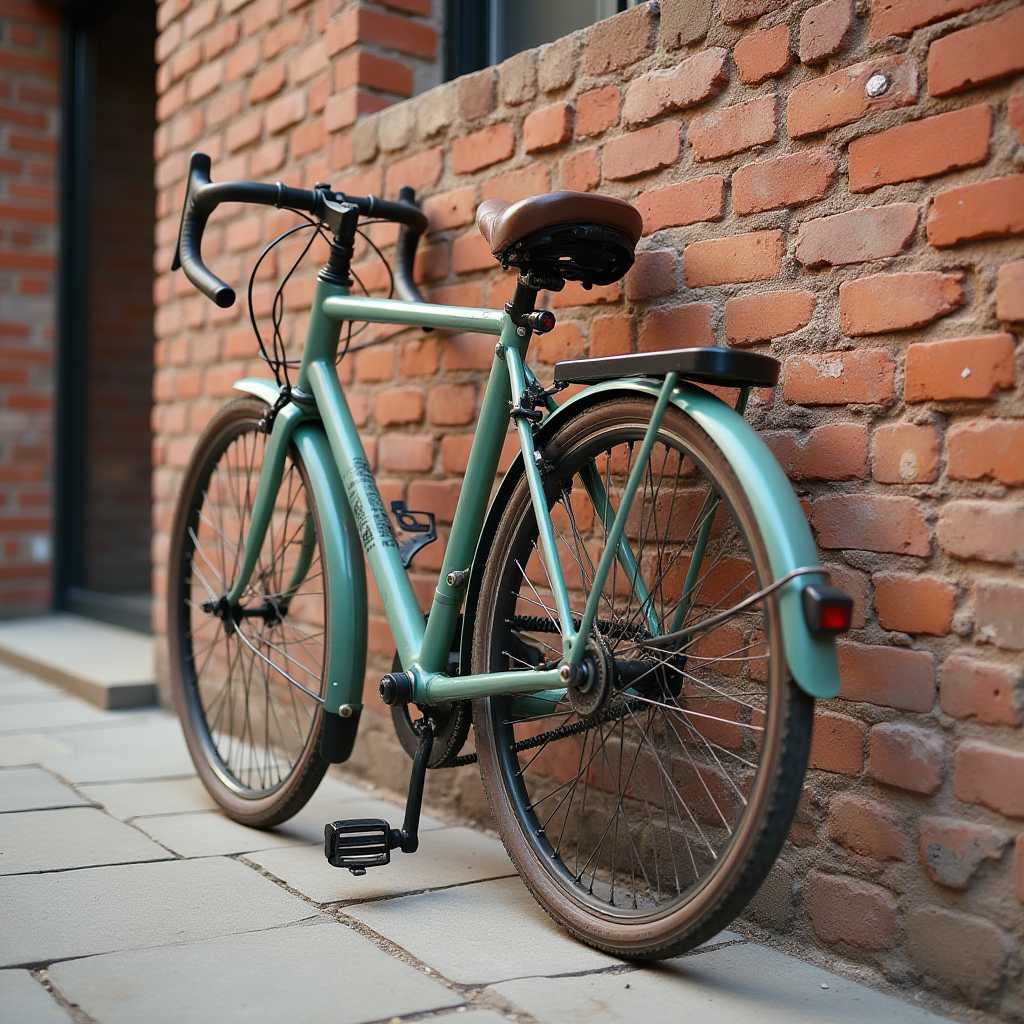} &
        \includegraphics[width=\imgwidth, height=\imgwidth]{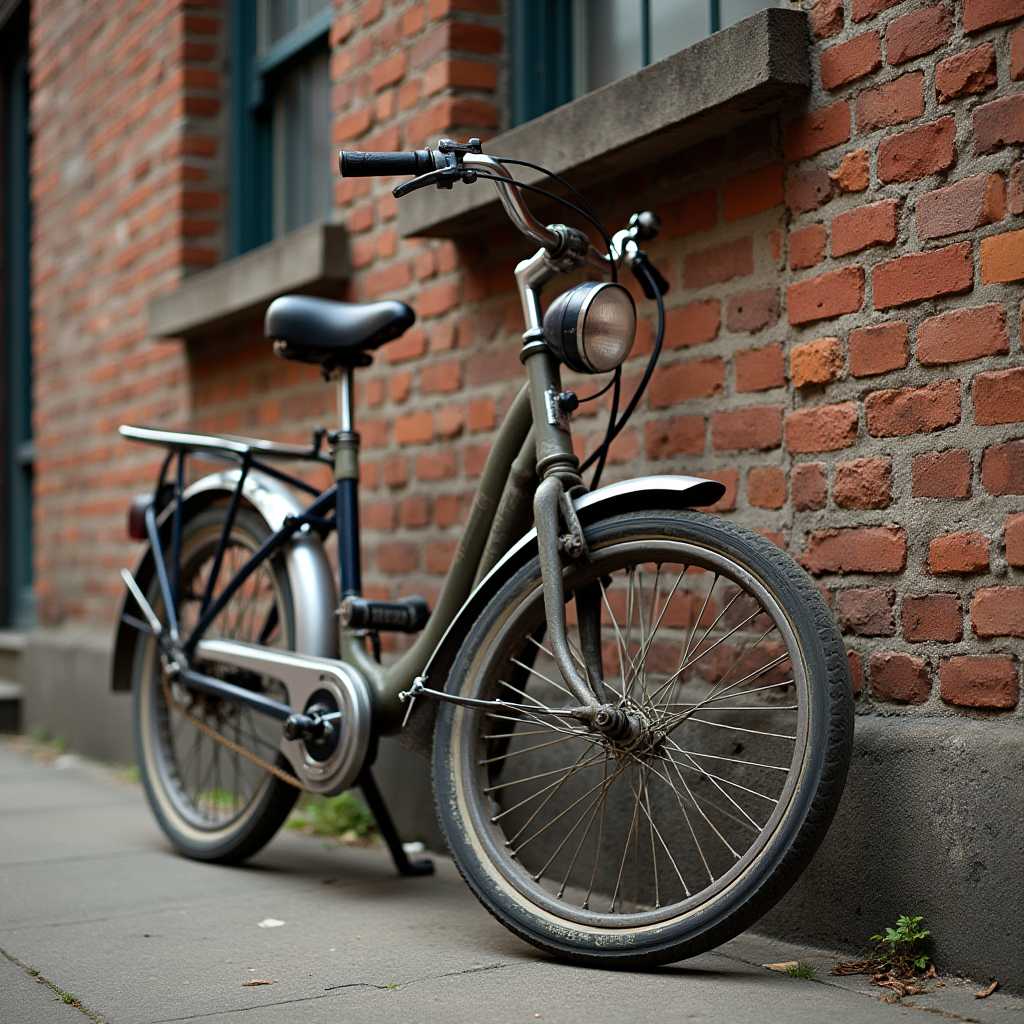} &
        \includegraphics[width=\imgwidth, height=\imgwidth]{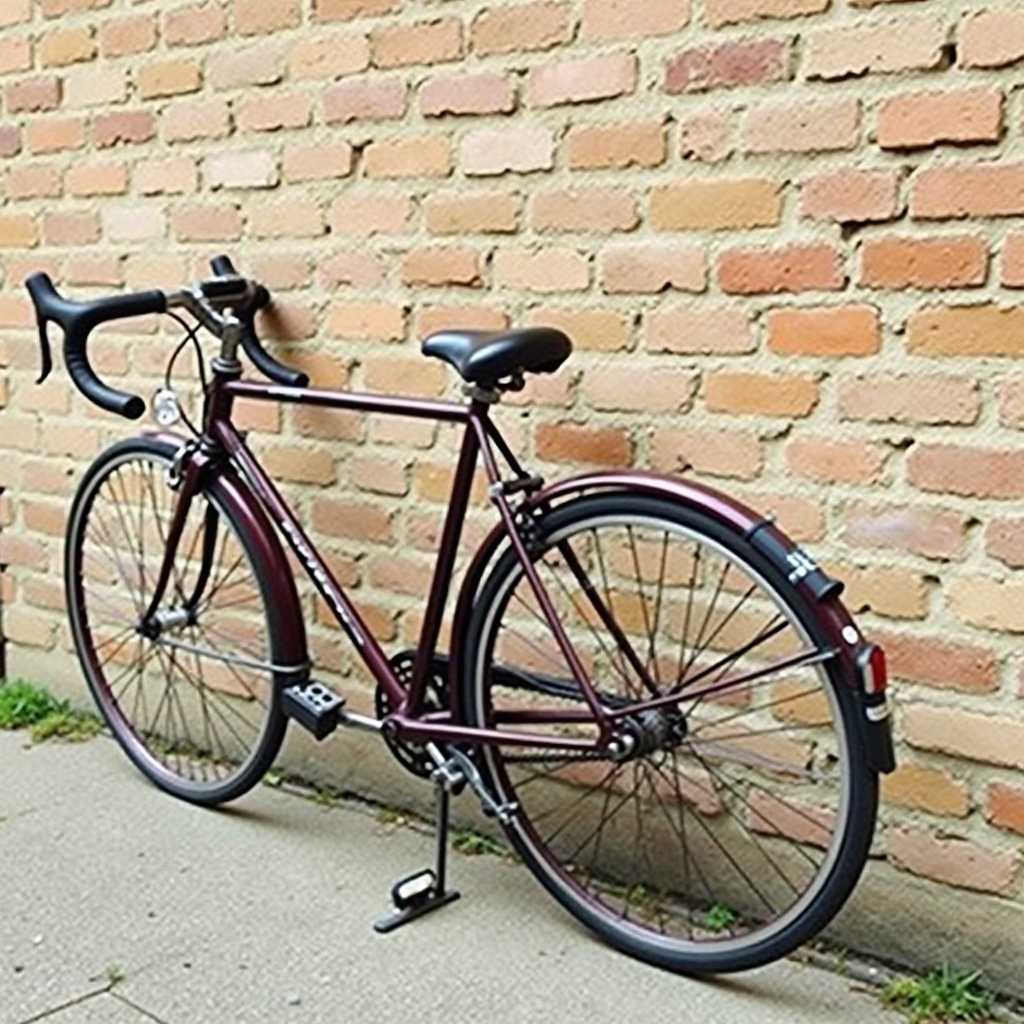} &
        \includegraphics[width=\imgwidth, height=\imgwidth]{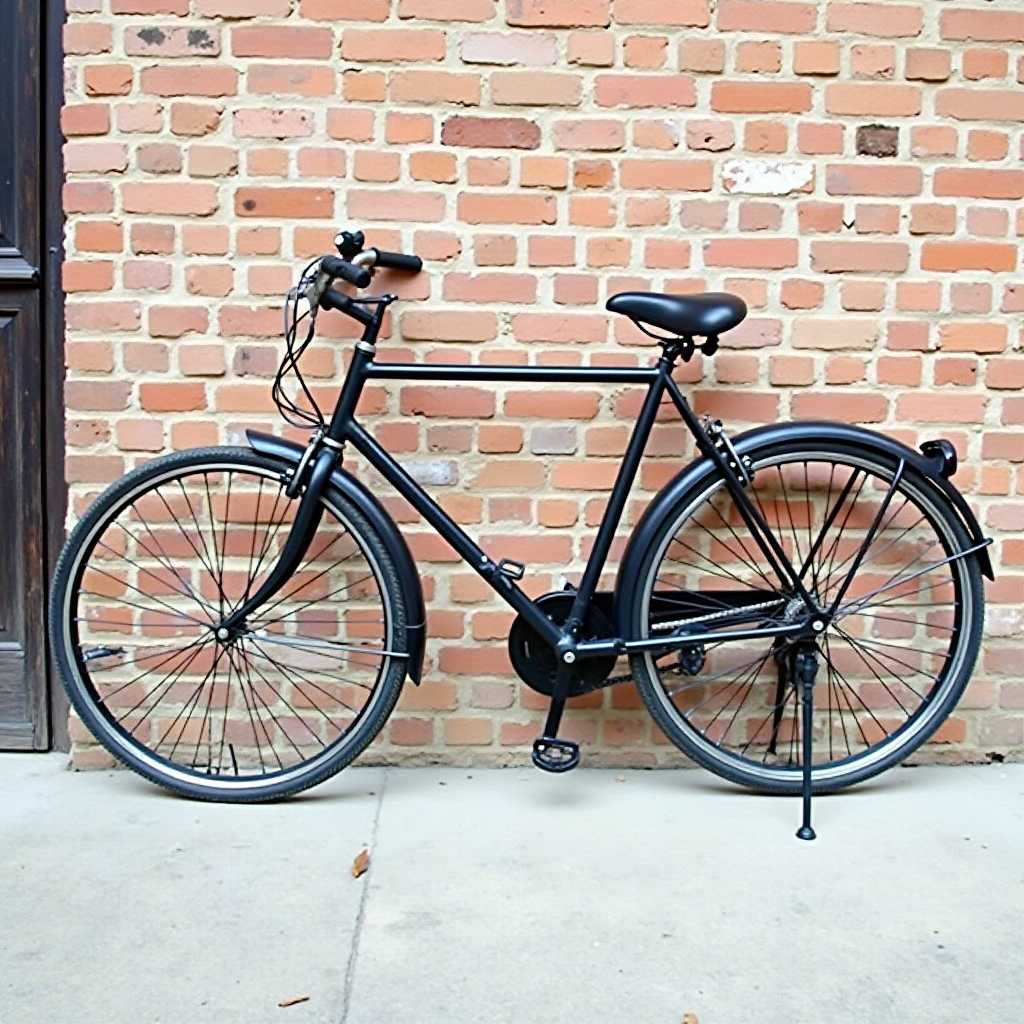} \\
        \multicolumn{9}{c}{\vspace{2pt}\small ``A classic bicycle leaned against an old brick wall'' \vspace{8pt}} \\

        \vertlabel{Flux} & 
        \includegraphics[width=\imgwidth, height=\imgwidth]{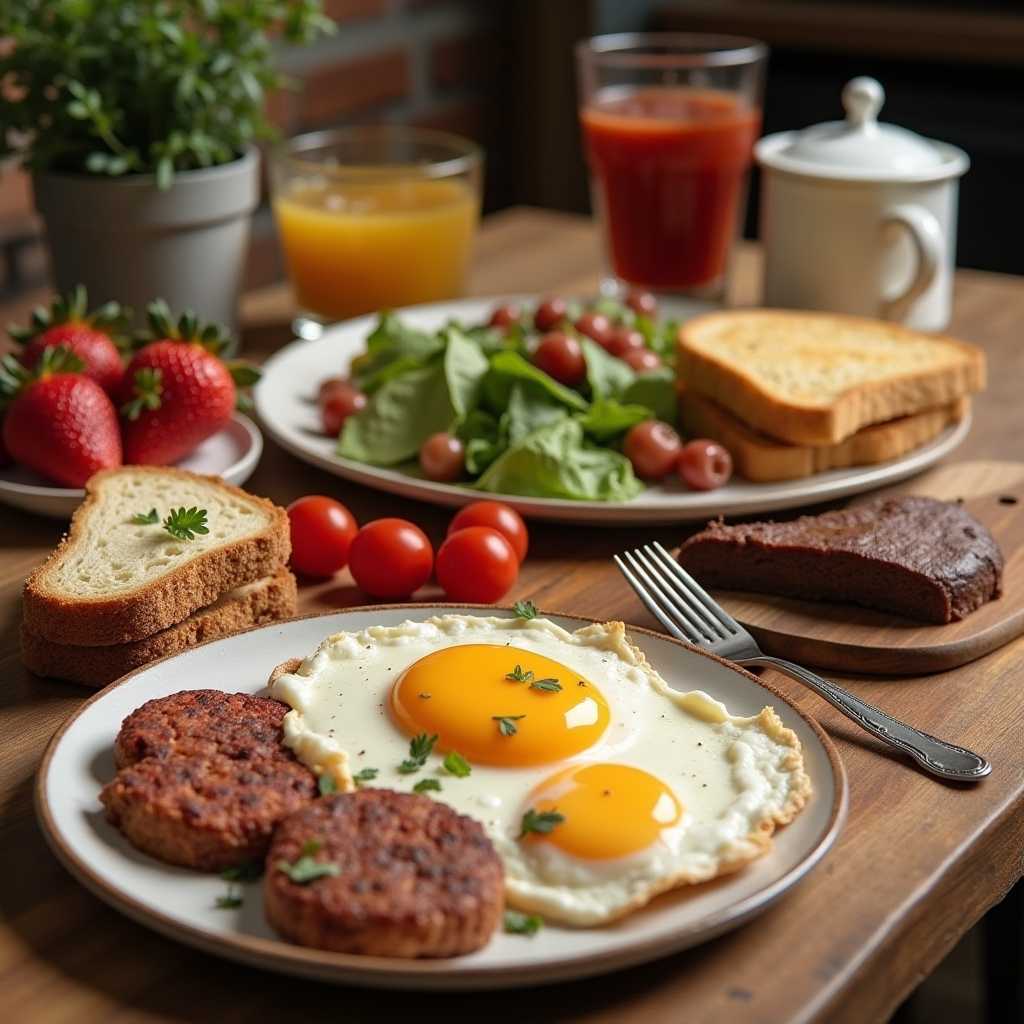} &
        \includegraphics[width=\imgwidth, height=\imgwidth]{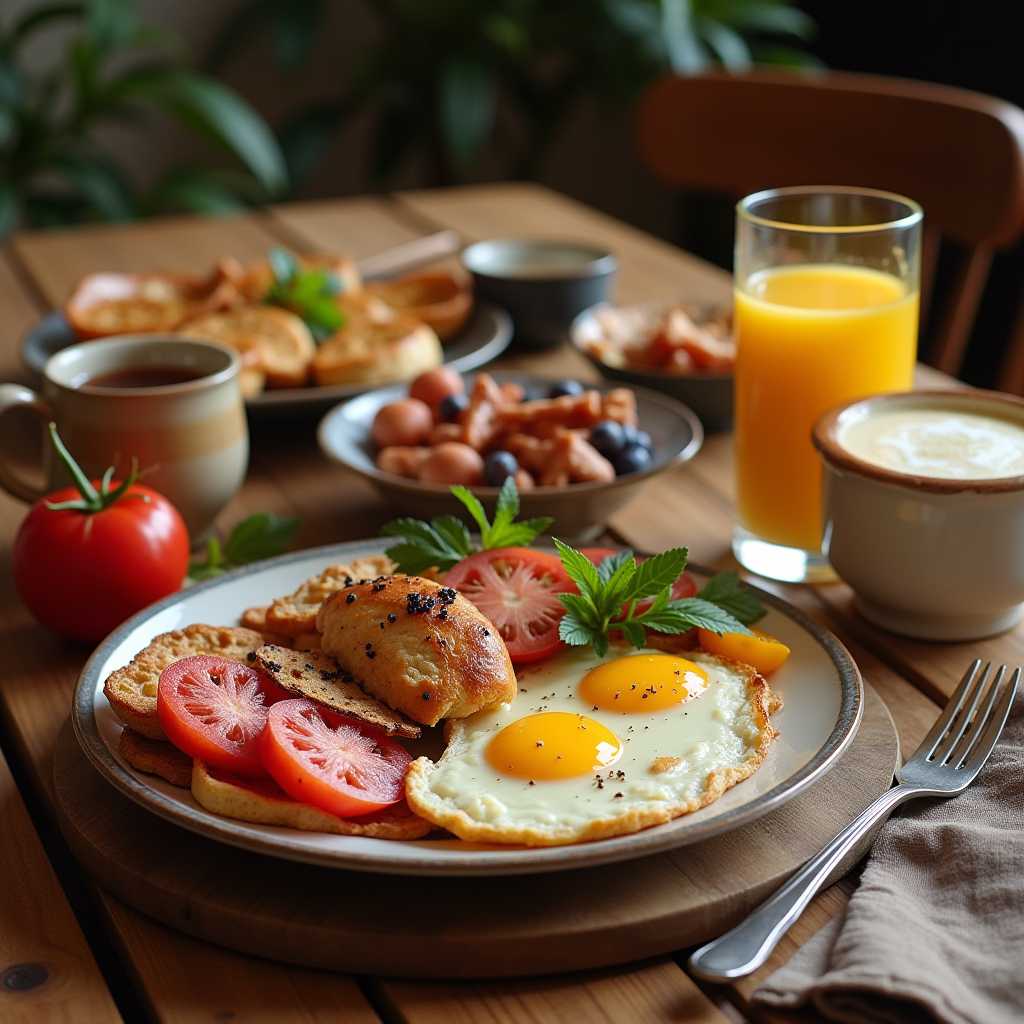} &
        \includegraphics[width=\imgwidth, height=\imgwidth]{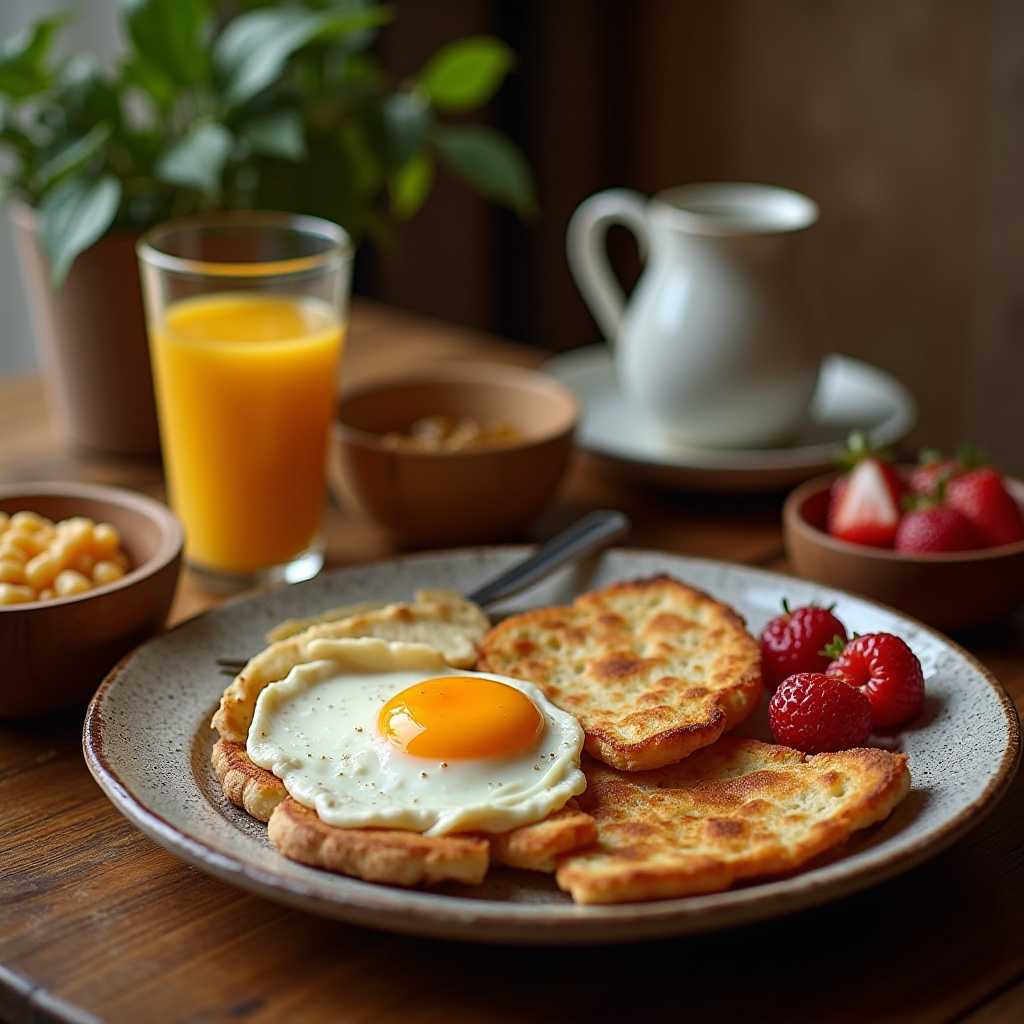} &
        \includegraphics[width=\imgwidth, height=\imgwidth]{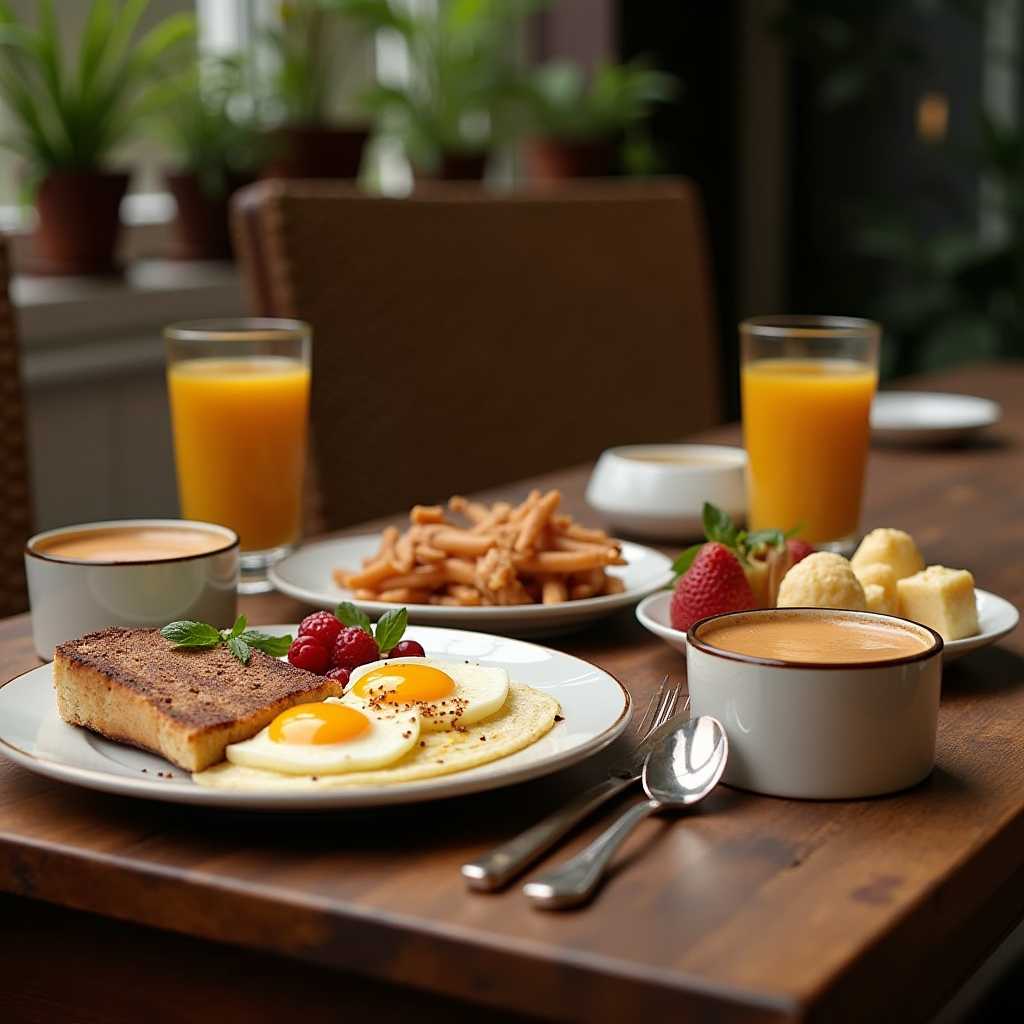} &
        \includegraphics[width=\imgwidth, height=\imgwidth]{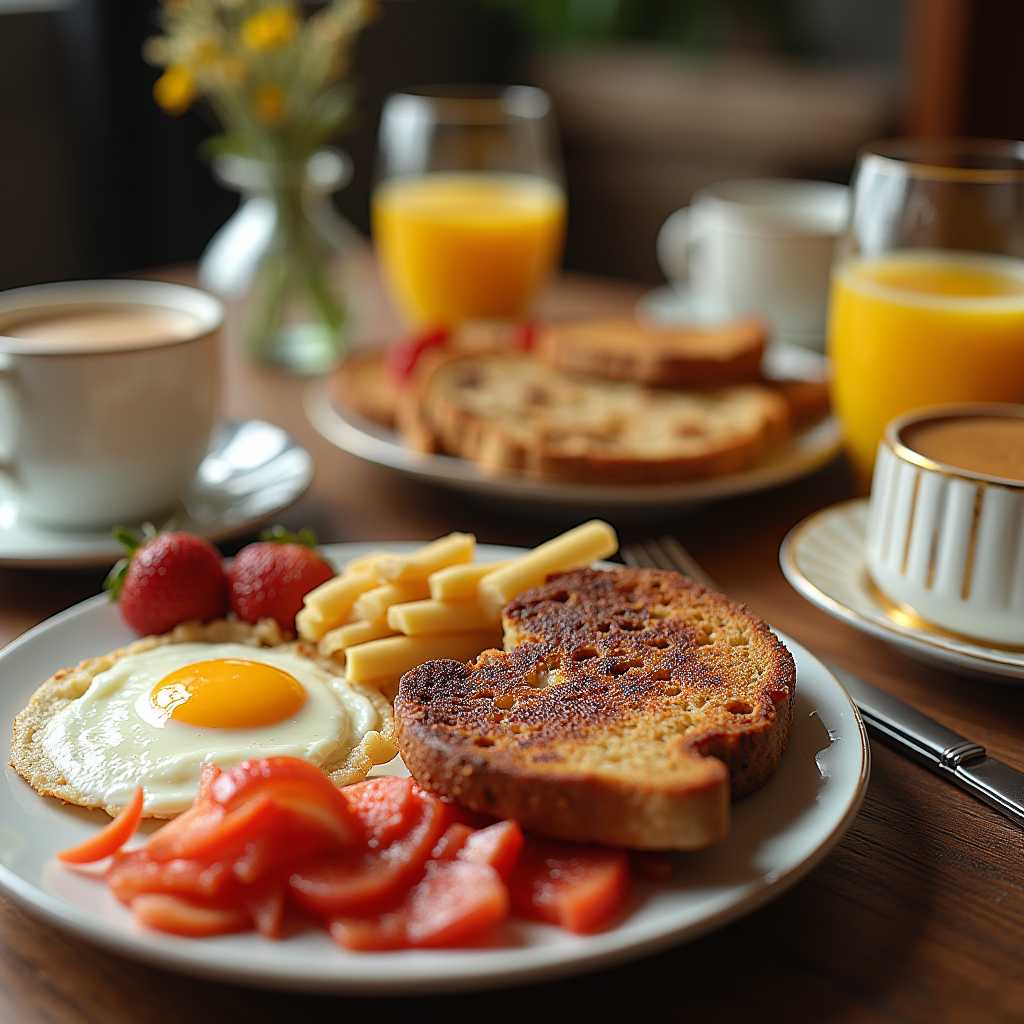} &
        \includegraphics[width=\imgwidth, height=\imgwidth]{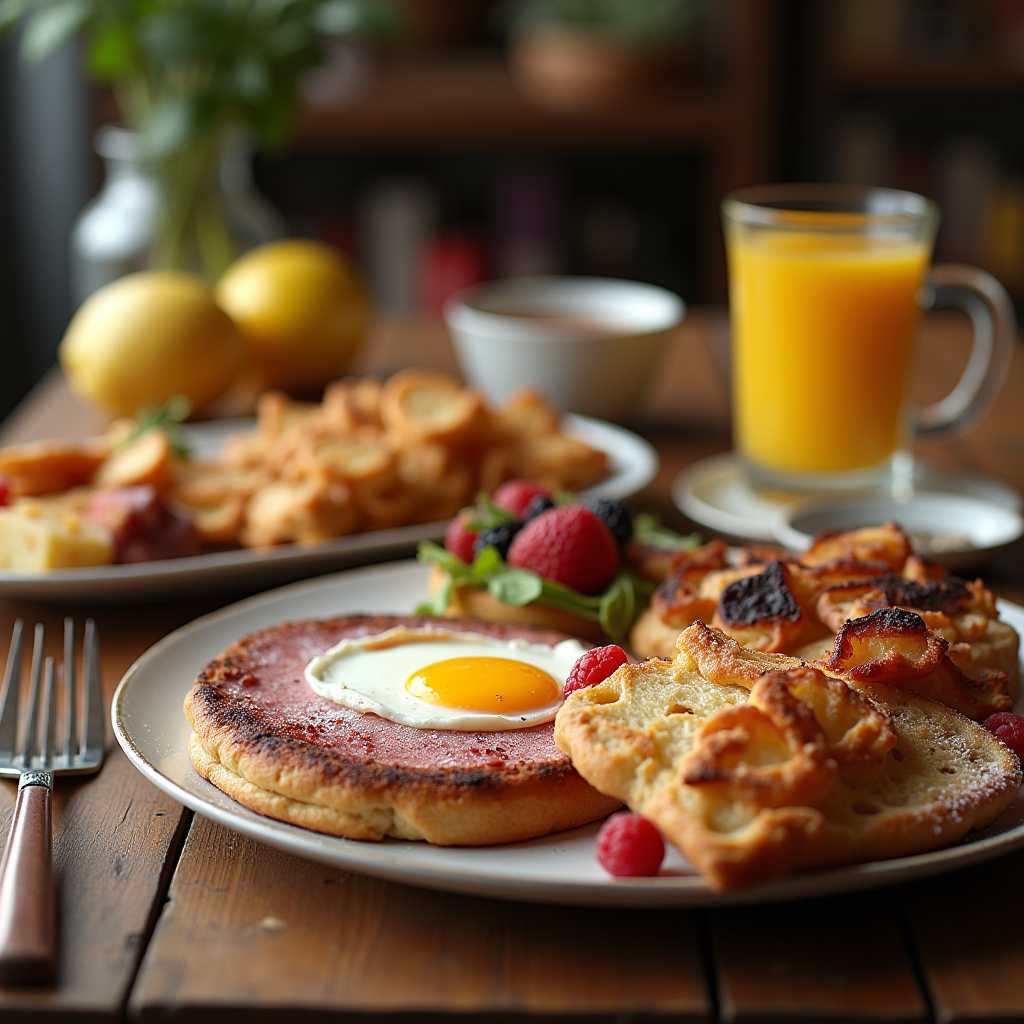} &
        \includegraphics[width=\imgwidth, height=\imgwidth]{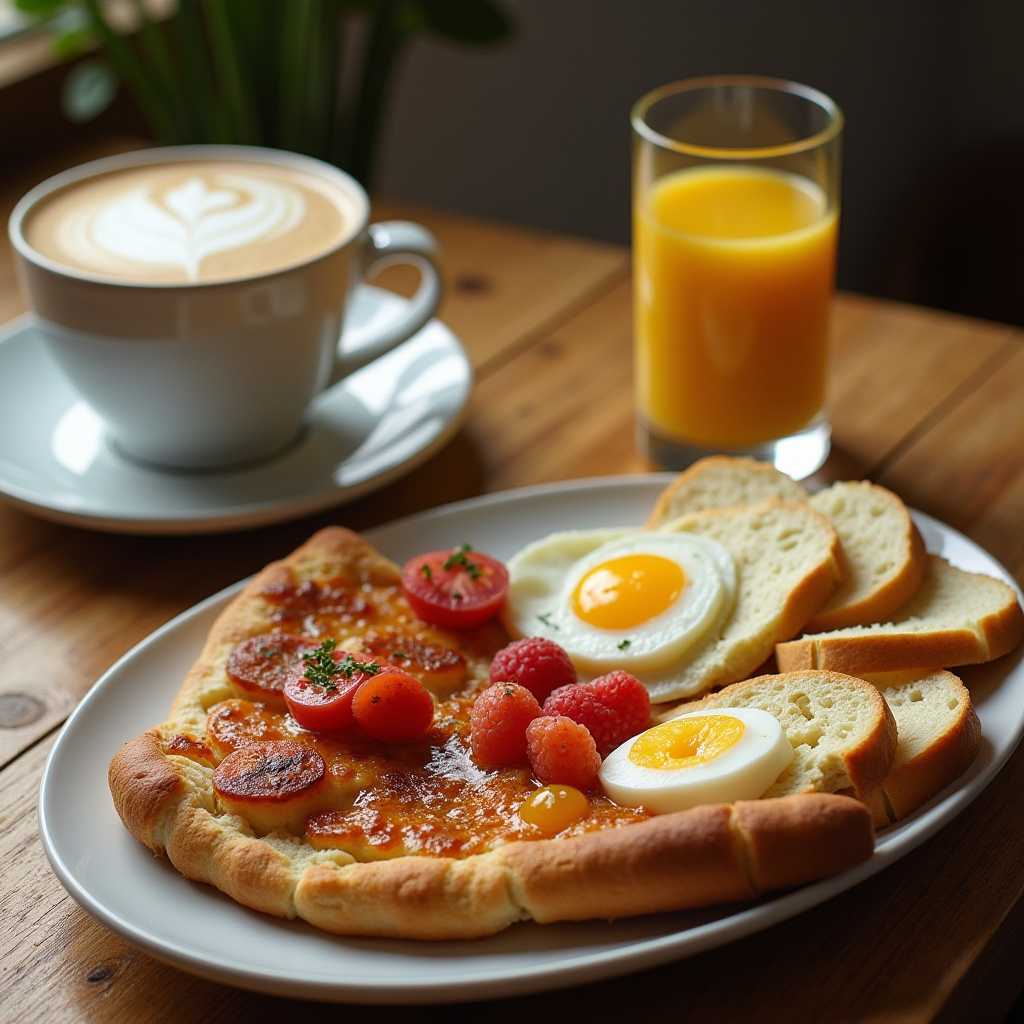} &
        \includegraphics[width=\imgwidth, height=\imgwidth]{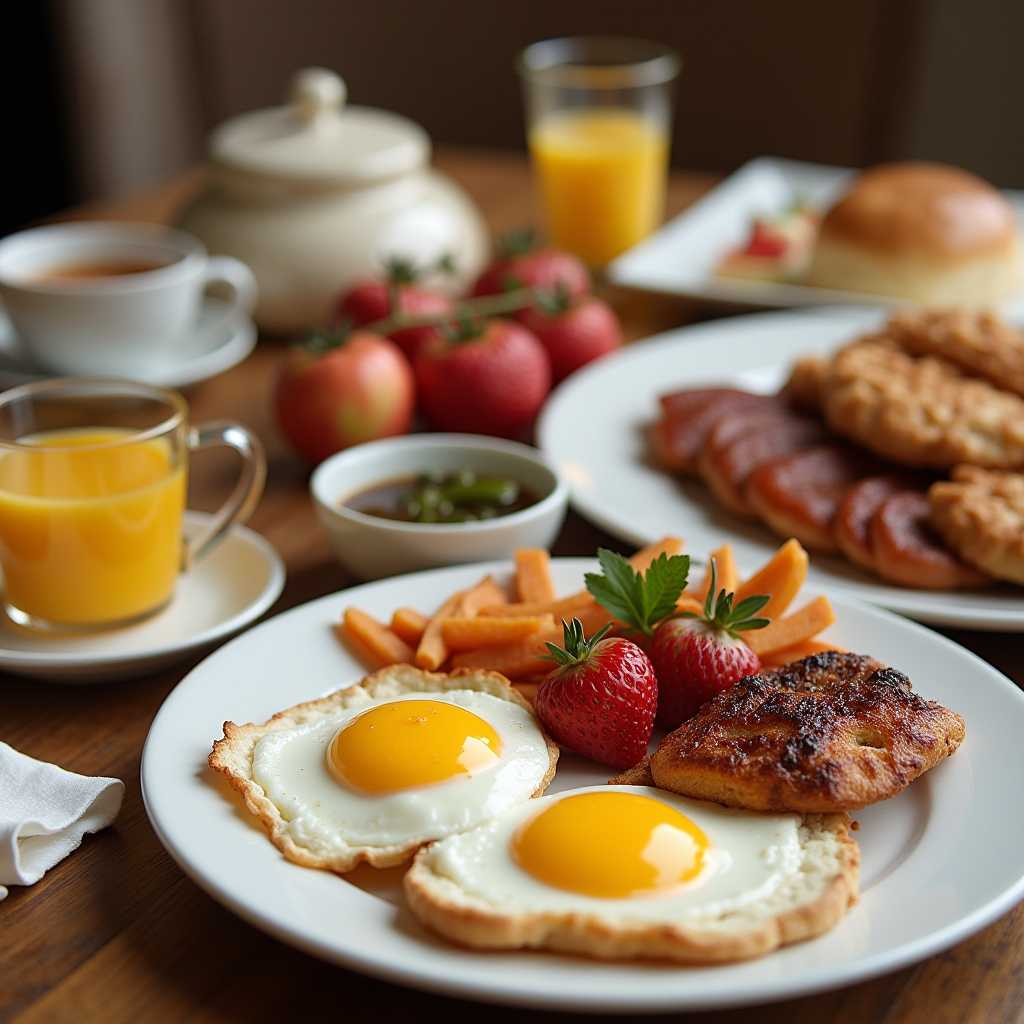} \\[-1pt]

        \vertlabel{Ours} & 
        \includegraphics[width=\imgwidth, height=\imgwidth]{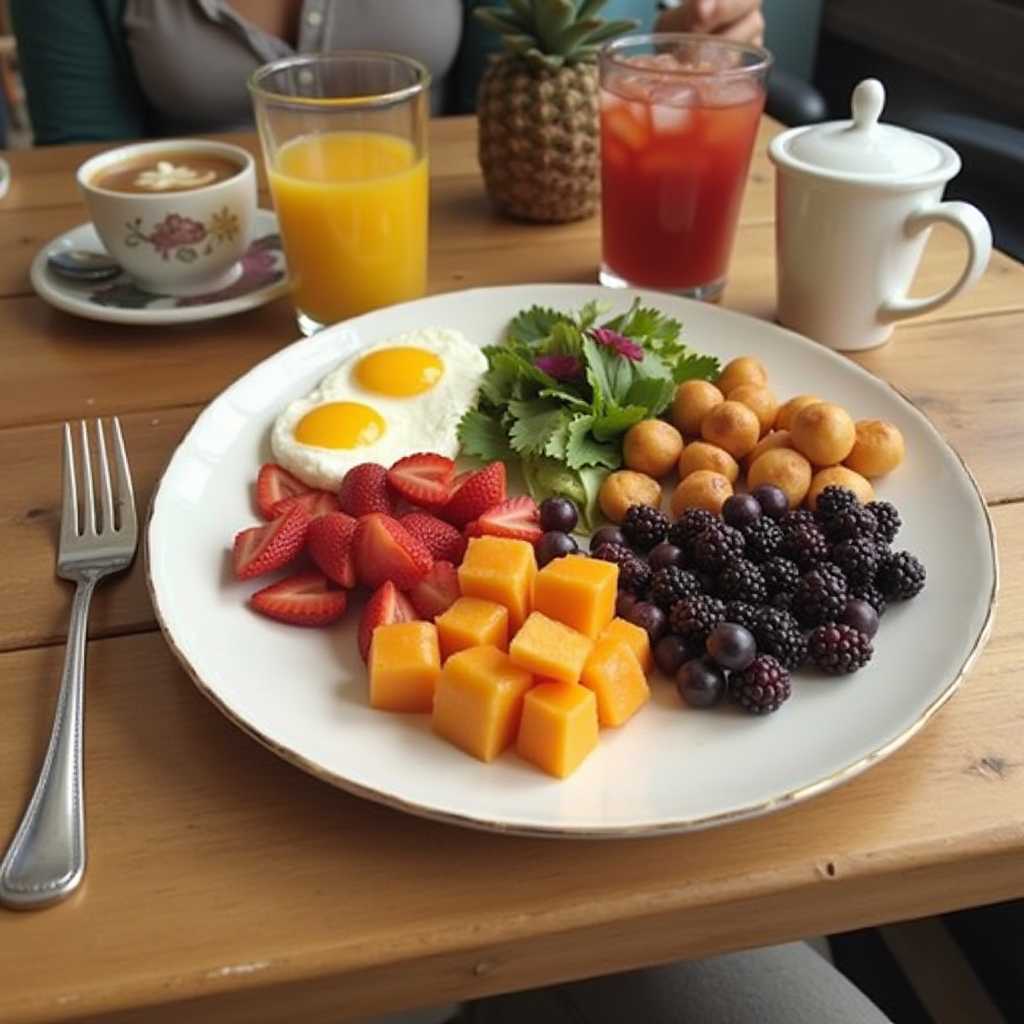} &
        \includegraphics[width=\imgwidth, height=\imgwidth]{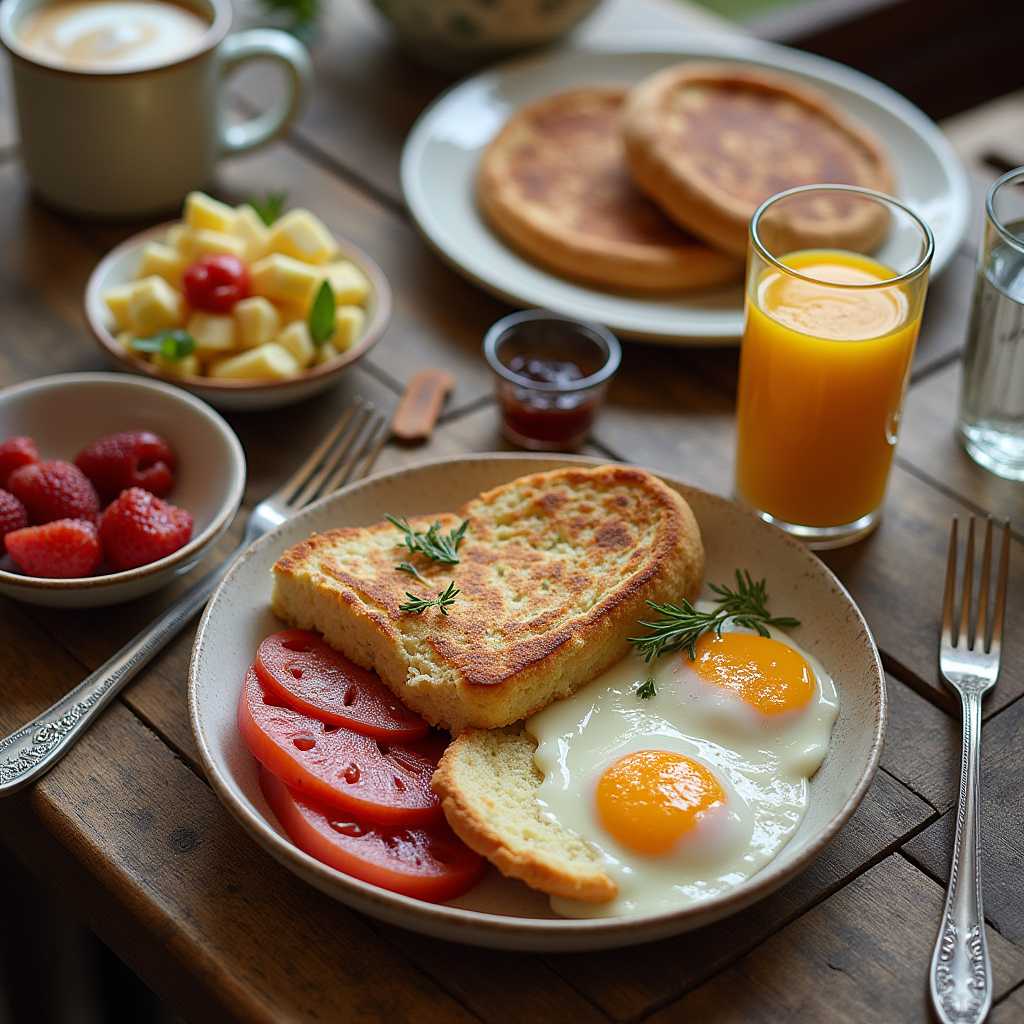} &
        \includegraphics[width=\imgwidth, height=\imgwidth]{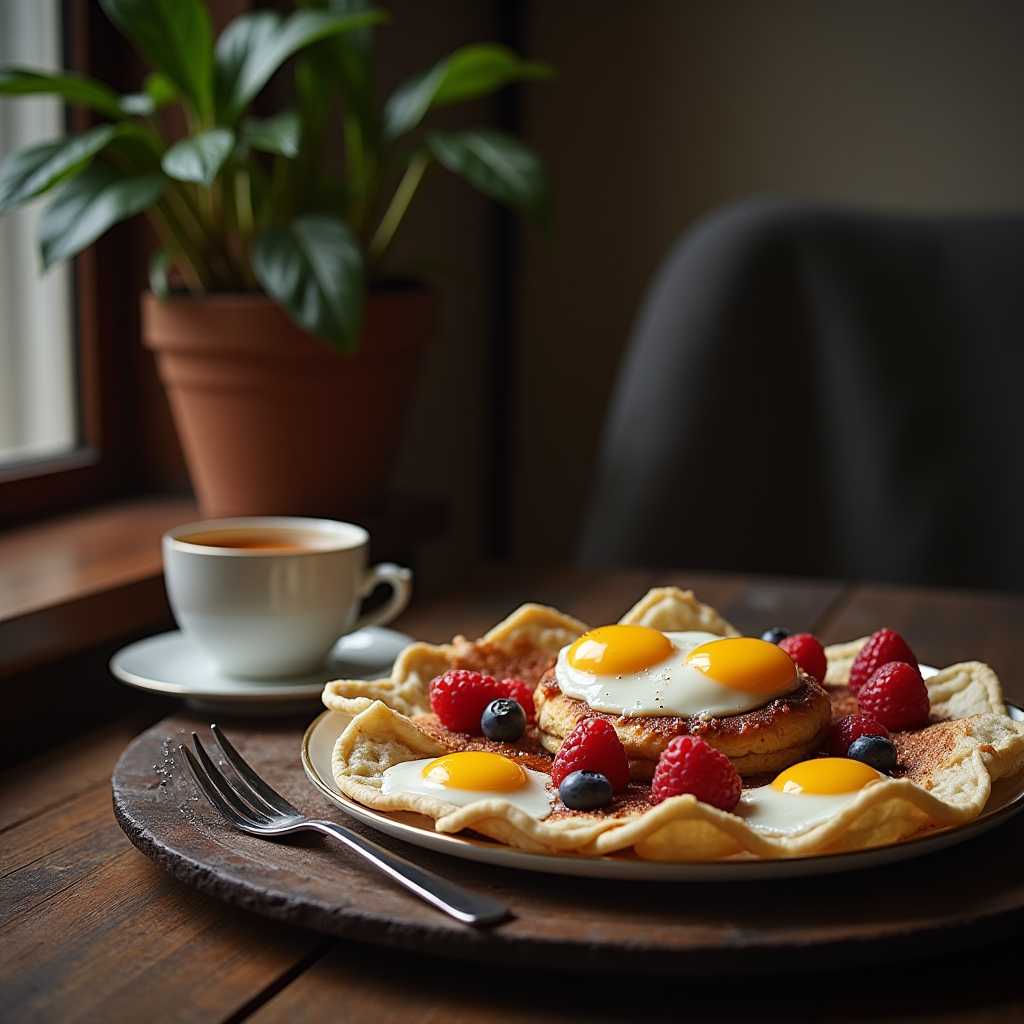} &
        \includegraphics[width=\imgwidth, height=\imgwidth]{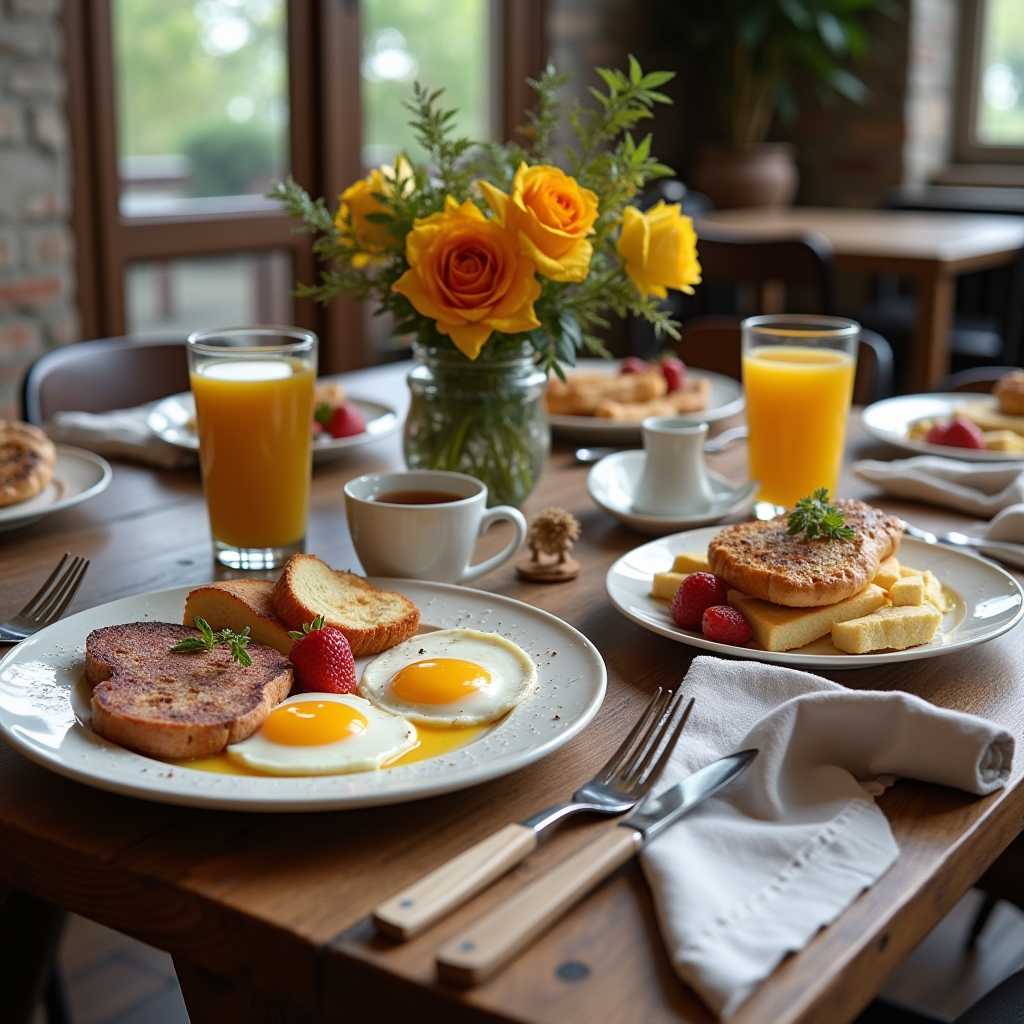} &
        \includegraphics[width=\imgwidth, height=\imgwidth]{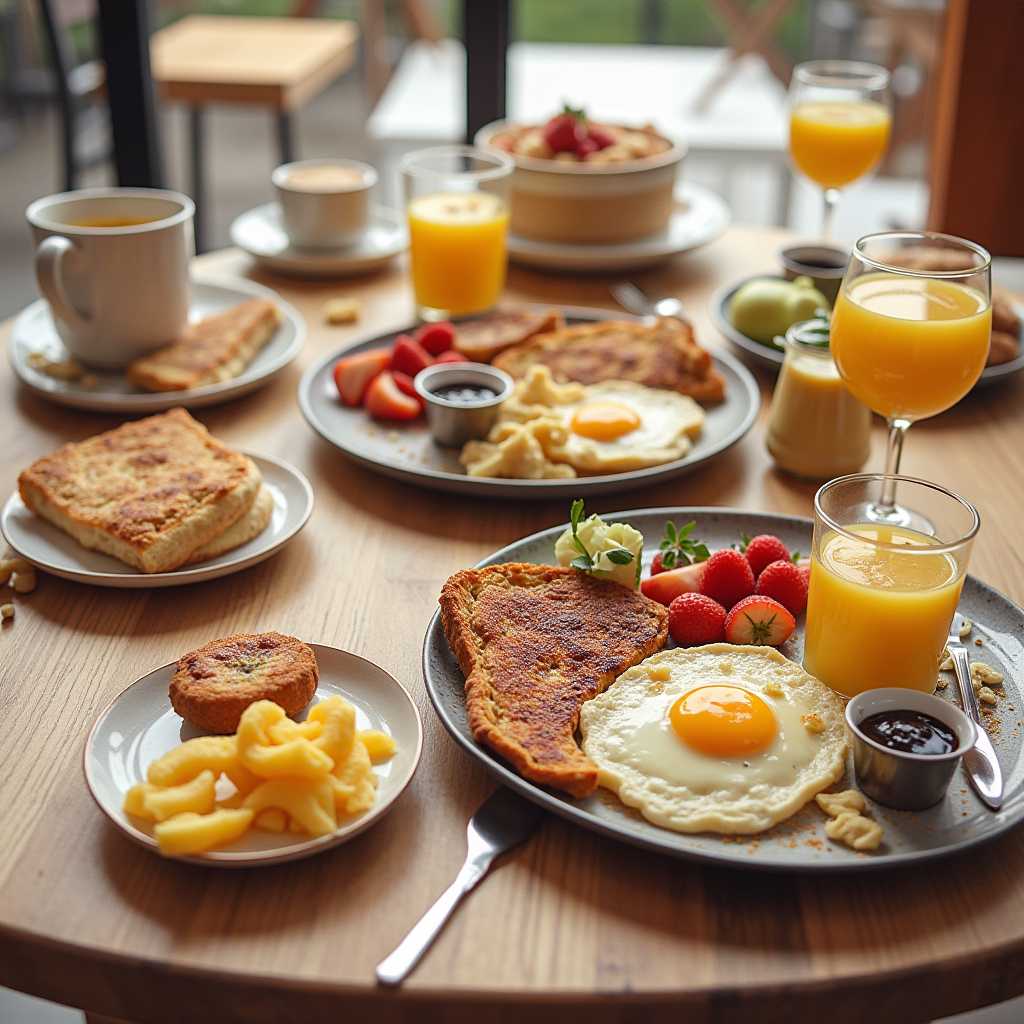} &
        \includegraphics[width=\imgwidth, height=\imgwidth]{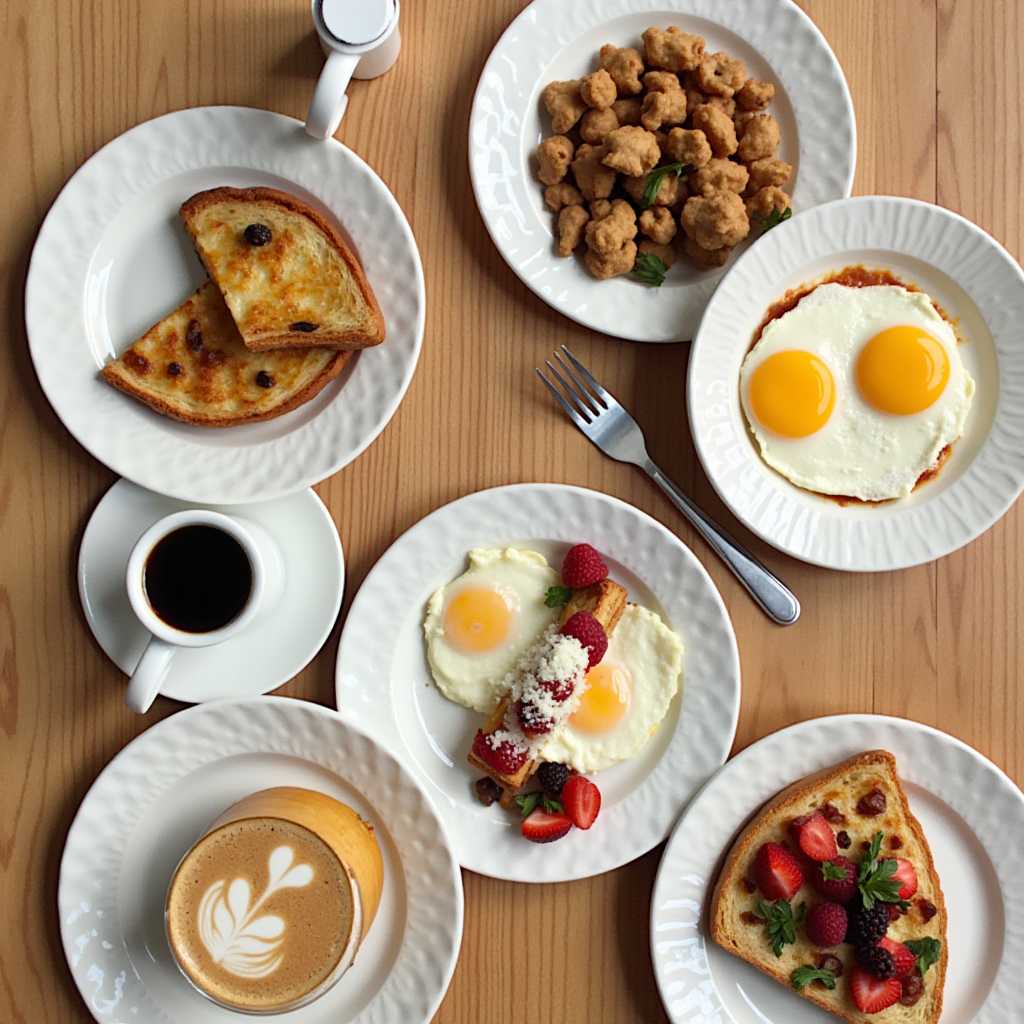} &
        \includegraphics[width=\imgwidth, height=\imgwidth]{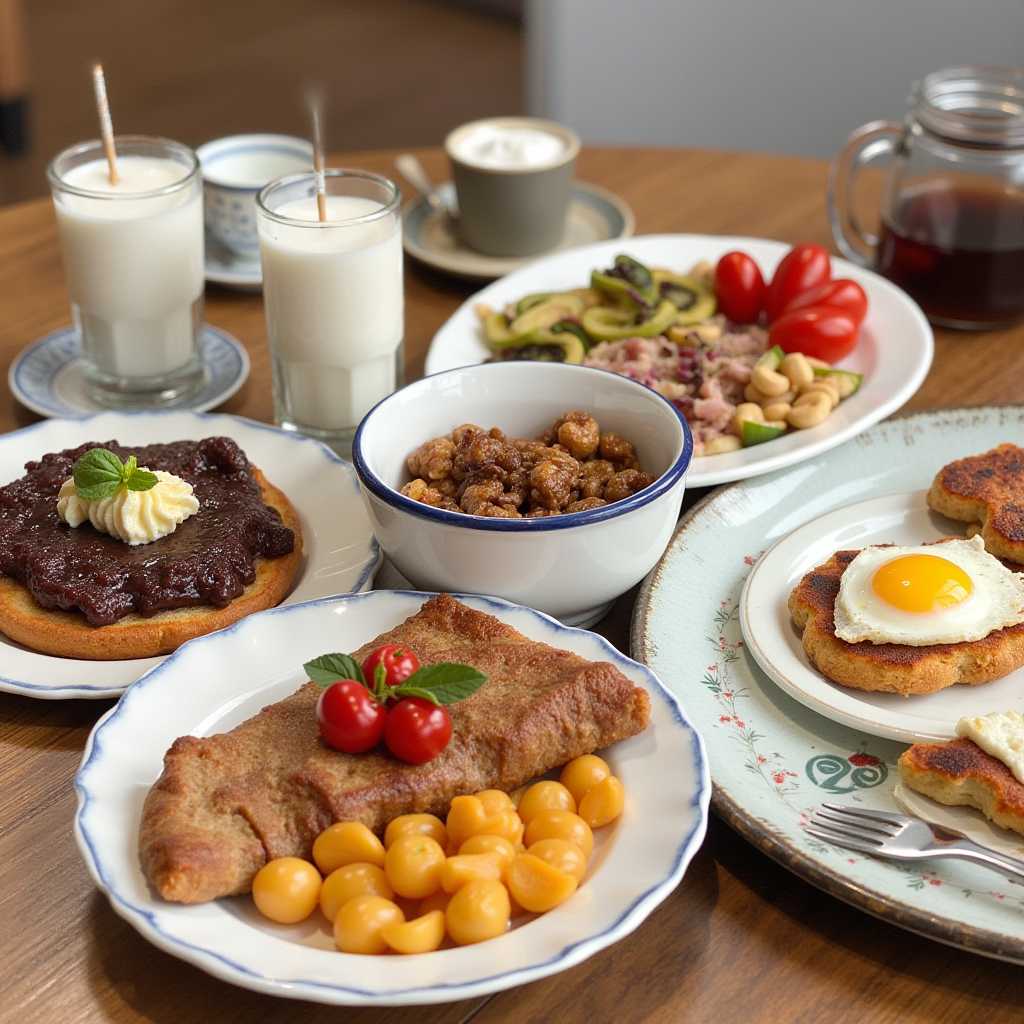} &
        \includegraphics[width=\imgwidth, height=\imgwidth]{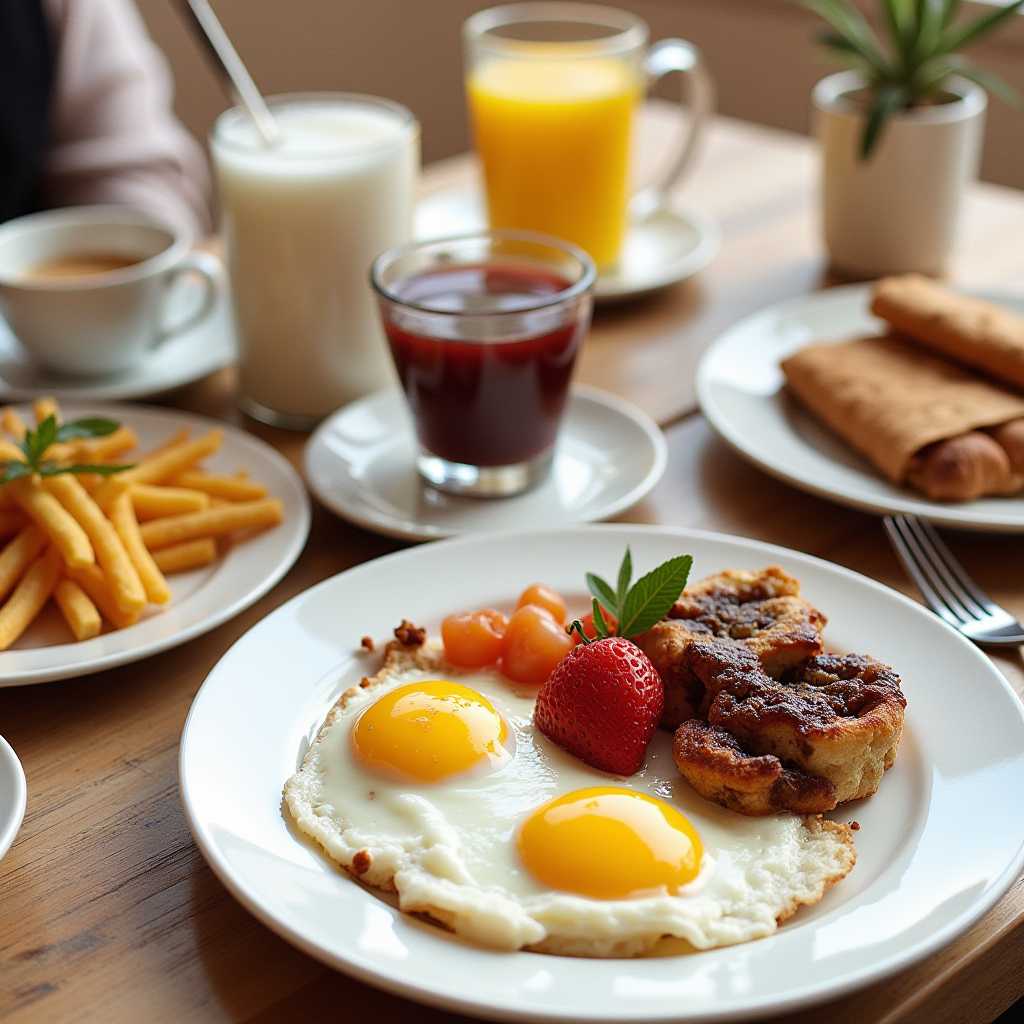} \\
        \multicolumn{9}{c}{\vspace{2pt}\small ``A delicious breakfast spread served on a wooden table'' \vspace{8pt}} \\

    \end{tabular}
    \caption{\textbf{Additional qualitative results on Flux-dev.} All batches were generated using the same random seed initialization.}\label{fig:extra_res3}
\end{figure*}

We present additional qualitative results of our method on SD3.5-Large (Figure~\ref{fig:large}), SD3.5-Turbo (Figure~\ref{fig:turbo}) and Flux-dev (Figures~\ref{fig:extra_res1}, \ref{fig:extra_res2}, and \ref{fig:extra_res3}).

\section{Additional Quantitative Results} 
\label{sec:additional_quntitative_results}

\paragraph{Additional comparisons.}
\label{par:additional_metrics}

\begin{table}[t] \centering \small 
\caption{\textbf{Detailed metrics for the Flux-dev Pareto frontiers in Figure~\ref{fig:flux_quant}.}}
\label{tab:flux_metrics} 
\resizebox{\linewidth}{!}{  \begin{tabular}{llcccc}
\toprule
Method &  & Vendi ($\uparrow$) & IR ($\uparrow$) \hspace{1pt} & VQA ($\uparrow$) \hspace{1pt} & KID $\times 10^{-4}$ ($\downarrow$) \\
\midrule
\multirow{1}{*}{Base Model} &  & 1.780 & 1.075 & 0.883 &  \\
\midrule

{\multirow{4}{*}{Ours}} & ${\eta = 5\cdot10^{6}}$ & {1.810} & {1.102} & {0.884} & {0.066} \\
{} & ${\eta = 1\cdot10^{7}}$ & {1.831} & {1.092} & {0.883} & {0.103} \\
{} & ${\eta = 1\cdot10^{8}}$ & {1.869} & {1.075} & {0.883} & {0.157} \\
{} & ${\eta = 5\cdot10^{8}}$ & {1.898} & {1.070} & {0.880} & {0.172} \\
\midrule

\multirow{4}{*}{CADS} 
 & $s = 10^{-20}$ & 1.908 & 0.377 & 0.719 & 0.558 \\
 & $s = 10^{-18}$ & 1.908 & 0.377 & 0.719 & 0.558 \\
 & $s = 10^{-12}$ & 1.910 & 0.303 & 0.699 & 0.530 \\
 & $s = 10^{-11}$ & 1.923 & 0.208 & 0.674 & 0.588 \\
\midrule

\multirow{3}{*}{PG} & $s = 1$ & 1.753 & 0.991 & 0.871 & 0.555 \\
 & $s = 80$ & 1.759 & 1.018 & 0.864 & 0.675 \\
 & $s = 150$ & 1.787 & 0.846 & 0.848 & 2.650 \\
\midrule
\multirow{4}{*}{SGI} & 8 Candidates & 1.778 & 1.152 & 0.875 & 0.440 \\
 & 16 Candidates & 1.829 & 1.085 & 0.873 & 0.461 \\
 & 32 Candidates & 1.860 & 1.063 & 0.872 & 0.289 \\
 & 64 Candidates & 1.916 & 1.042 & 0.872 & 0.297 \\
\midrule
\multirow{2}{*}{SPARKE} & $s = 0.01$ & 1.790 & 1.094 & 0.884 & 0.057 \\
 & $s = 0.02$ & 1.850 & 1.067 & 0.873 & 1.079 \\
\bottomrule \end{tabular}}  \end{table}

\begin{table}[t] \centering \small
\caption{\textbf{Detailed metrics for the SD3.5-Large Pareto frontiers in Figure~\ref{fig:sd_quant}.}}
\label{tab:sd_metrics}
\resizebox{\linewidth}{!}{  \begin{tabular}{llcccc}\toprule
Method &  & Vendi ($\uparrow$) \hspace{1pt} & IR ($\uparrow$) \hspace{1pt} & VQA ($\uparrow$) \hspace{1pt} & KID $\times 10^{-4}$ ($\downarrow$) \\
\midrule
\multirow{1}{*}{Base Model} &  & 1.819 & 1.051 & 0.905 &  \\
\midrule

{\multirow{4}{*}{Ours}} & ${\eta = 5\cdot10^{2}}$ & {1.851} & {1.018} & {0.904} & {0.619} \\
{} & ${\eta = 5\cdot10^{4}}$ & {1.878} & {1.012} & {0.904} & {0.627} \\
{} & ${\eta = 5\cdot10^{5}}$ & {1.941} & {0.988} & {0.900} & {0.625} \\
{} & ${\eta = 5\cdot10^{6}}$ & {1.980} & {0.940} & {0.890} & {0.445} \\
\midrule

\multirow{3}{*}{CADS} 
 & $s = 10^{-12}$ & 2.004 & 0.131 & 0.717 & 0.941 \\
 & $s = 10^{-10}$ & 2.025 & 0.051 & 0.692 & 0.953 \\
 & $s = 10^{-08}$ & 2.018 & 0.066 & 0.692 & 0.953 \\
\midrule

\multirow{3}{*}{PG} & $s = 1$ & 1.900 & 0.783 & 0.878 & 1.521 \\
 & $s = 60$ & 1.913 & 0.707 & 0.868 & 4.053 \\
 & $s = 80$ & 1.924 & 0.632 & 0.861 & 5.930 \\
\midrule
\multirow{4}{*}{SGI} & 8 Candidates & 1.828 & 1.050 & 0.903 & 0.465 \\
 & 16 Candidates & 1.862 & 1.025 & 0.902 & 0.455 \\
 & 32 Candidates & 1.883 & 1.030 & 0.902 & 0.429 \\
 & 64 Candidates & 1.915 & 1.004 & 0.901 & 0.421 \\
\midrule
\multirow{4}{*}{SPARKE} & $s = 0.01$ & 1.860 & 1.027 & 0.902 & 0.362 \\
 & $s = 0.02$ & 1.887 & 0.999 & 0.901 & 0.770 \\
 & $s = 0.03$ & 1.912 & 0.925 & 0.899 & 1.393 \\
 & $s = 0.04$ & 1.989 & 0.735 & 0.882 & 2.918 \\
\bottomrule \end{tabular}} \end{table}

\begin{table}[t] \centering \small 
\caption{\textbf{Detailed metrics for the SD3.5-Turbo Pareto frontiers in Figure~\ref{fig:turbo_quant}.}}
\label{tab:turbo_metrics}
\resizebox{\linewidth}{!}{  \begin{tabular}{llcccc}\toprule
Method &  & Vendi ($\uparrow$) \hspace{1pt} & IR ($\uparrow$) \hspace{1pt} & VQA ($\uparrow$) \hspace{1pt} & KID $\times 10^{-4}$ ($\downarrow$) \\
\midrule
\multirow{1}{*}{Base Model} &  & 1.724 & 0.978 & 0.891 &  \\
\midrule

{\multirow{4}{*}{Ours}} & ${\eta = 1\cdot10^{5}}$ & {1.819} & {0.914} & {0.887} & {1.796} \\
{} & ${\eta = 5\cdot10^{5}}$ & {1.879} & {0.899} & {0.884} & {1.786} \\
{} & ${\eta = 1\cdot10^{6}}$ & {1.914} & {0.864} & {0.876} & {1.897} \\
{} & ${\eta = 1\cdot10^{7}}$ & {2.079} & {0.562} & {0.822} & {1.914} \\
\midrule

\multirow{4}{*}{CADS} & $s = 0.1$ & 1.808 & 0.551 & 0.772 & 0.158 \\
 & $s = 0.5$ & 1.853 & 0.383 & 0.731 & 0.526 \\
 & $s = 0.8$ & 1.911 & 0.180 & 0.683 & 1.319 \\
 & $s = 0.9$ & 1.958 & 0.127 & 0.673 & 1.348 \\
\midrule
\multirow{3}{*}{PG} & $s = 2$ & 1.765 & 0.915 & 0.884 & 0.881 \\
 & $s = 10$ & 1.857 & 0.638 & 0.859 & 2.285 \\
 & $s = 40$ & 1.926 & 0.221 & 0.821 & 14.128 \\
\midrule
\multirow{5}{*}{SGI} & 4 Candidates & 1.707 & 0.962 & 0.888 & 0.078 \\
 & 8 Candidates & 1.775 & 0.944 & 0.889 & 0.079 \\
 & 16 Candidates & 1.829 & 0.933 & 0.883 & 0.005 \\
 & 32 Candidates & 1.853 & 0.923 & 0.884 & 0.028 \\
 & 64 Candidates & 1.879 & 0.913 & 0.886 & 0.120 \\
\midrule
\multirow{5}{*}{SPARKE} & $s = 0.04$ & 1.728 & 1.011 & 0.890 & 0.206 \\
 & $s = 0.08$ & 1.763 & 0.928 & 0.885 & 0.744 \\
 & $s = 0.1$ & 1.812 & 0.837 & 0.871 & 1.219 \\
 & $s = 0.12$ & 1.869 & 0.629 & 0.850 & 2.742 \\
 & $s = 0.14$ & 1.970 & 0.231 & 0.803 & 7.037 \\
\bottomrule \end{tabular}} \end{table}

We present additional quantitative comparisons on SD3.5-Large (Figure~\ref{fig:sd_quant}) and SD3.5-Turbo (Figure~\ref{fig:turbo_quant}). Our method achieves competitive quality-diversity trade-offs at a fraction of the computational cost required by SGI. Detailed metrics across all evaluated models are provided in Tables \ref{tab:flux_metrics}, \ref{tab:sd_metrics}, and \ref{tab:turbo_metrics}.

\paragraph{User study table}
\label{par:user_study_table}
We provide the full results of our user study in Table~\ref{tab:user_study_results}.

\paragraph{Evaluation on detailed prompts}
\label{par:detailed_prompts}While diversity is typically easier to achieve when prompts leave significant room for interpretation, we evaluate our method on the 100 longest prompts from the ``Complex'' and ``Fine-Grained Detail'' categories of PartiPrompts~\cite{yu2022scalingautoregressivemodelscontentrich} using Flux-dev. Even under these highly constrained conditions, our method increases diversity and human preference scores with a negligible impact on prompt alignment. Specifically, we observe an increase in Vendi score ($+0.08$) and ImageReward ($+0.05$), while VQAScore remains nearly constant ($-0.01$). These results demonstrate that intervening in the Contextual Space effectively identifies and navigates remaining semantic degrees of freedom, even in the presence of extensive conditioning.

\begin{table}[t]
\centering
\caption{\textbf{User study results comparing our method against five competing approaches across four evaluation metrics.} Values show the percentage of times users preferred our method (Ours), the competitor (Comp.), or rated both equally (Tie). Results are aggregated from 450 pairwise comparisons per metric.}
\label{tab:user_study_results}
\resizebox{\columnwidth}{!}{
\begin{tabular}{llcccccc}
\toprule
Metric & Choice & Base Model & CADS & SGI & PG & SPARKE & Average \\
\midrule
\multirow{3}{*}{Diversity} & Ours & 71.6 & 52.2 & 56.7 & 80.0 & 34.4 & 61.1 \\
                           & Comp. & 12.9 & 30.0 & 11.1 & 14.4 & 53.1 & 22.0 \\
                           & Tie & 15.5 & 17.8 & 32.2 & 5.6 & 12.5 & 16.9 \\
\midrule
\multirow{3}{*}{Quality}   & Ours  & 49.1 & 67.8 & 15.6 & 82.2 & 85.9 & 58.0 \\
                           & Comp. & 6.9  & 11.1 & 31.1 & 12.2 & 3.1  & 13.1 \\
                           & Tie   & 44.0 & 21.1 & 53.3 & 5.6  & 10.9 & 28.9 \\
\midrule
\multirow{3}{*}{Adherence} & Ours  & 25.0 & 74.4 & 13.3 & 67.8 & 79.7 & 48.9 \\
                           & Comp. & 15.5 & 11.1 & 22.2 & 13.3 & 4.7  & 14.0 \\
                           & Tie   & 59.5 & 14.4 & 64.4 & 18.9 & 15.6 & 37.1 \\
\midrule
\multirow{3}{*}{Overall} & Ours & 57.8 & 74.4 & 31.1 & 83.3 & 87.5 & 65.1 \\
                                   & Comp. & 13.8 & 15.6 & 27.8 & 10.0 & 9.4  & 15.6 \\
                                   & Tie   & 28.4 & 10.0 & 41.1 & 6.7  & 3.1  & 19.3 \\
\midrule
\multirow{3}{*}{All Metrics}      & Ours  & 50.9 & 67.2 & 29.2 & 78.3 & 71.9 & 58.3 \\
                                 & Comp. & 12.3 & 16.9 & 23.1 & 12.5 & 17.6 & 16.2 \\
                                 & Tie   & 36.9 & 15.8 & 47.8 & 9.2  & 10.5 & 25.6 \\
\bottomrule
\end{tabular}
}
\end{table}

\newpage
\section{Additional Ablation Studies}
\label{sec:more_ablations}
\paragraph{Attention score repulsion.}

We investigate the impact of applying the repulsion mechanism directly to the attention scores within Flux-dev. Despite utilizing FlexAttention~\cite{dong2024flex} for optimization, this approach introduced significant computational overhead while paradoxically yielding a reduction in both semantic diversity and human preference scores ($-0.075$ Vendi, $-0.178$ ImageReward).

\paragraph{Batch size ablation}
\begin{table}[t]
\centering
\small
\caption{\textbf{Scalability across batch sizes.} Quantitative results on SD3.5-Turbo for varying batch sizes. We report the average Vendi score per pair to normalize for batch size constraints.}
\label{tab:batch_size_ablation}
\begin{tabular}{lccc}
\toprule
Batch size & Vendi & Vendi (avg. pair) & ImageReward \\
\midrule
4  & 1.819 & 1.393 & 0.914 \\
8  & 2.295 & 1.401 & 0.923 \\
16 & 2.768 & 1.404 & 0.928 \\
\bottomrule
\end{tabular}
\end{table}

We examine the scalability of our method by evaluating performance across varying batch sizes on SD3.5-Turbo. To ensure a fair comparison across different sample counts, we report the average Vendi score per pair, as the raw Vendi score is inherently bounded by the batch size. As shown in Table~\ref{tab:batch_size_ablation}, our method exhibits a consistent positive trend across all evaluated metrics as the batch size increases. This suggests that the repulsion mechanism scales effectively and benefits from the denser representation of the conditional manifold provided by larger batches.

\paragraph{Timestep ablation}
\label{par:timestep_ablation}
\begin{table}[t]
\centering
\small
\caption{\textbf{Effect of the timestep interval on diversity and human preference.} We evaluate different intervention windows during the diffusion trajectory for SD3.5-Large and SD3.5-Turbo.}
\label{tab:timestep_ablation}

\begin{tabular}{llcc}
\toprule
Model & Timestep interval & Vendi & ImageReward \\
\midrule

SD3.5-Turbo & [0,1/4] & 1.764 & 0.829 \\
            & [1/4,2/4] & 1.776 & 0.811 \\
            & [2/4,3/4] & 1.809 & 0.745 \\
            & [3/4,1] & 1.988 & 0.660 \\
            & [0,1] & 2.064 & 0.501 \\

\midrule

SD3.5-Large & [0,1/7] & 1.849 & 0.942 \\
            & [1/7,2/7] & 1.854 & 0.942 \\
            & [2/7,3/7] & 1.849 & 0.946 \\
            & [3/7,4/7] & 1.847 & 0.932 \\
            & [4/7,5/7] & 1.848 & 0.954 \\
            & [5/7,6/7] & 1.900 & 0.919 \\
            & [6/7,1] & 1.960 & 0.852 \\
            & [0,1] & 2.135 & 0.535 \\

\bottomrule
\end{tabular}
\end{table}

We analyze the impact of the repulsion window across the diffusion trajectory by applying the intervention within specific timestep intervals while keeping all other hyperparameters constant. Table~\ref{tab:timestep_ablation} summarizes these results. For both SD3.5-Large and SD3.5-Turbo, applying repulsion later in the trajectory typically improves ImageReward at the expense of diversity. Conversely, maintaining the intervention throughout the entire trajectory yields the highest diversity but results in a more pronounced decline in fidelity and alignment scores.

\paragraph{Transformer block ablation}
\label{par:block_ablation}
\begin{table}[t]
\centering
\small
\caption{\textbf{Performance across different transformer block groups.} Results are reported for interventions applied to the first, middle, or last third of the blocks for SD3.5-Large and SD3.5-Turbo.}
\label{tab:block_ablation}

\begin{tabular}{lcccc}
\toprule
 & \multicolumn{2}{c}{SD3.5-Turbo} & \multicolumn{2}{c}{SD3.5-Large} \\
\cmidrule(r){2-3}\cmidrule(l){4-5}
Block group & Vendi & ImageReward & Vendi & ImageReward \\
\midrule

First third  & 1.878 & 0.774 & 1.887 & 0.895 \\
Middle third & 1.947 & 0.844 & 1.947 & 0.902 \\
Last third   & 1.765 & 0.913 & 1.835 & 0.985 \\
All blocks   & 1.764 & 0.829 & 1.960 & 0.852 \\

\bottomrule
\end{tabular}
\end{table}

We further investigate how the selection of transformer blocks influences performance by restricting the intervention to the first, middle, or last third of the architecture's blocks. As reported in Table~\ref{tab:block_ablation}, applying repulsion to the middle blocks yields the strongest diversity among the partitioned groups, while preserving high preference scores for both SD3.5-Large and SD3.5-Turbo.

\clearpage
\end{appendices}

\end{document}